%% file: apal_iclr.tex
\documentclass[]{article}
\usepackage{iclr2022_conference,times}
\pdfoutput=1 %tells arxiv to use pdflatex must be in the first 5 lines
\usepackage{placeins}
\usepackage{hyperref}
\usepackage{url}
\usepackage{graphicx}
\usepackage{amsmath}
\usepackage{amssymb}
\usepackage{amsthm}
\usepackage{multicol}
\usepackage{float}
\usepackage{algorithm,algorithmic} % @max added for algorithm support
\usepackage{algorithmic,eqparbox,array}% @max for aligned algorithm comments
\usepackage{tikz}
\usetikzlibrary{patterns,shapes.arrows}
\usetikzlibrary{shapes.symbols}
\usetikzlibrary{math}
\usepackage{wrapfig}
\usepackage{nicefrac}
\input{labpal/math_commands.tex}

\usepackage{layouts}
\usepackage{xargs} % new commands with more default values
\usepackage{pgfplots}

\pgfplotsset{compat=1.3}
\usepgfplotslibrary{external} % compile pgf and tikz plots in external instances
\tikzsetexternalprefix{labpal/tikz_figures/}
{apal_iclr}

\title{Using a one dimensional parabolic model of the full-batch loss to estimate learning rates during training}

\author{Maximus Mutschler  \&  Kevin Laube \& Andreas Zell \\
	University of T\"ubingen\\
	Sand 1, D-72076 T\"ubingen, Germany\\
	\texttt{\{maximus.mutschler, kevin.laube, andreas.zell\}@uni-tuebingen.de} \\
}
\iclrfinalcopy

\begin{document}
\input{commands.tex}
\maketitle

\begin{abstract}
	\vspace{-0.3cm}
   A fundamental challenge in Deep Learning is to find optimal step sizes for stochastic gradient descent automatically. In traditional optimization, line searches are a commonly used method to determine step sizes. One problem in Deep Learning is that finding appropriate step sizes on the full-batch loss is unfeasibly expensive. Therefore, classical line search approaches, designed for losses without inherent noise, are usually not applicable. Recent empirical findings suggest, inter alia, that the full-batch loss behaves locally parabolically in the direction of noisy update step directions. Furthermore, the trend of the optimal update step size changes slowly. By exploiting these and more findings, this work introduces a line-search method that approximates the full-batch loss with a parabola estimated over several mini-batches. Learning rates are derived from such parabolas during training. In the experiments conducted, our approach is on par with SGD with Momentum tuned with a piece-wise constant learning rate schedule and often outperforms other line search approaches for Deep Learning across models, datasets, and batch sizes on validation and test accuracy. In addition, our approach is the first line search approach for Deep Learning that samples a larger batch size over multiple inferences to still work in low-batch scenarios.
   %recent empirical findings suggest that the full-batch loss along noisy update step directions behaves almost parabolically and that the trend of the optimal update step size is changing slowly. Exploiting these observations, this work introduces a line-search,
\end{abstract}
%\vspace{-0.5cm}
\section{Introduction}
\vspace{-0.3cm}
Automatic determination of an appropriate and loss function-dependent learning rate schedule to train models with stochastic gradient descent (SGD) or similar optimizers is still not solved satisfactorily for Deep Learning tasks. The long-term goal is to design optimizers that work out-of-the-box for a wide range of Deep Learning problems without requiring hyper-parameter tuning. Therefore, although well-working hand-designed schedules such as \textit{piece-wise constant learning rate} or \textit{cosine decay} exist (see \citep{sgdwithwarmrestarts,cyclicallearningrates}), it is desired to infer problem depended and better learning rate schedules automatically.

This work builds on recent empirical findings; among those are that the full-batch loss tends to have a simple parabolic shape in SGD update step direction \citep{line_analysis,pal} (see  Figure \ref{labpal_fig_line_plots}) and that the trend of the optimal update step changes slowly \citep{line_analysis} (see Figure \ref{labpal_fig_strategy_metrics}). Exploiting these and more found observations, we introduce a line search approach, approximating the full-batch loss along lines in SGD update step direction with parabolas. One parabola is sampled over several batches to obtain a more exact approximation of the full-batch loss. The learning rate is then derived from this parabola. As the trend of the locally optimal update step-size on the full-batch loss changes slowly, the line search only needs to be performed occasionally; usually, every 1000th step. 
%This results in a theoretical computational overhead of less than 2\% in our experiments.

The major contribution of this work is the combination of recent empirical findings to derive a line search method, which is built upon real-world observations and less on theoretical assumptions. This method outperforms the most prominent line search approaches introduced for Deep Learning (\cite{sls,pal,gradientOnlyLineSearch,probabilisticLineSearch}) across models, datasets usually considered in optimization for Deep Learning, in almost all experiments. In addition, it is on par with SGD with momentum tuned with a piece-wise constant learning rate schedule. The second important contribution is that we are the first to analyze how the considered line searches perform under high gradient noise that originates from small batch sizes. While all considered line searches perform poorly -mostly because they rely on mini-batch losses only-, our approach adapts well to increasing gradient noise by iteratively sampling larger batch sizes over several inferences. 

The paper is organized as follows: Section \ref{labpal_sec_related_work} provides an overview of related work. Section \ref{labpal_sec_labpal_introduction} derives our line search approach and introduces its mathematical and empirical foundations in detail. In Section \ref{labpal_sec_empirical_analysis} we analyze the performance of our approach across datasets, models, and gradient noise levels. Also, a comprehensive hyper-parameter, runtime, and memory consumption analysis is performed. Finally, we end with discussion including limitations in Sections \ref{labpal_sec_limitations} \& \ref{labpal_sec_discussion}.

\section{Related Work}
\label{labpal_sec_related_work}
%\begin{tabular}{ c  }
%	\input{labpal/"pure_line_8000.pgf"}
%\end{tabular}

\input{labpal/figure2_line_plots.tex}
\textbf{Deterministic line searches:}
According to \cite[\textsection 3]{numerical_optimization}, line searches are considered a solved problem, in the noise-free case. However, such methods are not robust to gradient and loss noise and often fail in this scenario since they shrink the search space inadequately or use too inexact approximations of the loss. \cite[\textsection 3.5]{numerical_optimization} introduces a deterministic line search using parabolic and cubic approximations of the loss, which motivated our approach.\\
\textbf{Line searches on mini-batch and full-batch losses and why to favor the latter.}
The following motivates the goal of our work to introduce a simple, reasonably fast, and well-performing line-search approach that approximates full-batch loss.\\
Many exact and inexact line search approaches for Deep Learning are applied on mini-batch losses \citep{pal,L4_alternative,L4,hypergradientdescent,sls}.
\citep{pal} approximates an exact line search by estimating the minimum of the mini-batch loss along lines with a one-dimensional parabolic approximation. The other approaches perform inexact line searches by estimating positions of the mini-batch loss along lines, which fulfill specific conditions. Such, inter alia, are the Goldberg, Armijo, and Wolfe conditions  (see \cite{numerical_optimization}). For these, convergence on convex stochastic functions can be assured under the interpolation condition \citep{sls}, which holds if the gradient with respect to each batch converges to zero at the optimum of the convex function. Under this condition, the convergence rates match those of gradient descent on the full-batch loss for convex functions \citep{sls}. However, relying on those assumptions and on mini-batch losses only does not lead to robust optimization, especially not if the gradient noise is high, as will be shown in Section \ref{labpal_sec_empirical_analysis}. \citep{line_analysis,pal} even showed that exact line searches on mini-batch losses are not working at all.
Line searches on the non-stochastic full-batch loss show linear convergence on any deterministic function that is twice continuously differentiable, has a relative minimum, and only positive eigenvalues of the Hessian at the minimum (see \cite{nonlinear_programming}). In addition, they are independent of gradient noise.
Therefore, it is reasonable to consider line searches on the full-batch loss.
However, these are cost-intensive since a massive amount of mini-batch losses for multiple positions along a line must be determined to measure the full-batch loss.  \\Probabilistic Line Search (PLS) \citep{probabilisticLineSearch} addresses this problem by performing Gaussian Process Regressions, which result in multiple one-dimensional cubic splines. In addition, a probabilistic belief over the first  (aka Armijo condition) and second Wolfe condition is introduced to find appropriate update positions. The major drawback of this conceptually appealing method is its high complexity and slow training speed.
A different approach working on the full-batch loss is Gradient-only line search (GOLSI) \citep{gradientOnlyLineSearch}. It approximates a line search on the full-batch loss by considering consecutive noisy directional derivatives whose noise is considerably smaller than the noise of the mini-batch losses.\\
\textbf{Empirical properties of the loss landscape:}
In Deep Learning, loss landscapes of the true loss (over the whole distribution), the full-batch loss, and the mini-batch loss can, in general, be highly non-convex. However, to efficiently perform a line search, some properties of these losses have to be apparent. Little is known about such properties from a theoretical perspective; however, several works suggest that loss landscapes tend to be simple and have some properties:
 \cite{line_analysis,visualisationLossLandscape,walkwithsgd,pal,empericalLineSearchApproximations,probabilisticLineSearch,LinePlots,wedge_model,elasticband}. 
 \citep{probabilisticLineSearch,walkwithsgd,line_analysis,pal} suggest that the full-batch loss $\mathcal{L}$ along lines in negative gradient directions tend to exhibit a simple shape for a set of Deep Learning problems. This set includes at least classification tasks on CIFAR-10, CIFAR-100, and ImageNet. 
 \citep{line_analysis} sampled the full-batch loss along the lines in SGD update step directions. This was done for $10,000$ consecutive SGD and SGD with momentum update steps of a ResNet18's, ResNet20's and MobileNetv2's training process on a subset of CIFAR-10. Representative plots of their $10,000$ measured full-batch losses along lines are presented in Figure \ref{labpal_fig_line_plots}. Relevant insights and found properties of these works will be introduced and exploited to derive our algorithm in Section \ref{labpal_sec_labpal_introduction}. \\
 \textbf{Using the batch size to tackle gradient noise:}  % todo maybe \paragraph
 Besides decreasing the learning rate, increasing the batch size remains an important choice to tackle gradient noise.
 \citeauthor{ModelOfLargeBatchTraining} exploits empirical information to predict the largest piratical batch size over datasets and models. \citeauthor{bigBatchSGD} adaptively increases the batch size over update steps to assure that the negative gradient is a descent direction. 
 \citeauthor{bayesianPerspectiveSGD} introduces the \textit{noise scale}, which controls the magnitude of the random fluctuations of consecutive gradients interpreted as a differential equation. The latter leads to the observation that increasing the batch size has a similar effect as decreasing the learning rate \citep{increase_batch_size}, which is exploited by our algorithm.

\vspace{-0.3cm}
\section{Our approach: Large-Batch Parabolic approximation line search (LabPal)}
\label{labpal_sec_labpal_introduction}
\subsection{Mathematical Foundations}
\label{labpal_subsec_math_foundations}
In this subsection, we introduce the mathematical background relevant for line searches and challenges that must be solved in order to perform line searches in Deep Learning. \\\
We consider the problem of minimizing the full-batch loss $\L$,  which is the mean over a large amount of sample losses $L_d$:
\begin{equation}\label{labpal_eq_fb_loss}
	\L \;:\; \mathbb{R}^n \rightarrow  \mathbb{R}, \; \mathbf{\theta}  \mapsto \frac{1}{|\mathbb{D}|}\sum_{d \in \mathbb{D}}L_d(\mathbf{\theta}),\end{equation}
where $\mathbb{D}$ is a finite dataset and $\mathbf{\theta}$ are $n$ parameters to optimize. To increase training speed generally, a mini-batch loss $\mathcal{L}_{\mathbb{B}}$, which is a noisy estimate of $\L$, is considered:
\begin{equation}\label{labpal_eq_mb_loss}
	\BL[] \;:\; \mathbb{R}^n \rightarrow  \mathbb{R}, \; \mathbf{\theta}  \mapsto \frac{1}{|\mathbb{B}|}\sum_{d \in \mathbb{B} \subset \mathbb{D}}L_d(\mathbf{\theta}),\end{equation}
with $|\mathbb{B}|\ll|\mathbb{D}|$. We define the mini-batch gradient at step $t$ as $\mathbf{g}_{\mathbb{B},t} \in \mathbb{R}^n$ as $\nabla_{\mathbf{\theta}_t}\BL[](\mathbf{\theta}_t)$.

For our approach, we need the full-batch loss along the direction of the negative normalized gradient of a specific mini-batch loss.  At optimization step $t$ with current parameters $\mathbf{\theta}_t$ and a direction defining batch $\mathbb{B}_t$, $\BL[]$ along a line with origin $\mathbf{\theta}_t$ in the negative direction of the normalized batch gradient $\hat{\mathbf{g}}_{\mathbb{B},t}=\nicefrac{\mathbf{g}_{\mathbb{B},t}}{||\mathbf{g}_{\mathbb{B},t}||}$ is given as:
\begin{equation}\label{labpal_eq_batch_line_search}
 \Bl\;:\; \mathbb{R} \rightarrow  \mathbb{R}, \; s  \mapsto \BL[](\mathbf{\theta}_t+s\cdot-\hat{\mathbf{g}}_{\mathbb{B}_t,t}),
\end{equation}
where $s$ is the step size along the line.
The corresponding full-batch loss along the same line is given by: \begin{equation}\label{labpal_eq_full_batch_line_search}	l_t :\; \mathbb{R} \rightarrow  \mathbb{R}, \; s \mapsto \L (\mathbf{\theta}_t+s\cdot-\hat{\mathbf{g}}_{\mathbb{B}_t,t}).
\end{equation}Let the step size to the first encountered minimum of $l_t$ be $s_{\text{min},t}$. 

Two major challenges have to be solved in order to perform line searches on $\L$:
\begin{enumerate}
	\item To measure $l_t$ exactly  it is required to determine every $L_d(\mathbf{\theta}_t+s\cdot - \hat{\mathbf{g}}_{\mathbb{B}_t,t})$ for all $d \in \mathbb{D}$ and for all step sizes $s$ on a line.
	\item To assure convergence line searches have to be performed in a descent direction \citep{bigBatchSGD}. The simplest form is the direction of steepest descent \citep{nonlinear_programming}. Therefore, the full-batch gradient $\nabla_{\mathbf{\theta}}\L \;:\; \mathbb{R}^n \rightarrow  \mathbb{R}^n, \; \mathbf{\theta} \mapsto \frac{1}{|\mathbb{D}|}\sum\limits_{d \in \mathbb{D}}\nabla L_d(\mathbf{\theta}_t)$ has to be approximated.
\end{enumerate}
 To be efficient, $l_t$ has to be approximated sufficiently well with as little data points $d$ and steps $s$ as possible, and one has to use as little $d$ as possible to approximate $\nabla_{\mathbf{\theta}}\L$ approximated sufficiently well.
 Such approximations are highly dependent on properties of  $\L$. Due to the complex structure of Deep Neural Networks, little is known about such properties from a theoretical perspective. Thus, we fall back to empirical properties.

\vspace{-0.3cm}
\subsection{Deriving the algorithm}\label{labpal_sec_deriving_the_alg}
\vspace{-0.1cm}
\input{labpal/figure_strategy_metrics} 
\algsetup{linenosize=\scriptsize}
\begin{algorithm}[b!]
	\vspace{-0.1cm}
	\caption{LABPAL\&SGD. Simplified conceptional pseudo-code of our proposed algorithm, which estimates update steps on a parabolic approximation of the full-batch loss. See the published source code for technical details. Default values are given in parenthesis. For LABPAL\&NSGD  SGD is replaced with NSGD, and the update step is measured instead of the learning rate.}  %define names here}
	\begin{multicols}{2}
		\begin{algorithmic}[1]
			\scriptsize
			\label{labpal_alg:MBPAL}
			\renewcommand{\algorithmicrequire}{\textbf{Input:}}
			\newcommand\algorithmicprocedure{\textbf{procedure}}
			\newcommand{\algorithmicendprocedure}{\algorithmicend\ \algorithmicprocedure}
			\makeatletter
			\newcommand\PROCEDURE[3][default]{%
				\ALC@it
				\algorithmicprocedure\ \textsc{#2}(#3)%
				\ALC@com{#1}%
				\begin{ALC@prc}%
				}
				\newcommand\ENDPROCEDURE{%
				\end{ALC@prc}%
				\ifthenelse{\boolean{ALC@noend}}{}{%
					\ALC@it\algorithmicendprocedure
				}%
			}
			\newenvironment{ALC@prc}{\begin{ALC@g}}{\end{ALC@g}}
			\makeatother
			\REQUIRE Hyperparameters: \\
			- initial parameters $\theta_0$\\
			- approximation batch size $|\mathbb{B}_a|$ $(1280)$\\ 
			- inference batch size $|\mathbb{B}_i|$ $(128)$\\ 
			- SGD steps $n_{\text{SGD}}$ $(1000)$, \quad \# or NSGD steps\\
			- step size adaptation $\alpha > 1$ $(1.8)$\\
			- training steps $t_{max} (150000)$ \\
			- batch size schedule $k(t)= \begin{cases}
				1, & \text{if $t\leq \lfloor t_{max} \cdot 0.5\rfloor$}\\
				2, & \text{elif $t\leq \lfloor t_{max} \cdot 0.75\rfloor$}\\
				4, & \text{elif $t > \lfloor t_{max} \cdot 0.75\rfloor$}
			\end{cases}$ 
			\STATE \# Variables have global scope
			%last update steps to consider \textit{u} (10)
			\STATE \textit{sampledBatchSize} $\leftarrow0$
			\STATE \textit{performedSGDsteps} $\leftarrow$ 0
			\STATE \textit{learningRate} $\leftarrow$ 0
			\STATE  $\theta \leftarrow \theta_0$
			\STATE \textit{state} $\leftarrow$ 'line search'
			\STATE \textit{direction} $\leftarrow$ current batch gradient
			\STATE  $t \leftarrow 0$
			\WHILE{$t < t_{max}$}
			\IF{\textit{state} is 'line search'}
			\STATE \textsc{perform\_line\_search\_step()} 
			\ENDIF
			\IF{state is 'SGDTraining'}
			\STATE \textsc{perfrom\_large\_batch\_sgd\_step()}
			\ENDIF
			\ENDWHILE
			\RETURN $\theta$
			
			\PROCEDURE{perform\_line\_search\_step}{}
			\IF{\textit{sampledBatchSize} $< |\mathbb{B}_a|$ }
			\STATE update estimate $\hat{\mathcal{L}}$ of $\mathcal{L}$ with\\ over multiple inferences sampled  $\mathcal{L}_{\mathbb{B}_t,t}$  with \\ $|\mathbb{B}_t| = k(t)\cdot |\mathbb{B}_i|)$
			\STATE increase \textit{sampledBatchSize} by $|\mathbb{B}_t|$ and $t$ by $k(t)$
			\ELSE 
			\STATE \textit{learningRate} $\leftarrow$ perform parabolic approximation with 3 values of $\hat{\mathcal{L}}$ along the search direction and estimate the learning rate. 
			%\STATE updateStep $\leftarrow$ Average over \textit{u} last measured update steps excluding the first measurements for which the shape of the lines varies a lot.
			\STATE \textit{learningRate} $\leftarrow$ \textit{learningRate} $\cdot\alpha$
			\STATE set \textit{sampledBatchSize} and \textit{performedSGDsteps}  to $0$
			\STATE \textit{state} $\leftarrow$ 'SGDTraining'
			\ENDIF
			\ENDPROCEDURE
			\STATE
			\PROCEDURE{perfrom\_large\_batch\_sgd\_step}{}
			\IF{performedSGDsteps $<n_{\text{SGD}}$}
			\STATE $\theta$ $\leftarrow$ perform SGD update with \textit{learningRate} and over multiple inferences sampled $\mathcal{L}_{\mathbb{B}_t,t}$ \\ with $|\mathbb{B}_t| = k(t)\cdot |\mathbb{B}_i|)$
			\STATE increase $t$ by $k(t)$
			\STATE increase \textit{performedSGDsteps} by $1$
			\ELSE 
			\STATE \textit{direction} $\leftarrow$ current batch gradient
			\STATE \textit{state} $\leftarrow$ 'line search'
			\ENDIF
			\ENDPROCEDURE
		\end{algorithmic} 
		\normalsize
	\end{multicols}
\end{algorithm}
%\algsetup{linenosize=\scriptsize}
In the following, we derive our line search approach on the full-batch loss by iteratively exploiting empirically found observations of \citep{line_analysis} and solving the challenges 
for a line search on the full-batch loss (see Section \ref{labpal_subsec_math_foundations}).
Given default values are inferred from a detailed hyper parameter analysis (Section \ref{labpal_app_hyperparameter_analysis})\\
\textbf{Observation 1:} \textit{Minima of  $\Bl$  can be at significantly different points than minima of $l_t$ and can even lead to update steps, which increase $\L$ }(Figure \ref{labpal_fig_strategy_metrics} center, green and red curve). \\
\textbf{Derivation Step 1:} This consolidates that line searches on a too low mini-batch loss are unpromising. Consequently, we concentrate on a better way to approximate $l_t$.\\
\textbf{Observation 2:} \textit{$l_t$ can be approximated with parabolas of positive curvature, whose fitting errors are of less than $0.6\cdot 10^{-2}$ mean absolute distance} (exemplarily shown in Figure \ref{labpal_fig_line_plots}).\\
\textbf{Derivation Step 2:} We approximate $l_t$ with a parabola ($l(s)_t \approx a_ts^2+b_ts+c_t$ with $a_t>0$). A parabolic approximation needs three measurements of $l_t$.  However, already computing  $l_t$ for one $s$ only is computationally unfeasible. Assuming i.i.d sample losses, the standard error of $\Bl(s)$, decreases with $\nicefrac{1}{\sqrt{|\mathbb{B}|}}$. Thus,  $\Bl$ -with a reasonable large batch size- is already a good estimator for the full-batch loss parabola. Consequently, we approximate $l_t$  with $l_{\mathbb{B}_a,t}$  by averaging over multiple $l_{\mathbb{B}_i,t}$ measured with multiple inferences. Thus, the approximation batch size $\mathbb{B}_a$, is significantly larger as the, by GPU memory limited, possible batch size $\mathbb{B}_i$. In our experiments, $\mathbb{B}_a$ is usually chosen to be $1280$, which is $10$ times larger as $\mathbb{B}_i$. In detail, we measure $l_{\mathbb{B}_a,t}$ at the points $s=0,0.0001$ and $0.01$, then we simply infer the parabola's parameters and the update step to the minimum. These values of $s$ empirically lead to the best and numerically most stable approximations.\\
\textbf{Observation 3:} \textit{The trend of $s_{\text{min},t}$ of consecutive $l_t$ changes slowly and consecutive $l_t$ do not change locally significantly.} (Figure \ref{labpal_fig_strategy_metrics} left, red curve).\\ 
\textbf{Observation 4:} \textit{$s_{\text{min},t}$ and the direction defining batch's $||\mathbf{g}_{\mathbb{B}_t,t}||$ are almost proportional during training.} (Figure \ref{labpal_fig_strategy_metrics} right).\\
\textbf{Derivation Step 3:} Using measurements of $l_{\mathbb{B}_a,t}$ to approximate $l_t$ with a parabola is by far to slow to compete against SGD if done for each weight update. By exploiting Observation 3  we can approximate $l_t$ after a constant amount of steps and reuse the measured learning rate $\lambda$ or update step size $\s$ for subsequent steps. In this case, $\lambda$ is a factor multiplied by $\mathbf{g}_{\mathbb{B},t}$, whereas $\s$ is a factor multiplied by $\hat{\mathbf{g}}_{\mathbb{B},t}$. Consequently with $\lambda$ we perform a step in gradient direction (as SGD also does), whereas, with $\s$ we perform a step in normalized gradient direction, ignoring the norm of the gradient.  Observation 4 allows us to reuse $\lambda$.
In our experiments, it is sufficient to measure a new $\lambda$ or $\s$ every 1000 steps only.\\
\textbf{Derivation Step 4:} So far, we can approximate $l_t$ efficiently and, thus, overcome the first challenge (see Section \ref{labpal_subsec_math_foundations}). Now, we will overcome the second challenge; approximating the full batch loss gradient for each weight update step:\\
For this, we revisit \cite{increase_batch_size} who approximates the magnitude of random gradient fluctuations, that appear if training with a mini-batch gradient, by the \textit{noise scale} $\nu \in \mathbb{R}$:
\begin{equation}\label{labpal_eq_noise_scale}
	 \nu \approx \nicefrac{(\lambda|\mathbb{D}|)}{|\mathbb{B}|},
\end{equation}
where $\lambda$ is the learning rate, $|\mathbb{D}|$ the dataset size and $|\mathbb{B}|$ the batch size. If the random gradient fluctuations are reduced, the approximation of the gradient gets better. Since we want to estimate the learning rate automatically, the only tunable parameter to reduce the \textit{noise scale} is the batch size. \\
\textbf{Observation 5:}  \textit{The variance of consecutive $s_{\text{min},t}$  is low, however, it increases continuously during training} (Figure \ref{labpal_fig_strategy_metrics} left, red curve).  \\
\textbf{Derivation Step 5:} It stands to reason that the latter happens because the random gradient fluctuations increase.
Consequently, during training, we increase the batch size for weight updates by iteratively sampling a larger batch with multiple inferences. This reduces the variance of consecutive $s_{\text{min},t}$ and lets us reuse estimated the $\lambda$ or $\s$ for more steps. After experiencing unusable results with the approach of  \citep{bigBatchSGD} to determine appropriate batch sizes, we stick to a simple piece-wise constant batch size schedule doubling the batch size after two and after three-quarters of the training.\\
\textbf{Observation 6:} \textit{On a global perspective a $\s$ that overestimates $s_{\text{min},t}$ optimizes and generalizes better.}\\
\textbf{Derivation Step 6:} Thus, after estimating $\lambda$ (or $s_{\text{min}}$) we multiply it with a factor $\alpha \in ]1,2[$: \begin{equation}\label{labpal_eq_learning_rate}
	\lambda_{\text{new}} = \alpha \lambda =  \nicefrac{\alpha s_{\text{min},t}}{||\mathbf{g}_{\mathbb{B},t}||} \quad \text{or} \quad  \s=\alpha s_{\text{min},t}
\end{equation}
Note that under out parabolic property, the first wolfe condition $w_1$, which is commonly used for line searches, simply relates to $\alpha$:  $w_1=-\frac{\alpha}{2}+1$ (see Appendix \ref{labpal_app_wolfe}).

Combining all derivations leads to our line search named \textit{large-batch parabolic approximation line search} (LABPAL), which is given in  Algorithm \ref{labpal_alg:MBPAL}. It samples the desired batch size over multiple inferences to perform a close approximation of the full-batch loss and then reuses the estimated learning rate to train with SGD (LABPAL\&SGD), or it reuses the update step to train with SGD with a normalized gradient (LABPAL\&NSGD). While LABPAL\&SGD elaborates Observation 4, LABPAL\&NSGD completely ignores information from $||\mathbf{g}||$. 

\vspace{-0.3cm}
\section{Empirical Analysis}
\vspace{-0.1cm}
\label{labpal_sec_empirical_analysis}
Our two approaches are compared against other line search methods across several datasets and networks in the following.
\textbf{To reasonably compare different line search methods, we define a \textit{step} as the sampling of a new input batch.} Consequently, the steps/batches that LABPAL takes to estimate a new learning rate/step size are considered, and optimization processes are compared on their data efficiency.\\
Note that the base ideas of the introduced line search approaches can be applied upon any direction giving technique such as Momentum, Adagrad \citep{adagrad} or Adam \citep{adam}. Results are averaged over 3 runs.

\subsection{Performance analysis on ground truth full-batch loss and proof of concept}
\input{labpal/figure_training_processes_palsgd} 
% %
To analyze how well our approach approximates the full-batch loss along lines, we extended the experiments of \cite{line_analysis} by LABPAL. \cite{line_analysis} measured the full-batch loss along lines in SGD update step directions of a training process; thus, this data provides ground truth to test how well the approach approximates the full-batch loss.  In this scenario, LABPAL\&SGD uses the full-batch size to estimate the learning rate and reuses it for 100 steps. No update step adaptation is applied.
Figure \ref{labpal_fig_strategy_metrics}
shows that LABPAL\&SGD fits the update step sizes to the minimum of the full-batch loss and performs near-optimal local improvement. The same holds for LABPAL\&NSGD. \\
We now test how our approaches perform in a scenario for which we can assure that the used empirical observations hold. Therefore, we consider the optimization problem of \citep{line_analysis} from which all empirical observations were inferred, which is training a ResNet20 on 8\% of CIFAR10. $\mathbb{B}_a$ of 1280 is used for both approaches. Learning rates are reused for 100 steps, and $\alpha=1.8$ is considered. The batch size is doubled after $5000$ and $7500$ steps. For SGD  $\lambda$ is halved after the same steps. A grid search for the best $\lambda$ is performed. Figure \ref{labpal_fig_training_process_resnet-20_palsgd} shows that LABPAL\&NSGD with update step adaptation outperforms SGD, even though $9\%$ of the training steps are used to estimate new update step sizes. This shows that using the estimated learning rates and step sizes leads to better performance than keeping them constant or decaying them with a piece-wise schedule. Interestingly huge $\lambda$s of up to $80,000$ are estimated, whereas $\s$s are decreasing. LABPAL\&SGD shows similar performance as SGD; however, it seems beneficial to ignore gradient size information as the better performance of LABPAL\&NSGD shows.
\vspace{-0.3cm}
\subsection{Performance comparison to SGD and to other line search approaches}
\vspace{-0.1cm}
\label{labpal_sec_opt_comparison}
We compare the SGD and NSGD variants of our approach against PLS \citep{probabilisticLineSearch}, GOLSI \citep{gradientOnlyLineSearch}, PAL \citep{pal}, SLS \citep{sls} and SGD with Momentum \citep{grad_descent}. The latter is a commonly used optimizer for Deep Learning problems and can be reinterpreted as a parabolic approximation line search on mini-batch losses \citep{line_analysis}. 
PLS is of interest since it approximates the full-batch loss to perform line searches. PAL, GOLSI, SLS on the other hand are line searches optimizing on mini-batch losses directly. For SGD with Momentum, a piece-wise constant learning rate schedule divides the learning rate after two and again after three-quarters of the training by a factor of 10.

Comparison is done across several datasets and models. Specifically, we compare ResNet-20 \citep{resnet}, DenseNet-40 \citep{denseNet}, and MobileNetV2 \citep{mobilenet} trained on CIFAR-10 \citep{CIFAR-10}, CIFAR-100 \citep{CIFAR-100}, and SVHN \citep{SVHN}. %In addition, we compare MobileNetV3 \citep{mobilenetv3} trained on ImageNet \citep{IMAGENET}.
 We concentrate on classification problems since the empirical observations are inferred from a classification task and since those problems are usually considered to benchmark new optimization approaches.  %as well as a 3 Layer Fully-Connected network and a 3 layer convolutional network on  MNIST \citep{MNIST}, FashionMNIST \citep{Fashion-MNIST} and NETWORK TODO ImageNet \citep{IMAGENET}.

For each optimizer, we perform a comprehensive hyper-parameter grid search on ResNet-20 trained on CIFAR-10 (see Appendix \ref{labpal_app_hyperparams}). The best performing hyper-parameters on the validation set are then reused for all other experiments. The latter is done to check the robustness of the optimizer by handling all other datasets as if they were unknown, as is usually the case in practice. Our aim here is to show that satisfactory results can be achieved on new problems without any fine-tuning needed. Further experimental details are found in Appendix \ref{labpal_app_hyperparam_emp_loss_comp}.

Figure \ref{labpal_fig_optimizer_comparison_cifar100} as well as Appendix Figures \ref{labpal_fig_optimizer_comparison_cifar10}, \ref{labpal_fig_optimizer_comparison_svhn} show that both LABPAL approaches outperform PLS, GOLSI and PAL considering training loss, validation accuracy, and test accuracy.  LABPAL\&NSGD tends to perform more robust and better than LABPAL\&SGD. LABPAL\&NSGD is on pair with SGD with Momentum and challenges SLS on validation and test accuracy. The important result is that hyper-parameter tuning for LABPAL is not needed to achieve good results across several models and datasets. However, this also is true for pure SGD, which suggests that the simple rule of performing a step size proportional to the norm of the momentum term is sufficient to implement a well-performing line search. This also strengthens the observation of \citep{line_analysis}, which states that SGD, with the correct learning rate, is already performing an almost optimal line search.

The derived learning rate schedules of the LABPAL approaches are significantly different from a piece-wise constant schedule (Figure \ref{labpal_fig_optimizer_comparison_cifar100}, \ref{labpal_fig_optimizer_comparison_cifar10}, \ref{labpal_fig_optimizer_comparison_svhn} first row). Interestingly they show a strong \textit{warm up (increasing) phase} at the beginning of the training followed by a rather \textit{constant phase} which can show minor learning rate changes with an increasing trend. The NSGD variant sometimes shows a second increasing phase, when the batch size is changed. The \textit{warm up} phase is often seen in sophisticated learning rate schedules for SGD; however, usually combined with a \textit{cool down} phase. The latter is not apparent for LABPAL since we increase the batch size. LABPAL\&NSGD indirectly uses learning rates of up to $10^6$ but still trains robustly. Of further interest is that all line search approaches do not decrease the learning rate at the end of the training as significantly as SGD, which hinders the line searches to converge.

A comparison of training speed and memory consumption is given in Appendix \ref{labpal_app_wall_clock}. In short, LABPAL  has identical GPU memory consumption as SGD and is on average only $19.6\%$ slower. However, for SGD usually a grid search is needed to find a good $\lambda$, which makes LABPAL considerably cheap.
\input{labpal/performance_comparison_cifar100}

\FloatBarrier

\vspace{-0.3cm}
\subsection{Adaptation to varying gradient noise}
\vspace{-0.1cm}
\label{labpal_sec_low_batch_size}
Recent literature, E.g., \citep{pal}, \citep{sls}, \citep{gradientOnlyLineSearch} show that line searches work with a relatively large batch size of 128 and a training set size of approximately 40000 on CIFAR-10.  \textit{However, a major, yet not comprehensively considered problem is that line searches operating on the mini-batch loss vary their behavior with another batch- and training set sizes leading to varying gradient noise}.
E.g., Figure \ref{labpal_fig_performance_comp_batch_size} shows that training with PAL, GOLSI or PlS and a batch size of 8 on CIFAR-10 does not work at all. The reason is that the by mini-batches induced gradient noise, and with it the difference between the full-batch loss and the mini-batch loss, increases.
However, we can adapt LABPAL to work in these scenarios by holding the \textit{noise scale} it is exposed to approximately constant. As the learning rate is inferred directly, the batch size has to be adapted.  
Based on the linear approximation of the \textit{noise scale} (see Equation \ref{labpal_eq_noise_scale}), we directly estimate a noise adaptation factor $\epsilon \in \mathbb{R}$ to adapt LABPAL's hyperparameter:
\begin{equation}
	\label{labpal_eq_noise_factor1}
	\epsilon := \frac{\nu_{new}}{\nu_{ori}}= \frac{|\mathbb{B}_{ori}|}{|\mathbb{B}_{new}|} \frac{|\mathbb{D}_{new}|}{|\mathbb{D}_{ori}|} = \frac{128}{|\mathbb{B}_{new}|}\frac{|\mathbb{D}_{new}|}{40,000}
\end{equation}
The original batch size $|\mathbb{B}_{ori}|$ and the original dataset size $|\mathbb{D}_{ori}|$  originate from our search for best-performing hyperparameters on CIFAR-10 with a training set size of 40,000, a batch size of 128, and 150,000 training steps.
We set the number of training steps to $150,000\epsilon$ and multiply the batch sizes in the batch size schedule $k$ by $\epsilon$. This rule makes the approach fully parameter less in practice (at least for the image classification scenario), since hyperparameters do not have to be adapted across models, batch sizes and datasets.
Figure \ref{labpal_fig_performance_comp_batch_size} shows a performance comparison over different batch sizes. 
Hyperparameters are not changed. By changing the batch size the noise adaptation factor of the LABPAL approaches gets adapted, which lets them still perform well with low batch sizes since they iteratively sample larger batch sizes over multiple inferences. The performance of PLS, PAL and GOLSI decreases with lower batch size. SLS's performance stays similar but its learning rate schedule degenerates.
For the evaluation on ResNet-20 and MobileNet-V2 see Appendix Figure \ref{labpal_fig_opt_comparison_bs32} \& \ref{labpal_fig_opt_comparison_bs8}. 
We note that this batch size adaptation approach to keep the noise scale on a similar level could also be applied to all other line searches, however this will exceed the limits of this work.
\input{labpal/fig_batch_size_comparison}
\vspace{-0.3cm}
\subsection{Hyperparameter Sensitivity Analysis}
\label{labpal_sec_hyperparameter_analysis}
We performed a detailed hyperparameter sensitivity analysis for LABPAL\&SGD and LABPAL\&NSGD. To keep the calculation cost feasible, we investigated the influence of each hyperparameter, keeping all other hyperparameters fixed to the default values (see Algorithm \ref{labpal_alg:MBPAL}).
Appendix Figure \ref{labpal_fig_sensitivity_analysis_lappalsgd} and \ref{labpal_fig_sensitivity_analysis_lappalnsgd} show the following characteristics:
Estimating new $\s$ or $\lambda$ with $\mathbb{B}_a$ smaller than $640$ decreases the performance since $l_t$ is not fitted well enough (row 1). The performance also decreases if reusing the $\lambda$ (or $\s$) for more update steps (row 2), and if using a step size adaptation $\alpha$ of less than 1.8 (row 3, except for ResNet). This shows that optimizing for the locally optimal minimum in line direction is not beneficial. From a global perspective, a slight decrease of the loss by performing steps to the other side of the parabola shows more promise. Interestingly, even using $\alpha$ larger than two still leads to good results. \citep{line_analysis} showed that the loss valley in line direction becomes wider during training. This might be a reason why these update steps, which should actually increase the loss, work. Using a maximal step size of less than 1.5 (row 7) and increasing the noise adaptation factor $\epsilon$ (row 9) while keeping the batch size constant also decreases the performance. The latter indicates that the inherent noise of SGD is essential for optimization. 
In addition, we considered a momentum factor and conclude that a value between $0.4$ and $0.6$ increases the performance for both LABPAL approaches (row 5).

\vspace{-0.3cm}
\section{Limitations}
\vspace{-0.1cm}
\label{labpal_sec_limitations}
Our approach can only work if the empirically found properties we rely on are apparent or are still a well enough approximation.  In Section \ref{labpal_sec_opt_comparison}  we showed that this is valid for classification tasks. In additional sample experiments, we observed that our approach also works on regression tasks using the square loss. However, it tends to fail if different kinds of losses from significantly different heads of a model are added, as it is often the case for object detection and object segmentation.

A theoretical analysis is lacking since the optimization field still does not know the reason for the local parabolic behavior of $l_t$ is, and consequently, what an appropriate function space to consider for convergence is.

\vspace{-0.3cm}
\section{Discussion \& Outlook }
\vspace{-0.2cm}
\label{labpal_sec_discussion}
This work introduced a robust line search approach for Deep Learning problems based upon empirically found properties of the full-batch loss. Our approach estimates learning rates well across models, datasets, and batch sizes. It mostly surpasses other line search approaches and challenges SGD with Momentum tuned with a piece-wise constant learning rate schedule. We are the first line search work that analyses and adapts to varying gradient noise. In addition, we show that mini-batch gradient norm information is not necessary for training.  In future, we will analyze the causes for the local parabolic behavior of the full-batch loss along lines, to get a better understanding of DNN loss landscapes and especially of why and when specific optimization approaches work.

\section*{Reproducibility}
Experimental details including all hyperparameters used for the experiments presented in Section \ref{labpal_sec_empirical_analysis} are found in App. \ref{labpal_app_hyperparams}.
The source code to reproduce our experiments including our implementations of SLS, GOLSI, PLS, PAL and LABPAL
is provided in the supplementary materials.

\section*{Ethics Statement}
 Since we understand our work as basic research, it is extremely error-prone to estimate its \textit{specific} ethical aspects and future positive or negative social consequences. As optimization research influences the whole field of deep learning, we refer to the following works, which discuss the ethical aspects and social consequences of AI and Deep Learning in a comprehensive and general way:\cite{yudkowsky2008artificial,muehlhauser2012singularity,bostrom2014ethics}.

\FloatBarrier
\bibliography{apal_iclr}
\bibliographystyle{iclr2022_conference}
\vfill
\pagebreak

\appendix
\FloatBarrier

\section{Hyperparamter Sensitivity Analysis}
\label{labpal_app_hyperparameter_analysis}.

\input{labpal/figure_sensitivity_analysis_labpalsgd} 

\input{labpal/figure_sensitivity_analysis_labpalnsgd} 
\vfill
\clearpage
\section{Further performance comparisons}

\input{labpal/performance_comparison_cifar10}
\input{labpal/performance_comparison_svhn}

\FloatBarrier
\vfill
\clearpage
\section{Further results for batch sizes 32 and 8}

\input{labpal/batch_size_32}
\input{labpal/batch_size_8}
\FloatBarrier
\vfill
\clearpage
\section{Wall clock time and GPU memory comparison}
\label{labpal_app_wall_clock}
%Considering the training speed in wall-clock time, Appendix Figure \ref{labpal_fig_speed_comparison} shows that the LABPAL approaches perform up to 28\% (DenseNet) and 68\% (MobileNet) slower than SGD and the other line search approaches except for PLS. The primary reason for this is that LABPAL copies the gradient at each step to support iteratively measured larger batch sizes.
%\begin{wrapfigure}{l}{0pt}

\begin{figure}[h!]
	\tikzsetfigurename{labpal_speed_comparison}
	\centering
	\begin{tikzpicture}[scale=0.4]%0.4
		\begin{axis}[
			width= 1.0\linewidth,
			height= 0.7\linewidth,
			ylabel= training time on CIFAR-10 in h.,
			title = speed comparison,
			title style={font=\LARGE},
			ybar ,
			xtick=data, 
			%bar shift=0pt, % this must be after ybar to have effect
			symbolic x coords={DenseNet-121,MobileNet-V2,ResNet-20},
			enlarge x limits  = 0.25,
			ylabel style={font=\LARGE},
			xlabel style={font=\LARGE},
			x tick label style={font=\Large},
			y tick label style={font=\Large},
			legend style={font=\LARGE,at={(0.0,1.5)},anchor=north west},
			xtick=data,
			xtick pos=left,
			ytick pos=left,
			ymin=0.0,%ymax=0.87,
			legend columns=2,
			]
			\addplot[fill=blue] %LABPAL-SGD  TODO
			coordinates {(DenseNet-121,4.3) (MobileNet-V2,4.3) (ResNet-20,1.52)};
			\addplot[fill=red]   %LABPAL-NSGD TODO
			coordinates {(DenseNet-121,4.5) (MobileNet-V2,4.5) (ResNet-20,1.66)};
			\addplot[fill=green]  %SGD
			coordinates {(DenseNet-121,3.75) (MobileNet-V2,3.66) (ResNet-20,1.3)};
			\addplot[fill=black]  %SLS
			coordinates {(DenseNet-121,2.81) (MobileNet-V2,2.85) (ResNet-20,1.02)};
			\addplot[fill=orange]  %PAL
			coordinates {(DenseNet-121,2.9) (MobileNet-V2,3.08) (ResNet-20,1.01)};
			\addplot[fill=violet]  %GOLSI
			coordinates {(DenseNet-121,4.05) (MobileNet-V2,4.07) (ResNet-20,1.5)};
			\addplot[fill=brown]  %PLS
			coordinates {(DenseNet-121,14.6) (MobileNet-V2,14.7) (ResNet-20,4.2)};
			
			\legend{LABPAL-SGD,LABPAL-NSGD,SGD,SLS,PAL,GOLSI,PLS}
		\end{axis}
	\end{tikzpicture}\hspace{1cm}
		\begin{tikzpicture}[scale=0.4]%0.4
		\begin{axis}[
			width= 1.0\linewidth,
			height= 0.7\linewidth,
			ylabel= GPU memory on CIFAR-10 in GB,
			ybar ,
			title = max. GPU memory allocation,
			title style={font=\LARGE},
			xtick=data, 
			%bar shift=0pt, % this must be after ybar to have effect
			symbolic x coords={DenseNet-121,MobileNet-V2,ResNet-20},
			enlarge x limits  = 0.25,
			ylabel style={font=\LARGE},
			xlabel style={font=\LARGE},
			x tick label style={font=\Large},
			y tick label style={font=\Large},
			legend style={font=\LARGE,at={(0.0,1.4)},anchor=north west},
			xtick=data,
			xtick pos=left,
			ytick pos=left,
			ymin=0.0,%ymax=0.87,
			legend columns=2,
			]
			\addplot[fill=blue] %LABPAL-SGD  TODO
			coordinates {(DenseNet-121,2.68) (MobileNet-V2,3.181) (ResNet-20,0.547)};
			\addplot[fill=red]   %LABPAL-NSGD TODO
			coordinates {(DenseNet-121,2.68) (MobileNet-V2,3.181) (ResNet-20,0.547)};
			\addplot[fill=green]  %SGD
			coordinates {(DenseNet-121,2.649) (MobileNet-V2,3.166) (ResNet-20,0.545)};
			\addplot[fill=black]  %SLS
			coordinates {(DenseNet-121,2.696) (MobileNet-V2,3.185) (ResNet-20,0.545)};
			\addplot[fill=orange]  %PAL
			coordinates {(DenseNet-121,2.67) (MobileNet-V2,3.174) (ResNet-20,0.546)};
			\addplot[fill=violet]  %GOLSI
			coordinates {(DenseNet-121,2.692) (MobileNet-V2,3.184) (ResNet-20,0.55)};
			\addplot[fill=brown]  %PLS
			coordinates {(DenseNet-121,6.1) (MobileNet-V2,7.9) (ResNet-20,1.09)};
			
		%	\legend{LABPAL-SGD,LABPAL-NSGD,SGD,SLS,PAL,GOLSI,PLS}
		\end{axis}
	\end{tikzpicture}

	\caption{ \textbf{Left:} Training time comparison on CIFAR-10. SGD, SLS, and PAL show similar training times. GOLSI, and both variants of LABPAL are slightly slower (up to 19.6\%). However, a slightly longer training time is acceptable if less time has to be spent in hyper-parameter tuning.  PLS is significantly slower. Note that in comparison to SGD, the implementations of the other optimizers are not optimized on CUDA level. \textbf{Right:} Maximum allocated memory comparison on CIFAR-10. Except for PLS all approaches need approximately the same amount of memory.}
	\label{labpal_fig_speed_comparison}
\end{figure}
\FloatBarrier

\section{Theoretical considerations}
As the field does not know what the reason for the local parabolic behavior of the full-batch loss is and, thus, what an appropriate function space to consider for convergence is, we refer to the theoretical analysis of \citep{pal}. They show convergence on a quadratic loss. This is also valid for LABPAL, with the addition that each mini batch-loss can be of any form as long as the mean over these losses is a quadratic function. 

\section{Relation of update step adaptation $\alpha$ and the first wolfe constant $w_1$.}
\label{labpal_app_wolfe}
Let $f \;:\; \mathbb{R} \rightarrow  \mathbb{R}$ be of form $x \mapsto ay^2+by+c $.  We start with the first Wolfe condition (a.k.a. Armijo condition, sufficient decrease condition):
\begin{align}
	f(x_0+y)            & \leq f(x_0)-y \nabla f(x_0)w_1 & \text{in our case }x_0=0 , w_1 \text{ wolfe constant} \\
	f(y)                & \leq f(0)+ybw_1                &  \\
	ay^2+by+c           & \leq c+ybw_1                   & \text{use quadratic shape}, \nabla f(x_0)=b           \\
	ay^2+by-ybw_1       & \stackrel{!}{=}0               &  \\
	\frac{ay^2+by}{by}  & =\frac{ay}{b}+1=w_1            &  \\
	-\frac{\alpha}{2}+1 & =w_1                           & \text{set } y=\alpha \frac{-b}{2a}, \alpha \in [1,2)  \\
	-2w_1+2             & =\alpha                             &
\end{align}
\vfill
\FloatBarrier
\clearpage
\section{Further experimental details}
\label{labpal_app_hyperparam_emp_loss_comp}
Further experimental details for the optimizer comparison in Figure \ref{labpal_fig_optimizer_comparison_cifar10},\ref{labpal_fig_optimizer_comparison_cifar100},\ref{labpal_fig_optimizer_comparison_svhn},\ref{labpal_fig_performance_comp_batch_size10},\ref{labpal_fig_opt_comparison_bs10_same},\ref{labpal_fig_opt_comparison_bs10_best} of Sections \ref{labpal_sec_opt_comparison} \& \ref{labpal_sec_low_batch_size}.

\textbf{PLS:} We adapted the only available and empirically improved TensorFlow \citep{Tensorflow} implementation of PLS \citep{probabilisticLineSearchImpl}, which was transferred to PyTorch \citep{PyTorch} by \citep{sls}, to run on several state-of-the-art models and datasets.

The training  steps for the experiments in section Section \ref{labpal_sec_empirical_analysis} were 100,000 for DenseNet and 150,000 steps for MobileNetv2 and ResNet-20. Note that we define one training step as processing one input batch to keep line search approaches comparable.

The batch size was 128 for all experiments.
The validation/train set splits were: 
5,000/45,000 for CIFAR-10 and CIFAR-100 
20,000/45,000 for SVHN.

All images were normalized with a mean and standard deviation determined over the dataset.
We used random horizontal flips and random cropping of size 32. The padding of the random crop was 8 for CIFAR-100 and 4 for SVHN and CIFAR-10.

All trainings were performed on Nvidia Geforce 1080-TI GPUs.

Results were averaged over three runs initialized with three different seeds for each experiment.

For implementation details, refer to the source code provided at \\ \url{https://github.com/cogsys-tuebingen/LABPAL}.

\subsection{Hyperparameter grid search on CIFAR-10}
\label{labpal_app_hyperparams}
\noindent For our evaluation, we used all combinations out of the following hyperparameters.

\textit{SGD}:\\
\begin{tabular}{lll}
	\centering
	hyperparameter & symbol    & values                          \\ \hline
	learning rate              & $\lambda$ & $\{0.001,0.01,0.1,1.0\}$ \\
	momentum                   & $\alpha $ & $\{0,0.4,0.9\}$\\
		learning rate schedule                      &      & \parbox{7cm}{	$\begin{cases}
			\lambda, & \text{if $t\leq \lfloor t_{max} \cdot 0.5\rfloor$}\\
			\lambda/10, & \text{elif $t\leq \lfloor t_{max} \cdot 0.75\rfloor$}\\
			\lambda/100, & \text{elif $t > \lfloor t_{max} \cdot 0.75\rfloor$}\end{cases}$, \\where $t_{max}$ is the amount of training steps}\\
\end{tabular}\\\\

\textit{PAL}:\\
\begin{tabular}{lll}
	\centering
	hyperparameter  & symbol     & values                         \\ \hline
	measuring step size         & $\mu $     & $ \{0.01,0.1,1\}$                   \\
	direction adaptation factor & $\beta $   & $ \{ 0.0,0.4,0.9\} $                   \\
	update step adaptation      & $\alpha $  & $ \{1,1.66\}$                  \\
	maximum step size           & $s_{max} $ & $ \{3.16 \text{ }(\approx 10^{0.5})\}$
\end{tabular}\\\\

\textit{LABPAL (SGD and NSGD)}:\\
\begin{tabular}{lll}
	hyperparameter                           & symbol   & values                        \\ \hline
	step size adaptation                     & $\alpha$ & $\{1.0,1.8,1.9\}$                          \\
	momentum                   				 &  & $\{0,0.4,0.9\}$ 									\\
	SGD steps                                & $n_{SGD}$  & $\{ 1000, 5000 \}$                \\
	approximation step size 				 &  $|\mathbb{B}_a|$ & $\{640,1280\}$           				\\
	batch size schedule                      & $k(t)$     & \parbox{7cm}{	$\begin{cases}
			1, & \text{if $t\leq \lfloor t_{max} \cdot 0.5\rfloor$}\\
			2, & \text{elif $t\leq \lfloor t_{max} \cdot 0.75\rfloor$}\\
			4, & \text{elif $t > \lfloor t_{max} \cdot 0.75\rfloor$}\end{cases}$, \\where $t_{max}$ is the amount of training steps}\\
	measure points for $l_{\mathbb{B}_a,t}$	&	&  $\{(0,0.0001,0.01)\}$
\end{tabular}\\\\

\textit{GOLSI}:\\
\begin{tabular}{lll}
	hyperparameter                     & symbol  & values          \\ \hline
	initial step size                  & $\mu$   & $\{0.001,0.01,0.1,1.0\}$    \\
	momentum                           & $\beta$ & $\{ 0,0.4,0.9\}$      \\
	step size scaling parameter        & $\eta$  & $\{ 0.2,2.0\}$  \\
	modified wolfe condition parameter & $c2$    & $\{ 0.9,0.99\}$
\end{tabular}\\\\

\textit{PLS}:\\
\begin{tabular}{lll}
	hyperparameter                                 & symbol     & values                \\ \hline
	first wolfe condition parameter                & $c_1$      & $\{ 0.3,0.4\}$ \\
	acceptance threshold for the wolfe probability & $cW$       & $\{ 0.1,0.2\}$    \\
	initial step size                              & $\alpha_0$ & $\{0.001,0.01,0.1,1.0\}$\\
	momentum                           & $\beta$ & $\{ 0,0.4,0.9\}$      
\end{tabular}\\\\

\textit{SLS}:\\
\begin{tabular}{lll}
	hyperparameter   & symbol      & values                            \\ \hline
	initial step size & $\mu$       & $\{0.001,0.01,0.1,1.0\}$ 						\\
	step size decay   & $\beta$     & $\{ 0.9,0.99\}$                   \\
	step size reset   & $\gamma$    & $\{ 2.0\}$                     	 \\
	Armijo constant   & $c$         & $\{ 0.1,0.01\}$                     \\
	maximum step size & $\mu_{max}$ & $\{ 10.0\}$                  		     \\
\end{tabular}\\
For SLS no momentum term is considered since \cite{sls} already showed SLS variants using  momentum like acceleration methods to be non-beneficial.

\end{document}

%% file: labpal/math_commands.tex
\usepackage{amsmath,amsfonts,bm}

\def\eqref#1{equation~\ref{#1}}
\def\1{\bm{1}}

\DeclareMathAlphabet{\mathsfit}{\encodingdefault}{\sfdefault}{m}{sl}
\SetMathAlphabet{\mathsfit}{bold}{\encodingdefault}{\sfdefault}{bx}{n}

%% file: commands.tex
\newcommand{\de}{\partial}
\newcommand{\dk}{\tensor[^{\de}]{k}{}}
\newcommand{\ddk}{\tensor[^{\de^2}]{k}{}}
\newcommand{\ddkd}{\tensor[^{\de^2}]{k}{^{\de}}}
\newcommand{\dddk}{\tensor[^{\de^3}]{k}{}}
\newcommand{\dddkd}{\tensor[^{\de^3}]{k}{^{\de}}}
\newcommand{\kd}{\tensor{k}{^{\de}}}
\newcommand{\dkd}{\tensor[^{\de}]{k}{^{\de}}}
\renewcommand{\vec}{\boldsymbol} 
\newcommand{\Id}{\mathbf{I}}
\newcommand{\Trans}{^\top}
\newcommand{\bl}{l_{\mathbb{B},t}}
\newcommand{\fbl}{l_t}
\newcommand{\bld}{l^\prime_{\mathbb{B},t}}
\newcommand{\fbld}{l^\prime_{t}}
\newcommand{\gp}{$\mathcal{GP}$}
\newcommand{\pW}{p^\text{Wolfe}}
\newcommand{\gradk}[1]{\nabla f_{ik}(#1)}
\newcommand{\xk}{\mathbf{\theta}_{t}}
\newcommand{\xkk}{\mathbf{\theta}_{t+1}}
\newcommand{\fk}{\mathcal{L}_{\mathbb{B}_t}}
\newcommand{\normsq}[1]{\left\|#1\right\|^{2}}
\newcommand{\x}{\mathbf{\theta}}

\newcommand{\defequal}{\stackrel{!}{=}}

%PAL
\newcommandx{\BL}[1][1=t]{\mathcal{L}_{\mathbb{B}_{#1}}}
\newcommandx{\Bl}[1][1={,t}]{l_{\mathbb{B}{#1}}}
\newcommandx{\Blh}[1][1={,t}]{\hat{l}_{\mathbb{B}{#1}}}

\renewcommandx{\L}{\mathcal{L}}

%\newcommandx{\BL}[2][1=\space ,2=t]{\mathcal{L}_{\mathbb{B}_{#1},{#2}}}
\newcommandx{\s}[1][1=\space]{s_{\text{upd}_{#1}}}
\newcommandx{\so}[1][1=\space]{s_{\text{opt}_{#1}}}
\renewcommandx{\mathbf}{\boldsymbol}
\newcommand{\pal}{\textit{PAL}}

\newcommand{\tikzremake}{\tikzset{external/force remake}}
\newcommand{\tikzdisable}{\tikzexternaldisable}
\newcommand{\tikzenable}{\tikzexternalenable}

%% file: labpal/figure2_line_plots.tex
\begin{figure}[t!]
\tikzsetfigurename{labpal_line_plots}
\vspace{-0.3cm}
%\figureDataPath
\centering
%\LARGE
\def\scale{0.4}
\begin{tabular}{ c c c }
\scalebox{\scale}{\input{"labpal/figure_data/line_plots/CIFAR10_mom0_resnet20_augment_result/line_plots/bs_128_bs_ori_128_sgd_lrs_[0.01]_pal_mus_[0.1]_ri_100_c_apal_0.9_ori_sgd_lr_0.1/pure_line_10.pgf"}}&
\scalebox{\scale}{\input{"labpal/figure_data/line_plots/CIFAR10_mom0_resnet20_augment_result/line_plots/bs_128_bs_ori_128_sgd_lrs_[0.01]_pal_mus_[0.1]_ri_100_c_apal_0.9_ori_sgd_lr_0.1/pure_line_4000.pgf"}}&
\scalebox{\scale}{\input{"labpal/figure_data/line_plots/CIFAR10_mom0_resnet20_augment_result/line_plots/bs_128_bs_ori_128_sgd_lrs_[0.01]_pal_mus_[0.1]_ri_100_c_apal_0.9_ori_sgd_lr_0.1/pure_line_8000.pgf"}}
\end{tabular}

\caption{Losses along the lines of the SGD training processes exhibit a parabolic shape. The loss of the direction defining mini-batch (green) is excluded from the distribution of mini-batch losses to show that it is significantly different. This makes line searches on it unfavorable. In addition, the parabolic property articulates stronger for the full-batch loss (red); thus, this work aims to approximate it efficiently with a parabola. This introducing figure is created with code and data from \cite{line_analysis}.}
\label{labpal_fig_line_plots}
\end{figure}

%% file: labpal/figure_data/line_plots/CIFAR10_mom0_resnet20_augment_result/line_plots/bs_128_bs_ori_128_sgd_lrs_[0.01]_pal_mus_[0.1]_ri_100_c_apal_0.9_ori_sgd_lr_0.1/pure_line_10.pgf
% This file was created by tikzplotlib v0.9.6.
\begin{tikzpicture}
\begin{axis}[
yticklabel style={font=\large},
xticklabel style={font=\large},
ytick style={font=\Large},
xtick style={font=\Large},
ylabel style={font=\Large},
xlabel style={font=\Large},
title style={font=\Large},
width=10.5cm,
height=7.2cm,
legend cell align={left},
legend style={fill opacity=0.8, draw opacity=1, text opacity=1, draw=white!80!black,font=\large},
tick align=outside,
tick pos=left,
title={line number/t: 10},
x grid style={white!69.0196078431373!black},
xlabel={step on line (s)},
xmin=-0.5496, xmax=0.5416,
xtick style={color=black},
xtick={-0.6,-0.4,-0.2,0,0.2,0.4,0.6},
xticklabels={\(\displaystyle {-0.6}\),\(\displaystyle {-0.4}\),\(\displaystyle {-0.2}\),\(\displaystyle {0.0}\),\(\displaystyle {0.2}\),\(\displaystyle {0.4}\),\(\displaystyle {0.6}\)},
y grid style={white!69.0196078431373!black},
ylabel={loss},
ymin=1.45774640124291, ymax=4.08789222669366,
ytick style={color=black},
ytick={1,1.5,2,2.5,3,3.5,4,4.5},
yticklabels={\(\displaystyle {1.0}\),\(\displaystyle {1.5}\),\(\displaystyle {2.0}\),\(\displaystyle {2.5}\),\(\displaystyle {3.0}\),\(\displaystyle {3.5}\),\(\displaystyle {4.0}\),\(\displaystyle {4.5}\)}
]
\path [draw=blue, fill=blue, opacity=0.2]
(axis cs:-0.5,3.84815073035134)
--(axis cs:-0.5,3.1602260087966)
--(axis cs:-0.494,3.13971026074432)
--(axis cs:-0.488,3.11935540317791)
--(axis cs:-0.482,3.09915807186917)
--(axis cs:-0.476,3.07906927914883)
--(axis cs:-0.47,3.05900771904271)
--(axis cs:-0.464,3.03898699431738)
--(axis cs:-0.458,3.01901599142002)
--(axis cs:-0.452,2.99935887966421)
--(axis cs:-0.446,2.97990664144163)
--(axis cs:-0.44,2.96065695396101)
--(axis cs:-0.434,2.94144537230022)
--(axis cs:-0.428,2.92235194009845)
--(axis cs:-0.422,2.90347832650878)
--(axis cs:-0.416,2.88461403278052)
--(axis cs:-0.41,2.86600646772422)
--(axis cs:-0.404,2.8476850244333)
--(axis cs:-0.398,2.82943564304151)
--(axis cs:-0.392,2.81126313636196)
--(axis cs:-0.386,2.7932844266179)
--(axis cs:-0.38,2.77553806948708)
--(axis cs:-0.374,2.75789333053399)
--(axis cs:-0.368,2.74024525511777)
--(axis cs:-0.362,2.72273823258001)
--(axis cs:-0.356,2.70539375502267)
--(axis cs:-0.35,2.68819518361124)
--(axis cs:-0.344,2.67124674649676)
--(axis cs:-0.338,2.65450174419675)
--(axis cs:-0.332,2.63790338864783)
--(axis cs:-0.326,2.62139891658444)
--(axis cs:-0.32,2.60507081385003)
--(axis cs:-0.314,2.5889493834693)
--(axis cs:-0.308,2.57295028021326)
--(axis cs:-0.302,2.55708826688351)
--(axis cs:-0.296,2.54154667747207)
--(axis cs:-0.29,2.52633520233212)
--(axis cs:-0.284,2.51139367494034)
--(axis cs:-0.278,2.49663734098431)
--(axis cs:-0.272,2.48205179622164)
--(axis cs:-0.266,2.46769169182517)
--(axis cs:-0.26,2.45349366095616)
--(axis cs:-0.254,2.43950396083528)
--(axis cs:-0.248,2.42567669664277)
--(axis cs:-0.242,2.41194335493492)
--(axis cs:-0.236,2.39834392804187)
--(axis cs:-0.23,2.38491384743247)
--(axis cs:-0.224,2.37180767976679)
--(axis cs:-0.218,2.35905975243077)
--(axis cs:-0.212,2.34668821899686)
--(axis cs:-0.206,2.33461577398703)
--(axis cs:-0.2,2.3227746806806)
--(axis cs:-0.194,2.31111995212268)
--(axis cs:-0.188,2.29966705711558)
--(axis cs:-0.182,2.2884818171151)
--(axis cs:-0.176,2.27741191035602)
--(axis cs:-0.17,2.26663216855377)
--(axis cs:-0.164,2.25614513119217)
--(axis cs:-0.158,2.24596033908892)
--(axis cs:-0.152,2.23602445691358)
--(axis cs:-0.146,2.22620082437061)
--(axis cs:-0.14,2.2165997531265)
--(axis cs:-0.134,2.2071442627348)
--(axis cs:-0.128,2.19791812822223)
--(axis cs:-0.122,2.1889913326595)
--(axis cs:-0.116,2.17826003418304)
--(axis cs:-0.11,2.16648628469557)
--(axis cs:-0.104,2.15511106722988)
--(axis cs:-0.0979999999999996,2.14400913519785)
--(axis cs:-0.0919999999999996,2.13317109504715)
--(axis cs:-0.0859999999999996,2.12249988177791)
--(axis cs:-0.0799999999999996,2.11221022764221)
--(axis cs:-0.0739999999999996,2.10237132315524)
--(axis cs:-0.0679999999999996,2.09282927028835)
--(axis cs:-0.0619999999999996,2.08358405390754)
--(axis cs:-0.0559999999999996,2.0746038928628)
--(axis cs:-0.0499999999999996,2.06587764271535)
--(axis cs:-0.0439999999999996,2.05748084327206)
--(axis cs:-0.0379999999999996,2.04930698219687)
--(axis cs:-0.0319999999999996,2.04140782309696)
--(axis cs:-0.0259999999999996,2.03389589395374)
--(axis cs:-0.0199999999999996,2.02675684215501)
--(axis cs:-0.0139999999999996,2.01999342162162)
--(axis cs:-0.00799999999999956,2.01350977178663)
--(axis cs:-0.00199999999999956,2.00737159792334)
--(axis cs:0,2.0053972597234)
--(axis cs:0.006,1.99976893467829)
--(axis cs:0.012,1.99451382341795)
--(axis cs:0.018,1.9895728440024)
--(axis cs:0.024,1.98502840474248)
--(axis cs:0.03,1.98075918806717)
--(axis cs:0.036,1.97672038618475)
--(axis cs:0.042,1.97291176067665)
--(axis cs:0.048,1.9694197550416)
--(axis cs:0.054,1.96616752492264)
--(axis cs:0.06,1.96317852241918)
--(axis cs:0.066,1.96053985506296)
--(axis cs:0.072,1.95815377356485)
--(axis cs:0.078,1.95604792796075)
--(axis cs:0.084,1.95417981827632)
--(axis cs:0.09,1.95255119167268)
--(axis cs:0.096,1.95115533098578)
--(axis cs:0.102,1.95007706107572)
--(axis cs:0.108,1.94921790109947)
--(axis cs:0.114,1.94857269339263)
--(axis cs:0.12,1.94818181172013)
--(axis cs:0.126,1.9480184498243)
--(axis cs:0.132,1.94815633259714)
--(axis cs:0.138,1.94851536210626)
--(axis cs:0.144,1.94903932115994)
--(axis cs:0.15,1.94993944605812)
--(axis cs:0.156,1.95123474416323)
--(axis cs:0.162,1.95287650520913)
--(axis cs:0.168,1.95485955127515)
--(axis cs:0.174,1.95715027069673)
--(axis cs:0.18,1.95972527796403)
--(axis cs:0.186,1.96261190669611)
--(axis cs:0.192,1.96582816285081)
--(axis cs:0.198,1.9665061167907)
--(axis cs:0.204,1.96744645759463)
--(axis cs:0.21,1.9686076969374)
--(axis cs:0.216,1.97006760910153)
--(axis cs:0.222,1.97180933621712)
--(axis cs:0.228,1.9737585936673)
--(axis cs:0.234,1.97594300028868)
--(axis cs:0.24,1.97803572064731)
--(axis cs:0.246,1.97971825883724)
--(axis cs:0.249900922179222,1.98089692334179)
--(axis cs:0.252,1.9815647508949)
--(axis cs:0.258,1.98359832435381)
--(axis cs:0.264,1.98585310287308)
--(axis cs:0.27,1.98828240972944)
--(axis cs:0.276,1.99094630271429)
--(axis cs:0.282,1.99392069986789)
--(axis cs:0.288,1.99715669883881)
--(axis cs:0.294,2.00061874522362)
--(axis cs:0.3,2.0041876474279)
--(axis cs:0.306,2.00792500301031)
--(axis cs:0.312,2.01186356222024)
--(axis cs:0.318,2.01600233768113)
--(axis cs:0.324,2.02039518475067)
--(axis cs:0.33,2.0249572932953)
--(axis cs:0.336,2.02967241051374)
--(axis cs:0.342,2.03459701221436)
--(axis cs:0.348,2.03974131710129)
--(axis cs:0.354,2.04518059888505)
--(axis cs:0.36,2.0508278561756)
--(axis cs:0.366,2.05670305649983)
--(axis cs:0.372,2.06274555908749)
--(axis cs:0.378,2.06906410257216)
--(axis cs:0.384,2.07560760021443)
--(axis cs:0.39,2.08229767245939)
--(axis cs:0.396,2.08907338799327)
--(axis cs:0.402,2.09600828727707)
--(axis cs:0.408,2.1031346777163)
--(axis cs:0.414,2.1104484621028)
--(axis cs:0.42,2.11796188561129)
--(axis cs:0.426,2.12572741665645)
--(axis cs:0.432,2.13364245361299)
--(axis cs:0.438,2.14159622092848)
--(axis cs:0.444,2.14964558574138)
--(axis cs:0.45,2.15780605908367)
--(axis cs:0.456,2.16608162531338)
--(axis cs:0.462,2.1745086422743)
--(axis cs:0.468,2.18309907249932)
--(axis cs:0.474,2.19177530710294)
--(axis cs:0.48,2.20054112182697)
--(axis cs:0.486,2.20946073177038)
--(axis cs:0.492,2.21849069649761)
--(axis cs:0.492,2.73030347446911)
--(axis cs:0.492,2.73030347446911)
--(axis cs:0.486,2.71589839062653)
--(axis cs:0.48,2.70174275332829)
--(axis cs:0.474,2.68778049922548)
--(axis cs:0.468,2.67392959503923)
--(axis cs:0.462,2.6602634434239)
--(axis cs:0.456,2.64677235885756)
--(axis cs:0.45,2.63344120996771)
--(axis cs:0.444,2.6203055222868)
--(axis cs:0.438,2.60728650097735)
--(axis cs:0.432,2.59438243147451)
--(axis cs:0.426,2.58168919337913)
--(axis cs:0.42,2.5692246381077)
--(axis cs:0.414,2.55687405430945)
--(axis cs:0.408,2.5446961306734)
--(axis cs:0.402,2.53267233347287)
--(axis cs:0.396,2.52068605978275)
--(axis cs:0.39,2.50886807404459)
--(axis cs:0.384,2.49722250341438)
--(axis cs:0.378,2.48579389124643)
--(axis cs:0.372,2.47448683355469)
--(axis cs:0.366,2.46339823212475)
--(axis cs:0.36,2.45252327818889)
--(axis cs:0.354,2.44182272581384)
--(axis cs:0.348,2.43128584045917)
--(axis cs:0.342,2.42090349271894)
--(axis cs:0.336,2.41074517578818)
--(axis cs:0.33,2.40079929481726)
--(axis cs:0.324,2.39110558060929)
--(axis cs:0.318,2.38161668321118)
--(axis cs:0.312,2.37238264258485)
--(axis cs:0.306,2.36326776153874)
--(axis cs:0.3,2.3541908467887)
--(axis cs:0.294,2.34525611891877)
--(axis cs:0.288,2.33655965677463)
--(axis cs:0.282,2.32804454304278)
--(axis cs:0.276,2.31971536739729)
--(axis cs:0.27,2.31158020324074)
--(axis cs:0.264,2.30358596588485)
--(axis cs:0.258,2.29574691480957)
--(axis cs:0.252,2.28809546213597)
--(axis cs:0.249900922179222,2.28546491474845)
--(axis cs:0.246,2.28057918860577)
--(axis cs:0.24,2.27317836903967)
--(axis cs:0.234,2.26599255390465)
--(axis cs:0.228,2.25904660439119)
--(axis cs:0.222,2.25242299796082)
--(axis cs:0.216,2.24609054555185)
--(axis cs:0.21,2.23994932835922)
--(axis cs:0.204,2.23404365312308)
--(axis cs:0.198,2.22837456967682)
--(axis cs:0.192,2.22301277704537)
--(axis cs:0.186,2.2178719593212)
--(axis cs:0.18,2.21295338496566)
--(axis cs:0.174,2.2081973743625)
--(axis cs:0.168,2.20364573132247)
--(axis cs:0.162,2.19946242263541)
--(axis cs:0.156,2.19556087767705)
--(axis cs:0.15,2.19200191972777)
--(axis cs:0.144,2.18870069924742)
--(axis cs:0.138,2.18761233519763)
--(axis cs:0.132,2.189654964488)
--(axis cs:0.126,2.19184478279203)
--(axis cs:0.12,2.1942296740599)
--(axis cs:0.114,2.1968731819652)
--(axis cs:0.108,2.19978097826242)
--(axis cs:0.102,2.20288859633729)
--(axis cs:0.096,2.20627187890932)
--(axis cs:0.09,2.20996835594997)
--(axis cs:0.084,2.21394501114264)
--(axis cs:0.078,2.21816101763397)
--(axis cs:0.072,2.22263816744089)
--(axis cs:0.066,2.22752176737413)
--(axis cs:0.06,2.23266341304407)
--(axis cs:0.054,2.23806704720482)
--(axis cs:0.048,2.24368551932275)
--(axis cs:0.042,2.24961025221273)
--(axis cs:0.036,2.25580629566684)
--(axis cs:0.03,2.26230163825676)
--(axis cs:0.024,2.26914245029911)
--(axis cs:0.018,2.27614079229534)
--(axis cs:0.012,2.28343535307795)
--(axis cs:0.006,2.29100046399981)
--(axis cs:0,2.29889268660918)
--(axis cs:-0.00199999999999956,2.30159606272355)
--(axis cs:-0.00799999999999956,2.30988669348881)
--(axis cs:-0.0139999999999996,2.31838059239089)
--(axis cs:-0.0199999999999996,2.32712158514187)
--(axis cs:-0.0259999999999996,2.33604703936726)
--(axis cs:-0.0319999999999996,2.34531276347116)
--(axis cs:-0.0379999999999996,2.35488555533811)
--(axis cs:-0.0439999999999996,2.36476917844266)
--(axis cs:-0.0499999999999996,2.37499809078872)
--(axis cs:-0.0559999999999996,2.38554258504882)
--(axis cs:-0.0619999999999996,2.39628988038749)
--(axis cs:-0.0679999999999996,2.40733466716483)
--(axis cs:-0.0739999999999996,2.4186238180846)
--(axis cs:-0.0799999999999996,2.43026058701798)
--(axis cs:-0.0859999999999996,2.44222303712741)
--(axis cs:-0.0919999999999996,2.45432196231559)
--(axis cs:-0.0979999999999996,2.46674100938253)
--(axis cs:-0.104,2.47950850543566)
--(axis cs:-0.11,2.4925207993947)
--(axis cs:-0.116,2.50580606143922)
--(axis cs:-0.122,2.51945194229484)
--(axis cs:-0.128,2.53323614201508)
--(axis cs:-0.134,2.54712723684497)
--(axis cs:-0.14,2.56124278414063)
--(axis cs:-0.146,2.57572358590551)
--(axis cs:-0.152,2.59058574866503)
--(axis cs:-0.158,2.60573146631941)
--(axis cs:-0.164,2.62120130914263)
--(axis cs:-0.17,2.63694772054441)
--(axis cs:-0.176,2.65299422293901)
--(axis cs:-0.182,2.66931073390879)
--(axis cs:-0.188,2.68596101365983)
--(axis cs:-0.194,2.70291628804989)
--(axis cs:-0.2,2.72008187859319)
--(axis cs:-0.206,2.737499904586)
--(axis cs:-0.212,2.75504031032324)
--(axis cs:-0.218,2.77271668892354)
--(axis cs:-0.224,2.79059522063471)
--(axis cs:-0.23,2.80860415659845)
--(axis cs:-0.236,2.82688776263967)
--(axis cs:-0.242,2.84526621131226)
--(axis cs:-0.248,2.86381709191483)
--(axis cs:-0.254,2.88275157811586)
--(axis cs:-0.26,2.90200294321403)
--(axis cs:-0.266,2.92149527859874)
--(axis cs:-0.272,2.9411947906483)
--(axis cs:-0.278,2.96120997087564)
--(axis cs:-0.284,2.98135763907339)
--(axis cs:-0.29,3.00176849064883)
--(axis cs:-0.296,3.02248319564387)
--(axis cs:-0.302,3.04342579981312)
--(axis cs:-0.308,3.06462792388629)
--(axis cs:-0.314,3.08599445957225)
--(axis cs:-0.32,3.10755575227086)
--(axis cs:-0.326,3.12932392477524)
--(axis cs:-0.332,3.15131237695459)
--(axis cs:-0.338,3.17347852769308)
--(axis cs:-0.344,3.19596750696655)
--(axis cs:-0.35,3.21874927880708)
--(axis cs:-0.356,3.24170670297462)
--(axis cs:-0.362,3.26480530318804)
--(axis cs:-0.368,3.28811197169125)
--(axis cs:-0.374,3.31156011996791)
--(axis cs:-0.38,3.33529919711873)
--(axis cs:-0.386,3.35969098703936)
--(axis cs:-0.392,3.38470076998055)
--(axis cs:-0.398,3.40970750205452)
--(axis cs:-0.404,3.43466201341653)
--(axis cs:-0.41,3.45976015921042)
--(axis cs:-0.416,3.48493973336008)
--(axis cs:-0.422,3.51014649566787)
--(axis cs:-0.428,3.53534924397536)
--(axis cs:-0.434,3.56073805270717)
--(axis cs:-0.44,3.58632993535139)
--(axis cs:-0.446,3.61205331103702)
--(axis cs:-0.452,3.63793825016182)
--(axis cs:-0.458,3.66384064088925)
--(axis cs:-0.464,3.68990605478029)
--(axis cs:-0.47,3.71604879843653)
--(axis cs:-0.476,3.74224532698281)
--(axis cs:-0.482,3.76856440273696)
--(axis cs:-0.488,3.79499321259209)
--(axis cs:-0.494,3.82155781370238)
--(axis cs:-0.5,3.84815073035134)
--cycle;
\addlegendimage{area legend, draw=blue, fill=blue, opacity=0.2}
\addlegendentry{fill between min. and max. batch losses}

\addplot [semithick, black, forget plot]
table {%
-0.5 3.50321282331061
-0.494 3.47936451134774
-0.488 3.45585257961102
-0.482 3.43238509915136
-0.476 3.40903926184001
-0.47 3.38580047198593
-0.464 3.3626673646304
-0.458 3.33979386463034
-0.452 3.31712702915684
-0.446 3.2945859661304
-0.44 3.27199787200516
-0.434 3.24949322655084
-0.428 3.22718949623959
-0.422 3.20502081406448
-0.416 3.18297802182133
-0.41 3.16115326213185
-0.404 3.13943159928385
-0.398 3.11770028404499
-0.392 3.0960792064725
-0.386 3.0746071167996
-0.38 3.05320535615101
-0.374 3.03198221691855
-0.368 3.0108857003288
-0.362 2.98993450936541
-0.356 2.96917686169763
-0.35 2.9486242123894
-0.344 2.92839599551371
-0.338 2.90837913498399
-0.332 2.88839630471921
-0.326 2.8685490283533
-0.32 2.84886029233894
-0.314 2.82931609819207
-0.308 2.81008028540964
-0.302 2.7910461241845
-0.296 2.7722432317023
-0.29 2.75366339660832
-0.284 2.73524685445591
-0.278 2.71708329201647
-0.272 2.69918713389779
-0.266 2.68156026866927
-0.26 2.66419671625772
-0.254 2.64707428043766
-0.248 2.62987632557633
-0.242 2.61276627000188
-0.236 2.59588659289875
-0.23 2.57928265872761
-0.224 2.56284526998934
-0.218 2.54655373815331
-0.212 2.53053569963959
-0.206 2.51471007037617
-0.2 2.49921819131123
-0.194 2.48405233497033
-0.188 2.4693554143887
-0.182 2.45524924746132
-0.176 2.44137463875813
-0.17 2.42780558366212
-0.164 2.41452391399071
-0.158 2.4012101724511
-0.152 2.38717626204016
-0.146 2.37435277662007
-0.14 2.36207735416247
-0.134 2.34801920392783
-0.128 2.33319083147217
-0.122 2.31869100045878
-0.116 2.30438698135549
-0.11 2.29030583932763
-0.104 2.27678908623056
-0.0979999999999996 2.26391783804866
-0.0919999999999996 2.25140031735646
-0.0859999999999996 2.23958548309747
-0.0799999999999996 2.22816719301045
-0.0739999999999996 2.21700673585292
-0.0679999999999996 2.20618954551173
-0.0619999999999996 2.1961421414162
-0.0559999999999996 2.18632292095572
-0.0499999999999996 2.17678730451735
-0.0439999999999996 2.16752336162608
-0.0379999999999996 2.15857796306955
-0.0319999999999996 2.14991016715067
-0.0259999999999996 2.14156625024043
-0.0199999999999996 2.1335404717247
-0.0139999999999996 2.12583593523595
-0.00799999999999956 2.11848008650122
-0.00199999999999956 2.11147264559986
0 2.10922044998733
0.006 2.09999718307517
0.012 2.08939985954203
0.018 2.08467811736045
0.024 2.08029007737059
0.03 2.07557423424441
0.036 2.06971520505613
0.042 2.06417533464264
0.048 2.0588761944673
0.054 2.05418884078972
0.06 2.05185315734707
0.066 2.04581368673826
0.072 2.04109747573966
0.078 2.03779690992087
0.084 2.03481630719034
0.09 2.03211876144633
0.096 2.02985940797953
0.102 2.02862473018467
0.108 2.02772648970131
0.114 2.0270783820888
0.12 2.02698101184797
0.126 2.02800607559038
0.132 2.02965331601445
0.138 2.0316083743237
0.144 2.03396748017985
0.15 2.03651474323124
0.156 2.03805468673818
0.162 2.03933073265944
0.168 2.04143232246861
0.174 2.04200693638995
0.18 2.04329495178536
0.186 2.04434788716026
0.192 2.04457984643523
0.198 2.04570975818206
0.204 2.04740053281421
0.21 2.04963576153386
0.216 2.0513979740208
0.222 2.05341993039474
0.228 2.05571233690716
0.234 2.0582148678368
0.24 2.06094287964515
0.246 2.06387457984965
0.249900922179222 2.06587024801411
0.252 2.06695726432372
0.258 2.07016792765353
0.264 2.07359660736984
0.27 2.0778107563965
0.276 2.08220132190036
0.282 2.08692346187308
0.288 2.09190167789347
0.294 2.09703885926865
0.3 2.10208065033657
0.306 2.10781125311041
0.312 2.1133409277536
0.318 2.11898481252138
0.324 2.12487734359456
0.33 2.13095391885145
0.336 2.13726282742573
0.342 2.1438390527619
0.348 2.15066673397087
0.354 2.15777481367695
0.36 2.16589271256817
0.366 2.17388204907184
0.372 2.18136809770658
0.378 2.19042082346277
0.384 2.19836707675131
0.39 2.20653016190045
0.396 2.21478559344541
0.402 2.22318642120808
0.408 2.23176723578945
0.414 2.24048728152411
0.42 2.24938891342026
0.426 2.25849793289672
0.432 2.26772564035491
0.438 2.27703055462916
0.444 2.28651725719101
0.45 2.29613716996391
0.456 2.30594032534282
0.462 2.31591299068532
0.468 2.32597104556044
0.474 2.3361347989121
0.48 2.34642569796415
0.486 2.35686903633177
0.492 2.36744212926715
};
\addplot [semithick, black, forget plot]
table {%
-0.5 3.58638829228585
-0.494 3.56169749343826
-0.488 3.5371342645667
-0.482 3.51271861648274
-0.476 3.48836616493645
-0.47 3.46414428477146
-0.464 3.43999822360638
-0.458 3.41597420556354
-0.452 3.39207531949796
-0.446 3.36831518835243
-0.44 3.3447256429863
-0.434 3.32126871630317
-0.428 3.29787531610054
-0.422 3.2745266189595
-0.416 3.25138038318255
-0.41 3.22848084426369
-0.404 3.20610769087216
-0.398 3.18427740185871
-0.392 3.1625931050512
-0.386 3.14106808038196
-0.38 3.11968886028626
-0.374 3.09850046900101
-0.368 3.07747320318595
-0.362 3.05661374321789
-0.356 3.03597009851364
-0.35 3.01538447264466
-0.344 2.99489009645185
-0.338 2.97444701162749
-0.332 2.954050861008
-0.326 2.93386116303736
-0.32 2.91396004191483
-0.314 2.89432701011538
-0.308 2.87491418936406
-0.302 2.85575389961014
-0.296 2.83672853355529
-0.29 2.81789518607548
-0.284 2.79922977637034
-0.278 2.78027191152796
-0.272 2.76029120187741
-0.266 2.74053688562708
-0.26 2.7210465822136
-0.254 2.70173630712088
-0.248 2.68268070323393
-0.242 2.66381749301217
-0.236 2.64521250640973
-0.23 2.62682986131404
-0.224 2.60865056631155
-0.218 2.59074200876057
-0.212 2.5730992150493
-0.206 2.55566403875127
-0.2 2.53921178521705
-0.194 2.52319625639939
-0.188 2.50736911586137
-0.182 2.49174555856735
-0.176 2.47641450460651
-0.17 2.46119989603176
-0.164 2.44600117165828
-0.158 2.43102859513601
-0.152 2.41627094615251
-0.146 2.40181872318499
-0.14 2.38761124934535
-0.134 2.37365102017066
-0.128 2.35992205020739
-0.122 2.34636856854195
-0.116 2.33306263270788
-0.11 2.32002770976396
-0.104 2.30724157945951
-0.0979999999999996 2.2949587056064
-0.0919999999999996 2.2847580415546
-0.0859999999999996 2.27488492737757
-0.0799999999999996 2.26549875165801
-0.0739999999999996 2.25692772516049
-0.0679999999999996 2.24764858238632
-0.0619999999999996 2.23786225769436
-0.0559999999999996 2.22837732266635
-0.0499999999999996 2.21931548591238
-0.0439999999999996 2.20980657928158
-0.0379999999999996 2.20000636368059
-0.0319999999999996 2.18973860831466
-0.0259999999999996 2.18054727150593
-0.0199999999999996 2.17209284915589
-0.0139999999999996 2.16419650055468
-0.00799999999999956 2.15655070566572
-0.00199999999999956 2.14913439878728
0 2.14671526569873
0.006 2.139613754116
0.012 2.13276576669887
0.018 2.12622521934099
0.024 2.11990639474243
0.03 2.11380950524472
0.036 2.10806208662689
0.042 2.10334037104622
0.048 2.09889438282698
0.054 2.09477999736555
0.06 2.09095464809798
0.066 2.08747825701721
0.072 2.08439208264463
0.078 2.08165996079333
0.084 2.07921935268678
0.09 2.07708817813545
0.096 2.07499864883721
0.102 2.07224732148461
0.108 2.06900152121671
0.114 2.06594013795257
0.12 2.06391004705802
0.126 2.06140107125975
0.132 2.05893283919431
0.138 2.05703715863638
0.144 2.05692518688738
0.15 2.05721378326416
0.156 2.05911872803699
0.162 2.06112950609531
0.168 2.06265574682038
0.174 2.06412834208459
0.18 2.06586154340766
0.186 2.06784419086762
0.192 2.07004898111336
0.198 2.07250723312609
0.204 2.07468824996613
0.21 2.07759474008344
0.216 2.08130345464451
0.222 2.08578350942116
0.228 2.08922901714686
0.234 2.09229095210321
0.24 2.09512390149757
0.246 2.09910669969395
0.249900922179222 2.10184906132054
0.252 2.10336581745651
0.258 2.10789776372258
0.264 2.11271213577129
0.27 2.1178387820255
0.276 2.12325873068767
0.282 2.12834309064783
0.288 2.13360168744111
0.294 2.13907572673634
0.3 2.14476033387473
0.306 2.15070129465312
0.312 2.15684688073816
0.318 2.16375127772335
0.324 2.17102337803226
0.33 2.17845710885013
0.336 2.1860761285061
0.342 2.19392980303383
0.348 2.20199002907611
0.354 2.21019548771437
0.36 2.21858558710665
0.366 2.22715087793767
0.372 2.23587354499614
0.378 2.24481950455811
0.384 2.25397427211283
0.39 2.26315660055843
0.396 2.27228033475694
0.402 2.28158241027268
0.408 2.29098763465299
0.414 2.30053863421199
0.42 2.31025827684789
0.426 2.32016793484217
0.432 2.33023217643495
0.438 2.34044591899146
0.444 2.35096996897482
0.45 2.36199500551447
0.456 2.37320317531703
0.462 2.38456377194962
0.468 2.39608041028259
0.474 2.40782007238886
0.48 2.41974092231249
0.486 2.43182845335105
0.492 2.44398298424494
};
\addplot [semithick, black]
table {%
-0.5 3.71168761065292
-0.494 3.685759820557
-0.488 3.6599345309487
-0.482 3.63417483493868
-0.476 3.60852827463805
-0.47 3.58303364203857
-0.464 3.5577341960261
-0.458 3.53252567964591
-0.452 3.50742088322113
-0.446 3.48239261231629
-0.44 3.45749828522912
-0.434 3.43279033494855
-0.428 3.40836840267912
-0.422 3.38409393284383
-0.416 3.36001341434712
-0.41 3.33599452358794
-0.404 3.31214015026308
-0.398 3.28836721377229
-0.392 3.26472443911553
-0.386 3.24130944425451
-0.38 3.21801464691634
-0.374 3.19495270732114
-0.368 3.17200808831876
-0.362 3.14921602402137
-0.356 3.12659251520745
-0.35 3.10416757737949
-0.344 3.08190058928267
-0.338 3.0597060119926
-0.332 3.03768996397412
-0.326 3.01574930671995
-0.32 2.99392805282696
-0.314 2.97230334545748
-0.308 2.9509006313383
-0.302 2.92978369713819
-0.296 2.90892810081641
-0.29 2.8883187459578
-0.284 2.86793304802632
-0.278 2.84775220524898
-0.272 2.82781237382733
-0.266 2.8081683287819
-0.26 2.78867885793807
-0.254 2.76940852182815
-0.248 2.75030914987292
-0.242 2.73135649914911
-0.236 2.7126311282118
-0.23 2.69410033553868
-0.224 2.67584456547047
-0.218 2.657831839766
-0.212 2.64005708946206
-0.206 2.62256083797547
-0.2 2.60532364483515
-0.194 2.58821887716476
-0.188 2.57113309923443
-0.182 2.5542455574323
-0.176 2.53766526276013
-0.17 2.52135195527808
-0.164 2.5052644622192
-0.158 2.48877686096239
-0.152 2.47225484211231
-0.146 2.45728446298745
-0.14 2.44405628246022
-0.134 2.43008972943062
-0.128 2.41698744008318
-0.122 2.40337901079329
-0.116 2.39006213151151
-0.11 2.37701435136842
-0.104 2.36372148190276
-0.0979999999999996 2.35060076677473
-0.0919999999999996 2.33751298265997
-0.0859999999999996 2.32470401251339
-0.0799999999999996 2.31207417164114
-0.0739999999999996 2.29966367332963
-0.0679999999999996 2.28767571796197
-0.0619999999999996 2.27600476850057
-0.0559999999999996 2.26432247669436
-0.0499999999999996 2.25237169279717
-0.0439999999999996 2.24181989725912
-0.0379999999999996 2.2323351529194
-0.0319999999999996 2.22304979129694
-0.0259999999999996 2.21372034499655
-0.0199999999999996 2.2041292579961
-0.0139999999999996 2.1964039276354
-0.00799999999999956 2.18777274701279
-0.00199999999999956 2.17942848126404
0 2.17671787680592
0.006 2.16851551504806
0.012 2.16044495266397
0.018 2.15403328393586
0.024 2.14791422226699
0.03 2.14209990313975
0.036 2.13554626167752
0.042 2.12947113870177
0.048 2.12373313726857
0.054 2.11776027269661
0.06 2.11167363624554
0.066 2.10694635927211
0.072 2.10255784040783
0.078 2.0984321765136
0.084 2.09459915047046
0.09 2.09207091655117
0.096 2.09006236982532
0.102 2.08832898712717
0.108 2.08637701126281
0.114 2.08456434798427
0.12 2.08312670513988
0.126 2.08196586696431
0.132 2.08105004869867
0.138 2.0804000768112
0.144 2.08001213433454
0.15 2.08208132057916
0.156 2.08537380443886
0.162 2.08863028429914
0.168 2.08807779249037
0.174 2.08786364912521
0.18 2.08792093856027
0.186 2.08851497416617
0.192 2.09090057498543
0.198 2.09353100013686
0.204 2.09645993262529
0.21 2.09988622175297
0.216 2.10354344715597
0.222 2.10737667171634
0.228 2.11137331923237
0.234 2.11560430008103
0.24 2.12082219601143
0.246 2.12665875663515
0.249900922179222 2.13061028844095
0.252 2.13279257831164
0.258 2.13917459142976
0.264 2.14567689146497
0.27 2.15238196135033
0.276 2.15899961671676
0.282 2.1649912353314
0.288 2.17074604664231
0.294 2.1765948770626
0.3 2.18411774653941
0.306 2.19255956038251
0.312 2.20108949171845
0.318 2.20891678131011
0.324 2.21690503398713
0.33 2.22499847755535
0.336 2.23340447695227
0.342 2.24337812120211
0.348 2.25354253899422
0.354 2.26393544595339
0.36 2.27420356421499
0.366 2.28441291471245
0.372 2.29449645182467
0.378 2.3036502921459
0.384 2.31215317087481
0.39 2.32064429775346
0.396 2.32941680168733
0.402 2.33822215907276
0.408 2.34672666661208
0.414 2.3553323361557
0.42 2.36419662076514
0.426 2.37318233045517
0.432 2.38228656648425
0.438 2.39196283020283
0.444 2.40273406277993
0.45 2.41366439073317
0.456 2.4247512434813
0.462 2.43597914998099
0.468 2.44737310148048
0.474 2.45882565433203
0.48 2.47046111366944
0.486 2.48228099211337
0.492 2.49421693301338
};
\addlegendentry{quartiles of batch losses}
\addplot [very thick, green!50.1960784313725!black]
table {%
-0.5 3.9800013112108
-0.494 3.95183666601224
-0.488 3.92380011461501
-0.482 3.89589445278398
-0.476 3.86818251053046
-0.47 3.84054581301461
-0.464 3.81293522143096
-0.458 3.78541358752409
-0.452 3.75796692754375
-0.446 3.73067040878232
-0.44 3.7035126585688
-0.434 3.67656563130731
-0.428 3.64981131427339
-0.422 3.62315536735696
-0.416 3.5966487191763
-0.41 3.57032988657011
-0.404 3.54412084564683
-0.398 3.51791043656704
-0.392 3.49187276465818
-0.386 3.465893329354
-0.38 3.4400331255456
-0.374 3.41435167734744
-0.368 3.3887822861725
-0.362 3.36329624915379
-0.356 3.33801303434302
-0.35 3.3128633700544
-0.344 3.28790953909629
-0.338 3.26322786070523
-0.332 3.23872191170813
-0.326 3.21434058851446
-0.32 3.1899954068067
-0.314 3.16570687436615
-0.308 3.14165886334376
-0.302 3.11758196933079
-0.296 3.09362491735374
-0.29 3.06972447494627
-0.284 3.04597388769616
-0.278 3.02238863881212
-0.272 2.99893493385753
-0.266 2.97565472882707
-0.26 2.95254130091053
-0.254 2.92963708506431
-0.248 2.90684395667631
-0.242 2.88427365588723
-0.236 2.86193279869622
-0.23 2.83969466300914
-0.224 2.81763927341672
-0.218 2.79573369910941
-0.212 2.77389330300502
-0.206 2.75212710746564
-0.2 2.73061697359663
-0.194 2.70919562852941
-0.188 2.68799320305698
-0.182 2.66703601996414
-0.176 2.64622091746423
-0.17 2.62553385796491
-0.164 2.60504007863346
-0.158 2.58463027956896
-0.152 2.56442636449356
-0.146 2.54441909736488
-0.14 2.52448522567283
-0.134 2.50475657603238
-0.128 2.48526654101443
-0.122 2.46600270352792
-0.116 2.44685753493104
-0.11 2.42786689661443
-0.104 2.40903451526538
-0.0979999999999996 2.39030020497739
-0.0919999999999996 2.37182736955583
-0.0859999999999996 2.35363613511436
-0.0799999999999996 2.33561610640027
-0.0739999999999996 2.31783077330329
-0.0679999999999996 2.30020285048522
-0.0619999999999996 2.28283450985327
-0.0559999999999996 2.26577413990162
-0.0499999999999996 2.24885220034048
-0.0439999999999996 2.23205685825087
-0.0379999999999996 2.21554174087942
-0.0319999999999996 2.19932807656005
-0.0259999999999996 2.18334385356866
-0.0199999999999996 2.16760246548802
-0.0139999999999996 2.15205128700472
-0.00799999999999956 2.13665255601518
-0.00199999999999956 2.12146993284114
0 2.11645479477011
0.006 2.10165981249884
0.012 2.08716706419364
0.018 2.07304293150082
0.024 2.05935036134906
0.03 2.04612372373231
0.036 2.03338847239502
0.042 2.02103769523092
0.048 2.00908065517433
0.054 1.99757902417332
0.06 1.98647006275132
0.066 1.97574342368171
0.072 1.96544824354351
0.078 1.95554769295268
0.084 1.94610671838745
0.09 1.93707594554871
0.096 1.92841106001288
0.102 1.92009606212378
0.108 1.912178712897
0.114 1.90467872470617
0.12 1.89762550964952
0.126 1.89099032711238
0.132 1.88477906165645
0.138 1.87882076972164
0.144 1.87317559192888
0.15 1.86793848127127
0.156 1.8629983891733
0.162 1.8584494178649
0.168 1.85417394456454
0.174 1.85008367453702
0.18 1.84631250682287
0.186 1.84277488628868
0.192 1.83960908977315
0.198 1.83675955876242
0.204 1.83407149964478
0.21 1.83168778591789
0.216 1.82959311222658
0.222 1.82768766372465
0.228 1.82600567990448
0.234 1.82463178050239
0.24 1.82360551180318
0.246 1.82283615908818
0.249900922179222 1.82249802461592
0.252 1.82236593530979
0.258 1.82218300155364
0.264 1.82231083500665
0.27 1.82272674876731
0.276 1.82345506508136
0.282 1.82436527480604
0.288 1.82554034225177
0.294 1.82688681111904
0.3 1.82846318237716
0.306 1.8303070536931
0.312 1.83242947963299
0.318 1.83474777726224
0.324 1.83734127751086
0.33 1.84024149979814
0.336 1.84335348565946
0.342 1.84671927129966
0.348 1.85032985580619
0.354 1.85412422454101
0.36 1.85807192874199
0.366 1.86226188042201
0.372 1.86666574105038
0.378 1.87124498019693
0.384 1.87595344526926
0.39 1.88080620544497
0.396 1.88584733908647
0.402 1.89103088475531
0.408 1.89639497907774
0.414 1.90194682743459
0.42 1.90766648810677
0.426 1.91348918853328
0.432 1.91948344882985
0.438 1.92555063586769
0.444 1.93171308098681
0.45 1.93805055656412
0.456 1.94455973889853
0.462 1.95130504517147
0.468 1.95833459322603
0.474 1.96554004281643
0.48 1.97284558265164
0.486 1.98027115033983
0.492 1.98793509612369
};
\addlegendentry{loss of direction defining batch $l_{batch}(s;\mathbb{B}_t)_t$}
\addplot [very thick, red]
table {%
-0.5 3.59972865278951
-0.494 3.57491659924412
-0.488 3.55020748401216
-0.482 3.52562993600765
-0.476 3.50118318953349
-0.47 3.47683809944704
-0.464 3.45260919749398
-0.458 3.4284969975991
-0.452 3.40452171462101
-0.446 3.38069126841248
-0.44 3.35699639195582
-0.434 3.33344005152128
-0.428 3.30998985312777
-0.422 3.28664766751916
-0.416 3.26342731455378
-0.41 3.24034919657777
-0.404 3.21741721465469
-0.398 3.19462643456529
-0.392 3.1719766930197
-0.386 3.1494723051236
-0.38 3.12711368099701
-0.374 3.10492052057305
-0.368 3.08288286604898
-0.362 3.06100268579428
-0.356 3.03928450145099
-0.35 3.0177338821162
-0.344 2.99637355663401
-0.338 2.97518402597404
-0.332 2.95416544781665
-0.326 2.93332224881601
-0.32 2.91266051839557
-0.314 2.89216444440058
-0.308 2.87186028735186
-0.302 2.8517431986395
-0.296 2.83181833883782
-0.29 2.81208568075954
-0.284 2.79256555591815
-0.278 2.77323424682072
-0.272 2.75410296060932
-0.266 2.73518638076121
-0.26 2.71646766785261
-0.254 2.69798284401213
-0.248 2.6797143350409
-0.242 2.66165904479976
-0.236 2.64383023004984
-0.23 2.62622626268694
-0.224 2.60885594969523
-0.218 2.59171819035191
-0.212 2.57481622118212
-0.206 2.55814065584991
-0.2 2.54169459490549
-0.194 2.52545442805604
-0.188 2.50946358243937
-0.182 2.49373373213712
-0.176 2.47826191520198
-0.17 2.46306422662747
-0.164 2.44811580640453
-0.158 2.43339864807422
-0.152 2.41893394295494
-0.146 2.40470967347483
-0.14 2.39073236600234
-0.134 2.37702555544547
-0.128 2.36357269664131
-0.122 2.35037916846886
-0.116 2.33745413889665
-0.11 2.32478936530242
-0.104 2.31239498707146
-0.0979999999999996 2.30027312419952
-0.0919999999999996 2.28844731032405
-0.0859999999999996 2.27692005336939
-0.0799999999999996 2.26566350413668
-0.0739999999999996 2.25469957880887
-0.0679999999999996 2.24402207243475
-0.0619999999999996 2.23363522482032
-0.0559999999999996 2.22354390032342
-0.0499999999999996 2.21374436571932
-0.0439999999999996 2.20424185712181
-0.0379999999999996 2.19506535959226
-0.0319999999999996 2.18619632041373
-0.0259999999999996 2.17762158446567
-0.0199999999999996 2.16933813811193
-0.0139999999999996 2.16134865503045
-0.00799999999999956 2.15366091416217
-0.00199999999999956 2.14628242199251
0 2.14388450185652
0.006 2.13688219397591
0.012 2.1301984522288
0.018 2.12383950703952
0.024 2.11778729895741
0.03 2.11204386933241
0.036 2.10659928064706
0.042 2.10145956750057
0.048 2.0966120063531
0.054 2.09207749778579
0.06 2.08785358965542
0.066 2.08393445869297
0.072 2.08031561299867
0.078 2.07701526791061
0.084 2.07400096383208
0.09 2.07126398434775
0.096 2.06881718082877
0.102 2.06666110147489
0.108 2.06479437597591
0.114 2.06320830804907
0.12 2.06191271833814
0.126 2.06089030732255
0.132 2.06013055022777
0.138 2.05964400493394
0.144 2.05943603464402
0.15 2.05952159543813
0.156 2.05987471768458
0.162 2.0604859186642
0.168 2.06136791601602
0.174 2.06251497687481
0.18 2.06392633221549
0.186 2.06558559097903
0.192 2.06750742657823
0.198 2.06969540602222
0.204 2.07214222773473
0.21 2.0748387206404
0.216 2.07778690415216
0.222 2.08096080276664
0.228 2.084370234159
0.234 2.088022844433
0.24 2.09191760540125
0.246 2.09602344088671
0.249900922179222 2.09881865611715
0.252 2.10036220387337
0.258 2.10492885546046
0.264 2.10971839810736
0.27 2.11474343181908
0.276 2.11999571023625
0.282 2.12548001338473
0.288 2.13118125236178
0.294 2.13710096674185
0.3 2.14323095675354
0.306 2.14956421134957
0.312 2.1560876200565
0.318 2.16279801607925
0.324 2.16971272490628
0.33 2.17682434214203
0.336 2.18413071851955
0.342 2.19162951923772
0.348 2.1993118720693
0.354 2.20718784333985
0.36 2.21526237278476
0.366 2.2235232704179
0.372 2.23199031838885
0.378 2.24064393535446
0.384 2.24946585600628
0.39 2.25847311912094
0.396 2.26764878940048
0.402 2.27700005220595
0.408 2.2865066156287
0.414 2.29617739277955
0.42 2.30603033565171
0.426 2.31605662829224
0.432 2.32625074937187
0.438 2.33659611792086
0.444 2.34711253302362
0.45 2.35778979015458
0.456 2.36863781329112
0.462 2.37964663612479
0.468 2.39081729347186
0.474 2.40214011896705
0.48 2.41360279130129
0.486 2.42522497028233
0.492 2.43700221711936
};
\addlegendentry{full batch loss $l(s)_t$}
\end{axis}

\end{tikzpicture}

%% file: labpal/figure_data/line_plots/CIFAR10_mom0_resnet20_augment_result/line_plots/bs_128_bs_ori_128_sgd_lrs_[0.01]_pal_mus_[0.1]_ri_100_c_apal_0.9_ori_sgd_lr_0.1/pure_line_4000.pgf
% This file was created by tikzplotlib v0.9.6.
\begin{tikzpicture}

\begin{axis}[
yticklabel style={font=\large},
xticklabel style={font=\large},
ytick style={font=\Large},
xtick style={font=\Large},
ylabel style={font=\Large},
xlabel style={font=\Large},
title style={font=\Large},
width=10.5cm,
height=7.2cm,
tick align=outside,
tick pos=left,
title={line number/t: 4,000},
x grid style={white!69.0196078431373!black},
xlabel={step on line (s)},
xmin=-0.5496, xmax=0.5416,
xtick style={color=black},
xtick={-0.6,-0.4,-0.2,0,0.2,0.4,0.6},
y grid style={white!69.0196078431373!black},
ylabel={loss},
ymin=0.0117527576412298, ymax=0.6,
ytick style={color=black},
]
\path [draw=blue, fill=blue, opacity=0.2]
(axis cs:-0.5,0.747794152872685)
--(axis cs:-0.5,0.334687058789658)
--(axis cs:-0.494,0.328328630388702)
--(axis cs:-0.488,0.322066939499733)
--(axis cs:-0.482,0.315888969665496)
--(axis cs:-0.476,0.309779958459856)
--(axis cs:-0.47,0.303709704747812)
--(axis cs:-0.464,0.297707025973752)
--(axis cs:-0.458,0.291768751406416)
--(axis cs:-0.452,0.285890970831762)
--(axis cs:-0.446,0.280095639749488)
--(axis cs:-0.44,0.274368839596681)
--(axis cs:-0.434,0.268714985370825)
--(axis cs:-0.428,0.263170521162705)
--(axis cs:-0.422,0.257722235873635)
--(axis cs:-0.416,0.252346309971904)
--(axis cs:-0.41,0.247071805037068)
--(axis cs:-0.404,0.24189105006551)
--(axis cs:-0.398,0.236795224295432)
--(axis cs:-0.392,0.231809772732168)
--(axis cs:-0.386,0.226938760223145)
--(axis cs:-0.38,0.222177968545758)
--(axis cs:-0.374,0.217478898969113)
--(axis cs:-0.368,0.212833054797962)
--(axis cs:-0.362,0.208222700034986)
--(axis cs:-0.356,0.203672285806844)
--(axis cs:-0.35,0.199217553503576)
--(axis cs:-0.344,0.194874962123773)
--(axis cs:-0.338,0.190621745642495)
--(axis cs:-0.332,0.186484024030273)
--(axis cs:-0.326,0.182442396487117)
--(axis cs:-0.32,0.178439980611885)
--(axis cs:-0.314,0.174459612774897)
--(axis cs:-0.308,0.170535312864823)
--(axis cs:-0.302,0.166679665846439)
--(axis cs:-0.296,0.162929860426123)
--(axis cs:-0.29,0.159256012035795)
--(axis cs:-0.284,0.155659508190475)
--(axis cs:-0.278,0.152085640728543)
--(axis cs:-0.272,0.148571302029114)
--(axis cs:-0.266,0.145150309043014)
--(axis cs:-0.26,0.141806571252765)
--(axis cs:-0.254,0.138535099936195)
--(axis cs:-0.248,0.135323853867508)
--(axis cs:-0.242,0.13217241360456)
--(axis cs:-0.236,0.129146722984013)
--(axis cs:-0.23,0.1262541656293)
--(axis cs:-0.224,0.123465970352776)
--(axis cs:-0.218,0.120767341767425)
--(axis cs:-0.212,0.1181849288118)
--(axis cs:-0.206,0.115686779556592)
--(axis cs:-0.2,0.113266711589762)
--(axis cs:-0.194,0.110902553841132)
--(axis cs:-0.188,0.108621595086968)
--(axis cs:-0.182,0.106392653316217)
--(axis cs:-0.176,0.104227988071427)
--(axis cs:-0.17,0.102137708225359)
--(axis cs:-0.164,0.100114747537204)
--(axis cs:-0.158,0.0981720395584493)
--(axis cs:-0.152,0.0962901844031116)
--(axis cs:-0.146,0.0944816575309244)
--(axis cs:-0.14,0.0927182727132476)
--(axis cs:-0.134,0.0910139731393519)
--(axis cs:-0.128,0.0893628068210925)
--(axis cs:-0.122,0.087772178344877)
--(axis cs:-0.116,0.0858485513140916)
--(axis cs:-0.11,0.0835109198346497)
--(axis cs:-0.104,0.081251713004486)
--(axis cs:-0.0979999999999996,0.0790933115023677)
--(axis cs:-0.0919999999999996,0.0770292268163523)
--(axis cs:-0.0859999999999996,0.0750827363628538)
--(axis cs:-0.0799999999999996,0.0732285968148738)
--(axis cs:-0.0739999999999996,0.071474643127749)
--(axis cs:-0.0679999999999996,0.0698088809028463)
--(axis cs:-0.0619999999999996,0.0682159262205232)
--(axis cs:-0.0559999999999996,0.0666972194747601)
--(axis cs:-0.0499999999999996,0.0652560060403632)
--(axis cs:-0.0439999999999996,0.0638654225870733)
--(axis cs:-0.0379999999999996,0.0625284772430371)
--(axis cs:-0.0319999999999996,0.0612813788737038)
--(axis cs:-0.0259999999999996,0.0601040114469443)
--(axis cs:-0.0199999999999996,0.0590085616933061)
--(axis cs:-0.0139999999999996,0.0579854437730051)
--(axis cs:-0.00799999999999956,0.0570185157161861)
--(axis cs:-0.00199999999999956,0.056114765471366)
--(axis cs:0,0.0558226736157211)
--(axis cs:0.006,0.0549804714846213)
--(axis cs:0.012,0.0541928047212347)
--(axis cs:0.018,0.053463688214548)
--(axis cs:0.024,0.0527837799696431)
--(axis cs:0.03,0.0521596181768078)
--(axis cs:0.036,0.0515673036818842)
--(axis cs:0.042,0.0510157223225362)
--(axis cs:0.048,0.0505037938249877)
--(axis cs:0.054,0.0500311452663708)
--(axis cs:0.06,0.04959577541332)
--(axis cs:0.066,0.0492112897084371)
--(axis cs:0.072,0.0488511734024352)
--(axis cs:0.078,0.0485164886612625)
--(axis cs:0.084,0.0482248654728774)
--(axis cs:0.09,0.0479791400993085)
--(axis cs:0.096,0.0477832850384103)
--(axis cs:0.102,0.0476357312789133)
--(axis cs:0.108,0.0475443107804459)
--(axis cs:0.114,0.0474902515278302)
--(axis cs:0.12,0.0474671294142694)
--(axis cs:0.126,0.0474965971946861)
--(axis cs:0.132,0.0475479943830084)
--(axis cs:0.138,0.04762235505241)
--(axis cs:0.144,0.0477151367358097)
--(axis cs:0.15,0.0478304607165161)
--(axis cs:0.156,0.0479907747249062)
--(axis cs:0.162,0.0479489126312274)
--(axis cs:0.168,0.0474495739305105)
--(axis cs:0.174,0.0469696349902815)
--(axis cs:0.18,0.0465158404552017)
--(axis cs:0.186,0.0461050922419357)
--(axis cs:0.186163783073425,0.0460942847688326)
--(axis cs:0.192,0.0457223560949505)
--(axis cs:0.198,0.0453635714774254)
--(axis cs:0.204,0.0450479951308136)
--(axis cs:0.21,0.0447581165602866)
--(axis cs:0.216,0.0444979514905777)
--(axis cs:0.222,0.044259666259058)
--(axis cs:0.228,0.0440576897583642)
--(axis cs:0.234,0.0438862584584192)
--(axis cs:0.24,0.0437495016497311)
--(axis cs:0.246,0.0436498709201252)
--(axis cs:0.252,0.0435724176543704)
--(axis cs:0.258,0.0435206213706656)
--(axis cs:0.264,0.0435072727687262)
--(axis cs:0.27,0.0435381041945031)
--(axis cs:0.276,0.0435924336589202)
--(axis cs:0.282,0.0436800985498027)
--(axis cs:0.288,0.0438165660483736)
--(axis cs:0.294,0.0439860382154857)
--(axis cs:0.3,0.0441732797517314)
--(axis cs:0.306,0.0443646468798842)
--(axis cs:0.312,0.0445683861586876)
--(axis cs:0.318,0.0447898547420853)
--(axis cs:0.324,0.0450439037602526)
--(axis cs:0.33,0.0453282179462158)
--(axis cs:0.336,0.0456343378466531)
--(axis cs:0.342,0.0459800287021337)
--(axis cs:0.348,0.0463543426833337)
--(axis cs:0.354,0.0467680288070071)
--(axis cs:0.36,0.0472207187480618)
--(axis cs:0.366,0.047717026069527)
--(axis cs:0.372,0.0482496309146394)
--(axis cs:0.378,0.0488498326460838)
--(axis cs:0.384,0.0494956855238304)
--(axis cs:0.39,0.050180305225425)
--(axis cs:0.396,0.0509184699224225)
--(axis cs:0.402,0.051708993854374)
--(axis cs:0.408,0.0525440673288018)
--(axis cs:0.414,0.0534327088645079)
--(axis cs:0.42,0.0543835090913753)
--(axis cs:0.426,0.0554029628541097)
--(axis cs:0.432,0.0564785968255559)
--(axis cs:0.438,0.0576350177410814)
--(axis cs:0.444,0.0588525714447639)
--(axis cs:0.45,0.0601437122497477)
--(axis cs:0.456,0.0615122235303756)
--(axis cs:0.462,0.0629841297741804)
--(axis cs:0.468,0.0645473931661349)
--(axis cs:0.474,0.0662023959462341)
--(axis cs:0.48,0.0679489438970142)
--(axis cs:0.486,0.0697925504103969)
--(axis cs:0.492,0.0717483906196565)
--(axis cs:0.492,0.326116884539129)
--(axis cs:0.492,0.326116884539129)
--(axis cs:0.486,0.32099914467554)
--(axis cs:0.48,0.315972010006974)
--(axis cs:0.474,0.311062925644666)
--(axis cs:0.468,0.306278070229207)
--(axis cs:0.462,0.301609137078954)
--(axis cs:0.456,0.297062221713701)
--(axis cs:0.45,0.292582135622606)
--(axis cs:0.444,0.288173662933421)
--(axis cs:0.438,0.283827382956376)
--(axis cs:0.432,0.279578930880254)
--(axis cs:0.426,0.275458802307249)
--(axis cs:0.42,0.271423848446173)
--(axis cs:0.414,0.267484443143425)
--(axis cs:0.408,0.263639011728242)
--(axis cs:0.402,0.25983218195775)
--(axis cs:0.396,0.256073528486544)
--(axis cs:0.39,0.252389739937579)
--(axis cs:0.384,0.24880682891223)
--(axis cs:0.378,0.245330150602359)
--(axis cs:0.372,0.241969088307365)
--(axis cs:0.366,0.238684907385211)
--(axis cs:0.36,0.235457073479491)
--(axis cs:0.354,0.232296092863361)
--(axis cs:0.348,0.229206992220578)
--(axis cs:0.342,0.226217224982586)
--(axis cs:0.336,0.22332515852431)
--(axis cs:0.33,0.220515184126781)
--(axis cs:0.324,0.217757648209405)
--(axis cs:0.318,0.215045932437276)
--(axis cs:0.312,0.212372748852401)
--(axis cs:0.306,0.209749388581562)
--(axis cs:0.3,0.20715345848508)
--(axis cs:0.294,0.204598684602791)
--(axis cs:0.288,0.202101119147586)
--(axis cs:0.282,0.199666359994971)
--(axis cs:0.276,0.197284696302153)
--(axis cs:0.27,0.19495405031769)
--(axis cs:0.264,0.192677229097652)
--(axis cs:0.258,0.190463221949613)
--(axis cs:0.252,0.188303688100743)
--(axis cs:0.246,0.186211943703003)
--(axis cs:0.24,0.184147474081473)
--(axis cs:0.234,0.182096303599859)
--(axis cs:0.228,0.18007073575894)
--(axis cs:0.222,0.17809383681993)
--(axis cs:0.216,0.176125057277161)
--(axis cs:0.21,0.174183079864211)
--(axis cs:0.204,0.172248737101933)
--(axis cs:0.198,0.170325600679579)
--(axis cs:0.192,0.168428361213148)
--(axis cs:0.186163783073425,0.166637000394107)
--(axis cs:0.186,0.166587407729412)
--(axis cs:0.18,0.164775634824973)
--(axis cs:0.174,0.163006746739124)
--(axis cs:0.168,0.161255845213994)
--(axis cs:0.162,0.15953955379968)
--(axis cs:0.156,0.157841348357941)
--(axis cs:0.15,0.15616081942591)
--(axis cs:0.144,0.154491620826732)
--(axis cs:0.138,0.152834384345496)
--(axis cs:0.132,0.151215631243595)
--(axis cs:0.126,0.149643151037094)
--(axis cs:0.12,0.149476523065021)
--(axis cs:0.114,0.150131440716408)
--(axis cs:0.108,0.150843446541645)
--(axis cs:0.102,0.151607363571088)
--(axis cs:0.096,0.15247160209155)
--(axis cs:0.09,0.153425374031736)
--(axis cs:0.084,0.15443314420761)
--(axis cs:0.078,0.155538938192246)
--(axis cs:0.072,0.156751865811929)
--(axis cs:0.066,0.158080014774895)
--(axis cs:0.06,0.159486779466671)
--(axis cs:0.054,0.160981989460897)
--(axis cs:0.048,0.162576624232356)
--(axis cs:0.042,0.164287394426597)
--(axis cs:0.036,0.166111423337624)
--(axis cs:0.03,0.168009340648445)
--(axis cs:0.024,0.169997059234446)
--(axis cs:0.018,0.172071762173906)
--(axis cs:0.012,0.174274947288954)
--(axis cs:0.006,0.176580543753396)
--(axis cs:0,0.178987413044365)
--(axis cs:-0.00199999999999956,0.179817919983283)
--(axis cs:-0.00799999999999956,0.1823559865379)
--(axis cs:-0.0139999999999996,0.184999009193079)
--(axis cs:-0.0199999999999996,0.187772060870729)
--(axis cs:-0.0259999999999996,0.190654178094992)
--(axis cs:-0.0319999999999996,0.193630440728698)
--(axis cs:-0.0379999999999996,0.196757124855848)
--(axis cs:-0.0439999999999996,0.199999850825317)
--(axis cs:-0.0499999999999996,0.203360949634753)
--(axis cs:-0.0559999999999996,0.206822665710983)
--(axis cs:-0.0619999999999996,0.210396265124906)
--(axis cs:-0.0679999999999996,0.214072634671953)
--(axis cs:-0.0739999999999996,0.217823409055609)
--(axis cs:-0.0799999999999996,0.221623784004323)
--(axis cs:-0.0859999999999996,0.225506926274137)
--(axis cs:-0.0919999999999996,0.229513012357996)
--(axis cs:-0.0979999999999996,0.233597894715774)
--(axis cs:-0.104,0.237722794317825)
--(axis cs:-0.11,0.241953672626724)
--(axis cs:-0.116,0.246271108509158)
--(axis cs:-0.122,0.250676763019482)
--(axis cs:-0.128,0.255220284760046)
--(axis cs:-0.134,0.259812328462058)
--(axis cs:-0.14,0.264483023250463)
--(axis cs:-0.146,0.269200201208292)
--(axis cs:-0.152,0.273985183750717)
--(axis cs:-0.158,0.278826863321186)
--(axis cs:-0.164,0.283716466096015)
--(axis cs:-0.17,0.288694554850281)
--(axis cs:-0.176,0.29374164085178)
--(axis cs:-0.182,0.298843162412814)
--(axis cs:-0.188,0.304038782604587)
--(axis cs:-0.194,0.309327451313606)
--(axis cs:-0.2,0.314702625638237)
--(axis cs:-0.206,0.320163852840414)
--(axis cs:-0.212,0.325710382810136)
--(axis cs:-0.218,0.331334693569217)
--(axis cs:-0.224,0.337067967132676)
--(axis cs:-0.23,0.342935772188163)
--(axis cs:-0.236,0.348884841718592)
--(axis cs:-0.242,0.354966858839735)
--(axis cs:-0.248,0.361111053988402)
--(axis cs:-0.254,0.367344321607242)
--(axis cs:-0.26,0.373677921149274)
--(axis cs:-0.266,0.380059229482403)
--(axis cs:-0.272,0.386439588086139)
--(axis cs:-0.278,0.392902172016325)
--(axis cs:-0.284,0.399416434367506)
--(axis cs:-0.29,0.405971662336743)
--(axis cs:-0.296,0.412576671355435)
--(axis cs:-0.302,0.419261664368806)
--(axis cs:-0.308,0.425962839611815)
--(axis cs:-0.314,0.432679443878799)
--(axis cs:-0.32,0.439433447644626)
--(axis cs:-0.326,0.446263512965575)
--(axis cs:-0.332,0.453195754332179)
--(axis cs:-0.338,0.460199789173154)
--(axis cs:-0.344,0.46728987085356)
--(axis cs:-0.35,0.474451296170106)
--(axis cs:-0.356,0.481664655721898)
--(axis cs:-0.362,0.488935834713283)
--(axis cs:-0.368,0.496197959034043)
--(axis cs:-0.374,0.503496633980422)
--(axis cs:-0.38,0.510839681575328)
--(axis cs:-0.386,0.520803310827823)
--(axis cs:-0.392,0.532096901207553)
--(axis cs:-0.398,0.543508896907266)
--(axis cs:-0.404,0.554994038711396)
--(axis cs:-0.41,0.566513539079203)
--(axis cs:-0.416,0.578190827356096)
--(axis cs:-0.422,0.589926219419251)
--(axis cs:-0.428,0.601715446385082)
--(axis cs:-0.434,0.613569856112954)
--(axis cs:-0.44,0.62545851428256)
--(axis cs:-0.446,0.637350641856255)
--(axis cs:-0.452,0.649337965417883)
--(axis cs:-0.458,0.661373910662098)
--(axis cs:-0.464,0.67350283365592)
--(axis cs:-0.47,0.685657027800668)
--(axis cs:-0.476,0.697838376896333)
--(axis cs:-0.482,0.710113426603557)
--(axis cs:-0.488,0.722548585152363)
--(axis cs:-0.494,0.735109491818007)
--(axis cs:-0.5,0.747794152872685)
--cycle;

\addplot [semithick, black]
table {%
-0.5 0.432449021268146
-0.494 0.423549716079297
-0.488 0.414923930157567
-0.482 0.406496713687488
-0.476 0.398238391152988
-0.47 0.390476662030588
-0.464 0.382947887366993
-0.458 0.375594145476467
-0.452 0.36845312018034
-0.446 0.361517617405464
-0.44 0.354336396049027
-0.434 0.347283278437468
-0.428 0.340350977697653
-0.422 0.333431482816787
-0.416 0.326399658678619
-0.41 0.319528929907898
-0.404 0.312792147865372
-0.398 0.306208187702166
-0.392 0.299715407568767
-0.386 0.293328990451014
-0.38 0.287079005121806
-0.374 0.280971827445643
-0.368 0.274986982957359
-0.362 0.269136483510631
-0.356 0.263711467755838
-0.35 0.258236470250346
-0.344 0.252373612958843
-0.338 0.246612996971284
-0.332 0.240959565674973
-0.326 0.236069766582802
-0.32 0.231459002439912
-0.314 0.226933404553376
-0.308 0.222299679101287
-0.302 0.217860337550019
-0.296 0.213497753106523
-0.29 0.209136665890513
-0.284 0.204676357747514
-0.278 0.200074223229169
-0.272 0.19545587774172
-0.266 0.190910683750263
-0.26 0.186470249686316
-0.254 0.182111950463011
-0.248 0.178174832457846
-0.242 0.174346968882349
-0.236 0.170706252596695
-0.23 0.167126488983402
-0.224 0.163006037729315
-0.218 0.159728623118027
-0.212 0.156197124140409
-0.206 0.152813161053108
-0.2 0.149524243648969
-0.194 0.146335248170317
-0.188 0.14339352615023
-0.182 0.140593288222072
-0.176 0.138523830351216
-0.17 0.136915180455066
-0.164 0.135546531713201
-0.158 0.13432170017651
-0.152 0.133161697052559
-0.146 0.132077202490155
-0.14 0.131041088613926
-0.134 0.130064877610838
-0.128 0.129040384065994
-0.122 0.127602335581679
-0.116 0.125986439857213
-0.11 0.123827200294384
-0.104 0.121423438424368
-0.0979999999999996 0.119098359487932
-0.0919999999999996 0.116851172069912
-0.0859999999999996 0.114690718567329
-0.0799999999999996 0.112354297549142
-0.0739999999999996 0.109339031919071
-0.0679999999999996 0.106415280237649
-0.0619999999999996 0.103578456816553
-0.0559999999999996 0.101088875691257
-0.0499999999999996 0.0987660851925496
-0.0439999999999996 0.0969382617583613
-0.0379999999999996 0.0956909523876858
-0.0319999999999996 0.0945368829088482
-0.0259999999999996 0.0928426157631379
-0.0199999999999996 0.0911044162412514
-0.0139999999999996 0.0895517639945111
-0.00799999999999956 0.0880896667118783
-0.00199999999999956 0.0867344687500556
0 0.0863060543997968
0.006 0.0850870045765425
0.012 0.0809997639955411
0.018 0.0796227379457523
0.024 0.0789091832996337
0.03 0.0782527859234531
0.036 0.0776535007470757
0.042 0.0771148429660406
0.048 0.0766344536743976
0.054 0.076203507708658
0.06 0.0758147562058945
0.066 0.0754690992774535
0.072 0.0751639948665838
0.078 0.0749123597400108
0.084 0.0747078507141864
0.09 0.0745573724827305
0.096 0.0744644245586501
0.102 0.0744205184855764
0.108 0.0744012631489847
0.114 0.0743147506395605
0.12 0.0744669510910587
0.126 0.0744689356927131
0.132 0.0745238387523953
0.138 0.0746191342225497
0.144 0.0747341159446797
0.15 0.0748804337848316
0.156 0.0750709811533085
0.162 0.0753004594565083
0.168 0.0755711435855554
0.174 0.0758642224250686
0.18 0.0761790668057872
0.186 0.07652728388549
0.186163783073425 0.0765373623095373
0.192 0.0769152410686606
0.198 0.0773449024185097
0.204 0.0778088052175735
0.21 0.0783038367357073
0.216 0.0788259164337483
0.222 0.079395438465484
0.228 0.0800060526057662
0.234 0.0806289143117021
0.24 0.0812805248492235
0.246 0.0819647528128259
0.252 0.0826886605807723
0.258 0.0834416594346819
0.264 0.0842276671623992
0.27 0.0850547711236945
0.276 0.0859263546370077
0.282 0.086839045944743
0.288 0.0877976021381943
0.294 0.0887956222549919
0.3 0.0898191119052335
0.306 0.0909893476230671
0.312 0.0925937116687934
0.318 0.0936397695658398
0.324 0.0947221406271942
0.33 0.0958581587418157
0.336 0.0970495100203301
0.342 0.0983081874283879
0.348 0.0997022956090175
0.354 0.101144314917444
0.36 0.102613931330369
0.366 0.104133853985912
0.372 0.105724328904258
0.378 0.107380256280701
0.384 0.109217815564689
0.39 0.111302873899013
0.396 0.11348895785455
0.402 0.115764023157618
0.408 0.118131749372802
0.414 0.120586369406833
0.42 0.123319657396802
0.426 0.125880446356378
0.432 0.128520701893269
0.438 0.131319183521776
0.444 0.134745659589941
0.45 0.137146870577767
0.456 0.138901017903528
0.462 0.141308759248397
0.468 0.143799563234125
0.474 0.146354180084918
0.48 0.149016048595469
0.486 0.151776615191792
0.492 0.154874954817324
};
\addplot [semithick, black]
table {%
-0.5 0.505915780603592
-0.494 0.496645427923109
-0.488 0.487443326613912
-0.482 0.47836817279948
-0.476 0.469405086906526
-0.47 0.46059313962698
-0.464 0.451919545475302
-0.458 0.443371049257454
-0.452 0.434873890536653
-0.446 0.426485854070782
-0.44 0.418207619984458
-0.434 0.410029555683089
-0.428 0.401698945074098
-0.422 0.39325557222217
-0.416 0.384937717088428
-0.41 0.376744295085668
-0.404 0.368675061064732
-0.398 0.360739076934749
-0.392 0.35374783235723
-0.386 0.347129663551508
-0.38 0.340621973288482
-0.374 0.334232779854848
-0.368 0.327944235582524
-0.362 0.321401680759778
-0.356 0.314649380697212
-0.35 0.307702724035432
-0.344 0.300904588464386
-0.338 0.294251812320058
-0.332 0.287739899860681
-0.326 0.281358651605043
-0.32 0.275122084815312
-0.314 0.269062139669589
-0.308 0.263180238961538
-0.302 0.257463877378354
-0.296 0.251885385717697
-0.29 0.246428963004415
-0.284 0.241134512986905
-0.278 0.235988579559937
-0.272 0.230990227438579
-0.266 0.226139646301801
-0.26 0.221431912910743
-0.254 0.216684516022942
-0.248 0.211837534415049
-0.242 0.20712000148506
-0.236 0.20257391946618
-0.23 0.198146777641698
-0.224 0.193863411721168
-0.218 0.189708030129773
-0.212 0.185671720489499
-0.206 0.181761317679456
-0.2 0.17796769423969
-0.194 0.174278784390514
-0.188 0.170705370767788
-0.182 0.167824019600377
-0.176 0.165087933519139
-0.17 0.161894739788066
-0.164 0.158214738990748
-0.158 0.154723345631901
-0.152 0.152247740828094
-0.146 0.149817186404097
-0.14 0.147444415294389
-0.134 0.145073212628272
-0.128 0.141919859541559
-0.122 0.13897776354144
-0.116 0.136220764959387
-0.11 0.133234237417836
-0.104 0.130180677956236
-0.0979999999999996 0.127403986026528
-0.0919999999999996 0.125773012478782
-0.0859999999999996 0.124492059932897
-0.0799999999999996 0.123712448461445
-0.0739999999999996 0.123432130938971
-0.0679999999999996 0.123214686950993
-0.0619999999999996 0.122288936627583
-0.0559999999999996 0.121264138151559
-0.0499999999999996 0.120276991576263
-0.0439999999999996 0.118811101448294
-0.0379999999999996 0.117403408214205
-0.0319999999999996 0.116050810063779
-0.0259999999999996 0.114760409814093
-0.0199999999999996 0.113520283677178
-0.0139999999999996 0.11157182930758
-0.00799999999999956 0.109710690002917
-0.00199999999999956 0.105353433126124
0 0.104988572823518
0.006 0.103932603394882
0.012 0.103009264704906
0.018 0.102380661537847
0.024 0.101798103392257
0.03 0.100995368020909
0.036 0.10006465506554
0.042 0.0991799771470427
0.048 0.0985757433063404
0.054 0.0982171300860699
0.06 0.0978974017698394
0.066 0.0976195045666622
0.072 0.0974013623028398
0.078 0.0972233534121352
0.084 0.0970844788148173
0.09 0.0970006054657056
0.096 0.0969834690713247
0.102 0.0970002550648874
0.108 0.0972754983549286
0.114 0.0975829219186028
0.12 0.0977213763704106
0.126 0.0978727647876984
0.132 0.0979284890405317
0.138 0.0981166810586125
0.144 0.0979481759078243
0.15 0.0978146632787394
0.156 0.0977794206400157
0.162 0.0980470156496351
0.168 0.0983610731462446
0.174 0.098727682950836
0.18 0.099149382980085
0.186 0.099623367915653
0.186163783073425 0.0996368316556141
0.192 0.100139078753812
0.198 0.100714133460145
0.204 0.101210896084572
0.21 0.10161334434973
0.216 0.102068068294662
0.222 0.102584159673695
0.228 0.103151680879903
0.234 0.103763774475574
0.24 0.104409104550384
0.246 0.105105462843071
0.252 0.105888035900052
0.258 0.106756213736401
0.264 0.108052457421184
0.27 0.109611069279278
0.276 0.111224246191934
0.282 0.112872859683919
0.288 0.114538437178705
0.294 0.115843331361895
0.3 0.117206136825093
0.306 0.118370083924991
0.312 0.119300906623678
0.318 0.120282759071816
0.324 0.121330700073042
0.33 0.122861346262397
0.336 0.124670036764329
0.342 0.126560533817607
0.348 0.128525781023749
0.354 0.130557165735688
0.36 0.132678159608267
0.366 0.134886470943057
0.372 0.137180183483822
0.378 0.139552272242704
0.384 0.142001547815696
0.39 0.144515566843302
0.396 0.14711149435281
0.402 0.149782968880695
0.408 0.152527821616671
0.414 0.155920892222032
0.42 0.159544864332915
0.426 0.163210097328656
0.432 0.166954209129061
0.438 0.170768297567048
0.444 0.174090046222943
0.45 0.177336899919072
0.456 0.180658010888813
0.462 0.184018099159363
0.468 0.187660449781081
0.474 0.191877633665071
0.48 0.196173174220537
0.486 0.200074478382734
0.492 0.203540267001667
};
\addplot [semithick, black]
table {%
-0.5 0.599569076160312
-0.494 0.589045516754644
-0.488 0.578597436159068
-0.482 0.56817432632347
-0.476 0.557764485458553
-0.47 0.547451095719738
-0.464 0.537171267726732
-0.458 0.526720995768603
-0.452 0.516356840263559
-0.446 0.506153528158222
-0.44 0.496059844719814
-0.434 0.486045058835801
-0.428 0.476174393144877
-0.422 0.466457376918224
-0.416 0.45688370688038
-0.41 0.447402962667007
-0.404 0.438031300872476
-0.398 0.42873124933633
-0.392 0.419548380825312
-0.386 0.410498481887404
-0.38 0.401569237427624
-0.374 0.392792661436346
-0.368 0.38418441399798
-0.362 0.375910111030811
-0.356 0.368717568987281
-0.35 0.361653196453447
-0.344 0.354653658760184
-0.338 0.347695325921803
-0.332 0.340790879124113
-0.326 0.333944975955026
-0.32 0.327206218924065
-0.314 0.320573567495597
-0.308 0.314019161971769
-0.302 0.307537177867146
-0.296 0.30112490731895
-0.29 0.294778584423282
-0.284 0.28850795860192
-0.278 0.282326047988314
-0.272 0.276252344644174
-0.266 0.270249268591839
-0.26 0.264323529724418
-0.254 0.258515126550183
-0.248 0.252792192183297
-0.242 0.247127135315573
-0.236 0.241500755266982
-0.23 0.23595261970894
-0.224 0.230495669265903
-0.218 0.22517837573207
-0.212 0.221079710915506
-0.206 0.217536671548842
-0.2 0.21406672691733
-0.194 0.210694993992119
-0.188 0.207424275236206
-0.182 0.204228525082497
-0.176 0.201106696764804
-0.17 0.197915514202125
-0.164 0.194633137335389
-0.158 0.191339236513362
-0.152 0.188101925642782
-0.146 0.184975937825613
-0.14 0.18191082809704
-0.134 0.17882582774905
-0.128 0.17484238964647
-0.122 0.170997989961558
-0.116 0.166771235306762
-0.11 0.162718827381943
-0.104 0.15883762217795
-0.0979999999999996 0.155137168504251
-0.0919999999999996 0.151594757104111
-0.0859999999999996 0.148181930000184
-0.0799999999999996 0.144913184870194
-0.0739999999999996 0.14181278852154
-0.0679999999999996 0.138880044036474
-0.0619999999999996 0.136076456252965
-0.0559999999999996 0.133104983278824
-0.0499999999999996 0.130999869438501
-0.0439999999999996 0.129295305928081
-0.0379999999999996 0.127530439768277
-0.0319999999999996 0.125856817251361
-0.0259999999999996 0.125769584102513
-0.0199999999999996 0.125846722767141
-0.0139999999999996 0.125980354218613
-0.00799999999999956 0.12286646812669
-0.00199999999999956 0.122982810128399
0 0.123006145282339
0.006 0.123087968514463
0.012 0.123100302491094
0.018 0.123077275674938
0.024 0.121778556704547
0.03 0.120331053112428
0.036 0.118968592233566
0.042 0.117679554371261
0.048 0.116455463059279
0.054 0.115298216842444
0.06 0.114810063052768
0.066 0.114681697963997
0.072 0.114629918597887
0.078 0.114642809050798
0.084 0.114708215754143
0.09 0.11479428399899
0.096 0.114916636520132
0.102 0.115084139102085
0.108 0.115310442389896
0.114 0.115561738485231
0.12 0.115795471054505
0.126 0.115935922285577
0.132 0.116124908286695
0.138 0.116358323919389
0.144 0.11625467855227
0.15 0.116157751705738
0.156 0.116174041831293
0.162 0.116631613327539
0.168 0.11710879568811
0.174 0.117616070420673
0.18 0.118159989363423
0.186 0.118738972003174
0.186163783073425 0.118754830512098
0.192 0.119345732713141
0.198 0.120006266274653
0.204 0.120697479410756
0.21 0.12141213372504
0.216 0.122158001856195
0.222 0.122928856226498
0.228 0.123742083784327
0.234 0.124579894954984
0.24 0.125466685453463
0.246 0.126388304371304
0.252 0.127351049434494
0.258 0.128361522948593
0.264 0.129421654450615
0.27 0.130519185846002
0.276 0.131646722599387
0.282 0.132839531376498
0.288 0.134066116001159
0.294 0.135339295102288
0.3 0.136644262591653
0.306 0.137983880205509
0.312 0.139354330275822
0.318 0.141762563195942
0.324 0.144318238613033
0.33 0.146968998346838
0.336 0.149732394885022
0.342 0.152595140068109
0.348 0.155383135214906
0.354 0.15773814169441
0.36 0.160731263482238
0.366 0.164240977413613
0.372 0.16788866194217
0.378 0.171656039019842
0.384 0.175530307680643
0.39 0.179345068132818
0.396 0.183132494710976
0.402 0.187040674504588
0.408 0.191098862133453
0.414 0.194925699115057
0.42 0.198463395461217
0.426 0.20198589095645
0.432 0.205491148643477
0.438 0.209101259117318
0.444 0.212848043026819
0.45 0.216704221297191
0.456 0.220671199675101
0.462 0.22473110265309
0.468 0.228862669744898
0.474 0.233059136898365
0.48 0.237321023868322
0.486 0.241670181733782
0.492 0.246096483868029
};
\addplot [very thick, green!50.1960784313725!black]
table {%
-0.5 1.81654831450081
-0.494 1.79488425156851
-0.488 1.7731796398646
-0.482 1.75147033396886
-0.476 1.72969133409913
-0.47 1.70788456499873
-0.464 1.68608897148524
-0.458 1.66429973994609
-0.452 1.64244989085724
-0.446 1.62061107936356
-0.44 1.59869818699797
-0.434 1.57669312184844
-0.428 1.5546129997104
-0.422 1.53257707818481
-0.416 1.51055648010923
-0.41 1.48852027259376
-0.404 1.46649670990183
-0.398 1.44457439245043
-0.392 1.42270379354088
-0.386 1.40087550198812
-0.38 1.37904614877082
-0.374 1.35721579681333
-0.368 1.3354167273163
-0.362 1.3136386478351
-0.356 1.29175198286699
-0.35 1.26985584659199
-0.344 1.24791985333463
-0.338 1.22595699620722
-0.332 1.20390796064827
-0.326 1.181835434016
-0.32 1.15984119565212
-0.314 1.13778013184879
-0.308 1.11563029777986
-0.302 1.09348606351092
-0.296 1.07136029823302
-0.29 1.04922619989899
-0.284 1.02705899210973
-0.278 1.00486776706456
-0.272 0.982727865213672
-0.266 0.960498532611787
-0.26 0.938275161769539
-0.254 0.91609942169034
-0.248 0.893908888004471
-0.242 0.871822278129531
-0.236 0.849855426166158
-0.23 0.827890054523123
-0.224 0.805973489183539
-0.218 0.784086180341838
-0.212 0.762275958178875
-0.206 0.740436749340304
-0.2 0.718632917682814
-0.194 0.696888207435066
-0.188 0.675308650937165
-0.182 0.653773049105965
-0.176 0.632151303241551
-0.17 0.610545276873683
-0.164 0.589082747986287
-0.158 0.567724575316985
-0.152 0.54650121677647
-0.146 0.525425866610519
-0.14 0.50453144407947
-0.134 0.483865181782619
-0.128 0.463419296934342
-0.122 0.443177742113283
-0.116 0.423124104731073
-0.11 0.403301606622971
-0.104 0.383695837933525
-0.0979999999999996 0.364286863676289
-0.0919999999999996 0.34519061612772
-0.0859999999999996 0.326362813850054
-0.0799999999999996 0.307824426528537
-0.0739999999999996 0.289574604109478
-0.0679999999999996 0.271581048173554
-0.0619999999999996 0.253876689326804
-0.0559999999999996 0.236514165860893
-0.0499999999999996 0.219532063293576
-0.0439999999999996 0.202917852900791
-0.0379999999999996 0.186783207498474
-0.0319999999999996 0.171134624785145
-0.0259999999999996 0.155990058303359
-0.0199999999999996 0.141504981216179
-0.0139999999999996 0.127724640070521
-0.00799999999999956 0.114779911623668
-0.00199999999999956 0.102774616731623
0 0.0990003621658538
0.006 0.0883971110516545
0.012 0.0788439912941782
0.018 0.0703936979436309
0.024 0.0630186698145668
0.03 0.0566351623982307
0.036 0.0511362405260777
0.042 0.0464071644678236
0.048 0.0423462300885252
0.054 0.0388657820360753
0.06 0.035885158646585
0.066 0.0333300780627671
0.072 0.031116660033441
0.078 0.0291902899125168
0.084 0.0275026214605417
0.09 0.0260107465118437
0.096 0.0246964709803465
0.102 0.0235319861985396
0.108 0.0224927501765177
0.114 0.0215629756950048
0.12 0.0207264222808836
0.126 0.0199712144493273
0.132 0.019297925976504
0.138 0.0186956545772269
0.144 0.0181479834878455
0.15 0.0176516585654506
0.156 0.0172066148810212
0.162 0.0168065545803972
0.168 0.0164498268042432
0.174 0.016130572295572
0.18 0.0158492345822362
0.186 0.0156024149608071
0.186163783073425 0.0155961301890534
0.192 0.0153854267383202
0.198 0.0152024247530462
0.204 0.0150480207137328
0.21 0.0149232920605031
0.216 0.0148290667297403
0.222 0.0147602522152092
0.228 0.0147119661931378
0.234 0.0146909470515372
0.24 0.0146979703597611
0.246 0.0147325956074328
0.252 0.0147940872990967
0.258 0.0148800031255641
0.264 0.0149875768006481
0.27 0.0151178357979503
0.276 0.0152754657778388
0.282 0.0154655457064784
0.288 0.0156808878123046
0.294 0.0159234998100496
0.3 0.0161964128523108
0.306 0.0164963093690321
0.312 0.0168231881519535
0.318 0.0171776088940024
0.324 0.0175692758445652
0.33 0.0179910061603973
0.336 0.0184460651226249
0.342 0.0189355503769256
0.348 0.0194551332037139
0.354 0.0200102581960139
0.36 0.0206103086016211
0.366 0.0212603176034792
0.372 0.0219480283010504
0.378 0.0226828224946364
0.384 0.0234711864116707
0.39 0.0243042652297535
0.396 0.0251849469743027
0.402 0.0261173641261266
0.408 0.0270949828518486
0.414 0.02812442405156
0.42 0.0292050547219896
0.426 0.0303431132966082
0.432 0.0315434501175974
0.438 0.0328198294868399
0.444 0.0341643166222457
0.45 0.0355771319642678
0.456 0.0370554498703831
0.462 0.0386093269809676
0.468 0.0402347187438997
0.474 0.0419316985302663
0.48 0.0437125624376441
0.486 0.0455867065958538
0.492 0.0475498319335105
};
\addplot [very thick, red]
table {%
-0.5 0.547252369203009
-0.494 0.537674408892596
-0.488 0.528206258218136
-0.482 0.518850080312431
-0.476 0.509599216724226
-0.47 0.500464211770081
-0.464 0.491446500706564
-0.458 0.482544148517505
-0.452 0.473743313398937
-0.446 0.465055119855932
-0.44 0.456475873296001
-0.434 0.447999165022302
-0.428 0.439627587229356
-0.422 0.431376731618016
-0.416 0.423237752073602
-0.41 0.41520525635725
-0.404 0.40727900470135
-0.398 0.399468837990527
-0.392 0.39177640381629
-0.386 0.384205503847177
-0.38 0.376750536752103
-0.374 0.369412069480178
-0.368 0.362178762827818
-0.362 0.355053394105144
-0.356 0.348037106597301
-0.35 0.341128673576669
-0.344 0.334324101846004
-0.338 0.327622831382656
-0.332 0.321029316514733
-0.326 0.314537851221709
-0.32 0.308151066891934
-0.314 0.301861619174406
-0.308 0.295672975593725
-0.302 0.28958508251069
-0.296 0.2835917204406
-0.29 0.277700530059196
-0.284 0.27191424719357
-0.278 0.266224839797765
-0.272 0.260638590415536
-0.266 0.255143740931608
-0.26 0.249750573067088
-0.254 0.24446419091062
-0.248 0.2392794594717
-0.242 0.234194444760046
-0.236 0.22920686467293
-0.23 0.224315071718078
-0.224 0.21952518073516
-0.218 0.214838312549612
-0.212 0.210254454628518
-0.206 0.205757732905868
-0.2 0.201350024964334
-0.194 0.197034226085755
-0.188 0.192826630089917
-0.182 0.188714526590865
-0.176 0.184692553324008
-0.17 0.180761763814458
-0.164 0.176926868866674
-0.158 0.173187561374797
-0.152 0.169537278983161
-0.146 0.165975710595669
-0.14 0.162498288157687
-0.134 0.159107552324107
-0.128 0.155807834010142
-0.122 0.152595071997425
-0.116 0.149475216468474
-0.11 0.146448585069121
-0.104 0.14350877586251
-0.0979999999999996 0.140655423090879
-0.0919999999999996 0.137890474666702
-0.0859999999999996 0.135214446550098
-0.0799999999999996 0.132628162118304
-0.0739999999999996 0.130131787632991
-0.0679999999999996 0.127720829375246
-0.0619999999999996 0.1253957076961
-0.0559999999999996 0.123161027972775
-0.0499999999999996 0.121014027265634
-0.0439999999999996 0.118946019382789
-0.0379999999999996 0.116960887489074
-0.0319999999999996 0.11506584125201
-0.0259999999999996 0.113259297567443
-0.0199999999999996 0.111545137929851
-0.0139999999999996 0.109920134675761
-0.00799999999999956 0.108388356321871
-0.00199999999999956 0.10695684198169
0 0.106502281371484
0.006 0.105205352241637
0.012 0.104003064781447
0.018 0.102898738107897
0.024 0.101890911530345
0.03 0.100969361244007
0.036 0.100130199860444
0.042 0.0993709673214969
0.048 0.0986851655411085
0.054 0.098069366115942
0.06 0.0975161721093666
0.066 0.0970215995901996
0.072 0.0965860243666701
0.078 0.096208459376167
0.084 0.0958864047055384
0.09 0.0956162238523521
0.096 0.0953955294994732
0.102 0.0952269651401583
0.108 0.0951086971419347
0.114 0.0950369319114352
0.12 0.095007889080013
0.126 0.0950206566074796
0.132 0.0950802774211642
0.138 0.09518192314311
0.144 0.0953288734727501
0.15 0.09551908210161
0.156 0.0957528258402917
0.162 0.0960283438952537
0.168 0.0963458784382963
0.174 0.0967047272714569
0.18 0.0971051574042839
0.186 0.0975490976205335
0.186163783073425 0.0975617520146803
0.192 0.0980334916947331
0.198 0.0985627000287245
0.204 0.0991332918234889
0.21 0.0997426231740506
0.216 0.100396689467476
0.222 0.101094102013661
0.228 0.101836430407773
0.234 0.102624408400955
0.24 0.103462723176658
0.246 0.104352469766042
0.252 0.105296318891552
0.258 0.106290148021457
0.264 0.107338250323105
0.27 0.108441775092834
0.276 0.10960088453555
0.282 0.11081751915785
0.288 0.112092344232711
0.294 0.113427500250939
0.3 0.114830174110787
0.306 0.116293147328004
0.312 0.117815693096213
0.318 0.119401072846902
0.324 0.121050766744774
0.33 0.122769333823287
0.336 0.124559910099295
0.342 0.126420425455912
0.348 0.128350797743459
0.354 0.130354793032747
0.36 0.1324321580958
0.366 0.13458382017008
0.372 0.136820832061026
0.378 0.139144352026068
0.384 0.141554727256899
0.39 0.144050159565012
0.396 0.146631990817032
0.402 0.1493017758694
0.408 0.152062920604816
0.414 0.15491496260939
0.42 0.157850800951606
0.426 0.160867453126801
0.432 0.163973782063779
0.438 0.167174989581963
0.444 0.170471522816204
0.45 0.173855980820218
0.456 0.177330892888446
0.462 0.18089484642375
0.468 0.184551082972396
0.474 0.188299801036784
0.48 0.192147036489234
0.486 0.196092861116029
0.492 0.200137139313976
};
\end{axis}

\end{tikzpicture}

%% file: labpal/figure_data/line_plots/CIFAR10_mom0_resnet20_augment_result/line_plots/bs_128_bs_ori_128_sgd_lrs_[0.01]_pal_mus_[0.1]_ri_100_c_apal_0.9_ori_sgd_lr_0.1/pure_line_8000.pgf
% This file was created by tikzplotlib v0.9.6.
\begin{tikzpicture}

\begin{axis}[
yticklabel style={font=\large},
xticklabel style={font=\large},
ytick style={font=\Large},
xtick style={font=\Large},
ylabel style={font=\Large},
xlabel style={font=\Large},
title style={font=\Large},
width=10.5cm,
height=7.2cm,
tick align=outside,
tick pos=left,
title={line number/t: 8,000},
x grid style={white!69.0196078431373!black},
xlabel={step on line (s)},
xmin=-0.5496, xmax=0.5416,
xtick style={color=black},
xtick={-0.6,-0.4,-0.2,0,0.2,0.4,0.6},
y grid style={white!69.0196078431373!black},
y tick label style={/pgf/number format/.cd,scaled y ticks = false,set thousands separator={},fixed,precision=4},
ylabel={loss},
ymin=0.0020955911300681, ymax=0.3,
ytick style={color=black},
]
\path [draw=blue, fill=blue, opacity=0.2]
(axis cs:-0.5,0.408855287179886)
--(axis cs:-0.5,0.0593006076270419)
--(axis cs:-0.494,0.0577057829649172)
--(axis cs:-0.488,0.0561610260881773)
--(axis cs:-0.482,0.0546701486570313)
--(axis cs:-0.476,0.0532213531639484)
--(axis cs:-0.47,0.05182088036962)
--(axis cs:-0.464,0.050465964986877)
--(axis cs:-0.458,0.0491782571350491)
--(axis cs:-0.452,0.0479280728185416)
--(axis cs:-0.446,0.0463145631733928)
--(axis cs:-0.44,0.0432192531510511)
--(axis cs:-0.434,0.0402471950980737)
--(axis cs:-0.428,0.0373982316362805)
--(axis cs:-0.422,0.0346800178049707)
--(axis cs:-0.416,0.0321038183547512)
--(axis cs:-0.41,0.0296674904210902)
--(axis cs:-0.404,0.0273809324442676)
--(axis cs:-0.398,0.0252595337061975)
--(axis cs:-0.392,0.0233130910247716)
--(axis cs:-0.386,0.0215633012039246)
--(axis cs:-0.38,0.0200036937487732)
--(axis cs:-0.374,0.0186321388728318)
--(axis cs:-0.368,0.0174327401660339)
--(axis cs:-0.362,0.0164031129726016)
--(axis cs:-0.356,0.0155278891155381)
--(axis cs:-0.35,0.0147922306571714)
--(axis cs:-0.344,0.0141832915177503)
--(axis cs:-0.338,0.0136641776013223)
--(axis cs:-0.332,0.01321794834897)
--(axis cs:-0.326,0.0128382916560079)
--(axis cs:-0.32,0.0125177103222242)
--(axis cs:-0.314,0.0122413345126786)
--(axis cs:-0.308,0.0120028905654143)
--(axis cs:-0.302,0.011795041353886)
--(axis cs:-0.296,0.0116139430048016)
--(axis cs:-0.29,0.0114514653071646)
--(axis cs:-0.284,0.0113022836769751)
--(axis cs:-0.278,0.0111686833730015)
--(axis cs:-0.272,0.0110504624132288)
--(axis cs:-0.266,0.0109414613861964)
--(axis cs:-0.26,0.010835193179814)
--(axis cs:-0.254,0.0107311599984032)
--(axis cs:-0.248,0.0106324466209404)
--(axis cs:-0.242,0.0105365720435521)
--(axis cs:-0.236,0.0104470320096555)
--(axis cs:-0.23,0.0103626117274019)
--(axis cs:-0.224,0.010281547856001)
--(axis cs:-0.218,0.0102064961053006)
--(axis cs:-0.212,0.0101289925055999)
--(axis cs:-0.206,0.0100526185512649)
--(axis cs:-0.2,0.00998173220725962)
--(axis cs:-0.194,0.00991108094904292)
--(axis cs:-0.188,0.00985094511590645)
--(axis cs:-0.182,0.00979032996725093)
--(axis cs:-0.176,0.00972792992489502)
--(axis cs:-0.17,0.00966831640542665)
--(axis cs:-0.164,0.00961367929540813)
--(axis cs:-0.158,0.00956307486089603)
--(axis cs:-0.152,0.00933879258998277)
--(axis cs:-0.146,0.00901193444651999)
--(axis cs:-0.14,0.00871597050671447)
--(axis cs:-0.134,0.00845084683817449)
--(axis cs:-0.128,0.00820959001661303)
--(axis cs:-0.122,0.00798705102292385)
--(axis cs:-0.116,0.00778447817683836)
--(axis cs:-0.11,0.00760445070453908)
--(axis cs:-0.104,0.00744721001013027)
--(axis cs:-0.0979999999999996,0.00730866098051208)
--(axis cs:-0.0919999999999996,0.00718825408307466)
--(axis cs:-0.0859999999999996,0.00708779812190163)
--(axis cs:-0.0799999999999996,0.00700203395306553)
--(axis cs:-0.0739999999999996,0.00692887205617321)
--(axis cs:-0.0679999999999996,0.00686567807700922)
--(axis cs:-0.0619999999999996,0.00681058608727386)
--(axis cs:-0.0559999999999996,0.0067643339389834)
--(axis cs:-0.0499999999999996,0.00672992778078024)
--(axis cs:-0.0439999999999996,0.00670586289179043)
--(axis cs:-0.0379999999999996,0.00669697878083464)
--(axis cs:-0.0319999999999996,0.0066967315965783)
--(axis cs:-0.0259999999999996,0.00670701687420389)
--(axis cs:-0.0199999999999996,0.00670843040681479)
--(axis cs:-0.0139999999999996,0.00652301822471379)
--(axis cs:-0.00799999999999956,0.00635191442915195)
--(axis cs:-0.00199999999999956,0.00619438450987847)
--(axis cs:0,0.00614460560523217)
--(axis cs:0.006,0.00600475945648282)
--(axis cs:0.012,0.00587839878874685)
--(axis cs:0.018,0.00576682688955799)
--(axis cs:0.024,0.00567488441073161)
--(axis cs:0.03,0.00560195454213896)
--(axis cs:0.036,0.005547105341701)
--(axis cs:0.042,0.00551017485591732)
--(axis cs:0.048,0.00548900838480104)
--(axis cs:0.054,0.00549064994292892)
--(axis cs:0.06,0.00551281694395389)
--(axis cs:0.066,0.00555540066521154)
--(axis cs:0.072,0.00562416899859486)
--(axis cs:0.078,0.00572660886915977)
--(axis cs:0.0835044905543327,0.00585041128534514)
--(axis cs:0.084,0.00586307865817404)
--(axis cs:0.09,0.00603763694165693)
--(axis cs:0.096,0.00625543515078364)
--(axis cs:0.102,0.00651821806347425)
--(axis cs:0.108,0.006831474037298)
--(axis cs:0.114,0.0072009487919541)
--(axis cs:0.12,0.00763310011666152)
--(axis cs:0.126,0.00803117583453383)
--(axis cs:0.132,0.00808474085053168)
--(axis cs:0.138,0.00814575061439904)
--(axis cs:0.144,0.00821328522313858)
--(axis cs:0.15,0.00829424518806443)
--(axis cs:0.156,0.00839470235055784)
--(axis cs:0.162,0.00851313833413642)
--(axis cs:0.168,0.00864564568872073)
--(axis cs:0.174,0.0087898812095677)
--(axis cs:0.18,0.00895209316162671)
--(axis cs:0.186,0.00912467233047831)
--(axis cs:0.192,0.00931186809722906)
--(axis cs:0.198,0.00951305186601109)
--(axis cs:0.204,0.00973703385078106)
--(axis cs:0.21,0.00998508137143161)
--(axis cs:0.216,0.010254683981246)
--(axis cs:0.222,0.0105508264604607)
--(axis cs:0.228,0.0107909635316249)
--(axis cs:0.234,0.0109305592520782)
--(axis cs:0.24,0.011097829054982)
--(axis cs:0.246,0.0112931745360815)
--(axis cs:0.252,0.0115110157352457)
--(axis cs:0.258,0.0117672883621994)
--(axis cs:0.264,0.0120589522432409)
--(axis cs:0.27,0.0123825606205732)
--(axis cs:0.276,0.0127400588999125)
--(axis cs:0.282,0.0131305576786778)
--(axis cs:0.288,0.0135621693652512)
--(axis cs:0.294,0.0140348436112154)
--(axis cs:0.3,0.014544011337793)
--(axis cs:0.306,0.0151164147065599)
--(axis cs:0.312,0.0157491331746129)
--(axis cs:0.318,0.0164471521405634)
--(axis cs:0.324,0.0172126265590323)
--(axis cs:0.33,0.0180401708839695)
--(axis cs:0.336,0.0189385026401737)
--(axis cs:0.342,0.0199218663731669)
--(axis cs:0.348,0.0209764925586542)
--(axis cs:0.354,0.0221041160663081)
--(axis cs:0.36,0.0233173503398605)
--(axis cs:0.366,0.0246268715265779)
--(axis cs:0.372,0.0260210496732983)
--(axis cs:0.378,0.0275279899590187)
--(axis cs:0.384,0.0291659863327339)
--(axis cs:0.39,0.0309481958801777)
--(axis cs:0.396,0.0328898086063371)
--(axis cs:0.402,0.0349871935704332)
--(axis cs:0.408,0.0372105191674047)
--(axis cs:0.414,0.0395942070157208)
--(axis cs:0.42,0.0421355136439626)
--(axis cs:0.426,0.0448162247696174)
--(axis cs:0.432,0.0476414851620926)
--(axis cs:0.438,0.0506464933271696)
--(axis cs:0.444,0.0538305834185005)
--(axis cs:0.45,0.0571820612053083)
--(axis cs:0.456,0.0607177770723512)
--(axis cs:0.462,0.064446469859622)
--(axis cs:0.468,0.068091534301051)
--(axis cs:0.474,0.0714509005935848)
--(axis cs:0.48,0.0750041516943409)
--(axis cs:0.486,0.0787072604904719)
--(axis cs:0.492,0.0825763891153177)
--(axis cs:0.492,0.281809171148689)
--(axis cs:0.492,0.281809171148689)
--(axis cs:0.486,0.275177238087242)
--(axis cs:0.48,0.268573646624181)
--(axis cs:0.474,0.262031627971389)
--(axis cs:0.468,0.255585597717161)
--(axis cs:0.462,0.2492520727864)
--(axis cs:0.456,0.243003192542441)
--(axis cs:0.45,0.2368218501193)
--(axis cs:0.444,0.230650028113228)
--(axis cs:0.438,0.224580206092846)
--(axis cs:0.432,0.218629090262407)
--(axis cs:0.426,0.212804459199422)
--(axis cs:0.42,0.207088000416965)
--(axis cs:0.414,0.201455211361025)
--(axis cs:0.408,0.195899342765592)
--(axis cs:0.402,0.190476651530141)
--(axis cs:0.396,0.185200493617722)
--(axis cs:0.39,0.180000009476319)
--(axis cs:0.384,0.174914171050267)
--(axis cs:0.378,0.169935013039547)
--(axis cs:0.372,0.165060393920852)
--(axis cs:0.366,0.160320675079194)
--(axis cs:0.36,0.15570013300684)
--(axis cs:0.354,0.151196509423371)
--(axis cs:0.348,0.14683723510848)
--(axis cs:0.342,0.142617415594296)
--(axis cs:0.336,0.138508665619562)
--(axis cs:0.33,0.134517849084417)
--(axis cs:0.324,0.13067865299195)
--(axis cs:0.318,0.126951651668906)
--(axis cs:0.312,0.123316720164404)
--(axis cs:0.306,0.121160891439931)
--(axis cs:0.3,0.119114947618267)
--(axis cs:0.294,0.117095976344096)
--(axis cs:0.288,0.115096089301986)
--(axis cs:0.282,0.113071040268767)
--(axis cs:0.276,0.111065104757496)
--(axis cs:0.27,0.109087236483041)
--(axis cs:0.264,0.107129018072057)
--(axis cs:0.258,0.105172165402753)
--(axis cs:0.252,0.103228524081511)
--(axis cs:0.246,0.10130879476782)
--(axis cs:0.24,0.0993696490440313)
--(axis cs:0.234,0.0974200457782973)
--(axis cs:0.228,0.0954884700778496)
--(axis cs:0.222,0.0935802168947857)
--(axis cs:0.216,0.0916919754086794)
--(axis cs:0.21,0.0898440090829889)
--(axis cs:0.204,0.0880506203426537)
--(axis cs:0.198,0.0862978412904951)
--(axis cs:0.192,0.084590209341965)
--(axis cs:0.186,0.0829150443217678)
--(axis cs:0.18,0.0812523103186131)
--(axis cs:0.174,0.0796031950012186)
--(axis cs:0.168,0.0779731989249943)
--(axis cs:0.162,0.0763449798558583)
--(axis cs:0.156,0.074746417353033)
--(axis cs:0.15,0.0732042300672495)
--(axis cs:0.144,0.0725842031418087)
--(axis cs:0.138,0.072096192014769)
--(axis cs:0.132,0.07161740585849)
--(axis cs:0.126,0.0711476175097484)
--(axis cs:0.12,0.0706844762331135)
--(axis cs:0.114,0.0702244035246525)
--(axis cs:0.108,0.0697547163948903)
--(axis cs:0.102,0.0693014814269174)
--(axis cs:0.096,0.0688255889819313)
--(axis cs:0.09,0.0683654306964759)
--(axis cs:0.084,0.0679284760790609)
--(axis cs:0.0835044905543327,0.0678930064900954)
--(axis cs:0.078,0.0674720754186203)
--(axis cs:0.072,0.0670249038347183)
--(axis cs:0.066,0.0665887214163662)
--(axis cs:0.06,0.0661431314924302)
--(axis cs:0.054,0.0656621092994719)
--(axis cs:0.048,0.0652472565166516)
--(axis cs:0.042,0.0651779757918668)
--(axis cs:0.036,0.065153016679988)
--(axis cs:0.03,0.0651654823171398)
--(axis cs:0.024,0.0652201905533972)
--(axis cs:0.018,0.0653210027406798)
--(axis cs:0.012,0.0654522827653318)
--(axis cs:0.006,0.0656356106874161)
--(axis cs:0,0.0658555860682715)
--(axis cs:-0.00199999999999956,0.0659402898512982)
--(axis cs:-0.00799999999999956,0.0662187379763982)
--(axis cs:-0.0139999999999996,0.0665323586390139)
--(axis cs:-0.0199999999999996,0.0668658733887047)
--(axis cs:-0.0259999999999996,0.0672059652511408)
--(axis cs:-0.0319999999999996,0.0675628210216393)
--(axis cs:-0.0379999999999996,0.0679500344573437)
--(axis cs:-0.0439999999999996,0.0683985839305848)
--(axis cs:-0.0499999999999996,0.0688524012500452)
--(axis cs:-0.0559999999999996,0.0693054420501821)
--(axis cs:-0.0619999999999996,0.0697766642901602)
--(axis cs:-0.0679999999999996,0.0702684970334321)
--(axis cs:-0.0739999999999996,0.0708066550348188)
--(axis cs:-0.0799999999999996,0.0713791967805649)
--(axis cs:-0.0859999999999996,0.071984607699956)
--(axis cs:-0.0919999999999996,0.0726238674683775)
--(axis cs:-0.0979999999999996,0.0743342480816934)
--(axis cs:-0.104,0.0788656273932739)
--(axis cs:-0.11,0.0834111452188929)
--(axis cs:-0.116,0.0879458200655853)
--(axis cs:-0.122,0.0924315245670103)
--(axis cs:-0.128,0.0969604369091762)
--(axis cs:-0.134,0.101500584159956)
--(axis cs:-0.14,0.105973661011076)
--(axis cs:-0.146,0.110467710683444)
--(axis cs:-0.152,0.114948456931465)
--(axis cs:-0.158,0.119372527521832)
--(axis cs:-0.164,0.123762354534457)
--(axis cs:-0.17,0.128132910306784)
--(axis cs:-0.176,0.132484576110645)
--(axis cs:-0.182,0.136805001244327)
--(axis cs:-0.188,0.141078003711112)
--(axis cs:-0.194,0.145286961893512)
--(axis cs:-0.2,0.149525601857282)
--(axis cs:-0.206,0.153728013670504)
--(axis cs:-0.212,0.157910144606523)
--(axis cs:-0.218,0.162102578474821)
--(axis cs:-0.224,0.166252031142639)
--(axis cs:-0.23,0.170333068195743)
--(axis cs:-0.236,0.174382591562327)
--(axis cs:-0.242,0.178431762602533)
--(axis cs:-0.248,0.182403876718639)
--(axis cs:-0.254,0.186332867510263)
--(axis cs:-0.26,0.190228095298567)
--(axis cs:-0.266,0.194145602476865)
--(axis cs:-0.272,0.198002074472165)
--(axis cs:-0.278,0.201832040414656)
--(axis cs:-0.284,0.205656600868316)
--(axis cs:-0.29,0.209455651164829)
--(axis cs:-0.296,0.213312929532038)
--(axis cs:-0.302,0.21723903139202)
--(axis cs:-0.308,0.221197123632999)
--(axis cs:-0.314,0.225141861937792)
--(axis cs:-0.32,0.229093665256564)
--(axis cs:-0.326,0.233095986944497)
--(axis cs:-0.332,0.237135259952071)
--(axis cs:-0.338,0.241274546542102)
--(axis cs:-0.344,0.245510248088549)
--(axis cs:-0.35,0.249809005221556)
--(axis cs:-0.356,0.25421267041449)
--(axis cs:-0.362,0.258792961028744)
--(axis cs:-0.368,0.263563354116074)
--(axis cs:-0.374,0.26852011495416)
--(axis cs:-0.38,0.273660463257965)
--(axis cs:-0.386,0.278943583458423)
--(axis cs:-0.392,0.284457416522236)
--(axis cs:-0.398,0.290157708898652)
--(axis cs:-0.404,0.296049696218851)
--(axis cs:-0.41,0.302119865599448)
--(axis cs:-0.416,0.308322909741069)
--(axis cs:-0.422,0.314665284152583)
--(axis cs:-0.428,0.321174469178747)
--(axis cs:-0.434,0.32782381675064)
--(axis cs:-0.44,0.334579974338666)
--(axis cs:-0.446,0.341445404645457)
--(axis cs:-0.452,0.34843937267829)
--(axis cs:-0.458,0.355571776424231)
--(axis cs:-0.464,0.362821307666758)
--(axis cs:-0.47,0.370182316110222)
--(axis cs:-0.476,0.377687246443571)
--(axis cs:-0.482,0.385312141000146)
--(axis cs:-0.488,0.393032827464012)
--(axis cs:-0.494,0.400876922555417)
--(axis cs:-0.5,0.408855287179886)
--cycle;

\addplot [semithick, black]
table {%
-0.5 0.12204947864344
-0.494 0.116801394568799
-0.488 0.112363557592141
-0.482 0.108045239817626
-0.476 0.103843830318043
-0.47 0.0997483754359542
-0.464 0.0958343016114733
-0.458 0.092381176227272
-0.452 0.0890475958004029
-0.446 0.0858529073518871
-0.44 0.0827603910648226
-0.434 0.0797766206445574
-0.428 0.0768585603413732
-0.422 0.0740294314925385
-0.416 0.0712881029955651
-0.41 0.0686408268265067
-0.404 0.0660832366193859
-0.398 0.063610927083821
-0.392 0.0612203135008005
-0.386 0.0589125763974847
-0.38 0.0567469507478994
-0.374 0.0547123050521261
-0.368 0.0527268065208114
-0.362 0.0511054793677883
-0.356 0.0497616846240036
-0.35 0.0484901624610688
-0.344 0.0473265369691437
-0.338 0.0458432459224295
-0.332 0.0442795937503081
-0.326 0.0427849680451914
-0.32 0.0413534373702684
-0.314 0.0399695641969539
-0.308 0.038844616064737
-0.302 0.0380540710982266
-0.296 0.0374711495496342
-0.29 0.0368985715656808
-0.284 0.0363358135638715
-0.278 0.03567911941017
-0.272 0.0342719172111429
-0.266 0.0329406990041683
-0.26 0.0316836475410469
-0.254 0.0304836992194189
-0.248 0.0293722869088224
-0.242 0.0283780875828985
-0.236 0.0274323119664909
-0.23 0.0267184555311126
-0.224 0.02622408555591
-0.218 0.0252580896368748
-0.212 0.0243202047699526
-0.206 0.0235730162552169
-0.2 0.0228927774072613
-0.194 0.0224220939180583
-0.188 0.0214854488856579
-0.182 0.0207280026730266
-0.176 0.0201086548079329
-0.17 0.019589103462399
-0.164 0.0191081109298679
-0.158 0.0186661995476763
-0.152 0.0182617631131902
-0.146 0.0178880497774221
-0.14 0.0175441182082201
-0.134 0.0171524361159061
-0.128 0.0160867776240605
-0.122 0.0152056703697542
-0.116 0.0146039910344112
-0.11 0.0141530331134835
-0.104 0.0137393691424427
-0.0979999999999996 0.013369252283261
-0.0919999999999996 0.0130581754388073
-0.0859999999999996 0.0127598658361158
-0.0799999999999996 0.0124743373418131
-0.0739999999999996 0.0122054809185817
-0.0679999999999996 0.0119492262999833
-0.0619999999999996 0.0117017440781804
-0.0559999999999996 0.0114756366848576
-0.0499999999999996 0.0113856981052394
-0.0439999999999996 0.0113746188646453
-0.0379999999999996 0.0113727609487558
-0.0319999999999996 0.0113923199985983
-0.0259999999999996 0.0114325460440445
-0.0199999999999996 0.0114787525067639
-0.0139999999999996 0.0115298700258344
-0.00799999999999956 0.0115940461726499
-0.00199999999999956 0.0116774352444947
0 0.0117090336790205
0.006 0.0118206233794551
0.012 0.0109818462384777
0.018 0.0111103552032893
0.024 0.0112666070736912
0.03 0.0114455064091088
0.036 0.011646363908145
0.042 0.0119459909995124
0.048 0.0123049658553015
0.054 0.0123686900891595
0.06 0.0123920061714256
0.066 0.0126348337839392
0.072 0.0129296909052416
0.078 0.0132473664101973
0.0835044905543327 0.0135578256284778
0.084 0.013586833461505
0.09 0.0139606816581074
0.096 0.014400689693326
0.102 0.014878424603889
0.108 0.0153931787239842
0.114 0.0159474144807332
0.12 0.0165447752228968
0.126 0.0171745598199067
0.132 0.0178516120474669
0.138 0.0183045464246455
0.144 0.0186328183032166
0.15 0.0187589474878352
0.156 0.0192537729011215
0.162 0.0198859005436965
0.168 0.0202389742295461
0.174 0.0206311815361687
0.18 0.0210521322610516
0.186 0.02148266277407
0.192 0.0217319589184138
0.198 0.0220196204299996
0.204 0.0231404614051244
0.21 0.0237399869140024
0.216 0.0244022523498729
0.222 0.0253233798152027
0.228 0.0264152317098483
0.234 0.027883490380315
0.24 0.0294160417514299
0.246 0.0310020380714028
0.252 0.0326492471198035
0.258 0.0343471868426155
0.264 0.0360772172736532
0.27 0.0378632420296915
0.276 0.0396663474291256
0.282 0.0414990963042511
0.288 0.0433664121427498
0.294 0.045270646839747
0.3 0.0472273393414397
0.306 0.0492565093542119
0.312 0.0513404523460095
0.318 0.0530355592607267
0.324 0.0546579594625431
0.33 0.0563282900775755
0.336 0.0580203933388751
0.342 0.059770620011036
0.348 0.0616017019979153
0.354 0.0635052259874647
0.36 0.0652061297914352
0.366 0.0668927670205646
0.372 0.0686180238970014
0.378 0.0704603822789765
0.384 0.0728858273826778
0.39 0.0753681724140063
0.396 0.0779108808100702
0.402 0.0809190142632651
0.408 0.0839653814407448
0.414 0.0871775879206484
0.42 0.0905344815551258
0.426 0.0939811740388646
0.432 0.0977064706127626
0.438 0.10041317387974
0.444 0.103540687496055
0.45 0.106788021326905
0.456 0.110114375784118
0.462 0.113554445050653
0.468 0.117062606192069
0.474 0.12065124141731
0.48 0.124329877701141
0.486 0.128124766640173
0.492 0.132050176699828
};
\addplot [semithick, black]
table {%
-0.5 0.197853985577954
-0.494 0.192714123617542
-0.488 0.187615559795003
-0.482 0.181537889809567
-0.476 0.17568130582062
-0.47 0.170009059713788
-0.464 0.164510977411462
-0.458 0.159200447374043
-0.452 0.154076886791603
-0.446 0.149133699867694
-0.44 0.144371681020078
-0.434 0.139788667731185
-0.428 0.135372667266252
-0.422 0.129731687551599
-0.416 0.123985014066708
-0.41 0.118451775985807
-0.404 0.113174903439107
-0.398 0.107694179168263
-0.392 0.102331588484508
-0.386 0.0981628751939206
-0.38 0.0942634100949978
-0.374 0.0900664142064939
-0.368 0.0860666128787269
-0.362 0.0822429142632003
-0.356 0.0786094984951718
-0.35 0.0751421685998446
-0.344 0.0718491237689569
-0.338 0.0687260656886698
-0.332 0.0657655732692427
-0.326 0.0629525841873899
-0.32 0.0602756824638219
-0.314 0.0577464156236764
-0.308 0.0560390859884354
-0.302 0.0544435229265849
-0.296 0.0529574366201001
-0.29 0.0515781696194427
-0.284 0.0503004203012085
-0.278 0.0491081543320631
-0.272 0.0479759987408022
-0.266 0.0469525265789335
-0.26 0.0455102400776365
-0.254 0.0434184110372519
-0.248 0.0417012524018481
-0.242 0.0405757429767673
-0.236 0.0395242754047454
-0.23 0.0385399542893091
-0.224 0.0374036104055669
-0.218 0.0365968515884602
-0.212 0.0361239427324123
-0.206 0.0356624407554619
-0.2 0.0352304818137743
-0.194 0.0348441883500095
-0.188 0.0344820183866296
-0.182 0.0341538195407645
-0.176 0.0337614741288249
-0.17 0.0318373195544148
-0.164 0.0313300585116697
-0.158 0.0309606298156864
-0.152 0.0306093375547508
-0.146 0.0302772069625462
-0.14 0.0299661050229531
-0.134 0.0296907532221839
-0.128 0.0294313070728506
-0.122 0.0291862112283485
-0.116 0.0289744556330696
-0.11 0.0287921758952246
-0.104 0.0286294771941132
-0.0979999999999996 0.0284835007830575
-0.0919999999999996 0.0282561955883561
-0.0859999999999996 0.0277177529076501
-0.0799999999999996 0.0271999196062008
-0.0739999999999996 0.026708889782259
-0.0679999999999996 0.0262480016423995
-0.0619999999999996 0.0258163141703971
-0.0559999999999996 0.0254178485623364
-0.0499999999999996 0.0250482991667073
-0.0439999999999996 0.024739506980479
-0.0379999999999996 0.0248611514586221
-0.0319999999999996 0.0250090524199676
-0.0259999999999996 0.0251887539936064
-0.0199999999999996 0.0254030974231461
-0.0139999999999996 0.0256532010545304
-0.00799999999999956 0.0257469771064671
-0.00199999999999956 0.0242701978648088
0 0.0242524756130125
0.006 0.0242186824791702
0.012 0.0242118908672503
0.018 0.0242543279335031
0.024 0.0243406816237025
0.03 0.0244642604767417
0.036 0.0246165987711818
0.042 0.0248700169553098
0.048 0.0253295964126409
0.054 0.0258295059966424
0.06 0.0263629733712158
0.066 0.0269433550198693
0.072 0.0266514499842523
0.078 0.0263081574274961
0.0835044905543327 0.0260067969361533
0.084 0.0259796444899916
0.09 0.025650014542427
0.096 0.0253460440681126
0.102 0.0250647808805742
0.108 0.0248035108781588
0.114 0.0247049265705889
0.12 0.0251447984098199
0.126 0.0258040786645014
0.132 0.0268029961658501
0.138 0.027869537490256
0.144 0.0289856990355918
0.15 0.0301596372077776
0.156 0.0313763690967203
0.162 0.032637397116063
0.168 0.0338578515866461
0.174 0.0348741596007853
0.18 0.0359388009249224
0.186 0.0370431668418886
0.192 0.0381970557559092
0.198 0.0394003361418437
0.204 0.0406354832705066
0.21 0.0419037527997314
0.216 0.0432097710173327
0.222 0.0445413514502622
0.228 0.0458974791359344
0.234 0.0472458414371805
0.24 0.0481695420505376
0.246 0.049094307722112
0.252 0.0503039695641843
0.258 0.0519222567510931
0.264 0.0535856283492828
0.27 0.0558154960381912
0.276 0.0582514202536344
0.282 0.0607580147913848
0.288 0.0632560609436782
0.294 0.0646159299499404
0.3 0.066964567986943
0.306 0.0694258593681473
0.312 0.0719914060563996
0.318 0.0750438352686034
0.324 0.078473379399891
0.33 0.0820264205831084
0.336 0.085718934505688
0.342 0.0895548754689772
0.348 0.0924523515202497
0.354 0.0952851936867418
0.36 0.0981414066997967
0.366 0.100732891422125
0.372 0.103192197358809
0.378 0.10571771186794
0.384 0.10830714919942
0.39 0.111052147971527
0.396 0.11401394625717
0.402 0.117107469979106
0.408 0.120309816353897
0.414 0.123648967242923
0.42 0.127108690999076
0.426 0.130811974705138
0.432 0.135130228746489
0.438 0.139265926592188
0.444 0.142786518246
0.45 0.146373237457913
0.456 0.150042658966761
0.462 0.153770763611751
0.468 0.157252902479296
0.474 0.160702726247446
0.48 0.164239692294716
0.486 0.1678542188864
0.492 0.171544296264583
};
\addplot [semithick, black]
table {%
-0.5 0.230897405532882
-0.494 0.222966748774461
-0.488 0.217452797914731
-0.482 0.211703766427855
-0.476 0.204541791282343
-0.47 0.197954829143713
-0.464 0.19327991937298
-0.458 0.188671707030898
-0.452 0.183596459203694
-0.446 0.178418864429292
-0.44 0.173339889321706
-0.434 0.168356261861985
-0.428 0.163462891082794
-0.422 0.157858017255972
-0.416 0.152322585798958
-0.41 0.146929726964448
-0.404 0.142047455205203
-0.398 0.138592504424636
-0.392 0.135231684655106
-0.386 0.131969775703479
-0.38 0.128749674994051
-0.374 0.12557911097768
-0.368 0.122468909941422
-0.362 0.119459246904315
-0.356 0.116525977846248
-0.35 0.113639626039647
-0.344 0.110792965233917
-0.338 0.108034794263943
-0.332 0.105345218713636
-0.326 0.102724232387567
-0.32 0.100168745423718
-0.314 0.097674457693117
-0.308 0.0952272328969735
-0.302 0.0928338821071126
-0.296 0.0908074425243027
-0.29 0.0889979083233643
-0.284 0.087205708386351
-0.278 0.0854803308928055
-0.272 0.0838023265152121
-0.266 0.0821378247162176
-0.26 0.0805252048403944
-0.254 0.0789423996690155
-0.248 0.0773928751713094
-0.242 0.0753784872864787
-0.236 0.0733612274670398
-0.23 0.0698868469119194
-0.224 0.0661494076370704
-0.218 0.064017272610237
-0.212 0.0625597189257495
-0.206 0.0607420077599528
-0.2 0.0595215692831204
-0.194 0.0583687304697652
-0.188 0.0573123480943745
-0.182 0.0563570170773782
-0.176 0.0554928483187475
-0.17 0.05469970607182
-0.164 0.0539914806191031
-0.158 0.0533766544631483
-0.152 0.05272786202733
-0.146 0.0520877759147781
-0.14 0.0515259185801202
-0.134 0.0510238722550124
-0.128 0.0505772820791718
-0.122 0.0501157870098109
-0.116 0.0496577202750286
-0.11 0.0496831039685728
-0.104 0.0492216782492406
-0.0979999999999996 0.0487123670889167
-0.0919999999999996 0.0484153991761186
-0.0859999999999996 0.0481622172198463
-0.0799999999999996 0.0478350264570814
-0.0739999999999996 0.0474944347008086
-0.0679999999999996 0.0471868888692763
-0.0619999999999996 0.0469106575487837
-0.0559999999999996 0.0454431591512395
-0.0499999999999996 0.0431051096920075
-0.0439999999999996 0.0425631665821014
-0.0379999999999996 0.0425485653673203
-0.0319999999999996 0.0429451363446589
-0.0259999999999996 0.0432912655966948
-0.0199999999999996 0.0434380606103551
-0.0139999999999996 0.0417186976185641
-0.00799999999999956 0.0417958962004857
-0.00199999999999956 0.0412201833242389
0 0.0410082458853325
0.006 0.0403805861101842
0.012 0.0397507184737865
0.018 0.0391274425303744
0.024 0.0384964267076468
0.03 0.0380461852006125
0.036 0.0378944715555624
0.042 0.037763500407095
0.048 0.037679182121972
0.054 0.0376288125935372
0.06 0.0385048599727189
0.066 0.0394639719228303
0.072 0.0404739370450905
0.078 0.041540486482252
0.0835044905543327 0.0425523816174275
0.084 0.0426444720752567
0.09 0.043758640840357
0.096 0.0448770021282534
0.102 0.0460015269507667
0.108 0.0469979470267984
0.114 0.0476117736660793
0.12 0.0482434522156926
0.126 0.0488831052465122
0.132 0.0495619098209822
0.138 0.0502834027531544
0.144 0.0510252700084517
0.15 0.0518027127576135
0.156 0.0532860624584199
0.162 0.0550361786211151
0.168 0.0567260462153909
0.174 0.0584342985355188
0.18 0.0601779071440361
0.186 0.0619552553422945
0.192 0.0634689312214792
0.198 0.0647820133458434
0.204 0.0661665789506899
0.21 0.0674704385187507
0.216 0.0687464105136426
0.222 0.0700765046999498
0.228 0.0714703422348374
0.234 0.0729303023255315
0.24 0.0744780974967024
0.246 0.0761103371662373
0.252 0.0778313948676521
0.258 0.0796258314751512
0.264 0.0815092077745412
0.27 0.0834876386144975
0.276 0.0852283764954705
0.282 0.0868585952466117
0.288 0.088553977100684
0.294 0.0908723956074849
0.3 0.093864048126809
0.306 0.0968801972544849
0.312 0.0998576485365926
0.318 0.102770192676783
0.324 0.106134122411303
0.33 0.108995651449603
0.336 0.111291408317009
0.342 0.113692032298611
0.348 0.116225614535429
0.354 0.120466397652406
0.36 0.124837486764931
0.366 0.129337429238212
0.372 0.133950313399183
0.378 0.138359262585678
0.384 0.142589244039917
0.39 0.146888586307906
0.396 0.152281382079487
0.402 0.158947980821368
0.408 0.164267272691854
0.414 0.167398786778572
0.42 0.170602666012841
0.426 0.173860277854893
0.432 0.177461725675568
0.438 0.182562966755941
0.444 0.187745516188392
0.45 0.192338442359356
0.456 0.196976804370405
0.462 0.202299045886346
0.468 0.207714967803314
0.474 0.213261121174394
0.48 0.218905920324227
0.486 0.224657364480009
0.492 0.230488931314924
};
\addplot [very thick, green!50.1960784313725!black]
table {%
-0.5 1.26459834145471
-0.494 1.24967155694504
-0.488 1.23486521867274
-0.482 1.22006772595373
-0.476 1.20526993326321
-0.47 1.19044993786511
-0.464 1.17572344658345
-0.458 1.16107667861355
-0.452 1.1464582777276
-0.446 1.13189472233396
-0.44 1.11730055910484
-0.434 1.10275202483253
-0.428 1.08814678435709
-0.422 1.07350410250534
-0.416 1.05886932851126
-0.41 1.04419419764416
-0.404 1.02947443686197
-0.398 1.01474927975258
-0.392 0.99999299897559
-0.386 0.985187177557344
-0.38 0.970347236107156
-0.374 0.955422646805877
-0.368 0.94041509189463
-0.362 0.925240814925062
-0.356 0.910028129510746
-0.35 0.894882771009096
-0.344 0.879722373022553
-0.338 0.864615073122731
-0.332 0.849506069317801
-0.326 0.8343540061104
-0.32 0.819224508212499
-0.314 0.804060202788293
-0.308 0.788954813700092
-0.302 0.773872297078895
-0.296 0.758777437349285
-0.29 0.743656170412158
-0.284 0.728531096490707
-0.278 0.713533967762508
-0.272 0.698512376099932
-0.266 0.68344660851064
-0.26 0.668323823929388
-0.254 0.653191320624541
-0.248 0.638020257438208
-0.242 0.622729411191731
-0.236 0.607361825443254
-0.23 0.591822707948184
-0.224 0.576209345698565
-0.218 0.560531332443333
-0.212 0.544749210028699
-0.206 0.528869965771738
-0.2 0.512880968725436
-0.194 0.496798462350498
-0.188 0.480622550088584
-0.182 0.464402137005729
-0.176 0.448103342635392
-0.17 0.431830175457856
-0.164 0.415577063011182
-0.158 0.399224804444735
-0.152 0.382748639898871
-0.146 0.366295418503247
-0.14 0.349845095142762
-0.134 0.333395459401554
-0.128 0.316881311397857
-0.122 0.300317131870728
-0.116 0.283764375978468
-0.11 0.267177696869668
-0.104 0.250625376971533
-0.0979999999999996 0.233972685618754
-0.0919999999999996 0.21741048700095
-0.0859999999999996 0.20108120483222
-0.0799999999999996 0.184891431267091
-0.0739999999999996 0.169006218854942
-0.0679999999999996 0.153329859216511
-0.0619999999999996 0.137905062870629
-0.0559999999999996 0.122738270192282
-0.0499999999999996 0.107975412943331
-0.0439999999999996 0.0936264893938732
-0.0379999999999996 0.0799182123307866
-0.0319999999999996 0.0669436898635556
-0.0259999999999996 0.0547988072195052
-0.0199999999999996 0.0436874818855285
-0.0139999999999996 0.0338221394229957
-0.00799999999999956 0.0256332915145375
-0.00199999999999956 0.0192231545053049
0 0.0174660319420713
0.006 0.0132499121873358
0.012 0.0103536379287315
0.018 0.00839445663732041
0.024 0.00706417450534458
0.03 0.0061281280906258
0.036 0.00544417712960876
0.042 0.0049276554189579
0.048 0.00452764826148377
0.054 0.00420763295964011
0.06 0.00394762705119062
0.066 0.00373092142620124
0.072 0.00354965801221152
0.078 0.00339583769087171
0.0835044905543327 0.0032745322327693
0.084 0.00326437221848874
0.09 0.0031517938691078
0.096 0.00305588226446984
0.102 0.0029725126346577
0.108 0.00290049014586613
0.114 0.0028379743078692
0.12 0.00278387923313939
0.126 0.00273791561776515
0.132 0.00269974264003703
0.138 0.00266957808468221
0.144 0.0026463750125279
0.15 0.00263089693428703
0.156 0.00262214654149578
0.162 0.00261948891258512
0.168 0.00262208444234202
0.174 0.0026308749552747
0.18 0.00264564964246949
0.186 0.00266770541786271
0.192 0.00269728826893217
0.198 0.00273517120538219
0.204 0.00278184667211151
0.21 0.00283775230925426
0.216 0.00290198200200825
0.222 0.00297432191100327
0.228 0.00305688189629338
0.234 0.0031495124728983
0.24 0.00325334183496284
0.246 0.00336921776004673
0.252 0.00349587274477564
0.258 0.00363580646461653
0.264 0.00379287176596016
0.27 0.00396834905501159
0.276 0.00416091515650369
0.282 0.00437146539072109
0.288 0.00460195434982202
0.294 0.00485365907852764
0.3 0.00512575165767626
0.306 0.00542309167514815
0.312 0.00574994672153101
0.318 0.00610755420358117
0.324 0.00649925923585154
0.33 0.00692821252306947
0.336 0.00738424547167044
0.342 0.00787609752577706
0.348 0.00841156477938115
0.354 0.00899756361177134
0.36 0.00963598036926888
0.366 0.0103282620753282
0.372 0.0110792722318759
0.378 0.0118937228652777
0.384 0.0127659763775975
0.39 0.0137062454088556
0.396 0.0147287277096805
0.402 0.0158387866373982
0.408 0.0170522866187415
0.414 0.0183732775978605
0.42 0.0198174978952486
0.426 0.02138643926158
0.432 0.0230751813134663
0.438 0.0248845431288618
0.444 0.0268372210808863
0.45 0.0289201195274267
0.456 0.0311552249851037
0.462 0.033522250702853
0.468 0.0360350707968469
0.474 0.038716305242706
0.48 0.0415328686797805
0.486 0.0444927778648355
0.492 0.0476092602686536
};
\addplot [very thick, red]
table {%
-0.5 0.218895739442601
-0.494 0.213173858532925
-0.488 0.207573977761813
-0.482 0.202092607452161
-0.476 0.196733339604934
-0.47 0.19149191199753
-0.464 0.186377704022723
-0.458 0.181379730301635
-0.452 0.176491001490704
-0.446 0.171726056843667
-0.44 0.167090507126436
-0.434 0.162582627576042
-0.428 0.158188103782372
-0.422 0.153910197304133
-0.416 0.149746943920827
-0.41 0.14569818272131
-0.404 0.141771393755312
-0.398 0.137965109502478
-0.392 0.13427539179578
-0.386 0.130695942249794
-0.38 0.127222567731082
-0.374 0.123845792286087
-0.368 0.12056476780202
-0.362 0.117375106985429
-0.356 0.114274472670006
-0.35 0.111264467239291
-0.344 0.10834083537833
-0.338 0.105495894843597
-0.332 0.102728362425677
-0.326 0.100039818257552
-0.32 0.0974271974273425
-0.314 0.0948808214196837
-0.308 0.0924003994331749
-0.302 0.0899812992667012
-0.296 0.0876261347629553
-0.29 0.0853350523249361
-0.284 0.0830999735952005
-0.278 0.0809319499072482
-0.272 0.0788225025196909
-0.266 0.0767636210483903
-0.26 0.0747631026328601
-0.254 0.0728185759400951
-0.248 0.0709268295912058
-0.242 0.0690841131725349
-0.236 0.0672933065591306
-0.23 0.065553690828092
-0.224 0.0638616561521143
-0.218 0.0622177260535992
-0.212 0.0606214129133886
-0.206 0.0590697847879344
-0.2 0.0575596355003335
-0.194 0.0560914195542248
-0.188 0.0546668035203211
-0.182 0.053280048178939
-0.176 0.0519307106733451
-0.17 0.0506254893000182
-0.164 0.0493624433743979
-0.158 0.0481348582341806
-0.152 0.0469424735774148
-0.146 0.0457871081835782
-0.14 0.0446676080810649
-0.134 0.0435861258558124
-0.128 0.042530465215667
-0.122 0.0415026356037671
-0.116 0.0405021686487905
-0.11 0.0395239569697209
-0.104 0.038568076812927
-0.0979999999999996 0.0376310318761244
-0.0919999999999996 0.0367225643311418
-0.0859999999999996 0.0358450968313709
-0.0799999999999996 0.0349926645014948
-0.0739999999999996 0.0341698248876409
-0.0679999999999996 0.0333729028108857
-0.0619999999999996 0.032604218547195
-0.0559999999999996 0.0318729238260178
-0.0499999999999996 0.0311827995934043
-0.0439999999999996 0.0305345989721196
-0.0379999999999996 0.0299414778577791
-0.0319999999999996 0.0294058639563244
-0.0259999999999996 0.0289270635327812
-0.0199999999999996 0.0285095125723839
-0.0139999999999996 0.028156003107355
-0.00799999999999956 0.0278784701715353
-0.00199999999999956 0.0276792099994
0 0.0276304217908829
0.006 0.027531328451425
0.012 0.0274945787790638
0.018 0.027507564076019
0.024 0.027561095126358
0.03 0.0276476493019699
0.036 0.0277621408893879
0.042 0.0279025351825071
0.048 0.0280666554900398
0.054 0.0282550321695409
0.06 0.0284650478350943
0.066 0.0286973854970693
0.072 0.0289541240830324
0.078 0.0292366362516371
0.0835044905543327 0.0295200657218688
0.084 0.0295467056178015
0.09 0.0298844676521168
0.096 0.0302543809455583
0.102 0.0306580353168589
0.108 0.0310924185459195
0.114 0.0315618886399282
0.12 0.0320653094882403
0.126 0.0326046307921915
0.132 0.0331841995152073
0.138 0.0338055057199588
0.144 0.0344609213695772
0.15 0.0351583693323988
0.156 0.0359002321151829
0.162 0.0366843011353006
0.168 0.0375106995189208
0.174 0.0383770125600303
0.18 0.0392919916905227
0.186 0.0402497693826892
0.192 0.0412538757760653
0.198 0.0423059766105616
0.204 0.0434063380544736
0.21 0.0445569493965128
0.216 0.0457618822333032
0.222 0.0470191612738571
0.228 0.0483303007729888
0.234 0.0496981209556232
0.24 0.0511259167707215
0.246 0.052608579446048
0.252 0.0541492343965245
0.258 0.0557487474300923
0.264 0.0574111765975161
0.27 0.0591379769148296
0.276 0.0609264612269476
0.282 0.0627795610855051
0.288 0.0647041106031068
0.294 0.0666957088190766
0.3 0.0687544057406077
0.306 0.0708859074753244
0.312 0.0730899859720808
0.318 0.0753692434030293
0.324 0.0777221179331578
0.33 0.0801509929173605
0.336 0.0826607154475603
0.342 0.0852472749318472
0.348 0.0879153671108396
0.354 0.0906568182079317
0.36 0.0934800544314866
0.366 0.0963866598080503
0.372 0.0993696452958431
0.378 0.102435517532091
0.384 0.105587084195288
0.39 0.108827302006282
0.396 0.112155743730434
0.402 0.115575355623225
0.408 0.119092571794239
0.414 0.122708996363446
0.42 0.126421724499675
0.426 0.130224578580758
0.432 0.134116975379617
0.438 0.138109796903112
0.444 0.142197665140581
0.45 0.146387262007032
0.456 0.150680531146039
0.462 0.155080243135318
0.468 0.159580130614322
0.474 0.164189580156334
0.48 0.168901525319036
0.486 0.173717842350336
0.492 0.178640967070847
};
\end{axis}

\end{tikzpicture}

%% file: labpal/figure_strategy_metrics.tex
\begin{figure}[t!]
	\tikzsetfigurename{labpal_strategy_metrics}
	\vspace{-0.5cm}
	\centering
	\def\scale{0.4}
	%\def\basis{10e-5}
	%\scalebox{1.0}{
	%\input{"\figureTwoDataPath CIFAR10_mom0_resnet20_augment_result/statistics_plots/distance_matrix.pgf"}}\\
	
	\begin{tabular}{ c c c }
		\scalebox{\scale}{\input{"labpal/figure_data/line_plots/CIFAR10_mom0_resnet20_augment_result/statistics_plots/bs_128_bs_ori_128_sgd_lrs_[0.05]_pal_mus_[0.1]_ri_100_c_apal_0.9_ori_sgd_lr_0.1/step_step.pgf"}}&\hspace{-0.25cm}
		%https://tex.stackexchange.com/questions/447795/pgf-fpu-ill-formatted-floating-point-number-using-shapes-symbols-library	
		\scalebox{\scale}{\input{"labpal/figure_data/line_plots/CIFAR10_mom0_resnet20_augment_result/statistics_plots/bs_128_bs_ori_128_sgd_lrs_[0.05]_pal_mus_[0.1]_ri_100_c_apal_0.9_ori_sgd_lr_0.1/improvements_step_accumulated.pgf"}}&\hspace{-0.25cm}
		
		\scalebox{\scale}{\input{"labpal/figure_data/line_plots/CIFAR10_mom0_resnet20_augment_result/statistics_plots/bs_128_bs_ori_128_sgd_lrs_[0.05]_pal_mus_[0.1]_ri_100_c_apal_0.9_ori_sgd_lr_0.1/fb_min_step_ratio.pgf"}}
	\end{tabular}

	\caption{Several metrics to compare update step strategies on the  full-batch losses along 10,000 lines measured by \cite{line_analysis}: 1. update step sizes, 2. accumulated loss improvement per step given as:  $l(0)-l(s_{upd})$ where $s_{upd}$ is the update step of a specific optimizer. This is the locally optimal improvement to the minimum of the full-batch loss along a line. The right plot shows almost proportional behavior between the optimal update step and the negative gradient norm of the direction defining mini-batch loss. The LABPAL\&SGD version of our approach performs almost optimal on ground truth data. Results  LABPAL\&NSGD are almost identical and thus omitted. Plotting code based on \cite{line_analysis} with addition of our proposed approach.}
	\label{labpal_fig_strategy_metrics}
\end{figure}
%\begin{tikzpicture}
%	\def\basis{10e-6}
%	
%	%https://tex.stackexchange.com/questions/447795/pgf-fpu-ill-formatted-floating-point-number-using-shapes-symbols-library
%	\pgfplotsset
%	{
%		y coord trafo/.code={\begingroup\pgfmathparseFPU{symlog(#1,\basis)}\pgfmathsmuggle\pgfmathresult\endgroup},
%		y coord inv trafo/.code={\begingroup\pgfmathparseFPU{symexp(#1,\basis)}\pgfmathsmuggle\pgfmathresult\endgroup},
%		yticklabel style={/pgf/number format/.cd,int detect,precision=2},
%	}
%pdflatex --enable-write18 -shell-escape -interaction=nonstopmode %.tex

%% file: labpal/figure_data/line_plots/CIFAR10_mom0_resnet20_augment_result/statistics_plots/bs_128_bs_ori_128_sgd_lrs_[0.05]_pal_mus_[0.1]_ri_100_c_apal_0.9_ori_sgd_lr_0.1/fb_min_step_ratio.pgf
% This file was created by tikzplotlib v0.9.6.
\begin{tikzpicture}
\definecolor{color0}{rgb}{1,0.647058823529412,0}
\begin{axis}[
yticklabel style={font=\large},
xticklabel style={font=\large},
ytick style={font=\Large},
xtick style={font=\Large},
ylabel style={font=\large},
xlabel style={font=\Large},
grid=major,
major grid style={line width=.2pt,draw=gray!50},
legend style={at={(0.25,0.975)},fill opacity=0.8, draw opacity=1, text opacity=1, draw=white!80!black,font=\large},
minor xtick={},
minor ytick={},
tick align=outside,
tick pos=left,
title={\Large{ratio of step to full-batch min. and norm of dir. grad.}},
title style={xshift=-1cm},
width=10.5cm,height=8cm,
x grid style={white!69.0196078431373!black},
xlabel={line number/update step},
xmin=-468, xmax=9828,
xtick style={color=black},
xtick={-2000,0,2000,4000,6000,8000,10000},
y grid style={white!69.0196078431373!black},
y tick label style={/pgf/number format/.cd,scaled y ticks = false,set thousands separator={},fixed,precision=4},
ylabel={ratio full-batch min. norm of dir. grad.},
ymin=-0.1, ymax=0.3,
ytick style={color=black},
ytick={-0.1,-0.05,0,0.05,0.1,0.15,0.2,0.25,0.3}
]
\addplot [line width=2pt, color0]
table {%
0 0.0591258084066985
10 0.053115029788327
20 0.051465887183126
30 0.0511653295013136
40 0.052150547314046
50 0.0517028921161306
60 0.0525107795391379
70 0.0513017561933607
80 0.0511336930905917
90 0.0519181159154445
100 0.0523024207067717
110 0.0520482790584625
120 0.0514438848803062
130 0.0512999791387137
140 0.0512444848919709
150 0.0518203507104591
160 0.0508343438063396
170 0.0515613316129044
180 0.0516512564977597
190 0.0513139710383459
200 0.0506566261502988
210 0.0513413550079655
220 0.0509652265260604
230 0.0514445703363361
240 0.0520809850051535
250 0.0514000746372663
260 0.0512493693185794
270 0.0509219027944586
280 0.0504651063227884
290 0.0524493141644855
300 0.0533463877752577
310 0.0541350570821911
320 0.0535671191578709
330 0.0532328686060211
340 0.0527450808000696
350 0.0524706699692754
360 0.0521915876485484
370 0.0529198366277118
380 0.0527467926603667
390 0.0516072085916407
400 0.0513923074223341
410 0.0513223465304827
420 0.0520673300201787
430 0.0533729289292703
440 0.0524033018757341
450 0.0518443922803255
460 0.0520006135878134
470 0.0519548441240451
480 0.0521457238491919
490 0.05290361550021
500 0.0524148792783482
510 0.0521549627694944
520 0.0532780980702767
530 0.0532711768736706
540 0.054954858502583
550 0.0525767697705352
560 0.0536780346761044
570 0.051901115417292
580 0.0515345293117715
590 0.0508401316533419
600 0.0523484828131825
610 0.0533581026917282
620 0.0531851490710108
630 0.053320394987503
640 0.0522432980122931
650 0.0526065360149565
660 0.0543563562981788
670 0.0533718914141951
680 0.0521568829596345
690 0.0534278535847927
700 0.0538316203995256
710 0.054457727560507
720 0.0537285056112817
730 0.053736083924807
740 0.0536924844399707
750 0.0544072922735831
760 0.0534089540756886
770 0.0530423231829834
780 0.0534320693580247
790 0.0543713864046323
800 0.0544473886740284
810 0.0546439220756979
820 0.0525930885715399
830 0.0528872427451826
840 0.0536098319512737
850 0.0546606948348131
860 0.0553975764717785
870 0.0566925360742273
880 0.0551680547452349
890 0.0530276044332872
900 0.0545911937564424
910 0.0529415399889793
920 0.0538074027685214
930 0.0541325721547217
940 0.054520706201727
950 0.0544566678517443
960 0.0531694435434339
970 0.0547145197043392
980 0.0532433976308114
990 0.0536090718100924
1000 0.0532407958520075
1010 0.0555183318288536
1020 0.0555686180881836
1030 0.0550531364284259
1040 0.0543679471113992
1050 0.0539959559561355
1060 0.0539658226454155
1070 0.0526171945556881
1080 0.0538093640683888
1090 0.0522556338144809
1100 0.0533640496773829
1110 0.053936362390461
1120 0.054567646168239
1130 0.0542779453220102
1140 0.0543268363397968
1150 0.0532606001896677
1160 0.0539581740290312
1170 0.0539029801346938
1180 0.052598536286046
1190 0.0534183951913337
1200 0.053740785561449
1210 0.0546103198655419
1220 0.0549779584120495
1230 0.0542943091812588
1240 0.0538077436282723
1250 0.0537408497075131
1260 0.0538816924238461
1270 0.0543913290196011
1280 0.0530021997091534
1290 0.0532199355545959
1300 0.0531577044676842
1310 0.0535459859087564
1320 0.0533196669295034
1330 0.053540593400709
1340 0.0533707724232701
1350 0.0532894649958793
1360 0.0538466013216455
1370 0.0546023730678424
1380 0.0544279362781893
1390 0.0537512225222324
1400 0.054921686569608
1410 0.0539893139883035
1420 0.0546165817003669
1430 0.0529817272962077
1440 0.0537278429425815
1450 0.0547230922797416
1460 0.0549546368210193
1470 0.0552823179850699
1480 0.0552091863220897
1490 0.0540825607526136
1500 0.0547558218165911
1510 0.053299452567714
1520 0.0531946510793829
1530 0.0533548928857688
1540 0.0542199760192576
1550 0.053391144639825
1560 0.0522517864136048
1570 0.0533564707276677
1580 0.0536788291324675
1590 0.053460299779973
1600 0.0537524689744726
1610 0.0529198685165793
1620 0.0531050726433632
1630 0.053735633796596
1640 0.0541984828270184
1650 0.0540897016187488
1660 0.0534290002334539
1670 0.0537475366209502
1680 0.0530812788176464
1690 0.0521043457685243
1700 0.0533155511270907
1710 0.0527523168753926
1720 0.0521312240493468
1730 0.0522691596280412
1740 0.05256652409588
1750 0.0531492294834718
1760 0.0541246600169834
1770 0.0534762423229386
1780 0.0541554379532456
1790 0.0545535882645854
1800 0.0528164122783715
1810 0.0532208923132292
1820 0.0542348816453122
1830 0.0536925016094136
1840 0.0533214194723168
1850 0.0548860587986321
1860 0.0560541157598572
1870 0.0553539834579409
1880 0.0539464759486693
1890 0.0545864120597286
1900 0.054751398298841
1910 0.0549652784703225
1920 0.0543955618665716
1930 0.0543081259350056
1940 0.052199139989343
1950 0.0543211321962565
1960 0.0539920386168088
1970 0.0528563464802822
1980 0.0526602450999261
1990 0.0542503135386335
2000 0.0545109088987116
2010 0.0530849159301493
2020 0.05184145594721
2030 0.0534787579678471
2040 0.0540511924300091
2050 0.0547233694486347
2060 0.0554833154236498
2070 0.0543215912247924
2080 0.0554862830026175
2090 0.0537556574009518
2100 0.0521096036008192
2110 0.0534276517688974
2120 0.0543293835466175
2130 0.0547158711271262
2140 0.0545271250288845
2150 0.0540960348468691
2160 0.0540278583703703
2170 0.0540962576880541
2180 0.0522789695598262
2190 0.0525048754852705
2200 0.0525596894684525
2210 0.0552037028674056
2220 0.0547812169143442
2230 0.0532133419881238
2240 0.0530699196710597
2250 0.0537605434083902
2260 0.0538824452505705
2270 0.0545442083089291
2280 0.0561854056155748
2290 0.0539805262877725
2300 0.0544851582566775
2310 0.0545671661090588
2320 0.0536526273167058
2330 0.0524041905145158
2340 0.0519805096069387
2350 0.0533411120861873
2360 0.0542304697815723
2370 0.0549181256370013
2380 0.0525989819612067
2390 0.0532448663268204
2400 0.0542019032393412
2410 0.0555556119225174
2420 0.0518043267487542
2430 0.0527698267892855
2440 0.0531511428223779
2450 0.0545610972057927
2460 0.0545888978747311
2470 0.0534625869453841
2480 0.0527815919733017
2490 0.053537952660489
2500 0.053255025332832
2510 0.0532034697312274
2520 0.0535451406040248
2530 0.050856030739532
2540 0.0526118201453169
2550 0.0525758597973804
2560 0.0541114479447871
2570 0.0533661257592708
2580 0.0530829833826713
2590 0.0536007927411866
2600 0.0539356901446625
2610 0.0526737055560915
2620 0.0542218925533243
2630 0.0527331316578449
2640 0.0530599741715213
2650 0.0537386436391274
2660 0.0530354488139665
2670 0.0519064924593204
2680 0.0536235587936726
2690 0.052621787548565
2700 0.0546525037804526
2710 0.0549507028547793
2720 0.0542577283254512
2730 0.0555509872522458
2740 0.0541859653669008
2750 0.0528393120513664
2760 0.0537574882104298
2770 0.0533091300493547
2780 0.0534837611108374
2790 0.0531461660864687
2800 0.0527536826890175
2810 0.0534234888164557
2820 0.0536265288261547
2830 0.0522107562797995
2840 0.052643549346649
2850 0.0522277673133369
2860 0.0534793716885554
2870 0.0545771520514526
2880 0.0530279522704415
2890 0.0528332201062991
2900 0.052517719323357
2910 0.0519280476204131
2920 0.0526458839732785
2930 0.0516127263477064
2940 0.0521325277143283
2950 0.0513133058877337
2960 0.0544412211611695
2970 0.0539122765176387
2980 0.0517131773639165
2990 0.0533362152814782
3000 0.0541886146039094
3010 0.0539395955740149
3020 0.0523868500403448
3030 0.0508004621417474
3040 0.0537214285705266
3050 0.053562964144214
3060 0.0517707480421031
3070 0.0514356114914064
3080 0.0515012275045396
3090 0.0520796046766438
3100 0.0533375639875434
3110 0.05462021524019
3120 0.0522996166574302
3130 0.0539102219510374
3140 0.0523815216282307
3150 0.0524317176952601
3160 0.0523960349149204
3170 0.0523172414507008
3180 0.0534681266565723
3190 0.0534553080085083
3200 0.0532190739527778
3210 0.0526377533385713
3220 0.0536918281325233
3230 0.0523658037550515
3240 0.0524030732253588
3250 0.0518146801900238
3260 0.0532456304309001
3270 0.053973037650371
3280 0.0539856078133957
3290 0.0551325686922361
3300 0.0544037757306616
3310 0.0552239322442222
3320 0.0523006535363478
3330 0.0535763120243072
3340 0.0534802767547964
3350 0.053438631450351
3360 0.0553297690130558
3370 0.0564929168068557
3380 0.0556730032171684
3390 0.0548459131663074
3400 0.0536126320113936
3410 0.0527355462449183
3420 0.0501747530212582
3430 0.0526917903444824
3440 0.0512662329970501
3450 0.0515063543308735
3460 0.0531047084030267
3470 0.0525250681884466
3480 0.051278698599416
3490 0.0511304947496307
3500 0.0531391979617758
3510 0.0530510928152852
3520 0.0523666496743788
3530 0.0533479445548202
3540 0.0527473958125997
3550 0.0523749738572554
3560 0.0520448143197431
3570 0.0519862438081979
3580 0.0546427260115758
3590 0.0543272845012571
3600 0.0547257940747716
3610 0.0515198553180096
3620 0.0547197254647317
3630 0.0523647591122764
3640 0.0507770769528546
3650 0.0518007363583678
3660 0.0555012419123656
3670 0.0557387576303409
3680 0.0569423597847727
3690 0.0539178938687395
3700 0.052718136332778
3710 0.0512193526515682
3720 0.0523426963078575
3730 0.0532415494315525
3740 0.0554663713718076
3750 0.0556728059840882
3760 0.0549957577722518
3770 0.0548393676298763
3780 0.0532669272822035
3790 0.051567052289709
3800 0.0527575393443307
3810 0.0525434504266505
3820 0.0524772928331046
3830 0.0519247731061357
3840 0.0515776299521418
3850 0.0524322034894695
3860 0.0541916955706702
3870 0.0552324175799529
3880 0.0549184607623406
3890 0.0536849343089428
3900 0.0536086801796425
3910 0.0523114362393339
3920 0.0514840246746078
3930 0.0540914135448157
3940 0.0540548712256471
3950 0.0527158191374705
3960 0.052826926463061
3970 0.051769932956903
3980 0.0523277958044371
3990 0.05272748074688
4000 0.0522197934326556
4010 0.0520982579610403
4020 0.0536409029495741
4030 0.0521652329433763
4040 0.0527725217736758
4050 0.0525008562328436
4060 0.0535441867737342
4070 0.0544797506926968
4080 0.0536986429230506
4090 0.0539260704521462
4100 0.0542459814233906
4110 0.0557279570382014
4120 0.0523169696261166
4130 0.0505391788230986
4140 0.051416713791727
4150 0.0537094274522346
4160 0.0529237946407056
4170 0.0510974040006637
4180 0.0543582351286507
4190 0.0540485478889493
4200 0.0497547045216202
4210 0.050021013191624
4220 0.0517065410260603
4230 0.0532874801958757
4240 0.053980984199242
4250 0.0533447727615377
4260 0.0551038137556855
4270 0.0546609494303733
4280 0.0555507852846152
4290 0.0549429904073706
4300 0.0561875262657263
4310 0.0586255524426626
4320 0.0608231716666365
4330 0.0553322747064782
4340 0.0537937388054724
4350 0.0525501555711308
4360 0.0536922797816794
4370 0.0548230460998234
4380 0.054641663891207
4390 0.0566490378978554
4400 0.0557156815080153
4410 0.0558470392055693
4420 0.0542471115356313
4430 0.0540755265682941
4440 0.0530476541501711
4450 0.0543118710447828
4460 0.0540389676983401
4470 0.051855171505324
4480 0.0521469486750747
4490 0.0500994420991039
4500 0.0525283753133029
4510 0.0533102591663706
4520 0.0530610836572969
4530 0.0552434637839919
4540 0.0540523686139397
4550 0.0543980995148064
4560 0.053926211576745
4570 0.0513034660233517
4580 0.0524352467713636
4590 0.053473683296099
4600 0.0551247997364551
4610 0.0540365531738135
4620 0.0524935275906364
4630 0.0561499622261509
4640 0.055185703449231
4650 0.062036019737057
4660 0.0572441746721127
4670 0.0577719521764669
4680 0.0552292822418262
4690 0.0555392854948655
4700 0.0573563098387862
4710 0.0541211140850024
4720 0.0535847857610665
4730 0.050900020942335
4740 0.0572338075976507
4750 0.0604956490329412
4760 0.0614381828656928
4770 0.056787601185147
4780 0.0578585958725405
4790 0.055655497487463
4800 0.0550786660064406
4810 0.0561034182938453
4820 0.0577625452728003
4830 0.0553767283390233
4840 0.0576260667986297
4850 0.0601017864231097
4860 0.0587267440157125
4870 0.0606107639433821
4880 0.0582116936695304
4890 0.0546204319139573
4900 0.0571369624582671
4910 0.0571235357410119
4920 0.0555996040571595
4930 0.0539355162273918
4940 0.0600283117065168
4950 0.0600441842991264
4960 0.060270893783181
4970 0.0578161250468316
4980 0.0607152809369106
4990 0.0558665139308705
5000 0.0576335874999092
5010 0.0549140289518628
5020 0.0543965380340336
5030 0.0542595880209631
5040 0.0564381183925231
5050 0.0593041897689351
5060 0.0578162755234518
5070 0.0584478767479598
5080 0.0606748325247907
5090 0.05991058969358
5100 0.0569716502538325
5110 0.0636054930624441
5120 0.0645144015658365
5130 0.0688013279419845
5140 0.0719740452367684
5150 0.0681797227648671
5160 0.0566934286435617
5170 0.0553244727469588
5180 0.0572351955104644
5190 0.0572928374305221
5200 0.0526616849257043
5210 0.0522384168177393
5220 0.0508435164836602
5230 0.0561787556761551
5240 0.0600950812651863
5250 0.0580289410945397
5260 0.0519799897113778
5270 0.0527568356228489
5280 0.056711160432499
5290 0.0570562375748093
5300 0.0570888333945491
5310 0.0599825970727991
5320 0.0608438937296803
5330 0.057428615900627
5340 0.05416525991884
5350 0.0547159260830798
5360 0.0542859391805585
5370 0.0519638615304293
5380 0.0531103301857253
5390 0.0562804472878743
5400 0.0611873727151204
5410 0.0598360355510512
5420 0.0525208268788923
5430 0.0545793304023751
5440 0.0529803243425387
5450 0.0504753426153987
5460 0.0547836763418329
5470 0.0594354186641413
5480 0.068185266130114
5490 0.0691226250265459
5500 0.0619221966900388
5510 0.0595729999467067
5520 0.0526700617431774
5530 0.0650455796777348
5540 0.0720256713983369
5550 0.0694310517647233
5560 0.0608898848832311
5570 0.0580415608355185
5580 0.0554113720060392
5590 0.0555130600395283
5600 0.0573505691423695
5610 0.0585778683918979
5620 0.0686449038412434
5630 0.0615753582581815
5640 0.0595602592033506
5650 0.0548283301266209
5660 0.0599535286392828
5670 0.0605040872471415
5680 0.0616835068963529
5690 0.0629385805057375
5700 0.0609489596650795
5710 0.0644801420268843
5720 0.0613647756379453
5730 0.0680346662132456
5740 0.0626083644197407
5750 0.0592077198629438
5760 0.0558292818333324
5770 0.0606630978042856
5780 0.06381180553969
5790 0.0583582300355364
5800 0.0551993816961653
5810 0.0632136721521295
5820 0.0646159466192716
5830 0.0753406300668217
5840 0.0652344554571608
5850 0.0606141906195404
5860 0.058577669218741
5870 0.0598465115961426
5880 0.0555446640537172
5890 0.0572958509697322
5900 0.0532726085972999
5910 0.0576174895703285
5920 0.0563725920967567
5930 0.0587151577558403
5940 0.0585278345969927
5950 0.0605581975029982
5960 0.064898549601209
5970 0.0603976274460229
5980 0.0556014264993704
5990 0.056853478955014
6000 0.0503995287741977
6010 0.0570200490508718
6020 0.0588098614782469
6030 0.0583467442996858
6040 0.0629069464265716
6050 0.0660191893983116
6060 0.0660621989011342
6070 0.0672024203987727
6080 0.0598822346670766
6090 0.0534112349530328
6100 0.0567787346150075
6110 0.0589330392828192
6120 0.0565339903372484
6130 0.0617425992272416
6140 0.0599223676042475
6150 0.0615658687192575
6160 0.0651907458142524
6170 0.0738896953358778
6180 0.0689228274335938
6190 0.0729859152635038
6200 0.0638092154678646
6210 0.068893926681532
6220 0.0708268342417357
6230 0.0600020029460036
6240 0.0489407647990038
6250 0.0540939550149832
6260 0.0707118381911947
6270 0.0737707397900079
6280 0.0864934652568044
6290 0.0784052867682239
6300 0.0694967768438081
6310 0.0750761400947813
6320 0.0716049479111498
6330 0.068194857917838
6340 0.0633018700276419
6350 0.0702033735917888
6360 0.0763544587553404
6370 0.0710720557237508
6380 0.06943538130884
6390 0.051612555730629
6400 0.0568638932365484
6410 0.0567162515167036
6420 0.0575383051110291
6430 0.054589157728045
6440 0.05542564820865
6450 0.0646580831181231
6460 0.0669189331051563
6470 0.0657923383288445
6480 0.0716174091218064
6490 0.0670106583366639
6500 0.0669003397187184
6510 0.060231847658731
6520 0.0645943555228093
6530 0.0600604478789415
6540 0.0539561932819539
6550 0.0559706532670482
6560 0.0572576327541333
6570 0.0648490893979316
6580 0.0668871371396008
6590 0.0622231760027445
6600 0.0692501283252323
6610 0.0727388300312172
6620 0.073148519565841
6630 0.0786619849417352
6640 0.0713587827302852
6650 0.062110876205181
6660 0.0579326151426194
6670 0.0554063993578122
6680 0.0650014939559499
6690 0.0643173718092496
6700 0.0619496860735199
6710 0.0561639903142822
6720 0.0619288775608863
6730 0.062811264186453
6740 0.0640172410113489
6750 0.0637202407174399
6760 0.0679029194059882
6770 0.0676619413323982
6780 0.06769339870369
6790 0.0760362532424319
6800 0.0679491656511793
6810 0.0600134719095837
6820 0.0508743392439429
6830 0.0588829364592723
6840 0.0643767001296278
6850 0.0728109595520087
6860 0.0728202163419158
6870 0.0707553383425569
6880 0.0587204979355121
6890 0.0636471269313438
6900 0.0852031768367121
6910 0.0944970832231368
6920 0.0899977623417891
6930 0.0691292316107268
6940 0.0706927230916839
6950 0.0762380108481173
6960 0.0810335824032615
6970 0.0748396933153506
6980 0.0858796346732337
6990 0.0891707675302282
7000 0.0775264295908645
7010 0.06747364216561
7020 0.0920442635565269
7030 0.0891077529814466
7040 0.0821537627774947
7050 0.061785218665277
7060 0.0651887341513616
7070 0.066097290331061
7080 0.0674134306478882
7090 0.0928457934959695
7100 0.108344958679216
7110 0.0906509524791744
7120 0.0790624537561582
7130 0.0788870984507205
7140 0.0895334287751302
7150 0.0786813679652826
7160 0.0852351566776295
7170 0.0713882841052667
7180 0.0621036093870111
7190 0.0764577796530962
7200 0.0675821087896208
7210 0.0681609287992882
7220 0.0647353416100002
7230 0.0717099647900779
7240 0.0700950419333474
7250 0.0632436479652787
7260 0.0663009851070997
7270 0.0712150659346616
7280 0.0718864445715253
7290 0.060961889435567
7300 0.0725035605705657
7310 0.0875130426906415
7320 0.0865596196477944
7330 0.089680519184304
7340 0.0896115382540109
7350 0.0794162212971625
7360 0.0751422733246397
7370 0.0723285968731577
7380 0.0804344018837275
7390 0.0794277531717577
7400 0.0833440665291581
7410 0.0989382919099016
7420 0.0885016955000257
7430 0.0835347599295131
7440 0.0784420661986712
7450 0.0739388606493215
7460 0.0954134929273098
7470 0.114104223984233
7480 0.0907956261904473
7490 0.0920656255705165
7500 0.0971888085413874
7510 0.0849929830076486
7520 0.0938614102632758
7530 0.0854856861368809
7540 0.079008843902644
7550 0.0750704781924313
7560 0.0720047757201163
7570 0.0903595750146422
7580 0.0756856370028542
7590 0.0792109222166034
7600 0.0987592695926398
7610 0.0937201468720227
7620 0.0674150639439168
7630 0.0731476630258825
7640 0.0751107254113305
7650 0.0805251855225131
7660 0.0877742536069366
7670 0.0913898165245925
7680 0.0989126095121396
7690 0.0889616576279257
7700 0.0753355746658826
7710 0.074559505903545
7720 0.0811355013650163
7730 0.0881468939227282
7740 0.1021497145026
7750 0.116831471203735
7760 0.0836926189720565
7770 0.0674011394225059
7780 0.0547345446825913
7790 0.0665616200919012
7800 0.0739537992335266
7810 0.0791132335317939
7820 0.0749253948012772
7830 0.0757055238639798
7840 0.0780672513184158
7850 0.0683812274178942
7860 0.0605765725712705
7870 0.0591109916107223
7880 0.0635597665387222
7890 0.0676224040146148
7900 0.078729830631432
7910 0.0748762097803914
7920 0.0707943840573986
7930 0.0831542100144644
7940 0.0813821978873432
7950 0.0760316493196324
7960 0.0724183042464267
7970 0.0631613399344207
7980 0.0639824729061095
7990 0.06613291490961
8000 0.0607681694630282
8010 0.0649705558682244
8020 0.0637630053458599
8030 0.0652204746492284
8040 0.0652309778986066
8050 0.0708036158195209
8060 0.067358125077873
8070 0.0688254569225454
8080 0.0647801736517145
8090 0.0699274935853762
8100 0.073311122907837
8110 0.0908272777650375
8120 0.097032811466642
8130 0.0794183838403329
8140 0.0925634859837351
8150 0.0919255767568276
8160 0.109615939301961
8170 0.0826206548743644
8180 0.0942069335505966
8190 0.0841024974611406
8200 0.10098725388551
8210 0.0956174967553701
8220 0.100955348680841
8230 0.0886513809387703
8240 0.0781018979590938
8250 0.0919549457168535
8260 0.0929263974033454
8270 0.0980947474020726
8280 0.0728983661389841
8290 0.0971351893974134
8300 0.0851216941833244
8310 0.0866812537866661
8320 0.089734310513237
8330 0.0939887546485463
8340 0.0922353165270342
8350 0.0964001812114529
8360 0.100800393791753
8370 0.0996081636044825
8380 0.094316072691974
8390 0.085892590005288
8400 0.113334306352186
8410 0.108449739142387
8420 0.0961329668013515
8430 0.0848942705463503
8440 0.08637747430041
8450 0.085142073652298
8460 0.0756074505930092
8470 0.0772364745677918
8480 0.08211863060695
8490 0.0954585592859377
8500 0.100733141771794
8510 0.088022494970732
8520 0.111439655775925
8530 0.112116869737969
8540 0.0953705034760037
8550 0.0996194667568483
8560 0.108663674051607
8570 0.0897945394246979
8580 0.129894501172296
8590 0.117700492724753
8600 0.0856027854488924
8610 0.078583186002778
8620 0.0948443755903648
8630 0.0883207251999531
8640 0.0903317677521782
8650 0.0824327864916011
8660 0.0877009025215952
8670 0.0877450940915898
8680 0.0855448306447694
8690 0.121274681195856
8700 0.155922460846209
8710 0.110509926485885
8720 0.0874435038738204
8730 0.0868714588570778
8740 0.109568226698065
8750 0.0911114558067109
8760 0.100048743662026
8770 0.0829397988282407
8780 0.0789760313512773
8790 0.0765733635802305
8800 0.101158266557599
8810 0.0969160656158554
8820 0.0763601623538583
8830 0.092319272523292
8840 0.100201051436616
8850 0.0875724595852313
8860 0.122826450727572
8870 0.0965620223623934
8880 0.155601374359106
8890 0.154695721904051
8900 0.112943923706958
8910 0.122087020224772
8920 0.121381154896392
8930 0.137549353616502
8940 0.126602645643553
8950 0.0882431387233924
8960 0.0622054130458411
8970 0.0669597564046349
8980 0.0759724900372465
8990 0.0906652398639962
9000 0.0942935511808034
9010 0.0731183756086994
9020 0.0678439534671341
9030 0.076973583124316
9040 0.12498555728156
9050 0.125084243004395
9060 0.109361685694919
9070 0.0816821801641864
9080 0.0855672041813391
9090 0.0954582131708058
9100 0.133690744089743
9110 0.157625803465211
9120 0.159518839923205
9130 0.179571192278651
9140 0.157870531083217
9150 0.104464096520853
9160 0.127094289934984
9170 0.12422463494104
9180 0.156268886605724
9190 0.161956160276412
9200 0.132784576641429
9210 0.131245835041955
9220 0.141804060010177
9230 0.115906575998526
9240 0.0851783964411699
9250 0.0957302977719138
9260 0.0881216642424958
9270 0.0903869300912588
9280 0.0777430927733744
9290 0.124813655367993
9300 0.134946473438341
9310 0.128392688974739
9320 0.103275030120769
9330 0.0996768755037691
9340 0.0873819549570908
9350 0.0887562989363485
9360 0.0930037551541715
};
\addlegendentry{ratio}
\end{axis}

\end{tikzpicture}

%% file: labpal/figure_training_processes_palsgd.tex
\begin{figure}[b!]
	\tikzsetfigurename{labpal_training_process_resnet_20_palsgd}
	\vspace{-0.3cm}
	\centering
	\newcommand\picscale{0.4}
	\newcommand\lossmin{0}
	\newcommand\lossmax{2.3}
	\newcommand\evalmin{0.5}
	\newcommand\evalmax{0.75}
	\newcommand\epochs{280}
	\newcommand\picheight{0.5\linewidth}
	\newcommand\legendwidth{9cm}
	\newcommand\legendheight{3cm}
	%CIFAR-10:
		\begin{tikzpicture}[scale=\picscale] % ccoordinate scale , does not affect text
		\begin{axis}[
			width=\linewidth, % Scale the plot to \linewidth
			height=\picheight,
			grid=major, % Display a grid
			grid style={dashed,gray!30}, % Set the style, transperancz
			xlabel= training step $\cdot 10^{3}$, % Set the labels
			xlabel style={font=\LARGE},
			ylabel= train. loss,
			ylabel style={font=\LARGE},
			xmin=0,
			xmax=320,
			ymin=\lossmin,ymax=\lossmax,
			x coord trafo/.code={
				\pgflibraryfpuifactive{
					\pgfmathparse{(#1)*(31.25)*(0.001)}
				}{
					\pgfkeys{/pgf/fpu=true}
					\pgfmathparse{(#1)*(31.25)*(0.001)}
					\pgfkeys{/pgf/fpu=false}
				}
			},
			ymode=log,
			%xtick={-10,...,10},
			%ytick={-10,...,10},
			%x unit=\si{\volt}, % Set the respective units
			%y unit=\si{\ampere},
			%legend style={at={(-0.7,1)},anchor=north,font=\huge,minimum height= \legendheight,draw=none}, % Put the legend left of the plot
			legend style={font=\LARGE,fill=none}, % Put the legend left of the plot
			legend pos=south west, % Put the legend left of the plot
			legend cell align={left},
			x tick label style={rotate=0,anchor=near xticklabel,font=\LARGE}, % Display labels sideways
			y tick label style={font=\LARGE,fill opacity=0.8},
			p1/.style={draw=green!50.1960784313725!black,line width=2pt},
			p2/.style={draw=green!25!black,line width=2pt},
			%p2/.style={draw=red,line width=2pt},
			p3/.style={draw=green,line width=2pt},
			p4/.style={draw=red,line width=2pt},
			%p5/.style={draw=black,line width=2pt},
			p6/.style={draw=blue,line width=2pt},
			p7/.style={draw=green,line width=2pt},
			%title= training loss,
			title style={font=\LARGE},
			]
			\addplot [p1] table[x=Step,y=Value,col sep=comma] {labpal/figure_data/figure1/train_loss00.csv};
			%\addplot [p7] table[x=Step,y=Value,col sep=comma] {figure_data/figure1/train_loss09.csv};
			\addplot [p3] table[x=Step,y=Value,col sep=comma] {labpal/figure_data/figure1/train_loss_decay.csv};
			\addplot [p2] table[x=Step,y=Value,col sep=comma] {labpal/figure_data/figure1/sgd_mom_train_loss.csv};
			\addplot [p4] table[x=Step,y=Value,col sep=comma] {labpal/figure_data/figure1/palnsgd_train_loss1_8.csv};
			%\addplot [p5] table[x=Step,y=Value,col sep=comma] {figure_data/figure1/palsgd_train_loss1_0.csv};
			\addplot [p6] table[x=Step,y=Value,col sep=comma] {labpal/figure_data/figure1/palsgd_train_loss1_8.csv}; 
			\legend{SGD $\lambda=0.1$,SGD $\lambda=0.1$ decay,SGD $\lambda=0.1\beta=0.9$ decay, LABPAL\&NSGD, LABPAL\&SGD}
		\end{axis}
	\end{tikzpicture}\quad
	\begin{tikzpicture}[scale=\picscale] % ccoordinate scale , does not affect text
		\begin{axis}[
			width=\linewidth, % Scale the plot to \linewidth
			height=\picheight,
			grid=major, % Display a grid
			grid style={dashed,gray!30}, % Set the style, transperancz
			xlabel=training step $\cdot 10^{3}$, % Set the labels
			xlabel style={font=\LARGE},
			ylabel= val. accuracy,
			ylabel style={font=\LARGE},
			xmin=0,xmax=320,
			ymin=\evalmin,
			ymax=0.77,
			scaled x ticks=true,
			x coord trafo/.code={
				\pgflibraryfpuifactive{
					\pgfmathparse{(#1)*(31.25)*(0.001)}
				}{
					\pgfkeys{/pgf/fpu=true}
					\pgfmathparse{(#1)*(31.25)*(0.001)}
					\pgfkeys{/pgf/fpu=false}
				}
			},
			%xtick={-10,...,10},
			%ytick={-10,...,10},
			%x unit=\si{\volt}, % Set the respective units
			%y unit=\si{\ampere},
			%legend style={at={(-0.7,1)},anchor=north,font=\huge,minimum height= \legendheight,draw=none}, % Put the legend left of the plot
			%legend style={anchor=south west,at={(1,0.2)}}, % Put the legend left of the plot
			legend pos=north west, % Put the legend left of the plot
			x tick label style={rotate=0,anchor=near xticklabel,font=\LARGE}, % Display labels sideways
			y tick label style={font=\LARGE},
			p1/.style={draw=green!50.1960784313725!black,line width=2pt},
			p2/.style={draw=green!25!black,line width=2pt},
			p3/.style={draw=green,line width=2pt},
			p4/.style={draw=red,line width=2pt},
			%p5/.style={draw=black,line width=2pt},
			p6/.style={draw=blue,line width=2pt},
			%p7/.style={draw=green,line width=2pt},
			%title= validation accuracy,
			title style={font=\LARGE},
			]
			\addplot [p1] table[x=Step,y=Value,col sep=comma] {labpal/figure_data/figure1/val_acc00.csv};
			%\addplot [p2] table[x=Step,y=Value,col sep=comma] {figure_data/figure1/val_acc00_0,05.csv};
			%\addplot [p7] table[x=Step,y=Value,col sep=comma] {figure_data/figure1/val_acc09.csv};
			\addplot [p3] table[x=Step,y=Value,col sep=comma] {labpal/figure_data/figure1/val_acc_decay.csv};
			\addplot [p2] table[x=Step,y=Value,col sep=comma] {labpal/figure_data/figure1/sgd_mom_val_acc.csv};
			\addplot [p4] table[x=Step,y=Value,col sep=comma] {labpal/figure_data/figure1/palnsgd_val_acc_1_8.csv};
			%\addplot [p5] table[x=Step,y=Value,col sep=comma] {figure_data/figure1/palsgd_val_acc_1_0.csv};
			\addplot [p6] table[x=Step,y=Value,col sep=comma] {labpal/figure_data/figure1/palsgd_val_acc_1_8.csv}; 
			%\legend{SGD $\lambda=0.1$,SGD $\lambda=0.05$, MBPAL\&NSGD $\alpha=1.0$, MBPAL\&NSGD $\alpha=1.25$, MBPAL\&SGD $\alpha=1.0$, MBPAL\&SGD $\alpha=1.25$ }
		\end{axis}
	\end{tikzpicture}\\
	\begin{tikzpicture}[scale=\picscale] % ccoordinate scale , does not affect text
	\begin{axis}[
		width=\linewidth, % Scale the plot to \linewidth
		height=\picheight,
		ymode=log,
		grid=major, % Display a grid
		grid style={dashed,gray!30}, % Set the style, transperancz
		xlabel=training step $\cdot 10^{3}$, % Set the labels
		xlabel style={font=\LARGE},
		ylabel= learning rate,
		ylabel style={font=\LARGE},
		xmin=0,xmax=10000,
		scaled x ticks=true,
					x coord trafo/.code={
			\pgflibraryfpuifactive{
				\pgfmathparse{(#1)*(0.001)}
			}{
				\pgfkeys{/pgf/fpu=true}
				\pgfmathparse{(#1)*(0.001)}
				\pgfkeys{/pgf/fpu=false}
			}
		},
		%xtick={-10,...,10},
		%ytick={-10,...,10},
		%x unit=\si{\volt}, % Set the respective units
		%y unit=\si{\ampere},
		%legend style={at={(-0.7,1)},anchor=north,font=\huge,minimum height= \legendheight,draw=none}, % Put the legend left of the plot
		%legend style={anchor=south west,at={(1,0.2)}}, % Put the legend left of the plot
		legend pos=north west, % Put the legend left of the plot
		x tick label style={rotate=0,anchor=near xticklabel,font=\LARGE}, % Display labels sideways
		y tick label style={font=\LARGE},
		p1/.style={draw=green!50.1960784313725!black,line width=2pt},
		p2/.style={draw=green!25!black,line width=2pt},
		p3/.style={draw=green,line width=2pt},
		p4/.style={draw=red,line width=2pt},
		%p5/.style={draw=black,line width=2pt},
		p6/.style={draw=blue,line width=2pt},
		p7/.style={draw=green,line width=2pt},
		%title= learning rates,
		title style={font=\LARGE},
		]
		\addplot [p1] table[x=Step,y=Value,col sep=comma] {labpal/figure_data/figure1/lr00_0_1.csv};
		%\addplot [p2] table[x=Step,y=Value,col sep=comma] {figure_data/figure1/lr00_0_05.csv};
		%\addplot [p7] table[x=Step,y=Value,col sep=comma] {figure_data/figure1/lr00_0_1.csv};
		%\addplot [p3] table[x=Step,y=Value,col sep=comma] {labpal/figure_data/figure1/lr_decay.csv};
		\addplot [p2] table[x=Step,y=Value,col sep=comma] {labpal/figure_data/figure1/sgd_mom_lr.csv};
		\addplot [p4] table[x=Step,y=Value,col sep=comma] {labpal/figure_data/figure1/palnsgd_lr_1_8.csv};
		%\addplot [p5] table[x=Step,y=Value,col sep=comma] {figure_data/figure1/palsgd_lr_1_0.csv};
		\addplot [p6] table[x=Step,y=Value,col sep=comma] {labpal/figure_data/figure1/palsgd_lr_1_8.csv}; 
		%\legend{SGD $\lambda=0.1$,SGD $\lambda=0.05$, MBPAL\&NSGD $\alpha=1.0$, MBPAL\&NSGD $\alpha=1.25$, MBPAL\&SGD $\alpha=1.0$, MBPAL\&SGD $\alpha=1.25$ }
	\end{axis}
\end{tikzpicture}\quad
	\begin{tikzpicture}[scale=\picscale] % ccoordinate scale , does not affect text
	\begin{axis}[
		width=\linewidth, % Scale the plot to \linewidth
		height=\picheight,
		grid=major, % Display a grid
		grid style={dashed,gray!30}, % Set the style, transperancz
		xlabel=training step $\cdot 10^{3}$, % Set the labels
		xlabel style={font=\LARGE},
		ylabel= update step,
		ylabel style={font=\LARGE},
		xmin=0,xmax=10000,
		ymax=6.0,
		scaled x ticks=true,
				scaled x ticks=true,
		x coord trafo/.code={
			\pgflibraryfpuifactive{
				\pgfmathparse{(#1)*(0.001)}
			}{
				\pgfkeys{/pgf/fpu=true}
				\pgfmathparse{(#1)*(0.001)}
				\pgfkeys{/pgf/fpu=false}
			}
		},
		%xtick={-10,...,10},
		%ytick={-10,...,10},
		%x unit=\si{\volt}, % Set the respective units
		%y unit=\si{\ampere},
		%legend style={at={(-0.7,1)},anchor=north,font=\huge,minimum height= \legendheight,draw=none}, % Put the legend left of the plot
		%legend style={anchor=south west,at={(1,0.2)}}, % Put the legend left of the plot
		legend pos=north west, % Put the legend left of the plot
		x tick label style={rotate=0,anchor=near xticklabel,font=\LARGE}, % Display labels sideways
		y tick label style={font=\LARGE},
		p1/.style={draw=green!50.1960784313725!black,line width=2pt},
		p2/.style={draw=green!25!black,line width=2pt},
		p3/.style={draw=green,line width=2pt},
		p4/.style={draw=red,line width=2pt},
		%p5/.style={draw=black,line width=2pt},
		p6/.style={draw=blue,line width=2pt},
		p7/.style={draw=green,line width=2pt},
		%title= update steps (lr $\cdot$ gradient norm),
		title style={font=\LARGE},
		]
		\addplot [p1] table[x=Step,y=Value,col sep=comma] {labpal/figure_data/figure1/update_step_00_0_01.csv};
		\addplot [p2] table[x=Step,y=Value,col sep=comma] {labpal/figure_data/figure1/update_step_00_0_05.csv};
		%\addplot [p7] table[x=Step,y=Value,col sep=comma] {figure_data/figure1/update_step_09_0_01.csv};
		%\addplot [p5] table[x=Step,y=Value,col sep=comma] {figure_data/figure1/palsgd_upd_step_1_0.csv};
		\addplot [p6] table[x=Step,y=Value,col sep=comma] {labpal/figure_data/figure1/palsgd_upd_step_1_8.csv}; 
		\addplot [p3] table[x=Step,y=Value,col sep=comma] {labpal/figure_data/figure1/update_step_decay.csv};
		%\addplot [p2] table[x=Step,y=Value,col sep=comma] {figure_data/figure1/sgd_mom_step.csv};
		\addplot [p4] table[x=Step,y=Value,col sep=comma] {labpal/figure_data/figure1/palnsgd_upd_step_1_8.csv};

		%\legend{SGD $\lambda=0.1$,SGD $\lambda=0.05$, MBPAL\&NSGD $\alpha=1.0$, MBPAL\&NSGD $\alpha=1.25$, MBPAL\&SGD $\alpha=1.0$, MBPAL\&SGD $\alpha=1.25$ }
	\end{axis}
\end{tikzpicture}

	\caption{Training process on the problem of which the empirical observations were inferred ( ResNet-20 trained on 8\% of CIFAR-10 with SGD). LABPAL\&NSGD and LABPAL\&SGD outperform SGD. Interestingly LABPAL\&NSGD  estimated huge $\lambda$s, whereas $s_{upd}$s are decreasing}
	\label{labpal_fig_training_process_resnet-20_palsgd}
\end{figure}

%% file: labpal/performance_comparison_cifar100.tex
\begin{figure}[b!]
	\tikzsetfigurename{labpal_optimizer_comparison_cifar100}
	\vspace{-0.3cm}
	\centering
	\def\scale{0.4}
	\begin{tabular}{ c c c}	
		\hspace{-0.2cm}\scalebox{\scale}{\input{"labpal/figure_data/performance_comparison/CIFAR-100/CIFAR-100_ResNet-20_learning_rate.pgf"}}&	\hspace{-0.2cm} 
		\scalebox{\scale}{\input{"labpal/figure_data/performance_comparison/CIFAR-100/CIFAR-100_MobileNet-V2_learning_rate.pgf"}}& 	\hspace{-0.2cm} 
		\scalebox{\scale}{\input{"labpal/figure_data/performance_comparison/CIFAR-100/CIFAR-100_DenseNet-121_learning_rate.pgf"}}\\
		\hspace{-0.2cm}\scalebox{\scale}{\input{"labpal/figure_data/performance_comparison/CIFAR-100/CIFAR-100_ResNet-20_training_loss.pgf"}}&\hspace{-0.2cm}
		\scalebox{\scale}{\input{"labpal/figure_data/performance_comparison/CIFAR-100/CIFAR-100_MobileNet-V2_training_loss.pgf"}}&\hspace{-0.2cm}
		\hspace{-0.2cm}\scalebox{\scale}{\input{"labpal/figure_data/performance_comparison/CIFAR-100/CIFAR-100_DenseNet-121_training_loss.pgf"}}\\
		\scalebox{\scale}{\input{"labpal/figure_data/performance_comparison/CIFAR-100/CIFAR-100_ResNet-20_validation_accuracy.pgf"}}&\hspace{-0.2cm}
		\scalebox{\scale}{\input{"labpal/figure_data/performance_comparison/CIFAR-100/CIFAR-100_MobileNet-V2_validation_accuracy.pgf"}}&\hspace{-0.2cm}
		\hspace{-0.2cm}\scalebox{\scale}{\input{"labpal/figure_data/performance_comparison/CIFAR-100/CIFAR-100_DenseNet-121_validation_accuracy.pgf"}}\\	
		\scalebox{\scale}{\input{"labpal/figure_data/performance_comparison/CIFAR-100/CIFAR-100_ResNet-20_test_accuracy.pgf"}}&\hspace{-0.2cm}	
		\scalebox{\scale}{\input{"labpal/figure_data/performance_comparison/CIFAR-100/CIFAR-100_MobileNet-V2_test_accuracy.pgf"}}&\hspace{-0.2cm}	
		\scalebox{\scale}{\input{"labpal/figure_data/performance_comparison/CIFAR-100/CIFAR-100_DenseNet-121_test_accuracy.pgf"}}
	\end{tabular}
	\caption{Performance comparison on \textbf{CIFAR-100} of our approach LABPAL in the SGD and NSGD variants against several line searches and SGD with momentum. Optimal hyperparameters for CIFAR-10 ResNet-20 found with a grid search are reused (Appendix \ref{labpal_app_hyperparams}). Here, our approaches surpass the other approaches on training loss, validation, and test accuracy. Columns indicate different models. Rows indicate different metrics.  Results for CIFAR-10 and SVHN are given in appendix Figures \ref{labpal_fig_optimizer_comparison_cifar10}, and \ref{labpal_fig_optimizer_comparison_svhn}. The batch-size used is $128$. Due to a too high memory consumption we could not run PLS on MobileNet-V2. } 
	\label{labpal_fig_optimizer_comparison_cifar100}
\end{figure}

%% file: labpal/figure_data/performance_comparison/CIFAR-100/CIFAR-100_DenseNet-121_training_loss.pgf
% This file was created by tikzplotlib v0.9.8.
\begin{tikzpicture}

\definecolor{color0}{rgb}{0.647058823529412,0.164705882352941,0.164705882352941}
\definecolor{color1}{rgb}{0.933333333333333,0.509803921568627,0.933333333333333}
\definecolor{color2}{rgb}{1,0.647058823529412,0}

\begin{axis}[
yticklabel style={font=\large},
xticklabel style={font=\large},
ytick style={font=\Large},
xtick style={font=\Large},
ylabel style={font=\Large},
xlabel style={font=\Large},
title style={font=\Large},
grid = major,
major grid style={dotted},
log basis y={10},
minor xtick={},
minor ytick={0.0002,0.0003,0.0004,0.0005,0.0006,0.0007,0.0008,0.0009,0.002,0.003,0.004,0.005,0.006,0.007,0.008,0.009,0.02,0.03,0.04,0.05,0.06,0.07,0.08,0.09,0.2,0.3,0.4,0.5,0.6,0.7,0.8,0.9,2,3,4,5,6,7,8,9,20,30,40,50,60,70,80,90,200,300,400,500,600,700,800,900},
tick align=outside,
tick pos=left,
title={training loss CIFAR-100 DenseNet-121},
width=10.5cm,height=8cm,,
x grid style={white!69.0196078431373!black},
xlabel={epochs},
xmin=-14.2, xmax=298.2,
xtick style={color=black},
xtick={-50,0,50,100,150,200,250,300},
y grid style={white!69.0196078431373!black},
ylabel={training loss},
ymin=0.001, ymax=6.5,
ymode=log,
ytick style={color=black},
ytick={0.0001,0.001,0.01,0.1,1,10,100}
]
\addplot [line width=2.5pt, color0, opacity=1.0]
table {%
0 4.64078617095947
1 3.9434445699056
2 3.37345870335897
3 3.03932404518127
4 2.78901124000549
5 2.58772500356038
6 2.41139729817708
7 2.25596602757772
8 2.11948696772257
9 2.00041254361471
10 1.88955855369568
11 1.80027532577515
12 1.70765821139018
13 1.62998517354329
14 1.56324390570323
15 1.49085239569346
16 1.42670667171478
17 1.36725191275279
18 1.31794635454814
19 1.26676265398661
20 1.21254829565684
21 1.16219210624695
22 1.12156542142232
23 1.07011457284292
24 1.03495216369629
25 0.996322850386302
26 0.951188683509827
27 0.916031221548716
28 0.879382769266764
29 0.844218691190084
30 0.804973661899567
31 0.777230123678843
32 0.744409143924713
33 0.714467465877533
34 0.686062713464101
35 0.659921129544576
36 0.630391935507456
37 0.60122287273407
38 0.582013150056203
39 0.556438823541005
40 0.530074020226797
41 0.509582857290904
42 0.487969507773717
43 0.464256982008616
44 0.444502085447311
45 0.424716105063756
46 0.412039895852407
47 0.393963009119034
48 0.372929642597834
49 0.363521188497543
50 0.347328394651413
51 0.328213234742482
52 0.315528084834417
53 0.307584722836812
54 0.297729730606079
55 0.284566193819046
56 0.269375244776408
57 0.266481031974157
58 0.254192709922791
59 0.237902874747912
60 0.235280215740204
61 0.222758357723554
62 0.214316820104917
63 0.215715318918228
64 0.204850400487582
65 0.201068326830864
66 0.191026161114375
67 0.184202974041303
68 0.17618790268898
69 0.16951351861159
70 0.172311807672183
71 0.161257227261861
72 0.157424891988436
73 0.153713345527649
74 0.150330012043317
75 0.139304627974828
76 0.138760695854823
77 0.136647214492162
78 0.135515491167704
79 0.134810820221901
80 0.128076722224553
81 0.124043829739094
82 0.126220464706421
83 0.116767697036266
84 0.117481591800849
85 0.114511276284854
86 0.108488894999027
87 0.107815568645795
88 0.105396710336208
89 0.103014945983887
90 0.104547863205274
91 0.0951946750283241
92 0.102386285861333
93 0.0944345916310946
94 0.0928066000342369
95 0.110853865742683
96 0.114355062445005
97 0.113152871529261
98 0.109829562405745
99 0.111559249460697
100 0.103711637357871
101 0.0963243941466014
102 0.0975342219074567
103 0.0938062220811844
104 0.0896558736761411
105 0.0892367785175641
106 0.0891612122456233
107 0.12280132373174
108 0.122044128676256
109 0.117525105675062
110 0.117141790688038
111 0.110776163637638
112 0.106866538524628
113 0.10103011627992
114 0.0977376972635587
115 0.0981762396792571
116 0.0939177672068278
117 0.0910751298069954
118 0.0866602336366971
119 0.0799926395217578
120 0.0822060691813628
121 0.0765038703878721
122 0.0794614938398202
123 0.0727023271222909
124 0.0734462589025497
125 0.0733308494091034
126 0.0686664034922918
127 0.0744283919533094
128 0.0860101977984111
129 0.092261366546154
130 0.0916663594543934
131 0.0847827245791753
132 0.0874561804036299
133 0.0791849506398042
134 0.0786988586187363
135 0.0775394154091676
136 0.0749195925891399
137 0.0728865526616573
138 0.0719216763973236
139 0.0643871786693732
140 0.0951989516615868
141 0.181625415881475
142 0.204791225492954
143 0.162212443848451
144 0.133617614706357
145 0.117659876743952
146 0.107106796155373
147 0.099464874714613
148 0.0944397101799647
149 0.10229487468799
150 0.109394753972689
151 0.099633543441693
152 0.0978336681922277
153 0.0976140350103378
154 0.091512031853199
155 0.0929209279517333
156 0.0872375356654326
157 0.0811990362902482
158 0.0809379853308201
159 0.0794441240529219
160 0.0807348340749741
161 0.0748758651316166
162 0.069346085190773
163 0.0662442843119303
164 0.0661748598019282
165 0.0606977976858616
166 0.0627648967007796
167 0.0666095154980818
168 0.0779013161857923
169 0.0828075408935547
170 0.0807420151929061
171 0.0767441838979721
172 0.0746907206873099
173 0.0722410629192988
174 0.0697016815344493
175 0.0702229092518489
176 0.15300665423274
177 0.182787425816059
178 0.166292381783326
179 0.140303902328014
180 0.126047966380914
181 0.111656426141659
182 0.0975710662702719
183 0.0963445814947287
184 0.0862399265170097
185 0.0874915309250355
186 0.0851810611784458
187 0.0837995707988739
188 0.0816011540591717
189 0.0772477984428406
190 0.0732898476223151
191 0.0689252230028311
192 0.0686942401031653
193 0.0650112169484297
194 0.0651382779081662
195 0.0647852669159571
196 0.0603408689300219
197 0.0590828234950701
198 0.0598276642461618
199 0.0606444577376048
200 0.0582535589734713
201 0.0605306513607502
202 0.0547003112733364
203 0.0508968730767568
204 0.0514591659108798
205 0.069016261647145
206 0.0841022903720538
207 0.0888837923606237
208 0.0875844744344552
209 0.0742603515585264
210 0.0758051089942455
211 0.088465033719937
212 0.090495182822148
213 0.0850139707326889
214 0.0805188044905663
215 0.0725812713305155
216 0.0618084507683913
217 0.0675864753623803
218 0.0751542473832766
219 0.0786771389345328
220 0.0805913445850213
221 0.102082774043083
222 0.109474703669548
223 0.107252719501654
224 0.112229863802592
225 0.100866032143434
226 0.0986257257560889
227 0.0910572111606598
228 0.0862345124284426
229 0.0817765332758427
230 0.0762076117098331
231 0.0717191062867641
232 0.0657631432016691
233 0.0646887458860874
234 0.0705904004474481
235 0.067957499374946
236 0.0648660324513912
237 0.056079442302386
238 0.0561448335647583
239 0.0691646176079909
240 0.0884991064667702
241 0.107804505775372
242 0.150518926481406
243 0.229495391249657
244 0.215875272949537
245 0.184401233990987
246 0.157256089150906
247 0.134275443851948
248 0.125288546085358
249 0.122944451868534
250 0.104966096580029
251 0.105931123097738
252 0.0980308229724566
253 0.0923344567418098
254 0.0853292221824328
255 0.0818522721529007
256 0.0850482558210691
257 0.0838839809099833
258 0.0813161432743073
259 0.0828324345250924
260 0.0701586765547593
261 0.0681913979351521
262 0.0646872520446777
263 0.0821739211678505
264 0.104810006916523
265 0.102123382190863
266 0.0979065547386805
267 0.1030098721385
268 0.091030025233825
269 0.0820905628303687
270 0.0768061615526676
271 0.0760455578565598
272 0.073215081046025
273 0.0728924783567588
274 0.0631364087263743
275 0.0660044985512892
276 0.0625207585593065
277 0.0569941935439905
278 0.0582497281332811
279 0.0689409387608369
280 0.0765401969353358
281 0.0870778163274129
282 0.075312419484059
283 0.0764474508663018
284 0.0768784297009309
};
\addplot [line width=2.5pt, green!50.1960784313725!black, opacity=1.0]
table {%
0 4.64078617095947
1 3.87370737393697
2 3.17594186464945
3 2.78305729230245
4 2.51665385564168
5 2.26691826184591
6 2.06787570317586
7 1.90442256132762
8 1.76851014296214
9 1.6523562669754
10 1.53795226414998
11 1.44500736395518
12 1.35447005430857
13 1.27573319276174
14 1.20289198557536
15 1.13279207547506
16 1.0695804754893
17 1.00517278909683
18 0.948447406291962
19 0.893277664979299
20 0.838516116142273
21 0.794322490692139
22 0.755182445049286
23 0.698999206225077
24 0.662216107050578
25 0.629290560881297
26 0.582044621308645
27 0.555406391620636
28 0.522682686646779
29 0.486907929182053
30 0.462819943825404
31 0.433370471000671
32 0.40493697921435
33 0.381354848543803
34 0.365110029776891
35 0.345796843369802
36 0.330526252587636
37 0.310121804475784
38 0.291730533043544
39 0.274632960557938
40 0.259660730759303
41 0.252895136674245
42 0.239122405648232
43 0.229291478792826
44 0.218738595644633
45 0.20937517285347
46 0.195328027009964
47 0.192847559849421
48 0.182628278930982
49 0.172845462958018
50 0.170291046301524
51 0.157264307141304
52 0.155564318100611
53 0.156537815928459
54 0.145585884650548
55 0.138364921013514
56 0.13315453628699
57 0.125976391136646
58 0.120250637332598
59 0.119522010286649
60 0.118636680146058
61 0.113510037461917
62 0.113618552684784
63 0.110341568787893
64 0.106271920104822
65 0.0975656087199847
66 0.100489899516106
67 0.094848578174909
68 0.094611165424188
69 0.092450184126695
70 0.0891375268499057
71 0.0864556555946668
72 0.0846276755134265
73 0.0802592908342679
74 0.0751560752590497
75 0.0759070639808973
76 0.0769487793246905
77 0.0732252771655719
78 0.0725373650590579
79 0.0712208126982053
80 0.0668982019027074
81 0.0661707247296969
82 0.0660631010929743
83 0.0654898037513097
84 0.0653350651264191
85 0.0637663876016935
86 0.062585794677337
87 0.0630903666218122
88 0.0602504064639409
89 0.0600483479599158
90 0.0584108990927537
91 0.0544921693702539
92 0.0558276871840159
93 0.0576656349003315
94 0.0527738916377226
95 0.0491302534937859
96 0.0542960862318675
97 0.0503118087848028
98 0.0501502727468809
99 0.0501697510480881
100 0.0477829438944658
101 0.0451791534821192
102 0.0423533370097478
103 0.0449437461793423
104 0.0435427092015743
105 0.0426515576740106
106 0.0427975977460543
107 0.041874885559082
108 0.0394768168528875
109 0.0410652073721091
110 0.0400587283074856
111 0.0375805447498957
112 0.0364965870976448
113 0.0371214784681797
114 0.0392542307575544
115 0.0382901343206565
116 0.0392993502318859
117 0.0380348873635133
118 0.0378811632593473
119 0.0348074473440647
120 0.0343204761544863
121 0.0342202689498663
122 0.0348733340700467
123 0.0308725262681643
124 0.0309908023724953
125 0.0308708529919386
126 0.0299485238889853
127 0.0297178036222855
128 0.0305810905992985
129 0.0295218781878551
130 0.0291409573207299
131 0.0294220875948668
132 0.0304899290204048
133 0.0306999223927657
134 0.0299370723466078
135 0.028467978661259
136 0.0273557764788469
137 0.0266721037526925
138 0.0275004909684261
139 0.0276787367959817
140 0.0293493686864773
141 0.0277544111013412
142 0.026507871846358
143 0.0247018386920293
144 0.0144527908414602
145 0.0121192817265789
146 0.00946076090137164
147 0.00923031413306793
148 0.00802644559492668
149 0.0078542932557563
150 0.00677177822217345
151 0.00699053332209587
152 0.00680017812798421
153 0.00624938402324915
154 0.00605656625702977
155 0.00586476549506187
156 0.00510994779566924
157 0.0056799273006618
158 0.00533380747462312
159 0.00536053208634257
160 0.00506522149468462
161 0.00526726820195715
162 0.00477408726389209
163 0.00492941044891874
164 0.00464638586466511
165 0.00447195349261165
166 0.00459239492192864
167 0.0041344358275334
168 0.004471596951286
169 0.00432833128919204
170 0.00436163297854364
171 0.00433972881486018
172 0.00401099181423585
173 0.00401863416967293
174 0.00379961860987047
175 0.00384697830304503
176 0.00382689929877718
177 0.00371966472206016
178 0.00387084390968084
179 0.00359112645188967
180 0.00369461175675193
181 0.00365552857207755
182 0.00368392180340985
183 0.00366845067280034
184 0.00349025436056157
185 0.00333202094770968
186 0.00335438479669392
187 0.003273645726343
188 0.00329200931203862
189 0.00310640246607363
190 0.00346372975036502
191 0.00329830714811881
192 0.00334459907996158
193 0.00338911769601206
194 0.00318533461540937
195 0.00285543571226299
196 0.00294597509006659
197 0.00303869337464372
198 0.00304600495534639
199 0.00296380215634902
200 0.00312500547928115
201 0.00282304299374421
202 0.00276365938285987
203 0.00277512311004102
204 0.00289603128718833
205 0.00291418749839067
206 0.00318390619941056
207 0.0028656932990998
208 0.00274958283019563
209 0.00279914673107366
210 0.00256082221555213
211 0.00259967350090543
212 0.00275427685119212
213 0.00279342033900321
214 0.0026496258409073
215 0.0025150622241199
216 0.00244624749757349
217 0.00244935958956679
218 0.00241624362145861
219 0.00262497831135988
220 0.00255863508209586
221 0.00246360556532939
222 0.00257790818189581
223 0.00242063123732805
224 0.00276350493853291
225 0.0026522025000304
226 0.00265104222732286
227 0.00253029920471211
228 0.00266435548352698
229 0.00249109293023745
230 0.00260115026806792
231 0.00250782794319093
232 0.00252551080969473
233 0.00236421202619871
234 0.00252879364416003
235 0.00247109619279703
236 0.00254305979857842
237 0.00247059172640244
238 0.00257776374928653
239 0.00239425431936979
240 0.00236769020557404
241 0.00244602677412331
242 0.00266067955332498
243 0.00248462489495675
244 0.0023576773237437
245 0.00233108946122229
246 0.0027041716966778
247 0.00245548031913737
248 0.00239161960780621
249 0.00251152777733902
250 0.00250214012339711
251 0.0024825306609273
252 0.00257897757304211
253 0.00248874723911285
254 0.00238445385669669
255 0.00223848937700192
256 0.00236972553345064
257 0.00262397419040402
258 0.00240224118654927
259 0.00240628056538602
260 0.00254867970943451
261 0.00232064188458025
262 0.00245510235739251
263 0.00255400609845916
264 0.00230075667301814
265 0.00209368579089642
266 0.00246255984529853
267 0.00251603599948188
268 0.00262784065368275
269 0.00254951836541295
270 0.00228576618246734
271 0.00238012258584301
272 0.00250447130141159
273 0.00247777416370809
274 0.00265769784649213
275 0.00208514036300282
276 0.00242564136472841
277 0.00225283081332842
278 0.00235081277787685
279 0.00241366222811242
280 0.00244073189484576
281 0.00258353515528142
282 0.00233880539114277
283 0.00224327160200725
284 0.00262581409576039
};
\addplot [line width=2.5pt, black, opacity=1.0]
table {%
0 4.64078617095947
1 4.16854588190715
2 3.70032540957133
3 3.48739870389303
4 3.36798477172852
5 3.29790115356445
6 3.16434542338053
7 2.97040589650472
8 2.78154706954956
9 2.63589072227478
10 2.54670834541321
11 2.49384053548177
12 2.48733814557393
13 2.53607002894084
14 2.69875685373942
15 3.10837443669637
16 3.65894174575806
17 3.8342604637146
18 3.36612319946289
19 3.45093925793966
20 3.53074272473653
21 3.34082349141439
22 3.01511907577515
23 2.88260189692179
24 2.85940631230672
25 2.82711370786031
26 2.91411519050598
27 2.87758143742879
28 2.79544194539388
29 2.69144455591838
30 2.67716447512309
31 2.60281276702881
32 2.55867266654968
33 2.60959045092265
34 2.55109477043152
35 2.40948875745138
36 2.07642789681753
37 2.08863091468811
38 2.06375511487325
39 2.13385001818339
40 1.9166374206543
41 1.84149587154388
42 1.65680634975433
43 1.70820983250936
44 1.78387419382731
45 1.77963268756866
46 1.76825777689616
47 1.69198127587636
48 1.72266602516174
49 1.76848820845286
50 1.80698990821838
51 1.73403243223826
52 1.70912925402323
53 1.67083597183228
54 1.64085372289022
55 1.65141312281291
56 1.59425961971283
57 1.7063524723053
58 1.72115584214528
59 1.6202582915624
60 1.55462618668874
61 1.53490153948466
62 1.50988749663035
63 1.35616298516591
64 1.2898504336675
65 1.26799746354421
66 1.25286622842153
67 1.24474918842316
68 1.25156573454539
69 1.26802011330922
70 1.34083338578542
71 1.41018577416738
72 1.46743218104045
73 1.39799936612447
74 1.35231773058573
75 1.36541001001994
76 1.24772683779399
77 1.24378220240275
78 1.28425284226735
79 1.41295210520426
80 1.49159435431163
81 1.56676411628723
82 1.48971891403198
83 1.47582459449768
84 1.45694057146708
85 1.3934809366862
86 1.44389736652374
87 1.45182812213898
88 1.51288866996765
89 1.37186360359192
90 1.31124965349833
91 1.29654343922933
92 1.09728878736496
93 1.01835000514984
94 0.991322815418243
95 0.965241332848867
96 0.970262249310811
97 1.0109281539917
98 1.10971087217331
99 1.22397619485855
100 1.26101050774256
101 1.32377920548121
102 1.22651147842407
103 1.15276356538137
104 1.09555246432622
105 1.07103163003922
106 1.04075799385707
107 1.03297239542007
108 1.04599273204803
109 1.07765032847722
110 1.08216202259064
111 0.987544357776642
112 0.998863736788432
113 1.04723860820134
114 1.01826421419779
115 0.925485412279765
116 0.938605904579163
117 1.0134438474973
118 1.06072183450063
119 1.01428429285685
120 1.00821703672409
121 1.04319578409195
122 1.04237081607183
123 1.05489222208659
124 1.12700190146764
125 1.14828459421794
126 1.04140092929204
127 1.01969013611476
128 0.996570686499278
129 0.966399312019348
130 0.943961461385091
131 0.939096351464589
132 0.917073508103689
133 0.907214681307475
134 0.901623765627543
135 0.927195211251577
136 0.955148061116536
137 0.928125162919362
138 0.937407930692037
139 0.971639593442281
140 0.964671730995178
141 0.895141462484995
142 0.80158140261968
143 0.773453374703725
144 0.747324029604594
145 0.73331348101298
146 0.715671896934509
147 0.729370375474294
148 0.754641473293304
149 0.826486686865489
150 0.936057945092519
151 0.971822996934255
152 0.876407663027445
153 0.854041357835134
154 0.765349427858988
155 0.735881825288137
156 0.721597333749135
157 0.707875490188599
158 0.698373397191366
159 0.695290942986806
160 0.683479646841685
161 0.67990509668986
162 0.672700226306915
163 0.6773761510849
164 0.665636857350667
165 0.660127758979797
166 0.682162582874298
167 0.657750487327576
168 0.6299200852712
169 0.614659647146861
170 0.623033444086711
171 0.611369053522746
172 0.587774932384491
173 0.575717687606812
174 0.554295559724172
175 0.548842579126358
176 0.549727449814479
177 0.550914138555527
178 0.564429094394048
179 0.592968622843424
180 0.654318124055862
181 0.653536180655162
182 0.689834952354431
183 0.803228259086609
184 0.720951716105143
185 0.677853723367055
186 0.687874297300974
187 0.684733092784882
188 0.759320318698883
189 0.764562706152598
190 0.914235790570577
191 0.901038328806559
192 0.818882604440053
193 0.791353782018026
194 0.751139342784882
195 0.717383662859599
196 0.69233093659083
197 0.674737632274628
198 0.658148984114329
199 0.639734148979187
200 0.626793503761292
201 0.601870814959208
202 0.595559438069661
203 0.567580858866374
204 0.557990511258443
205 0.540596157312393
206 0.528567165136337
207 0.510373502969742
208 0.504516412814458
209 0.505262146393458
210 0.505775799353917
211 0.518425037463506
212 0.564789831638336
213 0.602881411711375
214 0.600341161092122
215 0.620163361231486
216 0.588915010293325
217 0.575747052828471
218 0.594404339790344
219 0.545626153548559
220 0.514559616645177
221 0.516734391450882
222 0.516066620747248
223 0.509803692499797
224 0.510581930478414
225 0.548867970705032
226 0.568269689877828
227 0.588299373785655
228 0.621570309003194
229 0.614162117242813
230 0.617666025956472
231 0.557515273491542
232 0.531458139419556
233 0.541830857594808
234 0.510547230641047
235 0.497056096792221
236 0.49250261982282
237 0.457223892211914
238 0.470573703447978
239 0.453408598899841
240 0.4480968217055
241 0.460418323675791
242 0.481714814901352
243 0.524081975221634
244 0.55565686027209
245 0.520408829053243
246 0.467055608828862
247 0.440031598011653
248 0.443423022826513
249 0.480115244785945
250 0.509515742460887
251 0.519051730632782
252 0.517552435398102
253 0.474440604448318
254 0.408342351516088
255 0.437093565861384
256 0.473856190840403
257 0.475377574563026
258 0.41333994269371
259 0.400644138455391
260 0.412369459867477
261 0.4580251177152
262 0.418464670578639
263 0.447457700967789
264 0.441297481457392
265 0.381517847379049
266 0.398879478375117
267 0.434952090183894
268 0.506789475679398
269 0.45797598361969
270 0.389475653568904
271 0.377266526222229
272 0.364760955174764
273 0.366250708699226
274 0.375644668936729
275 0.383250186840693
276 0.360693340500196
277 0.337788095076879
278 0.335057139396667
279 0.321994299689929
280 0.311770175894101
281 0.304017394781113
282 0.291828955213229
283 0.292764763037364
284 0.286294192075729
};
\addplot [line width=2.5pt, color1, opacity=1.0]
table {%
0 4.64078617095947
1 4.31787093480428
2 3.79490812619527
3 3.4923955599467
4 3.27567561467489
5 3.11357688903809
6 2.94388842582703
7 2.8278337319692
8 2.71911056836446
9 2.60397990544637
10 2.48792950312297
11 2.4185639222463
12 2.32019591331482
13 2.22726782162984
14 2.14739084243774
15 2.08098395665487
16 2.03089865048726
17 1.9598693450292
18 1.91578602790833
19 1.84308795134226
20 1.79983206590017
21 1.74957001209259
22 1.71121144294739
23 1.66476233800252
24 1.62944118181864
25 1.58218359947205
26 1.5512345234553
27 1.50769718488057
28 1.4926157395045
29 1.44563392798106
30 1.41438011328379
31 1.37991766134898
32 1.33924988905589
33 1.3198721408844
34 1.30512921015422
35 1.25801515579224
36 1.230055809021
37 1.21100886662801
38 1.18731852372487
39 1.15906699498494
40 1.13448309898376
41 1.10984595616659
42 1.08047389984131
43 1.07325216134389
44 1.04940263430278
45 1.02105005582174
46 1.01179869969686
47 0.993638475735982
48 0.964110016822815
49 0.938532829284668
50 0.927387277285258
51 0.894576052824656
52 0.893564780553182
53 0.863967577616374
54 0.86382665236791
55 0.837663014729818
56 0.815940896670024
57 0.807467758655548
58 0.785366733868917
59 0.771554430325826
60 0.756943305333455
61 0.7295241355896
62 0.721840679645538
63 0.705686132113139
64 0.698269883791606
65 0.682301898797353
66 0.665986061096191
67 0.658398012320201
68 0.640585839748383
69 0.628441333770752
70 0.614488422870636
71 0.605188171068827
72 0.586008330186208
73 0.58089272181193
74 0.567770799001058
75 0.550765832265218
76 0.544047911961873
77 0.533142328262329
78 0.528850317001343
79 0.522563874721527
80 0.507682879765828
81 0.500923156738281
82 0.492942263682683
83 0.472590684890747
84 0.473673323790232
85 0.468394974867503
86 0.459633737802505
87 0.444955448309581
88 0.435494512319565
89 0.4246686398983
90 0.430009325345357
91 0.410061985254288
92 0.416292995214462
93 0.400657067696253
94 0.398207157850266
95 0.389810095230738
96 0.384413907925288
97 0.378573109706243
98 0.372891833384832
99 0.366469273964564
100 0.363163898388545
101 0.351216504971186
102 0.345259378353755
103 0.342724084854126
104 0.341642498970032
105 0.334579716126124
106 0.327424238125483
107 0.324249108632406
108 0.321321656306585
109 0.315287212530772
110 0.315224786599477
111 0.308849563201268
112 0.302819947401683
113 0.29757230480512
114 0.292584518591563
115 0.290392408768336
116 0.287186056375504
117 0.288179536660512
118 0.282097299893697
119 0.279759794473648
120 0.274699807167053
121 0.277833104133606
122 0.267987291018168
123 0.261880546808243
124 0.263830721378326
125 0.259477853775024
126 0.248118569453557
127 0.253173321485519
128 0.251328647136688
129 0.245241850614548
130 0.248827556769053
131 0.245183075467745
132 0.24005534251531
133 0.227494612336159
134 0.233181854089101
135 0.229180271426837
136 0.235557769735654
137 0.228159894545873
138 0.225859825809797
139 0.221982901295026
140 0.222834428151449
141 0.218189626932144
142 0.218352923790614
143 0.21715102593104
144 0.209852615992228
145 0.20988737543424
146 0.208784267306328
147 0.208158572514852
148 0.205412497123082
149 0.202380682031314
150 0.201659445961316
151 0.1969286998113
152 0.191233222683271
153 0.195934464534124
154 0.190817847847939
155 0.193607737620672
156 0.193862651785215
157 0.189276615778605
158 0.184591218829155
159 0.190302575627963
160 0.189410055677096
161 0.178968151410421
162 0.179846187432607
163 0.179550444086393
164 0.17996634542942
165 0.181226740280787
166 0.177652105689049
167 0.175162956118584
168 0.16690394282341
169 0.171430915594101
170 0.170425067345301
171 0.175667429963748
172 0.171368052562078
173 0.169288461407026
174 0.166637654105822
175 0.160205741723379
176 0.167370662093163
177 0.162316873669624
178 0.157259846727053
179 0.157862861951192
180 0.160920858383179
181 0.161700631181399
182 0.155310476819674
183 0.156299923857053
184 0.15411739051342
185 0.155647014578184
186 0.145475914080938
187 0.155025323232015
188 0.149210959672928
189 0.146137371659279
190 0.149649019042651
191 0.149866362412771
192 0.14430396258831
193 0.146883214513461
194 0.147112146019936
195 0.143214464187622
196 0.14122141400973
197 0.142038787404696
198 0.139824365576108
199 0.141132007042567
200 0.142699589331945
201 0.139637584487597
202 0.140100210905075
203 0.13592321674029
204 0.134049490094185
205 0.136305287480354
206 0.134218876560529
207 0.136721735199293
208 0.132506648699443
209 0.131497234106064
210 0.131243775288264
211 0.13079788784186
212 0.130666876832644
213 0.129274437824885
214 0.129730602105459
215 0.129862303535144
216 0.131352126598358
217 0.127954920132955
218 0.127951137721539
219 0.125008997817834
220 0.122098699212074
221 0.12351389726003
222 0.122140203913053
223 0.124192734559377
224 0.11998587846756
225 0.119877482453982
226 0.123391804595788
227 0.120006357630094
228 0.120383463799953
229 0.124704644083977
230 0.118820396562417
231 0.115468174219131
232 0.116035312414169
233 0.114252532521884
234 0.117656916379929
235 0.117929975191752
236 0.11737289528052
237 0.113746864100297
238 0.112770510216554
239 0.114408023655415
240 0.118757747113705
241 0.110555882255236
242 0.111692550281684
243 0.110521264374256
244 0.109214218954245
245 0.111998476088047
246 0.10789663841327
247 0.109814522167047
248 0.110586787263552
249 0.107920703788598
250 0.111217699944973
251 0.105173928042253
252 0.105226039886475
253 0.104914948344231
254 0.103699572384357
255 0.108163433770339
256 0.103105631967386
257 0.102936454117298
258 0.105928584933281
259 0.106142714619637
260 0.103439465165138
261 0.102289594709873
262 0.105248386661212
263 0.101862107714017
264 0.102244528631369
265 0.103449150919914
266 0.105156220495701
267 0.10328254600366
268 0.104295864701271
269 0.100205436348915
270 0.0974720567464828
271 0.101238692800204
272 0.0933987870812416
273 0.100271083414555
274 0.0995764185984929
275 0.093807098766168
276 0.0937789008021355
277 0.0975874960422516
278 0.0952111010750135
279 0.0937479262550672
280 0.0963560889164607
281 0.0927068591117859
282 0.0974474499622981
283 0.0901439835627874
284 0.0927392443021138
};
\addplot [line width=2.5pt, red, opacity=1.0]
table {%
0 4.64078617095947
1 3.95971393585205
2 3.36203161875407
3 3.05695692698161
4 2.97605419158936
5 2.65950314203898
6 2.46027119954427
7 2.38809720675151
8 2.17025089263916
9 2.04607033729553
10 1.98406426111857
11 1.82642352581024
12 1.73724925518036
13 1.66582008202871
14 1.54900785287221
15 1.52063512802124
16 1.46609306335449
17 1.36645511786143
18 1.32447059949239
19 1.25509874025981
20 1.17780856291453
21 1.18843245506287
22 1.13464685281118
23 1.06495740016301
24 1.04055086771647
25 0.98300697406133
26 0.932397464911143
27 0.922910968462626
28 0.874794860680898
29 0.824543416500092
30 0.787359416484833
31 0.74926753838857
32 0.707831919193268
33 0.678189635276794
34 0.641900380452474
35 0.616007268428802
36 0.594235400358836
37 0.560228149096171
38 0.544666786988576
39 0.527407228946686
40 0.496792415777842
41 0.467662870883942
42 0.452483654022217
43 0.432795902093252
44 0.431906739870707
45 0.413629651069641
46 0.40665340423584
47 0.409453461567561
48 0.40177313486735
49 0.387142807245255
50 0.384388526280721
51 0.367420256137848
52 0.359724362691243
53 0.356117784976959
54 0.344796637694041
55 0.332405686378479
56 0.322956850131353
57 0.320509324471156
58 0.30005673567454
59 0.292755832274755
60 0.285541256268819
61 0.271850630640984
62 0.264407714207967
63 0.253780966003736
64 0.238000179330508
65 0.233109240730604
66 0.222245638569196
67 0.207436298330625
68 0.204544122020404
69 0.195481354991595
70 0.177561193704605
71 0.175609216094017
72 0.168624927600225
73 0.159887065490087
74 0.159449100494385
75 0.150346005956332
76 0.141799442470074
77 0.13789385308822
78 0.130409238239129
79 0.13084477186203
80 0.127291771272818
81 0.115001261234283
82 0.109171676139037
83 0.109927353759607
84 0.103937330345313
85 0.100516438484192
86 0.0978119472662608
87 0.0947936649123828
88 0.0938886428872744
89 0.0919270813465118
90 0.0830397456884384
91 0.080086608727773
92 0.0805719395478566
93 0.0775554105639458
94 0.073787705351909
95 0.0716366432607174
96 0.0677302690843741
97 0.0648196885983149
98 0.0615860658387343
99 0.0617834130922953
100 0.0606565065681934
101 0.0555128467579683
102 0.0527889604369799
103 0.0504294584194819
104 0.0494745336472988
105 0.0481190445522467
106 0.04484815026323
107 0.0429015625268221
108 0.0431492688755194
109 0.0415916591882706
110 0.0381784196943045
111 0.0376032336304585
112 0.0348736103624105
113 0.0324286948889494
114 0.0302386457721392
115 0.0313484643896421
116 0.0296943349142869
117 0.0279581770300865
118 0.029148825754722
119 0.0303234917422136
120 0.0311954207718372
121 0.0263403467833996
122 0.0257790517061949
123 0.0249999885757764
124 0.0233742902055383
125 0.0241394688685735
126 0.0222436416273316
127 0.023954451084137
128 0.0216053050632278
129 0.0217586206272244
130 0.0204072520136833
131 0.0211445512250066
132 0.0210232036188245
133 0.0194749707976977
134 0.0178765044547617
135 0.0173000890451173
136 0.0176720370848974
137 0.0169575260952115
138 0.0169526807342966
139 0.017087366587172
140 0.0168104694845776
141 0.0168691882863641
142 0.0160085350895921
143 0.0141332486333946
144 0.011551259085536
145 0.0111886848074694
146 0.0111854621985306
147 0.011465306704243
148 0.0104130099837979
149 0.0102549979152779
150 0.00985252500201265
151 0.00958474601308505
152 0.0101813668540368
153 0.0100493996093671
154 0.00998953240923584
155 0.00898020427363614
156 0.00837436659882466
157 0.0100809053207437
158 0.00919965739982823
159 0.00891983020119369
160 0.00965863062689702
161 0.00888262006143729
162 0.00845587796842059
163 0.00899830732184152
164 0.00921268276094148
165 0.00840594883387288
166 0.00849930511321872
167 0.00805665645748377
168 0.00828116239669422
169 0.00875631021335721
170 0.00803624248752991
171 0.00756495515815914
172 0.00823074470584591
173 0.00745189756465455
174 0.00837470226300259
175 0.00783119513653219
176 0.00818187991778056
177 0.00755480444058776
178 0.0060951296860973
179 0.00609397616547843
180 0.00561688809345166
181 0.00588248833082616
182 0.00566303147934377
183 0.00528478882430742
184 0.005974245024845
185 0.00541155599057674
186 0.00574876197303335
187 0.00543138417803372
188 0.00562833156436682
189 0.00493926574320843
190 0.00561179523356259
191 0.0054584148262317
192 0.00511322535263995
193 0.00532776528658966
194 0.00516215129755437
195 0.00405997503548861
196 0.00405746403460701
197 0.00399942121778925
198 0.00439366818560908
199 0.00389588267231981
200 0.00401808655199905
201 0.00377874448895454
202 0.00309168974248072
203 0.0036904172350963
204 0.00358213662790755
205 0.00373193990283956
206 0.00328323089828094
207 0.00328230597854902
208 0.00356150331208482
209 0.00347723943802218
210 0.00315092168360328
211 0.00286020788674553
212 0.00298426195513457
213 0.00322835186185936
214 0.00299772278716167
215 0.00256185175385326
216 0.00236810805896918
217 0.00234905067676057
218 0.00263137917499989
219 0.00235261756461114
220 0.00259137456305325
221 0.00227475352585316
222 0.00218226260039955
223 0.00210753122034172
224 0.0022422302281484
225 0.00207220039252813
226 0.00226480653509498
227 0.00225929201891025
228 0.00207894543806712
229 0.00247124411786596
230 0.00260549399536103
231 0.00217841181438416
232 0.00235645690311988
233 0.00230468138276289
234 0.00214099301956594
235 0.00203279196284711
236 0.00225834796826045
237 0.00217869621701539
238 0.00209102290682495
239 0.00217849815574785
240 0.00204582376560817
241 0.00233166439769169
242 0.00188442401122302
243 0.00199784070719033
244 0.00176833901787177
245 0.00196313737736394
246 0.0019175410658742
247 0.00190415225612621
248 0.00198756786994636
249 0.00183238678922256
250 0.00208271484977255
251 0.00182604607349883
252 0.00185898357691864
253 0.00221367118259271
254 0.00145776631931464
255 0.00196372647769749
256 0.00178443124362578
257 0.0019902012621363
258 0.00218085125864794
259 0.00170021748635918
260 0.00190021078257511
261 0.00161664797148357
262 0.00189688553412755
263 0.00154188841891785
264 0.00195857331467172
265 0.00201169451853881
266 0.00172158268590768
267 0.00169285394561787
268 0.00207651895470917
269 0.00205398467369378
270 0.00171784548244129
271 0.00194205766698966
272 0.00242110105076184
273 0.00188020030812671
274 0.00171280248711507
275 0.00155563947434227
276 0.00179155190320065
277 0.00199898611754179
278 0.0017287857675304
279 0.00188781728502363
280 0.00209518867389609
281 0.00189366463261346
282 0.00180393081003179
283 0.00213758290434877
284 0.00205234717577696
};
\addplot [line width=2.5pt, blue, opacity=1.0]
table {%
0 4.64078617095947
1 4.00082095464071
2 3.44155724843343
3 3.13301650683085
4 2.97396405537923
5 2.68530694643656
6 2.50474286079407
7 2.4120626449585
8 2.17360790570577
9 2.06395689646403
10 1.98581238587697
11 1.80937588214874
12 1.7462660074234
13 1.66837771733602
14 1.53353130817413
15 1.51019303003947
16 1.43186763922373
17 1.33019256591797
18 1.32196021080017
19 1.24682494004567
20 1.15622858206431
21 1.14102713267008
22 1.06424768765767
23 0.997162302335103
24 0.991235335667928
25 0.926438470681508
26 0.874607761700948
27 0.864786783854167
28 0.80463707447052
29 0.766077319780985
30 0.734830756982168
31 0.686465839544932
32 0.652099172274272
33 0.617668688297272
34 0.578778525193532
35 0.547174314657847
36 0.526911894480387
37 0.491631875435511
38 0.466461062431335
39 0.438462853431702
40 0.39852304259936
41 0.394751002391179
42 0.36693932612737
43 0.346828550100327
44 0.362717439730962
45 0.337909708420436
46 0.324471692244212
47 0.329063554604848
48 0.309694687525431
49 0.296187271674474
50 0.294607102870941
51 0.278222262859344
52 0.26087453464667
53 0.263291945060094
54 0.237410952647527
55 0.231450537840525
56 0.217712392409643
57 0.206254829963048
58 0.190296108524005
59 0.175053685903549
60 0.171962653597196
61 0.151467795173327
62 0.144707242647807
63 0.139137754837672
64 0.123516318698724
65 0.125483895341555
66 0.11823213348786
67 0.104870244860649
68 0.0961119532585144
69 0.0928297241528829
70 0.0790439148743947
71 0.0779711852471034
72 0.0734852949778239
73 0.0616766127447287
74 0.0606574428578218
75 0.056255366653204
76 0.0536954179406166
77 0.0505732136468093
78 0.0475191287696362
79 0.0460771061480045
80 0.0417201829453309
81 0.0391286785403887
82 0.0360181617240111
83 0.0350000038743019
84 0.0308563069750865
85 0.0293603514631589
86 0.0267880912870169
87 0.0270778816193342
88 0.0253205181409915
89 0.023160910854737
90 0.0210556741803885
91 0.0203478870292505
92 0.02065636528035
93 0.0177857372909784
94 0.0175897146885594
95 0.0161876011018952
96 0.0158734299863378
97 0.0144627199818691
98 0.0142540307715535
99 0.0127799483016133
100 0.011974369486173
101 0.0114758722484112
102 0.0117209047699968
103 0.0105159968758623
104 0.0106169128169616
105 0.0100861564278603
106 0.0100475403790673
107 0.00986223698904117
108 0.00953790297110876
109 0.0098825025682648
110 0.00962162980188926
111 0.00943387175599734
112 0.00887023750692606
113 0.00859223150958618
114 0.00839107576757669
115 0.00864394474774599
116 0.00813171608994404
117 0.00896906806156039
118 0.00990597003449996
119 0.0121762109920382
120 0.0129645966614286
121 0.0154455757389466
122 0.0169443249081572
123 0.0196467082326611
124 0.0204556506748001
125 0.021756975290676
126 0.0214687744155526
127 0.021111245577534
128 0.0211814558133483
129 0.0224313059200843
130 0.0225469535216689
131 0.0225389882301291
132 0.0214441757028302
133 0.0222818031907082
134 0.0228289728984237
135 0.0247487801437577
136 0.0247943097104629
137 0.0245351282258828
138 0.0211043764526645
139 0.0213783858343959
140 0.0227735647931695
141 0.0207550578440229
142 0.0248932745307684
143 0.0195275191217661
144 0.0118456138297915
145 0.00958355562761426
146 0.00875939552982648
147 0.0077825083086888
148 0.0069340355694294
149 0.00664394861087203
150 0.00680821404481928
151 0.00564714769522349
152 0.00572477281093597
153 0.00591048613811533
154 0.00568401487544179
155 0.00563350335384409
156 0.0058837008352081
157 0.00589726601416866
158 0.00548116955906153
159 0.00550044910050929
160 0.00544627383351326
161 0.00533488346263766
162 0.004595483886078
163 0.00550215823265413
164 0.0048486462328583
165 0.0050445853266865
166 0.00534753330672781
167 0.00541743465388815
168 0.00657022666806976
169 0.00680736064290007
170 0.00586744739363591
171 0.00647950638085604
172 0.00633701775223017
173 0.00670413843666514
174 0.00713444082066417
175 0.00655478968595465
176 0.00589585149039825
177 0.00607690106456478
178 0.00584899843670428
179 0.00539508461952209
180 0.0058568068780005
181 0.00600781881560882
182 0.00521250022575259
183 0.00557052677807709
184 0.0062083409478267
185 0.00583841597350935
186 0.00604639862043162
187 0.00565808576842149
188 0.00531654176302254
189 0.00601887369217972
190 0.00633455347269773
191 0.00605484369831781
192 0.00563905946910381
193 0.00615856051445007
194 0.00635166023857892
195 0.00767780203993122
196 0.00699924964768191
197 0.00653115892782807
198 0.00755597961445649
199 0.00688244968963166
200 0.00699440149279932
201 0.0070476436521858
202 0.00699491108146807
203 0.00755100324749947
204 0.00652502678955595
205 0.00718851131387055
206 0.00700744815791647
207 0.00780381712441643
208 0.00686444349897404
209 0.0072533218190074
210 0.00714483337166409
211 0.00711166866434117
212 0.00676386787866553
213 0.00667707730705539
214 0.00623921010022362
215 0.00527067707541088
216 0.0043367884742717
217 0.00394146454830964
218 0.00360131675067047
219 0.0036612794889758
220 0.00331252867666384
221 0.0029828993914028
222 0.00314461478653053
223 0.0031108926050365
224 0.00324376897575955
225 0.00316189493363102
226 0.00309702571636687
227 0.00317940833823135
228 0.00309003276440004
229 0.00324273093913992
230 0.00283729090976218
231 0.00290640501771122
232 0.00309912646965434
233 0.00295146969923129
234 0.00277692824602127
235 0.0026963409424449
236 0.00295156399564197
237 0.00259355471159021
238 0.00277021170283357
239 0.00266365466328959
240 0.00283276351789633
241 0.00257716972070436
242 0.0026154121151194
243 0.0028076411690563
244 0.00234411074779928
245 0.00247768937454869
246 0.00271399439467738
247 0.00285008417752882
248 0.00270511046983302
249 0.00246387436830749
250 0.00286355762121578
251 0.00225488709596296
252 0.00222163022651027
253 0.00241646100766957
254 0.00238469922138999
255 0.00231670398109903
256 0.00256834369308005
257 0.00248426711186767
258 0.00244745658710599
259 0.00222229208641996
260 0.0022768732936432
261 0.00227539738019307
262 0.00227445728766421
263 0.0023064889634649
264 0.00229153479449451
265 0.00199545198120177
266 0.00213538971729577
267 0.00214735104236752
268 0.0022578506032005
269 0.00211454966726402
270 0.00226354051847011
271 0.0021072601666674
272 0.00235612799103061
273 0.00227772428964575
274 0.00219960134321203
275 0.00208148135182758
276 0.00216362260592481
277 0.00247169060943027
278 0.00224724280027052
279 0.00205908230661104
280 0.00221467227675021
281 0.00210616101200382
282 0.00192203485251715
283 0.00184410888080796
284 0.00203766770816098
};
\addplot [line width=2.5pt, color2, opacity=1.0]
table {%
0 4.64078617095947
2 4.09980901082357
4 3.4193811416626
6 2.99245500564575
8 2.70182291666667
10 2.47215620676676
12 2.28625726699829
14 2.1281209786733
16 2.00643960634867
18 1.885369181633
20 1.78454347451528
22 1.68489642937978
24 1.59023404121399
26 1.50501235326131
28 1.42700970172882
30 1.34685949484507
32 1.26569604873657
34 1.16202084223429
36 1.07919812202454
38 0.985356311003367
40 0.893657286961873
42 0.812123556931814
44 0.733422180016836
46 0.659110208352407
48 0.595956524213155
50 0.536488254865011
52 0.480977733929952
54 0.436667482058207
56 0.392633646726608
58 0.363644997278849
60 0.332184731960297
62 0.305476148923238
64 0.283468117316564
66 0.269320021073023
68 0.248485401272774
70 0.23373248676459
72 0.223003094395002
74 0.207313929994901
76 0.197934329509735
78 0.188462346792221
80 0.177541812260946
82 0.172820935646693
84 0.163745611906052
86 0.160762170950572
88 0.148467868566513
90 0.144450480739276
92 0.143464947740237
94 0.136392995715141
96 0.132660458485285
98 0.126634021600087
100 0.123103862007459
102 0.120578892529011
104 0.118779473006725
106 0.112938210368156
108 0.11204765488704
110 0.10888143380483
112 0.105259304245313
114 0.101972386240959
116 0.0991441408793132
118 0.0983782783150673
120 0.0967540939648946
122 0.0939786086479823
124 0.0938907985885938
126 0.0939909666776657
128 0.0894170751174291
130 0.0884578948219617
132 0.0848558644453685
134 0.0842686866720517
136 0.0837219431996346
138 0.0827641387780507
140 0.0794454912344615
142 0.079469233751297
144 0.0759028792381287
146 0.0772748167316119
148 0.0729837144414584
150 0.0745721558729808
152 0.0716838240623474
154 0.0709903116027514
156 0.0682350744803746
158 0.0703433454036713
160 0.0685372625788053
162 0.0657889693975449
164 0.0679637317856153
166 0.0659819965561231
168 0.06478343034784
170 0.0644221343100071
172 0.0624253662923972
174 0.0632321201264858
176 0.0619191639125347
178 0.0615636358658473
180 0.0610585808753967
182 0.0591248658796152
184 0.0613241344690323
186 0.0583283553520838
188 0.0572033648689588
190 0.0600480685631434
192 0.0570502815147241
194 0.0569963989158471
196 0.0566393136978149
198 0.0555395372211933
200 0.0559744661053022
202 0.053838063031435
204 0.0545466194550196
206 0.0533737863103549
208 0.0541860200464725
210 0.054018201927344
212 0.0519808841248353
214 0.051687008390824
216 0.0507221780717373
218 0.0502634110550086
220 0.0500509701669216
222 0.0510101790229479
224 0.0497027772168318
226 0.0490432741741339
228 0.0475651149948438
230 0.0485469996929169
232 0.0495039410889149
234 0.0482560644547145
236 0.0478686168789864
238 0.0474534332752228
240 0.0467615562180678
242 0.0465962762633959
244 0.047072329868873
246 0.0462541555364927
248 0.0466048829257488
250 0.045686977605025
252 0.0441806763410568
254 0.0430667760471503
256 0.04420305788517
258 0.0424251208702723
260 0.0429752431809902
262 0.0444708466529846
264 0.0433803933362166
266 0.0424329241116842
268 0.0417793566981951
270 0.0425431802868843
272 0.043182734400034
274 0.0436992074052493
276 0.0425200946629047
278 0.0415876383582751
280 0.0424694046378136
282 0.0401366303364436
284 0.0409467183053493
};
\end{axis}

\end{tikzpicture}

%% file: labpal/figure_data/performance_comparison/CIFAR-100/CIFAR-100_DenseNet-121_validation_accuracy.pgf
% This file was created by tikzplotlib v0.9.8.
\begin{tikzpicture}

\definecolor{color0}{rgb}{0.647058823529412,0.164705882352941,0.164705882352941}
\definecolor{color1}{rgb}{0.933333333333333,0.509803921568627,0.933333333333333}
\definecolor{color2}{rgb}{1,0.647058823529412,0}

\begin{axis}[
yticklabel style={font=\large},
xticklabel style={font=\large},
ytick style={font=\Large},
xtick style={font=\Large},
ylabel style={font=\Large},
xlabel style={font=\Large},
title style={font=\Large},
grid = major,
major grid style={dotted},
minor xtick={},
minor ytick={},
tick align=outside,
tick pos=left,
title={validation accuracy CIFAR-100 DenseNet-121},
width=10.5cm,height=8cm,,
x grid style={white!69.0196078431373!black},
xlabel={epochs},
xmin=-13.15, xmax=298.15,
xtick style={color=black},
xtick={-50,0,50,100,150,200,250,300},
y grid style={white!69.0196078431373!black},
ylabel={validation accuracy},
ymin=0.35, ymax=0.74,
ytick style={color=black},
ytick={-0.1,0,0.1,0.2,0.3,0.4,0.5,0.6,0.7,0.8},
minor y tick num=3
]
\addplot [line width=2.5pt, color0, opacity=1.0]
table {%
1 0.154113243023554
2 0.195112183690071
3 0.256009618441264
4 0.290531516075134
5 0.330061435699463
6 0.354166666666667
7 0.372996797164281
8 0.406717409690221
9 0.407785793145498
10 0.432892630497615
11 0.452724357446035
12 0.463541666666667
13 0.476228624582291
14 0.485376606384913
15 0.494123925765355
16 0.493389417727788
17 0.507745732863744
18 0.515625
19 0.514890491962433
20 0.520900130271912
21 0.529113252957662
22 0.535389959812164
23 0.545339206854502
24 0.54667466878891
25 0.538928945859273
26 0.545138895511627
27 0.552817841370901
28 0.549612720807393
29 0.550547520319621
30 0.557625532150269
31 0.55528845389684
32 0.561965823173523
33 0.560830672581991
34 0.558293282985687
35 0.562967399756114
36 0.565104186534882
37 0.57298344373703
38 0.569845080375671
39 0.566105763117472
40 0.56764155626297
41 0.564636747042338
42 0.57011216878891
43 0.573250532150269
44 0.576388895511627
45 0.57605501015981
46 0.576522429784139
47 0.580128192901611
48 0.577123403549194
49 0.577190180619558
50 0.582064628601074
51 0.585603634516398
52 0.57605501015981
53 0.581196586290995
54 0.583600441614787
55 0.583600421746572
56 0.577457269032796
57 0.58553687731425
58 0.584334949652354
59 0.584001064300537
60 0.582398494084676
61 0.585870703061422
62 0.583667198816935
63 0.590611636638641
64 0.591613252957662
65 0.587473293145498
66 0.587807138760885
67 0.590878744920095
68 0.590344548225403
69 0.593616445859273
70 0.590411305427551
71 0.597155451774597
72 0.603098313013713
73 0.595285793145498
74 0.595886766910553
75 0.595419327418009
76 0.600827991962433
77 0.599091867605845
78 0.601495722929637
79 0.59107905626297
80 0.599893152713776
81 0.598958333333333
82 0.597489317258199
83 0.600360572338104
84 0.600894769032796
85 0.598357379436493
86 0.601428945859273
87 0.602297027905782
88 0.603498935699463
89 0.602230230967204
90 0.606436967849731
91 0.603699247042338
92 0.602096676826477
93 0.606770833333333
94 0.600293795267741
95 0.599559307098389
96 0.59702189763387
97 0.601896385351817
98 0.605435371398926
99 0.600160260995229
100 0.598958333333333
101 0.599959949652354
102 0.59889155626297
103 0.600227018197378
104 0.602029919624329
105 0.604700843493144
106 0.592948714892069
107 0.592748403549194
108 0.594017068545024
109 0.598157068093618
110 0.593015491962433
111 0.594284176826477
112 0.602096676826477
113 0.601562519868215
114 0.596020301183065
115 0.60176283121109
116 0.599893172581991
117 0.602096696694692
118 0.603231847286224
119 0.60877404610316
120 0.607371807098389
121 0.599425752957662
122 0.598424156506856
123 0.603899578253428
124 0.607638895511627
125 0.604967951774597
126 0.604700863361359
127 0.598691244920095
128 0.598157048225403
129 0.607505341370901
130 0.602297008037567
131 0.596754809220632
132 0.602764427661896
133 0.598157028357188
134 0.598691244920095
135 0.598023513952891
136 0.593482911586761
137 0.600427349408468
138 0.604901174704234
139 0.603298604488373
140 0.467147434751193
141 0.42828526844581
142 0.573717931906382
143 0.583800752957662
144 0.594951927661896
145 0.592748403549194
146 0.598758021990458
147 0.601428945859273
148 0.595886766910553
149 0.599091867605845
150 0.593082269032796
151 0.591479698816935
152 0.595753212769826
153 0.599358975887299
154 0.594618062178294
155 0.599492530028025
156 0.597489317258199
157 0.605168282985687
158 0.598424156506856
159 0.602297008037567
160 0.606036325295766
161 0.604500532150269
162 0.604567309220632
163 0.609107911586761
164 0.608907600243886
165 0.604099889596303
166 0.605368594328562
167 0.605235020319621
168 0.601095080375671
169 0.599425752957662
170 0.593616465727488
171 0.601829608281453
172 0.604366997877757
173 0.600494126478831
174 0.603699247042338
175 0.423944987356663
176 0.404513886819283
177 0.421006932854652
178 0.583600441614787
179 0.57091345389684
180 0.589276174704234
181 0.59168001015981
182 0.600761234760284
183 0.569577991962433
184 0.596888363361359
185 0.597622871398926
186 0.60483439763387
187 0.604366997877757
188 0.597088654836019
189 0.603165070215861
190 0.478231827418009
191 0.604099889596303
192 0.608707269032796
193 0.604300200939178
194 0.607705672581991
195 0.603565712769826
196 0.609107891718547
197 0.609174688657125
198 0.60610310236613
199 0.611912389596303
200 0.609375
201 0.608306606610616
202 0.551949779192607
203 0.610510150591532
204 0.608106315135956
205 0.415531524146597
206 0.602897961934408
207 0.595419327418009
208 0.461271375417709
209 0.60710471868515
210 0.60196312268575
211 0.443776714305083
212 0.599025110403697
213 0.415264417727788
214 0.601495722929637
215 0.600427329540253
216 0.603966335455577
217 0.600761214892069
218 0.415998935699463
219 0.603498935699463
220 0.597222228844961
221 0.392761759149532
222 0.415264432628949
223 0.590611656506856
224 0.572315692901611
225 0.596287389596303
226 0.398170421210428
227 0.400841350667179
228 0.600093483924866
229 0.602096676826477
230 0.602230230967204
231 0.608106315135956
232 0.606570502122243
233 0.600694457689921
234 0.604634086290995
235 0.604567329088847
236 0.607505341370901
237 0.602230250835419
238 0.593816796938578
239 0.564569969971975
240 0.587940692901611
241 0.592147429784139
242 0.216546465332309
243 0.0126869656766454
244 0.436832269032796
245 0.574986636638641
246 0.574719548225403
247 0.589877148469289
248 0.581263363361359
249 0.589610040187836
250 0.404046464090546
251 0.592881957689921
252 0.600494126478831
253 0.506744116544724
254 0.60096154610316
255 0.591145833333333
256 0.594484527905782
257 0.567240913709005
258 0.597556094328562
259 0.597556094328562
260 0.609508554140727
261 0.598958333333333
262 0.599826395511627
263 0.47809828321139
264 0.410657055986424
265 0.595352550347646
266 0.569978634516398
267 0.594484508037567
268 0.594751616319021
269 0.413995722929637
270 0.604233423868815
271 0.596754809220632
272 0.50687767068545
273 0.600560883680979
274 0.599492530028025
275 0.508613785107931
276 0.607572098573049
277 0.609708865483602
278 0.590411325295766
279 0.555088152488073
280 0.585336526234945
281 0.603298624356588
282 0.594083865483602
283 0.593215803305308
284 0.60096154610316
};
\addplot [line width=2.5pt, green!50.1960784313725!black, opacity=1.0]
table {%
1 0.162459939718246
2 0.244991987943649
3 0.297208865483602
4 0.351028313239415
5 0.376268694798152
6 0.411324779192607
7 0.445713142553965
8 0.468950321276983
9 0.468616455793381
10 0.491653303305308
11 0.513421475887299
12 0.516292730967204
13 0.533587058385213
14 0.547609508037567
15 0.539463142553965
16 0.553285260995229
17 0.563835461934408
18 0.568175733089447
19 0.570446034272512
20 0.574118594328562
21 0.577390491962433
22 0.578659196694692
23 0.573784728844961
24 0.579393684864044
25 0.583867530028025
26 0.585403303305308
27 0.583133002122243
28 0.590945521990458
29 0.590277751286825
30 0.592080652713776
31 0.599692841370901
32 0.594284176826477
33 0.591546475887299
34 0.600427349408468
35 0.600627680619558
36 0.593816757202148
37 0.60483439763387
38 0.612313032150269
39 0.599626064300537
40 0.601228634516398
41 0.60670405626297
42 0.606236656506856
43 0.608974357446035
44 0.610376596450806
45 0.609441777070363
46 0.613448202610016
47 0.610176285107931
48 0.615451395511627
49 0.610710461934408
50 0.618389407793681
51 0.618055562178294
52 0.613715271155039
53 0.618723293145498
54 0.619524598121643
55 0.621327459812164
56 0.61451655626297
57 0.619524558385213
58 0.62025906642278
59 0.619524578253428
60 0.620793282985687
61 0.617588142553965
62 0.619925220807393
63 0.620526174704234
64 0.620993594328562
65 0.618122319380442
66 0.626268684864044
67 0.623464206854502
68 0.623197118441264
69 0.627069969971975
70 0.625600973765055
71 0.619658132394155
72 0.622662941614787
73 0.626268704732259
74 0.629139959812164
75 0.623197118441264
76 0.620726505915324
77 0.62767094373703
78 0.626535793145498
79 0.623731315135956
80 0.620993594328562
81 0.629807690779368
82 0.623063564300537
83 0.624065160751343
84 0.625267108281453
85 0.627069969971975
86 0.626535793145498
87 0.62172810236613
88 0.630208313465118
89 0.627537389596303
90 0.631209949652354
91 0.629407048225403
92 0.634081204732259
93 0.630942841370901
94 0.630475421746572
95 0.632011214892069
96 0.627003192901611
97 0.62767094373703
98 0.632011214892069
99 0.630341867605845
100 0.627537389596303
101 0.630809287230174
102 0.630608975887299
103 0.634481827418009
104 0.631209929784139
105 0.633279939492544
106 0.632011214892069
107 0.635616978009542
108 0.630608956019084
109 0.63528311252594
110 0.631810883680979
111 0.639423072338104
112 0.635884086290995
113 0.631143152713776
114 0.634481827418009
115 0.633079588413239
116 0.634348293145498
117 0.631543795267741
118 0.634281516075134
119 0.640357911586761
120 0.634949247042338
121 0.636952439943949
122 0.634348293145498
123 0.631477018197378
124 0.633613785107931
125 0.641159176826477
126 0.636685351530711
127 0.639890491962433
128 0.636752128601074
129 0.640291134516398
130 0.636217951774597
131 0.636685371398926
132 0.631677369276682
133 0.633880893389384
134 0.640625
135 0.636084417502085
136 0.637353082497915
137 0.641025642553965
138 0.641426285107931
139 0.634147961934408
140 0.633547008037567
141 0.636284728844961
142 0.635950863361359
143 0.643963674704234
144 0.647369126478831
145 0.649372339248657
146 0.652310371398926
147 0.651842951774597
148 0.654847780863444
149 0.654447098573049
150 0.654914518197378
151 0.653912941614787
152 0.655248403549194
153 0.65831998984019
154 0.654580672581991
155 0.654714206854502
156 0.658386766910553
157 0.653979698816935
158 0.655982911586761
159 0.657785793145498
160 0.657118042310079
161 0.657051304976145
162 0.657451927661896
163 0.660189648469289
164 0.655448734760284
165 0.660256425539652
166 0.659722228844961
167 0.656383554140727
168 0.656784196694692
169 0.658119658629099
170 0.659655451774597
171 0.656850973765055
172 0.657318373521169
173 0.660323182741801
174 0.656517108281453
175 0.660256425539652
176 0.663595080375671
177 0.660523494084676
178 0.659388363361359
179 0.659922540187836
180 0.66159188747406
181 0.658787409464518
182 0.660657048225403
183 0.660256425539652
184 0.656116445859273
185 0.656316757202148
186 0.65892094373703
187 0.65872061252594
188 0.661124467849731
189 0.662326375643412
190 0.660256385803223
191 0.660189648469289
192 0.663261214892069
193 0.660723805427551
194 0.66466345389684
195 0.658520301183065
196 0.658453524112701
197 0.657118062178294
198 0.661191244920095
199 0.662126064300537
200 0.658854166666667
201 0.662927349408468
202 0.662192861239115
203 0.66099093357722
204 0.658854166666667
205 0.659188032150269
206 0.660523513952891
207 0.661925733089447
208 0.661258002122243
209 0.662593503793081
210 0.66159188747406
211 0.664463142553965
212 0.660122851530711
213 0.662660260995229
214 0.663661857446035
215 0.661324779192607
216 0.658587078253428
217 0.662860572338104
218 0.662727018197378
219 0.665130873521169
220 0.663127660751343
221 0.663995742797852
222 0.659788986047109
223 0.664596676826477
224 0.660122851530711
225 0.662860572338104
226 0.664463142553965
227 0.658854166666667
228 0.66346154610316
229 0.665531516075134
230 0.660857379436493
231 0.662459949652354
232 0.665998935699463
233 0.664196054140727
234 0.663862188657125
235 0.661191244920095
236 0.666199247042338
237 0.660256405671438
238 0.662059287230174
239 0.662259618441264
240 0.660924136638641
241 0.658186435699463
242 0.661124447981516
243 0.658520301183065
244 0.658253192901611
245 0.659722228844961
246 0.664262811342875
247 0.662526706854502
248 0.661057710647583
249 0.65892094373703
250 0.663127680619558
251 0.665464739004771
252 0.663261214892069
253 0.661458333333333
254 0.662593503793081
255 0.659588674704234
256 0.663327991962433
257 0.661591867605845
258 0.657986104488373
259 0.656917730967204
260 0.662393152713776
261 0.664196034272512
262 0.658987720807393
263 0.662660260995229
264 0.664863785107931
265 0.663394749164581
266 0.663862188657125
267 0.665130873521169
268 0.665798604488373
269 0.662660280863444
270 0.663194437821706
271 0.663261214892069
272 0.663928965727488
273 0.664396365483602
274 0.663795411586761
275 0.667467951774597
276 0.660990913709005
277 0.662927349408468
278 0.659922520319621
279 0.662593483924866
280 0.65952189763387
281 0.661324779192607
282 0.664129277070363
283 0.663261234760284
284 0.661725421746572
};
\addplot [line width=2.5pt, black, opacity=1.0]
table {%
1 0.0967548092206319
2 0.121060361464818
3 0.165331194798152
4 0.166533117492994
5 0.202524041136106
6 0.202924678723017
7 0.242922003070513
8 0.247662931680679
9 0.28185095389684
10 0.261485030253728
11 0.304620732863744
12 0.296340813239415
13 0.268095617492994
14 0.25260417163372
15 0.15798611442248
16 0.132211538652579
17 0.190705130497615
18 0.181290060281754
19 0.145432690779368
20 0.179887821276983
21 0.207732369502385
22 0.204861109455427
23 0.278912936647733
24 0.284121269981066
25 0.254340271155039
26 0.263688569267591
27 0.231971154610316
28 0.315838674704234
29 0.270900110403697
30 0.264890496929487
31 0.327724357446035
32 0.321247319380442
33 0.297943373521169
34 0.327256952722867
35 0.410523504018784
36 0.404113243023555
37 0.409321596225103
38 0.444444447755814
39 0.442508021990458
40 0.481637299060822
41 0.498998413483302
42 0.499332259098689
43 0.501068373521169
44 0.474425743023555
45 0.467214206854502
46 0.500066777070363
47 0.491653303305308
48 0.464610040187836
49 0.487580130497615
50 0.454126596450806
51 0.496995200713476
52 0.480769217014313
53 0.501736104488373
54 0.507077982028325
55 0.519965281089147
56 0.485710471868515
57 0.495125542084376
58 0.509081184864044
59 0.525373935699463
60 0.507545391718547
61 0.514690160751343
62 0.548477570215861
63 0.571247339248657
64 0.575320521990458
65 0.578458865483602
66 0.575186947981516
67 0.577857911586761
68 0.576989849408468
69 0.566239317258199
70 0.529780973990758
71 0.524505883455276
72 0.556891024112701
73 0.549212098121643
74 0.56002938747406
75 0.572115401426951
76 0.579660793145498
77 0.57238248984019
78 0.54914528131485
79 0.528912931680679
80 0.507278313239416
81 0.510550210873286
82 0.528044869502385
83 0.533587078253428
84 0.535723835229874
85 0.516760140657425
86 0.53465544184049
87 0.482438574234645
88 0.546875
89 0.565705120563507
90 0.554887811342875
91 0.576789538065592
92 0.60483439763387
93 0.609775642553965
94 0.615518172581991
95 0.611244658629099
96 0.601228614648183
97 0.588675220807393
98 0.575053413709005
99 0.524639407793681
100 0.547809839248657
101 0.556223293145498
102 0.569511214892069
103 0.585536857446035
104 0.589276174704234
105 0.594083865483602
106 0.598490913709005
107 0.596955120563507
108 0.586471676826477
109 0.577390491962433
110 0.603098293145498
111 0.602964739004771
112 0.598557690779368
113 0.58573716878891
114 0.610376596450806
115 0.61224627494812
116 0.598424136638641
117 0.412727027510603
118 0.591680030028025
119 0.592815180619558
120 0.584735572338104
121 0.587272961934408
122 0.588341355323792
123 0.585536857446035
124 0.568108975887299
125 0.588207801183065
126 0.587540070215861
127 0.592948695023855
128 0.601028303305308
129 0.603231847286224
130 0.601028323173523
131 0.605034708976746
132 0.610042730967204
133 0.60730501015981
134 0.603365401426951
135 0.606236636638641
136 0.605168263117472
137 0.599559287230174
138 0.596754809220632
139 0.603432158629099
140 0.596688032150269
141 0.62212872505188
142 0.627804497877757
143 0.631209929784139
144 0.634882469971975
145 0.635683755079905
146 0.633213142553965
147 0.625734508037567
148 0.608907580375671
149 0.598023494084676
150 0.49485844373703
151 0.606503744920095
152 0.609174688657125
153 0.61264689763387
154 0.623597741127014
155 0.624666154384613
156 0.62706998984019
157 0.627003212769826
158 0.630809287230174
159 0.633012811342875
160 0.633012811342875
161 0.631543815135956
162 0.629006405671438
163 0.631477038065592
164 0.629874467849731
165 0.625133554140727
166 0.628939628601074
167 0.637620190779368
168 0.633947630723318
169 0.63815438747406
170 0.642962078253428
171 0.635750532150269
172 0.640958865483602
173 0.640625
174 0.647569457689921
175 0.64389689763387
176 0.643763363361359
177 0.64022433757782
178 0.637019236882528
179 0.630942841370901
180 0.619123935699463
181 0.622061967849731
182 0.608373383680979
183 0.616720080375671
184 0.625467419624329
185 0.619991997877757
186 0.626335461934408
187 0.615651706854502
188 0.597222228844961
189 0.600560883680979
190 0.592080652713776
191 0.600894749164581
192 0.602630873521169
193 0.613848825295766
194 0.613848825295766
195 0.617788473765055
196 0.620659728844961
197 0.62827189763387
198 0.627604166666667
199 0.63221154610316
200 0.635750552018484
201 0.637887259324392
202 0.632411857446035
203 0.640892088413239
204 0.642427881558736
205 0.642227570215861
206 0.644497871398926
207 0.647035241127014
208 0.644163986047109
209 0.643896917502085
210 0.643963674704234
211 0.636551817258199
212 0.624265491962433
213 0.629006405671438
214 0.623597760995229
215 0.618723293145498
216 0.62747061252594
217 0.633747339248657
218 0.631343483924866
219 0.646501064300537
220 0.64269498984019
221 0.639423052469889
222 0.641493042310079
223 0.633547008037567
224 0.63221154610316
225 0.62439904610316
226 0.631677349408468
227 0.602430542310079
228 0.619658132394155
229 0.618055562178294
230 0.624332269032796
231 0.633814076582591
232 0.632545411586761
233 0.623931626478831
234 0.636485040187836
235 0.644497851530711
236 0.649372339248657
237 0.643095592657725
238 0.641893704732259
239 0.648103614648183
240 0.635817309220632
241 0.6328125
242 0.629340271155039
243 0.623197118441264
244 0.627737720807393
245 0.623530983924866
246 0.644364317258199
247 0.646901726722717
248 0.636418263117472
249 0.623998403549194
250 0.624332269032796
251 0.630809287230174
252 0.621861656506856
253 0.64636751015981
254 0.644164005915324
255 0.631209929784139
256 0.630408644676208
257 0.632278323173523
258 0.640558222929637
259 0.645633002122243
260 0.633547008037567
261 0.638822118441264
262 0.632678945859273
263 0.631543815135956
264 0.646834909915924
265 0.643763343493144
266 0.629473825295766
267 0.621060371398926
268 0.620325843493144
269 0.639957269032796
270 0.63855501015981
271 0.643963674704234
272 0.640625
273 0.637620190779368
274 0.641159176826477
275 0.641292730967204
276 0.640625
277 0.64122595389684
278 0.642361104488373
279 0.646300752957662
280 0.646501064300537
281 0.646367530028025
282 0.645032048225403
283 0.648170411586761
284 0.651909728844961
};
\addplot [line width=2.5pt, color1, opacity=1.0]
table {%
1 0.0917467946807543
2 0.134548609455427
3 0.160389959812164
4 0.201789528131485
5 0.231036325295766
6 0.251936435699463
7 0.26622595389684
8 0.285056094328562
9 0.308159718910853
10 0.324252128601074
11 0.339009076356888
12 0.364249467849731
13 0.357238243023555
14 0.3809428413709
15 0.406850963830948
16 0.40892094373703
17 0.426282048225403
18 0.433560371398926
19 0.442441244920095
20 0.452724357446035
21 0.455729166666667
22 0.468883544206619
23 0.475827982028325
24 0.486378212769826
25 0.484241465727488
26 0.493189086516698
27 0.495192309220632
28 0.488514959812164
29 0.516626616319021
30 0.506209949652354
31 0.518229166666667
32 0.522235572338104
33 0.522970080375671
34 0.530582269032796
35 0.537994126478831
36 0.535723825295766
37 0.539730250835419
38 0.547208885351817
39 0.543870190779368
40 0.543269236882528
41 0.552751084168752
42 0.547542750835419
43 0.550013363361359
44 0.561631937821706
45 0.56497061252594
46 0.555956204732259
47 0.564903855323792
48 0.568576375643412
49 0.565237720807393
50 0.572182138760885
51 0.572849889596303
52 0.570646345615387
53 0.576322118441264
54 0.578792730967204
55 0.586137811342875
56 0.577924688657125
57 0.582532048225403
58 0.578191757202148
59 0.584401706854502
60 0.592347760995229
61 0.579059839248657
62 0.587272981802622
63 0.577791134516398
64 0.582398494084676
65 0.587406516075134
66 0.586271365483602
67 0.587473273277283
68 0.593482911586761
69 0.590277791023254
70 0.592147429784139
71 0.590144236882528
72 0.596955120563507
73 0.595953524112701
74 0.596955140431722
75 0.590210994084676
76 0.591746787230174
77 0.597355743249257
78 0.605969548225403
79 0.591746787230174
80 0.596087078253428
81 0.596821586290995
82 0.60096154610316
83 0.595419347286224
84 0.596821586290995
85 0.603231847286224
86 0.596554478009542
87 0.606837610403697
88 0.601095080375671
89 0.60423344373703
90 0.599893172581991
91 0.602297008037567
92 0.59829060236613
93 0.598824779192607
94 0.604300220807393
95 0.602697650591532
96 0.605101486047109
97 0.608440160751343
98 0.604700843493144
99 0.600026706854502
100 0.602363804976145
101 0.597088674704234
102 0.599158644676208
103 0.603231827418009
104 0.599358975887299
105 0.604300200939178
106 0.606370190779368
107 0.611778835455577
108 0.604366978009542
109 0.606971164544423
110 0.605635682741801
111 0.607104698816935
112 0.609241445859273
113 0.60423344373703
114 0.607705652713776
115 0.610443373521169
116 0.598223845163981
117 0.607839206854502
118 0.609241465727488
119 0.605301817258199
120 0.603966335455577
121 0.601829588413239
122 0.596287389596303
123 0.609842419624329
124 0.60690438747406
125 0.611645301183065
126 0.604166666666667
127 0.605168263117472
128 0.61144498984019
129 0.601428965727488
130 0.61164532105128
131 0.609374980131785
132 0.609041134516398
133 0.605902771155039
134 0.610443373521169
135 0.611979166666667
136 0.60917466878891
137 0.609041134516398
138 0.604901174704234
139 0.608440180619558
140 0.606904367605845
141 0.605969548225403
142 0.608440160751343
143 0.60997595389684
144 0.609107911586761
145 0.612713674704234
146 0.604767620563507
147 0.614049156506856
148 0.611778855323792
149 0.615584929784139
150 0.616185903549194
151 0.610643684864044
152 0.613448182741801
153 0.612513363361359
154 0.609107911586761
155 0.608173072338104
156 0.608707249164581
157 0.609975973765055
158 0.606436987717946
159 0.618522961934408
160 0.613047540187836
161 0.614316244920095
162 0.613247871398926
163 0.619591335455577
164 0.611578524112701
165 0.609041134516398
166 0.613648513952891
167 0.619391024112701
168 0.615785260995229
169 0.613715271155039
170 0.615184287230174
171 0.609041134516398
172 0.620793263117472
173 0.616853634516398
174 0.61471688747406
175 0.614650130271912
176 0.620192289352417
177 0.618255893389384
178 0.611444969971975
179 0.617721696694692
180 0.617454588413239
181 0.611444969971975
182 0.616052349408468
183 0.614449779192607
184 0.61224627494812
185 0.614449799060822
186 0.608974357446035
187 0.614850441614787
188 0.616319457689921
189 0.615584949652354
190 0.616720100243886
191 0.609041134516398
192 0.615651726722717
193 0.619457801183065
194 0.61431622505188
195 0.619925200939178
196 0.616853614648183
197 0.614182670911153
198 0.611578524112701
199 0.615584949652354
200 0.612112720807393
201 0.611578504244486
202 0.618456184864044
203 0.621060351530711
204 0.625734488169352
205 0.624065160751343
206 0.614449799060822
207 0.619257469971975
208 0.61451655626297
209 0.611044327418009
210 0.62172810236613
211 0.618522961934408
212 0.618656516075134
213 0.619190692901611
214 0.620325843493144
215 0.617988785107931
216 0.616519769032796
217 0.619190712769826
218 0.619391004244486
219 0.623998403549194
220 0.616319437821706
221 0.61698716878891
222 0.618122339248657
223 0.619524558385213
224 0.614850421746572
225 0.616052349408468
226 0.613715291023254
227 0.617454608281453
228 0.617988785107931
229 0.615918795267741
230 0.616252680619558
231 0.616119126478831
232 0.618055562178294
233 0.620058755079905
234 0.619524578253428
235 0.622262299060822
236 0.622262279192607
237 0.624732911586761
238 0.621928413709005
239 0.623864849408468
240 0.618389447530111
241 0.616185903549194
242 0.614583333333333
243 0.6171875
244 0.619324266910553
245 0.619057158629099
246 0.621394236882528
247 0.614650090535482
248 0.615050752957662
249 0.62025906642278
250 0.614583333333333
251 0.617053945859273
252 0.619324247042338
253 0.617855230967204
254 0.618322650591532
255 0.616653323173523
256 0.617988765239716
257 0.616386195023855
258 0.620125532150269
259 0.62045939763387
260 0.618990381558736
261 0.623063544432322
262 0.61658654610316
263 0.620058755079905
264 0.616987188657125
265 0.620192309220632
266 0.624866445859273
267 0.622796475887299
268 0.620860040187836
269 0.619791666666667
270 0.617120722929637
271 0.622061947981516
272 0.621193925539652
273 0.624999980131785
274 0.626869658629099
275 0.623330652713776
276 0.615651706854502
277 0.618923604488373
278 0.618589739004771
279 0.619591355323792
280 0.623263875643412
281 0.626802881558736
282 0.627537389596303
283 0.624599357446035
284 0.619991997877757
};
\addplot [line width=2.5pt, red, opacity=1.0]
table {%
1 0.142294337352117
2 0.200721149643262
3 0.195980235934258
4 0.251535798112551
5 0.3106303413709
6 0.290865391492844
7 0.339943915605545
8 0.381543805201848
9 0.362713684638341
10 0.390892098347346
11 0.424679497877757
12 0.424746265014013
13 0.450587600469589
14 0.469818383455276
15 0.47462605436643
16 0.484308232863744
17 0.50627671678861
18 0.511418282985687
19 0.509748915831248
20 0.532318373521169
21 0.519497871398926
22 0.532585461934408
23 0.54153311252594
24 0.545940160751343
25 0.547208865483602
26 0.538661857446035
27 0.545739849408468
28 0.550814628601074
29 0.561631957689921
30 0.565972228844961
31 0.57298344373703
32 0.569310903549194
33 0.570779919624329
34 0.580328524112701
35 0.578859508037567
36 0.570779919624329
37 0.586004277070363
38 0.577323734760284
39 0.585002680619558
40 0.596354166666667
41 0.590211013952891
42 0.592147429784139
43 0.586338142553965
44 0.581330140431722
45 0.597155451774597
46 0.590811947981516
47 0.59107905626297
48 0.591212590535482
49 0.592147449652354
50 0.603231827418009
51 0.593215803305308
52 0.596020301183065
53 0.593282580375671
54 0.594350973765055
55 0.604300200939178
56 0.610376596450806
57 0.589342951774597
58 0.608106295267741
59 0.598090271155039
60 0.605168263117472
61 0.604166666666667
62 0.610510150591532
63 0.605034728844961
64 0.616119146347046
65 0.612847228844961
66 0.617855230967204
67 0.613782048225403
68 0.620459417502085
69 0.619858423868815
70 0.614249467849731
71 0.616386214892069
72 0.625066777070363
73 0.62540062268575
74 0.628138343493144
75 0.623798072338104
76 0.627069969971975
77 0.620392640431722
78 0.632545411586761
79 0.626268704732259
80 0.624332269032796
81 0.636885682741801
82 0.629407048225403
83 0.634281516075134
84 0.630475421746572
85 0.623464206854502
86 0.632946054140727
87 0.63528311252594
88 0.640758534272512
89 0.639423072338104
90 0.636685371398926
91 0.636885702610016
92 0.643162389596303
93 0.636551817258199
94 0.639890491962433
95 0.642761747042338
96 0.64369656642278
97 0.643162389596303
98 0.64269498984019
99 0.642227570215861
100 0.646367530028025
101 0.647903303305308
102 0.654981315135956
103 0.651375532150269
104 0.64576655626297
105 0.648103654384613
106 0.655381937821706
107 0.651442309220632
108 0.656984508037567
109 0.658520301183065
110 0.655048072338104
111 0.655181606610616
112 0.650106847286224
113 0.655181626478831
114 0.654246807098389
115 0.659722228844961
116 0.656116445859273
117 0.662059307098389
118 0.660189648469289
119 0.654714206854502
120 0.656450311342875
121 0.657518704732259
122 0.660323162873586
123 0.661057690779368
124 0.664797008037567
125 0.667200863361359
126 0.660323162873586
127 0.662059307098389
128 0.659254809220632
129 0.662259618441264
130 0.663728614648183
131 0.660256405671438
132 0.664396365483602
133 0.663728634516398
134 0.660924156506856
135 0.664863785107931
136 0.663060903549194
137 0.665531516075134
138 0.667401174704234
139 0.668269236882528
140 0.670539538065592
141 0.667067309220632
142 0.672075311342875
143 0.671674688657125
144 0.67494656642278
145 0.675080120563507
146 0.672809819380442
147 0.672609508037567
148 0.673477550347646
149 0.674345632394155
150 0.672609508037567
151 0.673344016075134
152 0.675080120563507
153 0.672342399756114
154 0.673076927661896
155 0.673744658629099
156 0.676014959812164
157 0.67454594373703
158 0.672008554140727
159 0.674011747042338
160 0.676148494084676
161 0.675547540187836
162 0.67414528131485
163 0.670739829540253
164 0.675013343493144
165 0.673143704732259
166 0.677150110403697
167 0.677216867605845
168 0.675681094328562
169 0.674946586290995
170 0.67721688747406
171 0.675881405671438
172 0.676816244920095
173 0.67641560236613
174 0.676282048225403
175 0.671006937821706
176 0.678952991962433
177 0.676749487717946
178 0.681824247042338
179 0.676014939943949
180 0.679553965727488
181 0.682024598121643
182 0.680689096450806
183 0.681557158629099
184 0.680221696694692
185 0.681490401426951
186 0.681490381558736
187 0.681156516075134
188 0.679553945859273
189 0.681089739004771
190 0.680956184864044
191 0.67948716878891
192 0.678552329540253
193 0.680488765239716
194 0.682425220807393
195 0.676682690779368
196 0.682091355323792
197 0.681356827418009
198 0.683693905671438
199 0.681891024112701
200 0.684161325295766
201 0.685830672581991
202 0.683360040187836
203 0.681089739004771
204 0.683159728844961
205 0.68255877494812
206 0.681623935699463
207 0.685630341370901
208 0.684628744920095
209 0.68235844373703
210 0.684161305427551
211 0.684228082497915
212 0.686698714892069
213 0.684161325295766
214 0.686364849408468
215 0.683627148469289
216 0.681757469971975
217 0.683693925539652
218 0.682425200939178
219 0.682358423868815
220 0.685830652713776
221 0.679487188657125
222 0.683226505915324
223 0.683092951774597
224 0.683426837126414
225 0.686030964056651
226 0.683760682741801
227 0.682491978009542
228 0.681824227174123
229 0.689369658629099
230 0.680021365483602
231 0.684895833333333
232 0.684228082497915
233 0.68215811252594
234 0.686164538065592
235 0.686164538065592
236 0.6875
237 0.686832249164581
238 0.683894236882528
239 0.683894236882528
240 0.683560351530711
241 0.681089758872986
242 0.684762279192607
243 0.683092971642812
244 0.684161325295766
245 0.686965823173523
246 0.683894236882528
247 0.683159708976746
248 0.68215811252594
249 0.681690692901611
250 0.68422810236613
251 0.686364849408468
252 0.687099357446035
253 0.687032580375671
254 0.683627148469289
255 0.680088142553965
256 0.685964186986287
257 0.683693885803223
258 0.683426817258199
259 0.688234508037567
260 0.68275906642278
261 0.688301285107931
262 0.684628744920095
263 0.682625532150269
264 0.682158132394155
265 0.685630321502686
266 0.68810095389684
267 0.686631957689921
268 0.683092971642812
269 0.685162921746572
270 0.685096164544423
271 0.68502938747406
272 0.687099357446035
273 0.683092931906382
274 0.686298092206319
275 0.687166154384613
276 0.68215811252594
277 0.687232891718547
278 0.685229698816935
279 0.68255877494812
280 0.686565180619558
281 0.687232911586761
282 0.684161325295766
283 0.686965823173523
284 0.683627148469289
};
\addplot [line width=2.5pt, blue, opacity=1.0]
table {%
1 0.143028850356738
2 0.184094548225403
3 0.196848288178444
4 0.23711271584034
5 0.309428413709005
6 0.295539528131485
7 0.320112178723017
8 0.387286325295766
9 0.366853644450506
10 0.410389959812164
11 0.434561967849731
12 0.437700321276983
13 0.463875532150269
14 0.488648504018784
15 0.462139417727788
16 0.504139959812164
17 0.518429478009542
18 0.51221955815951
19 0.530048092206319
20 0.521501064300537
21 0.530048072338104
22 0.55141560236613
23 0.53145033121109
24 0.541599889596303
25 0.561965803305308
26 0.54667466878891
27 0.563234508037567
28 0.567174156506856
29 0.576789518197378
30 0.562700311342875
31 0.569110572338104
32 0.568843464056651
33 0.579660793145498
34 0.569778303305308
35 0.582398494084676
36 0.59168001015981
37 0.597288986047109
38 0.576856315135956
39 0.59127938747406
40 0.593349357446035
41 0.593549688657125
42 0.595352550347646
43 0.598157068093618
44 0.594951907793681
45 0.601028303305308
46 0.591145833333333
47 0.587740401426951
48 0.597622851530711
49 0.583199779192607
50 0.600494126478831
51 0.610376596450806
52 0.598758002122243
53 0.603899558385213
54 0.61511751015981
55 0.608039538065592
56 0.618589758872986
57 0.60463406642278
58 0.617454588413239
59 0.62005877494812
60 0.614650110403697
61 0.621861656506856
62 0.626469035943349
63 0.623263875643412
64 0.627737700939178
65 0.628205120563507
66 0.624799688657125
67 0.624732911586761
68 0.637419879436493
69 0.642694969971975
70 0.643963674704234
71 0.640625
72 0.642027239004771
73 0.642561435699463
74 0.647569437821706
75 0.644764979680379
76 0.644164005915324
77 0.653712610403697
78 0.65297810236613
79 0.651442309220632
80 0.660523494084676
81 0.652243574460348
82 0.650507469971975
83 0.65418001015981
84 0.657986124356588
85 0.659989297389984
86 0.661458333333333
87 0.658386766910553
88 0.659789005915324
89 0.656984508037567
90 0.665932138760885
91 0.66733439763387
92 0.662059287230174
93 0.660590271155039
94 0.663795391718547
95 0.669871807098389
96 0.666866978009542
97 0.670739869276682
98 0.668002148469289
99 0.664396365483602
100 0.671474357446035
101 0.670739849408468
102 0.669270833333333
103 0.670606315135956
104 0.66920405626297
105 0.673744658629099
106 0.672609508037567
107 0.66633282105128
108 0.668068905671438
109 0.669070502122243
110 0.668803413709005
111 0.66940438747406
112 0.667868594328562
113 0.667935351530711
114 0.672676285107931
115 0.673611104488373
116 0.672542730967204
117 0.669070502122243
118 0.665865361690521
119 0.665331204732259
120 0.664196034272512
121 0.661858975887299
122 0.658319969971975
123 0.657919327418009
124 0.661458333333333
125 0.662059267361959
126 0.662192821502686
127 0.659121255079905
128 0.659188032150269
129 0.657385150591532
130 0.657184839248657
131 0.661792198816935
132 0.655982891718547
133 0.653512279192607
134 0.662192841370901
135 0.656784196694692
136 0.657251596450806
137 0.652911325295766
138 0.656583865483602
139 0.655916114648183
140 0.660657068093618
141 0.652310371398926
142 0.654380341370901
143 0.667267620563507
144 0.672876576582591
145 0.669070521990458
146 0.668402771155039
147 0.672475973765055
148 0.673744658629099
149 0.672342419624329
150 0.678218464056651
151 0.672943373521169
152 0.674746255079905
153 0.673611124356588
154 0.675280451774597
155 0.672208865483602
156 0.673944969971975
157 0.673544327418009
158 0.679821054140727
159 0.674813032150269
160 0.674612720807393
161 0.673878212769826
162 0.675480763117472
163 0.675614317258199
164 0.673878212769826
165 0.675547540187836
166 0.674145301183065
167 0.672542730967204
168 0.666065692901611
169 0.672943393389384
170 0.672943393389384
171 0.672542730967204
172 0.673677881558736
173 0.668936967849731
174 0.670806626478831
175 0.674078524112701
176 0.67394498984019
177 0.672409196694692
178 0.671875
179 0.670739849408468
180 0.668202459812164
181 0.674212078253428
182 0.67127404610316
183 0.675347228844961
184 0.67474627494812
185 0.673811415831248
186 0.670272449652354
187 0.673811435699463
188 0.667134086290995
189 0.673076927661896
190 0.676348805427551
191 0.671808222929637
192 0.671274026234945
193 0.673344035943349
194 0.670739849408468
195 0.672743042310079
196 0.66920405626297
197 0.667668263117472
198 0.665464739004771
199 0.669137299060822
200 0.667000532150269
201 0.67127404610316
202 0.674612700939178
203 0.668736636638641
204 0.667401174704234
205 0.665998915831248
206 0.671674688657125
207 0.669871787230174
208 0.667534728844961
209 0.670606315135956
210 0.670940160751343
211 0.667935351530711
212 0.671140491962433
213 0.666065692901611
214 0.670205652713776
215 0.673811435699463
216 0.67434561252594
217 0.673143684864044
218 0.675213674704234
219 0.673143704732259
220 0.671874980131785
221 0.676615913709005
222 0.673277239004771
223 0.6796875
224 0.67514689763387
225 0.68175748984019
226 0.675080120563507
227 0.675614337126414
228 0.676482359568278
229 0.677884618441264
230 0.671006937821706
231 0.674479186534882
232 0.678218503793081
233 0.675948182741801
234 0.67701655626297
235 0.675547520319621
236 0.68028845389684
237 0.67247595389684
238 0.678352038065592
239 0.677283664544423
240 0.673944969971975
241 0.677150110403697
242 0.673544327418009
243 0.674412409464518
244 0.674011747042338
245 0.674278835455577
246 0.678552369276682
247 0.678218483924866
248 0.673544327418009
249 0.673076907793681
250 0.676148494084676
251 0.674879809220632
252 0.680154899756114
253 0.676816244920095
254 0.675146917502085
255 0.676348825295766
256 0.675881405671438
257 0.671340803305308
258 0.677617530028025
259 0.67701655626297
260 0.677283664544423
261 0.678485572338104
262 0.672676285107931
263 0.67761751015981
264 0.673277258872986
265 0.67721688747406
266 0.677951395511627
267 0.675814628601074
268 0.675480763117472
269 0.675614297389984
270 0.673677901426951
271 0.674145301183065
272 0.678485592206319
273 0.675614297389984
274 0.674412389596303
275 0.678418795267741
276 0.675881405671438
277 0.679220080375671
278 0.676348825295766
279 0.674813032150269
280 0.676883021990458
281 0.676014959812164
282 0.675480763117472
283 0.678819437821706
284 0.67721688747406
};
\addplot [line width=2.5pt, color2, opacity=1.0]
table {%
2 0.134147971868515
4 0.197248935699463
6 0.238715281089147
8 0.300814648469289
10 0.348691234985987
12 0.367321034272512
14 0.349959939718246
16 0.415464739004771
18 0.423477560281754
20 0.437166134516398
22 0.456597228844961
24 0.482572118441264
26 0.503205140431722
28 0.505876084168752
30 0.511151194572449
32 0.533653855323792
34 0.542200843493144
36 0.553485572338104
38 0.560096144676208
40 0.568442841370901
42 0.575186967849731
44 0.586404919624329
46 0.590611636638641
48 0.596287389596303
50 0.600427369276682
52 0.60136216878891
54 0.605435351530711
56 0.606236636638641
58 0.606169859568278
60 0.607171475887299
62 0.61184561252594
64 0.61264689763387
66 0.610176285107931
68 0.613648513952891
70 0.607305030028025
72 0.610176265239716
74 0.614983995755514
76 0.615050752957662
78 0.612980763117472
80 0.614316244920095
82 0.609708865483602
84 0.616052349408468
86 0.613581736882528
88 0.617454608281453
90 0.614449799060822
92 0.613648494084676
94 0.620592951774597
96 0.616185903549194
98 0.616185903549194
100 0.617120722929637
102 0.616052349408468
104 0.610710461934408
106 0.615317841370901
108 0.622662941614787
110 0.620259086290995
112 0.615050752957662
114 0.616452991962433
116 0.618322650591532
118 0.617254277070363
120 0.616987188657125
122 0.616920391718547
124 0.618255873521169
126 0.623864829540253
128 0.618923604488373
130 0.614115913709005
132 0.615918795267741
134 0.617654919624329
136 0.623063564300537
138 0.622262279192607
140 0.621394236882528
142 0.615518152713776
144 0.617454588413239
146 0.621461013952891
148 0.625066777070363
150 0.618790070215861
152 0.619991978009542
154 0.621461013952891
156 0.621327459812164
158 0.620793263117472
160 0.621127128601074
162 0.625801265239716
164 0.620526154836019
166 0.619524558385213
168 0.620993594328562
170 0.620993594328562
172 0.61925748984019
174 0.621394236882528
176 0.623864869276682
178 0.61658654610316
180 0.620125532150269
182 0.62212872505188
184 0.621527771155039
186 0.616519769032796
188 0.622796475887299
190 0.621661325295766
192 0.617855230967204
194 0.623531003793081
196 0.623197118441264
198 0.620993594328562
200 0.620058755079905
202 0.620392640431722
204 0.621527791023254
206 0.619925220807393
208 0.627804478009542
210 0.622662921746572
212 0.619524558385213
214 0.620058755079905
216 0.621995190779368
218 0.622729698816935
220 0.622128744920095
222 0.623197138309479
224 0.621394217014313
226 0.620659708976746
228 0.622729698816935
230 0.623063564300537
232 0.618456204732259
234 0.617988785107931
236 0.622195502122243
238 0.620192309220632
240 0.622061967849731
242 0.625066777070363
244 0.623597741127014
246 0.61965811252594
248 0.617855250835419
250 0.619457801183065
252 0.618255873521169
254 0.623263895511627
256 0.621594548225403
258 0.626535793145498
260 0.62232905626297
262 0.624399026234945
264 0.618723293145498
266 0.623063564300537
268 0.621327459812164
270 0.625133554140727
272 0.618055562178294
274 0.624265491962433
276 0.627737700939178
278 0.625734508037567
280 0.621127148469289
282 0.617321034272512
284 0.624799688657125
};
\end{axis}

\end{tikzpicture}

%% file: labpal/figure_data/performance_comparison/CIFAR-100/CIFAR-100_ResNet-20_test_accuracy.pgf
% This file was created by tikzplotlib v0.9.8.
\begin{tikzpicture}

\definecolor{color0}{rgb}{0.647058823529412,0.164705882352941,0.164705882352941}
\definecolor{color1}{rgb}{0.933333333333333,0.509803921568627,0.933333333333333}
\definecolor{color2}{rgb}{1,0.647058823529412,0}

\begin{axis}[
yticklabel style={font=\large},
xticklabel style={font=\large},
ytick style={font=\Large},
xtick style={font=\Large},
ylabel style={font=\Large},
xlabel style={font=\Large},
title style={font=\Large},
grid = major,
major grid style={dotted},
legend cell align={left},
legend style={
  fill opacity=0.8,
  draw opacity=1,
  text opacity=1,
  at={(0.91,0.5)},
  anchor=east,
  draw=white!80!black
},
minor xtick={},
minor ytick={},
reverse legend, legend cell align={left}, legend style={ fill opacity=0.8, draw opacity=1, text opacity=1, at={(0.94,0.3)}, anchor=east, draw=white!80!black},
tick align=outside,
tick pos=left,
title={test accuracy CIFAR-100 ResNet-20},
width=10.5cm,height=8cm,,
x grid style={white!69.0196078431373!black},
xlabel={epoch with best val. acc.},
xmin=-0.055, xmax=0.055,
xtick style={color=black},
xmajorticks=false,
y grid style={white!69.0196078431373!black},
ylabel={test accuracy},
ymin=0.3, ymax=0.69,
ytick style={color=black},
minor y tick num=3
]
\addplot [draw=color0, fill=color0, mark=-, only marks, mark options={scale=3},line width=3pt]
table{%
x  y
-0.006 0.604800999164581
0 -1
};
\addlegendentry{PLS : $\alpha$: 0.1, $c_{W}$: 0.1, c1: 0.4, $\beta$: 0.0}
\addplot [draw=green!50.1960784313725!black, fill=green!50.1960784313725!black, mark=-, only marks, mark options={scale=3},line width=3pt]
table{%
x  y
0 0.630008002122243
0 -1
};
\addlegendentry{SGD : $\lambda$: 0.1, $\beta$: 0.9}
\addplot [draw=black, fill=black, mark=-, only marks, mark options={scale=3},line width=3pt]
table{%
x  y
0 0.557124733924866
0 -1
};
\addlegendentry{SLS : c: 0.1, $\beta$: 0.9, $\mu$: 0.1}
\addplot [draw=color1, fill=color1, mark=-, only marks, mark options={scale=3},line width=3pt]
table{%
x  y
0 0.604066510995229
0 -1
};
\addlegendentry{GOLSI : $c$: 0.99, $\eta$: 2.0, $\alpha$: 0.1, $\beta$: 0.4}
\addplot [draw=color2, fill=color2, mark=-, only marks, mark options={scale=3},line width=3pt]
table{%
x  y
0 0.572482645511627
0 -1
};
\addlegendentry{PAL : $\alpha$: 1.0, $\mu$: 1.0, $\beta$: 0.4 }
\addplot [draw=red, fill=red, mark=-, only marks, mark options={scale=3},line width=3pt]
table{%
x  y
0 0.61535123984019
0 -1
};
\addlegendentry{LABPAL-NSGD: $|\mathbb{B}_a|$: 1280, $\beta$: 0.0, $\alpha$: 1.9, n: 1000}
\addplot [draw=blue, fill=blue, mark=-, only marks, mark options={scale=3},line width=3pt]
table{%
x  y
-0.006 0.629440426826477
0 -1
};
\addlegendentry{LABPAL-SGD: $|\mathbb{B}_a|$: 1280, $\beta$: 0.0, $\alpha$: 1.9, n: 1000}s
\end{axis}

\end{tikzpicture}

%% file: labpal/figure_data/performance_comparison/CIFAR-100/CIFAR-100_MobileNet-V2_test_accuracy.pgf
% This file was created by tikzplotlib v0.9.8.
\begin{tikzpicture}

\definecolor{color1}{rgb}{0.933333333333333,0.509803921568627,0.933333333333333}
\definecolor{color2}{rgb}{1,0.647058823529412,0}

\begin{axis}[
yticklabel style={font=\large},
xticklabel style={font=\large},
ytick style={font=\Large},
xtick style={font=\Large},
ylabel style={font=\Large},
xlabel style={font=\Large},
title style={font=\Large},
grid = major,
major grid style={dotted},
minor xtick={},
minor ytick={},
tick align=outside,
tick pos=left,
title={test accuracy CIFAR-100 MobileNet-V2},
width=10.5cm,height=8cm,,
x grid style={white!69.0196078431373!black},
xlabel={epoch with best val. acc.},
xmin=-0.055, xmax=0.055,
xtick style={color=black},
xmajorticks=false,
y grid style={white!69.0196078431373!black},
ylabel={test accuracy},
ymin=0.55, ymax=0.75,
ytick style={color=black},
minor y tick num=3
]
\addplot [draw=green!50.1960784313725!black, fill=green!50.1960784313725!black, mark=-, only marks, mark options={scale=3},line width=3pt]
table{%
x  y
-0.006 0.730969548225403
0 -1
};
\addplot [draw=black, fill=black, mark=-, only marks, mark options={scale=3},line width=3pt]
table{%
x  y
0 0.517995456854502
0 -1
};
\addplot [draw=color1, fill=color1, mark=-, only marks, mark options={scale=3},line width=3pt]
table{%
x  y
-0.006 0.697115381558736
0 -1
};
\addplot [draw=red, fill=red, mark=-, only marks, mark options={scale=3},line width=3pt]
table{%
x  y
0 0.730235040187836
0 -1
};
\addplot [draw=blue, fill=blue, mark=-, only marks, mark options={scale=3},line width=3pt]
table{%
x  y
0 0.708900908629099
0 -1
};
\addplot [draw=color2, fill=color2, mark=-, only marks, mark options={scale=3},line width=3pt]
table{%
x  y
0 0.699953258037567
0 -1
};
\end{axis}

\end{tikzpicture}

%% file: labpal/figure_data/performance_comparison/CIFAR-100/CIFAR-100_DenseNet-121_test_accuracy.pgf
% This file was created by tikzplotlib v0.9.8.
\begin{tikzpicture}

\definecolor{color0}{rgb}{0.647058823529412,0.164705882352941,0.164705882352941}
\definecolor{color1}{rgb}{0.933333333333333,0.509803921568627,0.933333333333333}
\definecolor{color2}{rgb}{1,0.647058823529412,0}

\begin{axis}[
yticklabel style={font=\large},
xticklabel style={font=\large},
ytick style={font=\Large},
xtick style={font=\Large},
ylabel style={font=\Large},
xlabel style={font=\Large},
title style={font=\Large},
grid = major,
major grid style={dotted},
minor xtick={},
minor ytick={},
tick align=outside,
tick pos=left,
title={test accuracy CIFAR-100 DenseNet-121},
width=10.5cm,height=8cm,,
x grid style={white!69.0196078431373!black},
xlabel={epoch with best val. acc.},
xmin=-0.055, xmax=0.055,
xtick style={color=black},
xmajorticks=false,
y grid style={white!69.0196078431373!black},
ylabel={test accuracy},
ymin=0.55, ymax=0.75,
ytick style={color=black},
minor y tick num=3
]
\addplot [draw=color0, fill=color0, mark=-, only marks, mark options={scale=3},line width=3pt]
table{%
x  y
0 0.628939648469289
0 -1
};
%\addlegendentry{PLS : $\alpha$: 0.0001, $c-{W}$: 0.2, c1: 0.005, $\beta$: 0.0}
\addplot [draw=green!50.1960784313725!black, fill=green!50.1960784313725!black, mark=-, only marks, mark options={scale=3},line width=3pt]
table{%
x  y
0 0.686665336290995
0 -1
};
%\addlegendentry{SGD : $\lambda$: 0.1, $\beta$: 0.0}
\addplot [draw=black, fill=black, mark=-, only marks, mark options={scale=3},line width=3pt]
table{%
x  y
0 0.680889427661896
0 -1
};
%\addlegendentry{SLS : c: 0.1, $\beta$: 0.9, $\mu$: 0.1}
\addplot [draw=color1, fill=color1, mark=-, only marks, mark options={scale=3},line width=3pt]
table{%
x  y
0 0.641659994920095
0 -1
};
%\addlegendentry{GOLSI : $c$: 0.9, $\eta$: 2.0, $\alpha$: 0.0001, $\beta$: 0.0}
\addplot [draw=red, fill=red, mark=-, only marks, mark options={scale=3},line width=3pt]
table{%
x  y
0 0.703926285107931
0 -1
};
%\addlegendentry{LABPAL-NSGD : $\epsilon$: 1.0}
\addplot [draw=blue, fill=blue, mark=-, only marks, mark options={scale=3},line width=3pt]
table{%
x  y
0 0.692508021990458
0 -1
};
%\addlegendentry{LABPAL-SGD : $\epsilon$: 1.0}
\addplot [draw=color2, fill=color2, mark=-, only marks, mark options={scale=3},line width=3pt]
table{%
x  y
0 0.647369126478831
0 -1
};
%\addlegendentry{PAL : $\alpha$: 1.66, $\mu$: 1.0}
\end{axis}

\end{tikzpicture}

%% file: labpal/fig_batch_size_comparison.tex
\begin{figure}[t!]
	\tikzsetfigurename{labpal_performance_comp_batch_size10}
	\centering
	\vspace{-0.5cm}
	\def\scale{0.4}
	\begin{tabular}{ c c c}
		\scalebox{\scale}{\input{"labpal/figure_data/performance_comparison/CIFAR-10/CIFAR-10_DenseNet-121_learning_rate.pgf"}}&	
		\scalebox{\scale}{\input{"labpal/figure_data/performance_comparison_low_batch/plots_bs32/CIFAR-10_DenseNet-121_learning_rate2.pgf"}}&
		\scalebox{\scale}{\input{"labpal/figure_data/performance_comparison_low_batch/plots_bs8/CIFAR-10_DenseNet-121_learning_rate2.pgf"}}\\
		\scalebox{\scale}{\input{"labpal/figure_data/performance_comparison/CIFAR-10/CIFAR-10_DenseNet-121_validation_accuracy2.pgf"}}&
		\scalebox{\scale}{\input{"labpal/figure_data/performance_comparison_low_batch/plots_bs32/CIFAR-10_DenseNet-121_validation_accuracy2.pgf"}}&
		\scalebox{\scale}{\input{"labpal/figure_data/performance_comparison_low_batch/plots_bs8/CIFAR-10_DenseNet-121_validation_accuracy2.pgf"}}
	\end{tabular}
	\vspace{-0.3cm}
	\caption{Performance comparison of a \textbf{DenseNet-121}, trained on \textbf{CIFAR-10} for \textbf{batch sizes 128, 32 and 8}. Hyperparameters are not changed.   (For the evaluation on ResNet-20 and MobileNet-V2 see Appendix Figure \ref{labpal_fig_opt_comparison_bs32} \& \ref{labpal_fig_opt_comparison_bs8}). PLS curves are incomplete since training failed. Training steps are increased by a factor of 4 and 16 for batch size 32 and 8 respectively.}
	\label{labpal_fig_performance_comp_batch_size}
	\vspace{-0.5cm}
\end{figure}

%% file: labpal/figure_data/performance_comparison_low_batch/plots_bs32/CIFAR-10_DenseNet-121_learning_rate2.pgf
% This file was created by tikzplotlib v0.9.8.
\begin{tikzpicture}

\definecolor{color0}{rgb}{0.933333333333333,0.509803921568627,0.933333333333333}
\definecolor{color1}{rgb}{0.647058823529412,0.164705882352941,0.164705882352941}
\definecolor{color2}{rgb}{1,0.647058823529412,0}

\begin{axis}[
,
log basis y={10},
minor xtick={},
minor ytick={},
tick align=outside,
tick pos=left,
title={batch size 32},
title style={font=\Large,thick},
width=10.5cm,height=8cm,grid=major,major grid style={dotted},
reverse legend, legend cell align={left}, legend style={ fill opacity=0.8, draw opacity=1, text opacity=1, at={(0.98,0.28)}, anchor=east, draw=white!80!black},
x grid style={white!69.0196078431373!black},
xlabel={steps},
xmin=-19949.95, xmax=418970.95,
xtick style={color=black},
xtick={-50000,0,50000,100000,150000,200000,250000,300000,350000,400000,450000},
y grid style={white!69.0196078431373!black},
ylabel={learning rate},
ymin=10e-12, ymax=10e5,
ymode=log,
ytick style={color=black},
ytick={1e-14,1e-12,1e-10,1e-08,1e-06,0.0001,0.01,1,100,10000,1000000}
]
\addplot [line width=2.5pt, color0, opacity=1.0]
table {%
1 0.100000001490116
2777 1.38721084594727
5534 1.60026228427887
8313 0.673967063426971
11046 0.982671856880188
13759 0.97798103094101
16469 1.15661287307739
19178 1.15527653694153
21866 0.403303474187851
24549 0.862769901752472
27252 0.806399405002594
29915 1.05899012088776
32609 1.39318013191223
35243 0.638906121253967
37865 0.711150884628296
40507 1.04813981056213
43156 1.89725399017334
45788 1.59754800796509
48423 1.22970163822174
51025 1.1757345199585
53645 0.95950448513031
56297 1.78477644920349
58912 0.821375668048859
61565 0.859544098377228
64177 1.46984827518463
66816 1.11813986301422
69442 0.880754292011261
72067 0.943892776966095
74683 1.12780153751373
77289 1.14587390422821
79908 0.903947174549103
82498 2.04232907295227
85097 1.2383748292923
87698 1.42974078655243
90307 0.90019816160202
92906 1.09290146827698
95498 0.690349996089935
98094 1.54164326190948
100704 0.933300018310547
103300 1.35156869888306
105906 0.743186891078949
108500 0.874322533607483
111074 0.790430128574371
113668 0.537707448005676
116236 1.77401387691498
118820 1.36024129390717
121375 0.757489264011383
123978 1.21643459796906
126576 0.885730087757111
129146 1.72258472442627
131736 1.23126828670502
134310 1.03684592247009
136899 0.974213123321533
139467 1.11130940914154
142048 1.20939469337463
144623 1.04247891902924
147199 1.93937695026398
149777 2.66032218933105
152363 1.53151440620422
154921 0.941340386867523
157516 0.856026887893677
160133 1.44167864322662
162712 0.892515122890472
165341 0.7824667096138
167913 2.8176589012146
170456 2.65221285820007
173027 1.37965846061707
175625 1.09883522987366
178182 1.65511834621429
180748 1.24703335762024
183352 0.931830644607544
185920 0.931965470314026
188552 1.47914910316467
191145 1.73799216747284
193695 2.92869019508362
196284 3.25437331199646
198872 1.09675788879395
201470 2.56773519515991
204048 1.32815742492676
206631 7.56103134155273
209217 2.07933902740479
211786 0.475111156702042
214347 0.821169912815094
216942 2.5433452129364
219511 1.94269287586212
222123 1.37682926654816
224680 2.2419102191925
227299 2.05890226364136
229874 1.01553797721863
232471 3.50288200378418
235119 0.933994054794312
237732 0.941258728504181
240353 1.46326804161072
242992 2.37866044044495
245596 1.16441607475281
248185 1.69062840938568
250765 1.31647396087646
253375 1.90005481243134
255981 1.09865009784698
258584 1.59387755393982
261230 1.07727456092834
263877 2.88259053230286
266509 0.911364018917084
269163 0.0754771530628204
271771 1.74833345413208
274386 3.00834226608276
276998 2.10016512870789
279655 2.95104050636292
282321 0.733536601066589
284952 1.11500477790833
287594 9.99999993922529e-09
289676 9.99999993922529e-09
291766 9.99999993922529e-09
293849 9.99999993922529e-09
295936 9.99999993922529e-09
298029 9.99999993922529e-09
300129 9.99999993922529e-09
302228 9.99999993922529e-09
304316 9.99999993922529e-09
306396 9.99999993922529e-09
308496 9.99999993922529e-09
310586 9.99999993922529e-09
312666 9.99999993922529e-09
314773 9.99999993922529e-09
316858 9.99999993922529e-09
318938 9.99999993922529e-09
321024 9.99999993922529e-09
323118 9.99999993922529e-09
325197 9.99999993922529e-09
327283 9.99999993922529e-09
329385 9.99999993922529e-09
331479 9.99999993922529e-09
333583 9.99999993922529e-09
335667 9.99999993922529e-09
337762 9.99999993922529e-09
339860 9.99999993922529e-09
341949 9.99999993922529e-09
344040 9.99999993922529e-09
346110 9.99999993922529e-09
348202 9.99999993922529e-09
350284 9.99999993922529e-09
352362 9.99999993922529e-09
354433 9.99999993922529e-09
356511 9.99999993922529e-09
358600 9.99999993922529e-09
360701 9.99999993922529e-09
362796 9.99999993922529e-09
364884 9.99999993922529e-09
366959 9.99999993922529e-09
369057 9.99999993922529e-09
371145 9.99999993922529e-09
373235 9.99999993922529e-09
375333 9.99999993922529e-09
377406 9.99999993922529e-09
379503 9.99999993922529e-09
381598 9.99999993922529e-09
383673 9.99999993922529e-09
385760 9.99999993922529e-09
387860 9.99999993922529e-09
389948 9.99999993922529e-09
392044 9.99999993922529e-09
394134 9.99999993922529e-09
396234 9.99999993922529e-09
398327 9.99999993922529e-09
};
\addlegendentry{GOLSI : $c$: 0.99, $\eta$: 2.0, $\alpha$: 0.1, $\beta$: 0.4}
\addplot [line width=2.5pt, color1, opacity=1.0]
table {%
1 0
2750 0.235584909717242
4345 0.483907725972434
5882 2.279399116834
7036 3.16166504224141
8095 5.02711582183838
9229 5.39130369822184
10477 6.01416190465291
11895 7.06610790888468
13180 7.06610790888468
14408 7.06610790888468
15635 8.53778060277303
17226 9.58156855901082
19026 10.341699441274
20249 6.5814646879832
21249 7.74469335873922
22249 7.74469335873922
};
\addlegendentry{PLS : $\alpha$: 0.1, $c_{W}$: 0.1, c1: 0.4, $\beta$: 0.0}
\addplot [line width=2.5pt, color2, opacity=1.0]
table {%
2 0.0623696732024352
2002 2.8521430293719
4002 2.80771772066752
6002 2.10934221744537
8002 5.53824353218079
10002 2.16535111268361
12002 3.72148243586222
14002 3.72570443153381
16002 5.13554501533508
18002 3.75440621376038
20002 3.58941793441772
22002 2.83069588740667
24002 3.40243943532308
26002 2.57252317667007
28002 4.21447436014811
30002 1.96283562978109
32002 3.03545316060384
34002 3.13685043652852
36002 3.38537653287252
38002 3.06387921174367
40002 2.38904468218485
42002 3.52945033709208
44002 1.82444129387538
46002 2.3024609486262
48002 2.90450100104014
50002 1.99535282452901
52002 2.05859417716662
54002 1.21006976316373
56002 1.5117921034495
58002 2.55598851044973
60002 2.11390721797943
62002 1.7471094528834
64002 2.75682727495829
66002 1.32595018545787
68002 1.7887673775355
70002 1.55318876107534
72002 1.80467879772186
74002 1.87822461128235
76002 2.2381720940272
78002 1.60379175345103
80002 2.01481266816457
82002 1.07392591238022
84002 1.51752916971842
86002 1.26203634341558
88002 1.54924952983856
90002 1.277943054835
92002 1.51651702324549
94002 1.24283691247304
96002 1.92141052087148
98002 0.988869428634644
100002 1.01921153068542
102002 1.01621601978938
104002 1.2845952908198
106002 0.983288695414861
108002 1.09987314542135
110002 1.77665090560913
112002 1.28267272313436
114002 1.37258938948313
116002 1.82566010951996
118002 1.17869818210602
120002 1.09554813305537
122002 1.81438636779785
124002 1.03169500827789
126002 1.05195885896683
128002 1.99723293383916
130002 1.40263464053472
132002 1.14564088980357
134002 1.24003211657206
136002 1.16686060031255
138002 3.05071576436361
140002 1.4908371369044
142002 1.14773291349411
144002 1.1458881298701
146002 3.1065582036972
148002 1.29933772484461
150002 1.10556777318319
152002 1.75915977358818
154002 1.05029918750127
156002 5.21462168296178
158002 0.477855424086253
160002 1.34721054633458
162002 2.27647272745768
164002 1.19807318846385
166002 0.932421088218689
168002 0.541910285751025
170002 2.04665903250376
172002 2.50551984707514
174002 1.8195081949234
176002 2.2287202278773
178002 1.41230384508769
180002 3.41319364309311
182002 5.18060700098674
184002 3.17123254140218
186002 1.93754661083221
188002 1.59503936767578
190002 2.1969145933787
192002 0.784367342789968
194002 0.789279898007711
196002 0.835910618305206
198002 3.79894111553828
200002 4.58587241172791
202002 0.956696768601735
204002 1.6825248003006
206002 1.21305894851685
208002 2.27978285153707
210002 1.83060216903687
212002 5.41654102007548
214002 6.20732927322388
216002 4.24254369735718
218002 2.4130943218867
220002 3.16114844878515
222002 0.915796538194021
224002 0.898742437362671
226002 1.05133581161499
228002 2.60008760293325
230002 1.11645632982254
232002 1.50339901447296
234002 2.24126555522283
236002 0.883228758970896
238002 3.08848468462626
240002 1.94926549990972
242002 0.93618913491567
244002 5.5878381729126
246002 2.31108661492666
248002 1.01800239086151
250002 1.36005008220673
252002 2.14081064860026
254002 3.97190161546071
256002 2.68489807844162
258002 2.61391909917196
260002 3.8522390127182
262002 2.73476309577624
264002 1.32810996969541
266002 0.871612350145976
268002 2.21251275142034
270002 1.27858068545659
272002 4.44485851128896
274002 5.59790617227554
276002 4.10061691204707
278002 2.973978916804
280002 1.3393040895462
282002 4.1113575498263
284002 1.85389616092046
286002 2.25054359436035
288002 3.31631028652191
290002 1.82894893487295
292002 0.641110897064209
294002 2.84200127919515
296002 0.984409213066101
298002 0.875784079233805
300002 4.13979522387187
302002 1.25967820485433
304002 0.921518166859945
306002 3.27926882108053
308002 1.64634170134862
310002 2.28677868843079
312002 1.20418494939804
314002 3.13542755444845
316002 0.735578298568726
318002 3.79727919896444
320002 2.33287264903386
322002 0.589277565479279
324002 0.302757680416107
326002 1.9994576772054
328002 1.40248507261276
330002 2.13966675599416
332002 2.05940318107605
334002 5.73937400182088
336002 4.5270624756813
338002 1.29657697677612
340002 1.84940695762634
342002 9.67708750565847
344002 3.22831877072652
346002 0.697071929772695
348002 1.66028889020284
350002 0.727725982666016
352002 0.641598244508108
354002 1.10220354795456
356002 0.798902690410614
358002 4.33519568045934
360002 1.79733653863271
362002 2.04673361778259
364002 7.68455090125402
366002 2.94216016928355
368002 5.90727748473485
370002 1.50649237632751
372002 12.3695233662923
374002 1.00670659542084
376002 3.75359582901001
378002 1.29866156975428
380002 3.41389479239782
382002 1.17355269193649
384002 1.59191280603409
386002 2.70432813962301
388002 5.52570756276449
390002 1.45218481620153
392002 0.788146555423737
394002 26.1854600509008
396002 1.43707389136155
398002 2.1672588189443
};\addlegendentry{PAL : $\alpha$: 1.0, $\mu$: 1.0, $\beta$: 0.4 }
\addplot [line width=2.5pt, green!50.1960784313725!black, opacity=1.0]
table {%
1 0.100000001490116
1001 0.100000001490116
2001 0.100000001490116
3001 0.100000001490116
4001 0.100000001490116
5001 0.100000001490116
6001 0.100000001490116
7001 0.100000001490116
8001 0.100000001490116
9001 0.100000001490116
10001 0.100000001490116
11001 0.100000001490116
12001 0.100000001490116
13001 0.100000001490116
14001 0.100000001490116
15001 0.100000001490116
16001 0.100000001490116
17001 0.100000001490116
18001 0.100000001490116
19001 0.100000001490116
20001 0.100000001490116
21001 0.100000001490116
22001 0.100000001490116
23001 0.100000001490116
24001 0.100000001490116
25001 0.100000001490116
26001 0.100000001490116
27001 0.100000001490116
28001 0.100000001490116
29001 0.100000001490116
30001 0.100000001490116
31001 0.100000001490116
32001 0.100000001490116
33001 0.100000001490116
34001 0.100000001490116
35001 0.100000001490116
36001 0.100000001490116
37001 0.100000001490116
38001 0.100000001490116
39001 0.100000001490116
40001 0.100000001490116
41001 0.100000001490116
42001 0.100000001490116
43001 0.100000001490116
44001 0.100000001490116
45001 0.100000001490116
46001 0.100000001490116
47001 0.100000001490116
48001 0.100000001490116
49001 0.100000001490116
50001 0.100000001490116
51001 0.100000001490116
52001 0.100000001490116
53001 0.100000001490116
54001 0.100000001490116
55001 0.100000001490116
56001 0.100000001490116
57001 0.100000001490116
58001 0.100000001490116
59001 0.100000001490116
60001 0.100000001490116
61001 0.100000001490116
62001 0.100000001490116
63001 0.100000001490116
64001 0.100000001490116
65001 0.100000001490116
66001 0.100000001490116
67001 0.100000001490116
68001 0.100000001490116
69001 0.100000001490116
70001 0.100000001490116
71001 0.100000001490116
72001 0.100000001490116
73001 0.100000001490116
74001 0.100000001490116
75001 0.100000001490116
76001 0.100000001490116
77001 0.100000001490116
78001 0.100000001490116
79001 0.100000001490116
80001 0.100000001490116
81001 0.100000001490116
82001 0.100000001490116
83001 0.100000001490116
84001 0.100000001490116
85001 0.100000001490116
86001 0.100000001490116
87001 0.100000001490116
88001 0.100000001490116
89001 0.100000001490116
90001 0.100000001490116
91001 0.100000001490116
92001 0.100000001490116
93001 0.100000001490116
94001 0.100000001490116
95001 0.100000001490116
96001 0.100000001490116
97001 0.100000001490116
98001 0.100000001490116
99001 0.100000001490116
100001 0.100000001490116
101001 0.100000001490116
102001 0.100000001490116
103001 0.100000001490116
104001 0.100000001490116
105001 0.100000001490116
106001 0.100000001490116
107001 0.100000001490116
108001 0.100000001490116
109001 0.100000001490116
110001 0.100000001490116
111001 0.100000001490116
112001 0.100000001490116
113001 0.100000001490116
114001 0.100000001490116
115001 0.100000001490116
116001 0.100000001490116
117001 0.100000001490116
118001 0.100000001490116
119001 0.100000001490116
120001 0.100000001490116
121001 0.100000001490116
122001 0.100000001490116
123001 0.100000001490116
124001 0.100000001490116
125001 0.100000001490116
126001 0.100000001490116
127001 0.100000001490116
128001 0.100000001490116
129001 0.100000001490116
130001 0.100000001490116
131001 0.100000001490116
132001 0.100000001490116
133001 0.100000001490116
134001 0.100000001490116
135001 0.100000001490116
136001 0.100000001490116
137001 0.100000001490116
138001 0.100000001490116
139001 0.100000001490116
140001 0.100000001490116
141001 0.100000001490116
142001 0.100000001490116
143001 0.100000001490116
144001 0.100000001490116
145001 0.100000001490116
146001 0.100000001490116
147001 0.100000001490116
148001 0.100000001490116
149001 0.100000001490116
150001 0.100000001490116
151001 0.100000001490116
152001 0.100000001490116
153001 0.100000001490116
154001 0.100000001490116
155001 0.100000001490116
156001 0.100000001490116
157001 0.100000001490116
158001 0.100000001490116
159001 0.100000001490116
160001 0.100000001490116
161001 0.100000001490116
162001 0.100000001490116
163001 0.100000001490116
164001 0.100000001490116
165001 0.100000001490116
166001 0.100000001490116
167001 0.100000001490116
168001 0.100000001490116
169001 0.100000001490116
170001 0.100000001490116
171001 0.100000001490116
172001 0.100000001490116
173001 0.100000001490116
174001 0.100000001490116
175001 0.100000001490116
176001 0.100000001490116
177001 0.100000001490116
178001 0.100000001490116
179001 0.100000001490116
180001 0.100000001490116
181001 0.100000001490116
182001 0.100000001490116
183001 0.100000001490116
184001 0.100000001490116
185001 0.100000001490116
186001 0.100000001490116
187001 0.100000001490116
188001 0.100000001490116
189001 0.100000001490116
190001 0.100000001490116
191001 0.100000001490116
192001 0.100000001490116
193001 0.100000001490116
194001 0.100000001490116
195001 0.100000001490116
196001 0.100000001490116
197001 0.100000001490116
198001 0.100000001490116
199001 0.100000001490116
200001 0.00999999977648258
201001 0.00999999977648258
202001 0.00999999977648258
203001 0.00999999977648258
204001 0.00999999977648258
205001 0.00999999977648258
206001 0.00999999977648258
207001 0.00999999977648258
208001 0.00999999977648258
209001 0.00999999977648258
210001 0.00999999977648258
211001 0.00999999977648258
212001 0.00999999977648258
213001 0.00999999977648258
214001 0.00999999977648258
215001 0.00999999977648258
216001 0.00999999977648258
217001 0.00999999977648258
218001 0.00999999977648258
219001 0.00999999977648258
220001 0.00999999977648258
221001 0.00999999977648258
222001 0.00999999977648258
223001 0.00999999977648258
224001 0.00999999977648258
225001 0.00999999977648258
226001 0.00999999977648258
227001 0.00999999977648258
228001 0.00999999977648258
229001 0.00999999977648258
230001 0.00999999977648258
231001 0.00999999977648258
232001 0.00999999977648258
233001 0.00999999977648258
234001 0.00999999977648258
235001 0.00999999977648258
236001 0.00999999977648258
237001 0.00999999977648258
238001 0.00999999977648258
239001 0.00999999977648258
240001 0.00999999977648258
241001 0.00999999977648258
242001 0.00999999977648258
243001 0.00999999977648258
244001 0.00999999977648258
245001 0.00999999977648258
246001 0.00999999977648258
247001 0.00999999977648258
248001 0.00999999977648258
249001 0.00999999977648258
250001 0.00999999977648258
251001 0.00999999977648258
252001 0.00999999977648258
253001 0.00999999977648258
254001 0.00999999977648258
255001 0.00999999977648258
256001 0.00999999977648258
257001 0.00999999977648258
258001 0.00999999977648258
259001 0.00999999977648258
260001 0.00999999977648258
261001 0.00999999977648258
262001 0.00999999977648258
263001 0.00999999977648258
264001 0.00999999977648258
265001 0.00999999977648258
266001 0.00999999977648258
267001 0.00999999977648258
268001 0.00999999977648258
269001 0.00999999977648258
270001 0.00999999977648258
271001 0.00999999977648258
272001 0.00999999977648258
273001 0.00999999977648258
274001 0.00999999977648258
275001 0.00999999977648258
276001 0.00999999977648258
277001 0.00999999977648258
278001 0.00999999977648258
279001 0.00999999977648258
280001 0.00999999977648258
281001 0.00999999977648258
282001 0.00999999977648258
283001 0.00999999977648258
284001 0.00999999977648258
285001 0.00999999977648258
286001 0.00999999977648258
287001 0.00999999977648258
288001 0.00999999977648258
289001 0.00999999977648258
290001 0.00999999977648258
291001 0.00999999977648258
292001 0.00999999977648258
293001 0.00999999977648258
294001 0.00999999977648258
295001 0.00999999977648258
296001 0.00999999977648258
297001 0.00999999977648258
298001 0.00999999977648258
299001 0.00999999977648258
300001 0.00100000004749745
301001 0.00100000004749745
302001 0.00100000004749745
303001 0.00100000004749745
304001 0.00100000004749745
305001 0.00100000004749745
306001 0.00100000004749745
307001 0.00100000004749745
308001 0.00100000004749745
309001 0.00100000004749745
310001 0.00100000004749745
311001 0.00100000004749745
312001 0.00100000004749745
313001 0.00100000004749745
314001 0.00100000004749745
315001 0.00100000004749745
316001 0.00100000004749745
317001 0.00100000004749745
318001 0.00100000004749745
319001 0.00100000004749745
320001 0.00100000004749745
321001 0.00100000004749745
322001 0.00100000004749745
323001 0.00100000004749745
324001 0.00100000004749745
325001 0.00100000004749745
326001 0.00100000004749745
327001 0.00100000004749745
328001 0.00100000004749745
329001 0.00100000004749745
330001 0.00100000004749745
331001 0.00100000004749745
332001 0.00100000004749745
333001 0.00100000004749745
334001 0.00100000004749745
335001 0.00100000004749745
336001 0.00100000004749745
337001 0.00100000004749745
338001 0.00100000004749745
339001 0.00100000004749745
340001 0.00100000004749745
341001 0.00100000004749745
342001 0.00100000004749745
343001 0.00100000004749745
344001 0.00100000004749745
345001 0.00100000004749745
346001 0.00100000004749745
347001 0.00100000004749745
348001 0.00100000004749745
349001 0.00100000004749745
350001 0.00100000004749745
351001 0.00100000004749745
352001 0.00100000004749745
353001 0.00100000004749745
354001 0.00100000004749745
355001 0.00100000004749745
356001 0.00100000004749745
357001 0.00100000004749745
358001 0.00100000004749745
359001 0.00100000004749745
360001 0.00100000004749745
361001 0.00100000004749745
362001 0.00100000004749745
363001 0.00100000004749745
364001 0.00100000004749745
365001 0.00100000004749745
366001 0.00100000004749745
367001 0.00100000004749745
368001 0.00100000004749745
369001 0.00100000004749745
370001 0.00100000004749745
371001 0.00100000004749745
372001 0.00100000004749745
373001 0.00100000004749745
374001 0.00100000004749745
375001 0.00100000004749745
376001 0.00100000004749745
377001 0.00100000004749745
378001 0.00100000004749745
379001 0.00100000004749745
380001 0.00100000004749745
381001 0.00100000004749745
382001 0.00100000004749745
383001 0.00100000004749745
384001 0.00100000004749745
385001 0.00100000004749745
386001 0.00100000004749745
387001 0.00100000004749745
388001 0.00100000004749745
389001 0.00100000004749745
390001 0.00100000004749745
391001 0.00100000004749745
392001 0.00100000004749745
393001 0.00100000004749745
394001 0.00100000004749745
395001 0.00100000004749745
396001 0.00100000004749745
397001 0.00100000004749745
398001 0.00100000004749745
399001 0.00100000004749745
};
\addlegendentry{SGD : $\lambda$: 0.1, $\beta$: 0.9}
\addplot [line width=2.5pt, black, opacity=1.0]
table {%
1 0.100049309432507
2003 0.110145437220732
4003 0.180330981810888
6006 0.167528743545214
8007 0.250847501059373
10012 0.254680325587591
12015 0.273427426815033
14021 0.256926750143369
16021 0.356071357925733
18038 0.198253805438677
20038 0.284858867526054
22038 0.336605399847031
24050 0.21327926715215
26050 0.265361602107684
28050 0.394464448094368
30055 0.355927427609762
32064 0.261550868550936
34091 0.197479120145241
36091 0.158324678738912
38091 0.160196110606194
40091 0.176820253332456
42091 0.201589127381643
44091 0.317005544900894
46127 0.11392661742866
48127 0.139191267391046
50127 0.109238486737013
52127 0.158596908052762
54127 0.130658303697904
56127 0.158385172486305
58129 0.182393381992976
60133 0.181026871005694
62140 0.148198634386063
64159 0.117754459381104
66161 0.101097490638494
68161 0.0860518415768941
70161 0.0996937900781631
72172 0.0824359009663264
74186 0.0737614097694556
76198 0.0645380175362031
78198 0.05097881394128
80198 0.0646308821936448
82198 0.0674620792269707
84198 0.0733252912759781
86209 0.0549356391032537
88209 0.0634409586588542
90214 0.0627023453513781
92216 0.0698856338858604
94222 0.0668484941124916
96228 0.0621479203303655
98233 0.0639765647550424
100237 0.0641244339446227
102241 0.0616404811541239
104246 0.0627236093084017
106260 0.0352582323054473
108260 0.0463958693047365
110263 0.0389114040881395
112271 0.0378595888614655
114271 0.0473836548626423
116282 0.0421886617938677
118282 0.0448288495341937
120289 0.042085754374663
122297 0.0369032050172488
124297 0.0441900255779425
126300 0.0466440891226133
128305 0.0438725203275681
130308 0.0459223141272863
132316 0.0414804257452488
134320 0.0422283050914605
136326 0.042586375027895
138328 0.0449770124008258
140336 0.0374638866633177
142340 0.0382084331164757
144344 0.0383447663237651
146350 0.0370828000207742
148354 0.0370318864782651
150360 0.0333752694229285
152365 0.0362283481905858
154367 0.0398564375936985
156375 0.0327004262556632
158379 0.0301728459695975
160386 0.0325002893805504
162387 0.0358621372530858
164392 0.0327461299796899
166398 0.0353002833823363
168404 0.0307140511771043
170410 0.0306078058977922
172412 0.0321514718234539
174417 0.0312831544627746
176422 0.033122081309557
178426 0.0333742555230856
180431 0.030633774275581
182437 0.0290973894298077
184442 0.0292305499315262
186445 0.0307543929666281
188451 0.029016533245643
190456 0.0280518761525551
192459 0.0308754853904247
194466 0.0253813005983829
196470 0.0294611901044846
198481 0.0221205862859885
200481 0.0261239688843489
202483 0.0314146845291058
204491 0.0285294217367967
206507 0.0110754923714395
208507 0.00433113520952411
210507 0.007090969026254
212507 0.0116093903916359
214522 0.00391336998175485
216522 0.0064070001346392
218522 0.0104895917244604
220526 0.011267626060975
222532 0.00980372976892368
224536 0.0105308927795597
226541 0.0101807924805444
228548 0.00797228828378061
230549 0.0117470647302401
232556 0.00919879105906478
234560 0.00988108480174634
236566 0.00859732865481706
238568 0.0114012472353714
240576 0.00803519226973957
242582 0.00699125541146012
244585 0.00834423552599917
246589 0.00896314532617609
248595 0.00779864814614499
250599 0.00837709012469606
252601 0.0111091807736815
254606 0.0107398554945087
256613 0.00841007444813638
258615 0.0111529231243025
260621 0.00970392862360881
262626 0.00938132034671927
264630 0.010077153045732
266636 0.00876792337890726
268647 0.00450472411472113
270647 0.00737516956543005
272648 0.0108672181945381
274654 0.00945534249399983
276660 0.00822689883160382
278664 0.00883710510223405
280668 0.00949257241362151
282672 0.0101966566195998
284679 0.00798471087368497
286682 0.00952994825139033
288688 0.00829181201674068
290876 5.71288814865678e-11
292890 2.84565744810653e-11
294890 4.65893290381899e-11
296906 3.82824535960502e-11
298906 4.72546296074553e-11
300906 4.60018793890793e-11
302913 3.85539985353622e-11
304923 3.18740664649135e-11
306923 3.3667413475158e-11
308940 1.82937666751148e-11
310940 2.99506993280592e-11
312940 3.55785164061576e-11
314940 4.5311461180364e-11
316962 4.43932279536411e-11
318964 1.48772990454793e-11
320964 2.43572315095086e-11
322964 2.73843719691043e-11
324964 4.06442587919109e-11
326969 3.88835323892704e-11
328973 3.83783074764388e-11
330978 4.50104409762305e-11
332978 4.21873097798212e-11
334985 3.51351413241385e-11
336994 2.59833685239038e-11
339006 2.91951989204546e-11
341011 3.18756236683537e-11
343011 2.9273632996818e-11
345011 4.50813200427354e-11
347019 2.55643226495147e-11
349020 3.50292671027116e-11
351020 4.31745079502536e-11
353021 4.70163462560717e-11
355033 3.3455928999393e-11
357039 3.47271014028427e-11
359044 4.76580037008419e-11
361050 4.04277173080928e-11
363052 5.65202421901982e-11
365053 3.9689484337968e-11
367053 4.86134882349276e-11
369058 5.05893417435122e-11
371074 3.52225326451701e-11
373074 3.62115719354517e-11
375074 5.35523708711239e-11
377095 2.98924108845429e-11
379095 3.73415939373493e-11
381095 3.38701462709867e-11
383096 3.919457177796e-11
385100 5.10003243340644e-11
387106 5.04541651513701e-11
389123 2.71511800311227e-11
391123 2.61193549427885e-11
393123 3.28869674548482e-11
395123 4.45776193735981e-11
397128 5.08713430176962e-11
399141 3.06426591277035e-11
};
\addlegendentry{SLS : c: 0.1, $\beta$: 0.9, $\mu$: 0.1}
\addplot [line width=2.5pt, blue, opacity=1.0]
table {%
20 0.0638142228126526
1020 0.052600059658289
2020 0.052600059658289
3020 0.052600059658289
4020 0.052600059658289
5020 0.103502348065376
6020 0.103502348065376
7020 0.103502348065376
8020 0.103502348065376
9020 0.150411993265152
10020 0.150411993265152
11020 0.150411993265152
12020 0.150411993265152
13020 0.198586046695709
14020 0.198586046695709
15020 0.198586046695709
16020 0.198586046695709
17020 0.220505401492119
18020 0.220505401492119
19020 0.220505401492119
20020 0.220505401492119
21020 0.251788139343262
22020 0.251788139343262
23020 0.251788139343262
24020 0.251788139343262
25020 0.285890191793442
26020 0.285890191793442
27020 0.285890191793442
28020 0.285890191793442
29020 0.315260767936707
30020 0.315260767936707
31020 0.315260767936707
32020 0.315260767936707
33020 0.359337031841278
34020 0.359337031841278
35020 0.359337031841278
36020 0.359337031841278
37020 0.364026635885239
38020 0.364026635885239
39020 0.364026635885239
40020 0.364026635885239
41020 0.378974288702011
42020 0.378974288702011
43020 0.378974288702011
44020 0.378974288702011
45020 0.386409074068069
46020 0.386409074068069
47020 0.386409074068069
48020 0.386409074068069
49020 0.378196477890015
50020 0.378196477890015
51020 0.378196477890015
52020 0.378196477890015
53020 0.378116577863693
54020 0.378116577863693
55020 0.378116577863693
56020 0.378116577863693
57020 0.38107642531395
58020 0.38107642531395
59020 0.38107642531395
60020 0.38107642531395
61020 0.413988530635834
62020 0.413988530635834
63020 0.413988530635834
64020 0.413988530635834
65020 0.426286607980728
66020 0.426286607980728
67020 0.426286607980728
68020 0.426286607980728
69020 0.440522402524948
70020 0.440522402524948
71020 0.440522402524948
72020 0.440522402524948
73020 0.439496099948883
74020 0.439496099948883
75020 0.439496099948883
76020 0.439496099948883
77020 0.440635532140732
78020 0.440635532140732
79020 0.440635532140732
80020 0.440635532140732
81020 0.445022970438004
82020 0.445022970438004
83020 0.445022970438004
84020 0.445022970438004
85020 0.433503717184067
86020 0.433503717184067
87020 0.433503717184067
88020 0.433503717184067
89020 0.413608700037003
90020 0.413608700037003
91020 0.413608700037003
92020 0.413608700037003
93020 0.387716591358185
94020 0.387716591358185
95020 0.387716591358185
96020 0.387716591358185
97020 0.373207420110703
98020 0.373207420110703
99020 0.373207420110703
100020 0.373207420110703
101020 0.358298808336258
102020 0.358298808336258
103020 0.358298808336258
104020 0.358298808336258
105020 0.335995823144913
106020 0.335995823144913
107020 0.335995823144913
108020 0.335995823144913
109020 0.340781182050705
110020 0.340781182050705
111020 0.340781182050705
112020 0.340781182050705
113020 0.329207897186279
114020 0.329207897186279
115020 0.329207897186279
116020 0.329207897186279
117020 0.309711575508118
118020 0.309711575508118
119020 0.309711575508118
120020 0.309711575508118
121020 0.284049391746521
122020 0.284049391746521
123020 0.284049391746521
124020 0.284049391746521
125020 0.286866247653961
126020 0.286866247653961
127020 0.286866247653961
128020 0.286866247653961
129020 0.269594728946686
130020 0.269594728946686
131020 0.269594728946686
132020 0.269594728946686
133020 0.258042484521866
134020 0.258042484521866
135020 0.258042484521866
136020 0.258042484521866
137020 0.311275273561478
138020 0.311275273561478
139020 0.311275273561478
140020 0.311275273561478
141020 0.303622901439667
142020 0.303622901439667
143020 0.303622901439667
144020 0.303622901439667
145020 0.298938423395157
146020 0.298938423395157
147020 0.298938423395157
148020 0.298938423395157
149020 0.307735949754715
150020 0.307735949754715
151020 0.307735949754715
152020 0.307735949754715
153020 0.294962525367737
154020 0.294962525367737
155020 0.294962525367737
156020 0.294962525367737
157020 0.329672127962112
158020 0.329672127962112
159020 0.329672127962112
160020 0.329672127962112
161020 0.319820284843445
162020 0.319820284843445
163020 0.319820284843445
164020 0.319820284843445
165020 0.32157438993454
166020 0.32157438993454
167020 0.32157438993454
168020 0.32157438993454
169020 0.330822736024857
170020 0.330822736024857
171020 0.330822736024857
172020 0.330822736024857
173020 0.426197111606598
174020 0.426197111606598
175020 0.426197111606598
176020 0.426197111606598
177020 0.425680637359619
178020 0.425680637359619
179020 0.425680637359619
180020 0.425680637359619
181020 0.456744968891144
182020 0.456744968891144
183020 0.456744968891144
184020 0.456744968891144
185020 0.45198717713356
186020 0.45198717713356
187020 0.45198717713356
188020 0.45198717713356
189020 0.457675993442535
190020 0.457675993442535
191020 0.457675993442535
192020 0.457675993442535
193020 0.467437714338303
194020 0.467437714338303
195020 0.467437714338303
196020 0.467437714338303
197020 0.412060350179672
198020 0.412060350179672
199020 0.412060350179672
200020 0.412060350179672
201020 0.410495311021805
202020 0.410495311021805
203020 0.410495311021805
204020 0.410495311021805
205020 0.410495311021805
206020 0.410495311021805
207020 0.410495311021805
208020 0.410495311021805
209020 0.408468753099442
210020 0.408468753099442
211020 0.408468753099442
212020 0.408468753099442
213020 0.408468753099442
214020 0.408468753099442
215020 0.408468753099442
216020 0.408468753099442
217020 0.405989646911621
218020 0.405989646911621
219020 0.405989646911621
220020 0.405989646911621
221020 0.405989646911621
222020 0.405989646911621
223020 0.405989646911621
224020 0.405989646911621
225020 0.452391743659973
226020 0.452391743659973
227020 0.452391743659973
228020 0.452391743659973
229020 0.452391743659973
230020 0.452391743659973
231020 0.452391743659973
232020 0.452391743659973
233020 0.451100260019302
234020 0.451100260019302
235020 0.451100260019302
236020 0.451100260019302
237020 0.451100260019302
238020 0.451100260019302
239020 0.451100260019302
240020 0.451100260019302
241020 0.473604559898376
242020 0.473604559898376
243020 0.473604559898376
244020 0.473604559898376
245020 0.473604559898376
246020 0.473604559898376
247020 0.473604559898376
248020 0.473604559898376
249020 0.654541373252869
250020 0.654541373252869
251020 0.654541373252869
252020 0.654541373252869
253020 0.654541373252869
254020 0.654541373252869
255020 0.654541373252869
256020 0.654541373252869
257020 0.63267594575882
258020 0.63267594575882
259020 0.63267594575882
260020 0.63267594575882
261020 0.63267594575882
262020 0.63267594575882
263020 0.63267594575882
264020 0.63267594575882
265020 0.582283675670624
266020 0.582283675670624
267020 0.582283675670624
268020 0.582283675670624
269020 0.582283675670624
270020 0.582283675670624
271020 0.582283675670624
272020 0.582283675670624
273020 0.729400277137756
274020 0.729400277137756
275020 0.729400277137756
276020 0.729400277137756
277020 0.729400277137756
278020 0.729400277137756
279020 0.729400277137756
280020 0.729400277137756
281020 0.702348172664642
282020 0.702348172664642
283020 0.702348172664642
284020 0.702348172664642
285020 0.702348172664642
286020 0.702348172664642
287020 0.702348172664642
288020 0.702348172664642
289020 0.691676735877991
290020 0.691676735877991
291020 0.691676735877991
292020 0.691676735877991
293020 0.691676735877991
294020 0.691676735877991
295020 0.691676735877991
296020 0.691676735877991
297020 0.68524181842804
298020 0.68524181842804
299020 0.68524181842804
300020 0.68524181842804
301020 0.695001006126404
302020 0.695001006126404
303020 0.695001006126404
304020 0.695001006126404
305020 0.695001006126404
306020 0.695001006126404
307020 0.695001006126404
308020 0.695001006126404
309020 0.695001006126404
310020 0.695001006126404
311020 0.695001006126404
312020 0.695001006126404
313020 0.695001006126404
314020 0.695001006126404
315020 0.695001006126404
316020 0.695001006126404
317020 0.757639408111572
318020 0.757639408111572
319020 0.757639408111572
320020 0.757639408111572
321020 0.757639408111572
322020 0.757639408111572
323020 0.757639408111572
324020 0.757639408111572
325020 0.757639408111572
326020 0.757639408111572
327020 0.757639408111572
328020 0.757639408111572
329020 0.757639408111572
330020 0.757639408111572
331020 0.757639408111572
332020 0.757639408111572
333020 1.00965082645416
334020 1.00965082645416
335020 1.00965082645416
336020 1.00965082645416
337020 1.00965082645416
338020 1.00965082645416
339020 1.00965082645416
340020 1.00965082645416
341020 1.00965082645416
342020 1.00965082645416
343020 1.00965082645416
344020 1.00965082645416
345020 1.00965082645416
346020 1.00965082645416
347020 1.00965082645416
348020 1.00965082645416
349020 1.06449055671692
350020 1.06449055671692
351020 1.06449055671692
352020 1.06449055671692
353020 1.06449055671692
354020 1.06449055671692
355020 1.06449055671692
356020 1.06449055671692
357020 1.06449055671692
358020 1.06449055671692
359020 1.06449055671692
360020 1.06449055671692
361020 1.06449055671692
362020 1.06449055671692
363020 1.06449055671692
364020 1.06449055671692
365020 1.15304410457611
366020 1.15304410457611
367020 1.15304410457611
368020 1.15304410457611
369020 1.15304410457611
370020 1.15304410457611
371020 1.15304410457611
372020 1.15304410457611
373020 1.15304410457611
374020 1.15304410457611
375020 1.15304410457611
376020 1.15304410457611
377020 1.15304410457611
378020 1.15304410457611
379020 1.15304410457611
380020 1.15304410457611
381020 1.17879939079285
382020 1.17879939079285
383020 1.17879939079285
384020 1.17879939079285
385020 1.17879939079285
386020 1.17879939079285
387020 1.17879939079285
388020 1.17879939079285
389020 1.17879939079285
390020 1.17879939079285
391020 1.17879939079285
392020 1.17879939079285
393020 1.17879939079285
394020 1.17879939079285
395020 1.17879939079285
396020 1.17879939079285
397020 1.36329913139343
398020 1.36329913139343
399020 1.36329913139343
};
\addlegendentry{LABPAL-SGD : $\epsilon: 128/bs$ (others same as above)}
\addplot [line width=2.5pt, red, opacity=1.0]
table {%
20 0.0544980149716139
1020 0.0506910849362612
2020 0.0543158892542124
3020 0.060392377898097
4020 0.0566706843674183
5020 0.117775462567806
6020 0.12196858599782
7020 0.128394942730665
8020 0.141924113035202
9020 0.175884395837784
10020 0.206749141216278
11020 0.225401423871517
12020 0.227550469338894
13020 0.280481263995171
14020 0.341429129242897
15020 0.298750698566437
16020 0.318221688270569
17020 0.364757254719734
18020 0.450377255678177
19020 0.524550274014473
20020 0.525976195931435
21020 0.481768816709518
22020 0.638998627662659
23020 0.499094501137733
24020 0.534931004047394
25020 0.709545522928238
26020 0.600711017847061
27020 0.722618728876114
28020 0.725976169109344
29020 0.597132921218872
30020 0.720901221036911
31020 0.664112687110901
32020 0.753188461065292
33020 0.738119006156921
34020 0.861665338277817
35020 0.937308520078659
36020 0.588200986385345
37020 0.893023192882538
38020 0.928048312664032
39020 0.834814816713333
40020 0.808730095624924
41020 1.16887751221657
42020 0.999584496021271
43020 1.01347017288208
44020 1.68295294046402
45020 0.937935888767242
46020 1.07334733009338
47020 1.10943418741226
48020 1.16749531030655
49020 0.856108099222183
50020 0.969639807939529
51020 1.06498426198959
52020 1.12018746137619
53020 0.892938107252121
54020 1.19116121530533
55020 1.03246659040451
56020 1.4866731762886
57020 0.948221564292908
58020 1.11470663547516
59020 0.945331990718842
60020 1.16339391469955
61020 1.41830319166183
62020 0.837665140628815
63020 0.923907101154327
64020 1.57855892181396
65020 1.13307163119316
66020 1.24502152204514
67020 1.49590241909027
68020 1.16353815793991
69020 1.17714151740074
70020 1.32987651228905
71020 1.42214274406433
72020 1.41488260030746
73020 0.970477759838104
74020 1.25347471237183
75020 1.2658851146698
76020 1.40183877944946
77020 2.863325715065
78020 0.939962863922119
79020 1.13991656899452
80020 1.31267940998077
81020 1.37495082616806
82020 1.07703340053558
83020 2.56691586971283
84020 1.21149170398712
85020 1.25721019506454
86020 1.44348973035812
87020 1.33590525388718
88020 2.01336467266083
89020 1.25598028302193
90020 1.63934510946274
91020 1.68771064281464
92020 1.70045709609985
93020 1.31559348106384
94020 1.231818318367
95020 1.2151605784893
96020 1.05396384000778
97020 2.9871738255024
98020 1.52562892436981
99020 1.46457633376122
100020 0.95894181728363
101020 1.6305850148201
102020 2.08083897829056
103020 1.25190359354019
104020 1.054782807827
105020 2.49318027496338
106020 1.87549182772636
107020 1.48218256235123
108020 7.68150451779366
109020 1.20406419038773
110020 2.40544003248215
111020 2.68577206134796
112020 1.07903897762299
113020 1.2845256626606
114020 0.94949147105217
115020 1.08556494116783
116020 1.06917163729668
117020 2.68213596940041
118020 2.40727126598358
119020 1.97788065671921
120020 4.19820049405098
121020 22.5955818295479
122020 0.671730041503906
123020 1.64331549406052
124020 4.00281429290771
125020 10.4167091846466
126020 3.87630766630173
127020 2.50006866455078
128020 2.30857864022255
129020 1.22363582253456
130020 0.501404777169228
131020 0.45368330180645
132020 0.652102589607239
133020 0.835157364606857
134020 3.26677703857422
135020 1.53494438529015
136020 1.07089371979237
137020 3.3534597158432
138020 3.30405205488205
139020 1.58276188373566
140020 1.36136132478714
141020 8.81350481510162
142020 1.57139927148819
143020 25.2165019512177
144020 2.20276245474815
145020 26.8779344558716
146020 0.995668217539787
147020 5.93280649185181
148020 0.768967136740685
149020 122.374958232045
150020 5.11698812246323
151020 45.8248252868652
152020 0.237415201961994
153020 0.501953125
154020 5.08758282661438
155020 1.71095716953278
156020 5.27484047412872
157020 3.05319835245609
158020 31.2534618377686
159020 13.7601322233677
160020 146.383432388306
161020 8.45943367481232
162020 0.146177414804697
163020 22.8955995887518
164020 59.8301870524883
165020 85638.759765625
166020 92.6777081489563
167020 4.27313431352377
168020 84492.8715209961
169020 0.165330708026886
170020 183.288891792297
171020 2.41027367115021
172020 10.841026801616
173020 7.76973259449005
174020 92301.0421761125
175020 6691.83382415771
176020 40.920910179615
177020 2.0829558596015
178020 255.955903701484
179020 3516.77516956627
180020 89998.0538311005
181020 17.6787750720978
182020 229.824555397034
183020 392.555220961571
184020 49.5032272338867
185020 8391.73775315285
186020 14.5451887845993
187020 1036.49456787109
188020 80177.4765625
189020 703.573748245835
190020 63529.1030055732
191020 65581.6623535156
192020 614.914772033691
193020 1399.4920501709
194020 16242.6298828125
195020 29158.5070953369
196020 38.9231830835342
197020 265.319005966187
198020 1882.75942102075
199020 3785.20044648647
200020 27879.7640380859
201020 1621.02860867977
202020 2341.15128707886
203020 6261.13108238578
204020 48576.9015960693
205020 48603.8238372803
206020 33.5829188674688
207020 48408.7444680929
208020 9567.0613078028
209020 355.377181321383
210020 37166.8542685509
211020 28.3438355177641
212020 37165.7390368283
213020 37169.0111227036
214020 7464.05934429169
215020 0.237154763191938
216020 4.05981934070587
217020 36436.1463827044
218020 3192.00507031381
219020 36436.5680989027
220020 36436.1065635532
221020 24.9681911468506
222020 1073.67025184631
223020 11802.5197801888
224020 36455.4887638092
225020 122.221031188965
226020 523.392916604877
227020 4376.80901670456
228020 32963.541451931
229020 32984.5287132263
230020 32956.1384314001
231020 32957.8531552553
232020 32957.7682137489
233020 34085.8014936447
234020 34076.651446104
235020 23570.1565246582
236020 864.715620279312
237020 5870.40997666121
238020 15.2466688379645
239020 34076.5957214534
240020 131.507175624371
241020 209.024852409959
242020 131721.604901748
243020 34195.2673386335
244020 599.72900390625
245020 34194.6300016791
246020 34246.4213905334
247020 74.461600586772
248020 3324.59766483307
249020 16484.5104217529
250020 16297.9508428574
251020 0.157956690527499
252020 16444.70262146
253020 134.038597106934
254020 91.7711322158575
255020 1882.95368288457
256020 16298.093179971
257020 16244.0653043985
258020 0.249437997117639
259020 16682.7010461092
260020 27.9623991549015
261020 16682.8236039281
262020 16683.0254164636
263020 2933.39756831527
264020 9633.04264748096
265020 790.135734558105
266020 2354.56158447266
267020 15284.9148330688
268020 8318.09414672852
269020 17.1305894851685
270020 15216.5984296799
271020 15451.6536102295
272020 15354.5191955566
273020 1.68590413033962
274020 1016.64998580515
275020 1662.92541122437
276020 15559.7034301758
277020 14171.4333820343
278020 0.591159909963608
279020 15260.8715633601
280020 61.2904844284058
281020 15260.9549271911
282020 1342.52739620209
283020 15692.9615859985
284020 15640.2869358063
285020 1543.23529268801
286020 15669.847328186
287020 15634.4199801534
288020 54694.4929199219
289020 458.104309082031
290020 18.5846214294434
291020 197.128097057343
292020 196.316458702087
293020 65947.501953125
294020 12082.6164250374
295020 252.521911621094
296020 207.499101042747
297020 17194.7075195312
298020 11935.8501815796
299020 16180.2357940674
300020 13287.1555175781
301020 16161.7452793121
302020 15678.2680969238
303020 16142.1730666161
304020 95.1618435829878
305020 16469.877532959
306020 386.586845606565
307020 16150.4439592361
308020 12699.4364700317
309020 1420.87037789822
310020 12058.3662675768
311020 16140.0150086582
312020 1051.17601329088
313020 16158.9737377167
314020 2564.67634417117
315020 499.655913114548
316020 4246.38896942139
317020 16039.9250928313
318020 16039.8933590651
319020 499.962692260742
320020 4.50503444671631
321020 41.2024140357971
322020 15147.5302848816
323020 23062.2880859375
324020 21652.228515625
325020 134.63497005403
326020 484.406982421875
327020 2949.94036364555
328020 16050.3511314392
329020 107745.069602966
330020 761.070602416992
331020 4530.02368164062
332020 23532.515625
333020 103069.319000244
334020 16320.3801879883
335020 19001.0004882812
336020 15639.1117610931
337020 8826.82690429688
338020 54882.609375
339020 9277.93635329604
340020 5741.54125976562
341020 37637.0703125
342020 15631.1140717715
343020 15646.4455718994
344020 465.554825782776
345020 4575.98460388184
346020 81481.3920449317
347020 23.5597231835127
348020 893.373291492462
349020 22093.3109690547
350020 16667.2784423828
351020 88637.9234619141
352020 102948.579101562
353020 15661.5957641602
354020 10186.7916717529
355020 15428.0619183481
356020 63729.8869628906
357020 492.995185852051
358020 90633.3916015625
359020 73862.0556640625
360020 15430.3820867538
361020 15428.0708711147
362020 102616.948242188
363020 24091.8505859375
364020 240.936249747872
365020 15096.2548571602
366020 16694.9354248047
367020 5662.31480407715
368020 74451.6171875
369020 7708.02444458008
370020 74557.1083984375
371020 37173.2138671875
372020 74557.1083984375
373020 15125.1816577911
374020 2311.50085449219
375020 21992.07421875
376020 74557.1083984375
377020 74557.1083984375
378020 62541.8312988281
379020 60474.0427856445
380020 150.082107543945
381020 59251.8058063425
382020 9986.19624638557
383020 74347.8720703125
384020 15367.802734375
385020 60062.0706787109
386020 59285.1295318604
387020 33054.0908203125
388020 74347.8720703125
389020 3016.89144897461
390020 341.990898132324
391020 74347.8720703125
392020 15483.0335388184
393020 74347.8720703125
394020 88.9294852614403
395020 16508.2813720703
396020 85.6393814086914
397020 15616.3755745888
398020 53591.1015625
399020 62261.466796875
};
\addlegendentry{LABPAL-NSGD : $\epsilon: 128/bs$ (others same as above)}
\end{axis}

\end{tikzpicture}

%% file: labpal/figure_data/performance_comparison/CIFAR-10/CIFAR-10_DenseNet-121_validation_accuracy2.pgf
% This file was created by tikzplotlib v0.9.8.
\begin{tikzpicture}

\definecolor{color0}{rgb}{0.647058823529412,0.164705882352941,0.164705882352941}
\definecolor{color1}{rgb}{0.933333333333333,0.509803921568627,0.933333333333333}
\definecolor{color2}{rgb}{1,0.647058823529412,0}

\begin{axis}[
grid = major,
major grid style={dotted},
minor xtick={},
%minor ytick={},
tick align=outside,
tick pos=left,
width=10.5cm,height=8cm,,
x grid style={white!69.0196078431373!black},
xlabel={epochs},
xmin=-13.15, xmax=298.15,
xtick style={color=black},
xtick={-50,0,50,100,150,200,250,300},
y grid style={white!69.0196078431373!black},
ylabel={validation accuracy},
ymin=0.85, ymax=0.94,
ytick style={color=black},
ytick={0.85,0.86,0.87,0.88,0.89,0.9,0.91,0.92,0.93},
minor y tick num=1
]
\addplot [line width=2.5pt, color0, opacity=1.0]
table {%
1 0.460536847511927
2 0.574519236882528
3 0.641960461934408
4 0.695045411586761
5 0.707398494084676
6 0.724692841370901
7 0.760817309220632
8 0.772035260995229
9 0.771100401878357
10 0.789997339248657
11 0.787459929784139
12 0.801949799060822
13 0.80795939763387
14 0.817307690779368
15 0.82545405626297
16 0.824519236882528
17 0.83019498984019
18 0.827056606610616
19 0.825787921746572
20 0.835803965727488
21 0.823651174704234
22 0.836538473765055
23 0.842280964056651
24 0.848357359568278
25 0.842013895511627
26 0.833533644676208
27 0.845419347286224
28 0.852764427661896
29 0.849492530028025
30 0.852430562178294
31 0.840544859568278
32 0.849759618441264
33 0.854634086290995
34 0.859041134516398
35 0.855235040187836
36 0.855034728844961
37 0.857905964056651
38 0.851896365483602
39 0.855635662873586
40 0.845819969971975
41 0.854433755079905
42 0.83653845389684
43 0.854033132394155
44 0.857238252957662
45 0.854700863361359
46 0.853165050347646
47 0.852564116319021
48 0.854500532150269
49 0.856370190779368
50 0.847489317258199
51 0.841012279192607
52 0.855902771155039
53 0.858106315135956
54 0.858373383680979
55 0.858907580375671
56 0.851228634516398
57 0.847355763117472
58 0.851896365483602
59 0.853699247042338
60 0.836071054140727
61 0.852831184864044
62 0.860977570215861
63 0.858573714892069
64 0.860042730967204
65 0.865251064300537
66 0.860243062178294
67 0.857238233089447
68 0.861912389596303
69 0.746928413709005
70 0.852897981802622
71 0.857839206854502
72 0.859041134516398
73 0.836271385351817
74 0.847689648469289
75 0.847622871398926
76 0.847355763117472
77 0.847756425539652
78 0.85423344373703
79 0.861912389596303
80 0.857572098573049
81 0.85423344373703
82 0.857638895511627
83 0.849158644676208
84 0.857171475887299
85 0.856637299060822
86 0.867454588413239
87 0.850694437821706
88 0.870259086290995
89 0.869057158629099
90 0.868122319380442
91 0.859842419624329
92 0.863247851530711
93 0.858640491962433
94 0.871794859568278
95 0.860510150591532
96 0.860710461934408
97 0.872863233089447
98 0.869190692901611
99 0.857505341370901
100 0.867053965727488
101 0.86204594373703
102 0.868589739004771
103 0.868456204732259
104 0.85423344373703
105 0.850560903549194
106 0.861177881558736
107 0.619991992910703
108 0.872262299060822
109 0.869324247042338
110 0.872195502122243
111 0.870726505915324
112 0.878672540187836
113 0.877804497877757
114 0.881677329540253
115 0.353766031563282
116 0.864516536394755
117 0.874666134516398
118 0.869658132394155
119 0.872128744920095
120 0.870058755079905
121 0.878672540187836
122 0.877804478009542
123 0.883947630723318
124 0.869858463605245
125 0.876736124356588
126 0.879473825295766
127 0.87439904610316
128 0.883947670459747
129 0.876201907793681
130 0.875934839248657
131 0.882411857446035
132 0.887620190779368
133 0.886418282985687
134 0.626201927661896
135 0.635950853427251
136 0.871127148469289
137 0.76315438747406
138 0.872662941614787
139 0.881343464056651
140 0.620726488530636
141 0.873397429784139
142 0.872996787230174
143 0.751736104488373
144 0.869391004244486
145 0.871394236882528
146 0.865384618441264
147 0.877804478009542
148 0.85403311252594
149 0.884348293145498
150 0.839610040187836
151 0.559294859568278
152 0.874732931454976
153 0.872262279192607
154 0.878739337126414
155 0.879073182741801
156 0.872796475887299
157 0.876602570215861
158 0.871928413709005
159 0.869724889596303
160 0.883346696694692
161 0.880008002122243
162 0.878004809220632
163 0.872662941614787
164 0.888287941614787
165 0.867053945859273
166 0.876669347286224
167 0.885616978009542
168 0.883947670459747
169 0.872863252957662
170 0.888888895511627
171 0.879874467849731
172 0.879607379436493
173 0.886418263117472
174 0.879941244920095
175 0.884281516075134
176 0.753138363361359
177 0.877403835455577
178 0.877537409464518
179 0.876469016075134
180 0.878739297389984
181 0.887419859568278
182 0.879807690779368
183 0.601161867380142
184 0.877604166666667
185 0.879273494084676
186 0.868456204732259
187 0.88054217894872
188 0.883146365483602
189 0.883012811342875
190 0.70506144563357
191 0.87954060236613
192 0.889489849408468
193 0.767160793145498
194 0.873597760995229
195 0.883146385351817
196 0.728632479906082
197 0.78331998984019
198 0.875
199 0.668402776122093
200 0.868790070215861
201 0.606503742436568
202 0.870125552018484
203 0.876068373521169
204 0.789797027905782
205 0.755542198816935
206 0.653712605436643
207 0.877203524112701
208 0.874532580375671
209 0.871127148469289
210 0.886017640431722
211 0.894431094328562
212 0.849091867605845
213 0.865584929784139
214 0.844951927661896
215 0.867254277070363
216 0.788528303305308
217 0.874532600243886
218 0.529513890544573
219 0.839409728844961
220 0.855635682741801
221 0.684428413709005
222 0.881209929784139
223 0.88735310236613
224 0.887820521990458
225 0.880809307098389
226 0.770365913709005
227 0.623864861826102
228 0.757545411586761
229 0.615117520093918
230 0.634682153662046
231 0.874131957689921
232 0.875934839248657
233 0.880876044432322
234 0.66559828321139
235 0.824986636638641
236 0.873731315135956
237 0.844217419624329
238 0.882478654384613
239 0.862847228844961
240 0.778445521990458
241 0.607572123408318
242 0.613782040774822
243 0.567240911225478
244 0.530849357446035
245 0.621861654023329
246 0.625801295042038
247 0.547475951413314
248 0.625066791971525
};
\addplot [line width=2.5pt, green!50.1960784313725!black, opacity=1.0]
table {%
1 0.512152771155039
2 0.6015625
3 0.656583885351817
4 0.717548072338104
5 0.731570502122243
6 0.767561415831248
7 0.783186435699463
8 0.800480763117472
9 0.811164538065592
10 0.825988252957662
11 0.826388875643412
12 0.831263343493144
13 0.844751616319021
14 0.841479698816935
15 0.854033132394155
16 0.855635682741801
17 0.852564096450806
18 0.856637279192607
19 0.862179478009542
20 0.861845632394155
21 0.866519769032796
22 0.862513363361359
23 0.868923624356588
24 0.872863252957662
25 0.868856827418009
26 0.868990381558736
27 0.876335481802622
28 0.873864869276682
29 0.873664538065592
30 0.876335461934408
31 0.870860040187836
32 0.874131937821706
33 0.87827189763387
34 0.880008002122243
35 0.875267088413239
36 0.877136747042338
37 0.88568377494812
38 0.883346696694692
39 0.882011214892069
40 0.882946034272512
41 0.88608439763387
42 0.891960481802622
43 0.886551817258199
44 0.88241183757782
45 0.886151194572449
46 0.884214739004771
47 0.889756937821706
48 0.891292730967204
49 0.883213142553965
50 0.887152771155039
51 0.88528311252594
52 0.889222741127014
53 0.89122595389684
54 0.891025642553965
55 0.885750532150269
56 0.884949266910553
57 0.893629809220632
58 0.892694969971975
59 0.889022429784139
60 0.892427881558736
61 0.89289528131485
62 0.893563032150269
63 0.895365913709005
64 0.897502680619558
65 0.892761747042338
66 0.895699799060822
67 0.892761747042338
68 0.89556622505188
69 0.893763363361359
70 0.893830120563507
71 0.900373935699463
72 0.892895301183065
73 0.890758554140727
74 0.894163986047109
75 0.893563032150269
76 0.895900090535482
77 0.895499467849731
78 0.895032068093618
79 0.896434287230174
80 0.895699799060822
81 0.89309561252594
82 0.895699779192607
83 0.895966867605845
84 0.894965271155039
85 0.897636214892069
86 0.899305562178294
87 0.894097228844961
88 0.899973293145498
89 0.900841335455577
90 0.898571034272512
91 0.898704608281453
92 0.897702991962433
93 0.899172008037567
94 0.901575863361359
95 0.896968503793081
96 0.898771365483602
97 0.898704588413239
98 0.900774578253428
99 0.903445521990458
100 0.898103634516398
101 0.89903845389684
102 0.900974889596303
103 0.90110844373703
104 0.897769769032796
105 0.896701375643412
106 0.901308755079905
107 0.904113233089447
108 0.900707801183065
109 0.8984375
110 0.902644236882528
111 0.902844568093618
112 0.902243594328562
113 0.902310351530711
114 0.901375532150269
115 0.903178413709005
116 0.903044879436493
117 0.901842931906382
118 0.8984375
119 0.903111636638641
120 0.899105250835419
121 0.902110040187836
122 0.899105230967204
123 0.902911325295766
124 0.901509086290995
125 0.90170939763387
126 0.902510682741801
127 0.904780983924866
128 0.903645833333333
129 0.900974889596303
130 0.903645833333333
131 0.902243594328562
132 0.905048072338104
133 0.898303945859273
134 0.902777771155039
135 0.904847741127014
136 0.905649026234945
137 0.903311967849731
138 0.904447098573049
139 0.906917730967204
140 0.90397971868515
141 0.902644236882528
142 0.906917730967204
143 0.908186435699463
144 0.910389959812164
145 0.912860592206319
146 0.912393152713776
147 0.912459949652354
148 0.910790582497915
149 0.910657048225403
150 0.914396365483602
151 0.914930562178294
152 0.914529939492544
153 0.911458333333333
154 0.913394769032796
155 0.913862188657125
156 0.912059287230174
157 0.915731847286224
158 0.914663473765055
159 0.91346154610316
160 0.916466355323792
161 0.913728634516398
162 0.912459929784139
163 0.914997339248657
164 0.914329588413239
165 0.915865381558736
166 0.915197630723318
167 0.914730230967204
168 0.914463142553965
169 0.91386216878891
170 0.913928945859273
171 0.915998935699463
172 0.915865381558736
173 0.918336013952891
174 0.916332801183065
175 0.914863785107931
176 0.915531496206919
177 0.915731827418009
178 0.915598293145498
179 0.914797008037567
180 0.919204076131185
181 0.914730230967204
182 0.916332801183065
183 0.91386216878891
184 0.916132469971975
185 0.915731827418009
186 0.915998935699463
187 0.91673344373703
188 0.915665050347646
189 0.917000532150269
190 0.918335994084676
191 0.913595060507456
192 0.916266024112701
193 0.917067289352417
194 0.913928945859273
195 0.917267620563507
196 0.915397981802622
197 0.917000532150269
198 0.91860310236613
199 0.918870210647583
200 0.915865381558736
201 0.91613248984019
202 0.914396365483602
203 0.916800220807393
204 0.917735060056051
205 0.915731827418009
206 0.917000512282054
207 0.91613248984019
208 0.915731847286224
209 0.916199266910553
210 0.917868594328562
211 0.916199247042338
212 0.916599889596303
213 0.915531535943349
214 0.917200863361359
215 0.916332801183065
216 0.918135682741801
217 0.919070521990458
218 0.918135702610016
219 0.917200863361359
220 0.916800200939178
221 0.917200863361359
222 0.916065692901611
223 0.916800220807393
224 0.914463142553965
225 0.915464739004771
226 0.91673344373703
227 0.918402791023254
228 0.915798604488373
229 0.917467951774597
230 0.916266024112701
231 0.917935371398926
232 0.917267620563507
233 0.917267620563507
234 0.916800220807393
235 0.914930562178294
236 0.913995722929637
237 0.917401174704234
238 0.915932158629099
239 0.916266024112701
240 0.917935371398926
241 0.917267640431722
242 0.914129257202148
243 0.917067309220632
244 0.91713406642278
245 0.918669859568278
246 0.916599909464518
247 0.918068905671438
248 0.917467951774597
249 0.914930562178294
250 0.917000532150269
251 0.917267620563507
252 0.917668263117472
253 0.915665070215861
254 0.915464739004771
255 0.913728614648183
256 0.915598293145498
257 0.918002148469289
258 0.914329588413239
259 0.916599909464518
260 0.916332801183065
261 0.916599889596303
262 0.914930562178294
263 0.915197650591532
264 0.913728614648183
265 0.917735040187836
266 0.914396345615387
267 0.914997319380442
268 0.914730230967204
269 0.915865381558736
270 0.917534728844961
271 0.917334417502085
272 0.915197650591532
273 0.917601486047109
274 0.917534728844961
275 0.916332801183065
276 0.91713406642278
277 0.914596696694692
278 0.918269236882528
279 0.913795411586761
280 0.916466355323792
281 0.917267640431722
282 0.912593483924866
283 0.914596696694692
284 0.916599889596303
};
\addplot [line width=2.5pt, black, opacity=1.0]
table {%
1 0.416199256976445
2 0.394564638535182
3 0.488848825295766
4 0.574719548225403
5 0.62252938747406
6 0.614449799060822
7 0.706797540187836
8 0.737313032150269
9 0.703525642553965
10 0.729700843493144
11 0.75761216878891
12 0.762486636638641
13 0.769698182741801
14 0.791332801183065
15 0.788661857446035
16 0.804420411586761
17 0.800747871398926
18 0.81270033121109
19 0.772168795267741
20 0.790331204732259
21 0.821113765239716
22 0.807959417502085
23 0.837673624356588
24 0.831196586290995
25 0.820646385351817
26 0.842214186986287
27 0.831396917502085
28 0.859842419624329
29 0.841746787230174
30 0.81804221868515
31 0.820779899756114
32 0.844618062178294
33 0.853031516075134
34 0.847355763117472
35 0.865518152713776
36 0.86678683757782
37 0.855769217014313
38 0.837339739004771
39 0.840077459812164
40 0.827791134516398
41 0.861444969971975
42 0.874666134516398
43 0.870592951774597
44 0.865785241127014
45 0.860710461934408
46 0.860376596450806
47 0.857572118441264
48 0.84829060236613
49 0.854233463605245
50 0.845018704732259
51 0.847622871398926
52 0.852497339248657
53 0.863247871398926
54 0.884615381558736
55 0.885216335455577
56 0.879807710647583
57 0.867321054140727
58 0.875467419624329
59 0.873130341370901
60 0.862179478009542
61 0.838074266910553
62 0.869791666666667
63 0.862914005915324
64 0.858640491962433
65 0.868856827418009
66 0.871861656506856
67 0.878338674704234
68 0.879340271155039
69 0.871260682741801
70 0.855902771155039
71 0.866853614648183
72 0.867254277070363
73 0.875
74 0.871260682741801
75 0.844484508037567
76 0.85877404610316
77 0.847889959812164
78 0.838274578253428
79 0.851629277070363
80 0.875801285107931
81 0.877871255079905
82 0.879273513952891
83 0.879139979680379
84 0.876535793145498
85 0.867321054140727
86 0.88735310236613
87 0.890090823173523
88 0.889022449652354
89 0.886151174704234
90 0.887686967849731
91 0.893963674704234
92 0.893563032150269
93 0.891092399756114
94 0.888488252957662
95 0.892494658629099
96 0.888888895511627
97 0.894831717014313
98 0.894898494084676
99 0.894898494084676
100 0.897702991962433
101 0.899238785107931
102 0.897302349408468
103 0.900173604488373
104 0.899906516075134
105 0.901642620563507
106 0.902176817258199
107 0.903645833333333
108 0.905515511830648
109 0.904447118441264
110 0.904380321502686
111 0.904380341370901
112 0.904714186986287
113 0.908787389596303
114 0.905181626478831
115 0.90645033121109
116 0.907118062178294
117 0.907986104488373
118 0.908653855323792
119 0.905181626478831
120 0.907652219136556
121 0.909455140431722
122 0.908186435699463
123 0.907318373521169
124 0.908386747042338
125 0.909521917502085
126 0.909722228844961
127 0.906717419624329
128 0.910723845163981
129 0.907852570215861
130 0.910256405671438
131 0.909388343493144
132 0.909254809220632
133 0.908453524112701
134 0.908787389596303
135 0.909588674704234
136 0.911591867605845
137 0.910657048225403
138 0.911458333333333
139 0.909254809220632
140 0.911725421746572
141 0.912593483924866
142 0.909655451774597
143 0.912059287230174
144 0.911792198816935
145 0.909989337126414
146 0.911258002122243
147 0.905982891718547
148 0.907986124356588
149 0.909188032150269
150 0.907385150591532
151 0.912059287230174
152 0.909455140431722
153 0.910256405671438
154 0.910924136638641
155 0.908453524112701
156 0.909722228844961
157 0.911525110403697
158 0.909054497877757
159 0.911925733089447
160 0.910256405671438
161 0.909254809220632
162 0.910990913709005
163 0.909922540187836
164 0.910323182741801
165 0.911258002122243
166 0.908854166666667
167 0.911124467849731
168 0.909455120563507
169 0.911391576131185
170 0.910790582497915
171 0.909989297389984
172 0.909254809220632
173 0.913060903549194
174 0.912593483924866
175 0.910723825295766
176 0.912459929784139
177 0.911258021990458
178 0.912860572338104
179 0.90952189763387
180 0.910056074460348
181 0.911458353201548
182 0.910523494084676
183 0.913060903549194
184 0.911391576131185
185 0.911792198816935
186 0.91079060236613
187 0.910924136638641
188 0.912393152713776
189 0.911191244920095
190 0.912259618441264
191 0.910790582497915
192 0.910523494084676
193 0.907986104488373
194 0.91079060236613
195 0.911925752957662
196 0.91079060236613
197 0.913995722929637
198 0.911057690779368
199 0.909121255079905
200 0.910323182741801
201 0.912192821502686
202 0.910056094328562
203 0.910189648469289
204 0.90892094373703
205 0.911725421746572
206 0.910590271155039
207 0.91386216878891
208 0.912593483924866
209 0.912126064300537
210 0.910790582497915
211 0.909521917502085
212 0.910990913709005
213 0.912793795267741
214 0.910189648469289
215 0.910857379436493
216 0.91199251015981
217 0.913394749164581
218 0.911391576131185
219 0.911925733089447
220 0.913060903549194
221 0.911658644676208
222 0.911525090535482
223 0.913261195023855
224 0.912927349408468
225 0.911391576131185
226 0.912660260995229
227 0.912459949652354
228 0.912393172581991
229 0.91466345389684
230 0.912793795267741
231 0.913795391718547
232 0.912727038065592
233 0.911992530028025
234 0.910723825295766
235 0.912793815135956
236 0.910723825295766
237 0.912660260995229
238 0.910389959812164
239 0.911057690779368
240 0.9140625
241 0.911458333333333
242 0.910723825295766
243 0.911057690779368
244 0.913728634516398
245 0.914129277070363
246 0.912994106610616
247 0.913060903549194
248 0.914062480131785
249 0.910523513952891
250 0.912126064300537
251 0.912593464056651
252 0.911124467849731
253 0.911725441614787
254 0.914196034272512
255 0.91426283121109
256 0.912727018197378
257 0.912727038065592
258 0.912593464056651
259 0.912192841370901
260 0.913862188657125
261 0.912126064300537
262 0.913261214892069
263 0.914463142553965
264 0.912459949652354
265 0.90952189763387
266 0.910389959812164
267 0.912727038065592
268 0.912393152713776
269 0.911992530028025
270 0.913795411586761
271 0.911658644676208
272 0.912059307098389
273 0.911124487717946
274 0.909722228844961
275 0.911057690779368
276 0.913327991962433
277 0.912994126478831
278 0.910389939943949
279 0.909989317258199
280 0.911124467849731
281 0.91346154610316
282 0.910590271155039
283 0.911391576131185
284 0.913928945859273
};
\addplot [line width=2.5pt, color1, opacity=1.0]
table {%
1 0.348290592432022
2 0.436431626478831
3 0.490718493858973
4 0.555355230967204
5 0.575186967849731
6 0.596287389596303
7 0.626135150591532
8 0.645833333333333
9 0.664196034272512
10 0.697248935699463
11 0.708800752957662
12 0.727497319380442
13 0.736778855323792
14 0.741786857446035
15 0.730969568093618
16 0.769698162873586
17 0.776842951774597
18 0.775774578253428
19 0.785523513952891
20 0.789797008037567
21 0.795739829540253
22 0.806891024112701
23 0.807759086290995
24 0.81497061252594
25 0.811164518197378
26 0.811631937821706
27 0.830462058385213
28 0.830061455567678
29 0.82772437731425
30 0.824586013952891
31 0.82565438747406
32 0.835202991962433
33 0.84147971868515
34 0.84354966878891
35 0.840945502122243
36 0.842481315135956
37 0.84581998984019
38 0.852697650591532
39 0.855101486047109
40 0.851896365483602
41 0.853699247042338
42 0.854099869728088
43 0.84889155626297
44 0.856436947981516
45 0.850894769032796
46 0.865785241127014
47 0.861979166666667
48 0.862446586290995
49 0.860243062178294
50 0.860243062178294
51 0.858306626478831
52 0.868322670459747
53 0.862780431906382
54 0.866252680619558
55 0.871327439943949
56 0.868255893389384
57 0.868589758872986
58 0.868055562178294
59 0.867053965727488
60 0.863114317258199
61 0.868322630723318
62 0.869190712769826
63 0.876335481802622
64 0.87727032105128
65 0.869057138760885
66 0.874399026234945
67 0.878806094328562
68 0.875868062178294
69 0.873530983924866
70 0.880942821502686
71 0.880809287230174
72 0.870325843493144
73 0.880408644676208
74 0.882745703061422
75 0.874532580375671
76 0.874065160751343
77 0.881543795267741
78 0.878138363361359
79 0.878539005915324
80 0.875
81 0.876068373521169
82 0.882478654384613
83 0.882278323173523
84 0.880275110403697
85 0.879874467849731
86 0.886551817258199
87 0.88014155626297
88 0.883680562178294
89 0.884548604488373
90 0.883880873521169
91 0.885950863361359
92 0.884415070215861
93 0.883680562178294
94 0.88855501015981
95 0.885817309220632
96 0.888621807098389
97 0.881143152713776
98 0.885950843493144
99 0.884949247042338
100 0.885349909464518
101 0.890958865483602
102 0.889022429784139
103 0.884548604488373
104 0.884147981802622
105 0.886151174704234
106 0.891493062178294
107 0.885149578253428
108 0.883680562178294
109 0.889890491962433
110 0.888488252957662
111 0.888087590535482
112 0.889623403549194
113 0.886752128601074
114 0.887887279192607
115 0.889756957689921
116 0.895566244920095
117 0.888087630271912
118 0.895299136638641
119 0.89389689763387
120 0.888221164544423
121 0.88795405626297
122 0.883279919624329
123 0.891359508037567
124 0.889957249164581
125 0.889957269032796
126 0.889222741127014
127 0.887152751286825
128 0.892427901426951
129 0.888822118441264
130 0.896968464056651
131 0.892227570215861
132 0.890224357446035
133 0.892828524112701
134 0.889823714892069
135 0.892628212769826
136 0.886885662873586
137 0.893963674704234
138 0.89636751015981
139 0.893162409464518
140 0.89309561252594
141 0.89616721868515
142 0.896100441614787
143 0.891960481802622
144 0.891626616319021
145 0.890625
146 0.891493042310079
147 0.891225973765055
148 0.888488252957662
149 0.89042466878891
150 0.888087610403697
151 0.895232379436493
152 0.887620190779368
153 0.894297540187836
154 0.891092419624329
155 0.89516560236613
156 0.896033664544423
157 0.896233975887299
158 0.894097228844961
159 0.895833353201548
160 0.893362700939178
161 0.892828524112701
162 0.898170411586761
163 0.891493062178294
164 0.892561435699463
165 0.892561455567678
166 0.899505873521169
167 0.895232359568278
168 0.895299136638641
169 0.892227550347646
170 0.894097228844961
171 0.892227550347646
172 0.893362720807393
173 0.897435903549194
174 0.896300752957662
175 0.899772981802622
176 0.895833333333333
177 0.897502660751343
178 0.897168815135956
179 0.895633002122243
180 0.893362700939178
181 0.897903303305308
182 0.900173604488373
183 0.896167198816935
184 0.895900090535482
185 0.900173624356588
186 0.90050748984019
187 0.900106847286224
188 0.897102038065592
189 0.8984375
190 0.899238785107931
191 0.896768152713776
192 0.899105230967204
193 0.894364317258199
194 0.896233995755514
195 0.896233975887299
196 0.894431094328562
197 0.899772981802622
198 0.896968503793081
199 0.897168795267741
200 0.899973273277283
201 0.897970080375671
202 0.898838142553965
203 0.896901706854502
204 0.899639427661896
205 0.895833333333333
206 0.896033644676208
207 0.900707801183065
208 0.898904919624329
209 0.894564628601074
210 0.903178413709005
211 0.898504277070363
212 0.893629809220632
213 0.898504277070363
214 0.894965271155039
215 0.900707801183065
216 0.898170411586761
217 0.902510682741801
218 0.898437480131785
219 0.896768152713776
220 0.899372339248657
221 0.897703011830648
222 0.898170411586761
223 0.897102038065592
224 0.900841355323792
225 0.898838142553965
226 0.896367530028025
227 0.902510702610016
228 0.902377148469289
229 0.902176817258199
230 0.902978082497915
231 0.899439116319021
232 0.902510682741801
233 0.899572670459747
234 0.897035260995229
235 0.900173624356588
236 0.898504277070363
237 0.900507469971975
238 0.898370742797852
239 0.900307158629099
240 0.895299156506856
241 0.899105230967204
242 0.900440692901611
243 0.901842951774597
244 0.901976486047109
245 0.899973273277283
246 0.903712610403697
247 0.898303945859273
248 0.905782580375671
249 0.902577459812164
250 0.895432690779368
251 0.902711013952891
252 0.90170939763387
253 0.904046456019084
254 0.899906535943349
255 0.898437519868215
256 0.89863783121109
257 0.898504277070363
258 0.896968503793081
259 0.898771365483602
260 0.895900110403697
261 0.896968464056651
262 0.901575843493144
263 0.90150906642278
264 0.89903845389684
265 0.900307158629099
266 0.903512279192607
267 0.899505873521169
268 0.898237188657125
269 0.898704608281453
270 0.902243574460348
271 0.901175200939178
272 0.904781003793081
273 0.903111656506856
274 0.901776174704234
275 0.900307158629099
276 0.902310351530711
277 0.901041646798452
278 0.900707801183065
279 0.902443905671438
280 0.898838142553965
281 0.899772981802622
282 0.899238785107931
283 0.901776174704234
284 0.902711013952891
};
\addplot [line width=2.5pt, blue, opacity=1.0]
table {%
1 0.514222770929337
2 0.602163463830948
3 0.540464758872986
4 0.674078524112701
5 0.717848569154739
6 0.675180286169052
7 0.75040066242218
8 0.763822108507156
9 0.752704322338104
10 0.790665060281754
11 0.794170677661896
12 0.759415060281754
13 0.799779623746872
14 0.824018448591232
15 0.837339758872986
16 0.831730753183365
17 0.826221972703934
18 0.83573716878891
19 0.821013629436493
20 0.854467153549194
21 0.844250798225403
22 0.850661069154739
23 0.831229954957962
24 0.852664262056351
25 0.845853358507156
26 0.853265225887299
27 0.857071310281754
28 0.866586536169052
29 0.85526841878891
30 0.867087364196777
31 0.867487967014313
32 0.85176283121109
33 0.8671875
34 0.872796475887299
35 0.876402229070663
36 0.872395843267441
37 0.870292454957962
38 0.874198704957962
39 0.873297274112701
40 0.877503991127014
41 0.873697906732559
42 0.881911069154739
43 0.879407048225403
44 0.876001626253128
45 0.880308479070663
46 0.87459933757782
47 0.881710737943649
48 0.882011204957962
49 0.887920647859573
50 0.883112967014313
51 0.88671875
52 0.886117786169052
53 0.892628192901611
54 0.886017620563507
55 0.889322936534882
56 0.883513629436493
57 0.89042466878891
58 0.890725165605545
59 0.895532846450806
60 0.893028825521469
61 0.893629789352417
62 0.895633012056351
63 0.89863783121109
64 0.900841325521469
65 0.896233975887299
66 0.894030451774597
67 0.897636204957962
68 0.895633012056351
69 0.903345346450806
70 0.903145045042038
71 0.901542484760284
72 0.903846144676208
73 0.902544051408768
74 0.901241987943649
75 0.907051265239716
76 0.904447108507156
77 0.905248403549194
78 0.908954322338104
79 0.906750798225403
80 0.906850963830948
81 0.910757213830948
82 0.909154623746872
83 0.909254789352417
84 0.910056084394455
85 0.912359774112701
86 0.907451927661896
87 0.909054487943649
88 0.909054487943649
89 0.910056084394455
90 0.912259638309479
91 0.905949503183365
92 0.907151430845261
93 0.907051265239716
94 0.910657048225403
95 0.910456717014313
96 0.908653855323792
97 0.912059307098389
98 0.907451927661896
99 0.905949503183365
100 0.907652258872986
101 0.901842951774597
102 0.897235572338104
103 0.899839758872986
104 0.881911039352417
105 0.889022409915924
106 0.880408644676208
107 0.889823704957962
108 0.887219548225403
109 0.887419879436493
110 0.890124201774597
111 0.898838132619858
112 0.881310075521469
113 0.892528057098389
114 0.898737996816635
115 0.893830150365829
116 0.879607379436493
117 0.890825301408768
118 0.892628222703934
119 0.898737967014313
120 0.882011234760284
121 0.896935075521469
122 0.897235572338104
123 0.895833313465118
124 0.894330948591232
125 0.887820512056351
126 0.884114593267441
127 0.895232379436493
128 0.891726762056351
129 0.894130617380142
130 0.905448704957962
131 0.894631415605545
132 0.897636204957962
133 0.897335737943649
134 0.901442289352417
135 0.899539262056351
136 0.897135436534882
137 0.897636234760284
138 0.897836536169052
139 0.90254408121109
140 0.895933508872986
141 0.905749201774597
142 0.904547274112701
143 0.911258012056351
144 0.916666686534882
145 0.915164262056351
146 0.915464758872986
147 0.913361370563507
148 0.916766822338104
149 0.917267620563507
150 0.917167484760284
151 0.916165858507156
152 0.918970346450806
153 0.916766822338104
154 0.916666686534882
155 0.920773237943649
156 0.915564894676208
157 0.918770015239716
158 0.916566491127014
159 0.916866987943649
160 0.914863765239716
161 0.909154653549194
162 0.901241987943649
163 0.908954322338104
164 0.910857379436493
165 0.903645813465118
166 0.904547274112701
167 0.912159472703934
168 0.908553689718246
169 0.909154653549194
170 0.913461536169052
171 0.910957545042038
172 0.882211536169052
173 0.89453125
174 0.897335737943649
175 0.896033674478531
176 0.899038463830948
177 0.897936701774597
178 0.8984375
179 0.898938298225403
180 0.891526460647583
181 0.898838132619858
182 0.901542484760284
183 0.900340557098389
184 0.904046475887299
185 0.901642620563507
186 0.899338960647583
187 0.899939894676208
188 0.905348569154739
189 0.904847741127014
190 0.903044879436493
191 0.89863783121109
192 0.903345376253128
193 0.902243584394455
194 0.901141852140427
195 0.899639427661896
196 0.896834939718246
197 0.904647439718246
198 0.905248373746872
199 0.903445512056351
200 0.904146641492844
201 0.903846144676208
202 0.906350165605545
203 0.901041656732559
204 0.900941520929337
205 0.902544051408768
206 0.892728388309479
207 0.891726762056351
208 0.903044849634171
209 0.898737996816635
210 0.900240391492844
211 0.898137032985687
212 0.889723539352417
213 0.899238765239716
214 0.91015625
215 0.913161069154739
216 0.910356551408768
217 0.913161069154739
218 0.914763599634171
219 0.915965557098389
220 0.915564924478531
221 0.91756808757782
222 0.915965557098389
223 0.917568117380142
224 0.916967153549194
225 0.916065722703934
226 0.913962334394455
227 0.917467951774597
228 0.9140625
229 0.915264397859573
230 0.913962364196777
231 0.915164262056351
232 0.917267650365829
233 0.918068915605545
234 0.915264427661896
235 0.914963930845261
236 0.914162665605545
237 0.911558479070663
238 0.916065692901611
239 0.914863795042038
240 0.91776841878891
241 0.918269217014313
242 0.916466355323792
243 0.913561701774597
244 0.916165888309479
245 0.918469548225403
246 0.915765225887299
247 0.917568117380142
248 0.913461536169052
249 0.915364563465118
250 0.913962334394455
251 0.91776841878891
252 0.914362996816635
253 0.91446316242218
254 0.917768448591232
255 0.918870180845261
256 0.916866987943649
257 0.918369382619858
258 0.914663434028625
259 0.916466355323792
260 0.917668282985687
261 0.916366189718246
262 0.919371008872986
263 0.918569684028625
264 0.917067319154739
265 0.915064096450806
266 0.916866987943649
267 0.916866987943649
268 0.91836941242218
269 0.918569713830948
270 0.916065692901611
271 0.917167484760284
272 0.916266024112701
273 0.919070512056351
274 0.919871777296066
275 0.914663463830948
276 0.916466355323792
277 0.913261234760284
278 0.919471144676208
279 0.918669849634171
280 0.915464729070663
281 0.917067289352417
282 0.919070512056351
283 0.914463132619858
284 0.919571310281754
};
\addplot [line width=2.5pt, red, opacity=1.0]
table {%
1 0.508680562178294
2 0.57525372505188
3 0.558226486047109
4 0.660657048225403
5 0.706129809220632
6 0.667000552018484
7 0.713474889596303
8 0.759481847286224
9 0.766559839248657
10 0.774639427661896
11 0.784321586290995
12 0.791132469971975
13 0.804153323173523
14 0.810830652713776
15 0.813501596450806
16 0.819911857446035
17 0.82752404610316
18 0.829927881558736
19 0.846420923868815
20 0.845619658629099
21 0.841613252957662
22 0.851629257202148
23 0.852564116319021
24 0.849425733089447
25 0.851228634516398
26 0.858506937821706
27 0.867321054140727
28 0.864783664544423
29 0.863715271155039
30 0.869057158629099
31 0.881610572338104
32 0.86985844373703
33 0.875400642553965
34 0.87706998984019
35 0.873998383680979
36 0.867654919624329
37 0.877537409464518
38 0.886351486047109
39 0.884147981802622
40 0.880876044432322
41 0.884281516075134
42 0.88795405626297
43 0.882478654384613
44 0.883213142553965
45 0.881009618441264
46 0.889423072338104
47 0.889423072338104
48 0.89042466878891
49 0.889022429784139
50 0.88321312268575
51 0.887620190779368
52 0.890892088413239
53 0.889289518197378
54 0.892694969971975
55 0.895232359568278
56 0.896634618441264
57 0.897702972094218
58 0.898637811342875
59 0.901642620563507
60 0.893095632394155
61 0.90090811252594
62 0.896834929784139
63 0.900307178497314
64 0.902110040187836
65 0.902577459812164
66 0.903846144676208
67 0.902243594328562
68 0.901509086290995
69 0.903846164544423
70 0.905315180619558
71 0.90377938747406
72 0.906316777070363
73 0.904914518197378
74 0.908319969971975
75 0.902911325295766
76 0.905916114648183
77 0.90625
78 0.905248383680979
79 0.905248403549194
80 0.905715803305308
81 0.908854166666667
82 0.906517088413239
83 0.910924156506856
84 0.909722228844961
85 0.90831998984019
86 0.912660241127014
87 0.911591867605845
88 0.910056094328562
89 0.910523494084676
90 0.912727018197378
91 0.913528303305308
92 0.911324779192607
93 0.914129277070363
94 0.912860572338104
95 0.913528323173523
96 0.91346154610316
97 0.913995742797852
98 0.916399578253428
99 0.911525090535482
100 0.912727018197378
101 0.913728634516398
102 0.913528323173523
103 0.913728634516398
104 0.913661857446035
105 0.913795411586761
106 0.913261234760284
107 0.917534708976746
108 0.91366187731425
109 0.912593483924866
110 0.915798624356588
111 0.915397981802622
112 0.91653311252594
113 0.914997319380442
114 0.914196054140727
115 0.916599889596303
116 0.914730230967204
117 0.914863785107931
118 0.918469548225403
119 0.917534708976746
120 0.918269217014313
121 0.916933755079905
122 0.917868594328562
123 0.916332801183065
124 0.918068905671438
125 0.919938564300537
126 0.920005341370901
127 0.915932158629099
128 0.918202479680379
129 0.920138875643412
130 0.919938544432322
131 0.919871787230174
132 0.918536305427551
133 0.920606295267741
134 0.919871807098389
135 0.920205652713776
136 0.921340803305308
137 0.921140491962433
138 0.921607891718547
139 0.92207533121109
140 0.922542730967204
141 0.921808222929637
142 0.919471144676208
143 0.922943353652954
144 0.921941777070363
145 0.922743042310079
146 0.925948182741801
147 0.921140491962433
148 0.920539518197378
149 0.922676285107931
150 0.924212078253428
151 0.924679478009542
152 0.923811415831248
153 0.923477570215861
154 0.923344016075134
155 0.921674688657125
156 0.923944969971975
157 0.923410773277283
158 0.924612700939178
159 0.924679478009542
160 0.922876596450806
161 0.925213674704234
162 0.924479166666667
163 0.926482379436493
164 0.921941777070363
165 0.923744638760885
166 0.923477570215861
167 0.922876616319021
168 0.92207533121109
169 0.926215271155039
170 0.92227562268575
171 0.92414528131485
172 0.922609508037567
173 0.924011747042338
174 0.927350441614787
175 0.923677881558736
176 0.925747851530711
177 0.923210481802622
178 0.927417198816935
179 0.92434561252594
180 0.92641560236613
181 0.925347208976746
182 0.924345632394155
183 0.924278855323792
184 0.924345632394155
185 0.923811435699463
186 0.926282048225403
187 0.925681074460348
188 0.92394498984019
189 0.925614317258199
190 0.925614317258199
191 0.925080120563507
192 0.925080120563507
193 0.922342419624329
194 0.92434561252594
195 0.925547540187836
196 0.924479186534882
197 0.923010130723318
198 0.925948182741801
199 0.925213674704234
200 0.923210481802622
201 0.925213674704234
202 0.925146917502085
203 0.925747851530711
204 0.926348825295766
205 0.926682710647583
206 0.924479166666667
207 0.928619106610616
208 0.924479146798452
209 0.924679478009542
210 0.923677881558736
211 0.924345632394155
212 0.923277258872986
213 0.925413986047109
214 0.926682690779368
215 0.926749467849731
216 0.925547520319621
217 0.925814628601074
218 0.924679497877757
219 0.924345632394155
220 0.926014939943949
221 0.925547560056051
222 0.924612700939178
223 0.924813032150269
224 0.924412389596303
225 0.926883021990458
226 0.92454594373703
227 0.924145301183065
228 0.92454594373703
229 0.923744658629099
230 0.924212078253428
231 0.92434561252594
232 0.923744658629099
233 0.923744658629099
234 0.925547520319621
235 0.925480763117472
236 0.923076907793681
237 0.92434561252594
238 0.922676265239716
239 0.923677881558736
240 0.925881425539652
241 0.924011747042338
242 0.923410793145498
243 0.923878192901611
244 0.925814628601074
245 0.922676285107931
246 0.924011766910553
247 0.92414528131485
248 0.926816244920095
249 0.922743062178294
250 0.924612700939178
251 0.925547540187836
252 0.927751064300537
253 0.925280431906382
254 0.924879829088847
255 0.921874980131785
256 0.925280471642812
257 0.924212058385213
258 0.923210481802622
259 0.925413986047109
260 0.923544347286224
261 0.923410793145498
262 0.923210481802622
263 0.924879809220632
264 0.923544347286224
265 0.926682690779368
266 0.922409196694692
267 0.923076907793681
268 0.925747851530711
269 0.924545963605245
270 0.923811435699463
271 0.925881405671438
272 0.925681094328562
273 0.926215271155039
274 0.924813032150269
275 0.924011766910553
276 0.92247595389684
277 0.923344016075134
278 0.923477590084076
279 0.923143684864044
280 0.923477550347646
281 0.925080120563507
282 0.922208865483602
283 0.921073714892069
284 0.922008554140727
};
\addplot [line width=2.5pt, color2, opacity=1.0]
table {%
2 0.506810913483302
4 0.595285773277283
6 0.651909728844961
8 0.733306626478831
10 0.776375532150269
12 0.783987720807393
14 0.792801817258199
16 0.824385682741801
18 0.836672008037567
20 0.836538473765055
22 0.84702189763387
24 0.846888363361359
26 0.853498915831248
28 0.863848845163981
30 0.872128744920095
32 0.871127148469289
34 0.878739317258199
36 0.876602570215861
38 0.876201927661896
40 0.883613785107931
42 0.8828125
44 0.889489849408468
46 0.889823714892069
48 0.887887279192607
50 0.891960461934408
52 0.885817309220632
54 0.889222760995229
56 0.892895301183065
58 0.891025642553965
60 0.892294327418009
62 0.890624980131785
64 0.895365913709005
66 0.89516560236613
68 0.893429497877757
70 0.898170411586761
72 0.89823716878891
74 0.890558222929637
76 0.89516560236613
78 0.896167198816935
80 0.895699799060822
82 0.894163986047109
84 0.895032048225403
86 0.896233956019084
88 0.89576655626297
90 0.897702991962433
92 0.896634618441264
94 0.897235592206319
96 0.895633021990458
98 0.897836526234945
100 0.897569457689921
102 0.898971676826477
104 0.897302349408468
106 0.897703011830648
108 0.900307158629099
110 0.894965271155039
112 0.900307158629099
114 0.896768152713776
116 0.894898494084676
118 0.902510682741801
120 0.898036857446035
122 0.901041666666667
124 0.901642620563507
126 0.900040050347646
128 0.897035241127014
130 0.900307138760885
132 0.900841355323792
134 0.902710994084676
136 0.902176817258199
138 0.901642640431722
140 0.903445521990458
142 0.901909708976746
144 0.900841335455577
146 0.898971696694692
148 0.900240381558736
150 0.905982891718547
152 0.903044879436493
154 0.899305562178294
156 0.904647449652354
158 0.902377148469289
160 0.901776194572449
162 0.902911345163981
164 0.902911325295766
166 0.901509086290995
168 0.906583865483602
170 0.901375532150269
172 0.898971676826477
174 0.899505873521169
176 0.902710994084676
178 0.90050748984019
180 0.901709417502085
182 0.901041666666667
184 0.904180030028025
186 0.903445521990458
188 0.906183242797852
190 0.902978122234344
192 0.908186415831248
194 0.906784176826477
196 0.903912921746572
198 0.906183242797852
200 0.903378744920095
202 0.902176837126414
204 0.905715823173523
206 0.904781003793081
208 0.904780983924866
210 0.904246807098389
212 0.901976486047109
214 0.905916134516398
216 0.902844548225403
218 0.903712610403697
220 0.906984508037567
222 0.906517088413239
224 0.907451927661896
226 0.906116465727488
228 0.903779367605845
230 0.905114829540253
232 0.905181606610616
234 0.90317843357722
236 0.903512279192607
238 0.902043263117472
240 0.902043263117472
242 0.905849357446035
244 0.904847760995229
246 0.905114849408468
248 0.903979698816935
250 0.902777791023254
252 0.907385150591532
254 0.903311967849731
256 0.901642620563507
258 0.906984488169352
260 0.905782580375671
262 0.905448714892069
264 0.905515491962433
266 0.901976486047109
268 0.90625
270 0.906784196694692
272 0.910523513952891
274 0.905582269032796
276 0.904046475887299
278 0.904580652713776
280 0.905181626478831
282 0.906717419624329
284 0.905649026234945
};
\end{axis}

\end{tikzpicture}

%% file: labpal/figure_data/performance_comparison_low_batch/plots_bs32/CIFAR-10_DenseNet-121_validation_accuracy2.pgf
% This file was created by tikzplotlib v0.9.8.
\begin{tikzpicture}

\definecolor{color0}{rgb}{0.933333333333333,0.509803921568627,0.933333333333333}
\definecolor{color1}{rgb}{0.647058823529412,0.164705882352941,0.164705882352941}
\definecolor{color2}{rgb}{1,0.647058823529412,0}

\begin{axis}[
minor xtick={},
minor ytick={},
tick align=outside,
tick pos=left,
width=10.5cm,height=8cm,grid=major,major grid style={dotted},
x grid style={white!69.0196078431373!black},
xlabel={epochs},
xmin=-13.15, xmax=298.15,
xtick style={color=black},
xtick={-50,0,50,100,150,200,250,300},
y grid style={white!69.0196078431373!black},
ylabel={validation accuracy},
ymin=0.85, ymax=0.94,
ytick style={color=black},
ytick={0.85,0.86,0.87,0.88,0.89,0.9,0.91,0.92,0.93,0.94}
]
\addplot [line width=2.5pt, color0, opacity=1.0]
table {%
1 0.377003192901611
2 0.350761204957962
3 0.512019217014313
4 0.552684307098389
5 0.599959909915924
6 0.621594548225403
7 0.654847741127014
8 0.67988783121109
9 0.663461565971375
10 0.66386216878891
11 0.707131385803223
12 0.722355782985687
13 0.733173072338104
14 0.750801265239716
15 0.754006385803223
16 0.741185903549194
17 0.75741183757782
18 0.7734375
19 0.793669879436493
20 0.787660241127014
21 0.795673072338104
22 0.809294879436493
23 0.805288434028625
24 0.798277258872986
25 0.799879789352417
26 0.81991183757782
27 0.794471144676208
28 0.809895813465118
29 0.810096144676208
30 0.822916686534882
31 0.817307710647583
32 0.827323734760284
33 0.837339758872986
34 0.832331717014313
35 0.829126596450806
36 0.847355782985687
37 0.847956717014313
38 0.850160241127014
39 0.852163434028625
40 0.842748403549194
41 0.844751596450806
42 0.854967951774597
43 0.854967951774597
44 0.847556114196777
45 0.843149065971375
46 0.855568885803223
47 0.864783644676208
48 0.845953524112701
49 0.854366958141327
50 0.85897433757782
51 0.855969548225403
52 0.845953524112701
53 0.870993614196777
54 0.853165090084076
55 0.844951927661896
56 0.86678683757782
57 0.860176265239716
58 0.852964758872986
59 0.861578524112701
60 0.865184307098389
61 0.863381385803223
62 0.864983975887299
63 0.85897433757782
64 0.85957533121109
65 0.86738783121109
66 0.854567289352417
67 0.8671875
68 0.866185903549194
69 0.862379789352417
70 0.864983975887299
71 0.865384638309479
72 0.86678683757782
73 0.871995210647583
74 0.873397409915924
75 0.87479966878891
76 0.877003192901611
77 0.863782048225403
78 0.868589758872986
79 0.871794879436493
80 0.87479966878891
81 0.872996807098389
82 0.87540066242218
83 0.878605782985687
84 0.878605782985687
85 0.874399065971375
86 0.878405451774597
87 0.883413434028625
88 0.8828125
89 0.87479966878891
90 0.885616958141327
91 0.894230782985687
92 0.889423072338104
93 0.877604186534882
94 0.88301283121109
95 0.886017620563507
96 0.888020813465118
97 0.880608975887299
98 0.881810903549194
99 0.886217951774597
100 0.889423072338104
101 0.884415090084076
102 0.886418282985687
103 0.888621807098389
104 0.895432710647583
105 0.887820541858673
106 0.886418282985687
107 0.885817289352417
108 0.887019217014313
109 0.889623403549194
110 0.88301283121109
111 0.880608975887299
112 0.89042466878891
113 0.889823734760284
114 0.889623403549194
115 0.887620210647583
116 0.887219548225403
117 0.882211565971375
118 0.89102566242218
119 0.88261216878891
120 0.886818885803223
121 0.891626596450806
122 0.895232379436493
123 0.878205120563507
124 0.892628192901611
125 0.89102566242218
126 0.880208313465118
127 0.894030451774597
128 0.883413434028625
129 0.89022433757782
130 0.885016024112701
131 0.888421475887299
132 0.888621807098389
133 0.892828524112701
134 0.883814096450806
135 0.893229186534882
136 0.886618614196777
137 0.887219548225403
138 0.887219548225403
139 0.887419879436493
140 0.889623403549194
141 0.894631385803223
142 0.89082533121109
143 0.893629789352417
144 0.888421475887299
145 0.890625
146 0.878806114196777
147 0.888421475887299
148 0.893229186534882
149 0.89823716878891
150 0.883613765239716
151 0.889423072338104
152 0.896634638309479
153 0.884415090084076
154 0.899439096450806
155 0.891225934028625
156 0.893830120563507
157 0.886217951774597
158 0.888221144676208
159 0.876802861690521
160 0.887820541858673
161 0.886818885803223
162 0.893629789352417
163 0.883814096450806
164 0.893028855323792
165 0.892027258872986
166 0.896834909915924
167 0.900440692901611
168 0.886017620563507
169 0.892227590084076
170 0.885016024112701
171 0.891826927661896
172 0.89102566242218
173 0.893629789352417
174 0.883613765239716
175 0.897435903549194
176 0.893629789352417
177 0.894030451774597
178 0.897035241127014
179 0.884615361690521
180 0.899839758872986
181 0.895232379436493
182 0.890625
183 0.886618614196777
184 0.88321316242218
185 0.892227590084076
186 0.896634638309479
187 0.900641024112701
188 0.887219548225403
189 0.895432710647583
190 0.896834909915924
191 0.894831717014313
192 0.891626596450806
193 0.891626596450806
194 0.897435903549194
195 0.895633041858673
196 0.891426265239716
197 0.902043282985687
198 0.896634638309479
199 0.900440692901611
200 0.900240361690521
201 0.897035241127014
202 0.896033644676208
203 0.896233975887299
204 0.888221144676208
205 0.894831717014313
206 0.893429458141327
207 0.893028855323792
208 0.890625
209 0.892227590084076
210 0.888421475887299
211 0.893229186534882
212 0.894230782985687
213 0.884615361690521
214 0.897235572338104
215 0.890625
216 0.888221144676208
217 0.889022409915924
218 0.890625
219 0.89102566242218
220 0.889823734760284
221 0.889423072338104
222 0.892027258872986
223 0.886017620563507
224 0.893629789352417
225 0.891426265239716
226 0.895232379436493
227 0.893629789352417
228 0.889022409915924
229 0.897636234760284
230 0.886217951774597
231 0.896434307098389
232 0.888822138309479
233 0.891426265239716
234 0.89102566242218
235 0.890625
236 0.896634638309479
237 0.891426265239716
238 0.894230782985687
239 0.890024065971375
240 0.890024065971375
241 0.892828524112701
242 0.894431114196777
243 0.888621807098389
244 0.89022433757782
245 0.891826927661896
246 0.895432710647583
247 0.896834909915924
248 0.895633041858673
249 0.884214758872986
250 0.889423072338104
251 0.887219548225403
252 0.89022433757782
253 0.892227590084076
254 0.887019217014313
255 0.895432710647583
256 0.891225934028625
257 0.892227590084076
258 0.897235572338104
259 0.892628192901611
260 0.893429458141327
261 0.895232379436493
262 0.892027258872986
263 0.893229186534882
264 0.89823716878891
265 0.896634638309479
266 0.894230782985687
267 0.896233975887299
268 0.888020813465118
269 0.887820541858673
270 0.893629789352417
271 0.892027258872986
272 0.893229186534882
273 0.888221144676208
274 0.889222741127014
275 0.892427861690521
276 0.896033644676208
277 0.896834909915924
278 0.891826927661896
279 0.895032048225403
280 0.891426265239716
281 0.892628192901611
282 0.888221144676208
283 0.894631385803223
284 0.889423072338104
};
\addplot [line width=2.5pt, color1, opacity=1.0]
table {%
1 0.113381413122018
2 0.22228899349769
3 0.296407575408618
4 0.517962058385213
5 0.472155456741651
6 0.678285241127014
7 0.320512823760509
8 0.5172275553147
9 0.492588135103385
10 0.462406526009242
11 0.303218478957812
12 0.545539528131485
13 0.210403308272362
14 0.291933755079905
15 0.503739312291145
16 0.258279919624329
};
\addplot [line width=2.5pt, red, opacity=1.0]
table {%
1 0.493990391492844
2 0.566907048225403
3 0.579026460647583
4 0.620192319154739
5 0.695913463830948
6 0.683193117380142
7 0.70272433757782
8 0.762820512056351
9 0.733273237943649
10 0.761217951774597
11 0.795172274112701
12 0.780749201774597
13 0.802984774112701
14 0.800480753183365
15 0.814002394676208
16 0.828225165605545
17 0.833733975887299
18 0.843249201774597
19 0.84805691242218
20 0.853665858507156
21 0.850761204957962
22 0.849959939718246
23 0.852864563465118
24 0.853866189718246
25 0.866286039352417
26 0.866486400365829
27 0.872095346450806
28 0.871895045042038
29 0.871794879436493
30 0.876201927661896
31 0.867287665605545
32 0.879407048225403
33 0.881310105323792
34 0.882011204957962
35 0.881810903549194
36 0.869691520929337
37 0.887219548225403
38 0.888822108507156
39 0.887219548225403
40 0.887319713830948
41 0.884715557098389
42 0.894330948591232
43 0.890625
44 0.889923900365829
45 0.890324503183365
46 0.886117786169052
47 0.889923870563507
48 0.889423072338104
49 0.895532876253128
50 0.897435873746872
51 0.898737996816635
52 0.890625
53 0.888521641492844
54 0.898337364196777
55 0.900340527296066
56 0.889623403549194
57 0.896434277296066
58 0.893529653549194
59 0.899839729070663
60 0.89473158121109
61 0.896033674478531
62 0.900240391492844
63 0.901342153549194
64 0.899539262056351
65 0.902644217014313
66 0.903746008872986
67 0.900040060281754
68 0.903946310281754
69 0.906951129436493
70 0.901542454957962
71 0.904747605323792
72 0.909254789352417
73 0.907852590084076
74 0.904146641492844
75 0.908353358507156
76 0.903846174478531
77 0.907051295042038
78 0.908052861690521
79 0.90234375
80 0.908553689718246
81 0.906850963830948
82 0.909254789352417
83 0.909955948591232
84 0.911558508872986
85 0.905649036169052
86 0.908653855323792
87 0.915164262056351
88 0.91015625
89 0.907051265239716
90 0.909655451774597
91 0.91055691242218
92 0.909054487943649
93 0.914463132619858
94 0.91035658121109
95 0.910556882619858
96 0.911458343267441
97 0.911959111690521
98 0.912860572338104
99 0.9140625
100 0.913261234760284
101 0.913261204957962
102 0.913962334394455
103 0.917467951774597
104 0.913161039352417
105 0.919871777296066
106 0.917267620563507
107 0.919471144676208
108 0.918970346450806
109 0.919571310281754
110 0.920172274112701
111 0.917367786169052
112 0.918970346450806
113 0.920572936534882
114 0.920272439718246
115 0.919871807098389
116 0.919971972703934
117 0.918870180845261
118 0.921975135803223
119 0.920272439718246
120 0.919971942901611
121 0.922475963830948
122 0.923277229070663
123 0.925480753183365
124 0.924278855323792
125 0.923778057098389
126 0.919170677661896
127 0.922976762056351
128 0.921975165605545
129 0.923477590084076
130 0.921975165605545
131 0.920472741127014
132 0.919471144676208
133 0.920472741127014
134 0.919571310281754
135 0.925380617380142
136 0.918970346450806
137 0.92167466878891
138 0.921374201774597
139 0.920072108507156
140 0.920973539352417
141 0.920472770929337
142 0.925580948591232
143 0.922175496816635
144 0.92167466878891
145 0.923377394676208
146 0.925380617380142
147 0.921374201774597
148 0.922275632619858
149 0.922976762056351
150 0.923477560281754
151 0.924379020929337
152 0.924178689718246
153 0.922375798225403
154 0.923778057098389
155 0.920973569154739
156 0.924579352140427
157 0.922676265239716
158 0.92207533121109
159 0.920673072338104
160 0.921374201774597
161 0.920172274112701
162 0.924479156732559
163 0.921975135803223
164 0.922676295042038
165 0.923076927661896
166 0.920773237943649
167 0.921274065971375
168 0.921975135803223
169 0.926081746816635
170 0.924479156732559
171 0.921774834394455
172 0.922776430845261
173 0.926181882619858
174 0.924178689718246
175 0.920572936534882
176 0.921875
177 0.924879819154739
178 0.925180286169052
179 0.923577725887299
180 0.923878222703934
181 0.923577725887299
182 0.92578125
183 0.924679487943649
184 0.926282048225403
185 0.922576129436493
186 0.92227566242218
187 0.926081717014313
188 0.924579322338104
189 0.925080150365829
190 0.924679487943649
191 0.922876626253128
192 0.923577725887299
193 0.922976762056351
194 0.924379020929337
195 0.927383810281754
196 0.924178689718246
197 0.924779653549194
198 0.921574532985687
199 0.925881415605545
200 0.921875
201 0.921875
202 0.922676295042038
203 0.923778027296066
204 0.924078524112701
205 0.92167466878891
206 0.924679487943649
207 0.924679487943649
208 0.926282048225403
209 0.925280451774597
210 0.926282048225403
211 0.923778027296066
212 0.928485572338104
213 0.922175496816635
214 0.925080120563507
215 0.926181882619858
216 0.924479156732559
217 0.926782846450806
218 0.925280421972275
219 0.926482379436493
220 0.924879819154739
221 0.92598158121109
222 0.925380617380142
223 0.925080120563507
224 0.926983177661896
225 0.926382213830948
226 0.926081717014313
227 0.923277229070663
228 0.924078524112701
229 0.927383840084076
230 0.92558091878891
231 0.926081717014313
232 0.924679487943649
233 0.92618191242218
234 0.926081746816635
235 0.926682680845261
236 0.92598158121109
237 0.927884608507156
238 0.926282048225403
239 0.926181882619858
240 0.927083343267441
241 0.926682710647583
242 0.928084939718246
243 0.927083313465118
244 0.925881415605545
245 0.925881415605545
246 0.928485572338104
247 0.926582545042038
248 0.927483975887299
249 0.928685903549194
250 0.927984774112701
251 0.927183479070663
252 0.926081746816635
253 0.927183508872986
254 0.927984774112701
255 0.926682680845261
256 0.928886204957962
257 0.927684307098389
258 0.927483975887299
259 0.925681084394455
260 0.928685903549194
261 0.928685903549194
262 0.927083343267441
263 0.92538058757782
264 0.925981551408768
265 0.929186701774597
266 0.929286867380142
267 0.928485572338104
268 0.927383840084076
269 0.927584141492844
270 0.928285241127014
271 0.926482379436493
272 0.926482349634171
273 0.927283644676208
274 0.928385406732559
275 0.927283644676208
276 0.927784472703934
277 0.927483975887299
278 0.930989593267441
279 0.930689096450806
280 0.927984774112701
281 0.930288463830948
282 0.927684307098389
283 0.928385406732559
284 0.93008816242218
};
\addplot [line width=2.5pt, color2, opacity=1.0]
table {%
2 0.449385672807693
4 0.568977038065592
6 0.631477018197378
8 0.667467951774597
10 0.69511216878891
12 0.725293795267741
14 0.753939628601074
16 0.759682138760885
18 0.770032048225403
20 0.784188012282054
22 0.791466355323792
24 0.811298072338104
26 0.818108975887299
28 0.835202991962433
30 0.83059561252594
32 0.831463694572449
34 0.841947118441264
36 0.849626084168752
38 0.85650372505188
40 0.858640491962433
42 0.857171495755514
44 0.860777239004771
46 0.863181094328562
48 0.872262279192607
50 0.87560095389684
52 0.876135150591532
54 0.879273513952891
56 0.880208333333333
58 0.878872871398926
60 0.885016024112701
62 0.883213142553965
64 0.880008002122243
66 0.881543815135956
68 0.883279919624329
70 0.879807690779368
72 0.880208333333333
74 0.881477038065592
76 0.875467419624329
78 0.888822118441264
80 0.883947650591532
82 0.890558222929637
84 0.886885662873586
86 0.885750532150269
88 0.881944437821706
90 0.884548624356588
92 0.891893704732259
94 0.881744126478831
96 0.885216335455577
98 0.881410260995229
100 0.891159176826477
102 0.881076375643412
104 0.885149578253428
106 0.88054221868515
108 0.88488248984019
110 0.884949247042338
112 0.886752128601074
114 0.887753744920095
116 0.886217951774597
118 0.883680542310079
120 0.889289538065592
122 0.873597741127014
124 0.888287921746572
126 0.892227570215861
128 0.885950843493144
130 0.885616997877757
132 0.891693373521169
134 0.885216335455577
136 0.883747339248657
138 0.885416646798452
140 0.88775372505188
142 0.889489849408468
144 0.886752128601074
146 0.889356315135956
148 0.886485040187836
150 0.885750532150269
152 0.887086013952891
154 0.875934819380442
156 0.886551817258199
158 0.879273494084676
160 0.887419879436493
162 0.887686967849731
164 0.885950843493144
166 0.886952459812164
168 0.885817309220632
170 0.89329594373703
172 0.888488233089447
174 0.88568377494812
176 0.887954076131185
178 0.894898494084676
180 0.884615381558736
182 0.885950863361359
184 0.891159176826477
186 0.88488248984019
188 0.892027258872986
190 0.883480230967204
192 0.892160793145498
194 0.891292750835419
196 0.888488252957662
198 0.886885682741801
200 0.890892088413239
202 0.879941244920095
204 0.88341345389684
206 0.889155964056651
208 0.882411857446035
210 0.895833333333333
212 0.890624980131785
214 0.887686947981516
216 0.885750532150269
218 0.888688584168752
220 0.889222760995229
222 0.884281535943349
224 0.887820502122243
226 0.887820521990458
228 0.889957269032796
230 0.887820521990458
232 0.890691777070363
234 0.895165582497915
236 0.888488252957662
238 0.890224357446035
240 0.888020833333333
242 0.882745722929637
244 0.891025642553965
246 0.890691777070363
248 0.888955652713776
250 0.891693393389384
252 0.891159196694692
254 0.893763363361359
256 0.894230763117472
258 0.892227550347646
260 0.896167198816935
262 0.895900110403697
264 0.892895301183065
266 0.898771365483602
268 0.88488248984019
270 0.892160793145498
272 0.893830120563507
274 0.894965271155039
276 0.888621807098389
278 0.897369106610616
280 0.887620190779368
282 0.893696586290995
284 0.894163986047109
};
\addplot [line width=2.5pt, green!50.1960784313725!black, opacity=1.0]
table {%
1 0.462473293145498
2 0.556290070215861
3 0.620058755079905
4 0.679220100243886
5 0.702991445859273
6 0.730835994084676
7 0.76008282105128
8 0.78104966878891
9 0.793068905671438
10 0.792200843493144
11 0.762753744920095
12 0.793469548225403
13 0.786124467849731
14 0.819911857446035
15 0.82792466878891
16 0.832932670911153
17 0.845152239004771
18 0.842881937821706
19 0.845753212769826
20 0.855502128601074
21 0.849225421746572
22 0.857572118441264
23 0.857371807098389
24 0.86144498984019
25 0.863715271155039
26 0.862847208976746
27 0.867254257202148
28 0.870058755079905
29 0.874999980131785
30 0.870592931906382
31 0.87252938747406
32 0.872996787230174
33 0.87706998984019
34 0.869724889596303
35 0.871594548225403
36 0.870592951774597
37 0.881610572338104
38 0.873197138309479
39 0.879807690779368
40 0.873798092206319
41 0.880608975887299
42 0.875801265239716
43 0.883146365483602
44 0.880208333333333
45 0.882946034272512
46 0.881677369276682
47 0.87974093357722
48 0.883747319380442
49 0.884882469971975
50 0.881209949652354
51 0.883547008037567
52 0.883680542310079
53 0.883880893389384
54 0.883079588413239
55 0.891626616319021
56 0.884081184864044
57 0.8828125
58 0.887553413709005
59 0.887686967849731
60 0.885817309220632
61 0.887686967849731
62 0.888688564300537
63 0.892628212769826
64 0.888020833333333
65 0.89082533121109
66 0.893763343493144
67 0.889489849408468
68 0.892361104488373
69 0.885884086290995
70 0.885149578253428
71 0.891426285107931
72 0.89536593357722
73 0.894163986047109
74 0.892160773277283
75 0.892027239004771
76 0.888020833333333
77 0.889957249164581
78 0.890825311342875
79 0.895566244920095
80 0.890090823173523
81 0.893229166666667
82 0.892561435699463
83 0.893028855323792
84 0.893830120563507
85 0.894564628601074
86 0.89122595389684
87 0.89309561252594
88 0.891760150591532
89 0.885683755079905
90 0.893629809220632
91 0.892160813013713
92 0.890958885351817
93 0.892895301183065
94 0.890958865483602
95 0.894831736882528
96 0.889890472094218
97 0.895432690779368
98 0.888421495755514
99 0.895633021990458
100 0.890892108281453
101 0.894364317258199
102 0.897369126478831
103 0.88795405626297
104 0.893629809220632
105 0.891826927661896
106 0.892494658629099
107 0.889890491962433
108 0.893563032150269
109 0.887086013952891
110 0.892561435699463
111 0.891225973765055
112 0.892828524112701
113 0.892962058385213
114 0.89042466878891
115 0.896434287230174
116 0.890424688657125
117 0.889489849408468
118 0.890224357446035
119 0.893963674704234
120 0.894831736882528
121 0.89596688747406
122 0.894631425539652
123 0.890157580375671
124 0.896100421746572
125 0.897235572338104
126 0.895833333333333
127 0.899572650591532
128 0.895299136638641
129 0.89596688747406
130 0.896701375643412
131 0.893563032150269
132 0.89783654610316
133 0.898704588413239
134 0.895566244920095
135 0.897970080375671
136 0.892962078253428
137 0.896968483924866
138 0.886017640431722
139 0.892494658629099
140 0.889423072338104
141 0.894631405671438
142 0.895699799060822
143 0.905916134516398
144 0.905381957689921
145 0.907518704732259
146 0.906049688657125
147 0.905715803305308
148 0.905715823173523
149 0.906984508037567
150 0.907852570215861
151 0.909722208976746
152 0.909989317258199
153 0.909321586290995
154 0.906450311342875
155 0.909054478009542
156 0.907852570215861
157 0.90604966878891
158 0.908520301183065
159 0.908453524112701
160 0.908520301183065
161 0.910323182741801
162 0.908653835455577
163 0.907051304976145
164 0.90685095389684
165 0.910456736882528
166 0.912526706854502
167 0.910924136638641
168 0.910189628601074
169 0.912059287230174
170 0.909989317258199
171 0.911925752957662
172 0.910590271155039
173 0.910323182741801
174 0.913261214892069
175 0.908587058385213
176 0.910924136638641
177 0.907785793145498
178 0.911658664544423
179 0.91199251015981
180 0.91159188747406
181 0.908587078253428
182 0.91346154610316
183 0.913060903549194
184 0.911992530028025
185 0.913862188657125
186 0.911057690779368
187 0.909588674704234
188 0.911858975887299
189 0.909655451774597
190 0.910657048225403
191 0.913928945859273
192 0.912860572338104
193 0.907919347286224
194 0.910990913709005
195 0.90892094373703
196 0.90831998984019
197 0.911124467849731
198 0.910990913709005
199 0.913261214892069
200 0.909989317258199
201 0.912860572338104
202 0.910456736882528
203 0.911124467849731
204 0.912727038065592
205 0.912593483924866
206 0.913194457689921
207 0.91426283121109
208 0.910924156506856
209 0.910924136638641
210 0.913327972094218
211 0.910657048225403
212 0.911258002122243
213 0.909054478009542
214 0.913862188657125
215 0.913928965727488
216 0.913194457689921
217 0.910924136638641
218 0.908987700939178
219 0.916466335455577
220 0.913728634516398
221 0.91446312268575
222 0.909054478009542
223 0.913194457689921
224 0.91426283121109
225 0.909922540187836
226 0.910456736882528
227 0.910857359568278
228 0.912192841370901
229 0.913261214892069
230 0.911925733089447
231 0.910723825295766
232 0.912526706854502
233 0.912727038065592
234 0.910723825295766
235 0.913194437821706
236 0.916132469971975
237 0.911858975887299
238 0.913595100243886
239 0.911725421746572
240 0.914863785107931
241 0.911525110403697
242 0.913261214892069
243 0.90892094373703
244 0.912326395511627
245 0.910657048225403
246 0.911658644676208
247 0.911258002122243
248 0.912192841370901
249 0.911258021990458
250 0.913795391718547
251 0.911792198816935
252 0.911525090535482
253 0.910657048225403
254 0.910056074460348
255 0.91079060236613
256 0.913595080375671
257 0.911792198816935
258 0.911658644676208
259 0.911658644676208
260 0.912793795267741
261 0.912393152713776
262 0.909588674704234
263 0.910389959812164
264 0.913127660751343
265 0.913595080375671
266 0.91119126478831
267 0.908253192901611
268 0.910523494084676
269 0.908653835455577
270 0.910590271155039
271 0.914797008037567
272 0.915197650591532
273 0.912326375643412
274 0.912459949652354
275 0.912459929784139
276 0.911925752957662
277 0.911591867605845
278 0.911525110403697
279 0.914463142553965
280 0.910323202610016
281 0.910657048225403
282 0.911124467849731
283 0.910189648469289
284 0.915665050347646
};
\addplot [line width=2.5pt, black, opacity=1.0]
table {%
1 0.458333333333333
2 0.500801275173823
3 0.59602028131485
4 0.638087610403697
5 0.685096164544423
6 0.732905983924866
7 0.749465823173523
8 0.73110310236613
9 0.775106847286224
10 0.791666666666667
11 0.79440438747406
12 0.802283644676208
13 0.827590823173523
14 0.82605501015981
15 0.826989849408468
16 0.827991465727488
17 0.825921456019084
18 0.827791134516398
19 0.838207801183065
20 0.839009086290995
21 0.850828011830648
22 0.842080652713776
23 0.856436967849731
24 0.863314648469289
25 0.866786857446035
26 0.872863272825877
27 0.878806094328562
28 0.878472248713175
29 0.880876044432322
30 0.881944437821706
31 0.869324266910553
32 0.865050752957662
33 0.883346676826477
34 0.883413473765055
35 0.88034188747406
36 0.890357891718547
37 0.888822118441264
38 0.89022437731425
39 0.884081184864044
40 0.885616978009542
41 0.886885682741801
42 0.889823714892069
43 0.884615401426951
44 0.881076375643412
45 0.890691757202148
46 0.895365913709005
47 0.896634618441264
48 0.901308755079905
49 0.899505873521169
50 0.900307158629099
51 0.896434307098389
52 0.898971696694692
53 0.901509086290995
54 0.903445502122243
55 0.902510662873586
56 0.901909708976746
57 0.899172008037567
58 0.903912921746572
59 0.901241997877757
60 0.900440692901611
61 0.903645833333333
62 0.909455120563507
63 0.905649026234945
64 0.904781003793081
65 0.904647429784139
66 0.905315160751343
67 0.902510702610016
68 0.903378744920095
69 0.904847741127014
70 0.905181626478831
71 0.903712610403697
72 0.903044859568278
73 0.905315160751343
74 0.904246807098389
75 0.906784196694692
76 0.907719016075134
77 0.90625
78 0.904513895511627
79 0.905048072338104
80 0.908453524112701
81 0.906116445859273
82 0.908052881558736
83 0.909922540187836
84 0.903979698816935
85 0.905715803305308
86 0.906650642553965
87 0.908119658629099
88 0.90564904610316
89 0.909254789352417
90 0.905114849408468
91 0.907785793145498
92 0.907184819380442
93 0.905982911586761
94 0.909121255079905
95 0.906784196694692
96 0.90685095389684
97 0.909855763117472
98 0.908987700939178
99 0.907986124356588
100 0.908253212769826
101 0.908520301183065
102 0.905448714892069
103 0.910256405671438
104 0.910389959812164
105 0.90932156642278
106 0.910590271155039
107 0.909655431906382
108 0.910189628601074
109 0.910389959812164
110 0.907251596450806
111 0.909321586290995
112 0.909722208976746
113 0.909254809220632
114 0.909989317258199
115 0.910256405671438
116 0.909054497877757
117 0.908787389596303
118 0.906784176826477
119 0.910323182741801
120 0.908987720807393
121 0.908453524112701
122 0.909788986047109
123 0.910256405671438
124 0.908453524112701
125 0.909455140431722
126 0.910189648469289
127 0.907585461934408
128 0.907251596450806
129 0.911258021990458
130 0.910122871398926
131 0.910590271155039
132 0.910323202610016
133 0.91079060236613
134 0.910389959812164
135 0.912860592206319
136 0.90892094373703
137 0.912593483924866
138 0.91159188747406
139 0.910723825295766
140 0.909655451774597
141 0.912727018197378
142 0.912393172581991
143 0.912459929784139
144 0.910189628601074
145 0.910456736882528
146 0.907919347286224
147 0.908520301183065
148 0.914930542310079
149 0.908520301183065
150 0.910924136638641
151 0.911191244920095
152 0.912459929784139
153 0.911992530028025
154 0.913795411586761
155 0.91159188747406
156 0.90912127494812
157 0.910122871398926
158 0.913595080375671
159 0.910790582497915
160 0.911525130271912
161 0.910456736882528
162 0.91099093357722
163 0.911725421746572
164 0.911124447981516
165 0.910456736882528
166 0.909188032150269
167 0.910523494084676
168 0.910056094328562
169 0.910723825295766
170 0.910122851530711
171 0.910323162873586
172 0.908119658629099
173 0.912727018197378
174 0.914596676826477
175 0.909922560056051
176 0.909788986047109
177 0.909388363361359
178 0.910590291023254
179 0.911792198816935
180 0.912393172581991
181 0.909388363361359
182 0.910924136638641
183 0.910456736882528
184 0.910256385803223
185 0.911124447981516
186 0.912192841370901
187 0.911792198816935
188 0.911124447981516
189 0.908854166666667
190 0.911458313465118
191 0.909254809220632
192 0.913060903549194
193 0.909054478009542
194 0.912393172581991
195 0.909588674704234
196 0.906917730967204
197 0.912192841370901
198 0.910323202610016
199 0.912927349408468
200 0.910723825295766
201 0.913795391718547
202 0.911191244920095
203 0.912126084168752
204 0.910456717014313
205 0.912526706854502
206 0.910924136638641
207 0.909588654836019
208 0.912593483924866
209 0.911725421746572
210 0.910256405671438
211 0.910523494084676
212 0.91139155626297
213 0.909188032150269
214 0.911725421746572
215 0.911792198816935
216 0.909722228844961
217 0.908453524112701
218 0.911124447981516
219 0.912192841370901
220 0.910590271155039
221 0.912994126478831
222 0.913194437821706
223 0.909722228844961
224 0.910056074460348
225 0.91159188747406
226 0.91079060236613
227 0.910456736882528
228 0.913595080375671
229 0.909722208976746
230 0.91079060236613
231 0.911458333333333
232 0.91139155626297
233 0.912927329540253
234 0.912459949652354
235 0.911324779192607
236 0.911858975887299
237 0.911725421746572
238 0.908052901426951
239 0.909521877765656
240 0.91159188747406
241 0.911258021990458
242 0.908854166666667
243 0.909655451774597
244 0.909655431906382
245 0.913528323173523
246 0.913728634516398
247 0.909054497877757
248 0.91179221868515
249 0.910790582497915
250 0.90892094373703
251 0.910189648469289
252 0.912126064300537
253 0.909855782985687
254 0.909922560056051
255 0.909789005915324
256 0.910990913709005
257 0.911591867605845
258 0.910456717014313
259 0.909788986047109
260 0.913461526234945
261 0.911792198816935
262 0.910857359568278
263 0.913194457689921
264 0.913127660751343
265 0.912994126478831
266 0.911725421746572
267 0.910990913709005
268 0.912126064300537
269 0.911925752957662
270 0.911258021990458
271 0.909722208976746
272 0.911658644676208
273 0.909788986047109
274 0.912326375643412
275 0.910924136638641
276 0.912326395511627
277 0.910122871398926
278 0.912192841370901
279 0.911258002122243
280 0.910122851530711
281 0.911258002122243
282 0.909855782985687
283 0.911925752957662
284 0.912727018197378
};
\addplot [line width=2.5pt, blue, opacity=1.0]
table {%
1 0.501201927661896
2 0.571714758872986
3 0.521233975887299
4 0.647035241127014
5 0.690304458141327
6 0.626402258872986
7 0.720152258872986
8 0.762820541858673
9 0.769831717014313
10 0.789863765239716
11 0.796875
12 0.785456717014313
13 0.821113765239716
14 0.828926265239716
15 0.837540090084076
16 0.821514427661896
17 0.832131385803223
18 0.84415066242218
19 0.847355782985687
20 0.849559307098389
21 0.848758041858673
22 0.844951927661896
23 0.852964758872986
24 0.856971144676208
25 0.844951927661896
26 0.860376596450806
27 0.861378192901611
28 0.858774065971375
29 0.878405451774597
30 0.870392620563507
31 0.877203524112701
32 0.864182710647583
33 0.877604186534882
34 0.871794879436493
35 0.873597741127014
36 0.853165090084076
37 0.877203524112701
38 0.89102566242218
39 0.882011234760284
40 0.875
41 0.876602590084076
42 0.876201927661896
43 0.889222741127014
44 0.880608975887299
45 0.877003192901611
46 0.869791686534882
47 0.891225934028625
48 0.882011234760284
49 0.879807710647583
50 0.886818885803223
51 0.888822138309479
52 0.886818885803223
53 0.88241183757782
54 0.887219548225403
55 0.892427861690521
56 0.88261216878891
57 0.888421475887299
58 0.899639427661896
59 0.886618614196777
60 0.881810903549194
61 0.89022433757782
62 0.899238765239716
63 0.89823716878891
64 0.895232379436493
65 0.901442289352417
66 0.908052861690521
67 0.894030451774597
68 0.897235572338104
69 0.902243614196777
70 0.900841355323792
71 0.902844548225403
72 0.900440692901611
73 0.900440692901611
74 0.900440692901611
75 0.90604966878891
76 0.905448734760284
77 0.901442289352417
78 0.899038434028625
79 0.900040090084076
80 0.90665066242218
81 0.896233975887299
82 0.903245210647583
83 0.90604966878891
84 0.908253192901611
85 0.908854186534882
86 0.909455120563507
87 0.903846144676208
88 0.907051265239716
89 0.900240361690521
90 0.911858975887299
91 0.90604966878891
92 0.907251596450806
93 0.913261234760284
94 0.909254789352417
95 0.911258041858673
96 0.907852590084076
97 0.909054458141327
98 0.910056114196777
99 0.905248403549194
100 0.904046475887299
101 0.905248403549194
102 0.904847741127014
103 0.905448734760284
104 0.90645033121109
105 0.909254789352417
106 0.902644217014313
107 0.905448734760284
108 0.911057710647583
109 0.90604966878891
110 0.910657048225403
111 0.908653855323792
112 0.90645033121109
113 0.901442289352417
114 0.902844548225403
115 0.914663434028625
116 0.904046475887299
117 0.904647409915924
118 0.907451927661896
119 0.908052861690521
120 0.908453524112701
121 0.909254789352417
122 0.902443885803223
123 0.908453524112701
124 0.900841355323792
125 0.895232379436493
126 0.89823716878891
127 0.907251596450806
128 0.889623403549194
129 0.895232379436493
130 0.899238765239716
131 0.899639427661896
132 0.896233975887299
133 0.89823716878891
134 0.904647409915924
135 0.90584933757782
136 0.896434307098389
137 0.901842951774597
138 0.901442289352417
139 0.896834909915924
140 0.897636234760284
141 0.900841355323792
142 0.902243614196777
143 0.914863765239716
144 0.911458313465118
145 0.912860572338104
146 0.915464758872986
147 0.91386216878891
148 0.911458313465118
149 0.915865361690521
150 0.914863765239716
151 0.916065692901611
152 0.914663434028625
153 0.913261234760284
154 0.919471144676208
155 0.916866958141327
156 0.915464758872986
157 0.917467951774597
158 0.91366183757782
159 0.915865361690521
160 0.911057710647583
161 0.916666686534882
162 0.917668282985687
163 0.912660241127014
164 0.912660241127014
165 0.918669879436493
166 0.913461565971375
167 0.916065692901611
168 0.909054458141327
169 0.919671475887299
170 0.915464758872986
171 0.915865361690521
172 0.915665090084076
173 0.916666686534882
174 0.915264427661896
175 0.907451927661896
176 0.915064096450806
177 0.912459909915924
178 0.915665090084076
179 0.912459909915924
180 0.91446316242218
181 0.90665066242218
182 0.911458313465118
183 0.9140625
184 0.910256385803223
185 0.909655451774597
186 0.911057710647583
187 0.909455120563507
188 0.912259638309479
189 0.91446316242218
190 0.909655451774597
191 0.912059307098389
192 0.912059307098389
193 0.91366183757782
194 0.910456717014313
195 0.912059307098389
196 0.907652258872986
197 0.909254789352417
198 0.903846144676208
199 0.90665066242218
200 0.90625
201 0.910256385803223
202 0.911458313465118
203 0.907852590084076
204 0.907451927661896
205 0.909254789352417
206 0.907652258872986
207 0.909455120563507
208 0.904447138309479
209 0.910857379436493
210 0.908653855323792
211 0.911258041858673
212 0.913060903549194
213 0.912459909915924
214 0.918068885803223
215 0.920272409915924
216 0.914663434028625
217 0.921073734760284
218 0.914863765239716
219 0.918469548225403
220 0.920072138309479
221 0.919070541858673
222 0.916666686534882
223 0.919070541858673
224 0.919471144676208
225 0.91446316242218
226 0.918068885803223
227 0.918669879436493
228 0.916065692901611
229 0.919871807098389
230 0.917467951774597
231 0.918669879436493
232 0.920272409915924
233 0.916666686534882
234 0.920673072338104
235 0.920072138309479
236 0.916866958141327
237 0.922676265239716
238 0.917668282985687
239 0.919270813465118
240 0.918669879436493
241 0.91426283121109
242 0.91426283121109
243 0.916866958141327
244 0.919671475887299
245 0.920873403549194
246 0.917668282985687
247 0.916065692901611
248 0.915264427661896
249 0.919871807098389
250 0.91386216878891
251 0.91366183757782
252 0.920472741127014
253 0.916466355323792
254 0.917467951774597
255 0.916065692901611
256 0.914663434028625
257 0.921274065971375
258 0.916666686534882
259 0.915264427661896
260 0.916866958141327
261 0.915264427661896
262 0.915464758872986
263 0.920272409915924
264 0.909655451774597
265 0.917267620563507
266 0.913261234760284
267 0.919671475887299
268 0.916266024112701
269 0.915064096450806
270 0.917067289352417
271 0.917668282985687
272 0.919270813465118
273 0.918469548225403
274 0.91426283121109
275 0.919070541858673
276 0.919471144676208
277 0.912660241127014
278 0.918068885803223
279 0.914863765239716
280 0.910056114196777
281 0.911258041858673
282 0.911057710647583
283 0.912860572338104
284 0.912459909915924
};
\end{axis}

\end{tikzpicture}

%% file: labpal/figure_data/performance_comparison_low_batch/plots_bs8/CIFAR-10_DenseNet-121_validation_accuracy2.pgf
% This file was created by tikzplotlib v0.9.8.
\begin{tikzpicture}

\definecolor{color0}{rgb}{0.933333333333333,0.509803921568627,0.933333333333333}
\definecolor{color1}{rgb}{0.647058823529412,0.164705882352941,0.164705882352941}
\definecolor{color2}{rgb}{1,0.647058823529412,0}

\begin{axis}[
minor xtick={},
minor ytick={},
tick align=outside,
tick pos=left,
width=10.5cm,height=8cm,grid=major,major grid style={dotted},
x grid style={white!69.0196078431373!black},
xlabel={epochs},
xmin=0,  xmax=298.15,
xtick style={color=black},
xtick={-50,0,50,100,150,200,250,300},
y grid style={white!69.0196078431373!black},
ylabel={validation accuracy},
ymin=0.85, ymax=0.94,
ytick style={color=black},
ytick={0.85,0.86,0.87,0.88,0.89,0.9,0.91,0.92,0.93}
]
\addplot [line width=2.5pt, color0, opacity=1.0]
table {%
1 0.30579999089241
2 0.351599991321564
3 0.463200002908707
4 0.499799996614456
5 0.514999985694885
6 0.554000020027161
7 0.560800015926361
8 0.579999983310699
9 0.609600007534027
10 0.613200008869171
11 0.585600018501282
12 0.631399989128113
13 0.630800008773804
14 0.657599985599518
15 0.629000008106232
16 0.69080001115799
17 0.693000018596649
18 0.698599994182587
19 0.717999994754791
20 0.70959997177124
21 0.7185999751091
22 0.717000007629395
23 0.737200021743774
24 0.756600022315979
25 0.732999980449677
26 0.736599981784821
27 0.752399981021881
28 0.763800024986267
29 0.773599982261658
30 0.776000022888184
31 0.769999980926514
32 0.761600017547607
33 0.7882000207901
34 0.742200016975403
35 0.765999972820282
36 0.785600006580353
37 0.763599991798401
38 0.785600006580353
39 0.77700001001358
40 0.771000027656555
41 0.77539998292923
42 0.792800009250641
43 0.786199986934662
44 0.750400006771088
45 0.806400001049042
46 0.779999971389771
47 0.823199987411499
48 0.804400026798248
49 0.799199998378754
50 0.786199986934662
51 0.819800019264221
52 0.783800005912781
53 0.793200016021729
54 0.801999986171722
55 0.771200001239777
56 0.79040002822876
57 0.80919998884201
58 0.810599982738495
59 0.777999997138977
60 0.787199974060059
61 0.832400023937225
62 0.835799992084503
63 0.817799985408783
64 0.821799993515015
65 0.840799987316132
66 0.83380001783371
67 0.843200027942657
68 0.829200029373169
69 0.838199973106384
70 0.844200015068054
71 0.805999994277954
};
\addplot [line width=2.5pt, color1, opacity=1.0]
table {%
1 0.0981999983390172
2 0.102066665887833
3 0.0967333341638247
4 0.100733334819476
5 0.0990666647752126
6 0.102199998994668
7 0.100066664318244
8 0.101799999674161
9 0.0998666658997536
10 0.0999999990065892
11 0.102666666110357
12 0.100466666122278
13 0.100466666122278
14 0.0964666654666265
15 0.102933332324028
16 0.0993333334724108
17 0.101199999451637
18 0.0986000001430511
19 0.108600000540415
20 0.101333332558473
21 0.103133330742518
22 0.101599998772144
23 0.10080000013113
24 0.0961333339413007
25 0.103600000341733
26 0.0977333337068558
27 0.103866666555405
28 0.0993999987840652
29 0.101333332558473
30 0.104266665875912
31 0.0989999994635582
32 0.102666666110357
33 0.104399998982747
34 0.0993333334724108
35 0.10266666362683
36 0.0993999987840652
37 0.0991333325703939
38 0.0973333319028219
39 0.0986000001430511
40 0.110466664036115
41 0.0953333328167597
42 0.103866664071878
43 0.100266665220261
44 0.0996666674812635
45 0.100733332335949
46 0.103133330742518
47 0.104066664973895
48 0.102733331422011
49 0.0977333337068558
50 0.100399998327096
51 0.103199998537699
52 0.0996000021696091
53 0.0961999992529551
54 0.0941333323717117
55 0.103399999439716
56 0.102133331199487
57 0.101266667246819
58 0.100666667024295
59 0.0991333325703939
60 0.0946666672825813
61 0.0993999987840652
62 0.0994000012675921
63 0.101933332780997
64 0.0986000001430511
65 0.0997999981045723
66 0.100333333015442
67 0.102533333003521
68 0.100133332113425
69 0.100199999908606
70 0.0996666649977366
71 0.10099999854962
72 0.098133330543836
73 0.100199999908606
74 0.0973333319028219
75 0.0995999996860822
76 0.112066666285197
77 0.103133330742518
78 0.100599999229113
79 0.10793333252271
80 0.101533330976963
81 0.100266665220261
82 0.0994666665792465
83 0.0994666665792465
84 0.102333332101504
85 0.094200000166893
86 0.100933330754439
87 0.108199998736382
88 0.0987333332498868
89 0.100733332335949
90 0.0969333325823148
91 0.101466665665309
92 0.101799997190634
93 0.100933330754439
94 0.100000001490116
95 0.097800001502037
96 0.100599999229113
97 0.0992000003655751
98 0.0997333327929179
99 0.0964000001549721
100 0.100733332335949
101 0.0961999992529551
102 0.100199997425079
103 0.103533332546552
104 0.102066665887833
105 0.106666666766008
106 0.100399998327096
107 0.10099999854962
108 0.0997333327929179
109 0.0979333346088727
110 0.103066665430864
111 0.0993333334724108
112 0.0992000003655751
113 0.102133333683014
114 0.101066663861275
115 0.10080000013113
116 0.0999999990065892
117 0.0980666652321815
118 0.102866664528847
119 0.102733333905538
120 0.100466666122278
121 0.102133333683014
122 0.0988000010450681
123 0.101866667469343
124 0.0997999981045723
125 0.0953333328167597
126 0.0992666656772296
127 0.104333331187566
128 0.101866664985816
129 0.102866667012374
130 0.101266664763292
131 0.100866665442785
132 0.108133333424727
133 0.105199997623761
134 0.102933332324028
135 0.0995333318909009
136 0.102399999896685
137 0.104599999884764
138 0.101400000353654
139 0.0995333318909009
140 0.0989999994635582
141 0.100133334596952
142 0.10099999854962
143 0.102266664306323
144 0.0993333334724108
145 0.0998666658997536
146 0.10173333187898
147 0.10266666362683
148 0.0957333321372668
149 0.100199997425079
150 0.0996666649977366
151 0.0885999997456868
152 0.111466663579146
153 0.101000001033147
154 0.0961333339413007
155 0.101799999674161
156 0.100333333015442
157 0.104333331187566
158 0.0997999981045723
159 0.113866664469242
160 0.102600000798702
161 0.0961333339413007
162 0.0983333339293798
163 0.100199997425079
164 0.103399999439716
165 0.103733330965042
166 0.0964666654666265
167 0.102133333683014
168 0.101333332558473
169 0.0988000010450681
170 0.102933332324028
171 0.0957333346207937
172 0.104266665875912
173 0.10099999854962
174 0.0993999987840652
175 0.1047333329916
};
\addplot [line width=2.5pt, red, opacity=1.0]
table {%
1 0.470399990677834
2 0.533300012350082
3 0.578600019216537
4 0.620099991559982
5 0.648299992084503
6 0.675200015306473
7 0.693200021982193
8 0.738299995660782
9 0.73879998922348
10 0.751899987459183
11 0.776799976825714
12 0.736600011587143
13 0.77240002155304
14 0.794699996709824
15 0.800799995660782
16 0.807500004768372
17 0.8158999979496
18 0.801899999380112
19 0.808799982070923
20 0.840700000524521
21 0.836199998855591
22 0.83840000629425
23 0.857600003480911
24 0.847100019454956
25 0.864199995994568
26 0.856099992990494
27 0.862499982118607
28 0.870800018310547
29 0.870799988508224
30 0.854900002479553
31 0.860199987888336
32 0.870800018310547
33 0.878200024366379
34 0.875500023365021
35 0.866699993610382
36 0.87389999628067
37 0.879599988460541
38 0.876399993896484
39 0.883700013160706
40 0.881600022315979
41 0.881300002336502
42 0.883500009775162
43 0.882699996232986
44 0.879400014877319
45 0.883199989795685
46 0.881099998950958
47 0.874200016260147
48 0.880299985408783
49 0.881999999284744
50 0.870799988508224
51 0.887799978256226
52 0.887100011110306
53 0.894700020551682
54 0.893500000238419
55 0.880899995565414
56 0.890100002288818
57 0.884100019931793
58 0.894199997186661
59 0.891400009393692
60 0.888300001621246
61 0.892599999904633
62 0.879200011491776
63 0.901500016450882
64 0.900499999523163
65 0.898999989032745
66 0.90090000629425
67 0.899100005626678
68 0.897599995136261
69 0.899399995803833
70 0.90090000629425
71 0.901600003242493
72 0.904699981212616
73 0.90149998664856
74 0.9060999751091
75 0.902900010347366
76 0.89750000834465
77 0.901800006628036
78 0.904400020837784
79 0.904799997806549
80 0.906499981880188
81 0.905600011348724
82 0.901699990034103
83 0.904500007629395
84 0.911200016736984
85 0.903400003910065
86 0.907999992370605
87 0.907499998807907
88 0.90939998626709
89 0.905400007963181
90 0.911100000143051
91 0.909299999475479
92 0.909899979829788
93 0.90719997882843
94 0.912699997425079
95 0.911399990320206
96 0.913700014352798
97 0.911900013685226
98 0.910699993371964
99 0.91389998793602
100 0.915899991989136
101 0.90829998254776
102 0.909600019454956
103 0.913699984550476
104 0.916299998760223
105 0.914700001478195
106 0.91770002245903
107 0.914900004863739
108 0.912999987602234
109 0.913599997758865
110 0.914899975061417
111 0.912400007247925
112 0.914800018072128
113 0.913800001144409
114 0.912299990653992
115 0.916900008916855
116 0.91609999537468
117 0.912699997425079
118 0.918599992990494
119 0.914600014686584
120 0.918699979782104
121 0.915100008249283
122 0.918899983167648
123 0.918799996376038
124 0.918000012636185
125 0.91839998960495
126 0.918500006198883
127 0.918099999427795
128 0.919400006532669
129 0.918899983167648
130 0.919099986553192
131 0.917499989271164
132 0.922499984502792
133 0.919699996709824
134 0.920100003480911
135 0.917999982833862
136 0.919900000095367
137 0.918499976396561
138 0.920899987220764
139 0.919200003147125
140 0.921199977397919
141 0.920499980449677
142 0.921000003814697
143 0.921400010585785
144 0.92059999704361
145 0.922800004482269
146 0.91949999332428
147 0.921600013971329
148 0.920400023460388
149 0.920700013637543
150 0.923799991607666
151 0.919300019741058
152 0.922799974679947
153 0.92069998383522
154 0.92179998755455
155 0.923699975013733
156 0.920000016689301
157 0.919299989938736
158 0.922399997711182
159 0.919699996709824
160 0.916700005531311
161 0.918900012969971
162 0.922000020742416
163 0.922600001096725
164 0.922800004482269
165 0.919799983501434
166 0.918300002813339
167 0.920800000429153
168 0.92220002412796
169 0.919999986886978
170 0.921399980783463
171 0.92059999704361
172 0.924199998378754
173 0.927299976348877
174 0.923500001430511
175 0.925200015306473
176 0.924199998378754
177 0.922299981117249
178 0.924699991941452
179 0.924499988555908
180 0.923299998044968
181 0.921000003814697
182 0.921499997377396
183 0.921399980783463
184 0.92399999499321
185 0.922199994325638
186 0.922800004482269
187 0.923700004816055
188 0.923799991607666
189 0.923299998044968
190 0.922399997711182
191 0.923799991607666
192 0.92560002207756
193 0.92519998550415
194 0.922199994325638
195 0.925200015306473
196 0.928300023078918
197 0.925799995660782
198 0.922399997711182
199 0.924299985170364
200 0.923500001430511
201 0.924100011587143
202 0.924699991941452
203 0.925300002098083
204 0.92629998922348
205 0.923600018024445
206 0.926100015640259
207 0.92629998922348
208 0.924199998378754
209 0.926399976015091
210 0.923899978399277
211 0.922499984502792
212 0.925299972295761
213 0.927900016307831
214 0.928299993276596
215 0.924499988555908
216 0.927300006151199
217 0.927100002765656
218 0.92560002207756
219 0.925300002098083
220 0.92509999871254
221 0.92620000243187
222 0.923900008201599
223 0.92509999871254
224 0.924400001764297
225 0.928000003099442
226 0.926499992609024
227 0.924400001764297
228 0.928000003099442
229 0.926499992609024
230 0.924700021743774
231 0.925799995660782
232 0.928300023078918
233 0.926899999380112
234 0.924899995326996
235 0.927700012922287
236 0.927699983119965
237 0.926599979400635
238 0.927300006151199
239 0.924899995326996
240 0.926599979400635
241 0.924699991941452
242 0.925700008869171
243 0.927300006151199
244 0.925699979066849
245 0.924199998378754
246 0.924899995326996
247 0.925900012254715
248 0.928299993276596
249 0.926799982786179
250 0.925899982452393
251 0.925799995660782
252 0.927700012922287
253 0.927100002765656
254 0.926499992609024
255 0.928799986839294
256 0.926400005817413
257 0.926400005817413
258 0.926799982786179
259 0.924899995326996
260 0.926600009202957
261 0.925799995660782
262 0.922399997711182
263 0.925300002098083
264 0.930399984121323
265 0.926800012588501
266 0.926899999380112
267 0.926300019025803
268 0.928100019693375
269 0.92620000243187
270 0.925799995660782
271 0.926599979400635
272 0.924699991941452
273 0.925000011920929
274 0.929900020360947
275 0.924899995326996
276 0.926799982786179
277 0.927100002765656
278 0.929499983787537
279 0.926099985837936
280 0.926699995994568
281 0.926699995994568
282 0.924899995326996
283 0.926699995994568
284 0.928500026464462
};
\addplot [line width=2.5pt, color2, opacity=1.0]
table {%
2 0.103799998760223
4 0.10226666678985
6 0.0992000003655751
8 0.100466666122278
10 0.100466668605804
12 0.101066666344802
14 0.101000001033147
16 0.0989333316683769
18 0.0975999981164932
20 0.0984666645526886
22 0.100266667703787
24 0.100666669507821
26 0.0993333334724108
28 0.100466666122278
30 0.100266665220261
32 0.10099999854962
34 0.0980666677157084
36 0.100533333917459
38 0.0964000001549721
40 0.0994666665792465
42 0.102933332324028
44 0.0986666679382324
46 0.100066664318244
48 0.094200000166893
50 0.104066664973895
52 0.102866667012374
54 0.0953333328167597
56 0.0972666665911674
58 0.100733332335949
60 0.100266667703787
62 0.100466666122278
64 0.0986666654547056
66 0.101600001255671
68 0.096600001056989
70 0.0968666672706604
72 0.0987999985615412
74 0.0968666672706604
76 0.101133331656456
78 0.102733333905538
80 0.0960666661461194
82 0.101399997870127
84 0.0961333339413007
86 0.101066666344802
88 0.0989999994635582
90 0.096600001056989
92 0.0961333339413007
94 0.0979999999205271
96 0.101000001033147
98 0.0984000017245611
100 0.0997333327929179
102 0.100933333237966
104 0.101133331656456
106 0.101266664763292
108 0.0995999996860822
110 0.101133331656456
112 0.0966666638851166
114 0.0965333332618078
116 0.100266665220261
118 0.103666665653388
120 0.098600002626578
122 0.100466666122278
124 0.101333332558473
126 0.0961999992529551
128 0.0998666658997536
130 0.103399999439716
132 0.0984666645526886
134 0.0991333350539207
136 0.0988666663567225
138 0.0972666665911674
140 0.0999999990065892
142 0.103866664071878
144 0.103866666555405
146 0.101400000353654
148 0.102933332324028
150 0.099333330988884
152 0.095400000611941
154 0.0981999983390172
156 0.0981333330273628
158 0.101666664083799
160 0.10246666520834
162 0.101066666344802
164 0.102533333003521
166 0.0981333330273628
168 0.0993333334724108
170 0.102266664306323
172 0.0999999990065892
174 0.0961999992529551
176 0.101066663861275
178 0.101599998772144
180 0.104266665875912
182 0.0994000012675921
184 0.101199999451637
186 0.0982666661341985
188 0.0989333316683769
190 0.100933333237966
192 0.0989333316683769
194 0.0997999981045723
196 0.0973333343863487
198 0.09740000218153
200 0.0980666652321815
202 0.0997333327929179
204 0.101266664763292
206 0.102866667012374
208 0.100399998327096
210 0.103866666555405
212 0.100266665220261
214 0.0997999981045723
216 0.0976666659116745
218 0.102133333683014
220 0.0980666652321815
222 0.100066666801771
224 0.101199999451637
226 0.100133334596952
228 0.0982000008225441
230 0.101933332780997
232 0.10099999854962
234 0.0963333323597908
236 0.0993999987840652
238 0.0991333325703939
240 0.101199999451637
242 0.101666666567326
244 0.102000000576178
246 0.0982666661341985
248 0.0974666650096575
250 0.100333333015442
252 0.0995333318909009
254 0.102399997413158
256 0.100466666122278
258 0.0973999996980031
260 0.102199998994668
262 0.0981333330273628
264 0.103466667234898
266 0.103133330742518
268 0.102399999896685
270 0.100666667024295
272 0.100333333015442
274 0.0990666647752126
276 0.0992666656772296
278 0.100599999229113
280 0.101333332558473
282 0.101400000353654
284 0.100333333015442
};
\addplot [line width=2.5pt, green!50.1960784313725!black, opacity=1.0]
table {%
1 0.364466667175293
2 0.489933321873347
3 0.567799985408783
4 0.59826668103536
5 0.619599997997284
6 0.63726665576299
7 0.628800014654795
8 0.689733326435089
9 0.660866677761078
10 0.638999988635381
11 0.697933336098989
12 0.719400008519491
13 0.705333332220713
14 0.736733337243398
15 0.740599989891052
16 0.771599988142649
17 0.78766667842865
18 0.750733315944672
19 0.77593332529068
20 0.796466668446859
21 0.712266663710276
22 0.783666670322418
23 0.809800008932749
24 0.790866653124491
25 0.785666664441427
26 0.799999992052714
27 0.777600010236104
28 0.795399983723958
29 0.795799990495046
30 0.816799998283386
31 0.818399985631307
32 0.817533334096273
33 0.829599996407827
34 0.831933339436849
35 0.806800007820129
36 0.842733323574066
37 0.849733332792918
38 0.828666667143504
39 0.834733347098033
40 0.825133343537649
41 0.846466680367788
42 0.751600007216136
43 0.796333332856496
44 0.801066676775614
45 0.803599993387858
46 0.823466658592224
47 0.824933330217997
48 0.659000019232432
49 0.707466671864192
50 0.741266667842865
51 0.732933322588603
52 0.779533326625824
53 0.789266665776571
54 0.79666668176651
55 0.767866671085358
56 0.798666656017303
57 0.816733340422312
58 0.79093333085378
59 0.815533339977264
60 0.803800006707509
61 0.825000007947286
62 0.798199991385142
63 0.827133317788442
64 0.830666661262512
65 0.837333341439565
66 0.831999977429708
67 0.828266680240631
68 0.848266661167145
69 0.847333331902822
70 0.841733336448669
71 0.844400008519491
72 0.856266657511393
73 0.8392666776975
74 0.854200005531311
75 0.853133320808411
76 0.845933338006338
77 0.85073333978653
78 0.853866656621297
79 0.844866673151652
80 0.851333340009054
81 0.84253333012263
82 0.842399994532267
83 0.861733317375183
84 0.842133323351542
85 0.856666664282481
86 0.84553333123525
87 0.846599996089935
88 0.854666670163473
89 0.838133335113525
90 0.851533313592275
91 0.857999980449677
92 0.856333335240682
93 0.859800000985463
94 0.846200009187063
95 0.852933327356974
96 0.84253333012263
97 0.831533332665761
98 0.860733330249786
99 0.855999986330668
100 0.863333324591319
101 0.856333335240682
102 0.86053333679835
103 0.864466647307078
104 0.870266656080882
105 0.863333344459534
106 0.859666645526886
107 0.856666684150696
108 0.871533354123433
109 0.867266654968262
110 0.874466677506765
111 0.864866673946381
112 0.861866652965546
113 0.85233332713445
114 0.86520000298818
115 0.875333348910014
116 0.870333313941956
117 0.857866684595744
118 0.872933328151703
119 0.878866652647654
120 0.875333329041799
121 0.870133340358734
122 0.86353333791097
123 0.870733320713043
124 0.867800017197927
125 0.877400000890096
126 0.864600002765656
127 0.878399988015493
128 0.876866678396861
129 0.868533333142599
130 0.875666658083598
131 0.860466678937276
132 0.87199999888738
133 0.874466677506765
134 0.869399984677633
135 0.86573334534963
136 0.865066667397817
137 0.874666670958201
138 0.868399997552236
139 0.869199991226196
140 0.874066670735677
141 0.867866675059001
142 0.862866659959157
143 0.881933331489563
144 0.878866672515869
145 0.882399996121724
146 0.888800005118052
147 0.88700000445048
148 0.886533339818319
149 0.876199980576833
150 0.889933327833811
151 0.885266661643982
152 0.886799991130829
153 0.892866671085358
154 0.899133324623108
155 0.889799992243449
156 0.896733323733012
157 0.892933328946432
158 0.888533333937327
159 0.884266674518585
160 0.895400007565816
161 0.882933338483175
162 0.890199999014537
163 0.895333329836527
164 0.892199993133545
165 0.896333336830139
166 0.898333330949148
167 0.896399994691213
168 0.89053334792455
169 0.895066658655802
170 0.884866654872894
171 0.895799994468689
172 0.88453334569931
173 0.895266671975454
174 0.88866666952769
175 0.896599988142649
176 0.896399994691213
177 0.897800008455912
178 0.893533329168955
179 0.888800005118052
180 0.892066657543182
181 0.89466667175293
182 0.890333334604899
183 0.89766667286555
184 0.898666659990946
185 0.899399995803833
186 0.899133324623108
187 0.896066685517629
188 0.896200001239777
189 0.889799992243449
190 0.898999989032745
191 0.896399994691213
192 0.896399994691213
193 0.891266663869222
194 0.89300000667572
195 0.898866673310598
196 0.902266681194305
197 0.897866666316986
198 0.898600002129873
199 0.898533324400584
200 0.897533337275187
201 0.898799975713094
202 0.893466651439667
203 0.900866667429606
204 0.899599989255269
205 0.903933346271515
206 0.896999994913737
207 0.900866667429606
208 0.900600016117096
209 0.896600008010864
210 0.896733343601227
211 0.893599987030029
212 0.897866666316986
213 0.891800006230672
214 0.892333348592122
215 0.901066660881042
216 0.889533340930939
217 0.898400008678436
218 0.901199996471405
219 0.903200010458628
220 0.890666683514913
221 0.894466678301493
222 0.903733332951864
223 0.903800010681152
224 0.892800013224284
225 0.894600013891856
226 0.898533344268799
227 0.898200015227
228 0.899600009123484
229 0.892800013224284
230 0.896733323733012
231 0.895533343156179
232 0.896200021107992
233 0.901733338832855
234 0.900266667207082
235 0.898400008678436
236 0.894733349482218
237 0.902933339277903
238 0.893466651439667
239 0.90146666765213
240 0.902133325735728
241 0.896933337052663
242 0.899399995803833
243 0.897266646226247
244 0.902133325735728
245 0.89520001411438
246 0.900533318519592
247 0.90013333161672
248 0.90119997660319
249 0.89766667286555
250 0.902399996916453
251 0.895600020885468
252 0.896066665649414
253 0.901266674200694
254 0.898133317629496
255 0.901599983374278
256 0.896199981371562
257 0.898533324400584
258 0.895599981149038
259 0.89546666542689
260 0.898600002129873
261 0.895199994246165
262 0.900733331839244
263 0.898933331171672
264 0.897666652997335
265 0.892133335272471
266 0.896133343378703
267 0.896466672420502
268 0.895066658655802
269 0.898533324400584
270 0.901266674200694
271 0.898733337720235
272 0.891933341821035
273 0.887600004673004
274 0.900733331839244
275 0.894733309745789
276 0.893933316071828
277 0.903066674868266
278 0.899399995803833
279 0.900333325068156
280 0.90146666765213
281 0.897266666094462
282 0.902066667874654
283 0.899933338165283
284 0.898866673310598
};
\addplot [line width=2.5pt, black, opacity=1.0]
table {%
1 0.439266661802928
2 0.568400005499522
3 0.635066668192546
4 0.655066668987274
5 0.679533322652181
6 0.705933332443237
7 0.705666681130727
8 0.734399994214376
9 0.749800006548564
10 0.762199997901917
11 0.775800009568532
12 0.762333313624064
13 0.788133323192596
14 0.816266675790151
15 0.809066673119863
16 0.836066683133443
17 0.837799986203512
18 0.8392666776975
19 0.844066679477692
20 0.853600005308787
21 0.848933339118958
22 0.862866679827372
23 0.86133333047231
24 0.864600002765656
25 0.871333340803782
26 0.872133334477743
27 0.875466664632161
28 0.875133335590363
29 0.878199994564056
30 0.877666672070821
31 0.887200017770131
32 0.881800015767415
33 0.88266666730245
34 0.887866655985514
35 0.888800005118052
36 0.892066677411397
37 0.890599985917409
38 0.891200006008148
39 0.898466686407725
40 0.895400007565816
41 0.89926666021347
42 0.895266652107239
43 0.895999987920125
44 0.899666666984558
45 0.902400016784668
46 0.895533343156179
47 0.900599996248881
48 0.900266667207082
49 0.899866660435995
50 0.898600002129873
51 0.902066667874654
52 0.903533339500427
53 0.904333313306173
54 0.903866648674011
55 0.904666682084401
56 0.904933333396912
57 0.903733332951864
58 0.906199991703033
59 0.904733339945475
60 0.9050000111262
61 0.904933333396912
62 0.901199996471405
63 0.902866661548615
64 0.904599984486898
65 0.905733327070872
66 0.904333333174388
67 0.904600004355113
68 0.908333341280619
69 0.903933346271515
70 0.903200010458628
71 0.905799984931946
72 0.905533333619436
73 0.904866655667623
74 0.903066674868266
75 0.905399998029073
76 0.905999998251597
77 0.906533320744832
78 0.90366667509079
79 0.90173331896464
80 0.905533333619436
81 0.904399991035461
82 0.905066668987274
83 0.904999991257985
84 0.904133339722951
85 0.903799990812937
86 0.901866674423218
87 0.9050000111262
88 0.904866655667623
89 0.904733339945475
90 0.903533319632212
91 0.902200003465017
92 0.904333333174388
93 0.905466675758362
94 0.907199998696645
95 0.904000004132589
96 0.90446666876475
97 0.907133340835571
98 0.907000005245209
99 0.905266662438711
100 0.906400005022685
101 0.905133346716563
102 0.903333346048991
103 0.902599990367889
104 0.907266656557719
105 0.90366667509079
106 0.903799990812937
107 0.906266669432322
108 0.90420001745224
109 0.906599998474121
110 0.90666667620341
111 0.904400010903676
112 0.903733332951864
113 0.903666655222575
114 0.905066668987274
115 0.905933340390523
116 0.907533327738444
117 0.902733325958252
118 0.903866668542226
119 0.905199984709422
120 0.902800003687541
121 0.905999998251597
122 0.901933312416077
123 0.906733334064484
124 0.90666667620341
125 0.904866675535838
126 0.902933339277903
127 0.903866668542226
128 0.906066675980886
129 0.904399991035461
130 0.904866655667623
131 0.906066675980886
132 0.905266662438711
133 0.905266662438711
134 0.90613333384196
135 0.904266675313314
136 0.906199991703033
137 0.907133340835571
138 0.907799998919169
139 0.902933339277903
140 0.905466675758362
141 0.903866648674011
142 0.905066668987274
143 0.906466643015544
144 0.902733325958252
145 0.905799984931946
146 0.904933333396912
147 0.906466662883759
148 0.902599990367889
149 0.902999997138977
150 0.905933340390523
151 0.9039333264033
152 0.907000005245209
153 0.90066667397817
154 0.902800003687541
155 0.903199990590413
156 0.905466675758362
157 0.902066667874654
158 0.906733334064484
159 0.905066668987274
160 0.90146666765213
161 0.903733313083649
162 0.90446666876475
163 0.903800010681152
164 0.903266668319702
165 0.903799990812937
166 0.903066655000051
167 0.907533327738444
168 0.903866688410441
169 0.904266675313314
170 0.903933346271515
171 0.902399996916453
172 0.905999998251597
173 0.905333340167999
174 0.905200004577637
175 0.906533340613047
176 0.905533333619436
177 0.905333340167999
178 0.907066663106283
179 0.903733352820078
180 0.905599991480509
181 0.905999998251597
182 0.906799991925557
183 0.905799984931946
184 0.901466687520345
185 0.903266668319702
186 0.905466675758362
187 0.906199991703033
188 0.907066663106283
189 0.904533326625824
190 0.903666655222575
191 0.905600011348724
192 0.904466648896535
193 0.902066667874654
194 0.905933320522308
195 0.904533326625824
196 0.90613333384196
197 0.908866663773855
198 0.905599991480509
199 0.903799990812937
200 0.901933352152506
201 0.903133352597555
202 0.903933346271515
203 0.906866669654846
204 0.907266676425934
205 0.906333327293396
206 0.907066663106283
207 0.904266655445099
208 0.9039333264033
209 0.901933352152506
210 0.904133339722951
211 0.905066668987274
212 0.905666649341583
213 0.905200004577637
214 0.905333340167999
215 0.904000004132589
216 0.904133339722951
217 0.904199997584025
218 0.906199991703033
219 0.90200001001358
220 0.903799990812937
221 0.905533313751221
222 0.904533326625824
223 0.902133325735728
224 0.905200004577637
225 0.904733339945475
226 0.901600003242493
227 0.903733332951864
228 0.904733339945475
229 0.904333333174388
230 0.906733334064484
231 0.904000004132589
232 0.905933340390523
233 0.904399991035461
234 0.90773332118988
235 0.901400009791056
236 0.907066663106283
237 0.905200004577637
238 0.904799997806549
239 0.905733327070872
240 0.906200011571248
241 0.904533346494039
242 0.906333347161611
243 0.906666656335195
244 0.90446666876475
245 0.903199990590413
246 0.904066661993662
247 0.905333320299784
248 0.901733338832855
249 0.905733327070872
250 0.906800011793772
251 0.903799990812937
252 0.905200004577637
253 0.905399998029073
254 0.90613333384196
255 0.904599984486898
256 0.903866668542226
257 0.904999991257985
258 0.906733314196269
259 0.905266662438711
260 0.907066663106283
261 0.900333325068156
262 0.907199998696645
263 0.903333326180776
264 0.903800010681152
265 0.904666682084401
266 0.903533339500427
267 0.907933334509532
268 0.903599997361501
269 0.906866669654846
270 0.904799997806549
271 0.905466655890147
272 0.906333347161611
273 0.905466655890147
274 0.905466675758362
275 0.906333327293396
276 0.903799990812937
277 0.903533339500427
278 0.906333347161611
279 0.904199997584025
280 0.903666655222575
281 0.904999991257985
282 0.901799996693929
283 0.905666669209798
284 0.905133326848348
};
\addplot [line width=2.5pt, blue, opacity=1.0]
table {%
1 0.451466659704844
2 0.535333335399628
3 0.559800008932749
4 0.619199991226196
5 0.642666677633921
6 0.67933334906896
7 0.689466675122579
8 0.73333332935969
9 0.72846664985021
10 0.763800005118052
11 0.773533324400584
12 0.752933343251546
13 0.787999987602234
14 0.804266671339671
15 0.812599996725718
16 0.810466667016347
17 0.823466658592224
18 0.823133329550425
19 0.824266672134399
20 0.838533341884613
21 0.826533317565918
22 0.84086666504542
23 0.844533324241638
24 0.838133335113525
25 0.846933325131734
26 0.840266684691111
27 0.854800005753835
28 0.860466678937276
29 0.848800003528595
30 0.85646665096283
31 0.855800012747447
32 0.859733323256175
33 0.860933323701223
34 0.867933332920074
35 0.872999986012777
36 0.870066662629445
37 0.872466663519541
38 0.874466677506765
39 0.876399993896484
40 0.8739333152771
41 0.876866658528646
42 0.877933343251546
43 0.875466664632161
44 0.879599988460541
45 0.88266666730245
46 0.879266659418742
47 0.874399999777476
48 0.879799981911977
49 0.882399996121724
50 0.885133345921834
51 0.885599990685781
52 0.880400002002716
53 0.881999989350637
54 0.887133340040843
55 0.878866672515869
56 0.884733339150747
57 0.884400010108948
58 0.890533328056335
59 0.888133327166239
60 0.88346666097641
61 0.891266663869222
62 0.889000018437703
63 0.892333328723907
64 0.887333333492279
65 0.888800005118052
66 0.899066666762034
67 0.891266663869222
68 0.895800014336904
69 0.894800007343292
70 0.894866685072581
71 0.898333330949148
72 0.89053334792455
73 0.890733341375987
74 0.901199996471405
75 0.899866660435995
76 0.899533331394196
77 0.895333329836527
78 0.897933344046275
79 0.900466660658518
80 0.900799989700317
81 0.900866667429606
82 0.905533333619436
83 0.903200010458628
84 0.902666668097178
85 0.904666662216187
86 0.900733351707458
87 0.903799990812937
88 0.903466661771139
89 0.900733351707458
90 0.902866661548615
91 0.900800009568532
92 0.904133319854736
93 0.906266669432322
94 0.909733335177104
95 0.906200011571248
96 0.909533321857452
97 0.911133329073588
98 0.908733328183492
99 0.906600018342336
100 0.908466657002767
101 0.907600005467733
102 0.907933334509532
103 0.910599986712138
104 0.909133354822795
105 0.910999993483225
106 0.909800012906392
107 0.9121333360672
108 0.91293332974116
109 0.912266671657562
110 0.911266664663951
111 0.910999993483225
112 0.912999987602234
113 0.911266684532166
114 0.912799994150797
115 0.912733316421509
116 0.915333350499471
117 0.910333355267843
118 0.913999994595846
119 0.91539998849233
120 0.914133330186208
121 0.910666664441427
122 0.913399994373322
123 0.913066645463308
124 0.912000020345052
125 0.913533329963684
126 0.910266677538554
127 0.913599987824758
128 0.914333323637644
129 0.913933336734772
130 0.915533324082693
131 0.917199989159902
132 0.91619998216629
133 0.916133344173431
134 0.913999994595846
135 0.915800015131632
136 0.917066673437754
137 0.916333337624868
138 0.915600001811981
139 0.919933319091797
140 0.916600008805593
141 0.912999987602234
142 0.913333336512248
143 0.919399996598562
144 0.918866654237111
145 0.921666661898295
146 0.920533339182536
147 0.920066674550374
148 0.917400002479553
149 0.918733338514964
150 0.918266673882802
151 0.921266655127207
152 0.920133352279663
153 0.921266655127207
154 0.920200010140737
155 0.917199989159902
156 0.920333325862885
157 0.91759999593099
158 0.920933345953623
159 0.915999988714854
160 0.919533332188924
161 0.915066679318746
162 0.920466661453247
163 0.918333331743876
164 0.917400002479553
165 0.917266666889191
166 0.91646667321523
167 0.917733331521352
168 0.918999989827474
169 0.917333324750265
170 0.918666660785675
171 0.917866667111715
172 0.914733350276947
173 0.914266645908356
174 0.916399995485942
175 0.915799995263418
176 0.919399996598562
177 0.915600001811981
178 0.915666659673055
179 0.917199989159902
180 0.919466654459635
181 0.915666659673055
182 0.918799976507823
183 0.917066653569539
184 0.916999995708466
185 0.917866667111715
186 0.919399976730347
187 0.918400009473165
188 0.920866668224335
189 0.915800015131632
190 0.915600001811981
191 0.915400008360545
192 0.918266673882802
193 0.917400002479553
194 0.919066667556763
195 0.917000015576681
196 0.918866674105326
197 0.916866679986318
198 0.916266659895579
199 0.916666666666667
200 0.915666659673055
201 0.91593333085378
202 0.917466660340627
203 0.913599987824758
204 0.912933349609375
205 0.915199995040894
206 0.911333342393239
207 0.904533326625824
208 0.906999985376994
209 0.911800007025401
210 0.910399993260702
211 0.909999986489614
212 0.907866676648458
213 0.912466684977214
214 0.914533336957296
215 0.91813333829244
216 0.918466667334239
217 0.919666667779287
218 0.919999996821086
219 0.922733326752981
220 0.919800003369649
221 0.920266648133596
222 0.921333332856496
223 0.919599990049998
224 0.920999983946482
225 0.917133331298828
226 0.921333332856496
227 0.920866668224335
228 0.922066648801168
229 0.920333345731099
230 0.921000003814697
231 0.921199997266134
232 0.920133332411448
233 0.922333339850108
234 0.919200003147125
235 0.921466668446859
236 0.920600016911825
237 0.921400010585785
238 0.922733326752981
239 0.922000010808309
240 0.920866668224335
241 0.920666674772898
242 0.92113333940506
243 0.923733333746592
244 0.920066654682159
245 0.920066674550374
246 0.922333339850108
247 0.92166668176651
248 0.921533346176147
249 0.920733332633972
250 0.922066668669383
251 0.920200010140737
252 0.9189333319664
253 0.920200010140737
254 0.92086664835612
255 0.919600009918213
256 0.922666668891907
257 0.922133326530457
258 0.919800003369649
259 0.922599991162618
260 0.921199997266134
261 0.917933344841003
262 0.921600004037221
263 0.922399997711182
264 0.923466662565867
265 0.921933313210805
266 0.923333326975505
267 0.921866655349731
268 0.920133332411448
269 0.920600016911825
270 0.922533333301544
271 0.920866668224335
272 0.923066675662994
273 0.920266668001811
274 0.920266668001811
275 0.921200017134349
276 0.921266674995422
277 0.919999996821086
278 0.920733312765757
279 0.920266668001811
280 0.919599990049998
281 0.918466667334239
282 0.9189333319664
283 0.919399996598562
284 0.921466668446859
};
\end{axis}

\end{tikzpicture}

%% file: labpal/figure_sensitivity_analysis_labpalsgd.tex
\begin{figure}[h!]
	\vspace{-1.0cm}
	\centering
	\def\scale{0.28}
	\begin{tabular}{ c c c}	
		\textbf{ResNet-20}&\textbf{MobileNet-V2} &\textbf{DenseNet-121}\\
		\scalebox{\scale}{\includegraphics{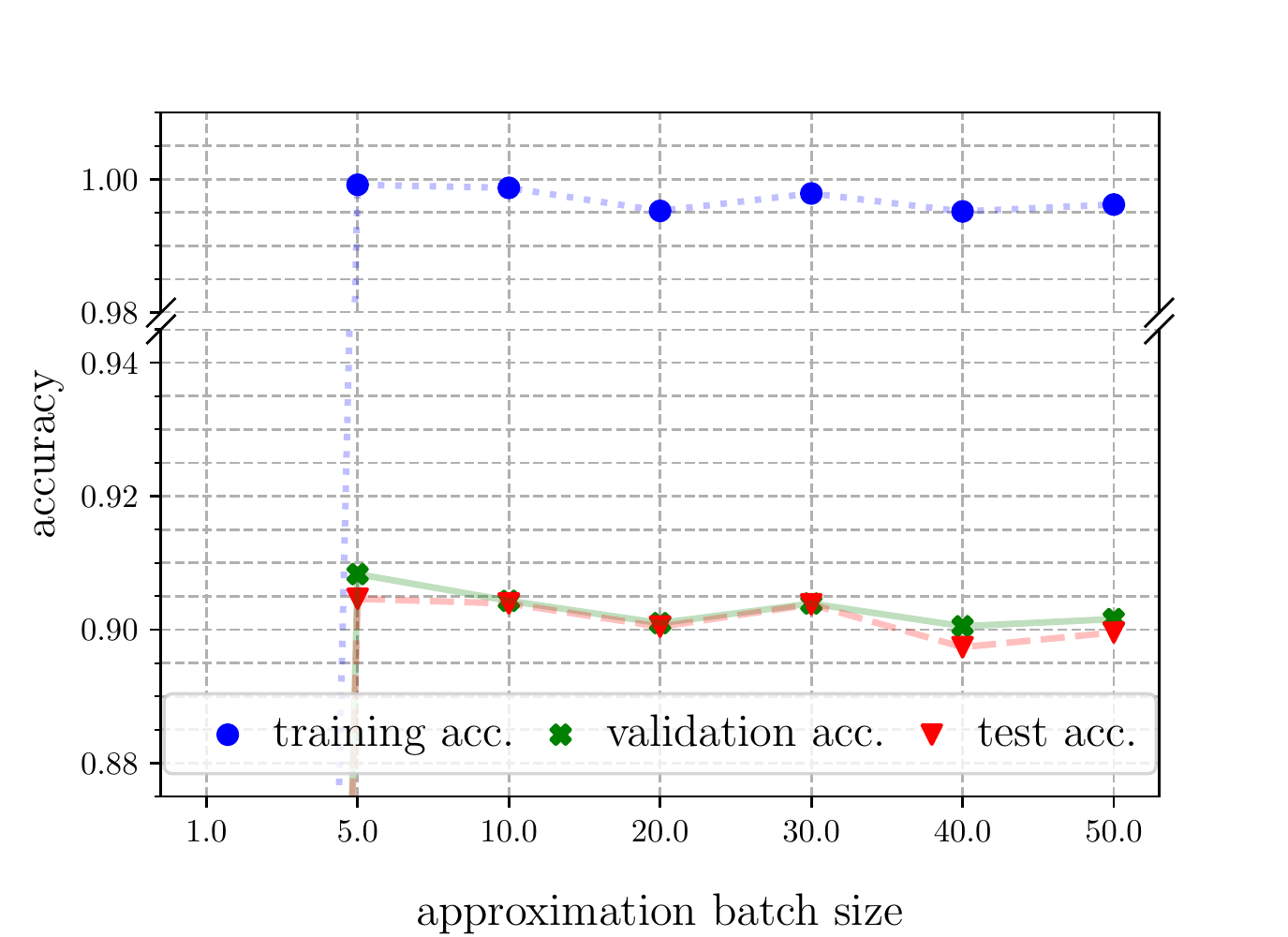}}&
		\scalebox{\scale}{\includegraphics{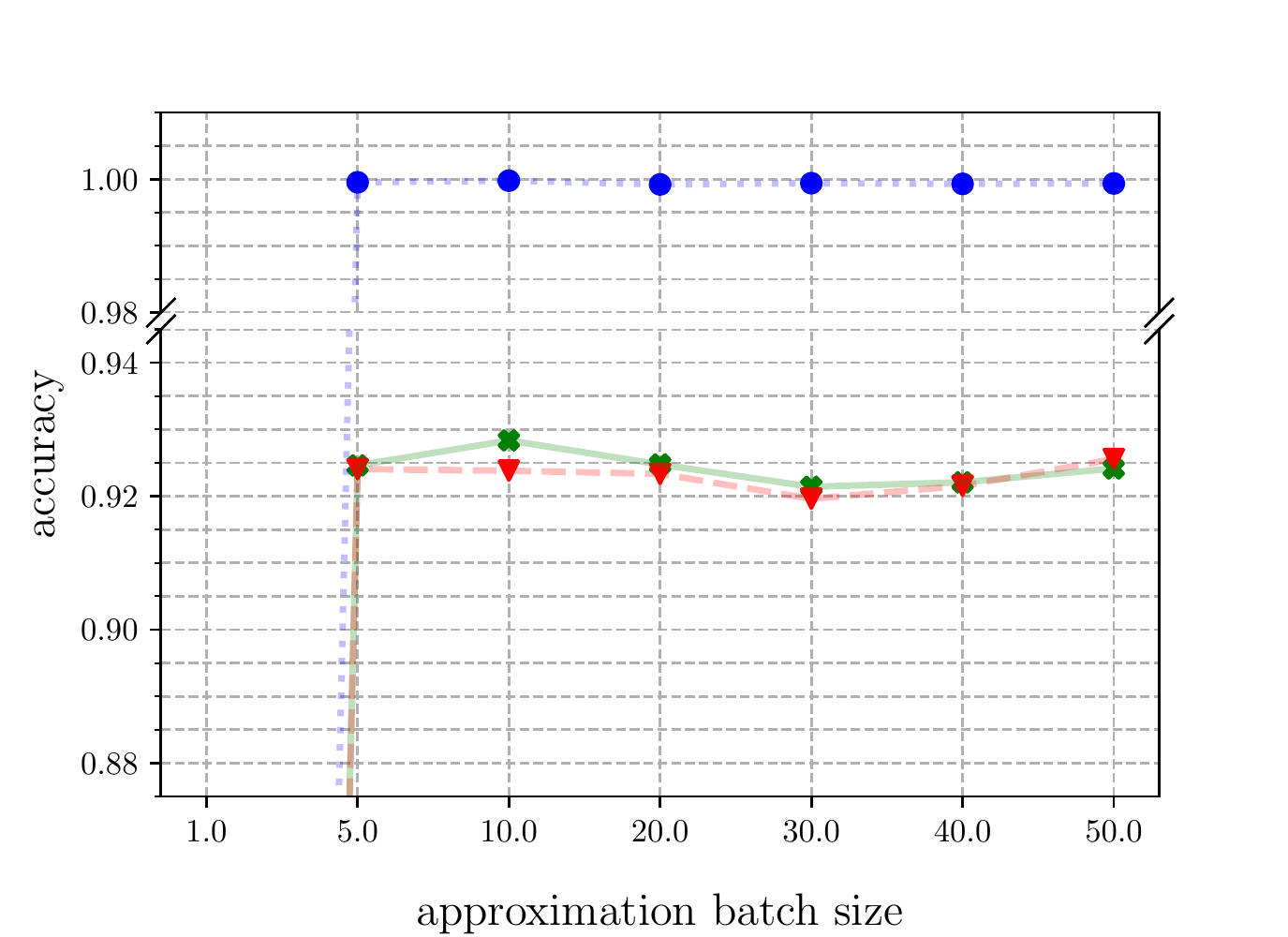}}&
		\scalebox{\scale}{\includegraphics{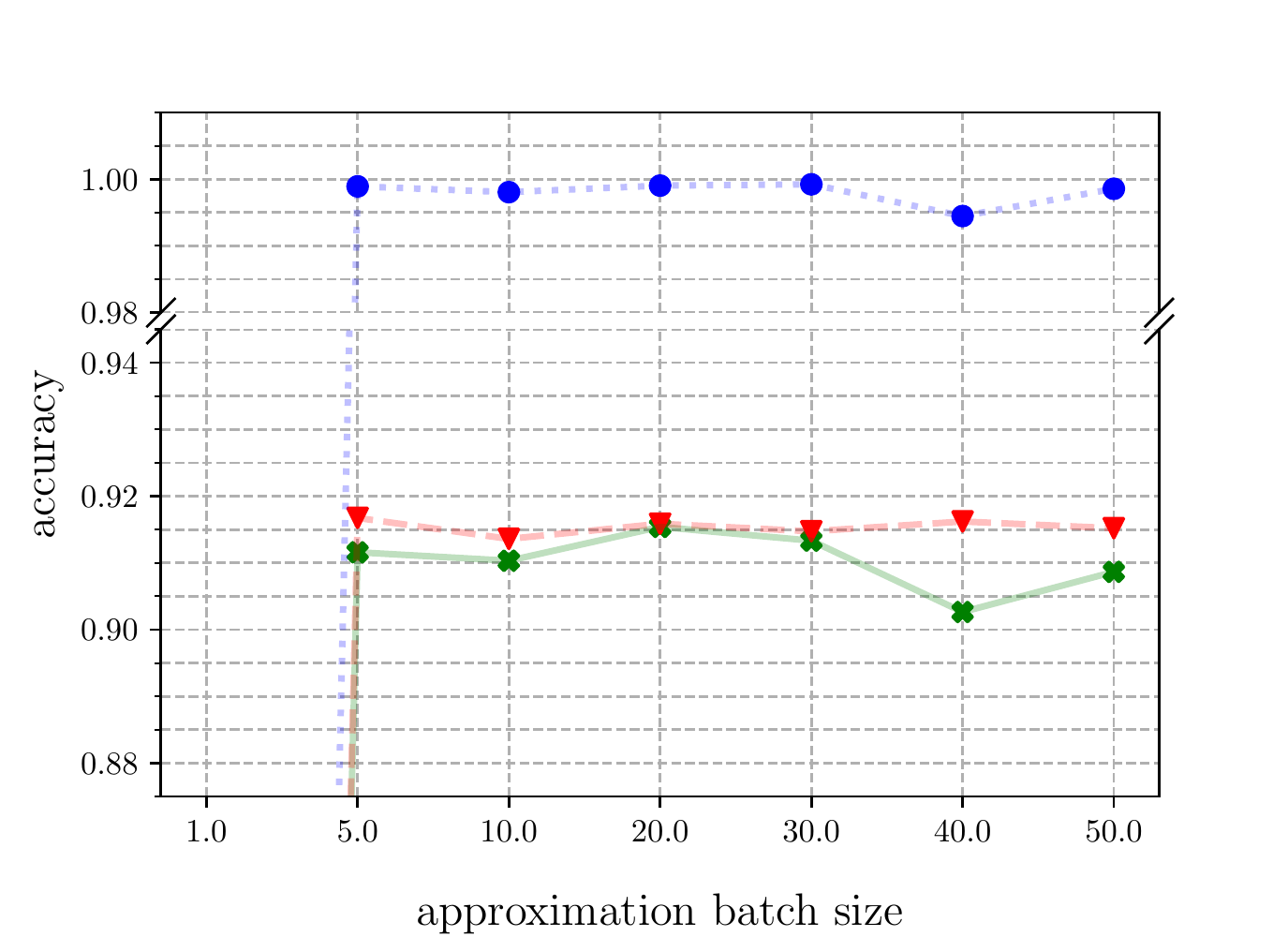}}\\
		\scalebox{\scale}{\includegraphics{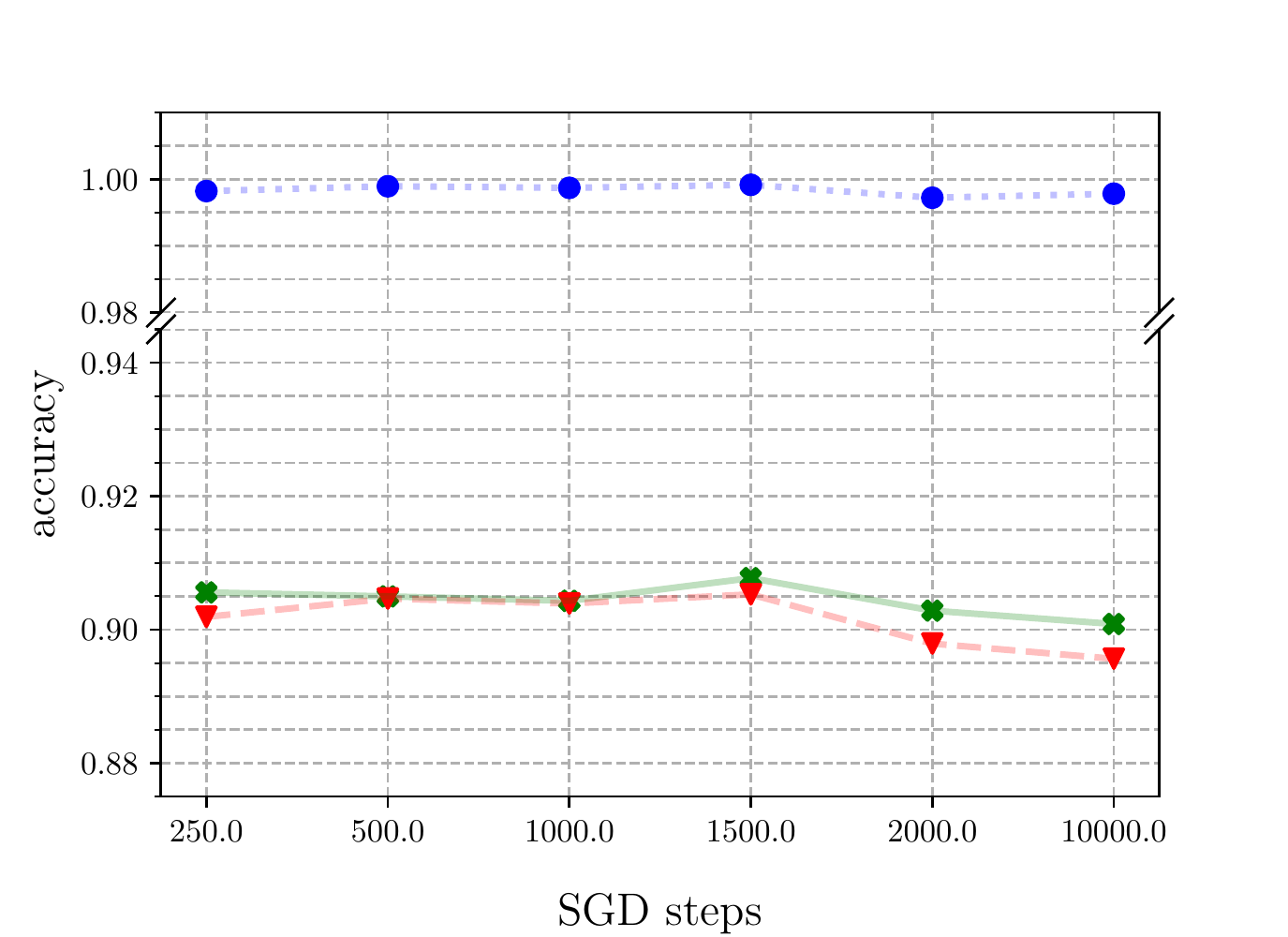}}&
		\scalebox{\scale}{\includegraphics{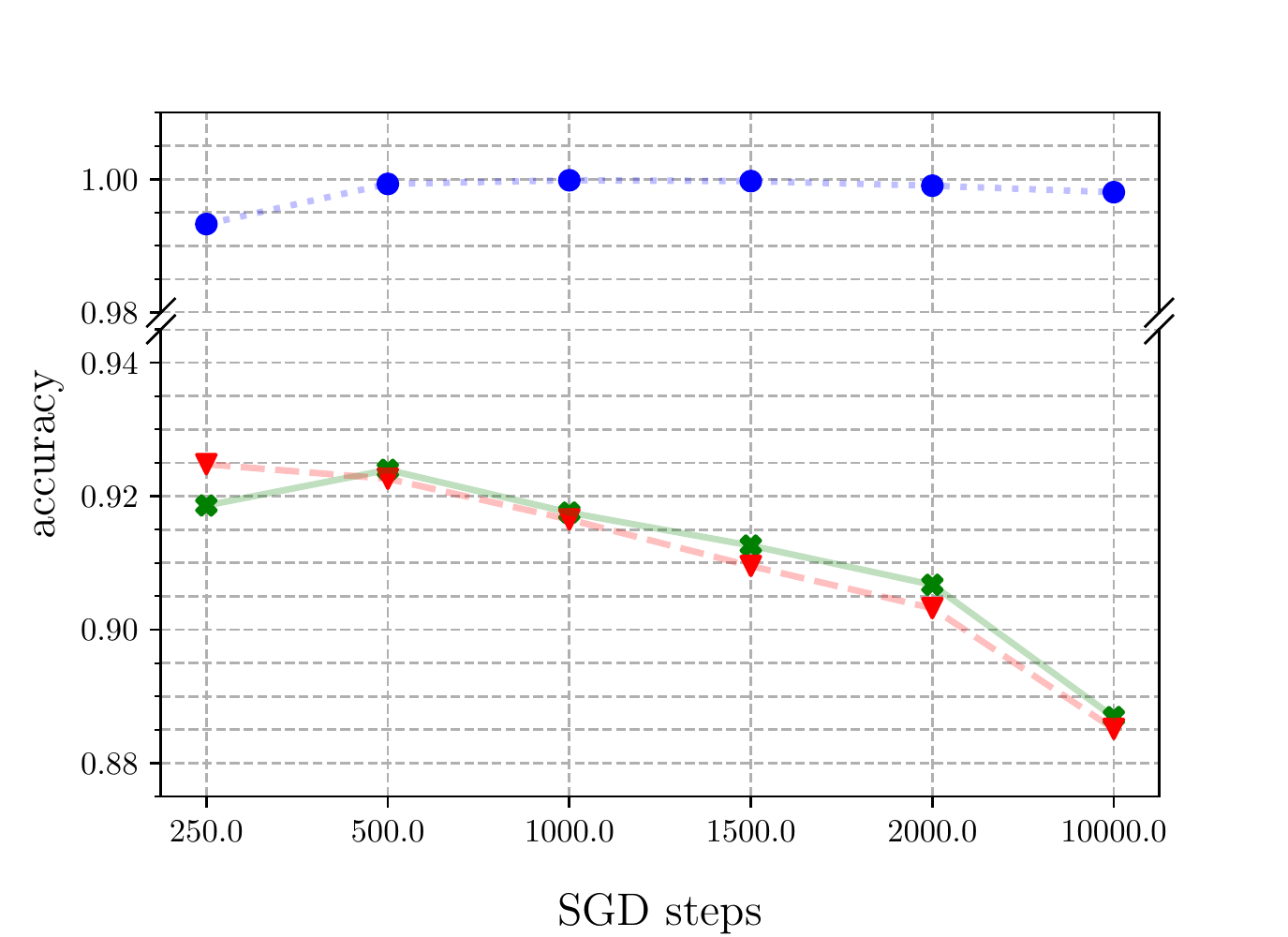}}&
		\scalebox{\scale}{\includegraphics{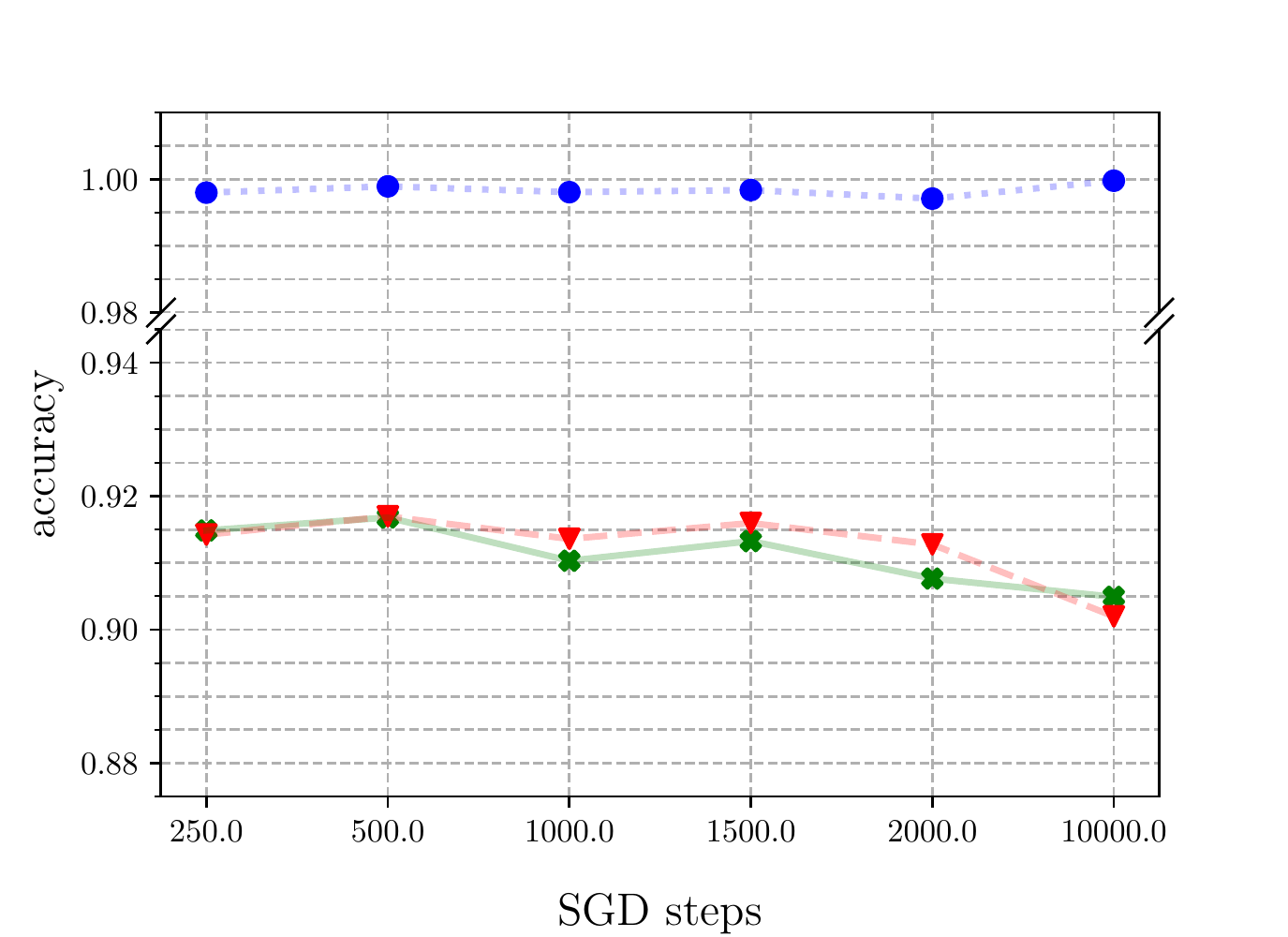}}\\
		\scalebox{\scale}{\includegraphics{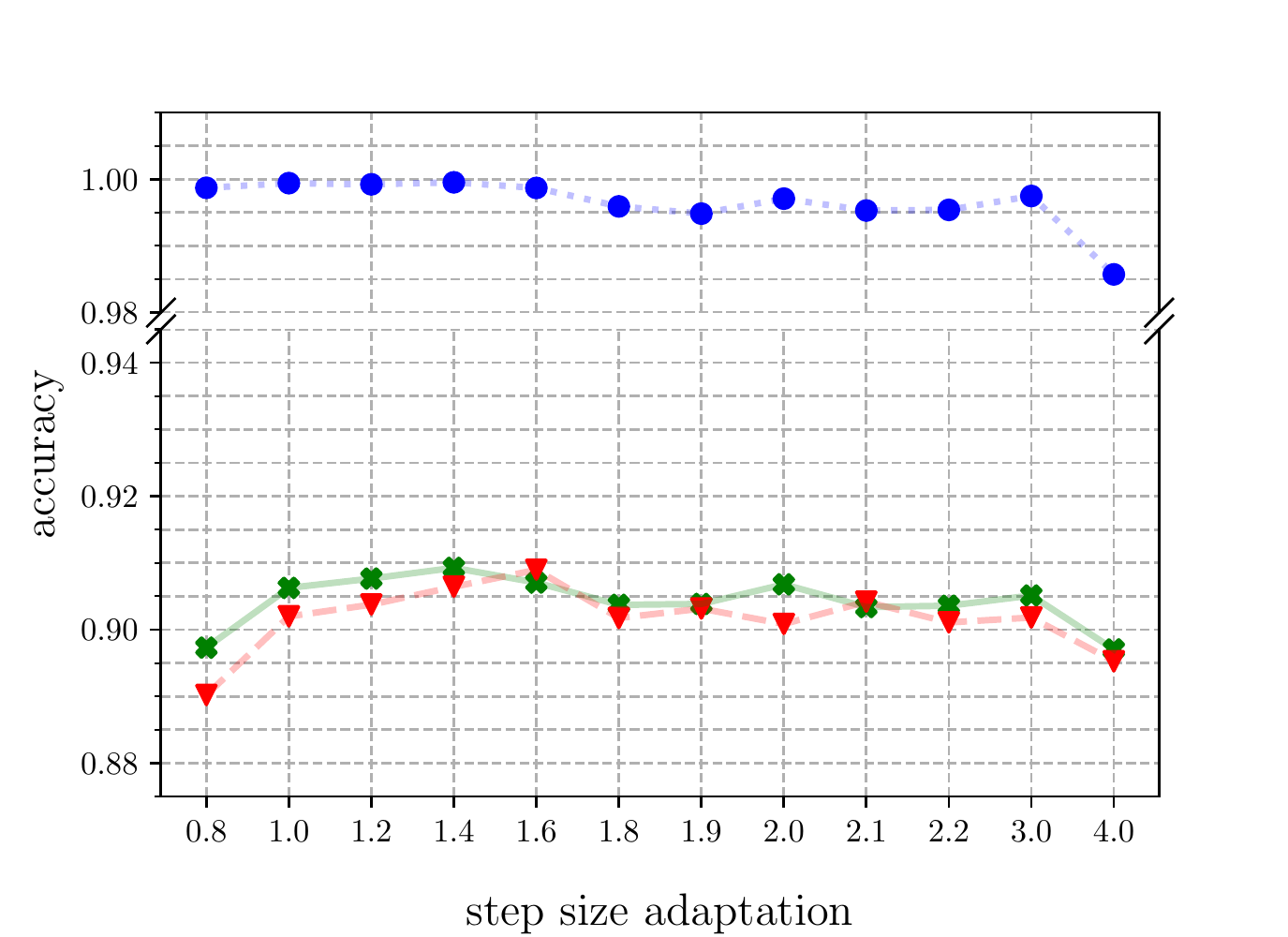}}&
		\scalebox{\scale}{\includegraphics{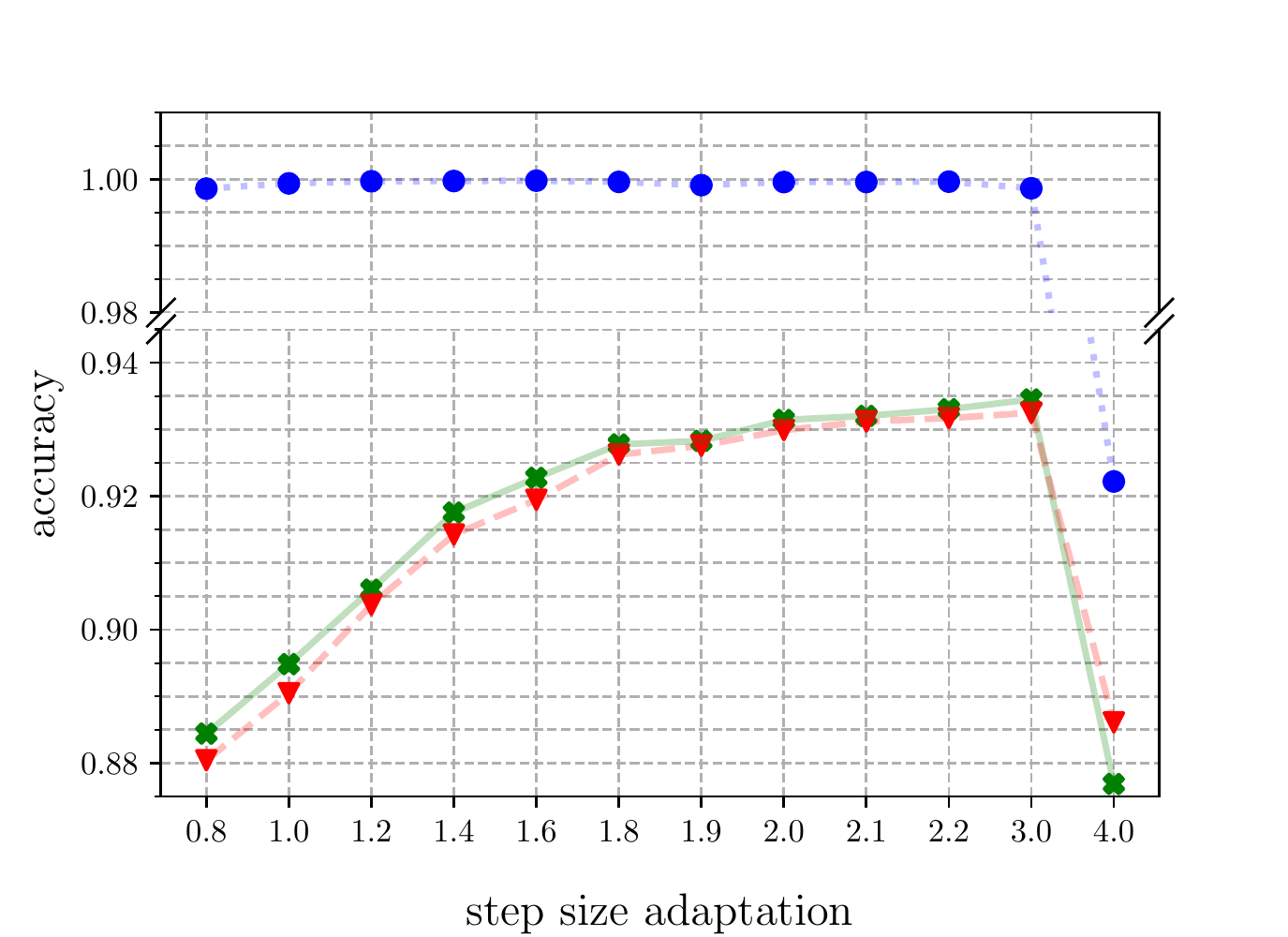}}&
		\scalebox{\scale}{\includegraphics{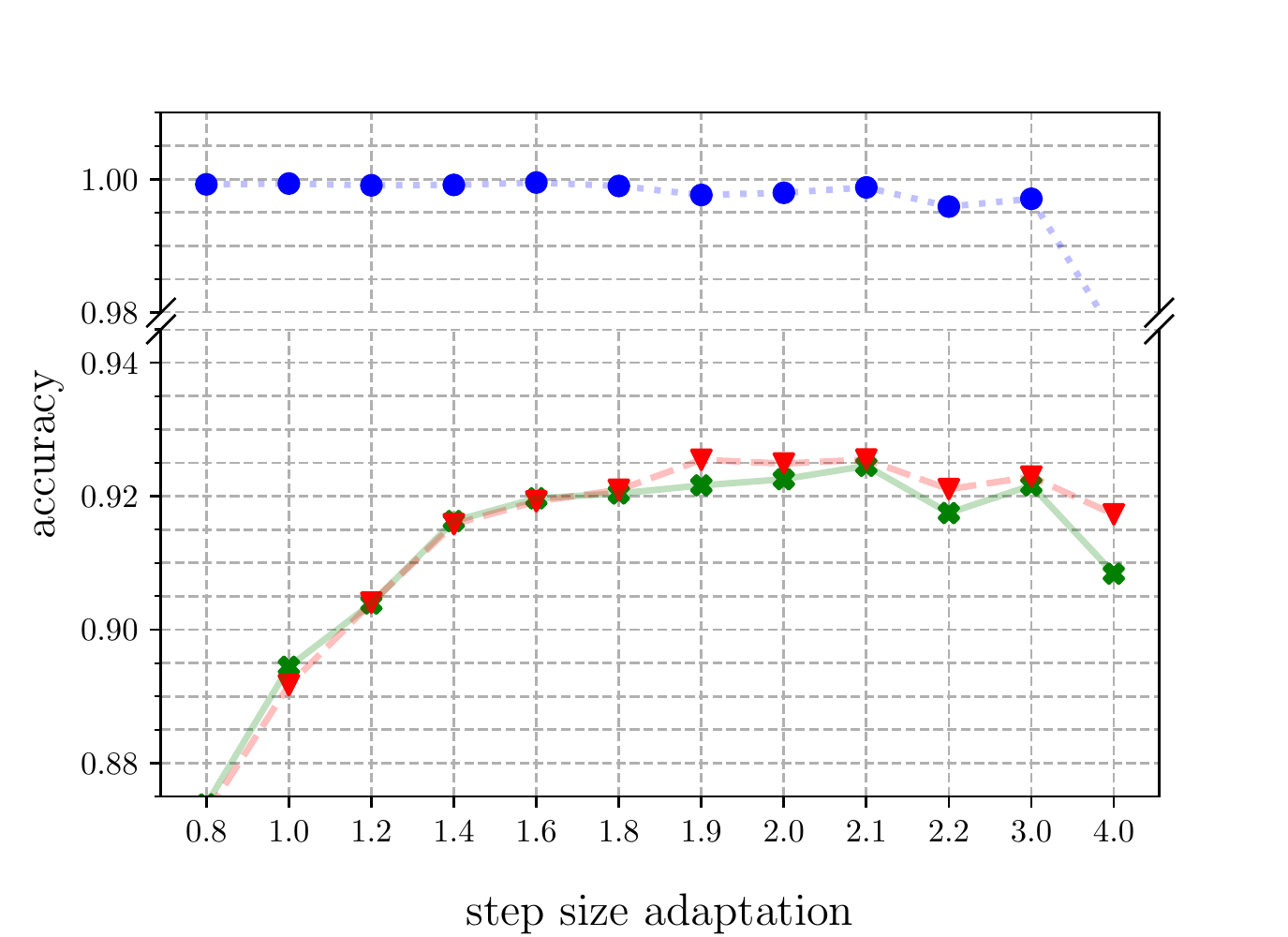}}\\
		\scalebox{\scale}{\includegraphics{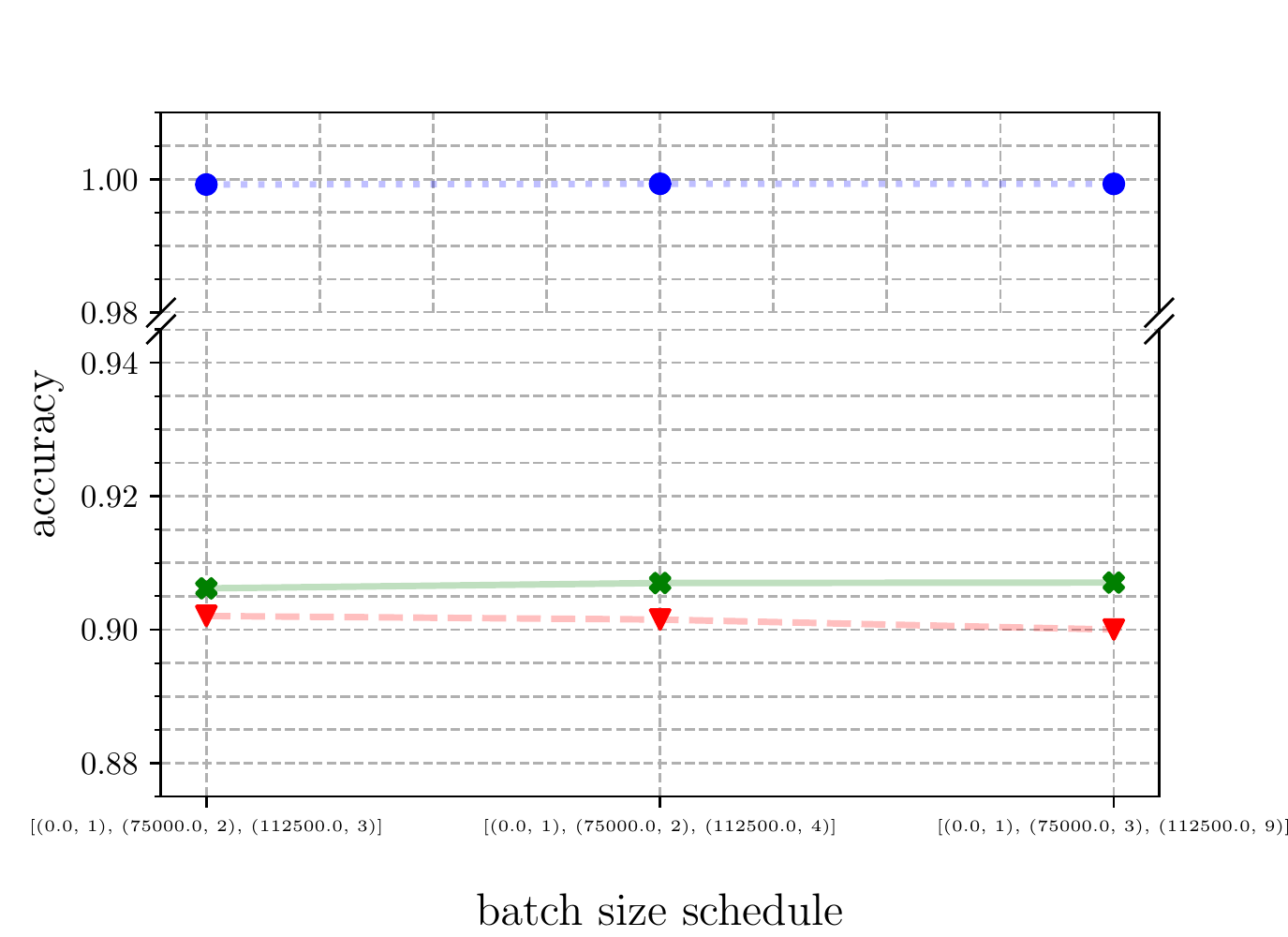}}&
		\scalebox{\scale}{\includegraphics{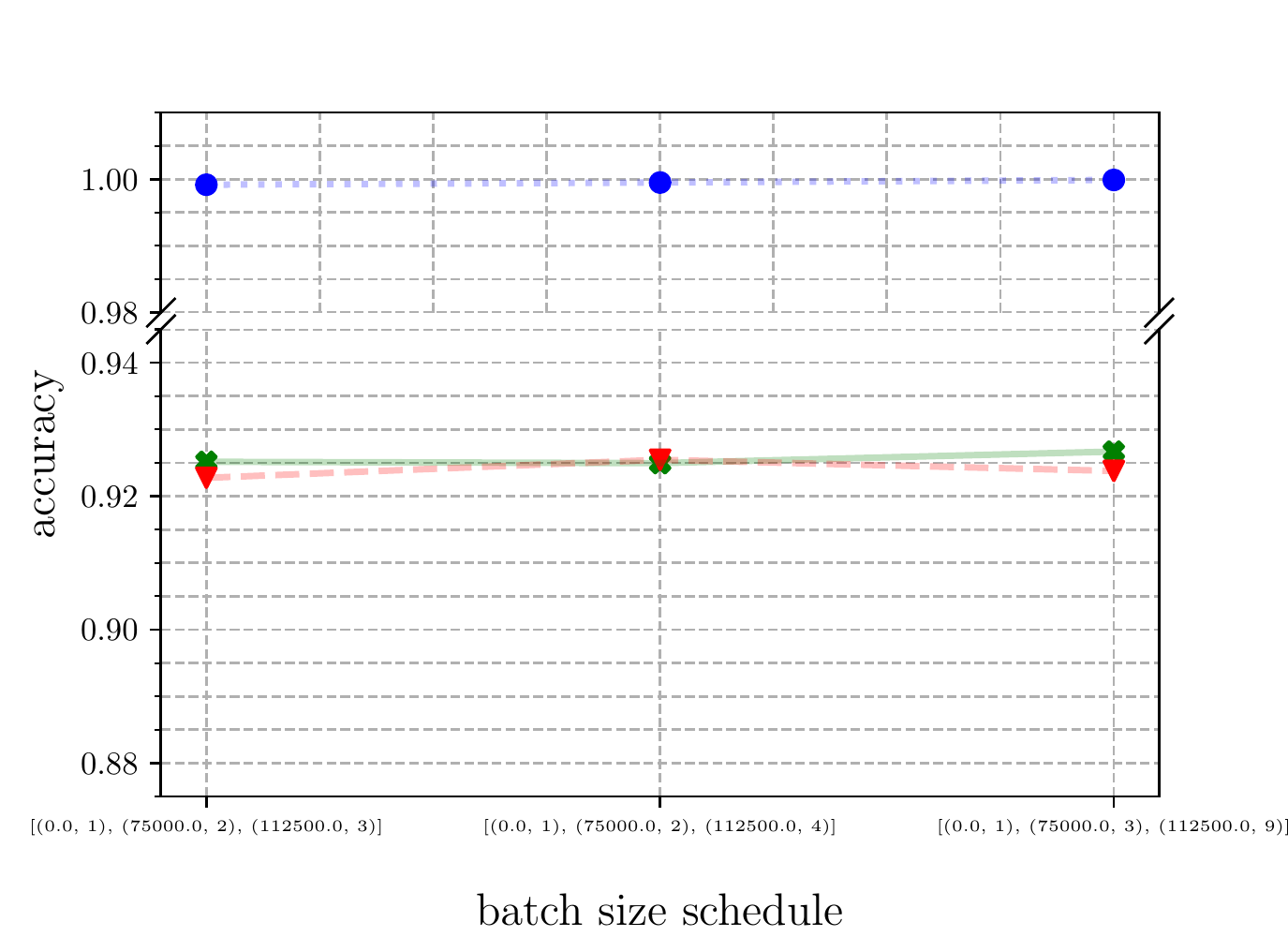}}&
		\scalebox{\scale}{\includegraphics{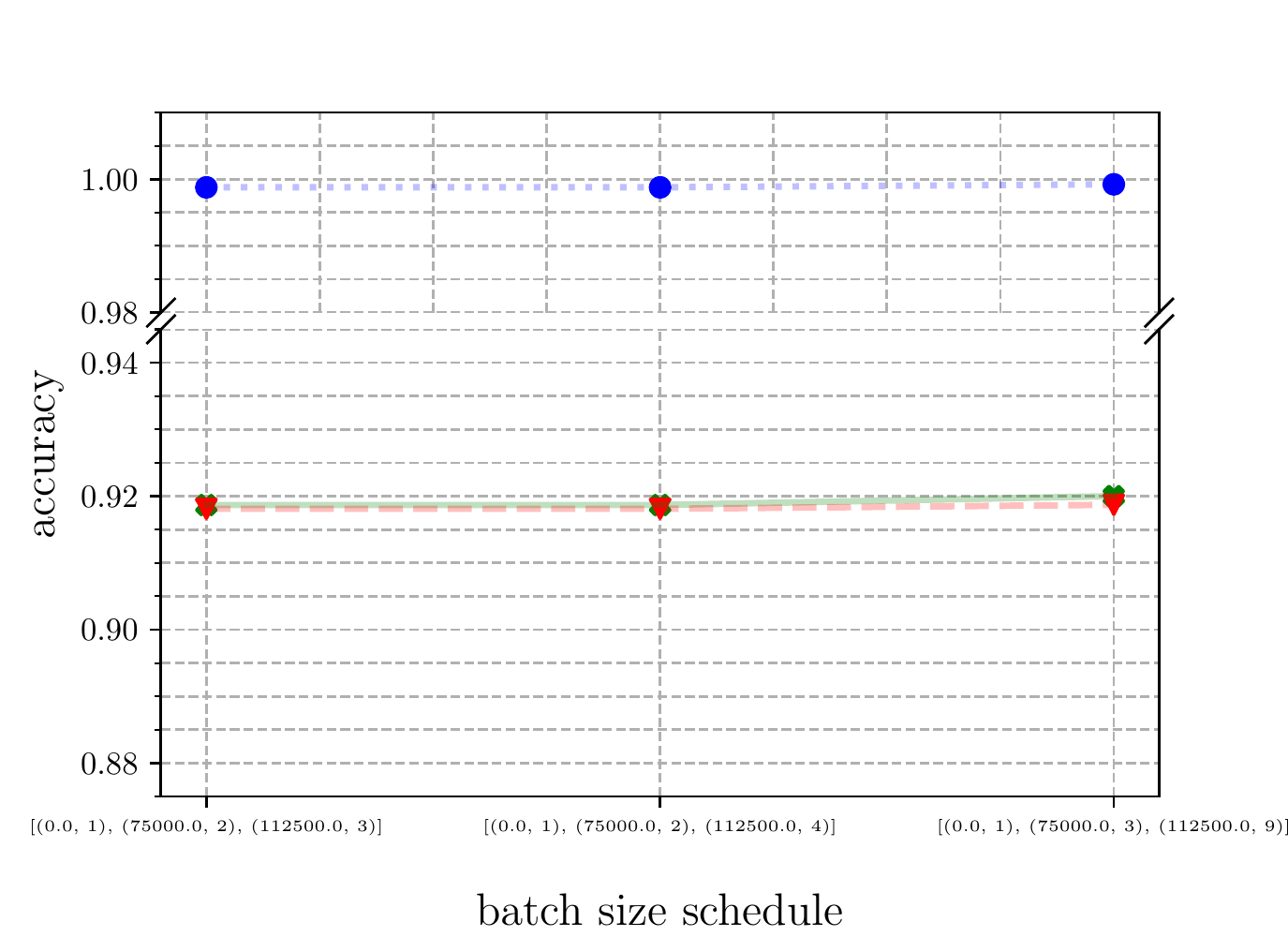}}\\
		\scalebox{\scale}{\includegraphics{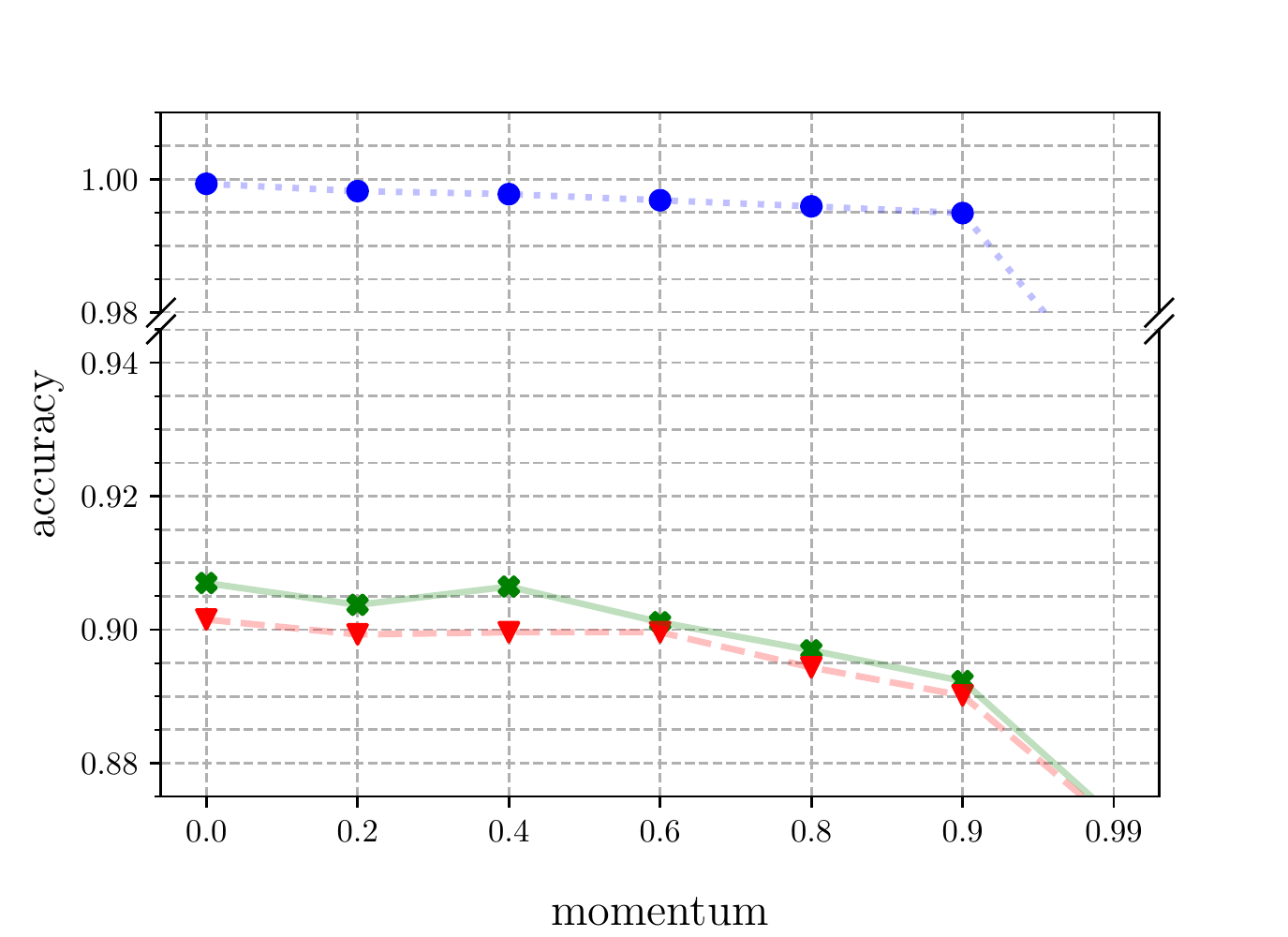}}&
		\scalebox{\scale}{\includegraphics{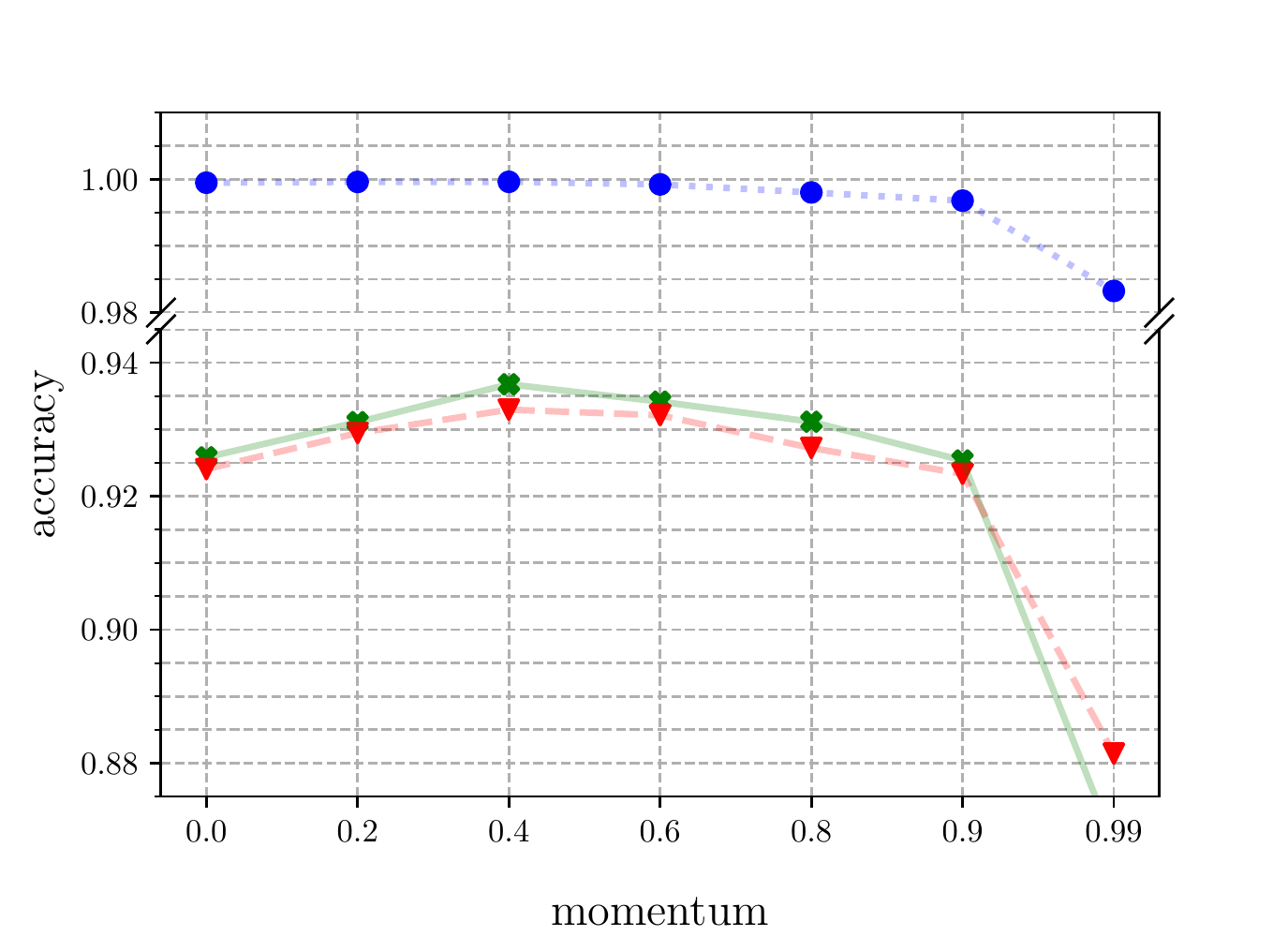}}&
		\scalebox{\scale}{\includegraphics{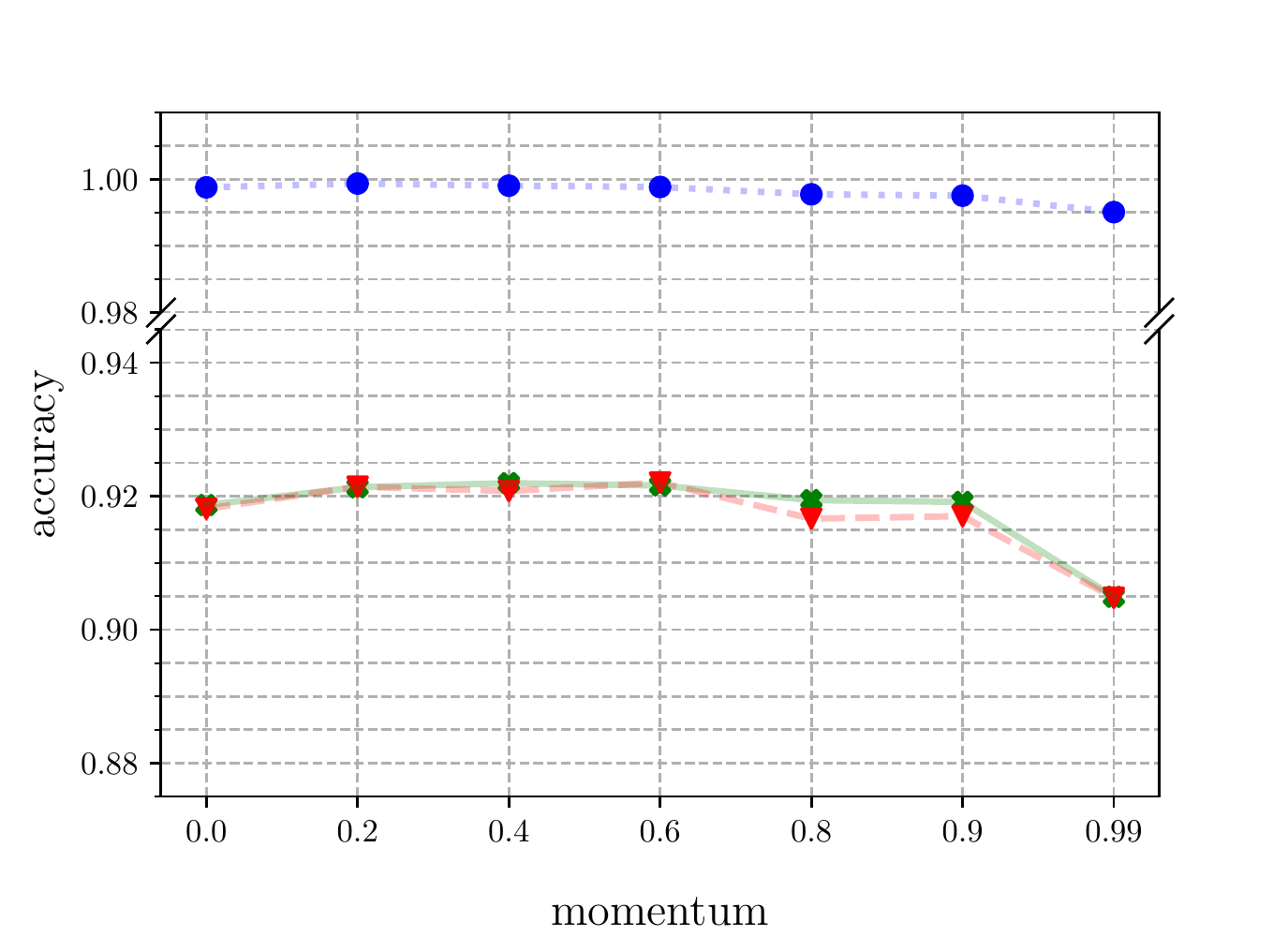}}\\
		\scalebox{\scale}{\includegraphics{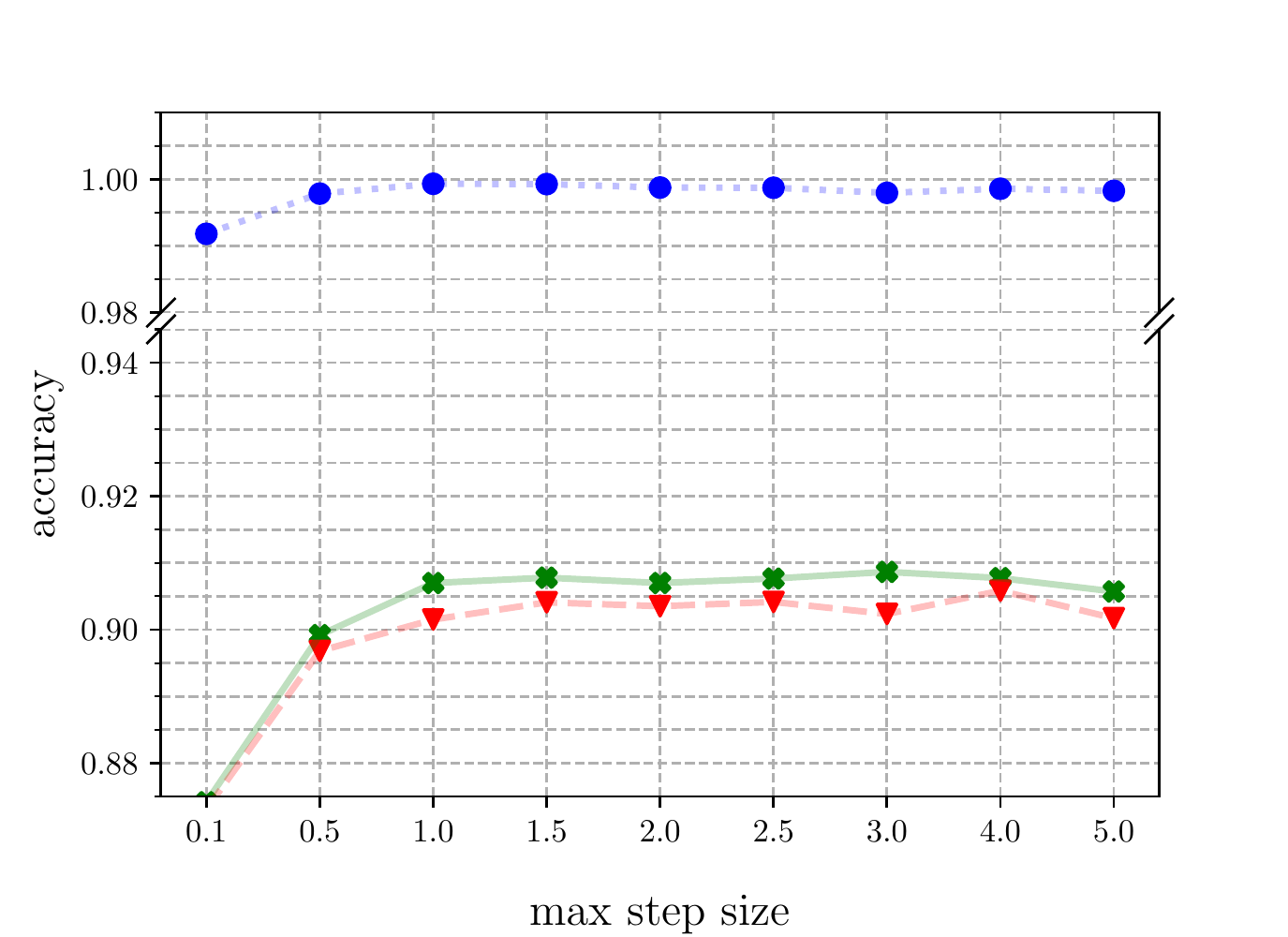}}&
		\scalebox{\scale}{\includegraphics{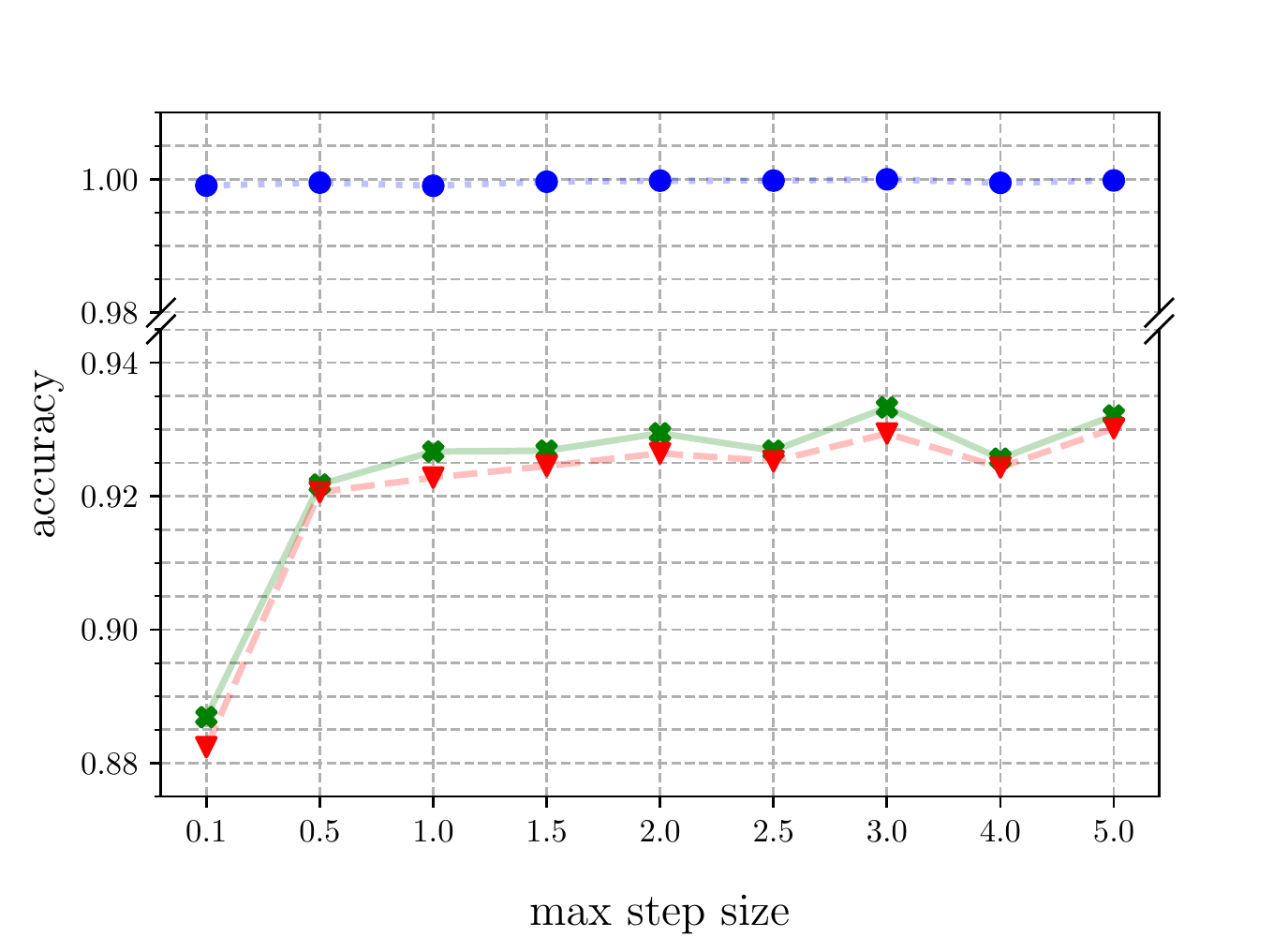}}&
		\scalebox{\scale}{\includegraphics{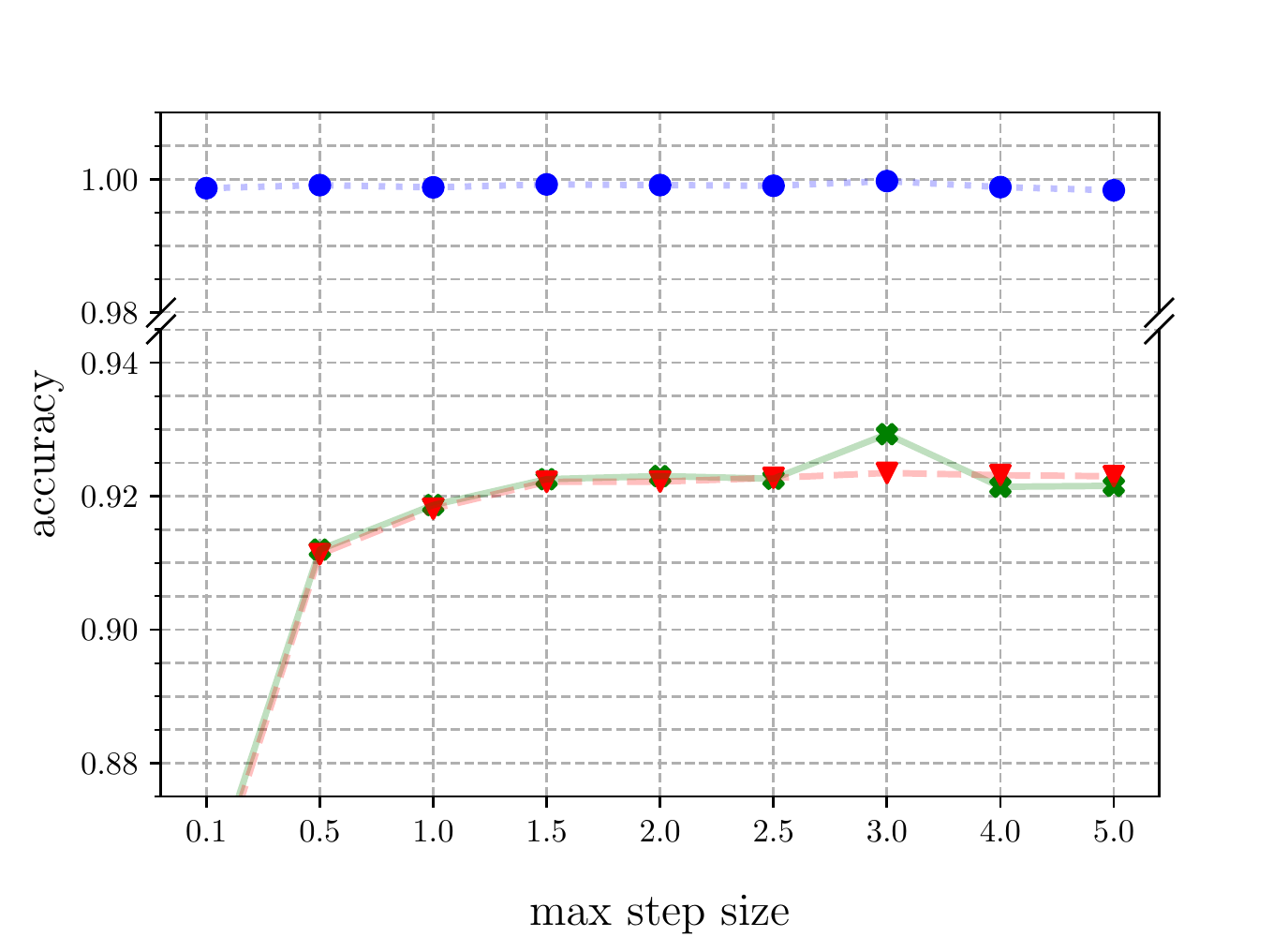}}\\
		\scalebox{\scale}{\includegraphics{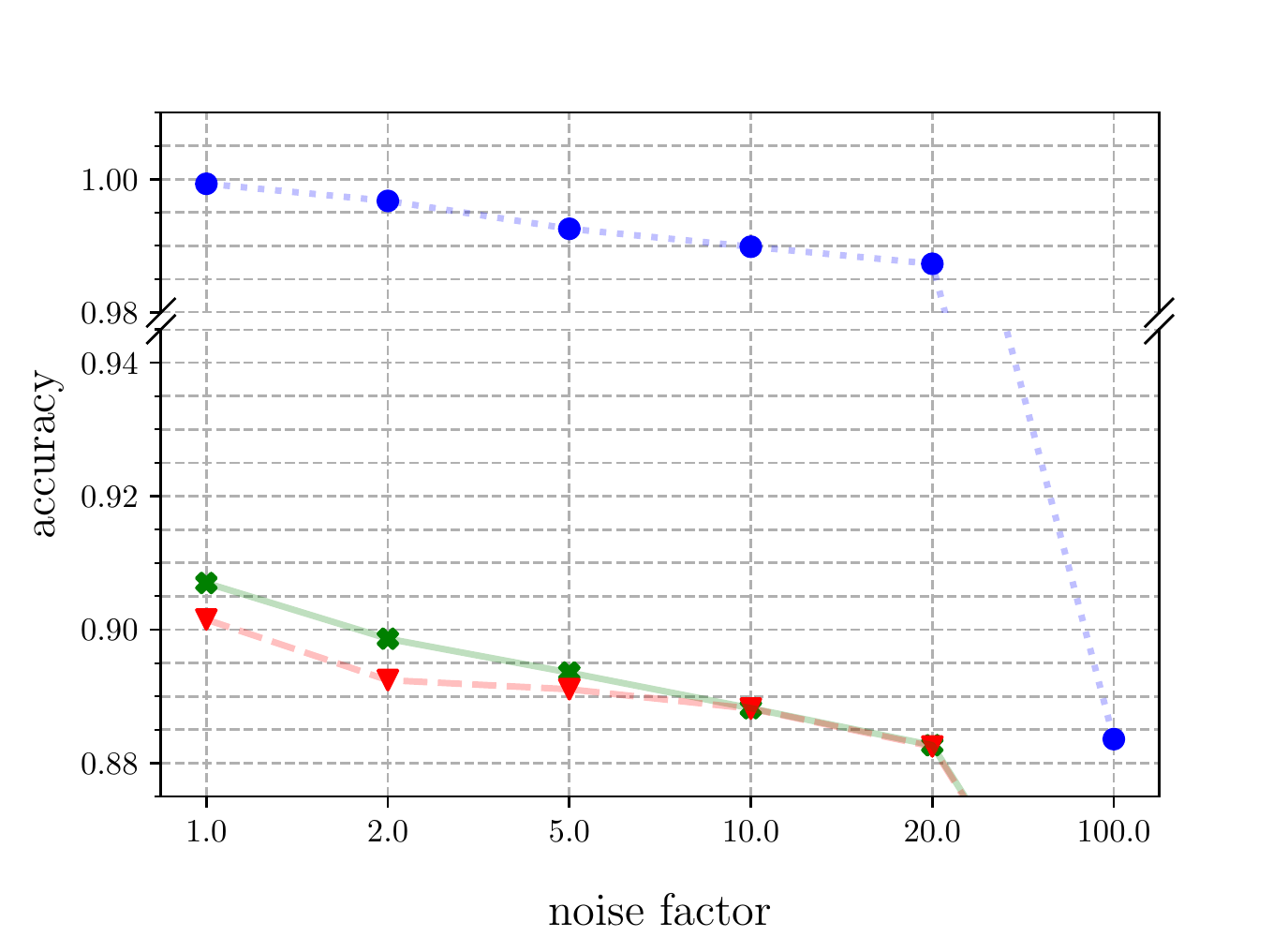}}&
		\scalebox{\scale}{\includegraphics{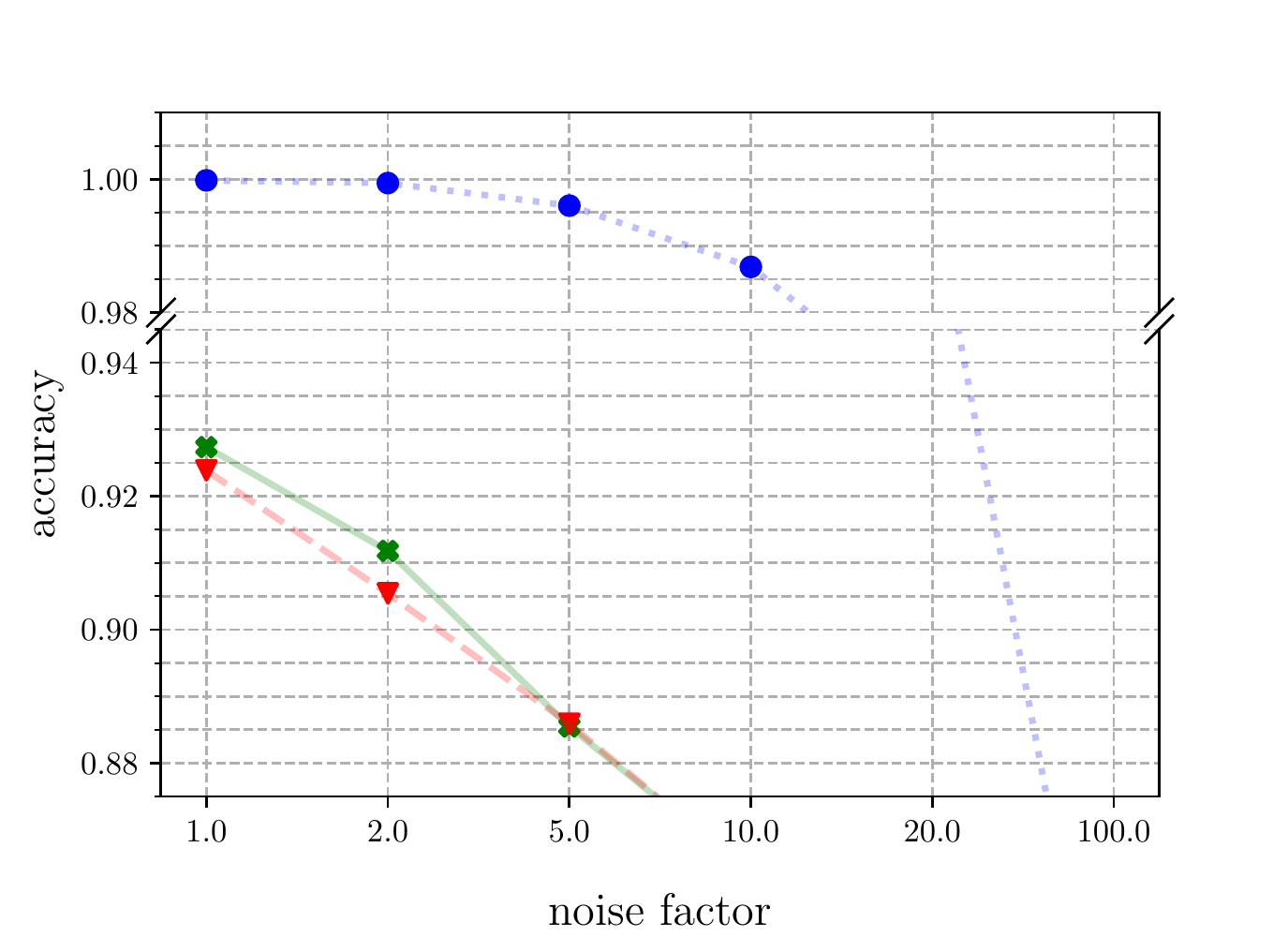}}&
		\scalebox{\scale}{\includegraphics{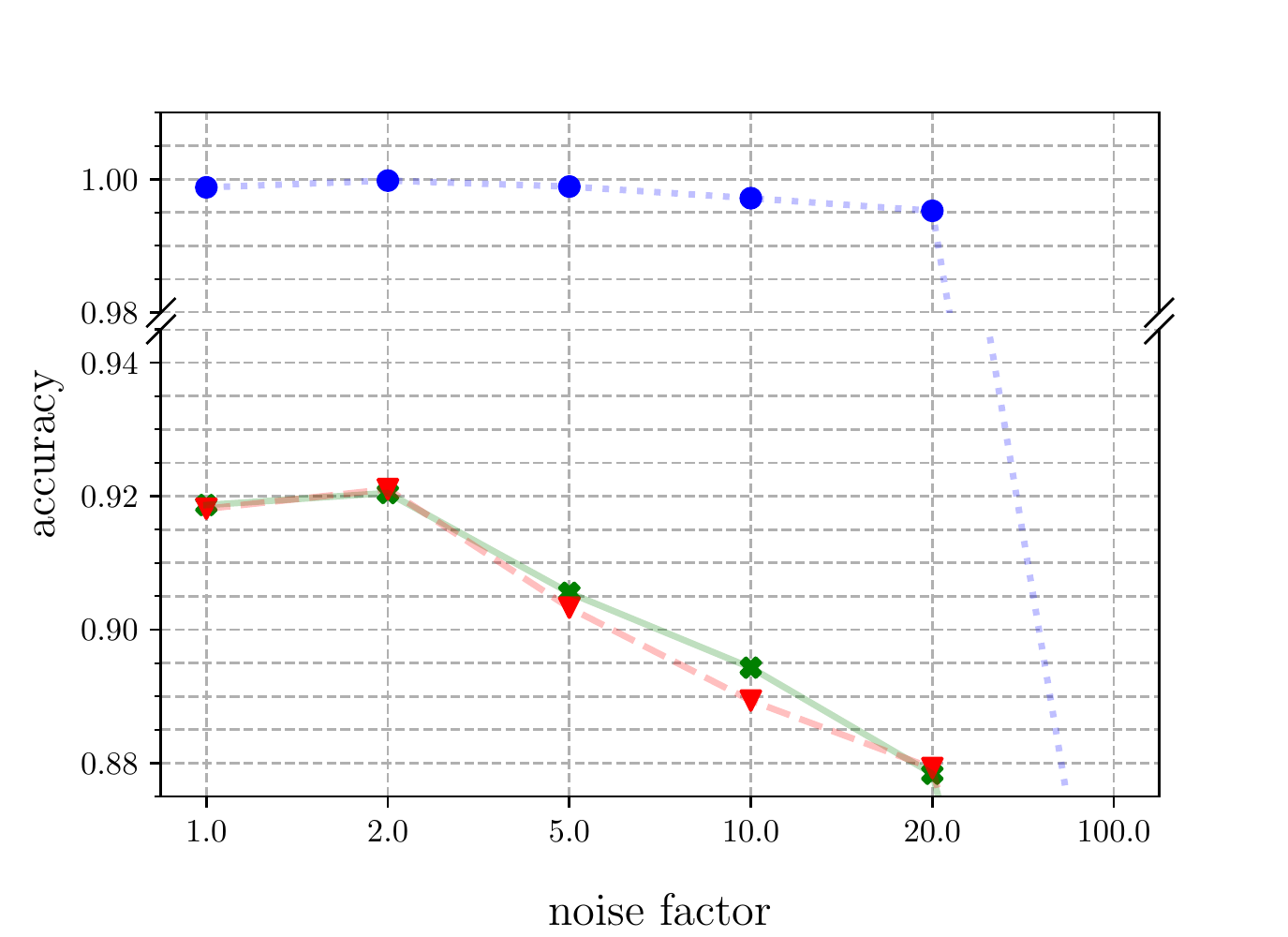}}
	\end{tabular}
	\caption{Sensitivity analysis of parameters of \textbf{LABPAL\&SGD}.  The observations of LABPAL\&NSGD described in Figure \ref{labpal_fig_sensitivity_analysis_lappalnsgd} are also valid for LABPAL\&NSGD.}
	\label{labpal_fig_sensitivity_analysis_lappalsgd}
\end{figure}

%% file: labpal/figure_sensitivity_analysis_labpalnsgd.tex
\begin{figure}[h!]
	\centering
	\vspace{-1cm}
	\def\scale{0.28}
	\begin{tabular}{ c c c}	
		\textbf{ResNet-20}&\textbf{MobileNet-V2} &\textbf{DenseNet-121}\\
		\scalebox{\scale}{\includegraphics{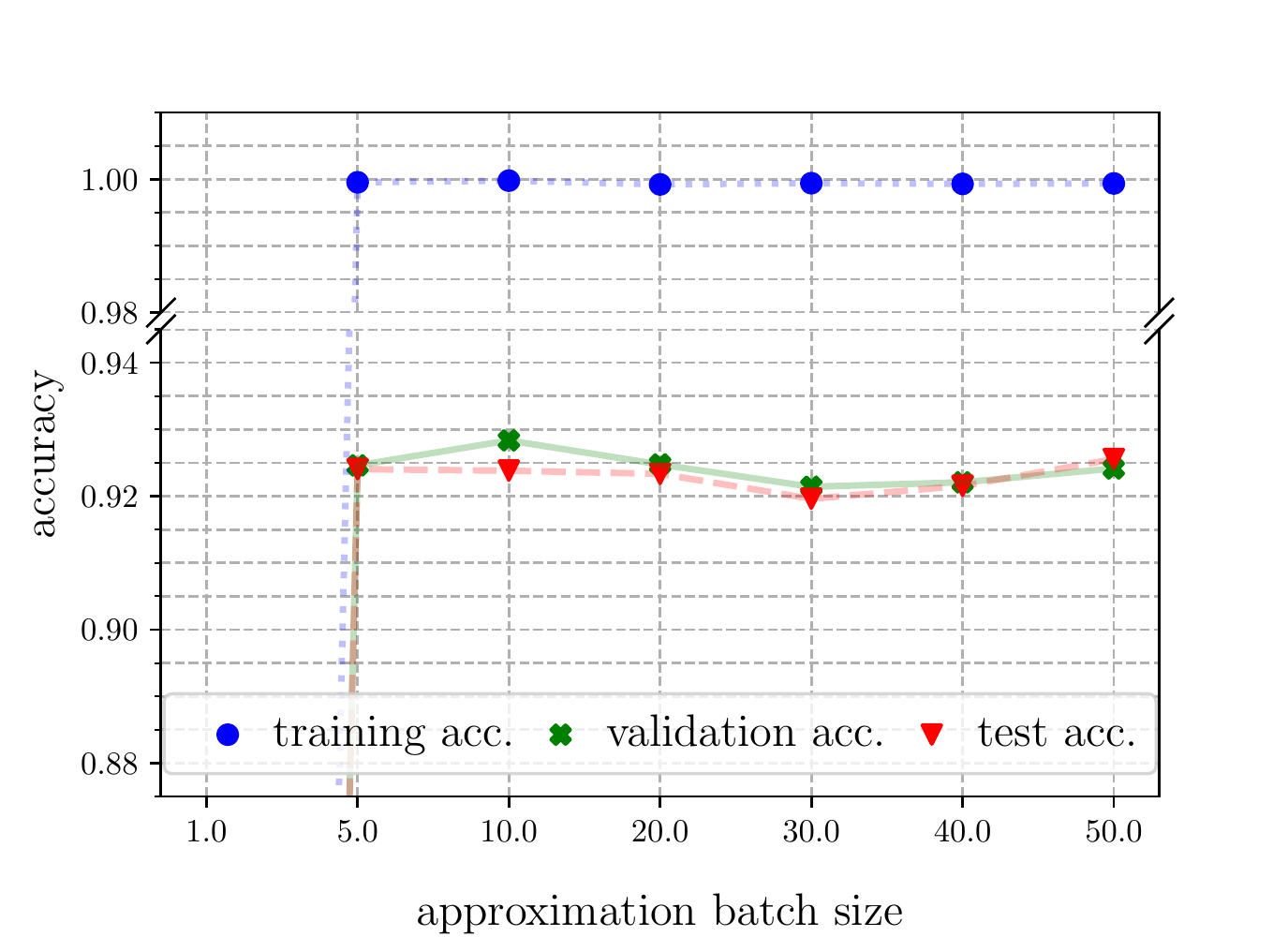}}&
		\scalebox{\scale}{\includegraphics{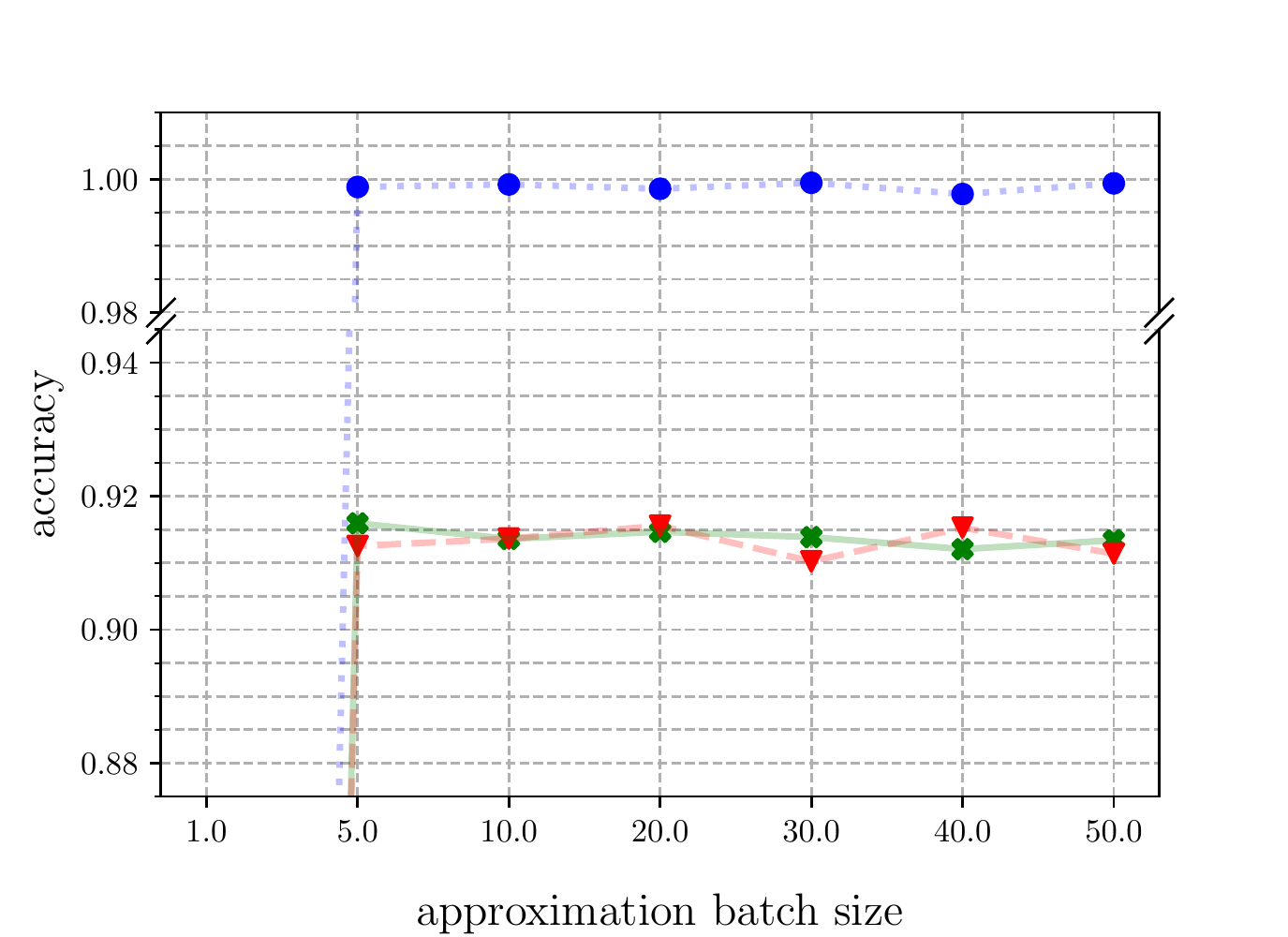}}&
		\scalebox{\scale}{\includegraphics{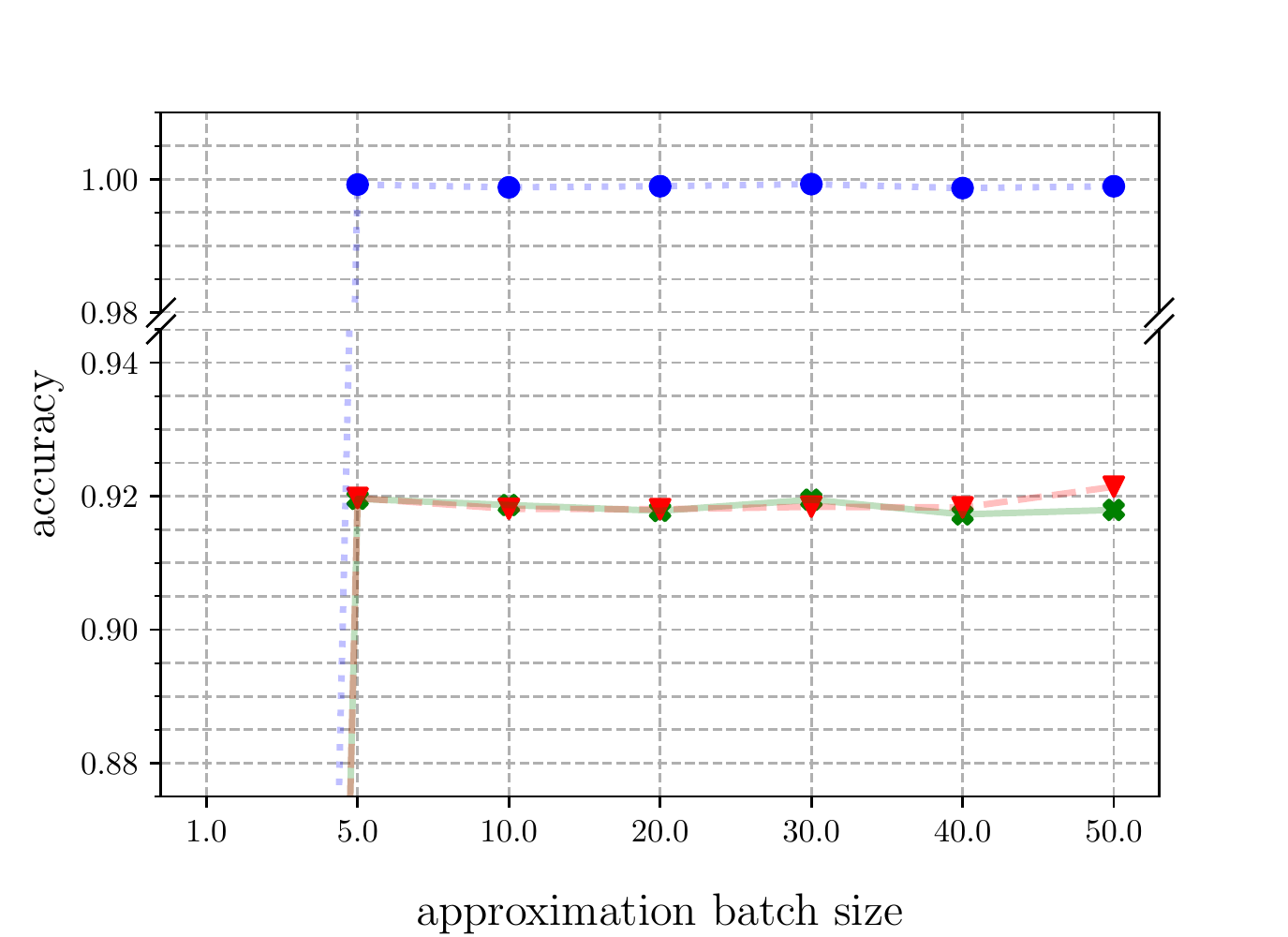}}\\
		\scalebox{\scale}{\includegraphics{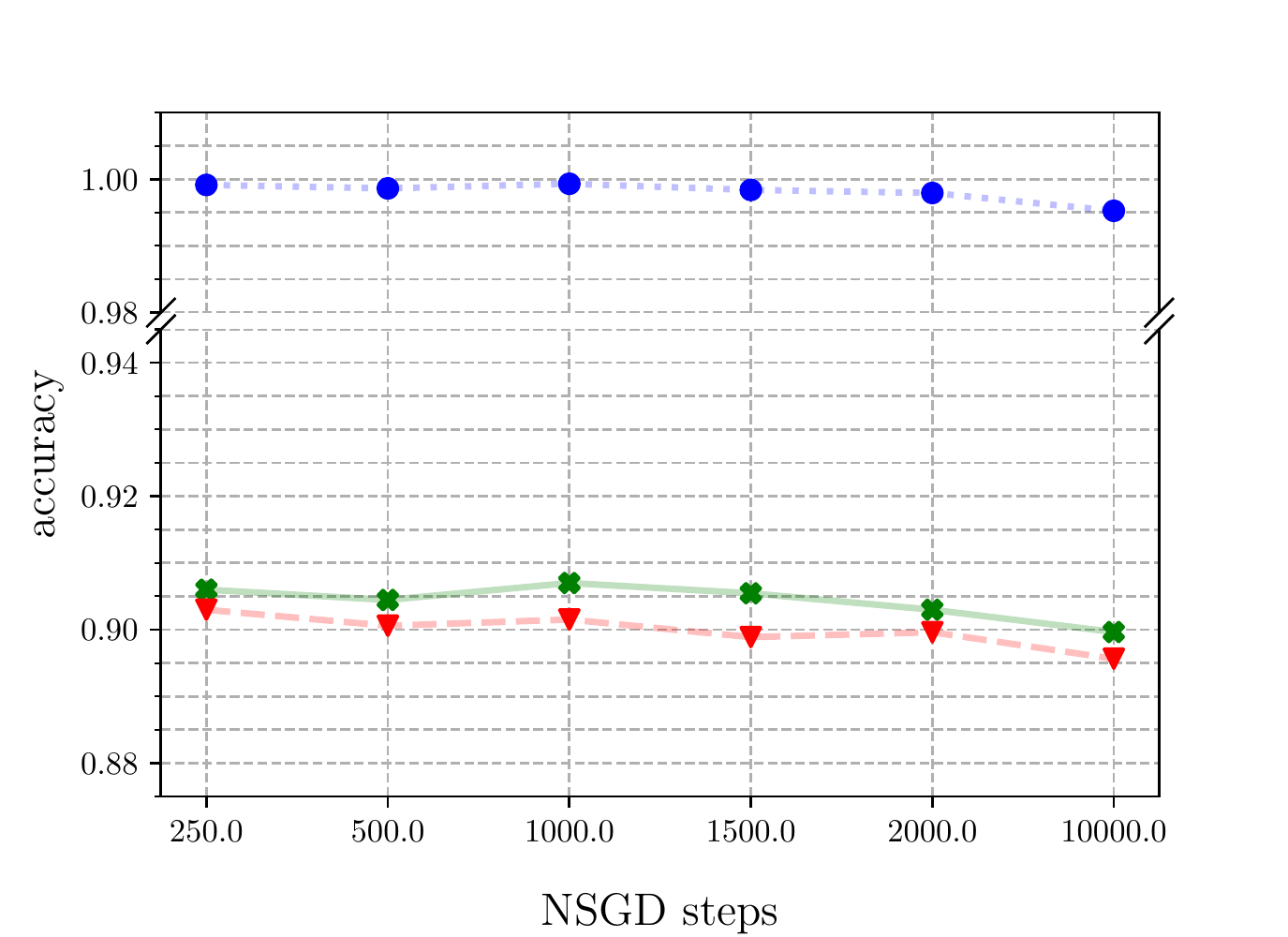}}&
		\scalebox{\scale}{\includegraphics{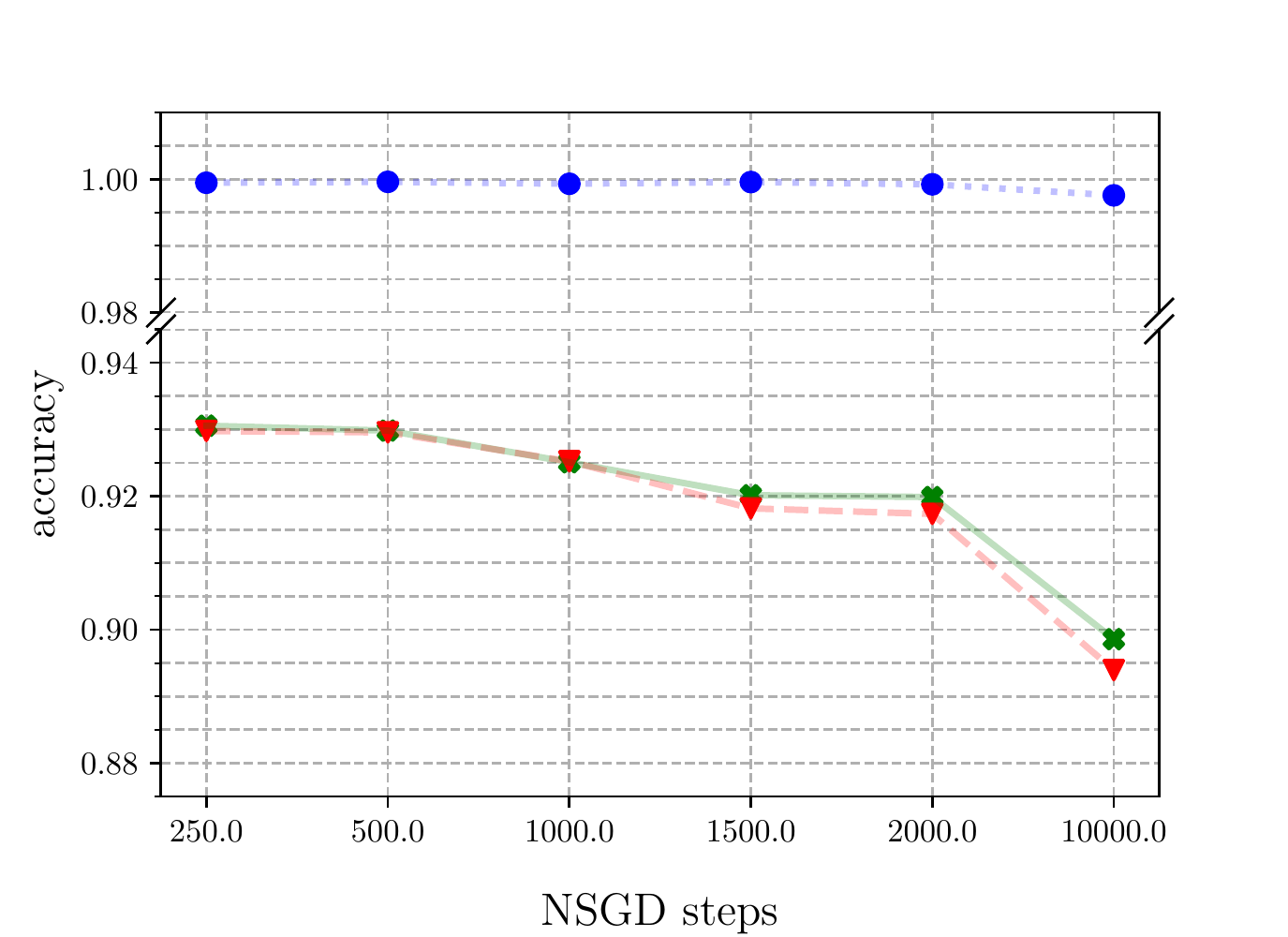}}&
		\scalebox{\scale}{\includegraphics{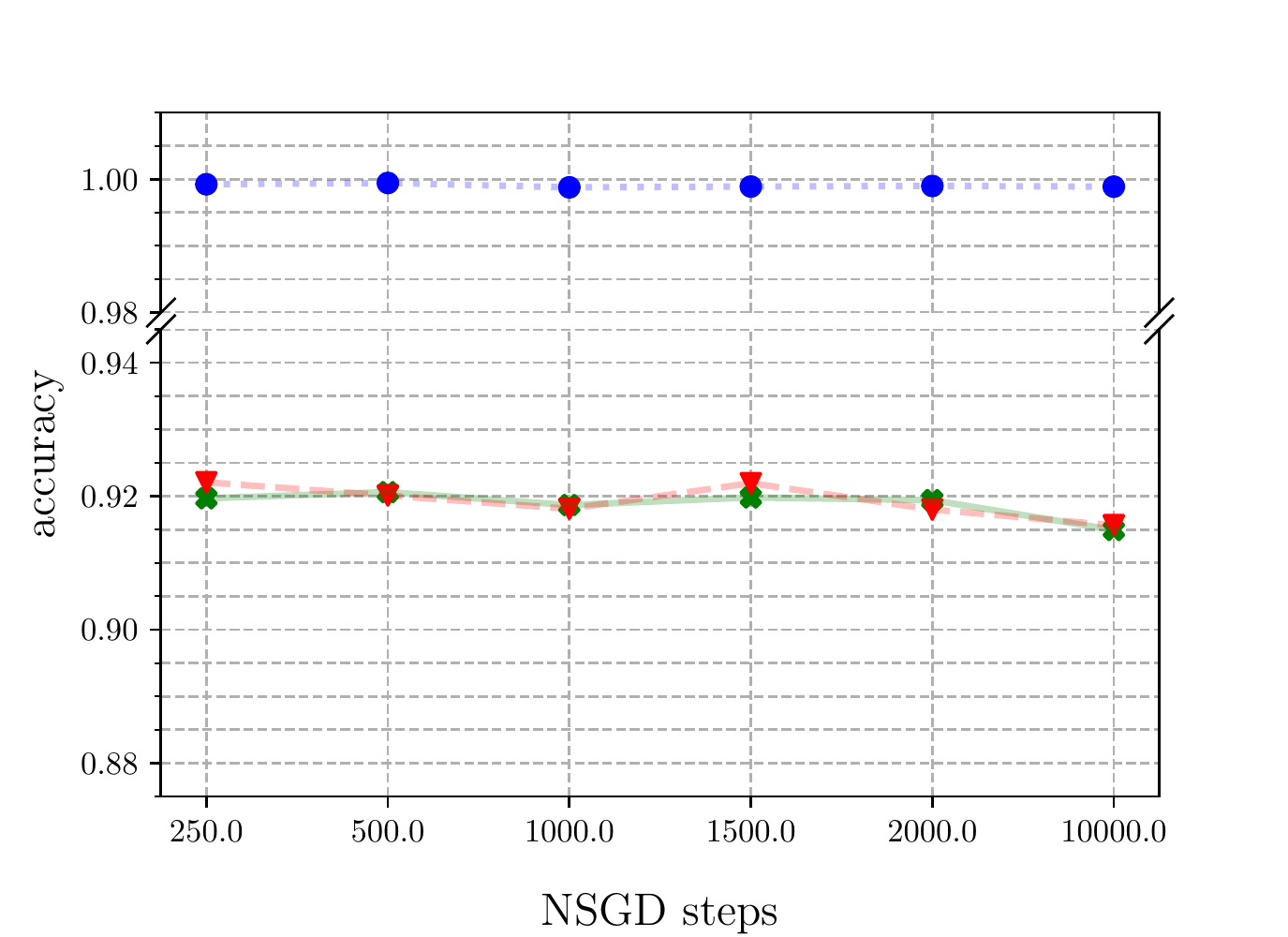}}\\
		\scalebox{\scale}{\includegraphics{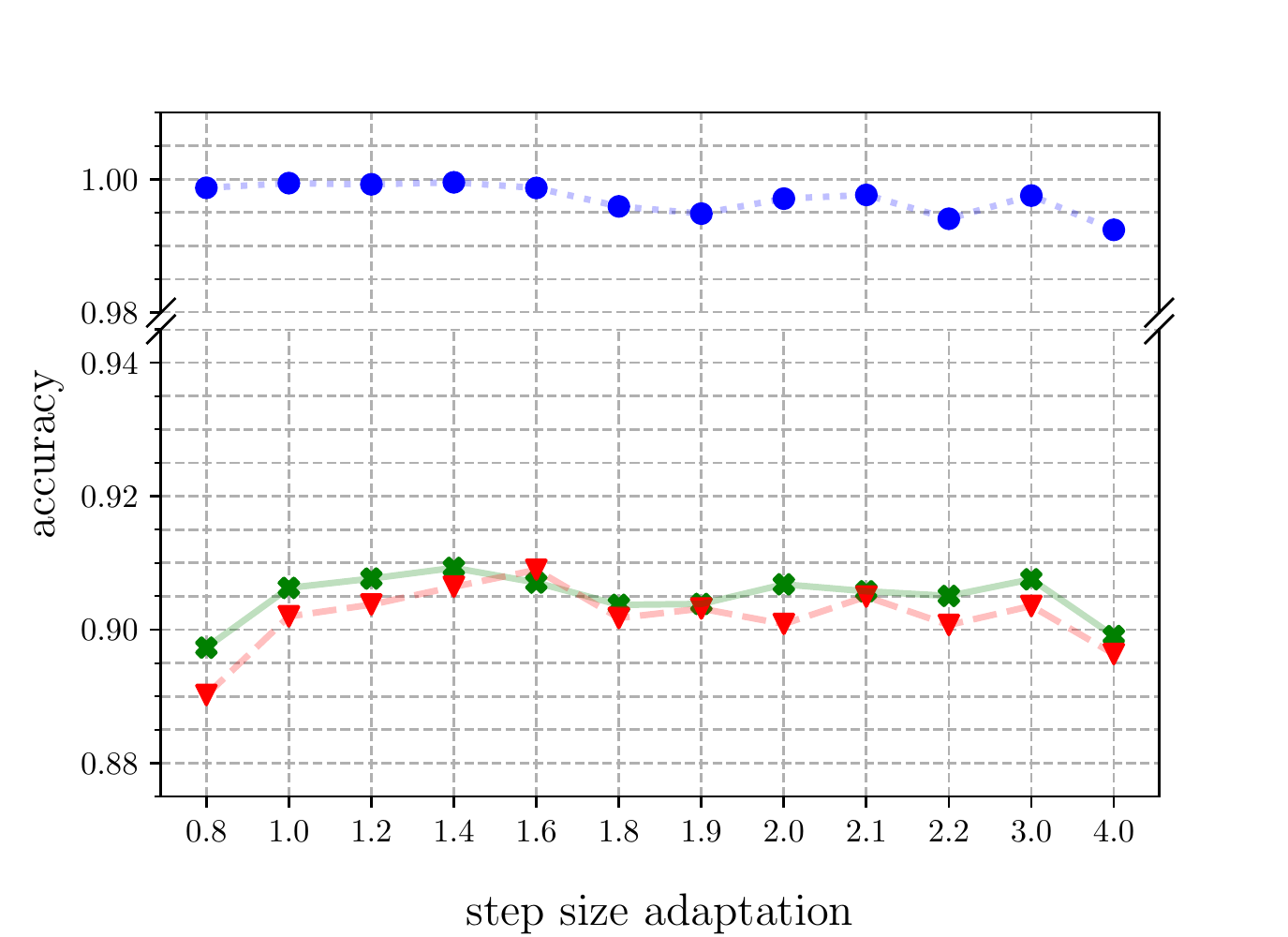}}&
		\scalebox{\scale}{\includegraphics{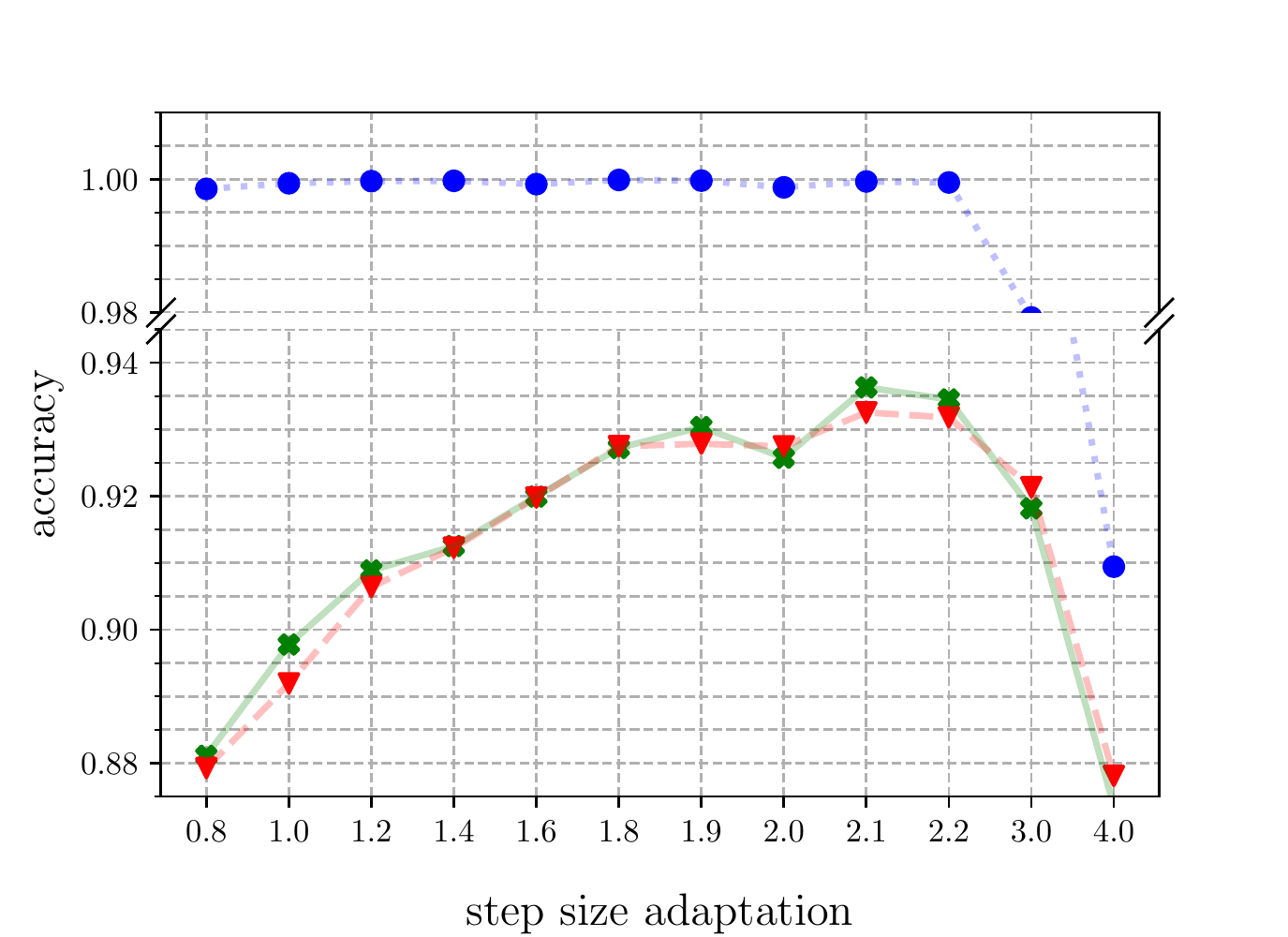}}&
		\scalebox{\scale}{\includegraphics{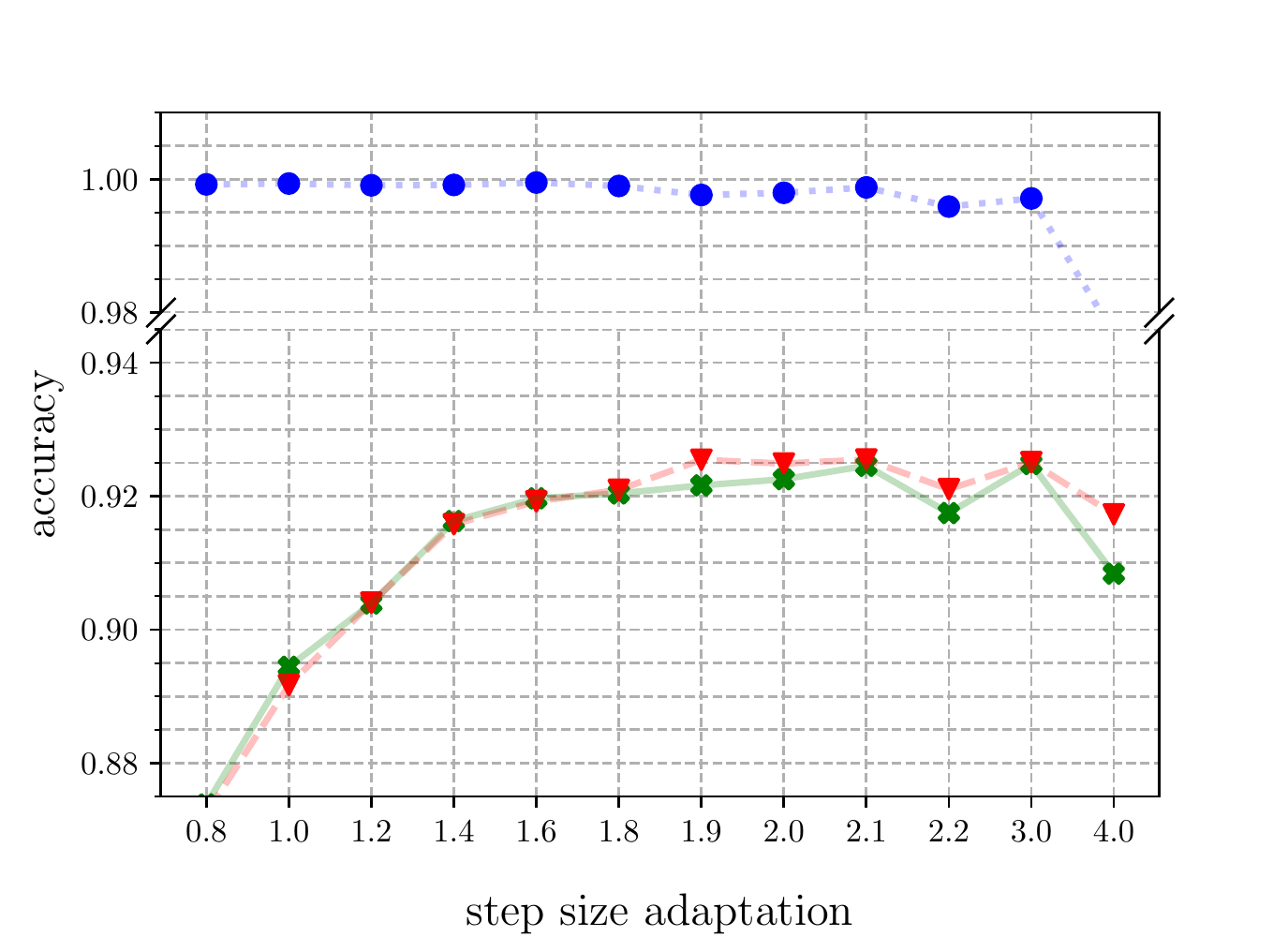}}\\
		\scalebox{\scale}{\includegraphics{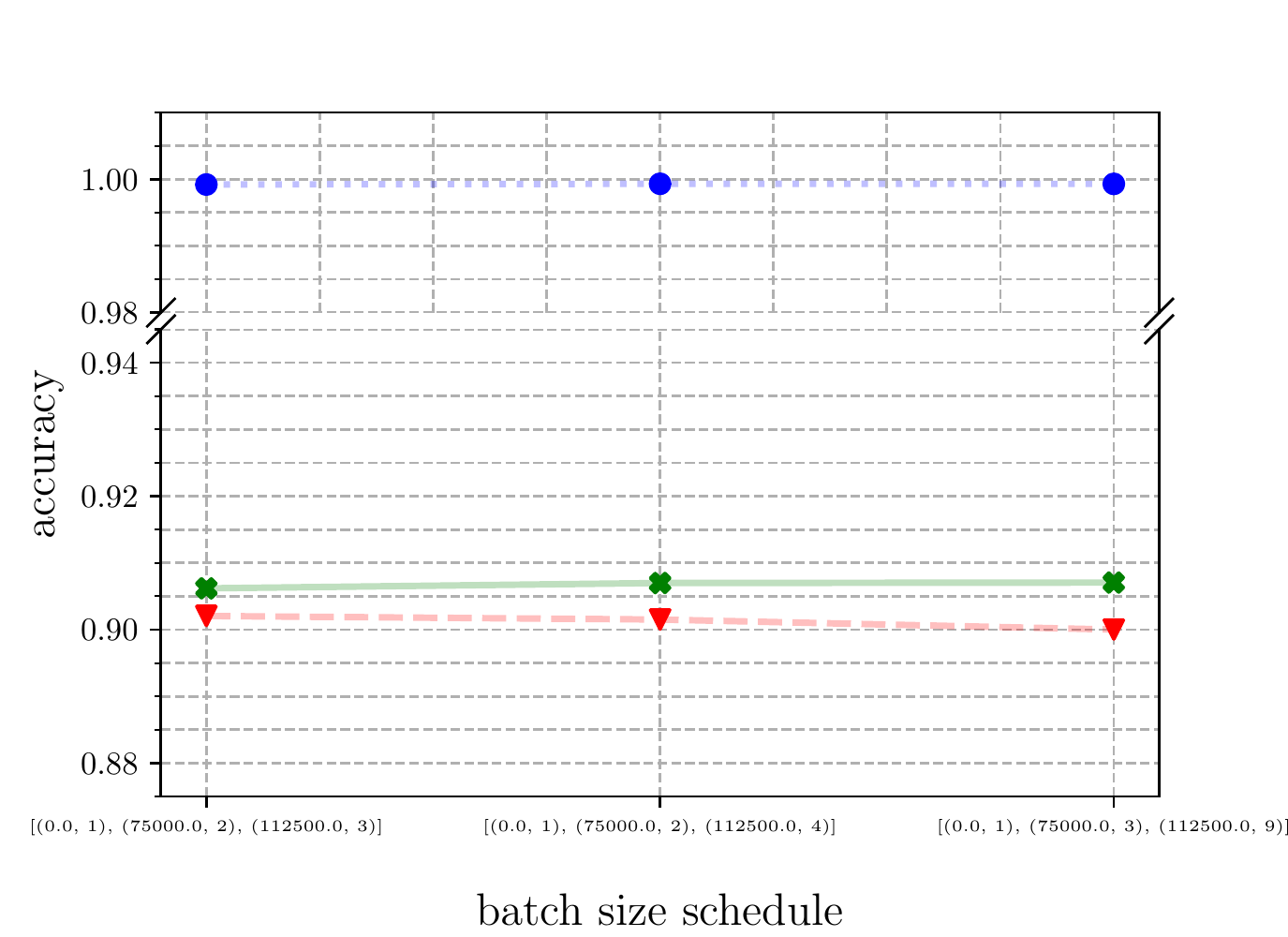}}&
		\scalebox{\scale}{\includegraphics{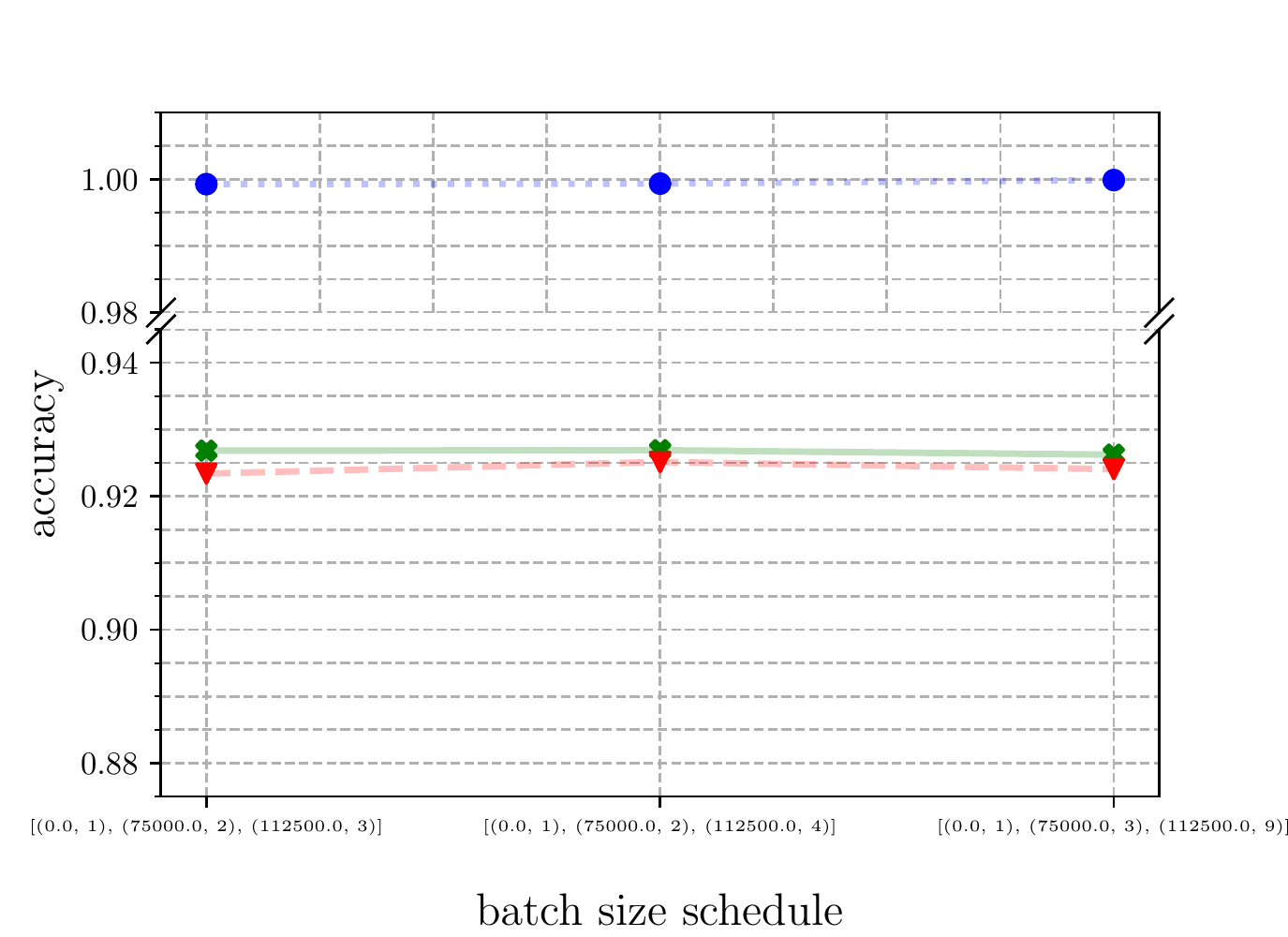}}&
		\scalebox{\scale}{\includegraphics{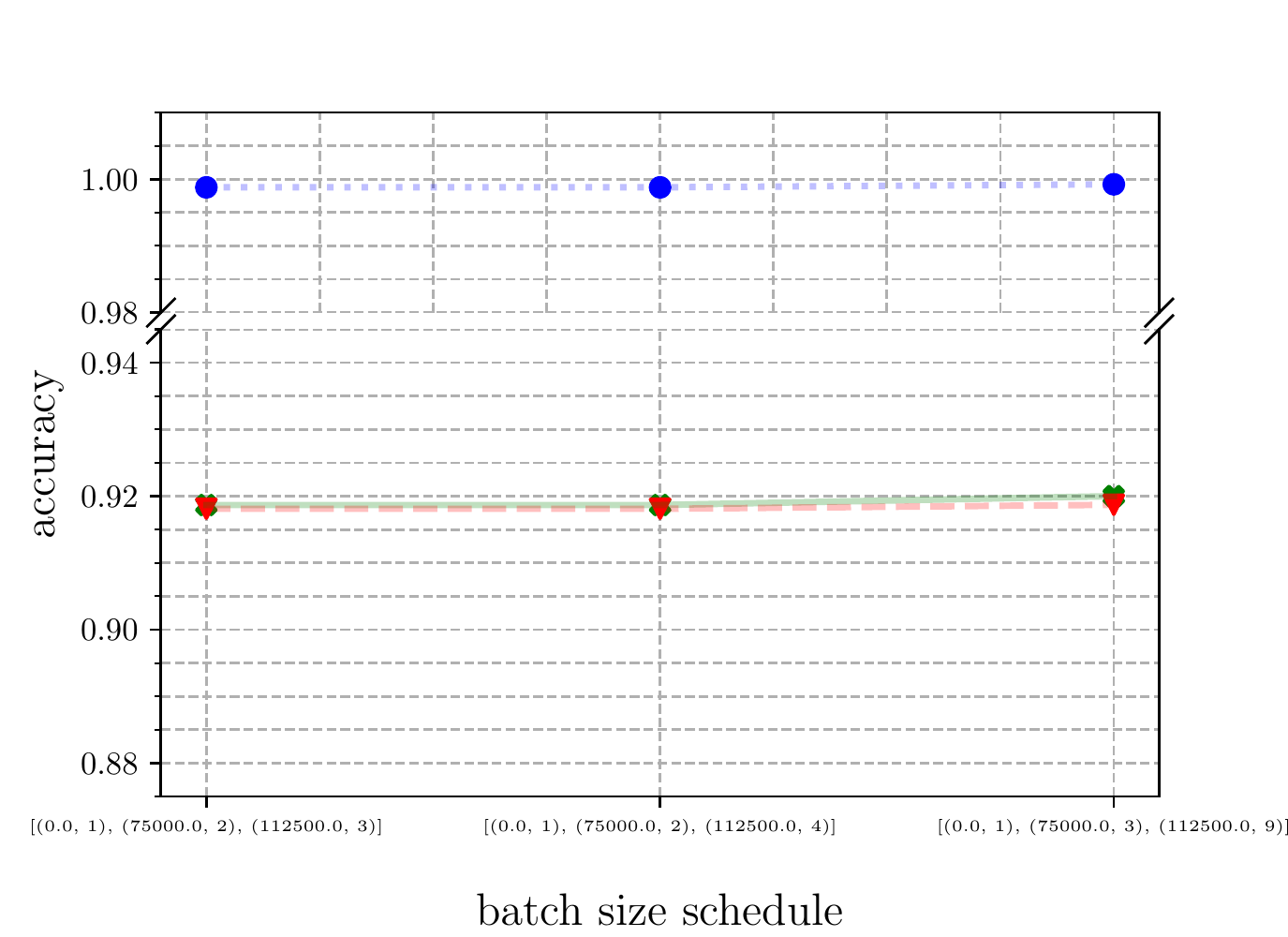}}\\
		\scalebox{\scale}{\includegraphics{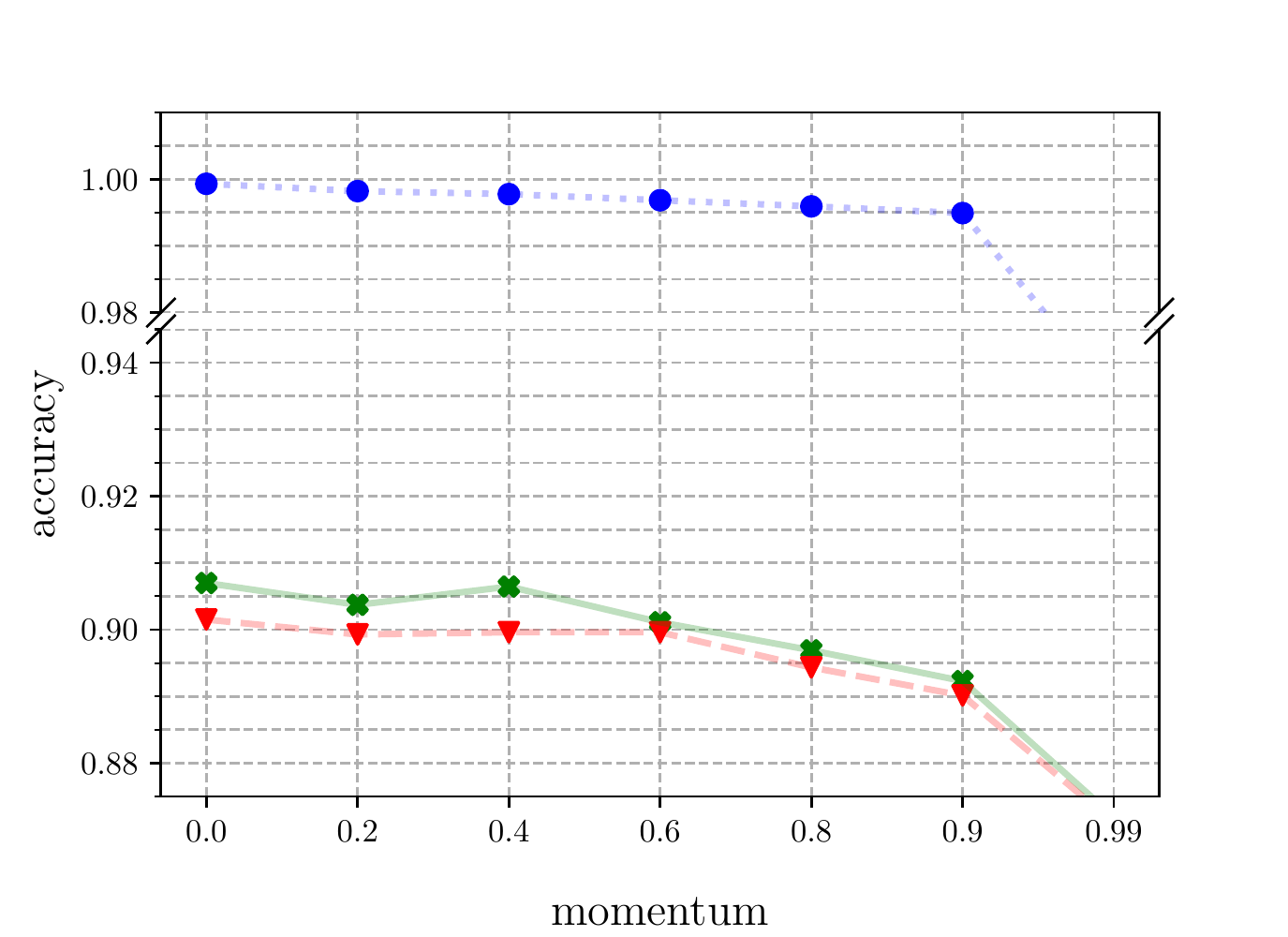}}&
		\scalebox{\scale}{\includegraphics{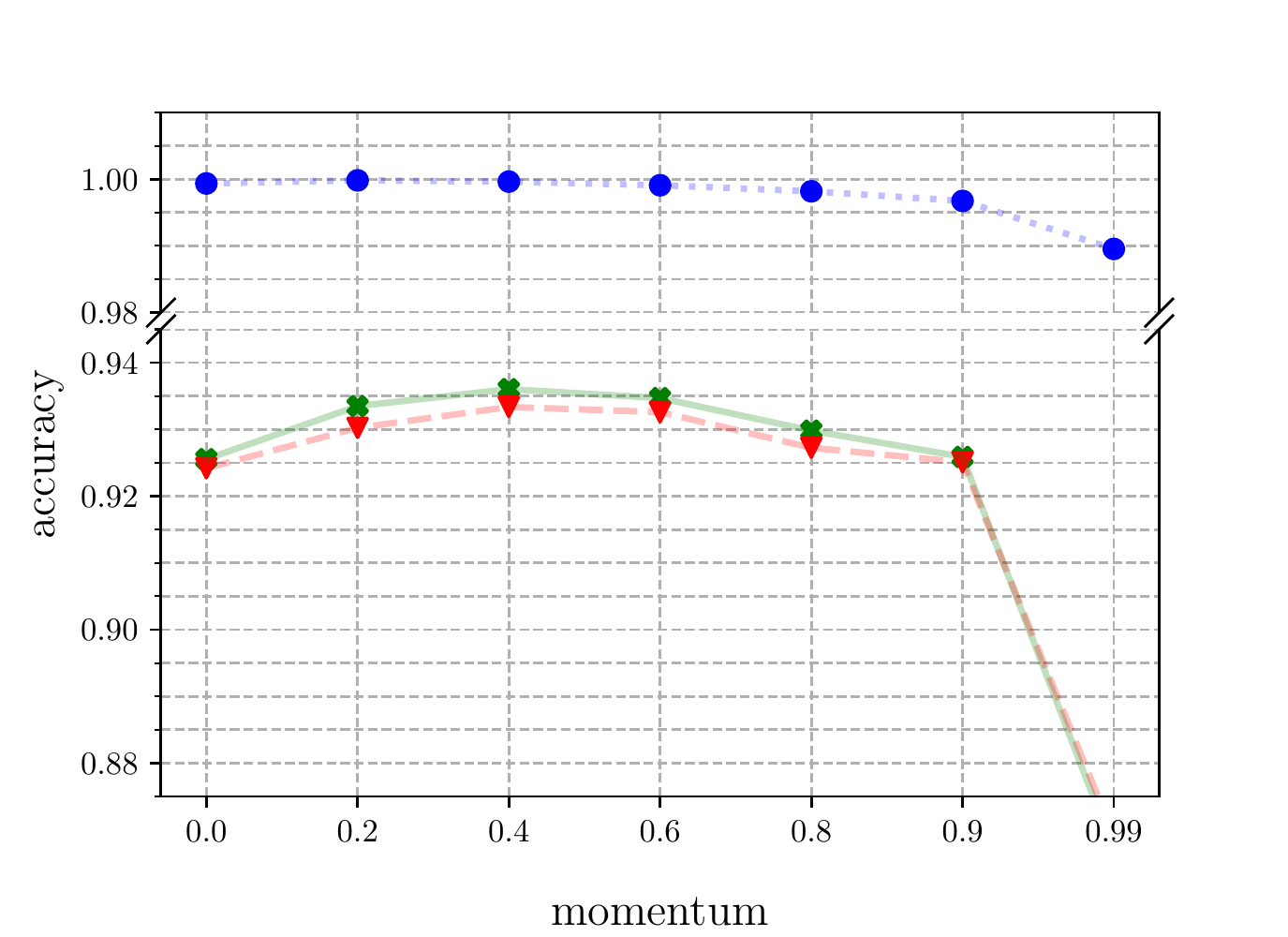}}&
		\scalebox{\scale}{\includegraphics{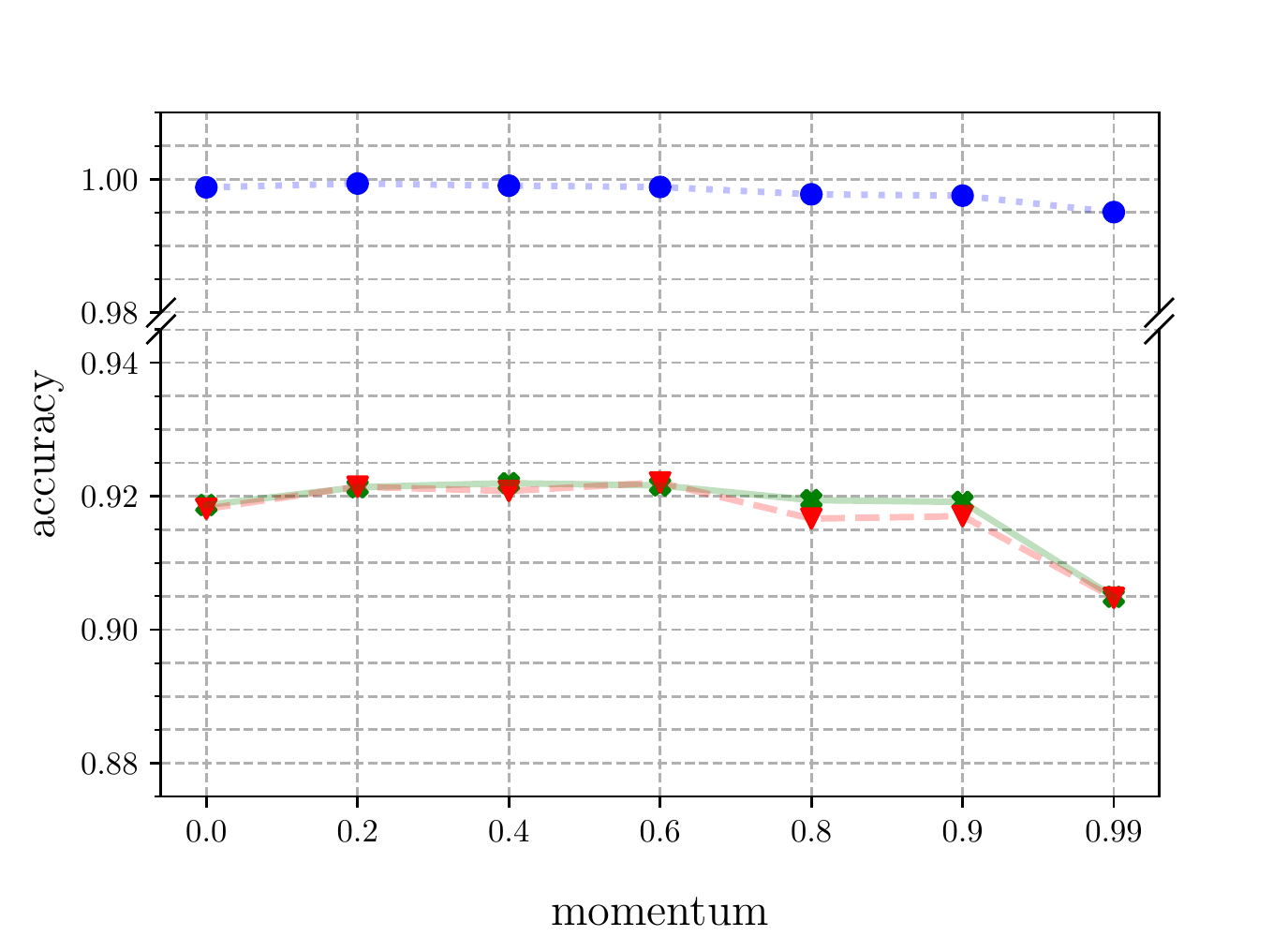}}\\
		\scalebox{\scale}{\includegraphics{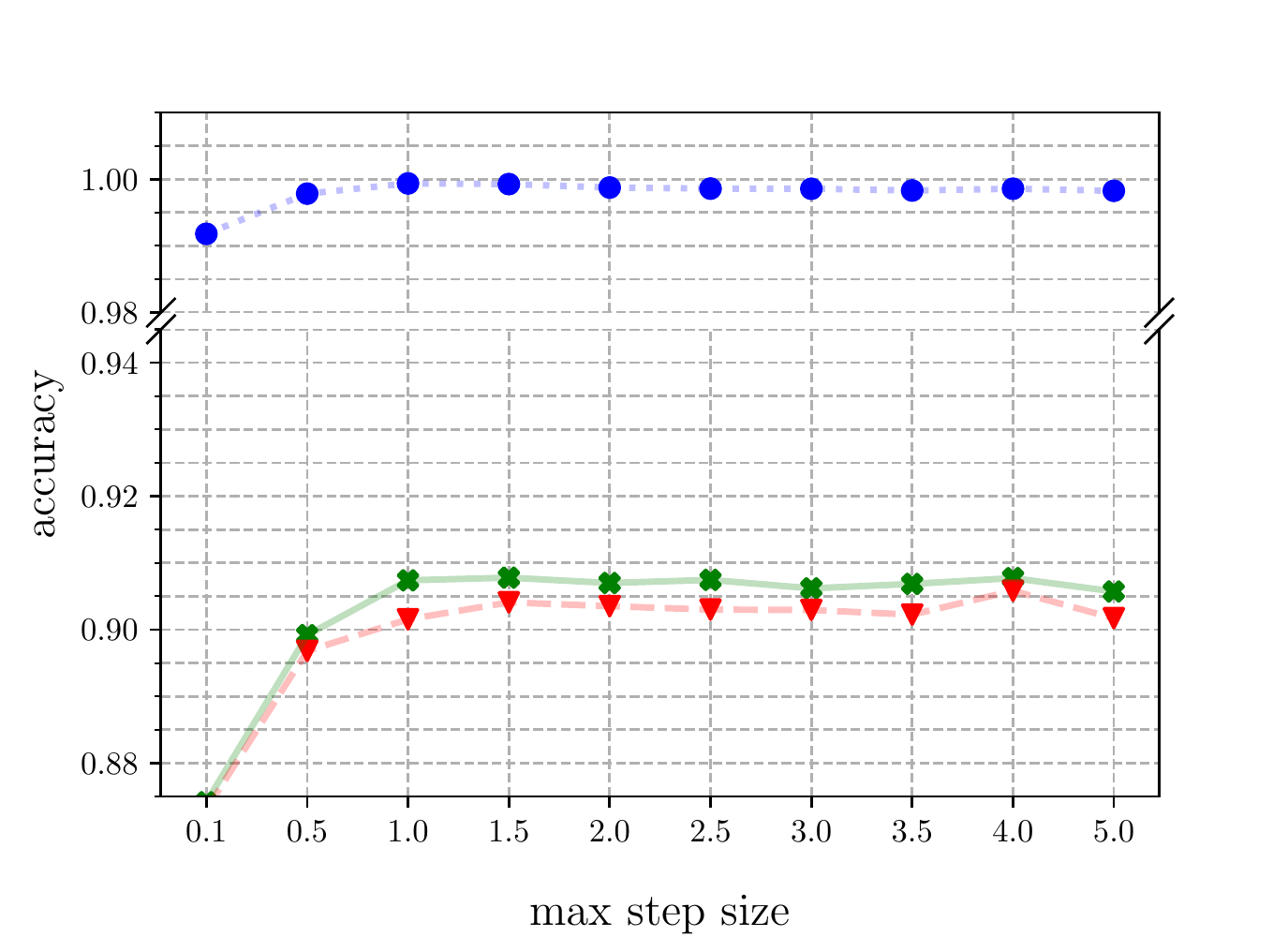}}&	
		\scalebox{\scale}{\includegraphics{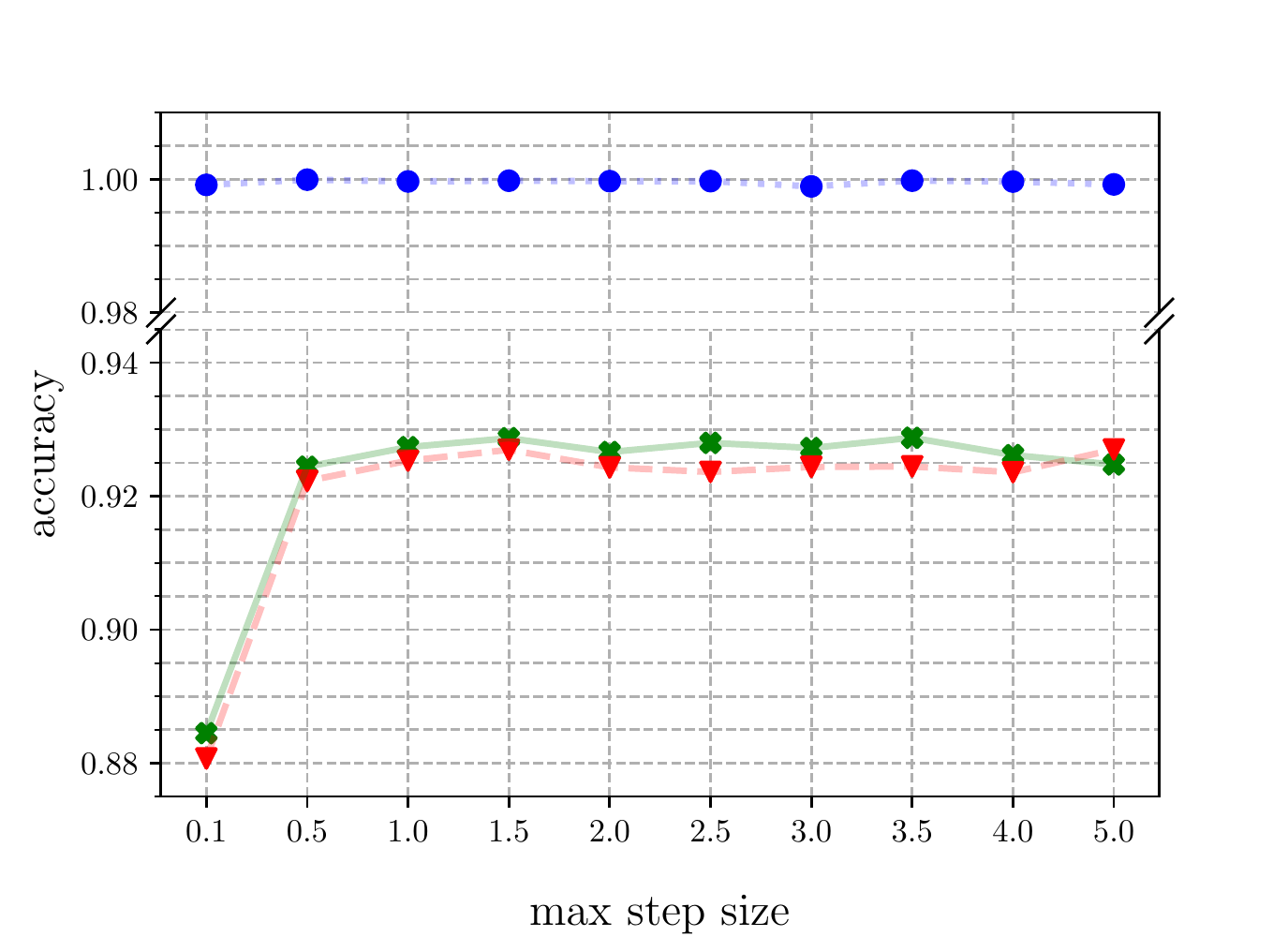}}&
		\scalebox{\scale}{\includegraphics{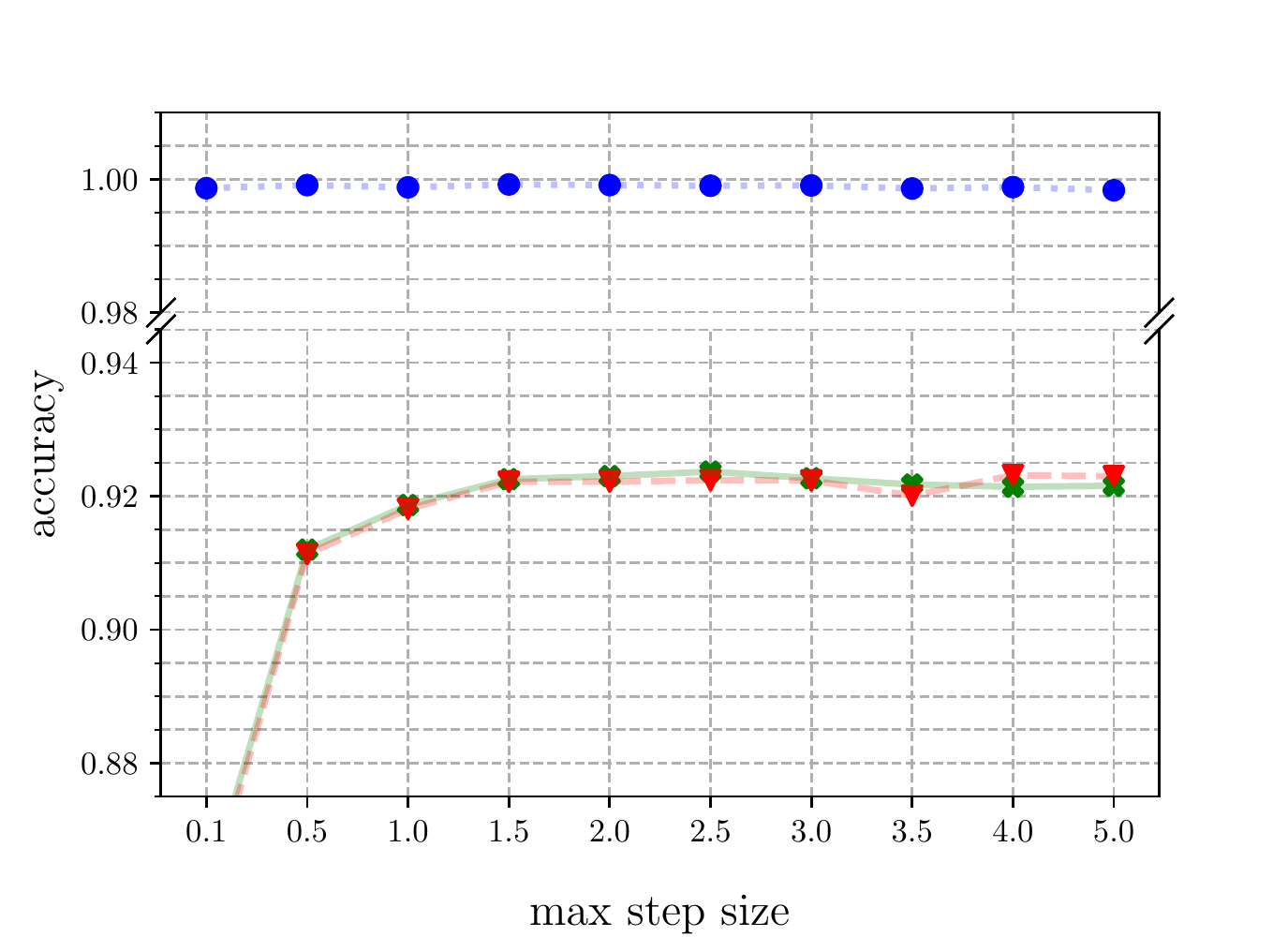}}\\
		\scalebox{\scale}{\includegraphics{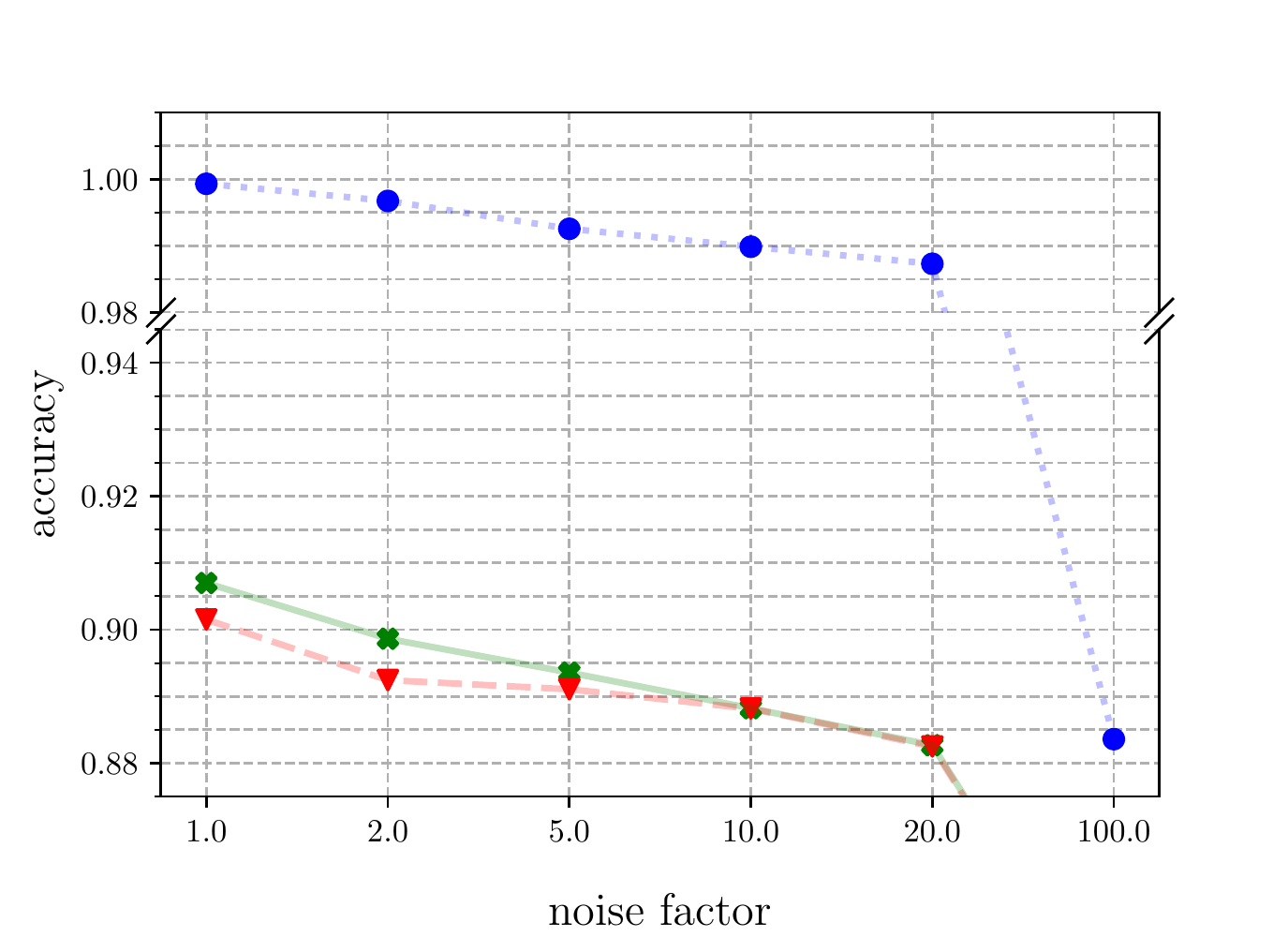}}&	\scalebox{\scale}{\includegraphics{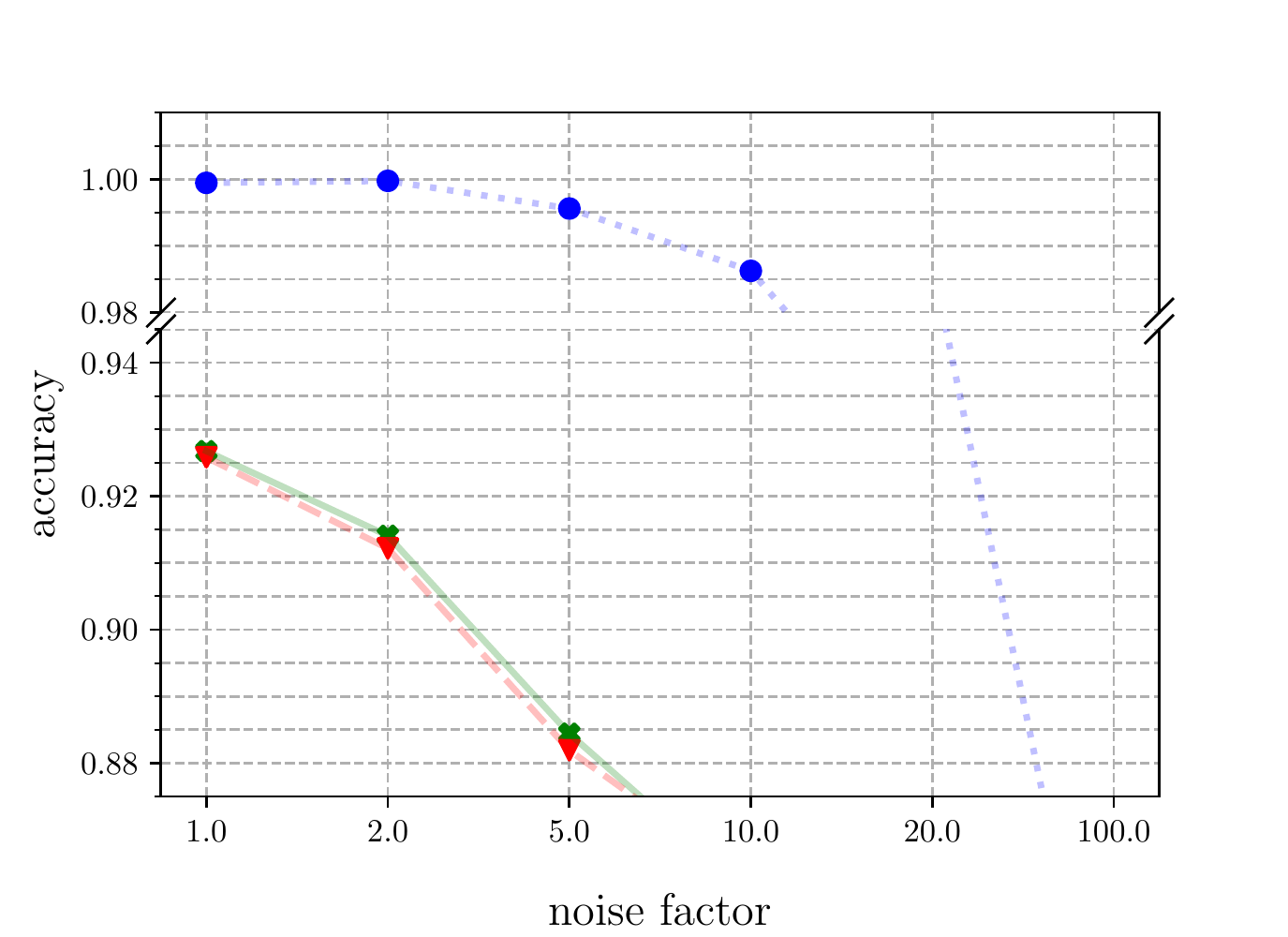}}&
		\scalebox{\scale}{\includegraphics{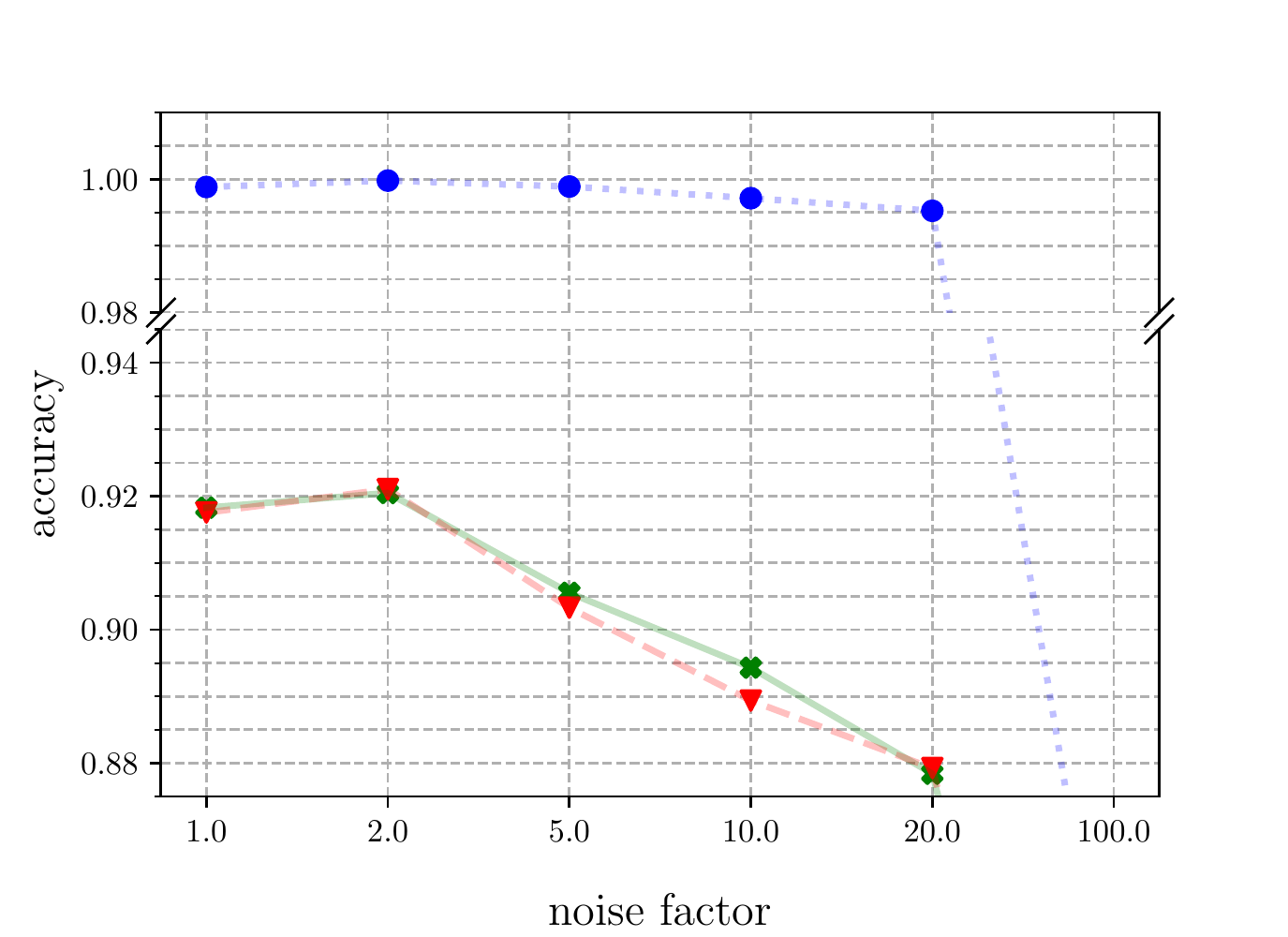}}
	\end{tabular}
	\caption{Sensitivity analysis of parameters of \textbf{LABPAL\&NSGD}. The default parameters are: approximation batch size $\mathbb{B}_a = 1280$, SGD steps $s=1000$, step size adaptation $\alpha = 1.8$, batch size schedule $k=$ ({0:1, 75000:2, 112500:4}), momentum $\beta=0$, maximal step size $=1.0$, noise-factor $\epsilon=1$. For $\mathbb{B}_a$ the factor 128 is multiplied with is given on the x axis.}
	\label{labpal_fig_sensitivity_analysis_lappalnsgd}
\end{figure}

%% file: labpal/performance_comparison_cifar10.tex
\begin{figure}[h!]
	\tikzsetfigurename{labpal_optimizer_comparison_cifar10}
	\centering
	\def\scale{0.4}

	\begin{tabular}{ c c c}	
		\scalebox{\scale}{\input{"labpal/figure_data/performance_comparison/CIFAR-10/CIFAR-10_ResNet-20_learning_rate.pgf"}}&
		\scalebox{\scale}{\input{"labpal/figure_data/performance_comparison/CIFAR-10/CIFAR-10_MobileNet-V2_learning_rate.pgf"}}&
		\scalebox{\scale}{\input{"labpal/figure_data/performance_comparison/CIFAR-10/CIFAR-10_DenseNet-121_learning_rate.pgf"}}\\
		\scalebox{\scale}{\input{"labpal/figure_data/performance_comparison/CIFAR-10/CIFAR-10_ResNet-20_training_loss.pgf"}}&
		\scalebox{\scale}{\input{"labpal/figure_data/performance_comparison/CIFAR-10/CIFAR-10_MobileNet-V2_training_loss.pgf"}}&
		\scalebox{\scale}{\input{"labpal/figure_data/performance_comparison/CIFAR-10/CIFAR-10_DenseNet-121_training_loss.pgf"}}\\
		\scalebox{\scale}{\input{"labpal/figure_data/performance_comparison/CIFAR-10/CIFAR-10_ResNet-20_validation_accuracy.pgf"}}&
		\scalebox{\scale}{\input{"labpal/figure_data/performance_comparison/CIFAR-10/CIFAR-10_MobileNet-V2_validation_accuracy.pgf"}}&
		\scalebox{\scale}{\input{"labpal/figure_data/performance_comparison/CIFAR-10/CIFAR-10_DenseNet-121_validation_accuracy.pgf"}}\\
		\scalebox{\scale}{\input{"labpal/figure_data/performance_comparison/CIFAR-10/CIFAR-10_ResNet-20_test_accuracy.pgf"}}&
		\scalebox{\scale}{\input{"labpal/figure_data/performance_comparison/CIFAR-10/CIFAR-10_MobileNet-V2_test_accuracy.pgf"}}&
		\scalebox{\scale}{\input{"labpal/figure_data/performance_comparison/CIFAR-10/CIFAR-10_DenseNet-121_test_accuracy.pgf"}}
	\end{tabular}
		\caption{Performance comparison on \textbf{CIFAR-10} of our approach LABPAL in the SGD and NSGD variants against several line searches and SGD. Optimal hyperparameters are found with an elaborate grid search. Our approaches challenge and sometimes outperform the other approaches on training loss, validation, and test accuracy. Columns indicate different models. Rows indicate different metrics.} 
	\label{labpal_fig_optimizer_comparison_cifar10}
\end{figure}

%% file: labpal/figure_data/performance_comparison/CIFAR-10/CIFAR-10_DenseNet-121_training_loss.pgf
% This file was created by tikzplotlib v0.9.8.
\begin{tikzpicture}

\definecolor{color0}{rgb}{0.647058823529412,0.164705882352941,0.164705882352941}
\definecolor{color1}{rgb}{0.933333333333333,0.509803921568627,0.933333333333333}
\definecolor{color2}{rgb}{1,0.647058823529412,0}

\begin{axis}[
grid = major,
major grid style={dotted},
log basis y={10},
minor xtick={},
minor ytick={2e-05,3e-05,4e-05,5e-05,6e-05,7e-05,8e-05,9e-05,0.0002,0.0003,0.0004,0.0005,0.0006,0.0007,0.0008,0.0009,0.002,0.003,0.004,0.005,0.006,0.007,0.008,0.009,0.02,0.03,0.04,0.05,0.06,0.07,0.08,0.09,0.2,0.3,0.4,0.5,0.6,0.7,0.8,0.9,2,3,4,5,6,7,8,9,20,30,40,50,60,70,80,90,200,300,400,500,600,700,800,900},
tick align=outside,
tick pos=left,
title={training loss CIFAR-10 DenseNet-121},
width=10.5cm,height=8cm,,
x grid style={white!69.0196078431373!black},
xlabel={epochs},
xmin=-14.2, xmax=298.2,
xtick style={color=black},
xtick={-50,0,50,100,150,200,250,300},
y grid style={white!69.0196078431373!black},
ylabel={training loss},
ymin=0.0001, ymax=6.0,
ymode=log,
ytick style={color=black},
ytick={1e-05,0.0001,0.001,0.01,0.1,1,10,100}
]
\addplot [line width=2.5pt, color0, opacity=1.0]
table {%
0 2.34695760409037
1 1.61670561631521
2 1.19927632808685
3 0.995419283707937
4 0.857371926307678
5 0.755861540635427
6 0.682371338208516
7 0.622405608495077
8 0.573267797629038
9 0.526595850785573
10 0.491704046726227
11 0.46042142311732
12 0.430224825938543
13 0.408631483713786
14 0.385092775026957
15 0.361077318588893
16 0.345200598239899
17 0.325146694978078
18 0.308075179656347
19 0.288648039102554
20 0.278831551472346
21 0.262620329856873
22 0.247196068366369
23 0.240302314360936
24 0.225005318721135
25 0.216889744003614
26 0.206298296650251
27 0.191812992095947
28 0.186614056428274
29 0.179475689927737
30 0.169556096196175
31 0.162781938910484
32 0.1548197666804
33 0.144019683202108
34 0.141497487823168
35 0.132500559091568
36 0.128478323419889
37 0.125612316032251
38 0.122204797963301
39 0.111584963897864
40 0.107389137148857
41 0.101941178242366
42 0.106058364113172
43 0.154278039932251
44 0.132505014538765
45 0.123246086140474
46 0.11836301535368
47 0.109126754105091
48 0.106994032859802
49 0.100624320407708
50 0.146054970721404
51 0.133867345750332
52 0.12467618783315
53 0.117097735404968
54 0.10781749834617
55 0.0990608260035515
56 0.136338969071706
57 0.125689590970675
58 0.113712139427662
59 0.130357588330905
60 0.124777540564537
61 0.111934795975685
62 0.1078750466307
63 0.0949756950139999
64 0.0937402000029882
65 0.0879919032255809
66 0.0962531318267186
67 0.0974230021238327
68 0.0910681163271268
69 0.0903154412905375
70 0.102798911432425
71 0.0922861744960149
72 0.0903032099207242
73 0.0829952384034793
74 0.157736892501513
75 0.129734625418981
76 0.132619388401508
77 0.122776016592979
78 0.125701149304708
79 0.113198491434256
80 0.134758283694585
81 0.121609029670556
82 0.109016291797161
83 0.102936362226804
84 0.0995136375228564
85 0.118207807342211
86 0.106132045388222
87 0.100456928213437
88 0.0915524090329806
89 0.0889196172356606
90 0.080897939701875
91 0.123234927654266
92 0.10447808355093
93 0.0979158009092013
94 0.0862654608984788
95 0.114300658305486
96 0.100867807865143
97 0.104870175321897
98 0.107923448085785
99 0.0961489031712214
100 0.0915624101956685
101 0.0879355495174726
102 0.12368256598711
103 0.0891300191481908
104 0.119635790586472
105 0.109086098770301
106 0.0995021189252536
107 0.101501243809859
108 0.104123217364152
109 0.092822772761186
110 0.0892617404460907
111 0.0831430479884148
112 0.0735074120263259
113 0.0690124407410622
114 0.0660090235372384
115 0.270521031071742
116 0.150966294109821
117 0.089667696505785
118 0.111436385661364
119 0.104990196724733
120 0.091080995897452
121 0.0900647093852361
122 0.081861600279808
123 0.0738481506705284
124 0.086046372850736
125 0.0822304040193558
126 0.077351450920105
127 0.0718776434659958
128 0.0706965699791908
129 0.0632382209102313
130 0.0578673196335634
131 0.0613072973986467
132 0.0590691852072875
133 0.0622054052849611
134 0.0589506675799688
135 0.0611302219331264
136 0.0791350205739339
137 0.0720265880227089
138 0.0761150841911634
139 0.0687782590587934
140 0.077295313278834
141 0.07960128535827
142 0.0790636340777079
143 0.0957869316140811
144 0.0834236492713292
145 0.076685257256031
146 0.087392528851827
147 0.0831173310677211
148 0.0720196440815926
149 0.0710298344492912
150 0.0830855369567871
151 0.123527482151985
152 0.123548338810603
153 0.1178906361262
154 0.10280787696441
155 0.0931106035908063
156 0.0883714656035105
157 0.0806499992807706
158 0.0868047848343849
159 0.114767737686634
160 0.0981281648079554
161 0.0899501467744509
162 0.0832449421286583
163 0.0815820284187794
164 0.0744682426253955
165 0.0850794166326523
166 0.0898730456829071
167 0.0803118323286374
168 0.0798205062747002
169 0.0735635980963707
170 0.0670453384518623
171 0.0627630054950714
172 0.0612507537007332
173 0.0623983765641848
174 0.0619048985342185
175 0.0622859783470631
176 0.0628923289477825
177 0.0892218599716822
178 0.0871841448048751
179 0.0858069583773613
180 0.0885840356349945
181 0.0795708944400152
182 0.0794829751054446
183 0.09866780291001
184 0.113974931339423
185 0.10432992130518
186 0.0984179029862086
187 0.0911060397823652
188 0.0829811319708824
189 0.0833853930234909
190 0.0750036984682083
191 0.0750023623307546
192 0.0717116792996724
193 0.0749442800879478
194 0.101276357968648
195 0.0946644221742948
196 0.100062722961108
197 0.0923024863004684
198 0.090074627349774
199 0.0875388905405998
200 0.0838764856259028
201 0.0915000289678574
202 0.0956261505683263
203 0.127316298584143
204 0.104107491672039
205 0.100925736129284
206 0.0943741674224536
207 0.0891385575135549
208 0.0847446918487549
209 0.0818626806139946
210 0.0778487647573153
211 0.0711166659990946
212 0.0947618260979652
213 0.110802285373211
214 0.112884496649106
215 0.11699253693223
216 0.10793857773145
217 0.103789983938138
218 0.0979653149843216
219 0.10252495855093
220 0.0956149473786354
221 0.0956241562962532
222 0.0937127744158109
223 0.0880430191755295
224 0.0839834933479627
225 0.0813831202685833
226 0.0751950691143672
227 0.0707445243994395
228 0.0741705944140752
229 0.210501273473104
230 0.116746673981349
231 0.147171219189962
232 0.115816608071327
233 0.110093819598357
234 0.121889670689901
235 0.149600305904945
236 0.125386137515306
237 0.121277816593647
238 0.111958588163058
239 0.124116616944472
240 0.140002583463987
241 0.521917395293713
242 22590462634.7561
243 2883307874986.75
244 33.4154546434681
245 33.4080274154743
246 33.4111053322752
247 33.4061459501584
248 33.3979019472996
};
\addplot [line width=2.5pt, green!50.1960784313725!black, opacity=1.0]
table {%
0 2.34695752461751
1 1.64194512367249
2 1.1509123245875
3 0.961859087149302
4 0.794875999291738
5 0.688123603661855
6 0.631134490172068
7 0.551484982172648
8 0.497089942296346
9 0.454014360904694
10 0.420146117607753
11 0.388490508000056
12 0.361743152141571
13 0.340387254953384
14 0.313789894183477
15 0.296213209629059
16 0.27774781982104
17 0.259652058283488
18 0.246247296531995
19 0.231645335753759
20 0.220893050233523
21 0.205808331569036
22 0.193590973814329
23 0.18312269449234
24 0.1729563275973
25 0.16768904030323
26 0.15443366765976
27 0.146252155303955
28 0.137978598475456
29 0.131219844023387
30 0.12446295718352
31 0.118280306458473
32 0.111689512928327
33 0.108029708266258
34 0.103176807363828
35 0.0951915656526883
36 0.0915550614396731
37 0.0876224438349406
38 0.087183286746343
39 0.0815805643796921
40 0.0776084288954735
41 0.073661744594574
42 0.0714776342113813
43 0.066740870475769
44 0.0635172439118226
45 0.063656210899353
46 0.059149693697691
47 0.056891872237126
48 0.0548096237083276
49 0.0517667680978775
50 0.0509260358909766
51 0.0499249498049418
52 0.0467126456399759
53 0.0449196919798851
54 0.0454605606695016
55 0.0425910949707031
56 0.0433051722745101
57 0.0406255573034286
58 0.038187064230442
59 0.0395916079481443
60 0.0365971997380257
61 0.0369105910261472
62 0.0352985002100468
63 0.0334564360479514
64 0.0327757646640142
65 0.032000258564949
66 0.0291535096863906
67 0.0301937765131394
68 0.0293124398837487
69 0.0293016911794742
70 0.0295891432712475
71 0.0261472972730796
72 0.0284498712668816
73 0.0256562276432912
74 0.0242047961801291
75 0.026793818299969
76 0.02476933101813
77 0.0246293476472298
78 0.0234149650981029
79 0.0222528384377559
80 0.0228337186078231
81 0.0233948361128569
82 0.021747650578618
83 0.0200971166292826
84 0.0204570833593607
85 0.018614146237572
86 0.0195495504885912
87 0.0195313369234403
88 0.0186421126127243
89 0.0170876746997237
90 0.0188344170649846
91 0.016531340777874
92 0.0162294044469794
93 0.0179234097401301
94 0.0156176978101333
95 0.0149760923037926
96 0.0170944202691317
97 0.0164478905498981
98 0.0169849048058192
99 0.015080274703602
100 0.0141409079854687
101 0.0139654045924544
102 0.0138806353012721
103 0.0143749359995127
104 0.0137342022111019
105 0.0143753591304024
106 0.014011489537855
107 0.012435607612133
108 0.0121545232832432
109 0.0115566380942861
110 0.0147370525325338
111 0.011569704550008
112 0.0104905090605219
113 0.0123758753761649
114 0.012030658001701
115 0.0114271289979418
116 0.0124869756400585
117 0.0107700498774648
118 0.0118958012511333
119 0.0126533918082714
120 0.0107163482656082
121 0.0117197722817461
122 0.0120365352680286
123 0.011177602224052
124 0.0104494709521532
125 0.0107106855139136
126 0.0092875015301009
127 0.00997892674058676
128 0.0109678820396463
129 0.0106207920859257
130 0.00829483894631267
131 0.0110636437311769
132 0.00850914356609186
133 0.0102380809063713
134 0.0092036387262245
135 0.00953644824524721
136 0.0119614905367295
137 0.00858464774986108
138 0.00790928474937876
139 0.00882081920281053
140 0.0077790132102867
141 0.00990133453160524
142 0.00895835304011901
143 0.00897396169602871
144 0.00358853062304358
145 0.00287719994472961
146 0.00235923061457773
147 0.00200313318055123
148 0.00202148908283561
149 0.00172102280581991
150 0.00153382844291627
151 0.00132061987339209
152 0.00118763442151248
153 0.00114929136664917
154 0.00124447218452891
155 0.00108504159531246
156 0.00111612374894321
157 0.000968767970334738
158 0.000938411802053452
159 0.000874791119713336
160 0.000958457298111171
161 0.000852033728733659
162 0.00097758456831798
163 0.000857953797094524
164 0.000807858731908103
165 0.000756655160027246
166 0.000722883210983127
167 0.000785900842553625
168 0.000719987690293541
169 0.00064802528747047
170 0.000749692922302832
171 0.000536832861447086
172 0.0006351320965526
173 0.000620281284985443
174 0.000658620129494617
175 0.000649949516324947
176 0.000563545909244567
177 0.000533683603862301
178 0.000533934614698713
179 0.000722033728379756
180 0.000629231838199
181 0.00058834506974866
182 0.000728147919289768
183 0.000504119729157537
184 0.000551933713722974
185 0.000638714816886932
186 0.000551347581980129
187 0.000446874221476416
188 0.000578804756514728
189 0.000602120281352351
190 0.000528838951140642
191 0.00041987753744858
192 0.000548311353971561
193 0.00051467137139601
194 0.000507109808192278
195 0.000392818047354619
196 0.000432756797332938
197 0.000432611714738111
198 0.000526006556659316
199 0.000357973399028803
200 0.000417747670629372
201 0.000404617972283935
202 0.000346036123422285
203 0.000372053143413117
204 0.000387641688575968
205 0.000441021955339238
206 0.000320802874436292
207 0.000413403759011999
208 0.000332120417927702
209 0.000363555562216789
210 0.000449754870108639
211 0.000374933801746617
212 0.000361249520210549
213 0.000357689365046099
214 0.000399175216443837
215 0.000378538767108694
216 0.000299240520689636
217 0.000292782778463637
218 0.000426335725933313
219 0.000398253789171576
220 0.000313853558812601
221 0.000394154980313033
222 0.000361346998640026
223 0.000295284233288839
224 0.000313442151915903
225 0.000307666341541335
226 0.000298314495012164
227 0.000357951367429147
228 0.000331676448695362
229 0.000357777035484711
230 0.000402678648242727
231 0.000366822661211093
232 0.000312890226875121
233 0.000303219625493512
234 0.000378196438153585
235 0.000315678335027769
236 0.000367405475117266
237 0.000339203582067663
238 0.000255906925303861
239 0.000416164208824436
240 0.000373699207557365
241 0.000265280318368847
242 0.000372610966830204
243 0.000403953813171635
244 0.000383138074539602
245 0.000347214450205987
246 0.000276882211134459
247 0.00027880606163914
248 0.000286440554191358
249 0.000287826483448346
250 0.000262687928625382
251 0.000290635041892529
252 0.000337935809511691
253 0.000348670514843737
254 0.000296711640354867
255 0.000306231871945783
256 0.000378347894487282
257 0.00032711877914456
258 0.000346968959396084
259 0.000367473694495857
260 0.000266244896920398
261 0.000341377347164477
262 0.000260099540658606
263 0.000382011019003888
264 0.000302306827506982
265 0.000318952331629892
266 0.000324393360642716
267 0.000267953020132457
268 0.000315039001482849
269 0.000312844999522592
270 0.000372234993847087
271 0.000326882252314438
272 0.000326375302392989
273 0.000409121237074335
274 0.000333301364056145
275 0.000302832166198641
276 0.000292918014262492
277 0.000266851614772653
278 0.00033049385334986
279 0.000336100580170751
280 0.000308548129396513
281 0.000261540301532174
282 0.000308356567984447
283 0.000342894073886176
284 0.000307298837772881
};
\addplot [line width=2.5pt, black, opacity=1.0]
table {%
0 2.34695752461751
1 1.82345374425252
2 1.59911155700684
3 1.42445194721222
4 1.24658079942067
5 1.09002161026001
6 0.958185871442159
7 0.849413494269053
8 0.764978567759196
9 0.745026171207428
10 0.692104240258535
11 0.59488449494044
12 0.591902732849121
13 0.589233299096425
14 0.539422820011775
15 0.49745370944341
16 0.515173504749934
17 0.469891160726547
18 0.463972588380178
19 0.485155522823334
20 0.498765856027603
21 0.481402297814687
22 0.440083901087443
23 0.426175465186437
24 0.371835430463155
25 0.381867289543152
26 0.349536329507828
27 0.33721591035525
28 0.325053364038467
29 0.309637308120728
30 0.338938782612483
31 0.365802139043808
32 0.309261411428452
33 0.307691688338916
34 0.304262121518453
35 0.303152501583099
36 0.263470366597176
37 0.261380662520727
38 0.288236717383067
39 0.314414282639821
40 0.306323230266571
41 0.289509803056717
42 0.252396017313004
43 0.215565159916878
44 0.216710771123568
45 0.236695751547813
46 0.245687291026115
47 0.223842536409696
48 0.238374004761378
49 0.26176606118679
50 0.257053509354591
51 0.272753790020943
52 0.242992535233498
53 0.249940107266108
54 0.195297832290332
55 0.152646780014038
56 0.155114114284515
57 0.172613064448039
58 0.178240497907003
59 0.166692917545637
60 0.190197249253591
61 0.237194533149401
62 0.208248575528463
63 0.185297176241875
64 0.209370464086533
65 0.204912985364596
66 0.185692553718885
67 0.177938312292099
68 0.149431829651197
69 0.164830520749092
70 0.191560094555219
71 0.194708272814751
72 0.207053576906522
73 0.158942033847173
74 0.15357168763876
75 0.173717553416888
76 0.193631075322628
77 0.196993467708429
78 0.212052551408609
79 0.20868069678545
80 0.163565874099731
81 0.14874163766702
82 0.144333203633626
83 0.128549225628376
84 0.129854641854763
85 0.145244414607684
86 0.119753080109755
87 0.0911051332950592
88 0.0823802190522353
89 0.0835213884711266
90 0.0875262965758642
91 0.0843533972899119
92 0.0758422476549943
93 0.0787758603692055
94 0.0828317925333977
95 0.0703231866161029
96 0.0750406421720982
97 0.0750026317934195
98 0.0708786019434531
99 0.0578107324739297
100 0.0535003698120515
101 0.058630791803201
102 0.0504731815308332
103 0.0428874380886555
104 0.0363048644115527
105 0.0391043623288473
106 0.0305042968442043
107 0.0277070657660564
108 0.0269027755906185
109 0.0246415873989463
110 0.02039141052713
111 0.0173212876543403
112 0.0190047395105163
113 0.0182078738386432
114 0.0162963528806965
115 0.0155892549082637
116 0.0146409707764784
117 0.01419022710373
118 0.0147512126713991
119 0.0131227572758993
120 0.0120578855276108
121 0.0123425427203377
122 0.0117574110627174
123 0.010156343691051
124 0.0104312410888573
125 0.00865101255476475
126 0.00843578903004527
127 0.0090626726547877
128 0.0094874898592631
129 0.00951312885930141
130 0.00810794765129685
131 0.00782698160037398
132 0.00846714293584228
133 0.00666349567472935
134 0.00767725535357992
135 0.00736856398483117
136 0.00708589935675263
137 0.00636360685651501
138 0.00656318136801322
139 0.00731468014419079
140 0.00667708786204457
141 0.00724547713374098
142 0.00588674734657009
143 0.00625760154798627
144 0.00649642727027337
145 0.00603993299106757
146 0.00598691357299685
147 0.00567496195435524
148 0.00554493799184759
149 0.00617174478247762
150 0.00501328675697247
151 0.00514019172017773
152 0.00531880091875792
153 0.00482540701826413
154 0.00587345085417231
155 0.00543802402292689
156 0.00502157816663384
157 0.00529617490246892
158 0.00449801896077891
159 0.00469885223234693
160 0.00458430343617996
161 0.00446032080799341
162 0.00438866298645735
163 0.00484328142677744
164 0.00441715695584814
165 0.0047023817896843
166 0.0038168605727454
167 0.00402697618119419
168 0.00452059507369995
169 0.00396510648230712
170 0.00367390162621935
171 0.00386442748519282
172 0.0036255131320407
173 0.003166544639195
174 0.00335782176504532
175 0.00296104365649323
176 0.00330474049163361
177 0.00370625220239162
178 0.00355703833823403
179 0.00389925336154799
180 0.00361614309561749
181 0.00381724451047679
182 0.00299656655018528
183 0.00335423143890997
184 0.00312544327850143
185 0.00323695309149722
186 0.00285132182762027
187 0.00376840176371237
188 0.00344962820721169
189 0.00332293255875508
190 0.00327132168846826
191 0.00313392677344382
192 0.00299555125335852
193 0.00348262760477761
194 0.0030828946425269
195 0.00323302205651999
196 0.00323386746458709
197 0.00342938893785079
198 0.00314744507583479
199 0.00313693665278455
200 0.00294145006531229
201 0.00228457513730973
202 0.00239735927122335
203 0.00292230125827094
204 0.00245057178350786
205 0.00312215563220282
206 0.00280346240227421
207 0.00254607604195674
208 0.00277282382982473
209 0.0029221340858688
210 0.00284999779736002
211 0.00208732450846583
212 0.00272767016819368
213 0.00239453953690827
214 0.00285175574632982
215 0.00290482312751313
216 0.00273915562623491
217 0.00250575020133207
218 0.00200936799713721
219 0.00248591982138654
220 0.00246242554082225
221 0.00274053860145311
222 0.0024421406754603
223 0.00222180147344867
224 0.00208003035125633
225 0.00222948527274032
226 0.00205849545697371
227 0.00231882246832053
228 0.00240559596568346
229 0.00208274414762855
230 0.0021381708405291
231 0.00222033165240039
232 0.00195911433547735
233 0.00201962445862591
234 0.00203539011999965
235 0.00199163301537434
236 0.00224598097459724
237 0.00221696953910092
238 0.00206269281140218
239 0.00210342284602424
240 0.00195034527375052
241 0.00180821764903764
242 0.00216910064530869
243 0.00206653451702247
244 0.00197046253985415
245 0.00234868501623472
246 0.00184476127227147
247 0.00176954773875574
248 0.00178260720955829
249 0.00216215034015477
250 0.00200862150328855
251 0.00172537231507401
252 0.00180183331637333
253 0.00182537513319403
254 0.00162575056310743
255 0.00166733834582071
256 0.00176631655388822
257 0.00197456191138675
258 0.0018579267586271
259 0.00211219703002522
260 0.00181602158894142
261 0.00203798885922879
262 0.00202658477549752
263 0.00158468439864616
264 0.00163011943611006
265 0.00178273996183028
266 0.00134769085949908
267 0.00176443229429424
268 0.00158713109946499
269 0.00183070237593104
270 0.00171518597441415
271 0.00170562940184027
272 0.0017597833648324
273 0.00156562278668086
274 0.00152700223649542
275 0.00150112889241427
276 0.00170957483351231
277 0.00161084028271337
278 0.00166192476172
279 0.00140554888639599
280 0.0013462845236063
281 0.00161147067168107
282 0.00164477065360794
283 0.00141948640036086
284 0.00175456947181374
};
\addplot [line width=2.5pt, color1, opacity=1.0]
table {%
0 2.34695752461751
1 2.19595630963643
2 1.65167633692424
3 1.43805464108785
4 1.2901363770167
5 1.18117833137512
6 1.09456916650136
7 1.02899158000946
8 0.974132001399994
9 0.918973008791606
10 0.848468005657196
11 0.798413932323456
12 0.747447729110718
13 0.722701489925385
14 0.678243656953176
15 0.64038473367691
16 0.630048990249634
17 0.595372537771861
18 0.573260148366292
19 0.555262784163157
20 0.534680386384328
21 0.513893713553747
22 0.49622564514478
23 0.488286177317301
24 0.470440169175466
25 0.453517337640127
26 0.445520063241323
27 0.435561656951904
28 0.41518830259641
29 0.409812966982524
30 0.400405367215474
31 0.381319791078568
32 0.37553882598877
33 0.359581838051478
34 0.358547657728195
35 0.347768406073252
36 0.338793307542801
37 0.336244364579519
38 0.318978170553843
39 0.310506095488866
40 0.308612863222758
41 0.307746340831121
42 0.288035303354263
43 0.287819743156433
44 0.28316127260526
45 0.272587577501933
46 0.270397315422694
47 0.263754268487294
48 0.259938458601634
49 0.254519442717234
50 0.244590764244397
51 0.243871172269185
52 0.23988597591718
53 0.233113616704941
54 0.227610309918722
55 0.223089238007863
56 0.219252412517866
57 0.211276014645894
58 0.208621631066004
59 0.201892619331678
60 0.201224848628044
61 0.196857293446859
62 0.19880399107933
63 0.191447978218397
64 0.189759105443954
65 0.180367658535639
66 0.178507377703985
67 0.176683699091276
68 0.173748527963956
69 0.165898531675339
70 0.167061602075895
71 0.163301095366478
72 0.160864348212878
73 0.162221401929855
74 0.154137363036474
75 0.154041469097137
76 0.148598944147428
77 0.149453848600388
78 0.143635114034017
79 0.141159906983376
80 0.139740467071533
81 0.142415821552277
82 0.137249072392782
83 0.133051678538322
84 0.13128166894118
85 0.128862197200457
86 0.128472775220871
87 0.128935133417447
88 0.125484727323055
89 0.117455152173837
90 0.121199771761894
91 0.116419444481532
92 0.114794358611107
93 0.113960007826487
94 0.114316309491793
95 0.114167682826519
96 0.10906845331192
97 0.108859951297442
98 0.106040718654792
99 0.101766422390938
100 0.102458740274111
101 0.104257054626942
102 0.0995381946365039
103 0.096317616601785
104 0.0984761193394661
105 0.0933503583073616
106 0.0949716046452522
107 0.0936186288793882
108 0.0924728612105052
109 0.0913538957635562
110 0.0893045788009961
111 0.0914759958783786
112 0.0910603553056717
113 0.0864450807372729
114 0.088465874393781
115 0.0870844026406606
116 0.0859138444066048
117 0.0842491686344147
118 0.0846422786513964
119 0.0807591329018275
120 0.0829070582985878
121 0.0804936066269875
122 0.0788850411772728
123 0.0787951697905858
124 0.0777422984441121
125 0.0751846159497897
126 0.0715375145276387
127 0.075435571372509
128 0.0767374038696289
129 0.0765308067202568
130 0.0758824124932289
131 0.0716829945643743
132 0.0739890312155088
133 0.0702071686585744
134 0.0700585593779882
135 0.0680197924375534
136 0.0682478820284208
137 0.0686228921016057
138 0.0693467780947685
139 0.0643770371874173
140 0.0665437926848729
141 0.0682966262102127
142 0.0686891923348109
143 0.0642033020655314
144 0.0645834356546402
145 0.0641228469709555
146 0.0667043477296829
147 0.0647316053509712
148 0.0624186731874943
149 0.0623607958356539
150 0.0621809400618076
151 0.0602481563886007
152 0.0601655431091785
153 0.0610088321069876
154 0.0608719239632289
155 0.0578162868817647
156 0.0602363099654516
157 0.0554840875168641
158 0.0579313337802887
159 0.0585478643576304
160 0.0584606441358725
161 0.0560579101244609
162 0.0562082131703695
163 0.0551194623112679
164 0.0533269097407659
165 0.0553956453998884
166 0.0538077503442764
167 0.0530851831038793
168 0.0542771182954311
169 0.052419060220321
170 0.0511849808196227
171 0.0527241826057434
172 0.0509484993914763
173 0.0519010735054811
174 0.0508149402836959
175 0.0512969096501668
176 0.0491279177367687
177 0.051031693816185
178 0.0489634697635969
179 0.0500495557983716
180 0.0472651446859042
181 0.0505516442159812
182 0.0502667563656966
183 0.0510893252988656
184 0.050665241976579
185 0.0458824336528778
186 0.0464688166975975
187 0.0451569780707359
188 0.0471451121071974
189 0.045541883756717
190 0.0454164321223895
191 0.0472323670983315
192 0.0454010305305322
193 0.0420350854595502
194 0.047883577644825
195 0.0470189986129602
196 0.0446308180689812
197 0.0435674314697584
198 0.043092696617047
199 0.044013445576032
200 0.0441380639870962
201 0.0438161467512449
202 0.0426856279373169
203 0.0423152993122737
204 0.0422750214735667
205 0.0433291171987851
206 0.0419294486443202
207 0.042293260494868
208 0.0408034026622772
209 0.0413830876350403
210 0.0427561650673548
211 0.0395783483982086
212 0.041442501048247
213 0.0420176088809967
214 0.0411098922292391
215 0.0403282480935256
216 0.0404463075101376
217 0.0416526968280474
218 0.0380997198323409
219 0.0384937028090159
220 0.0377847142517567
221 0.0378321409225464
222 0.0379380757610003
223 0.0403548503915469
224 0.0387885409096877
225 0.0388381791611513
226 0.0374712472160657
227 0.0376583685477575
228 0.039574229468902
229 0.0384346420566241
230 0.0367445883651574
231 0.0362131459017595
232 0.0367796371380488
233 0.0387556379040082
234 0.0368944567938646
235 0.0350445533792178
236 0.0371953013042609
237 0.0352369658648968
238 0.0327605940401554
239 0.0354868893822034
240 0.0342364298800627
241 0.0365230652193228
242 0.0330400963624318
243 0.0335810296237469
244 0.0371833853423595
245 0.0361483804881573
246 0.0356861303249995
247 0.0345295642813047
248 0.0341202629109224
249 0.0359372409681479
250 0.0357759284476439
251 0.0361027630666892
252 0.0326286684721708
253 0.0360465881725152
254 0.0338160035510858
255 0.0349642646809419
256 0.0334167908877134
257 0.0344820364067952
258 0.0344395364324252
259 0.0340984513362249
260 0.032011770332853
261 0.0348161918421586
262 0.0328502704699834
263 0.0321158214161793
264 0.031302135437727
265 0.0319297909736633
266 0.0325109430899223
267 0.0321794437865416
268 0.031345467393597
269 0.0301467888057232
270 0.0308273086945216
271 0.029437342658639
272 0.0301803536713123
273 0.0293494326372941
274 0.0303298520545165
275 0.0302131213247776
276 0.0304459227869908
277 0.0317315465460221
278 0.0317954855660597
279 0.0277510397136211
280 0.0300246855864922
281 0.0305154155939817
282 0.0283165064950784
283 0.0278792704145114
284 0.0297048836946487
};
\addplot [line width=2.5pt, blue, opacity=1.0]
table {%
0 2.32015824317932
1 1.64786499738693
2 1.20469743013382
3 1.03877305984497
4 0.978794068098068
5 0.790702372789383
6 0.712385684251785
7 0.670787841081619
8 0.584952592849731
9 0.538325935602188
10 0.513512268662453
11 0.456485360860825
12 0.487303659319878
13 0.439590618014336
14 0.382244065403938
15 0.371026679873466
16 0.350400552153587
17 0.324302941560745
18 0.311842814087868
19 0.294823408126831
20 0.274795040488243
21 0.261736258864403
22 0.248483970761299
23 0.233341619372368
24 0.227266438305378
25 0.213816896080971
26 0.203177466988564
27 0.199194692075253
28 0.185153819620609
29 0.176152259111404
30 0.167607292532921
31 0.160211510956287
32 0.152088202536106
33 0.149091608822346
34 0.141095831990242
35 0.128755673766136
36 0.120393767952919
37 0.116566587239504
38 0.111135113984346
39 0.101596847176552
40 0.0988941453397274
41 0.0930339731276035
42 0.0842555575072765
43 0.0820641480386257
44 0.0878755562007427
45 0.0818915106356144
46 0.0735425241291523
47 0.0761022232472897
48 0.0735873579978943
49 0.0740173794329166
50 0.0642811320722103
51 0.0647030789405107
52 0.0633645709604025
53 0.057014461606741
54 0.057261573150754
55 0.0487558711320162
56 0.0504751689732075
57 0.0447826124727726
58 0.0388754326850176
59 0.0402266010642052
60 0.036677299067378
61 0.0354723315685987
62 0.0320414192974567
63 0.0352472402155399
64 0.0268665524199605
65 0.0263426853343844
66 0.0257008327171206
67 0.0222902009263635
68 0.0223377766087651
69 0.0211490951478481
70 0.0165945161134005
71 0.0150205222889781
72 0.0163638866506517
73 0.0117543404921889
74 0.01235072594136
75 0.0121657275594771
76 0.0101310783065856
77 0.0109986048191786
78 0.00912659708410501
79 0.00824518268927932
80 0.0068259104155004
81 0.00763638434000313
82 0.00612427806481719
83 0.00679265079088509
84 0.00586887798272073
85 0.00482346140779555
86 0.00584279745817184
87 0.00517078093253076
88 0.00512554636225104
89 0.00548319541849196
90 0.00439224240835756
91 0.00609100074507296
92 0.005782226100564
93 0.00599208078347147
94 0.005895861890167
95 0.00657698139548302
96 0.0042619458399713
97 0.00400872784666717
98 0.00450659357011318
99 0.00700613739900291
100 0.00713837204966694
101 0.012116901576519
102 0.0211287997663021
103 0.0165934092365205
104 0.073401500005275
105 0.0523482849821448
106 0.0487296031787992
107 0.0499788764864206
108 0.0473002791404724
109 0.0471354741603136
110 0.0401018867269158
111 0.0375486765988171
112 0.0428262078203261
113 0.0487317191436887
114 0.0424179984256625
115 0.0402003177441657
116 0.0449312808923423
117 0.0381130152381957
118 0.0404871199280024
119 0.0391116929240525
120 0.034945769701153
121 0.0378725058399141
122 0.037153322249651
123 0.0338261746801436
124 0.0324563444592059
125 0.0331457355059683
126 0.0328682265244424
127 0.0346430782228708
128 0.0350017677992582
129 0.0315064219757915
130 0.0317153753712773
131 0.0298666497692466
132 0.0284482245333493
133 0.028948915656656
134 0.0237731714732945
135 0.0247880909591913
136 0.0214458787813783
137 0.0197745468467474
138 0.0221039028838277
139 0.0232694866135716
140 0.0213637263514102
141 0.0197383044287562
142 0.019524740986526
143 0.0147359820548445
144 0.00503220409154892
145 0.00379662308841944
146 0.00316473277052864
147 0.0028816144913435
148 0.00295353383990005
149 0.00167433923343197
150 0.00203595997299999
151 0.00186565739568323
152 0.0014621720474679
153 0.0013925401144661
154 0.00129806643235497
155 0.00121024789405055
156 0.00114505339297466
157 0.00116195806185715
158 0.00104384176665917
159 0.00118754300638102
160 0.00128107127966359
161 0.00493554183049127
162 0.0138493739650585
163 0.0122747436980717
164 0.00951648136833683
165 0.00950611743610352
166 0.00998603628249839
167 0.0096985554555431
168 0.00820652151014656
169 0.00704425311414525
170 0.00799315568292513
171 0.0105032315477729
172 0.0295402142219245
173 0.0279865590855479
174 0.0254639592021704
175 0.0242738449014723
176 0.0213362039066851
177 0.0213477928191423
178 0.0223943376913667
179 0.0268844561651349
180 0.0200825063511729
181 0.0206801877357066
182 0.0182581427507102
183 0.0191540024243295
184 0.0178290284238756
185 0.0191477625630796
186 0.0180413285270333
187 0.0228828759863973
188 0.01868881797418
189 0.0164790726266801
190 0.017541628330946
191 0.0184823414310813
192 0.0177973108366132
193 0.0144831147044897
194 0.0158432768657804
195 0.0157352141104639
196 0.0148282810114324
197 0.0153765385039151
198 0.0133883692324162
199 0.0129318507388234
200 0.014821648132056
201 0.0130933565087616
202 0.0151974060572684
203 0.0132193905301392
204 0.013048738706857
205 0.0142172989435494
206 0.019234768114984
207 0.0200164248235524
208 0.0189876989461482
209 0.0170253030955791
210 0.0146146505139768
211 0.0151501405052841
212 0.0174457849934697
213 0.016704517416656
214 0.0131045924499631
215 0.00523760030046105
216 0.00323615653906018
217 0.00277034053578973
218 0.00193431630032137
219 0.00191776524297893
220 0.00181923981290311
221 0.00145552220055833
222 0.00134722265647724
223 0.00159257953055203
224 0.00106579228304327
225 0.00120712257921696
226 0.00100873783230782
227 0.00121662038145587
228 0.00132761348504573
229 0.00129973574075848
230 0.00124403520021588
231 0.0010815110290423
232 0.000874268473125994
233 0.000714310706825927
234 0.000692561123287305
235 0.000503926392411813
236 0.000943468709010631
237 0.000831187178846449
238 0.000910817354451865
239 0.00103524344740435
240 0.00128105242038146
241 0.000930359004996717
242 0.00117660319665447
243 0.000994063215330243
244 0.000967314903391525
245 0.000818407075712457
246 0.000712510343873873
247 0.000895912467967719
248 0.00115868676220998
249 0.0015754361229483
250 0.00109960744157434
251 0.000725414865883067
252 0.00106095528462902
253 0.000736393325496465
254 0.00085816066712141
255 0.000870214600581676
256 0.00105114653706551
257 0.000642036029603332
258 0.000621201586909592
259 0.000515863328473642
260 0.000770845654187724
261 0.000847735820570961
262 0.000591809657635167
263 0.000708480656612664
264 0.000818832631921396
265 0.000890738243469968
266 0.000736289905034937
267 0.000507922391989268
268 0.000552418161532842
269 0.000636413489701226
270 0.000824229355202988
271 0.000679699674947187
272 0.00052154999866616
273 0.000430658546974882
274 0.000504828785778955
275 0.00053586924332194
276 0.000406242143071722
277 0.000674566545058042
278 0.000910832051886246
279 0.00072132857167162
280 0.000673046029987745
281 0.000587526941671968
282 0.000645366235403344
283 0.000622034829575568
284 0.000538208405487239
};
\addplot [line width=2.5pt, red, opacity=1.0]
table {%
0 2.34695752461751
1 1.60923250516256
2 1.22800918420156
3 1.05411914984385
4 1.01339737574259
5 0.855525374412537
6 0.768288890520732
7 0.742043574651082
8 0.645585278669993
9 0.588924447695414
10 0.555639266967773
11 0.501476993163427
12 0.467120806376139
13 0.446352223555247
14 0.411875466505686
15 0.393335580825806
16 0.377900878588359
17 0.352992445230484
18 0.340713262557983
19 0.322880526383718
20 0.302404959996541
21 0.293143699566523
22 0.275377551714579
23 0.26126566529274
24 0.249132956067721
25 0.23728820681572
26 0.221990242600441
27 0.210192466775576
28 0.198859229683876
29 0.187675808866819
30 0.176385248700778
31 0.168824990590413
32 0.161727413535118
33 0.154797747731209
34 0.147233078877131
35 0.139624600609144
36 0.135979597767194
37 0.13117291033268
38 0.122901516656081
39 0.1186581676205
40 0.112966902554035
41 0.105805757145087
42 0.105341918766499
43 0.0997704143325488
44 0.0995656574765841
45 0.0966412176688512
46 0.0936959733565648
47 0.0910370200872421
48 0.0877477129300435
49 0.0831879799564679
50 0.0803647090991338
51 0.0793308516343435
52 0.073483519256115
53 0.0714057559768359
54 0.0676461420953274
55 0.0630386956036091
56 0.0603957076867421
57 0.0578745404879252
58 0.0527722463011742
59 0.0507370233535767
60 0.0487131476402283
61 0.0436051574846109
62 0.0400574244558811
63 0.0414252392947674
64 0.0346453872819742
65 0.0332896523177624
66 0.0341682291279236
67 0.0300958659499884
68 0.0293470813582341
69 0.0270468611270189
70 0.0260787233710289
71 0.0237040395538012
72 0.0237457373489936
73 0.0239885902653138
74 0.0238428463538488
75 0.0226772328217824
76 0.0229600698997577
77 0.0234142175565163
78 0.0209782564391692
79 0.0191573646540443
80 0.0198567857344945
81 0.0185340810567141
82 0.018137751147151
83 0.0173034301648537
84 0.0150763106842836
85 0.0135403471067548
86 0.0154845137149096
87 0.0141592336197694
88 0.0143793802708387
89 0.0138033165906866
90 0.0121904763703545
91 0.0112728026385109
92 0.013377252034843
93 0.0142054778213302
94 0.0172376104940971
95 0.0149360792711377
96 0.0155491139739752
97 0.0137058896943927
98 0.0139416682844361
99 0.0136849842965603
100 0.0140951567639907
101 0.0141729296495517
102 0.0147509165108204
103 0.0129583105444908
104 0.0128018443162243
105 0.0127152812977632
106 0.0124521687005957
107 0.0146690020337701
108 0.0141483265906572
109 0.0129972845315933
110 0.0124394195154309
111 0.0118352004016439
112 0.0123856039717793
113 0.0124145457521081
114 0.0123206203182538
115 0.0105830254033208
116 0.0114502881964048
117 0.00926075394575795
118 0.00843102515985568
119 0.0080395988188684
120 0.00836996718620261
121 0.00948987637336055
122 0.00611859870453676
123 0.00700454988206426
124 0.00561684602871537
125 0.00505264522507787
126 0.00509854305225114
127 0.0050155750165383
128 0.00481926611003776
129 0.00539121241308749
130 0.00507299869786948
131 0.00401710936178764
132 0.00308902169732998
133 0.00319229547555248
134 0.00379813889351984
135 0.0036980053409934
136 0.00309903380305817
137 0.00317660100214804
138 0.00302464173485835
139 0.0028950494597666
140 0.00330257877552261
141 0.00326766046540191
142 0.00312266302838301
143 0.00245091010583565
144 0.00287490685392792
145 0.00259218333909909
146 0.00276366126490757
147 0.00249422252333413
148 0.0017388224756966
149 0.00137983715588537
150 0.00178662581796137
151 0.00137640169123188
152 0.00118919095257297
153 0.000511034081379573
154 0.00090663135900589
155 0.00136286955482016
156 0.00104052209644578
157 0.000853704133381446
158 0.000879391824128106
159 0.000915313517907634
160 0.00074407701807407
161 0.000547111694080134
162 0.000834083058483278
163 0.00141622833325528
164 0.000598321620297308
165 0.000882595389460524
166 0.00114833484015738
167 0.000591970980167389
168 0.000688356260070577
169 0.000814218185648012
170 0.000439741682688085
171 0.000964709276255841
172 0.00156736007193103
173 0.00113051472483979
174 0.00107727873061473
175 0.00081218299116396
176 0.000896582593365262
177 0.00111413205740973
178 0.000563083667657338
179 0.000855598763640349
180 0.000764012065095206
181 0.000727816180490966
182 0.000540669435091938
183 0.000882776036936169
184 0.000580187634720157
185 0.000529161635010193
186 0.000812254666622418
187 0.000961870954294379
188 0.000906701437391651
189 0.000805272643143932
190 0.000650053218123503
191 0.000572622768231668
192 0.000640300172866167
193 0.000824125803774223
194 0.000735503155738115
195 0.000971590144520936
196 0.000929125827193881
197 0.00133957629926347
198 0.00067170437250752
199 0.00127732095036966
200 0.000607588027075205
201 0.000698438379913568
202 0.00111776735866442
203 0.00102157434836651
204 0.00150516513773861
205 0.000996933895900535
206 0.00107287170006506
207 0.00102841118738676
208 0.000691950767456243
209 0.000585794866007442
210 0.000453985914646182
211 0.00108701615439107
212 0.00106557074468583
213 0.000560618713886167
214 0.000662113850315412
215 0.000741139325934152
216 0.000794445756279553
217 0.000546942141985104
218 0.000581127053010277
219 0.000496537754467378
220 0.000196050081285648
221 0.000636985759986904
222 0.000237373880130084
223 0.000676115412109842
224 0.000811134309818347
225 0.000662811638903804
226 0.000757157627958804
227 0.000827247797739498
228 0.00124916152950997
229 0.000861860423659285
230 0.000984638376394287
231 0.0016692885546945
232 0.00142633459472563
233 0.000646996214830627
234 0.000520725225214846
235 0.00148185365348278
236 0.0011271235125605
237 0.000730824804729006
238 0.000674403807655229
239 0.000874574965564534
240 0.000529623849918911
241 0.00117079933246108
242 0.000874990175361745
243 0.00103334553568857
244 0.000750599535422225
245 0.00132951012831957
246 0.000508348729150991
247 0.000955724176795532
248 0.000948378544611235
249 0.000367645627799599
250 0.000951391077251174
251 0.00104982424212115
252 0.000682856812394069
253 0.000746439277463651
254 0.000668293287162669
255 0.000912281383837884
256 0.0014907809527358
257 0.0010039095262376
258 0.00132982949920309
259 0.000487128663129018
260 0.00112089256678397
261 0.000774798105339869
262 0.000980682397009029
263 0.00129532759941261
264 0.000503050134284422
265 0.00115847509005107
266 0.00120221361673127
267 0.00222528822875271
268 0.000991988252887192
269 0.000628438336813512
270 0.00115255366351145
271 0.00111146635996799
272 0.000743710620251174
273 0.00123916136120291
274 0.0009440481189813
275 0.00087859504719745
276 0.000969097764027538
277 0.00105785934094153
278 0.000926193490158767
279 0.00117509190749843
280 0.0010131985861032
281 0.00113521569680112
282 0.00147914377491058
283 0.001080388824145
284 0.000577264141005192
};
\addplot [line width=2.5pt, color2, opacity=1.0]
table {%
0 2.34695752461751
2 1.65748711427053
4 1.22534267107646
6 0.96406888961792
8 0.78715980052948
10 0.667544980843862
12 0.590535223484039
14 0.537873029708862
16 0.477516770362854
18 0.428540786107381
20 0.388660252094269
22 0.351737588644028
24 0.322697877883911
26 0.297877162694931
28 0.273334344228109
30 0.250598097840945
32 0.230103159944216
34 0.213660558064779
36 0.197703510522842
38 0.180394530296326
40 0.169610833128293
42 0.156736180186272
44 0.148392935593923
46 0.136519849300385
48 0.127737159530322
50 0.121657006442547
52 0.115414130191008
54 0.108224088946978
56 0.103828447560469
58 0.100140452384949
60 0.0931066522995631
62 0.0899229794740677
64 0.085261861483256
66 0.0814911350607872
68 0.0797899837295214
70 0.0750762522220612
72 0.0741956308484077
74 0.0716430420676867
76 0.0690462961792946
78 0.0653620834151904
80 0.065381333231926
82 0.0614294894039631
84 0.0618386566638947
86 0.0582856585582097
88 0.0564295500516891
90 0.0568576628963153
92 0.0548021234571934
94 0.0539080339173476
96 0.0525579291085402
98 0.0509329947332541
100 0.0481601233283679
102 0.0494169505933921
104 0.04894108325243
106 0.0476026795804501
108 0.0448345405360063
110 0.0452566432456175
112 0.0451063103973866
114 0.0441313982009888
116 0.0418651675184568
118 0.0422661950190862
120 0.0431486566861471
122 0.0421702899038792
124 0.0405742824077606
126 0.038886908441782
128 0.0384591991702716
130 0.0381126354138056
132 0.0380597027639548
134 0.0365442919234435
136 0.0368728339672089
138 0.0361761040985584
140 0.035907369107008
142 0.035524179538091
144 0.0359036599596341
146 0.0342334434390068
148 0.0345083363354206
150 0.0335538884003957
152 0.0338561882575353
154 0.0340376993020376
156 0.0330189963181814
158 0.0316871503988902
160 0.0305010036875804
162 0.0314124034096797
164 0.030896912018458
166 0.0323351509869099
168 0.0308072064071894
170 0.0309235534320275
172 0.0295262976239125
174 0.0299935583025217
176 0.030551511173447
178 0.0281363769123952
180 0.0299935123572747
182 0.0286314804106951
184 0.0274715181440115
186 0.0276969696084658
188 0.0280176022400459
190 0.0269152522087097
192 0.0277270525693893
194 0.026959599306186
196 0.0265907216817141
198 0.02689270551006
200 0.0275189491609732
202 0.0267932880669832
204 0.0257031929989656
206 0.0269127792368333
208 0.0268362741917372
210 0.0251944245149692
212 0.0253065551320712
214 0.0250348187983036
216 0.0248856066415707
218 0.0244843444476525
220 0.024456844975551
222 0.0238929794480403
224 0.0246765731523434
226 0.0238142032176256
228 0.0230344068259001
230 0.0241803911825021
232 0.0233700927346945
234 0.0226349445680777
236 0.0221508207420508
238 0.0230176380525033
240 0.023708808546265
242 0.0237101130187511
244 0.0220344495028257
246 0.0218420370171467
248 0.0226004365831614
250 0.0211063598593076
252 0.0220243769387404
254 0.0222140041490396
256 0.0213974341750145
258 0.0208871165911357
260 0.020510150740544
262 0.0208879597485065
264 0.0217269795636336
266 0.0207544676959515
268 0.022242151821653
270 0.0209704482307037
272 0.0209286858638128
274 0.0200732946395874
276 0.0209795068949461
278 0.0191986591865619
280 0.0207943047086398
282 0.0211553120364745
284 0.0212990405658881
};
\end{axis}

\end{tikzpicture}

%% file: labpal/figure_data/performance_comparison/CIFAR-10/CIFAR-10_DenseNet-121_validation_accuracy.pgf
% This file was created by tikzplotlib v0.9.8.
\begin{tikzpicture}

\definecolor{color0}{rgb}{0.647058823529412,0.164705882352941,0.164705882352941}
\definecolor{color1}{rgb}{0.933333333333333,0.509803921568627,0.933333333333333}
\definecolor{color2}{rgb}{1,0.647058823529412,0}

\begin{axis}[
grid = major,
major grid style={dotted},
minor xtick={},
%minor ytick={},
tick align=outside,
tick pos=left,
title={validation accuracy CIFAR-10 DenseNet-121},
width=10.5cm,height=8cm,,
x grid style={white!69.0196078431373!black},
xlabel={epochs},
xmin=-13.15, xmax=298.15,
xtick style={color=black},
xtick={-50,0,50,100,150,200,250,300},
y grid style={white!69.0196078431373!black},
ylabel={validation accuracy},
ymin=0.8, ymax=0.945,
ytick style={color=black},
minor y tick num=1
]
\addplot [line width=2.5pt, color0, opacity=1.0]
table {%
1 0.460536847511927
2 0.574519236882528
3 0.641960461934408
4 0.695045411586761
5 0.707398494084676
6 0.724692841370901
7 0.760817309220632
8 0.772035260995229
9 0.771100401878357
10 0.789997339248657
11 0.787459929784139
12 0.801949799060822
13 0.80795939763387
14 0.817307690779368
15 0.82545405626297
16 0.824519236882528
17 0.83019498984019
18 0.827056606610616
19 0.825787921746572
20 0.835803965727488
21 0.823651174704234
22 0.836538473765055
23 0.842280964056651
24 0.848357359568278
25 0.842013895511627
26 0.833533644676208
27 0.845419347286224
28 0.852764427661896
29 0.849492530028025
30 0.852430562178294
31 0.840544859568278
32 0.849759618441264
33 0.854634086290995
34 0.859041134516398
35 0.855235040187836
36 0.855034728844961
37 0.857905964056651
38 0.851896365483602
39 0.855635662873586
40 0.845819969971975
41 0.854433755079905
42 0.83653845389684
43 0.854033132394155
44 0.857238252957662
45 0.854700863361359
46 0.853165050347646
47 0.852564116319021
48 0.854500532150269
49 0.856370190779368
50 0.847489317258199
51 0.841012279192607
52 0.855902771155039
53 0.858106315135956
54 0.858373383680979
55 0.858907580375671
56 0.851228634516398
57 0.847355763117472
58 0.851896365483602
59 0.853699247042338
60 0.836071054140727
61 0.852831184864044
62 0.860977570215861
63 0.858573714892069
64 0.860042730967204
65 0.865251064300537
66 0.860243062178294
67 0.857238233089447
68 0.861912389596303
69 0.746928413709005
70 0.852897981802622
71 0.857839206854502
72 0.859041134516398
73 0.836271385351817
74 0.847689648469289
75 0.847622871398926
76 0.847355763117472
77 0.847756425539652
78 0.85423344373703
79 0.861912389596303
80 0.857572098573049
81 0.85423344373703
82 0.857638895511627
83 0.849158644676208
84 0.857171475887299
85 0.856637299060822
86 0.867454588413239
87 0.850694437821706
88 0.870259086290995
89 0.869057158629099
90 0.868122319380442
91 0.859842419624329
92 0.863247851530711
93 0.858640491962433
94 0.871794859568278
95 0.860510150591532
96 0.860710461934408
97 0.872863233089447
98 0.869190692901611
99 0.857505341370901
100 0.867053965727488
101 0.86204594373703
102 0.868589739004771
103 0.868456204732259
104 0.85423344373703
105 0.850560903549194
106 0.861177881558736
107 0.619991992910703
108 0.872262299060822
109 0.869324247042338
110 0.872195502122243
111 0.870726505915324
112 0.878672540187836
113 0.877804497877757
114 0.881677329540253
115 0.353766031563282
116 0.864516536394755
117 0.874666134516398
118 0.869658132394155
119 0.872128744920095
120 0.870058755079905
121 0.878672540187836
122 0.877804478009542
123 0.883947630723318
124 0.869858463605245
125 0.876736124356588
126 0.879473825295766
127 0.87439904610316
128 0.883947670459747
129 0.876201907793681
130 0.875934839248657
131 0.882411857446035
132 0.887620190779368
133 0.886418282985687
134 0.626201927661896
135 0.635950853427251
136 0.871127148469289
137 0.76315438747406
138 0.872662941614787
139 0.881343464056651
140 0.620726488530636
141 0.873397429784139
142 0.872996787230174
143 0.751736104488373
144 0.869391004244486
145 0.871394236882528
146 0.865384618441264
147 0.877804478009542
148 0.85403311252594
149 0.884348293145498
150 0.839610040187836
151 0.559294859568278
152 0.874732931454976
153 0.872262279192607
154 0.878739337126414
155 0.879073182741801
156 0.872796475887299
157 0.876602570215861
158 0.871928413709005
159 0.869724889596303
160 0.883346696694692
161 0.880008002122243
162 0.878004809220632
163 0.872662941614787
164 0.888287941614787
165 0.867053945859273
166 0.876669347286224
167 0.885616978009542
168 0.883947670459747
169 0.872863252957662
170 0.888888895511627
171 0.879874467849731
172 0.879607379436493
173 0.886418263117472
174 0.879941244920095
175 0.884281516075134
176 0.753138363361359
177 0.877403835455577
178 0.877537409464518
179 0.876469016075134
180 0.878739297389984
181 0.887419859568278
182 0.879807690779368
183 0.601161867380142
184 0.877604166666667
185 0.879273494084676
186 0.868456204732259
187 0.88054217894872
188 0.883146365483602
189 0.883012811342875
190 0.70506144563357
191 0.87954060236613
192 0.889489849408468
193 0.767160793145498
194 0.873597760995229
195 0.883146385351817
196 0.728632479906082
197 0.78331998984019
198 0.875
199 0.668402776122093
200 0.868790070215861
201 0.606503742436568
202 0.870125552018484
203 0.876068373521169
204 0.789797027905782
205 0.755542198816935
206 0.653712605436643
207 0.877203524112701
208 0.874532580375671
209 0.871127148469289
210 0.886017640431722
211 0.894431094328562
212 0.849091867605845
213 0.865584929784139
214 0.844951927661896
215 0.867254277070363
216 0.788528303305308
217 0.874532600243886
218 0.529513890544573
219 0.839409728844961
220 0.855635682741801
221 0.684428413709005
222 0.881209929784139
223 0.88735310236613
224 0.887820521990458
225 0.880809307098389
226 0.770365913709005
227 0.623864861826102
228 0.757545411586761
229 0.615117520093918
230 0.634682153662046
231 0.874131957689921
232 0.875934839248657
233 0.880876044432322
234 0.66559828321139
235 0.824986636638641
236 0.873731315135956
237 0.844217419624329
238 0.882478654384613
239 0.862847228844961
240 0.778445521990458
241 0.607572123408318
242 0.613782040774822
243 0.567240911225478
244 0.530849357446035
245 0.621861654023329
246 0.625801295042038
247 0.547475951413314
248 0.625066791971525
};
\addplot [line width=2.5pt, green!50.1960784313725!black, opacity=1.0]
table {%
1 0.512152771155039
2 0.6015625
3 0.656583885351817
4 0.717548072338104
5 0.731570502122243
6 0.767561415831248
7 0.783186435699463
8 0.800480763117472
9 0.811164538065592
10 0.825988252957662
11 0.826388875643412
12 0.831263343493144
13 0.844751616319021
14 0.841479698816935
15 0.854033132394155
16 0.855635682741801
17 0.852564096450806
18 0.856637279192607
19 0.862179478009542
20 0.861845632394155
21 0.866519769032796
22 0.862513363361359
23 0.868923624356588
24 0.872863252957662
25 0.868856827418009
26 0.868990381558736
27 0.876335481802622
28 0.873864869276682
29 0.873664538065592
30 0.876335461934408
31 0.870860040187836
32 0.874131937821706
33 0.87827189763387
34 0.880008002122243
35 0.875267088413239
36 0.877136747042338
37 0.88568377494812
38 0.883346696694692
39 0.882011214892069
40 0.882946034272512
41 0.88608439763387
42 0.891960481802622
43 0.886551817258199
44 0.88241183757782
45 0.886151194572449
46 0.884214739004771
47 0.889756937821706
48 0.891292730967204
49 0.883213142553965
50 0.887152771155039
51 0.88528311252594
52 0.889222741127014
53 0.89122595389684
54 0.891025642553965
55 0.885750532150269
56 0.884949266910553
57 0.893629809220632
58 0.892694969971975
59 0.889022429784139
60 0.892427881558736
61 0.89289528131485
62 0.893563032150269
63 0.895365913709005
64 0.897502680619558
65 0.892761747042338
66 0.895699799060822
67 0.892761747042338
68 0.89556622505188
69 0.893763363361359
70 0.893830120563507
71 0.900373935699463
72 0.892895301183065
73 0.890758554140727
74 0.894163986047109
75 0.893563032150269
76 0.895900090535482
77 0.895499467849731
78 0.895032068093618
79 0.896434287230174
80 0.895699799060822
81 0.89309561252594
82 0.895699779192607
83 0.895966867605845
84 0.894965271155039
85 0.897636214892069
86 0.899305562178294
87 0.894097228844961
88 0.899973293145498
89 0.900841335455577
90 0.898571034272512
91 0.898704608281453
92 0.897702991962433
93 0.899172008037567
94 0.901575863361359
95 0.896968503793081
96 0.898771365483602
97 0.898704588413239
98 0.900774578253428
99 0.903445521990458
100 0.898103634516398
101 0.89903845389684
102 0.900974889596303
103 0.90110844373703
104 0.897769769032796
105 0.896701375643412
106 0.901308755079905
107 0.904113233089447
108 0.900707801183065
109 0.8984375
110 0.902644236882528
111 0.902844568093618
112 0.902243594328562
113 0.902310351530711
114 0.901375532150269
115 0.903178413709005
116 0.903044879436493
117 0.901842931906382
118 0.8984375
119 0.903111636638641
120 0.899105250835419
121 0.902110040187836
122 0.899105230967204
123 0.902911325295766
124 0.901509086290995
125 0.90170939763387
126 0.902510682741801
127 0.904780983924866
128 0.903645833333333
129 0.900974889596303
130 0.903645833333333
131 0.902243594328562
132 0.905048072338104
133 0.898303945859273
134 0.902777771155039
135 0.904847741127014
136 0.905649026234945
137 0.903311967849731
138 0.904447098573049
139 0.906917730967204
140 0.90397971868515
141 0.902644236882528
142 0.906917730967204
143 0.908186435699463
144 0.910389959812164
145 0.912860592206319
146 0.912393152713776
147 0.912459949652354
148 0.910790582497915
149 0.910657048225403
150 0.914396365483602
151 0.914930562178294
152 0.914529939492544
153 0.911458333333333
154 0.913394769032796
155 0.913862188657125
156 0.912059287230174
157 0.915731847286224
158 0.914663473765055
159 0.91346154610316
160 0.916466355323792
161 0.913728634516398
162 0.912459929784139
163 0.914997339248657
164 0.914329588413239
165 0.915865381558736
166 0.915197630723318
167 0.914730230967204
168 0.914463142553965
169 0.91386216878891
170 0.913928945859273
171 0.915998935699463
172 0.915865381558736
173 0.918336013952891
174 0.916332801183065
175 0.914863785107931
176 0.915531496206919
177 0.915731827418009
178 0.915598293145498
179 0.914797008037567
180 0.919204076131185
181 0.914730230967204
182 0.916332801183065
183 0.91386216878891
184 0.916132469971975
185 0.915731827418009
186 0.915998935699463
187 0.91673344373703
188 0.915665050347646
189 0.917000532150269
190 0.918335994084676
191 0.913595060507456
192 0.916266024112701
193 0.917067289352417
194 0.913928945859273
195 0.917267620563507
196 0.915397981802622
197 0.917000532150269
198 0.91860310236613
199 0.918870210647583
200 0.915865381558736
201 0.91613248984019
202 0.914396365483602
203 0.916800220807393
204 0.917735060056051
205 0.915731827418009
206 0.917000512282054
207 0.91613248984019
208 0.915731847286224
209 0.916199266910553
210 0.917868594328562
211 0.916199247042338
212 0.916599889596303
213 0.915531535943349
214 0.917200863361359
215 0.916332801183065
216 0.918135682741801
217 0.919070521990458
218 0.918135702610016
219 0.917200863361359
220 0.916800200939178
221 0.917200863361359
222 0.916065692901611
223 0.916800220807393
224 0.914463142553965
225 0.915464739004771
226 0.91673344373703
227 0.918402791023254
228 0.915798604488373
229 0.917467951774597
230 0.916266024112701
231 0.917935371398926
232 0.917267620563507
233 0.917267620563507
234 0.916800220807393
235 0.914930562178294
236 0.913995722929637
237 0.917401174704234
238 0.915932158629099
239 0.916266024112701
240 0.917935371398926
241 0.917267640431722
242 0.914129257202148
243 0.917067309220632
244 0.91713406642278
245 0.918669859568278
246 0.916599909464518
247 0.918068905671438
248 0.917467951774597
249 0.914930562178294
250 0.917000532150269
251 0.917267620563507
252 0.917668263117472
253 0.915665070215861
254 0.915464739004771
255 0.913728614648183
256 0.915598293145498
257 0.918002148469289
258 0.914329588413239
259 0.916599909464518
260 0.916332801183065
261 0.916599889596303
262 0.914930562178294
263 0.915197650591532
264 0.913728614648183
265 0.917735040187836
266 0.914396345615387
267 0.914997319380442
268 0.914730230967204
269 0.915865381558736
270 0.917534728844961
271 0.917334417502085
272 0.915197650591532
273 0.917601486047109
274 0.917534728844961
275 0.916332801183065
276 0.91713406642278
277 0.914596696694692
278 0.918269236882528
279 0.913795411586761
280 0.916466355323792
281 0.917267640431722
282 0.912593483924866
283 0.914596696694692
284 0.916599889596303
};
\addplot [line width=2.5pt, black, opacity=1.0]
table {%
1 0.416199256976445
2 0.394564638535182
3 0.488848825295766
4 0.574719548225403
5 0.62252938747406
6 0.614449799060822
7 0.706797540187836
8 0.737313032150269
9 0.703525642553965
10 0.729700843493144
11 0.75761216878891
12 0.762486636638641
13 0.769698182741801
14 0.791332801183065
15 0.788661857446035
16 0.804420411586761
17 0.800747871398926
18 0.81270033121109
19 0.772168795267741
20 0.790331204732259
21 0.821113765239716
22 0.807959417502085
23 0.837673624356588
24 0.831196586290995
25 0.820646385351817
26 0.842214186986287
27 0.831396917502085
28 0.859842419624329
29 0.841746787230174
30 0.81804221868515
31 0.820779899756114
32 0.844618062178294
33 0.853031516075134
34 0.847355763117472
35 0.865518152713776
36 0.86678683757782
37 0.855769217014313
38 0.837339739004771
39 0.840077459812164
40 0.827791134516398
41 0.861444969971975
42 0.874666134516398
43 0.870592951774597
44 0.865785241127014
45 0.860710461934408
46 0.860376596450806
47 0.857572118441264
48 0.84829060236613
49 0.854233463605245
50 0.845018704732259
51 0.847622871398926
52 0.852497339248657
53 0.863247871398926
54 0.884615381558736
55 0.885216335455577
56 0.879807710647583
57 0.867321054140727
58 0.875467419624329
59 0.873130341370901
60 0.862179478009542
61 0.838074266910553
62 0.869791666666667
63 0.862914005915324
64 0.858640491962433
65 0.868856827418009
66 0.871861656506856
67 0.878338674704234
68 0.879340271155039
69 0.871260682741801
70 0.855902771155039
71 0.866853614648183
72 0.867254277070363
73 0.875
74 0.871260682741801
75 0.844484508037567
76 0.85877404610316
77 0.847889959812164
78 0.838274578253428
79 0.851629277070363
80 0.875801285107931
81 0.877871255079905
82 0.879273513952891
83 0.879139979680379
84 0.876535793145498
85 0.867321054140727
86 0.88735310236613
87 0.890090823173523
88 0.889022449652354
89 0.886151174704234
90 0.887686967849731
91 0.893963674704234
92 0.893563032150269
93 0.891092399756114
94 0.888488252957662
95 0.892494658629099
96 0.888888895511627
97 0.894831717014313
98 0.894898494084676
99 0.894898494084676
100 0.897702991962433
101 0.899238785107931
102 0.897302349408468
103 0.900173604488373
104 0.899906516075134
105 0.901642620563507
106 0.902176817258199
107 0.903645833333333
108 0.905515511830648
109 0.904447118441264
110 0.904380321502686
111 0.904380341370901
112 0.904714186986287
113 0.908787389596303
114 0.905181626478831
115 0.90645033121109
116 0.907118062178294
117 0.907986104488373
118 0.908653855323792
119 0.905181626478831
120 0.907652219136556
121 0.909455140431722
122 0.908186435699463
123 0.907318373521169
124 0.908386747042338
125 0.909521917502085
126 0.909722228844961
127 0.906717419624329
128 0.910723845163981
129 0.907852570215861
130 0.910256405671438
131 0.909388343493144
132 0.909254809220632
133 0.908453524112701
134 0.908787389596303
135 0.909588674704234
136 0.911591867605845
137 0.910657048225403
138 0.911458333333333
139 0.909254809220632
140 0.911725421746572
141 0.912593483924866
142 0.909655451774597
143 0.912059287230174
144 0.911792198816935
145 0.909989337126414
146 0.911258002122243
147 0.905982891718547
148 0.907986124356588
149 0.909188032150269
150 0.907385150591532
151 0.912059287230174
152 0.909455140431722
153 0.910256405671438
154 0.910924136638641
155 0.908453524112701
156 0.909722228844961
157 0.911525110403697
158 0.909054497877757
159 0.911925733089447
160 0.910256405671438
161 0.909254809220632
162 0.910990913709005
163 0.909922540187836
164 0.910323182741801
165 0.911258002122243
166 0.908854166666667
167 0.911124467849731
168 0.909455120563507
169 0.911391576131185
170 0.910790582497915
171 0.909989297389984
172 0.909254809220632
173 0.913060903549194
174 0.912593483924866
175 0.910723825295766
176 0.912459929784139
177 0.911258021990458
178 0.912860572338104
179 0.90952189763387
180 0.910056074460348
181 0.911458353201548
182 0.910523494084676
183 0.913060903549194
184 0.911391576131185
185 0.911792198816935
186 0.91079060236613
187 0.910924136638641
188 0.912393152713776
189 0.911191244920095
190 0.912259618441264
191 0.910790582497915
192 0.910523494084676
193 0.907986104488373
194 0.91079060236613
195 0.911925752957662
196 0.91079060236613
197 0.913995722929637
198 0.911057690779368
199 0.909121255079905
200 0.910323182741801
201 0.912192821502686
202 0.910056094328562
203 0.910189648469289
204 0.90892094373703
205 0.911725421746572
206 0.910590271155039
207 0.91386216878891
208 0.912593483924866
209 0.912126064300537
210 0.910790582497915
211 0.909521917502085
212 0.910990913709005
213 0.912793795267741
214 0.910189648469289
215 0.910857379436493
216 0.91199251015981
217 0.913394749164581
218 0.911391576131185
219 0.911925733089447
220 0.913060903549194
221 0.911658644676208
222 0.911525090535482
223 0.913261195023855
224 0.912927349408468
225 0.911391576131185
226 0.912660260995229
227 0.912459949652354
228 0.912393172581991
229 0.91466345389684
230 0.912793795267741
231 0.913795391718547
232 0.912727038065592
233 0.911992530028025
234 0.910723825295766
235 0.912793815135956
236 0.910723825295766
237 0.912660260995229
238 0.910389959812164
239 0.911057690779368
240 0.9140625
241 0.911458333333333
242 0.910723825295766
243 0.911057690779368
244 0.913728634516398
245 0.914129277070363
246 0.912994106610616
247 0.913060903549194
248 0.914062480131785
249 0.910523513952891
250 0.912126064300537
251 0.912593464056651
252 0.911124467849731
253 0.911725441614787
254 0.914196034272512
255 0.91426283121109
256 0.912727018197378
257 0.912727038065592
258 0.912593464056651
259 0.912192841370901
260 0.913862188657125
261 0.912126064300537
262 0.913261214892069
263 0.914463142553965
264 0.912459949652354
265 0.90952189763387
266 0.910389959812164
267 0.912727038065592
268 0.912393152713776
269 0.911992530028025
270 0.913795411586761
271 0.911658644676208
272 0.912059307098389
273 0.911124487717946
274 0.909722228844961
275 0.911057690779368
276 0.913327991962433
277 0.912994126478831
278 0.910389939943949
279 0.909989317258199
280 0.911124467849731
281 0.91346154610316
282 0.910590271155039
283 0.911391576131185
284 0.913928945859273
};
\addplot [line width=2.5pt, color1, opacity=1.0]
table {%
1 0.348290592432022
2 0.436431626478831
3 0.490718493858973
4 0.555355230967204
5 0.575186967849731
6 0.596287389596303
7 0.626135150591532
8 0.645833333333333
9 0.664196034272512
10 0.697248935699463
11 0.708800752957662
12 0.727497319380442
13 0.736778855323792
14 0.741786857446035
15 0.730969568093618
16 0.769698162873586
17 0.776842951774597
18 0.775774578253428
19 0.785523513952891
20 0.789797008037567
21 0.795739829540253
22 0.806891024112701
23 0.807759086290995
24 0.81497061252594
25 0.811164518197378
26 0.811631937821706
27 0.830462058385213
28 0.830061455567678
29 0.82772437731425
30 0.824586013952891
31 0.82565438747406
32 0.835202991962433
33 0.84147971868515
34 0.84354966878891
35 0.840945502122243
36 0.842481315135956
37 0.84581998984019
38 0.852697650591532
39 0.855101486047109
40 0.851896365483602
41 0.853699247042338
42 0.854099869728088
43 0.84889155626297
44 0.856436947981516
45 0.850894769032796
46 0.865785241127014
47 0.861979166666667
48 0.862446586290995
49 0.860243062178294
50 0.860243062178294
51 0.858306626478831
52 0.868322670459747
53 0.862780431906382
54 0.866252680619558
55 0.871327439943949
56 0.868255893389384
57 0.868589758872986
58 0.868055562178294
59 0.867053965727488
60 0.863114317258199
61 0.868322630723318
62 0.869190712769826
63 0.876335481802622
64 0.87727032105128
65 0.869057138760885
66 0.874399026234945
67 0.878806094328562
68 0.875868062178294
69 0.873530983924866
70 0.880942821502686
71 0.880809287230174
72 0.870325843493144
73 0.880408644676208
74 0.882745703061422
75 0.874532580375671
76 0.874065160751343
77 0.881543795267741
78 0.878138363361359
79 0.878539005915324
80 0.875
81 0.876068373521169
82 0.882478654384613
83 0.882278323173523
84 0.880275110403697
85 0.879874467849731
86 0.886551817258199
87 0.88014155626297
88 0.883680562178294
89 0.884548604488373
90 0.883880873521169
91 0.885950863361359
92 0.884415070215861
93 0.883680562178294
94 0.88855501015981
95 0.885817309220632
96 0.888621807098389
97 0.881143152713776
98 0.885950843493144
99 0.884949247042338
100 0.885349909464518
101 0.890958865483602
102 0.889022429784139
103 0.884548604488373
104 0.884147981802622
105 0.886151174704234
106 0.891493062178294
107 0.885149578253428
108 0.883680562178294
109 0.889890491962433
110 0.888488252957662
111 0.888087590535482
112 0.889623403549194
113 0.886752128601074
114 0.887887279192607
115 0.889756957689921
116 0.895566244920095
117 0.888087630271912
118 0.895299136638641
119 0.89389689763387
120 0.888221164544423
121 0.88795405626297
122 0.883279919624329
123 0.891359508037567
124 0.889957249164581
125 0.889957269032796
126 0.889222741127014
127 0.887152751286825
128 0.892427901426951
129 0.888822118441264
130 0.896968464056651
131 0.892227570215861
132 0.890224357446035
133 0.892828524112701
134 0.889823714892069
135 0.892628212769826
136 0.886885662873586
137 0.893963674704234
138 0.89636751015981
139 0.893162409464518
140 0.89309561252594
141 0.89616721868515
142 0.896100441614787
143 0.891960481802622
144 0.891626616319021
145 0.890625
146 0.891493042310079
147 0.891225973765055
148 0.888488252957662
149 0.89042466878891
150 0.888087610403697
151 0.895232379436493
152 0.887620190779368
153 0.894297540187836
154 0.891092419624329
155 0.89516560236613
156 0.896033664544423
157 0.896233975887299
158 0.894097228844961
159 0.895833353201548
160 0.893362700939178
161 0.892828524112701
162 0.898170411586761
163 0.891493062178294
164 0.892561435699463
165 0.892561455567678
166 0.899505873521169
167 0.895232359568278
168 0.895299136638641
169 0.892227550347646
170 0.894097228844961
171 0.892227550347646
172 0.893362720807393
173 0.897435903549194
174 0.896300752957662
175 0.899772981802622
176 0.895833333333333
177 0.897502660751343
178 0.897168815135956
179 0.895633002122243
180 0.893362700939178
181 0.897903303305308
182 0.900173604488373
183 0.896167198816935
184 0.895900090535482
185 0.900173624356588
186 0.90050748984019
187 0.900106847286224
188 0.897102038065592
189 0.8984375
190 0.899238785107931
191 0.896768152713776
192 0.899105230967204
193 0.894364317258199
194 0.896233995755514
195 0.896233975887299
196 0.894431094328562
197 0.899772981802622
198 0.896968503793081
199 0.897168795267741
200 0.899973273277283
201 0.897970080375671
202 0.898838142553965
203 0.896901706854502
204 0.899639427661896
205 0.895833333333333
206 0.896033644676208
207 0.900707801183065
208 0.898904919624329
209 0.894564628601074
210 0.903178413709005
211 0.898504277070363
212 0.893629809220632
213 0.898504277070363
214 0.894965271155039
215 0.900707801183065
216 0.898170411586761
217 0.902510682741801
218 0.898437480131785
219 0.896768152713776
220 0.899372339248657
221 0.897703011830648
222 0.898170411586761
223 0.897102038065592
224 0.900841355323792
225 0.898838142553965
226 0.896367530028025
227 0.902510702610016
228 0.902377148469289
229 0.902176817258199
230 0.902978082497915
231 0.899439116319021
232 0.902510682741801
233 0.899572670459747
234 0.897035260995229
235 0.900173624356588
236 0.898504277070363
237 0.900507469971975
238 0.898370742797852
239 0.900307158629099
240 0.895299156506856
241 0.899105230967204
242 0.900440692901611
243 0.901842951774597
244 0.901976486047109
245 0.899973273277283
246 0.903712610403697
247 0.898303945859273
248 0.905782580375671
249 0.902577459812164
250 0.895432690779368
251 0.902711013952891
252 0.90170939763387
253 0.904046456019084
254 0.899906535943349
255 0.898437519868215
256 0.89863783121109
257 0.898504277070363
258 0.896968503793081
259 0.898771365483602
260 0.895900110403697
261 0.896968464056651
262 0.901575843493144
263 0.90150906642278
264 0.89903845389684
265 0.900307158629099
266 0.903512279192607
267 0.899505873521169
268 0.898237188657125
269 0.898704608281453
270 0.902243574460348
271 0.901175200939178
272 0.904781003793081
273 0.903111656506856
274 0.901776174704234
275 0.900307158629099
276 0.902310351530711
277 0.901041646798452
278 0.900707801183065
279 0.902443905671438
280 0.898838142553965
281 0.899772981802622
282 0.899238785107931
283 0.901776174704234
284 0.902711013952891
};
\addplot [line width=2.5pt, blue, opacity=1.0]
table {%
1 0.514222770929337
2 0.602163463830948
3 0.540464758872986
4 0.674078524112701
5 0.717848569154739
6 0.675180286169052
7 0.75040066242218
8 0.763822108507156
9 0.752704322338104
10 0.790665060281754
11 0.794170677661896
12 0.759415060281754
13 0.799779623746872
14 0.824018448591232
15 0.837339758872986
16 0.831730753183365
17 0.826221972703934
18 0.83573716878891
19 0.821013629436493
20 0.854467153549194
21 0.844250798225403
22 0.850661069154739
23 0.831229954957962
24 0.852664262056351
25 0.845853358507156
26 0.853265225887299
27 0.857071310281754
28 0.866586536169052
29 0.85526841878891
30 0.867087364196777
31 0.867487967014313
32 0.85176283121109
33 0.8671875
34 0.872796475887299
35 0.876402229070663
36 0.872395843267441
37 0.870292454957962
38 0.874198704957962
39 0.873297274112701
40 0.877503991127014
41 0.873697906732559
42 0.881911069154739
43 0.879407048225403
44 0.876001626253128
45 0.880308479070663
46 0.87459933757782
47 0.881710737943649
48 0.882011204957962
49 0.887920647859573
50 0.883112967014313
51 0.88671875
52 0.886117786169052
53 0.892628192901611
54 0.886017620563507
55 0.889322936534882
56 0.883513629436493
57 0.89042466878891
58 0.890725165605545
59 0.895532846450806
60 0.893028825521469
61 0.893629789352417
62 0.895633012056351
63 0.89863783121109
64 0.900841325521469
65 0.896233975887299
66 0.894030451774597
67 0.897636204957962
68 0.895633012056351
69 0.903345346450806
70 0.903145045042038
71 0.901542484760284
72 0.903846144676208
73 0.902544051408768
74 0.901241987943649
75 0.907051265239716
76 0.904447108507156
77 0.905248403549194
78 0.908954322338104
79 0.906750798225403
80 0.906850963830948
81 0.910757213830948
82 0.909154623746872
83 0.909254789352417
84 0.910056084394455
85 0.912359774112701
86 0.907451927661896
87 0.909054487943649
88 0.909054487943649
89 0.910056084394455
90 0.912259638309479
91 0.905949503183365
92 0.907151430845261
93 0.907051265239716
94 0.910657048225403
95 0.910456717014313
96 0.908653855323792
97 0.912059307098389
98 0.907451927661896
99 0.905949503183365
100 0.907652258872986
101 0.901842951774597
102 0.897235572338104
103 0.899839758872986
104 0.881911039352417
105 0.889022409915924
106 0.880408644676208
107 0.889823704957962
108 0.887219548225403
109 0.887419879436493
110 0.890124201774597
111 0.898838132619858
112 0.881310075521469
113 0.892528057098389
114 0.898737996816635
115 0.893830150365829
116 0.879607379436493
117 0.890825301408768
118 0.892628222703934
119 0.898737967014313
120 0.882011234760284
121 0.896935075521469
122 0.897235572338104
123 0.895833313465118
124 0.894330948591232
125 0.887820512056351
126 0.884114593267441
127 0.895232379436493
128 0.891726762056351
129 0.894130617380142
130 0.905448704957962
131 0.894631415605545
132 0.897636204957962
133 0.897335737943649
134 0.901442289352417
135 0.899539262056351
136 0.897135436534882
137 0.897636234760284
138 0.897836536169052
139 0.90254408121109
140 0.895933508872986
141 0.905749201774597
142 0.904547274112701
143 0.911258012056351
144 0.916666686534882
145 0.915164262056351
146 0.915464758872986
147 0.913361370563507
148 0.916766822338104
149 0.917267620563507
150 0.917167484760284
151 0.916165858507156
152 0.918970346450806
153 0.916766822338104
154 0.916666686534882
155 0.920773237943649
156 0.915564894676208
157 0.918770015239716
158 0.916566491127014
159 0.916866987943649
160 0.914863765239716
161 0.909154653549194
162 0.901241987943649
163 0.908954322338104
164 0.910857379436493
165 0.903645813465118
166 0.904547274112701
167 0.912159472703934
168 0.908553689718246
169 0.909154653549194
170 0.913461536169052
171 0.910957545042038
172 0.882211536169052
173 0.89453125
174 0.897335737943649
175 0.896033674478531
176 0.899038463830948
177 0.897936701774597
178 0.8984375
179 0.898938298225403
180 0.891526460647583
181 0.898838132619858
182 0.901542484760284
183 0.900340557098389
184 0.904046475887299
185 0.901642620563507
186 0.899338960647583
187 0.899939894676208
188 0.905348569154739
189 0.904847741127014
190 0.903044879436493
191 0.89863783121109
192 0.903345376253128
193 0.902243584394455
194 0.901141852140427
195 0.899639427661896
196 0.896834939718246
197 0.904647439718246
198 0.905248373746872
199 0.903445512056351
200 0.904146641492844
201 0.903846144676208
202 0.906350165605545
203 0.901041656732559
204 0.900941520929337
205 0.902544051408768
206 0.892728388309479
207 0.891726762056351
208 0.903044849634171
209 0.898737996816635
210 0.900240391492844
211 0.898137032985687
212 0.889723539352417
213 0.899238765239716
214 0.91015625
215 0.913161069154739
216 0.910356551408768
217 0.913161069154739
218 0.914763599634171
219 0.915965557098389
220 0.915564924478531
221 0.91756808757782
222 0.915965557098389
223 0.917568117380142
224 0.916967153549194
225 0.916065722703934
226 0.913962334394455
227 0.917467951774597
228 0.9140625
229 0.915264397859573
230 0.913962364196777
231 0.915164262056351
232 0.917267650365829
233 0.918068915605545
234 0.915264427661896
235 0.914963930845261
236 0.914162665605545
237 0.911558479070663
238 0.916065692901611
239 0.914863795042038
240 0.91776841878891
241 0.918269217014313
242 0.916466355323792
243 0.913561701774597
244 0.916165888309479
245 0.918469548225403
246 0.915765225887299
247 0.917568117380142
248 0.913461536169052
249 0.915364563465118
250 0.913962334394455
251 0.91776841878891
252 0.914362996816635
253 0.91446316242218
254 0.917768448591232
255 0.918870180845261
256 0.916866987943649
257 0.918369382619858
258 0.914663434028625
259 0.916466355323792
260 0.917668282985687
261 0.916366189718246
262 0.919371008872986
263 0.918569684028625
264 0.917067319154739
265 0.915064096450806
266 0.916866987943649
267 0.916866987943649
268 0.91836941242218
269 0.918569713830948
270 0.916065692901611
271 0.917167484760284
272 0.916266024112701
273 0.919070512056351
274 0.919871777296066
275 0.914663463830948
276 0.916466355323792
277 0.913261234760284
278 0.919471144676208
279 0.918669849634171
280 0.915464729070663
281 0.917067289352417
282 0.919070512056351
283 0.914463132619858
284 0.919571310281754
};
\addplot [line width=2.5pt, red, opacity=1.0]
table {%
1 0.508680562178294
2 0.57525372505188
3 0.558226486047109
4 0.660657048225403
5 0.706129809220632
6 0.667000552018484
7 0.713474889596303
8 0.759481847286224
9 0.766559839248657
10 0.774639427661896
11 0.784321586290995
12 0.791132469971975
13 0.804153323173523
14 0.810830652713776
15 0.813501596450806
16 0.819911857446035
17 0.82752404610316
18 0.829927881558736
19 0.846420923868815
20 0.845619658629099
21 0.841613252957662
22 0.851629257202148
23 0.852564116319021
24 0.849425733089447
25 0.851228634516398
26 0.858506937821706
27 0.867321054140727
28 0.864783664544423
29 0.863715271155039
30 0.869057158629099
31 0.881610572338104
32 0.86985844373703
33 0.875400642553965
34 0.87706998984019
35 0.873998383680979
36 0.867654919624329
37 0.877537409464518
38 0.886351486047109
39 0.884147981802622
40 0.880876044432322
41 0.884281516075134
42 0.88795405626297
43 0.882478654384613
44 0.883213142553965
45 0.881009618441264
46 0.889423072338104
47 0.889423072338104
48 0.89042466878891
49 0.889022429784139
50 0.88321312268575
51 0.887620190779368
52 0.890892088413239
53 0.889289518197378
54 0.892694969971975
55 0.895232359568278
56 0.896634618441264
57 0.897702972094218
58 0.898637811342875
59 0.901642620563507
60 0.893095632394155
61 0.90090811252594
62 0.896834929784139
63 0.900307178497314
64 0.902110040187836
65 0.902577459812164
66 0.903846144676208
67 0.902243594328562
68 0.901509086290995
69 0.903846164544423
70 0.905315180619558
71 0.90377938747406
72 0.906316777070363
73 0.904914518197378
74 0.908319969971975
75 0.902911325295766
76 0.905916114648183
77 0.90625
78 0.905248383680979
79 0.905248403549194
80 0.905715803305308
81 0.908854166666667
82 0.906517088413239
83 0.910924156506856
84 0.909722228844961
85 0.90831998984019
86 0.912660241127014
87 0.911591867605845
88 0.910056094328562
89 0.910523494084676
90 0.912727018197378
91 0.913528303305308
92 0.911324779192607
93 0.914129277070363
94 0.912860572338104
95 0.913528323173523
96 0.91346154610316
97 0.913995742797852
98 0.916399578253428
99 0.911525090535482
100 0.912727018197378
101 0.913728634516398
102 0.913528323173523
103 0.913728634516398
104 0.913661857446035
105 0.913795411586761
106 0.913261234760284
107 0.917534708976746
108 0.91366187731425
109 0.912593483924866
110 0.915798624356588
111 0.915397981802622
112 0.91653311252594
113 0.914997319380442
114 0.914196054140727
115 0.916599889596303
116 0.914730230967204
117 0.914863785107931
118 0.918469548225403
119 0.917534708976746
120 0.918269217014313
121 0.916933755079905
122 0.917868594328562
123 0.916332801183065
124 0.918068905671438
125 0.919938564300537
126 0.920005341370901
127 0.915932158629099
128 0.918202479680379
129 0.920138875643412
130 0.919938544432322
131 0.919871787230174
132 0.918536305427551
133 0.920606295267741
134 0.919871807098389
135 0.920205652713776
136 0.921340803305308
137 0.921140491962433
138 0.921607891718547
139 0.92207533121109
140 0.922542730967204
141 0.921808222929637
142 0.919471144676208
143 0.922943353652954
144 0.921941777070363
145 0.922743042310079
146 0.925948182741801
147 0.921140491962433
148 0.920539518197378
149 0.922676285107931
150 0.924212078253428
151 0.924679478009542
152 0.923811415831248
153 0.923477570215861
154 0.923344016075134
155 0.921674688657125
156 0.923944969971975
157 0.923410773277283
158 0.924612700939178
159 0.924679478009542
160 0.922876596450806
161 0.925213674704234
162 0.924479166666667
163 0.926482379436493
164 0.921941777070363
165 0.923744638760885
166 0.923477570215861
167 0.922876616319021
168 0.92207533121109
169 0.926215271155039
170 0.92227562268575
171 0.92414528131485
172 0.922609508037567
173 0.924011747042338
174 0.927350441614787
175 0.923677881558736
176 0.925747851530711
177 0.923210481802622
178 0.927417198816935
179 0.92434561252594
180 0.92641560236613
181 0.925347208976746
182 0.924345632394155
183 0.924278855323792
184 0.924345632394155
185 0.923811435699463
186 0.926282048225403
187 0.925681074460348
188 0.92394498984019
189 0.925614317258199
190 0.925614317258199
191 0.925080120563507
192 0.925080120563507
193 0.922342419624329
194 0.92434561252594
195 0.925547540187836
196 0.924479186534882
197 0.923010130723318
198 0.925948182741801
199 0.925213674704234
200 0.923210481802622
201 0.925213674704234
202 0.925146917502085
203 0.925747851530711
204 0.926348825295766
205 0.926682710647583
206 0.924479166666667
207 0.928619106610616
208 0.924479146798452
209 0.924679478009542
210 0.923677881558736
211 0.924345632394155
212 0.923277258872986
213 0.925413986047109
214 0.926682690779368
215 0.926749467849731
216 0.925547520319621
217 0.925814628601074
218 0.924679497877757
219 0.924345632394155
220 0.926014939943949
221 0.925547560056051
222 0.924612700939178
223 0.924813032150269
224 0.924412389596303
225 0.926883021990458
226 0.92454594373703
227 0.924145301183065
228 0.92454594373703
229 0.923744658629099
230 0.924212078253428
231 0.92434561252594
232 0.923744658629099
233 0.923744658629099
234 0.925547520319621
235 0.925480763117472
236 0.923076907793681
237 0.92434561252594
238 0.922676265239716
239 0.923677881558736
240 0.925881425539652
241 0.924011747042338
242 0.923410793145498
243 0.923878192901611
244 0.925814628601074
245 0.922676285107931
246 0.924011766910553
247 0.92414528131485
248 0.926816244920095
249 0.922743062178294
250 0.924612700939178
251 0.925547540187836
252 0.927751064300537
253 0.925280431906382
254 0.924879829088847
255 0.921874980131785
256 0.925280471642812
257 0.924212058385213
258 0.923210481802622
259 0.925413986047109
260 0.923544347286224
261 0.923410793145498
262 0.923210481802622
263 0.924879809220632
264 0.923544347286224
265 0.926682690779368
266 0.922409196694692
267 0.923076907793681
268 0.925747851530711
269 0.924545963605245
270 0.923811435699463
271 0.925881405671438
272 0.925681094328562
273 0.926215271155039
274 0.924813032150269
275 0.924011766910553
276 0.92247595389684
277 0.923344016075134
278 0.923477590084076
279 0.923143684864044
280 0.923477550347646
281 0.925080120563507
282 0.922208865483602
283 0.921073714892069
284 0.922008554140727
};
\addplot [line width=2.5pt, color2, opacity=1.0]
table {%
2 0.506810913483302
4 0.595285773277283
6 0.651909728844961
8 0.733306626478831
10 0.776375532150269
12 0.783987720807393
14 0.792801817258199
16 0.824385682741801
18 0.836672008037567
20 0.836538473765055
22 0.84702189763387
24 0.846888363361359
26 0.853498915831248
28 0.863848845163981
30 0.872128744920095
32 0.871127148469289
34 0.878739317258199
36 0.876602570215861
38 0.876201927661896
40 0.883613785107931
42 0.8828125
44 0.889489849408468
46 0.889823714892069
48 0.887887279192607
50 0.891960461934408
52 0.885817309220632
54 0.889222760995229
56 0.892895301183065
58 0.891025642553965
60 0.892294327418009
62 0.890624980131785
64 0.895365913709005
66 0.89516560236613
68 0.893429497877757
70 0.898170411586761
72 0.89823716878891
74 0.890558222929637
76 0.89516560236613
78 0.896167198816935
80 0.895699799060822
82 0.894163986047109
84 0.895032048225403
86 0.896233956019084
88 0.89576655626297
90 0.897702991962433
92 0.896634618441264
94 0.897235592206319
96 0.895633021990458
98 0.897836526234945
100 0.897569457689921
102 0.898971676826477
104 0.897302349408468
106 0.897703011830648
108 0.900307158629099
110 0.894965271155039
112 0.900307158629099
114 0.896768152713776
116 0.894898494084676
118 0.902510682741801
120 0.898036857446035
122 0.901041666666667
124 0.901642620563507
126 0.900040050347646
128 0.897035241127014
130 0.900307138760885
132 0.900841355323792
134 0.902710994084676
136 0.902176817258199
138 0.901642640431722
140 0.903445521990458
142 0.901909708976746
144 0.900841335455577
146 0.898971696694692
148 0.900240381558736
150 0.905982891718547
152 0.903044879436493
154 0.899305562178294
156 0.904647449652354
158 0.902377148469289
160 0.901776194572449
162 0.902911345163981
164 0.902911325295766
166 0.901509086290995
168 0.906583865483602
170 0.901375532150269
172 0.898971676826477
174 0.899505873521169
176 0.902710994084676
178 0.90050748984019
180 0.901709417502085
182 0.901041666666667
184 0.904180030028025
186 0.903445521990458
188 0.906183242797852
190 0.902978122234344
192 0.908186415831248
194 0.906784176826477
196 0.903912921746572
198 0.906183242797852
200 0.903378744920095
202 0.902176837126414
204 0.905715823173523
206 0.904781003793081
208 0.904780983924866
210 0.904246807098389
212 0.901976486047109
214 0.905916134516398
216 0.902844548225403
218 0.903712610403697
220 0.906984508037567
222 0.906517088413239
224 0.907451927661896
226 0.906116465727488
228 0.903779367605845
230 0.905114829540253
232 0.905181606610616
234 0.90317843357722
236 0.903512279192607
238 0.902043263117472
240 0.902043263117472
242 0.905849357446035
244 0.904847760995229
246 0.905114849408468
248 0.903979698816935
250 0.902777791023254
252 0.907385150591532
254 0.903311967849731
256 0.901642620563507
258 0.906984488169352
260 0.905782580375671
262 0.905448714892069
264 0.905515491962433
266 0.901976486047109
268 0.90625
270 0.906784196694692
272 0.910523513952891
274 0.905582269032796
276 0.904046475887299
278 0.904580652713776
280 0.905181626478831
282 0.906717419624329
284 0.905649026234945
};
\end{axis}

\end{tikzpicture}

%% file: labpal/figure_data/performance_comparison/CIFAR-10/CIFAR-10_ResNet-20_test_accuracy.pgf
% This file was created by tikzplotlib v0.9.8.
\begin{tikzpicture}

\definecolor{color0}{rgb}{0.647058823529412,0.164705882352941,0.164705882352941}
\definecolor{color1}{rgb}{0.933333333333333,0.509803921568627,0.933333333333333}
\definecolor{color2}{rgb}{1,0.647058823529412,0}

\begin{axis}[
grid = major,
major grid style={dotted},
legend cell align={left},
legend style={
  fill opacity=0.8,
  draw opacity=1,
  text opacity=1,
  at={(0.91,0.5)},
  anchor=east,
  draw=white!80!black
},
minor xtick={},
%minor ytick={},
reverse legend, legend cell align={left}, legend style={ fill opacity=0.8, draw opacity=1, text opacity=1, at={(0.93,0.3)}, anchor=east, draw=white!80!black},,
tick align=outside,
tick pos=left,
title={test accuracy CIFAR-10 ResNet-20},
width=10.5cm,height=8cm,,
x grid style={white!69.0196078431373!black},
xlabel={epoch with best val. acc.},
xmin=-0.055, xmax=0.055,
xtick style={color=black},
xmajorticks=false,
y grid style={white!69.0196078431373!black},
ylabel={test accuracy},
ymin=0.82, ymax=0.92,
ytick style={color=black},
ytick={0.88,0.89,0.9,0.91,92},
minor y tick num=3,
grid = major,
]
\addplot [draw=color0, fill=color0, mark=-, only marks, mark options={scale=3},line width=3pt]
table{%
x  y
0 0.884949266910553
0 -1
};
\addlegendentry{PLS : $\alpha$: 0.1, $c_{W}$: 0.1, c1: 0.4, $\beta$: 0.0}
\addplot [draw=green!50.1960784313725!black, fill=green!50.1960784313725!black, mark=-, only marks, mark options={scale=3},line width=3pt]
table{%
x  y
0 0.902677635351817
0 -1
};
\addlegendentry{SGD : $\lambda$: 0.1, $\beta$: 0.9}
\addplot [draw=black, fill=black, mark=-, only marks, mark options={scale=3,opacity=0.5},line width=3pt]
table{%
x  y
0 0.905248403549194
0 -1
};
\addlegendentry{SLS : c: 0.1, $\beta$: 0.9, $\mu$: 0.1}
\addplot [draw=color1, fill=color1, mark=-, only marks, mark options={scale=3},line width=3pt]
table{%
x  y
0 0.891493062178294
0 -1
};
\addlegendentry{GOLSI : $c$: 0.99, $\eta$: 2.0, $\alpha$: 0.1, $\beta$: 0.4}
\addplot [draw=color2, fill=color2, mark=-, only marks, mark options={scale=3},line width=3pt]
table{%
x  y
0 0.891526460647583
0 -1
};
\addlegendentry{PAL : $\alpha$: 1.0, $\mu$: 1.0, $\beta$: 0.4 }
\addplot [draw=blue, fill=blue, mark=-, only marks, mark options={scale=3},line width=3pt]
table{%
x  y
0 0.910306483507156
0 -1
};
\addlegendentry{LABPAL-SGD: $|\mathbb{B}_a|$: 1280, $\beta$: 0.0, $\alpha$: 1.9, n: 1000}
\addplot [draw=red, fill=red, mark=-, only marks, mark options={scale=3,opacity=0.5},line width=3pt]
table{%
x  y
0 0.904146631558736
0 -1
};
\addlegendentry{LABPAL-NSGD: $|\mathbb{B}_a|$: 1280, $\beta$: 0.0, $\alpha$: 1.9, n: 1000}

\end{axis}

\end{tikzpicture}

%% file: labpal/figure_data/performance_comparison/CIFAR-10/CIFAR-10_MobileNet-V2_test_accuracy.pgf
% This file was created by tikzplotlib v0.9.8.
\begin{tikzpicture}

\definecolor{color1}{rgb}{0.933333333333333,0.509803921568627,0.933333333333333}
\definecolor{color2}{rgb}{1,0.647058823529412,0}

\begin{axis}[
grid = major,
major grid style={dotted},
minor xtick={},
%minor ytick={},
tick align=outside,
tick pos=left,
title={test accuracy CIFAR-10 MobileNet-V2},
width=10.5cm,height=8cm,,
x grid style={white!69.0196078431373!black},
xlabel={epoch with best val. acc.},
xmin=-0.055, xmax=0.055,
xtick style={color=black},
xmajorticks=false,
y grid style={white!69.0196078431373!black},
ylabel={test accuracy},
ymin=0.89, ymax=0.94,
ytick style={color=black},
minor y tick num=3,
grid = major,
]
%\addlegendentry{PLS : $\alpha$: 0.0001, $c-{W}$: 0.2, c1: 0.005, $\beta$: 0.0}
\addplot [draw=green!50.1960784313725!black, fill=green!50.1960784313725!black, mark=-, only marks, mark options={scale=3},line width=3pt]
table{%
x  y
0 0.92311030626297
0 -1
};
%\addlegendentry{SGD : $\lambda$: 0.1, $\beta$: 0.0}
\addplot [draw=black, fill=black, mark=-, only marks, mark options={scale=3},line width=3pt]
table{%
x  y
0 0.921641290187836
0 -1
};
%\addlegendentry{SLS : c: 0.1, $\beta$: 0.9, $\mu-{init}$: 0.1}
\addplot [draw=color1, fill=color1, mark=-, only marks, mark options={scale=3},line width=3pt]
table{%
x  y
0 0.915631671746572
0 -1
};
%\addlegendentry{GOLSI : $c$: 0.9, $\eta$: 2.0, $\alpha$: 0.0001, $\beta$: 0.0}
\addplot [draw=blue, fill=blue, mark=-, only marks, mark options={scale=3},line width=3pt]
table{%
x  y
0 0.915114164352417
0 -1
};
%\addlegendentry{LABPAL-SGD : $\epsilon$: 1.0}
\addplot [draw=red, fill=red, mark=-, only marks, mark options={scale=3},line width=3pt]
table{%
x  y
0 0.929887811342875
0 -1
};
%\addlegendentry{LABPAL-NSGD : $\epsilon$: 1.0}
\addplot [draw=color2, fill=color2, mark=-, only marks, mark options={scale=3},line width=3pt]
table{%
x  y
0 0.919370989004771
0 -1
};
%\addlegendentry{PAL : $\alpha$: 1.66, $\mu$: 1.0}
\end{axis}

\end{tikzpicture}

%% file: labpal/figure_data/performance_comparison/CIFAR-10/CIFAR-10_DenseNet-121_test_accuracy.pgf
% This file was created by tikzplotlib v0.9.8.
\begin{tikzpicture}

\definecolor{color0}{rgb}{0.933333333333333,0.509803921568627,0.933333333333333}
\definecolor{color1}{rgb}{1,0.647058823529412,0}

\begin{axis}[
grid = major,
major grid style={dotted},
minor xtick={},
%minor ytick={},
tick align=outside,
tick pos=left,
title={test accuracy CIFAR-10 DenseNet-121},
width=10.5cm,height=8cm,,
x grid style={white!69.0196078431373!black},
xlabel={epoch with best val. acc.},
xmin=-0.055, xmax=0.055,
xtick style={color=black},
xmajorticks=false,
y grid style={white!69.0196078431373!black},
ylabel={test accuracy},
ymin=0.89, ymax=0.94,
ytick style={color=black},
minor y tick num=3,
grid = major,
]
\addplot [draw=green!50.1960784313725!black, fill=green!50.1960784313725!black, mark=-, only marks, mark options={scale=3},line width=3pt]
table{%
x  y
0 0.915130873521169
0 -1
};
%\addlegendentry{SGD : $\lambda$: 0.1, $\beta$: 0.0}
\addplot [draw=black, fill=black, mark=-, only marks, mark options={scale=3},line width=3pt]
table{%
x  y
0 0.913995722929637
0 -1
};
%\addlegendentry{SLS : c: 0.1, $\beta$: 0.9, $\mu-{init}$: 0.1}
\addplot [draw=color0, fill=color0, mark=-, only marks, mark options={scale=3},line width=3pt]
table{%
x  y
0 0.903445502122243
0 -1
};
%\addlegendentry{GOLSI : $c$: 0.9, $\eta$: 2.0, $\alpha$: 0.0001, $\beta$: 0.0}
\addplot [draw=blue, fill=blue, mark=-, only marks, mark options={scale=3},line width=3pt]
table{%
x  y
0 0.918269246816635
0 -1
};
%\addlegendentry{LABPAL-SGD : $\epsilon$: 1.0}
\addplot [draw=red, fill=red, mark=-, only marks, mark options={scale=3},line width=3pt]
table{%
x  y
0 0.927350441614787
0 -1
};
%\addlegendentry{LABPAL-NSGD : $\epsilon$: 1.0}
\addplot [draw=color1, fill=color1, mark=-, only marks, mark options={scale=3},line width=3pt]
table{%
x  y
0 0.910823980967204
0 -1
};
%\addlegendentry{PAL : $\alpha$: 1.66, $\mu$: 1.0}
\end{axis}

\end{tikzpicture}

%% file: labpal/performance_comparison_svhn.tex
\begin{figure}[h!]
	\tikzsetfigurename{labpal_optimizer_comparison_svhn}
	\centering
	\def\scale{0.4}
	\begin{tabular}{ c c c}	
		\scalebox{\scale}{\input{"labpal/figure_data/performance_comparison/SVHN/SVHN_ResNet-20_learning_rate.pgf"}}&
		\scalebox{\scale}{\input{"labpal/figure_data/performance_comparison/SVHN/SVHN_MobileNet-V2_learning_rate.pgf"}}&
		\scalebox{\scale}{\input{"labpal/figure_data/performance_comparison/SVHN/SVHN_DenseNet-121_learning_rate.pgf"}}\\
		\scalebox{\scale}{\input{"labpal/figure_data/performance_comparison/SVHN/SVHN_ResNet-20_training_loss.pgf"}}&
		\scalebox{\scale}{\input{"labpal/figure_data/performance_comparison/SVHN/SVHN_MobileNet-V2_training_loss.pgf"}}&
		\scalebox{\scale}{\input{"labpal/figure_data/performance_comparison/SVHN/SVHN_DenseNet-121_training_loss.pgf"}}\\
		\scalebox{\scale}{\input{"labpal/figure_data/performance_comparison/SVHN/SVHN_ResNet-20_validation_accuracy.pgf"}}&
		\scalebox{\scale}{\input{"labpal/figure_data/performance_comparison/SVHN/SVHN_MobileNet-V2_validation_accuracy.pgf"}}&
		\scalebox{\scale}{\input{"labpal/figure_data/performance_comparison/SVHN/SVHN_DenseNet-121_validation_accuracy.pgf"}}\\	
		\scalebox{\scale}{\input{"labpal/figure_data/performance_comparison/SVHN/SVHN_ResNet-20_test_accuracy.pgf"}}&	
		\scalebox{\scale}{\input{"labpal/figure_data/performance_comparison/SVHN/SVHN_MobileNet-V2_test_accuracy.pgf"}}&	
		\scalebox{\scale}{\input{"labpal/figure_data/performance_comparison/SVHN/SVHN_DenseNet-121_test_accuracy.pgf"}}

	\end{tabular}
		\caption{Performance comparison on \textbf{SVHN} of our approach LABPAL in the SGD and NSGD variants against several line search and SGD. Optimal hyperparameters found with a detailed grid search for CIFAR-10 are reused. Our approaches challenge and sometimes surpass the other approaches on training loss, validation, and test accuracy. Columns indicate different models. Rows indicate different metrics.} 
		\label{labpal_fig_optimizer_comparison_svhn}
\end{figure}

%% file: labpal/figure_data/performance_comparison/SVHN/SVHN_DenseNet-121_training_loss.pgf
% This file was created by tikzplotlib v0.9.8.
\begin{tikzpicture}

\definecolor{color0}{rgb}{0.647058823529412,0.164705882352941,0.164705882352941}
\definecolor{color1}{rgb}{1,0.647058823529412,0}
\definecolor{color2}{rgb}{0.933333333333333,0.509803921568627,0.933333333333333}

\begin{axis}[
grid = major,
major grid style={dotted},
log basis y={10},
minor xtick={},
minor ytick={},
tick align=outside,
tick pos=left,
title={training loss SVHN DenseNet-121},
width=10.5cm,height=8cm,,
x grid style={white!69.0196078431373!black},
xlabel={epochs},
xmin=-14.2, xmax=298.2,
xtick style={color=black},
xtick={-50,0,50,100,150,200,250,300},
y grid style={white!69.0196078431373!black},
ylabel={training loss},
ymin=1e-4, ymax=5.5,
ymode=log,
ytick style={color=black},
ytick={1e-05,0.0001,0.001,0.01,0.1,1,10,100}
]
\addplot [line width=2.5pt, color0, opacity=1.0]
table {%
0 2.34695760409037
1 1.61670561631521
2 1.19927632808685
3 0.995419283707937
4 0.857371926307678
5 0.755861540635427
6 0.682371338208516
7 0.622405608495077
8 0.573267797629038
9 0.526595850785573
10 0.491704046726227
11 0.46042142311732
12 0.430224825938543
13 0.408631483713786
14 0.385092775026957
15 0.361077318588893
16 0.345200598239899
17 0.325146694978078
18 0.308075179656347
19 0.288648039102554
20 0.278831551472346
21 0.262620329856873
22 0.247196068366369
23 0.240302314360936
24 0.225005318721135
25 0.216889744003614
26 0.206298296650251
27 0.191812992095947
28 0.186614056428274
29 0.179475689927737
30 0.169556096196175
31 0.162781938910484
32 0.1548197666804
33 0.144019683202108
34 0.141497487823168
35 0.132500559091568
36 0.128478323419889
37 0.125612316032251
38 0.122204797963301
39 0.111584963897864
40 0.107389137148857
41 0.101941178242366
42 0.106058364113172
43 0.154278039932251
44 0.132505014538765
45 0.123246086140474
46 0.11836301535368
47 0.109126754105091
48 0.106994032859802
49 0.100624320407708
50 0.146054970721404
51 0.133867345750332
52 0.12467618783315
53 0.117097735404968
54 0.10781749834617
55 0.0990608260035515
56 0.136338969071706
57 0.125689590970675
58 0.113712139427662
59 0.130357588330905
60 0.124777540564537
61 0.111934795975685
62 0.1078750466307
63 0.0949756950139999
64 0.0937402000029882
65 0.0879919032255809
66 0.0962531318267186
67 0.0974230021238327
68 0.0910681163271268
69 0.0903154412905375
70 0.102798911432425
71 0.0922861744960149
72 0.0903032099207242
73 0.0829952384034793
74 0.157736892501513
75 0.129734625418981
76 0.132619388401508
77 0.122776016592979
78 0.125701149304708
79 0.113198491434256
80 0.134758283694585
81 0.121609029670556
82 0.109016291797161
83 0.102936362226804
84 0.0995136375228564
85 0.118207807342211
86 0.106132045388222
87 0.100456928213437
88 0.0915524090329806
89 0.0889196172356606
90 0.080897939701875
91 0.123234927654266
92 0.10447808355093
93 0.0979158009092013
94 0.0862654608984788
95 0.114300658305486
96 0.100867807865143
97 0.104870175321897
98 0.107923448085785
99 0.0961489031712214
100 0.0915624101956685
101 0.0879355495174726
102 0.12368256598711
103 0.0891300191481908
104 0.119635790586472
105 0.109086098770301
106 0.0995021189252536
107 0.101501243809859
108 0.104123217364152
109 0.092822772761186
110 0.0892617404460907
111 0.0831430479884148
112 0.0735074120263259
113 0.0690124407410622
114 0.0660090235372384
115 0.270521031071742
116 0.150966294109821
117 0.089667696505785
118 0.111436385661364
119 0.104990196724733
120 0.091080995897452
121 0.0900647093852361
122 0.081861600279808
123 0.0738481506705284
124 0.086046372850736
125 0.0822304040193558
126 0.077351450920105
127 0.0718776434659958
128 0.0706965699791908
129 0.0632382209102313
130 0.0578673196335634
131 0.0613072973986467
132 0.0590691852072875
133 0.0622054052849611
134 0.0589506675799688
135 0.0611302219331264
136 0.0791350205739339
137 0.0720265880227089
138 0.0761150841911634
139 0.0687782590587934
140 0.077295313278834
141 0.07960128535827
142 0.0790636340777079
143 0.0957869316140811
144 0.0834236492713292
145 0.076685257256031
146 0.087392528851827
147 0.0831173310677211
148 0.0720196440815926
149 0.0710298344492912
150 0.0830855369567871
151 0.123527482151985
152 0.123548338810603
153 0.1178906361262
154 0.10280787696441
155 0.0931106035908063
156 0.0883714656035105
157 0.0806499992807706
158 0.0868047848343849
159 0.114767737686634
160 0.0981281648079554
161 0.0899501467744509
162 0.0832449421286583
163 0.0815820284187794
164 0.0744682426253955
165 0.0850794166326523
166 0.0898730456829071
167 0.0803118323286374
168 0.0798205062747002
169 0.0735635980963707
170 0.0670453384518623
171 0.0627630054950714
172 0.0612507537007332
173 0.0623983765641848
174 0.0619048985342185
175 0.0622859783470631
176 0.0628923289477825
177 0.0892218599716822
178 0.0871841448048751
179 0.0858069583773613
180 0.0885840356349945
181 0.0795708944400152
182 0.0794829751054446
183 0.09866780291001
184 0.113974931339423
185 0.10432992130518
186 0.0984179029862086
187 0.0911060397823652
188 0.0829811319708824
189 0.0833853930234909
190 0.0750036984682083
191 0.0750023623307546
192 0.0717116792996724
193 0.0749442800879478
194 0.101276357968648
195 0.0946644221742948
196 0.100062722961108
197 0.0923024863004684
198 0.090074627349774
199 0.0875388905405998
200 0.0838764856259028
201 0.0915000289678574
202 0.0956261505683263
203 0.127316298584143
204 0.104107491672039
205 0.100925736129284
206 0.0943741674224536
207 0.0891385575135549
208 0.0847446918487549
209 0.0818626806139946
210 0.0778487647573153
211 0.0711166659990946
212 0.0947618260979652
213 0.110802285373211
214 0.112884496649106
215 0.11699253693223
216 0.10793857773145
217 0.103789983938138
218 0.0979653149843216
219 0.10252495855093
220 0.0956149473786354
221 0.0956241562962532
222 0.0937127744158109
223 0.0880430191755295
224 0.0839834933479627
225 0.0813831202685833
226 0.0751950691143672
227 0.0707445243994395
228 0.0741705944140752
229 0.210501273473104
230 0.116746673981349
231 0.147171219189962
232 0.115816608071327
233 0.110093819598357
234 0.121889670689901
235 0.149600305904945
236 0.125386137515306
237 0.121277816593647
238 0.111958588163058
239 0.124116616944472
240 0.140002583463987
241 0.521917395293713
242 22590462634.7561
243 2883307874986.75
244 33.4154546434681
245 33.4080274154743
246 33.4111053322752
247 33.4061459501584
248 33.3979019472996
};
\addplot [line width=2.5pt, green!50.1960784313725!black, opacity=1.0]
table {%
0 2.35028998057048
1 2.28566137949626
2 1.62092808882395
3 1.02872937917709
4 0.671792407830556
5 0.540906051794688
6 0.432882120211919
7 0.402039448420207
8 0.333423137664795
9 0.309179832537969
10 0.285211046536763
11 0.269508729378382
12 0.252812812725703
13 0.242203041911125
14 0.2324391802152
15 0.222626869877179
16 0.214913621544838
17 0.206548084815343
18 0.197660903135935
19 0.192658896247546
20 0.18592543900013
21 0.180774326125781
22 0.173960755268733
23 0.175601457556089
24 0.165421133240064
25 0.15765468776226
26 0.154434641202291
27 0.150969341397285
28 0.147030651569366
29 0.140537430842717
30 0.135923698544502
31 0.134064137935638
32 0.128595059116681
33 0.125289395451546
34 0.122129601736863
35 0.118403054773808
36 0.119314782321453
37 0.111258089542389
38 0.108772334953149
39 0.104056191941102
40 0.101932143171628
41 0.101303247114023
42 0.0956257556875547
43 0.0938435594240824
44 0.0910978093743324
45 0.0888641898830732
46 0.0870847205320994
47 0.0877646331985791
48 0.0816853145758311
49 0.0807750423749288
50 0.0761487757166227
51 0.0749272753794988
52 0.0714526648322741
53 0.0704639268418153
54 0.0662077528735002
55 0.0660865679383278
56 0.0638517948488394
57 0.0637715545793374
58 0.0592693537473679
59 0.0580132169028123
60 0.0572256445884705
61 0.0553214984635512
62 0.0557651991645495
63 0.0588255288700263
64 0.0638863705098629
65 0.0532394461333752
66 0.0523680932819843
67 0.0483450070023537
68 0.0446491974095503
69 0.0454412437975407
70 0.044572364538908
71 0.0508425583442052
72 0.0431452703972658
73 0.0388390024503072
74 0.0387493371963501
75 0.035776998847723
76 0.0364402259389559
77 0.0347106878956159
78 0.0335167795419693
79 0.0337256838877996
80 0.0353604418536027
81 0.0317960716784
82 0.0307672197620074
83 0.0327139565100273
84 0.0280017604430517
85 0.0296647778401772
86 0.0275910298029582
87 0.0280187626679738
88 0.0257375575602055
89 0.0266882516443729
90 0.0244523622095585
91 0.0245206331213315
92 0.0249645821750164
93 0.0228340085595846
94 0.022910891721646
95 0.0239633526653051
96 0.0284925928960244
97 0.0211431961506605
98 0.0211770863582691
99 0.0208312788357337
100 0.0189311125626167
101 0.0202242986609538
102 0.0198185921957095
103 0.0243073056141535
104 0.0260408576577902
105 0.0225293754289548
106 0.0202820667376121
107 0.0204382296651602
108 0.0189930691073338
109 0.0194700658321381
110 0.0155380858729283
111 0.0163084197168549
112 0.0161247026796142
113 0.0145707620928685
114 0.0165030428518852
115 0.0154678542166948
116 0.0152758741751313
117 0.0159917650744319
118 0.0157104041427374
119 0.0147674235825737
120 0.0157513013109565
121 0.0146078408385317
122 0.0134095931425691
123 0.0131377521902323
124 0.0132486845056216
125 0.0132397205258409
126 0.0130386728172501
127 0.0105272913351655
128 0.0127115966752172
129 0.0124153783544898
130 0.0125625391180317
131 0.0119778150692582
132 0.010903337970376
133 0.0103831589221954
134 0.01105996966362
135 0.0103343476851781
136 0.00967872825761636
137 0.0118316533043981
138 0.0109821890170376
139 0.00996720169981321
140 0.00959298604478439
141 0.00897574579964081
142 0.00995009082059065
143 0.00905955738077561
144 0.00463854878519972
145 0.00354745650353531
146 0.00297002059717973
147 0.0023501361720264
148 0.00215557566843927
149 0.00208030641078949
150 0.00173428631387651
151 0.00184685278994342
152 0.00136528202953438
153 0.00150824171335747
154 0.00147850570889811
155 0.00128518952988088
156 0.0015036742358158
157 0.00126776087563485
158 0.00110566819785163
159 0.00115588404393444
160 0.00123129296116531
161 0.00109681972147276
162 0.00114149844739586
163 0.00104127405211329
164 0.000918472146925827
165 0.000975584106830259
166 0.000943334870195637
167 0.00104559977383663
168 0.000941532023716718
169 0.000811794889159501
170 0.000810042178879182
171 0.000773139433780064
172 0.000804627605248243
173 0.000862802611663938
174 0.000755378743633628
175 0.000834148920451601
176 0.000750243217529108
177 0.000715538936977585
178 0.000745264599875857
179 0.000794318155385554
180 0.000751930463593453
181 0.000679601546532164
182 0.000604909747683754
183 0.000689151318510994
184 0.000648138578981161
185 0.000659453372160594
186 0.000557010343375926
187 0.00054367170863164
188 0.000666930999917289
189 0.000521478184964508
190 0.000565497990464792
191 0.000593413695848236
192 0.000592765184895446
193 0.000568142868966485
194 0.000633362020986776
195 0.000486943812575191
196 0.000568338359395663
197 0.000612163431166361
198 0.000505755383831759
199 0.000479654379887506
200 0.000512339097137252
201 0.000504796247696504
202 0.000604349285519371
203 0.000492886271482954
204 0.000532446021679789
205 0.000593207621326049
206 0.000490091246319935
207 0.000454580731457099
208 0.000468788455085208
209 0.00047954268908749
210 0.000438258895883337
211 0.000428788024388875
212 0.000468177410463492
213 0.000450112856924534
214 0.000479033663092802
215 0.00042360292475981
216 0.000540837926867728
217 0.000393678880451868
218 0.000366880150977522
219 0.000445722204555447
220 0.000472165741181622
221 0.000365030573448166
222 0.000407465122407302
223 0.000398173375288025
224 0.000435323182803889
225 0.000386221538065001
226 0.000418225778654839
227 0.000389978503032277
228 0.000419119790118809
229 0.000402340616953249
230 0.000437894903977091
231 0.000398864028587316
232 0.000419467251049355
233 0.000496619313101595
234 0.000372727818709488
235 0.000404395366786048
236 0.000355035930018251
237 0.000391187631369879
238 0.000366804228785137
239 0.000380340643459931
240 0.000443386631862571
241 0.000412046754111846
242 0.000413493224186823
243 0.000408775270140419
244 0.000346824070826794
245 0.000443564912226672
246 0.000402208534069359
247 0.000355661992216483
248 0.00039218453457579
249 0.0004002948019964
250 0.000362702819984406
251 0.0004673334187828
252 0.000436493758267413
253 0.000413954597509777
254 0.000440169067587703
255 0.000436768459621817
256 0.000411515104739616
257 0.00036179925276277
258 0.000384511785038436
259 0.000448701330848659
260 0.000358008721377701
261 0.0003973238732821
262 0.000388891232432798
263 0.000403811485739425
264 0.000358059798600152
265 0.000378609848363946
266 0.000399899310044323
267 0.000454577452425535
268 0.000349780534937357
269 0.000372713014561062
270 0.000409952859627083
271 0.000422483775764704
272 0.000365970520457874
273 0.000450548987525205
274 0.000397546362364665
275 0.000389515063337361
276 0.000410419219406322
277 0.000375804210004086
278 0.000459744924834619
279 0.000404432833117122
280 0.0003746795700863
281 0.000459490159604078
282 0.000402871936482067
283 0.000394071054567272
284 0.000360274706811955
};
\addplot [line width=2.5pt, black, opacity=1.0]
table {%
0 2.35028998057048
1 2.33497635523478
2 2.03793207804362
3 1.60199745496114
4 1.23449516296387
5 0.867821017901103
6 0.669563015302022
7 0.583227694034576
8 0.525551776091258
9 0.469172984361649
10 0.420419494311015
11 0.386112650235494
12 0.381525953610738
13 0.347657829523087
14 0.321594337622325
15 0.321995089451472
16 0.295915464560191
17 0.285431375106176
18 0.302109559377034
19 0.278466959794362
20 0.252874374389648
21 0.276991426944733
22 0.246171673138936
23 0.219636743267377
24 0.207606251041094
25 0.225498124957085
26 0.249724060297012
27 0.260889237125715
28 0.207431068023046
29 0.204502686858177
30 0.196288034319878
31 0.214492991566658
32 0.193704555432002
33 0.205647205313047
34 0.210560257236163
35 0.177913904190063
36 0.1707873493433
37 0.179109245538712
38 0.196250403920809
39 0.197513724366824
40 0.188211704293887
41 0.194768751660983
42 0.185378084580104
43 0.174070705970128
44 0.175647556781769
45 0.163453603784243
46 0.14474448064963
47 0.146824548641841
48 0.131491002937158
49 0.139837143321832
50 0.163097634911537
51 0.173842151959737
52 0.175523723165194
53 0.139344801505407
54 0.133331562081973
55 0.12732923279206
56 0.133080554505189
57 0.136479233702024
58 0.150285904606183
59 0.146978897353013
60 0.149796289702257
61 0.142938437561194
62 0.140042985479037
63 0.135826456050078
64 0.125926526884238
65 0.112710947791735
66 0.113162847856681
67 0.126314635078112
68 0.127338307599227
69 0.125512103239695
70 0.130317104359468
71 0.111570231616497
72 0.0984825243552526
73 0.0850052411357562
74 0.0863169332345327
75 0.0875966797272364
76 0.0986048057675362
77 0.110833033919334
78 0.113703948756059
79 0.117394737899303
80 0.119960432251294
81 0.115862431625525
82 0.123250623544057
83 0.123516658941905
84 0.109213891128699
85 0.116194528837999
86 0.105633191764355
87 0.0898135205109914
88 0.0810082405805588
89 0.0768493836124738
90 0.0663630117972692
91 0.0614218488335609
92 0.0638879413406054
93 0.0655276998877525
94 0.0661570069690545
95 0.060597068319718
96 0.0613103744884332
97 0.0544856985410055
98 0.0505429692566395
99 0.0485584028065205
100 0.0440279127409061
101 0.0384304635226727
102 0.0358900142212709
103 0.0337635104854902
104 0.0316161854813496
105 0.0293134339153767
106 0.0313984546810389
107 0.0280457915117343
108 0.0266383619358142
109 0.0265571158379316
110 0.0235336807866891
111 0.0226653615633647
112 0.0196420028805733
113 0.0204240701471766
114 0.0186503163228432
115 0.0180277737478415
116 0.0155913298949599
117 0.0165306857476632
118 0.0153616045912107
119 0.0140477238843838
120 0.0142961932967106
121 0.0132305218527714
122 0.0124013805761933
123 0.0126451660568515
124 0.0126578345273932
125 0.0119833250840505
126 0.0114097582797209
127 0.0106764870385329
128 0.0111345595990618
129 0.00979265353331963
130 0.00980508079131444
131 0.00866233774771293
132 0.00889698198686043
133 0.00928254922231038
134 0.00868085181961457
135 0.00889840194334587
136 0.00860283632452289
137 0.00854598265141249
138 0.00751169367382924
139 0.00785340384269754
140 0.00811350888883074
141 0.00805172758797804
142 0.00745155305291216
143 0.00729761319234967
144 0.00657777891804775
145 0.00685904749358694
146 0.00715993794923027
147 0.00620957681288322
148 0.00606417547290524
149 0.00652351059640447
150 0.00681187445297837
151 0.00604394574960073
152 0.00569293337563674
153 0.00636301267271241
154 0.0054910130177935
155 0.00524855653444926
156 0.0051060700789094
157 0.00525960015753905
158 0.00502332082639138
159 0.00487198090801636
160 0.00514650375892719
161 0.00485742495705684
162 0.00469649699516594
163 0.00556659822662671
164 0.0049537254186968
165 0.00444751931354403
166 0.0053755013893048
167 0.00476563706373175
168 0.00453220245738824
169 0.00412518500039975
170 0.00466608908027411
171 0.00405495644857486
172 0.00442479737102985
173 0.00391454179771245
174 0.00424351744974653
175 0.00383012881502509
176 0.00336796751556297
177 0.00407944301453729
178 0.00392060191370547
179 0.00398572386863331
180 0.00395914089555542
181 0.00383524682062368
182 0.00365303906922539
183 0.00361295067705214
184 0.0038845775804172
185 0.00339764488550524
186 0.00338487435753147
187 0.00347027795699735
188 0.00327159914498528
189 0.00319760277246435
190 0.00341163536844154
191 0.00293884263373911
192 0.00262104611222943
193 0.00312687611828248
194 0.0033268427165846
195 0.00329958150784175
196 0.00323462090454996
197 0.00296148362879952
198 0.00273546638588111
199 0.00305320625193417
200 0.00263456099977096
201 0.00301071008046468
202 0.00272711568201582
203 0.00270242787276705
204 0.00305312685668468
205 0.00264089282912513
206 0.00280786429842313
207 0.00268595293164253
208 0.00230948564906915
209 0.0030723907208691
210 0.00267077214084566
211 0.00278117026512822
212 0.00269359528707961
213 0.00247194400678078
214 0.00254322829035421
215 0.00274854021457334
216 0.00235830216358105
217 0.00201333155079434
218 0.00241297665828218
219 0.00246196457495292
220 0.00248965923674405
221 0.00227848114445806
222 0.00229764225271841
223 0.00245348922908306
224 0.00225000564629833
225 0.00264597800560296
226 0.00220203271601349
227 0.00221121970874568
228 0.00244446354918182
229 0.00211257537982116
230 0.00250721001066267
231 0.0019590004424875
232 0.00180027874497076
233 0.00241906622735163
234 0.00219089592186113
235 0.00201956745392332
236 0.00212010143635174
237 0.00192251549257586
238 0.00220161113732805
239 0.00215044377061228
240 0.00161329237744212
241 0.00198902748525143
242 0.00188033337083956
243 0.0021800606045872
244 0.00221696201091011
245 0.00207715951061497
246 0.00182247821552058
247 0.0021320494900768
248 0.00185296575849255
249 0.00200014130678028
250 0.00183361528130869
251 0.00217730943889668
252 0.00194700042872379
253 0.00147823108515392
254 0.00187012615303199
255 0.00160592359801133
256 0.00198012505037089
257 0.00141798456509908
258 0.00200691792027404
259 0.00141291308682412
260 0.00153346266597509
261 0.00201681264055272
262 0.00175374369913091
263 0.00167936746341487
264 0.00182479670426498
265 0.00186114203340063
266 0.00154232617933303
267 0.00144869438372552
268 0.00154567486606538
269 0.00158661266323179
270 0.00168544818491985
271 0.00159319463030746
272 0.00174866438222428
273 0.00144672843938073
274 0.00193145715941985
275 0.00167681408735613
276 0.00153880395616094
277 0.00150823503887902
278 0.00131538095107923
279 0.00145932539210965
280 0.00134680261059354
281 0.00150611410693576
282 0.0018003962856407
283 0.0016690664148579
284 0.00143679557368159
};
\addplot [line width=2.5pt, color1, opacity=1.0]
table {%
0 2.35028998057048
1 2.54536207516988
2 2.27055668830872
3 2.27452365557353
4 2.27774373690287
5 2.27792310714722
6 2.27590012550354
7 2.2757154305776
8 2.27443695068359
9 2.26719792683919
10 2.23090362548828
11 2.04665859540304
12 1.82441540559133
13 1.587020556132
14 1.44212990999222
15 1.33638739585876
16 1.24136865139008
17 1.1661353011926
18 1.11992489298185
19 1.08893038829168
20 1.06200939416885
21 1.03901834289233
22 1.02279246846835
23 1.01508338252703
24 1.00120333830516
25 0.989125241835912
26 0.978453507026037
27 0.949450016021729
28 0.806583046913147
29 0.604917675256729
30 0.477491160233815
31 0.430549154678981
32 0.393593688805898
33 0.373580773671468
34 0.346766889095306
35 0.334514270226161
36 0.323228359222412
37 0.312438289324443
38 0.303639714916547
39 0.29774571955204
40 0.289719705780347
41 0.280941714843114
42 0.275424212217331
43 0.27254843711853
44 0.257400626937548
45 0.258662685751915
46 0.258402119080226
47 0.247341071565946
48 0.245186254382133
49 0.243353719512622
50 0.240978131691615
51 0.230002284049988
52 0.223549048105876
53 0.227711141109467
54 0.224456955989202
55 0.220955232779185
56 0.218737706542015
57 0.21221661567688
58 0.213673656185468
59 0.202956428130468
60 0.203781142830849
61 0.199915875991186
62 0.196939771374067
63 0.192398662368457
64 0.191072523593903
65 0.187346324324608
66 0.184995104869207
67 0.185961087544759
68 0.180421312650045
69 0.180452967683474
70 0.179879014690717
71 0.173136368393898
72 0.179191946983337
73 0.172886773943901
74 0.170618737737338
75 0.166587377587954
76 0.16850571334362
77 0.165766184528669
78 0.164581298828125
79 0.157857899864515
80 0.158278067906698
81 0.15875780582428
82 0.152277161677678
83 0.154291719198227
84 0.150375445683797
85 0.147215157747269
86 0.147104452053706
87 0.14700835943222
88 0.143818587064743
89 0.141539101799329
90 0.142666091521581
91 0.136682080725829
92 0.138788186013699
93 0.132147781550884
94 0.131851034859816
95 0.13538729896148
96 0.132019231716792
97 0.123612592617671
98 0.133541372915109
99 0.125990654031436
100 0.124750842650731
101 0.123326636850834
102 0.123955696821213
103 0.122279124955336
104 0.121644360323747
105 0.119184692700704
106 0.122183273235957
107 0.118236574033896
108 0.117915980517864
109 0.114095521469911
110 0.112774364650249
111 0.112465031445026
112 0.105924320717653
113 0.108178483943144
114 0.110615675648053
115 0.105924559136232
116 0.103037692606449
117 0.104912661015987
118 0.105299818019072
119 0.0977810074885686
120 0.100358170767625
121 0.0967327058315277
122 0.0980440974235535
123 0.0962346717715263
124 0.101305397848288
125 0.0943538347880046
126 0.0947112167874972
127 0.0916715562343597
128 0.0941319838166237
129 0.0925902177890142
130 0.0882525568207105
131 0.087915134926637
132 0.0888896783192952
133 0.0895477359493573
134 0.0905402675271034
135 0.086160883307457
136 0.0848312601447105
137 0.0851012244820595
138 0.0810662706693013
139 0.0849574729800224
140 0.0815438628196716
141 0.0792512446641922
142 0.082273652156194
143 0.0789840792616208
144 0.0785333116849263
145 0.0804996068278949
146 0.0793265302975972
147 0.0766698767741521
148 0.0735291503369808
149 0.0780330325166384
150 0.0739461754759153
151 0.0739215835928917
152 0.0712429421643416
153 0.0716946708659331
154 0.0708815368513266
155 0.0713241075476011
156 0.0724422037601471
157 0.0690259660283725
158 0.0696873118480047
159 0.0681971025963624
160 0.0655889896055063
161 0.069024865825971
162 0.064495629320542
163 0.0693470537662506
164 0.0641837889949481
165 0.063230998814106
166 0.0642813183367252
167 0.0590622959037622
168 0.0614245298008124
169 0.0622474526365598
170 0.0623590188721816
171 0.0559675606588523
172 0.0602617487311363
173 0.0583487637341022
174 0.0607994000116984
175 0.0611790766318639
176 0.059730239212513
177 0.056529230127732
178 0.0573680847883224
179 0.0580548085272312
180 0.0560743200282256
181 0.0542074479162693
182 0.0528408922255039
183 0.0553232058882713
184 0.052977924545606
185 0.0544861468176047
186 0.0514098244408766
187 0.0545991410811742
188 0.0537553032239278
189 0.0493683728079001
190 0.0522569790482521
191 0.0492001796762149
192 0.0506854938964049
193 0.050567248215278
194 0.0475234054028988
195 0.0500389002263546
196 0.0485551816721757
197 0.0489497780799866
198 0.0493169228235881
199 0.0463955961167812
200 0.0497680952151616
201 0.0508509961267312
202 0.0441789366304874
203 0.0459131921331088
204 0.0491821182270845
205 0.0466054317851861
206 0.0448873452842236
207 0.0429004716376464
208 0.0444503612816334
209 0.0450994955996672
210 0.0456888986130555
211 0.0423349539438883
212 0.0431778132915497
213 0.0441658621033033
214 0.0444555679957072
215 0.0444521506627401
216 0.041605236629645
217 0.0404902597268422
218 0.0427881367504597
219 0.0415552134315173
220 0.0443027565876643
221 0.0392584366103013
222 0.040487935145696
223 0.0412236638367176
224 0.0401291574041049
225 0.0395549573004246
226 0.0402759425342083
227 0.0404042266309261
228 0.0390392231444518
229 0.0397106595337391
230 0.0372675408919652
231 0.0383252936104933
232 0.040110753228267
233 0.037542184193929
234 0.0402195776502291
235 0.0385771716634432
236 0.038883063942194
237 0.0362853904565175
238 0.0353456897040208
239 0.0379755074779193
240 0.03550852338473
241 0.0363472929845254
242 0.0338265591611465
243 0.0369821712374687
244 0.0358475719889005
245 0.0354211218655109
246 0.0366200916469097
247 0.0351465021570524
248 0.0348500174780687
249 0.0347439832985401
250 0.0338623089094957
251 0.0351582057774067
252 0.0357349452873071
253 0.0320990731318792
254 0.0348586762944857
255 0.035354845225811
256 0.0342033170163631
257 0.0326106405506531
258 0.0313903292020162
259 0.0323404123385747
260 0.032265467569232
261 0.0334639238814513
262 0.0327303434411685
263 0.0333041213452816
264 0.0321330229441325
265 0.0322795535127322
266 0.0325981273005406
267 0.0326570942997932
268 0.0311570471773545
269 0.0319593443224827
270 0.0299321115016937
271 0.0331113114953041
272 0.0339119601994753
273 0.0293375570327044
274 0.0288136452436447
275 0.0296059908966223
276 0.0321258908758561
277 0.0324435941874981
278 0.0319383672128121
279 0.0294041062394778
280 0.029385178660353
281 0.030749707793196
282 0.0293082986027002
283 0.0284875718255838
284 0.0302490573376417
};
\addplot [line width=2.5pt, blue, opacity=1.0]
table {%
0 2.3312691450119
1 2.14450097084045
2 1.46099412441254
3 0.987222790718079
4 0.673083454370499
5 0.47105972468853
6 0.406601637601852
7 0.37680085003376
8 0.333486109972
9 0.316641628742218
10 0.298769101500511
11 0.276413217186928
12 0.267513781785965
13 0.26062884926796
14 0.242799058556557
15 0.233627930283546
16 0.225250877439976
17 0.213766798377037
18 0.214671410620213
19 0.207317121326923
20 0.194753833115101
21 0.211646497249603
22 0.200202479958534
23 0.189703024923801
24 0.20211636275053
25 0.187262840569019
26 0.178316682577133
27 0.173704333603382
28 0.164939031004906
29 0.162524960935116
30 0.153566382825375
31 0.1496322453022
32 0.143683284521103
33 0.139555595815182
34 0.133499547839165
35 0.129340261220932
36 0.127653922885656
37 0.122263297438622
38 0.121381562203169
39 0.115660261362791
40 0.111960090696812
41 0.107948631048203
42 0.103851318359375
43 0.102287255227566
44 0.103187762200832
45 0.102231539785862
46 0.0963307730853558
47 0.447587300091982
48 0.431221354752779
49 0.171613022685051
50 0.142402123659849
51 0.127136807888746
52 0.117319822311401
53 0.10881282761693
54 0.100683715194464
55 0.0969556011259556
56 0.0930662080645561
57 0.08822350949049
58 0.0854651629924774
59 0.0809470675885677
60 0.0789561346173286
61 0.0793071407824755
62 0.0804109424352646
63 0.0728014968335629
64 0.0805845856666565
65 0.0768515039235353
66 0.0745750833302736
67 0.0685861967504025
68 0.0635555479675531
69 0.0637434758245945
70 0.056126344949007
71 0.0545102264732122
72 0.0557731315493584
73 0.0566324051469564
74 0.0544658713042736
75 0.0513576399534941
76 0.0472673177719116
77 0.0470444615930319
78 0.0461523532867432
79 0.0433830581605434
80 0.0430848263204098
81 0.0387428468093276
82 0.0361134083941579
83 0.0356225026771426
84 0.0329560209065676
85 0.0334341637790203
86 0.032651380635798
87 0.0325365476310253
88 0.0324152633547783
89 0.0301078311167657
90 0.0287603568285704
91 0.0284705953672528
92 0.0278936019167304
93 0.0270671132020652
94 0.0248756115324795
95 0.0279677775688469
96 0.0221556099131703
97 0.0219549760222435
98 0.023882913403213
99 0.0207668230868876
100 0.0206921640783548
101 0.0185383665375412
102 0.0182193592190742
103 0.0138070040848106
104 0.0176753452979028
105 0.0187831786461174
106 0.0152355181053281
107 0.0128180594183505
108 0.0107203237712383
109 0.0087062013335526
110 0.00692245620302856
111 0.00831932085566223
112 0.00767158577218652
113 0.0109739257022738
114 0.0117228082381189
115 0.00852044997736812
116 0.00911728385835886
117 0.0109724602662027
118 0.00961893261410296
119 0.00936768855899572
120 0.00886958162300289
121 0.0147413108497858
122 0.0190000394359231
123 0.0193331621121615
124 0.0272448537871242
125 0.0257233511656523
126 0.0251879617571831
127 0.024928892031312
128 0.0249807583168149
129 0.0227616000920534
130 0.0218910481780767
131 0.0206101099029183
132 0.0224136607721448
133 0.0203983765095472
134 0.0202404148876667
135 0.0161210657097399
136 0.0159695604816079
137 0.018748864531517
138 0.0170217743143439
139 0.0170842884108424
140 0.014500908087939
141 0.0231305076740682
142 0.0256949784234166
143 0.0149575644172728
144 0.00669325888156891
145 0.00493426143657416
146 0.00458078272640705
147 0.00411908549722284
148 0.00407195987645537
149 0.00355614034924656
150 0.00401105300989002
151 0.00463083648355678
152 0.00476235873065889
153 0.00499801337718964
154 0.00478078401647508
155 0.00349470507353544
156 0.00437941169366241
157 0.0043515752768144
158 0.00416158506413922
159 0.00374310556799173
160 0.00331501662731171
161 0.00374185899272561
162 0.00354285020148382
163 0.00295736733824015
164 0.00336375215556473
165 0.00380406487965956
166 0.00458537053782493
167 0.00500114227179438
168 0.00418317969888449
169 0.00478656252380461
170 0.00498347368557006
171 0.00392158166505396
172 0.0242940799798816
173 0.0181996580213308
174 0.0153648196719587
175 0.016378138680011
176 0.0151158454827964
177 0.0140688326209784
178 0.0133205116726458
179 0.0133297764696181
180 0.0113692958839238
181 0.0139787637162954
182 0.0127095107454807
183 0.0128662651404738
184 0.0104987227823585
185 0.00862841005437076
186 0.0102530340664089
187 0.008211086736992
188 0.00855695107020438
189 0.0396430781111121
190 0.0190850213402882
191 0.0183155726408586
192 0.015889557893388
193 0.0166505292290822
194 0.0153026165207848
195 0.0193578093312681
196 0.0158037588698789
197 0.0145191862247884
198 0.0155498931417242
199 0.016874443506822
200 0.0151162903057411
201 0.0142467102268711
202 0.0142184538999572
203 0.0113506963243708
204 0.0129424456972629
205 0.0144562768982723
206 0.00994886946864426
207 0.0119606114458293
208 0.0139945169212297
209 0.0126522821374238
210 0.0104990198742598
211 0.0110757692018524
212 0.0104004357126541
213 0.0112628220813349
214 0.0101250765146688
215 0.00351444230182096
216 0.00269310578005388
217 0.00170577218523249
218 0.001622008101549
219 0.00170896609779447
220 0.00145207077730447
221 0.00174995654379018
222 0.00114048545947298
223 0.00142105064878706
224 0.00118282850598916
225 0.00135677440266591
226 0.00109616399277002
227 0.00100296306482051
228 0.0012714011609205
229 0.00121312942064833
230 0.000865203910507262
231 0.00102122144016903
232 0.000944378320127726
233 0.00119962255121209
234 0.000811848542070948
235 0.0012215237293276
236 0.000678788921504747
237 0.000801962880359497
238 0.000862966262502596
239 0.000938216937356628
240 0.000702911158441566
241 0.000794749445049092
242 0.0011082958735642
243 0.000900910628843121
244 0.00117812336975476
245 0.000887952803168446
246 0.0007877708558226
247 0.000724274133972358
248 0.000666178959363606
249 0.000601429921516683
250 0.00120227912702831
251 0.00112700834870338
252 0.00131619788589887
253 0.000856070342706516
254 0.00094615196576342
255 0.00102153382613324
256 0.00156814715592191
257 0.00176002029911615
258 0.00163171993335709
259 0.00229111855151132
260 0.00235595542471856
261 0.00212520943023264
262 0.00278442190028727
263 0.00254908413626254
264 0.00267311697825789
265 0.00253765098750591
266 0.00250723282806575
267 0.00292530143633485
268 0.00380148948170245
269 0.00392251752782613
270 0.00219420436769724
271 0.00213468878064305
272 0.00264772109221667
273 0.00257384090218693
274 0.00256063893903047
275 0.00245606567477807
276 0.00248064153129235
277 0.00223635497968644
278 0.00307092466391623
279 0.00324003130663186
280 0.00225118780508637
281 0.00247246702201664
282 0.00268710684031248
283 0.00209560070652515
284 0.00178596249315888
};
\addplot [line width=2.5pt, red, opacity=1.0]
table {%
0 2.35028998057048
1 2.13261286417643
2 1.35797917842865
3 0.87364387512207
4 0.659817516803741
5 0.496957043806712
6 0.424989620844523
7 0.404787798722585
8 0.36013525724411
9 0.335566093524297
10 0.320433249076207
11 0.297929614782333
12 0.284065196911494
13 0.278493712345759
14 0.262783293922742
15 0.248939951260885
16 0.241361126303673
17 0.228521342078845
18 0.233914469679197
19 0.231416846315066
20 0.219469795624415
21 0.217710748314857
22 0.20993560552597
23 0.202201008796692
24 0.211468741297722
25 0.197849462429682
26 0.193402344981829
27 0.193136021494865
28 0.183455223838488
29 0.178908223907153
30 0.176074395577113
31 0.171795343359311
32 0.166227986415227
33 0.163091431061427
34 0.160037234425545
35 0.152298370997111
36 0.154220268130302
37 0.143995443979899
38 0.140185209612052
39 0.134032408396403
40 0.130572142700354
41 0.124403176208337
42 0.121180941661199
43 0.118237820764383
44 0.118226659794648
45 0.116587157050769
46 0.115383587777615
47 0.116707061727842
48 0.111810587346554
49 0.117488766709963
50 0.107146290441354
51 0.10418609281381
52 0.101384135584037
53 0.0988749985893567
54 0.0962186604738235
55 0.0936044851938884
56 0.0902722924947739
57 0.0894926562905312
58 0.0858059624830882
59 0.0864048004150391
60 0.0806802734732628
61 0.0792849076290925
62 0.0773219888408979
63 0.0757646386822065
64 0.0712999602158864
65 0.0678548067808151
66 0.0662808914979299
67 0.0648846564193567
68 0.0624014424781005
69 0.0628172556559245
70 0.0576771174867948
71 0.0565637573599815
72 0.0541239554683367
73 0.0500517624119918
74 0.0488097791870435
75 0.0449164801587661
76 0.0446595412989457
77 0.0438811170558135
78 0.0422779861837626
79 0.0402293652296066
80 0.0395252096156279
81 0.0352417069176833
82 0.0357897486537695
83 0.0376522385825713
84 0.0323481137553851
85 0.0311210434883833
86 0.0327977513273557
87 0.0312910272429387
88 0.0323754145453374
89 0.0298827812075615
90 0.0308569471041362
91 0.0281097205976645
92 0.0305090496937434
93 0.0303871364643176
94 0.0281639285385609
95 0.0281213180472453
96 0.0254315783580144
97 0.0251218061894178
98 0.023200485855341
99 0.0226329242189725
100 0.0226775420208772
101 0.0223375540226698
102 0.0214158532520135
103 0.0213449976096551
104 0.0185939104606708
105 0.0227496400475502
106 0.020925747230649
107 0.0235000488658746
108 0.0238950711985429
109 0.0228712968528271
110 0.0228133245060841
111 0.0250912625342607
112 0.0235175055762132
113 0.0220483839511871
114 0.0219680558269223
115 0.0214462721099456
116 0.0175596177577972
117 0.019654493778944
118 0.0200171150888006
119 0.0210147891193628
120 0.0232703865816196
121 0.0165215594073137
122 0.0176506421218316
123 0.0204393891617656
124 0.0194900076215466
125 0.0220573522771398
126 0.0221945888673266
127 0.0258688516914845
128 0.0263799925645192
129 0.0252796386679014
130 0.0261313480635484
131 0.0318347873787085
132 0.0317094201842944
133 0.030466158563892
134 0.0311357397586107
135 0.0333126944800218
136 0.0345033866663774
137 0.0381758461395899
138 0.0381175627311071
139 0.0353533116479715
140 0.0370921685049931
141 0.0396659970283508
142 0.0419848182549079
143 0.0646878611296415
144 0.0289967879652977
145 0.0243302981058757
146 0.0259325066581368
147 0.0221085650846362
148 0.0232285475358367
149 0.0228027322640022
150 0.0293192047004898
151 0.0285957992697756
152 0.0660062463333209
153 0.0356097913657626
154 0.0287271710112691
155 0.0231689081216852
156 0.0280762097487847
157 0.0279376491283377
158 0.0279679661616683
159 0.0273184484491746
160 0.0279695090527336
161 0.0258409694458048
162 0.0291360591848691
163 0.0300746193776528
164 0.0292739250386755
165 0.0270430659875274
166 0.0278189666569233
167 0.0245123726005356
168 0.0262256897985935
169 0.0223832555736105
170 0.0253031412139535
171 0.0265631914759676
172 0.0316114295274019
173 0.0250786772618691
174 0.0244934344664216
175 0.0281834717219075
176 0.0256534187744061
177 0.0244864750032624
178 0.0266494426565866
179 0.0259450180456042
180 0.0323315017546217
181 0.0300948622946938
182 0.0307660849454502
183 0.0319081513832013
184 0.0491670149688919
185 0.0476873766941329
186 0.047536107711494
187 0.0498577561229467
188 0.0481237452477217
189 0.064773067055891
190 0.0630305817661186
191 0.0702108396217227
192 0.0534877544268966
193 0.054522688811024
194 0.0560764353722334
195 0.0610162761683265
196 0.0657384898513556
197 0.062762185310324
198 0.0630335155874491
199 0.0517966958383719
200 0.0632563003649314
201 0.062446892571946
202 0.06071294285357
203 0.0564164016395807
204 0.0529858972877264
205 0.0510466533402602
206 0.0504481767614683
207 0.0477130903551976
208 0.0427805424357454
209 0.0395032335072756
210 0.0433235044280688
211 0.0396526729067167
212 0.0400575132419666
213 0.0436435878897707
214 0.0382780488580465
215 0.0287085759143035
216 0.0242071558410923
217 0.0211954548334082
218 0.0174639588221908
219 0.0192533635223905
220 0.019305929231147
221 0.018282770489653
222 0.0169275483737389
223 0.0148654862617453
224 0.0151400758574406
225 0.0161118137960633
226 0.0134797394275665
227 0.0128954049820701
228 0.0117896612112721
229 0.0122720853736003
230 0.0119553642968337
231 0.0113187053551277
232 0.0109240838016073
233 0.0103195297221343
234 0.0112956992040078
235 0.00984740722924471
236 0.0117258196696639
237 0.00997502729296684
238 0.0112340475122134
239 0.00940455356612802
240 0.00950045417994261
241 0.00937117791424195
242 0.00990582009156545
243 0.0110981619606415
244 0.00914101892461379
245 0.0091342666807274
246 0.00935823377221823
247 0.00879749873032173
248 0.0072632619800667
249 0.00808756317322453
250 0.00980804984768232
251 0.00752782480170329
252 0.00813519535586238
253 0.0078027150593698
254 0.00854884351914128
255 0.00763541553169489
256 0.00790985999628901
257 0.00887730236475666
258 0.00982825209697088
259 0.00812071841210127
260 0.00959197866419951
261 0.011026149460425
262 0.0110495781215529
263 0.0111360149458051
264 0.010303139531364
265 0.0100631235788266
266 0.00886893707017104
267 0.00886784711231788
268 0.00829570678373178
269 0.0090096794689695
270 0.00906736853842934
271 0.00749284137661258
272 0.00835126296927532
273 0.00926451229800781
274 0.00795223874350389
275 0.00932761499037345
276 0.00831332492331664
277 0.00820190894107024
278 0.00830076495185494
279 0.00793923034022252
280 0.00765263351301352
};
\addplot [line width=2.5pt, color2, opacity=1.0]
table {%
0 2.35028998057048
2 1.94272895654043
4 0.939676662286123
6 0.572696844736735
8 0.448777745167414
10 0.392280211051305
12 0.354969014724096
14 0.330157419045766
16 0.306536952654521
18 0.290643682082494
20 0.272132347027461
22 0.25736207763354
24 0.247568512956301
26 0.232402846217155
28 0.224933207035065
30 0.216056962807973
32 0.208177129427592
34 0.198070704936981
36 0.189366122086843
38 0.180261313915253
40 0.172133843104045
42 0.168541073799133
44 0.165446951985359
46 0.156599854429563
48 0.150110145409902
50 0.141942059000333
52 0.137895623842875
54 0.132468362649282
56 0.127392575144768
58 0.122152174512545
60 0.117685315509637
62 0.112587290505568
64 0.106176979839802
66 0.103366603453954
68 0.0984775498509407
70 0.0944852853814761
72 0.091382309794426
74 0.0865070149302483
76 0.0847250918547312
78 0.08038579672575
80 0.0749710301558177
82 0.075958251953125
84 0.0722439189751943
86 0.0694683119654655
88 0.0672032063206037
90 0.0648171094556649
92 0.0625758158663909
94 0.0601819331447283
96 0.0578438465793928
98 0.0550334205230077
100 0.0543043203651905
102 0.0528822280466557
104 0.0496907544632753
106 0.0493170768022537
108 0.0482505386074384
110 0.0462653425832589
112 0.0450955964624882
114 0.0440755660335223
116 0.0440923199057579
118 0.0402890791495641
120 0.0414324427644412
122 0.0397089955707391
124 0.0396895185112953
126 0.0374290446440379
128 0.0373244273165862
130 0.0363658517599106
132 0.035143606364727
134 0.0355543829500675
136 0.0347515617807706
138 0.034062930693229
140 0.0337701787551244
142 0.0332437170048555
144 0.033057210346063
146 0.0314379576593637
148 0.0311570595949888
150 0.0300248209387064
152 0.0301564397911231
154 0.030074971417586
156 0.0301571742941936
158 0.0295701511204243
160 0.0291604455560446
162 0.0281740197290977
164 0.0278508104383945
166 0.028492928793033
168 0.0265567575891813
170 0.0256839046875636
172 0.0261730471005042
174 0.025618930036823
176 0.0255815827598174
178 0.0264366250485182
180 0.0254384900132815
182 0.02551762573421
184 0.0245843188216289
186 0.023880056415995
188 0.0250024255365133
190 0.0250078892956177
192 0.0237645376473665
194 0.0232565496116877
196 0.0231471502532562
198 0.0234538931399584
200 0.0221209737161795
202 0.0228796352942785
204 0.0224339210738738
206 0.0225626404086749
208 0.0225858644892772
210 0.0218818013866742
212 0.0225925905009111
214 0.021405091509223
216 0.0214783934255441
218 0.0212941020727158
220 0.0212404696891705
222 0.0218184192975362
224 0.0206772952030102
226 0.0212442638973395
228 0.0213360457370679
230 0.0206312003235022
232 0.0199504780272643
234 0.0197138339281082
236 0.0205021457125743
238 0.0197543340424697
240 0.0211453884840012
242 0.0203030606110891
244 0.0193439659972986
246 0.0197171835849683
248 0.0193701299528281
250 0.0195317671944698
252 0.0189219936728477
254 0.0193986135224501
256 0.0196591621885697
258 0.0184363008787235
260 0.0192541150997082
262 0.0184716569880644
264 0.0184258129447699
266 0.0185252713660399
268 0.0186748125900825
270 0.0173226855695248
272 0.0174870540698369
274 0.0184012111276388
276 0.0175612270832062
278 0.0179789544393619
280 0.017946803321441
282 0.0174420742938916
284 0.0172534411152204
};
\end{axis}

\end{tikzpicture}

%% file: labpal/figure_data/performance_comparison/SVHN/SVHN_DenseNet-121_validation_accuracy.pgf
% This file was created by tikzplotlib v0.9.8.
\begin{tikzpicture}

\definecolor{color0}{rgb}{0.647058823529412,0.164705882352941,0.164705882352941}
\definecolor{color1}{rgb}{1,0.647058823529412,0}
\definecolor{color2}{rgb}{0.933333333333333,0.509803921568627,0.933333333333333}

\begin{axis}[
grid = major,
major grid style={dotted},
minor xtick={},
minor ytick={},
tick align=outside,
tick pos=left,
title={validation accuracy SVHN DenseNet-121},
width=10.5cm,height=8cm,,
x grid style={white!69.0196078431373!black},
xlabel={epochs},
xmin=-13.15, xmax=298.15,
xtick style={color=black},
xtick={-50,0,50,100,150,200,250,300},
y grid style={white!69.0196078431373!black},
ylabel={validation accuracy},
ymin=0.9, ymax=0.96,
ytick style={color=black},
minor y tick num=1,
]
\addplot [line width=2.5pt, color0, opacity=1.0]
table {%
1 0.460536847511927
2 0.574519236882528
3 0.641960461934408
4 0.695045411586761
5 0.707398494084676
6 0.724692841370901
7 0.760817309220632
8 0.772035260995229
9 0.771100401878357
10 0.789997339248657
11 0.787459929784139
12 0.801949799060822
13 0.80795939763387
14 0.817307690779368
15 0.82545405626297
16 0.824519236882528
17 0.83019498984019
18 0.827056606610616
19 0.825787921746572
20 0.835803965727488
21 0.823651174704234
22 0.836538473765055
23 0.842280964056651
24 0.848357359568278
25 0.842013895511627
26 0.833533644676208
27 0.845419347286224
28 0.852764427661896
29 0.849492530028025
30 0.852430562178294
31 0.840544859568278
32 0.849759618441264
33 0.854634086290995
34 0.859041134516398
35 0.855235040187836
36 0.855034728844961
37 0.857905964056651
38 0.851896365483602
39 0.855635662873586
40 0.845819969971975
41 0.854433755079905
42 0.83653845389684
43 0.854033132394155
44 0.857238252957662
45 0.854700863361359
46 0.853165050347646
47 0.852564116319021
48 0.854500532150269
49 0.856370190779368
50 0.847489317258199
51 0.841012279192607
52 0.855902771155039
53 0.858106315135956
54 0.858373383680979
55 0.858907580375671
56 0.851228634516398
57 0.847355763117472
58 0.851896365483602
59 0.853699247042338
60 0.836071054140727
61 0.852831184864044
62 0.860977570215861
63 0.858573714892069
64 0.860042730967204
65 0.865251064300537
66 0.860243062178294
67 0.857238233089447
68 0.861912389596303
69 0.746928413709005
70 0.852897981802622
71 0.857839206854502
72 0.859041134516398
73 0.836271385351817
74 0.847689648469289
75 0.847622871398926
76 0.847355763117472
77 0.847756425539652
78 0.85423344373703
79 0.861912389596303
80 0.857572098573049
81 0.85423344373703
82 0.857638895511627
83 0.849158644676208
84 0.857171475887299
85 0.856637299060822
86 0.867454588413239
87 0.850694437821706
88 0.870259086290995
89 0.869057158629099
90 0.868122319380442
91 0.859842419624329
92 0.863247851530711
93 0.858640491962433
94 0.871794859568278
95 0.860510150591532
96 0.860710461934408
97 0.872863233089447
98 0.869190692901611
99 0.857505341370901
100 0.867053965727488
101 0.86204594373703
102 0.868589739004771
103 0.868456204732259
104 0.85423344373703
105 0.850560903549194
106 0.861177881558736
107 0.619991992910703
108 0.872262299060822
109 0.869324247042338
110 0.872195502122243
111 0.870726505915324
112 0.878672540187836
113 0.877804497877757
114 0.881677329540253
115 0.353766031563282
116 0.864516536394755
117 0.874666134516398
118 0.869658132394155
119 0.872128744920095
120 0.870058755079905
121 0.878672540187836
122 0.877804478009542
123 0.883947630723318
124 0.869858463605245
125 0.876736124356588
126 0.879473825295766
127 0.87439904610316
128 0.883947670459747
129 0.876201907793681
130 0.875934839248657
131 0.882411857446035
132 0.887620190779368
133 0.886418282985687
134 0.626201927661896
135 0.635950853427251
136 0.871127148469289
137 0.76315438747406
138 0.872662941614787
139 0.881343464056651
140 0.620726488530636
141 0.873397429784139
142 0.872996787230174
143 0.751736104488373
144 0.869391004244486
145 0.871394236882528
146 0.865384618441264
147 0.877804478009542
148 0.85403311252594
149 0.884348293145498
150 0.839610040187836
151 0.559294859568278
152 0.874732931454976
153 0.872262279192607
154 0.878739337126414
155 0.879073182741801
156 0.872796475887299
157 0.876602570215861
158 0.871928413709005
159 0.869724889596303
160 0.883346696694692
161 0.880008002122243
162 0.878004809220632
163 0.872662941614787
164 0.888287941614787
165 0.867053945859273
166 0.876669347286224
167 0.885616978009542
168 0.883947670459747
169 0.872863252957662
170 0.888888895511627
171 0.879874467849731
172 0.879607379436493
173 0.886418263117472
174 0.879941244920095
175 0.884281516075134
176 0.753138363361359
177 0.877403835455577
178 0.877537409464518
179 0.876469016075134
180 0.878739297389984
181 0.887419859568278
182 0.879807690779368
183 0.601161867380142
184 0.877604166666667
185 0.879273494084676
186 0.868456204732259
187 0.88054217894872
188 0.883146365483602
189 0.883012811342875
190 0.70506144563357
191 0.87954060236613
192 0.889489849408468
193 0.767160793145498
194 0.873597760995229
195 0.883146385351817
196 0.728632479906082
197 0.78331998984019
198 0.875
199 0.668402776122093
200 0.868790070215861
201 0.606503742436568
202 0.870125552018484
203 0.876068373521169
204 0.789797027905782
205 0.755542198816935
206 0.653712605436643
207 0.877203524112701
208 0.874532580375671
209 0.871127148469289
210 0.886017640431722
211 0.894431094328562
212 0.849091867605845
213 0.865584929784139
214 0.844951927661896
215 0.867254277070363
216 0.788528303305308
217 0.874532600243886
218 0.529513890544573
219 0.839409728844961
220 0.855635682741801
221 0.684428413709005
222 0.881209929784139
223 0.88735310236613
224 0.887820521990458
225 0.880809307098389
226 0.770365913709005
227 0.623864861826102
228 0.757545411586761
229 0.615117520093918
230 0.634682153662046
231 0.874131957689921
232 0.875934839248657
233 0.880876044432322
234 0.66559828321139
235 0.824986636638641
236 0.873731315135956
237 0.844217419624329
238 0.882478654384613
239 0.862847228844961
240 0.778445521990458
241 0.607572123408318
242 0.613782040774822
243 0.567240911225478
244 0.530849357446035
245 0.621861654023329
246 0.625801295042038
247 0.547475951413314
248 0.625066791971525
};
\addplot [line width=2.5pt, green!50.1960784313725!black, opacity=1.0]
table {%
1 0.301929454008738
2 0.535984843969345
3 0.727047840754191
4 0.744093279043833
5 0.844401021798452
6 0.868158141771952
7 0.886624077955882
8 0.893134474754333
9 0.904876887798309
10 0.911008516947428
11 0.909682750701904
12 0.917069137096405
13 0.919756174087524
14 0.917376895745595
15 0.920608441034953
16 0.925733884175619
17 0.925698379675547
18 0.922644416491191
19 0.92339015007019
20 0.929450750350952
21 0.93030301729838
22 0.931356529394786
23 0.929154833157857
24 0.929865062236786
25 0.932634949684143
26 0.929604649543762
27 0.93209042151769
28 0.930409550666809
29 0.931770821412404
30 0.930741012096405
31 0.931261837482452
32 0.933771312236786
33 0.932552099227905
34 0.934256613254547
35 0.926598012447357
36 0.934161941210429
37 0.933842341105143
38 0.934469699859619
39 0.93585463364919
40 0.932729641596476
41 0.933191279570262
42 0.936446487903595
43 0.934860328833262
44 0.936754246552785
45 0.936517516771952
46 0.933723966280619
47 0.935049712657928
48 0.934966862201691
49 0.935061554114024
50 0.934280296166738
51 0.934966842333476
52 0.933321495850881
53 0.934955020745595
54 0.9337357878685
55 0.936044037342072
56 0.936541199684143
57 0.935203611850739
58 0.934812963008881
59 0.937926133473714
60 0.935108880201975
61 0.934706449508667
62 0.93645832935969
63 0.932125965754191
64 0.936233421166738
65 0.934339503447215
66 0.936209758122762
67 0.936564862728119
68 0.935795466105143
69 0.935606062412262
70 0.930255671342214
71 0.936718742052714
72 0.935570557912191
73 0.936233421166738
74 0.936789770921071
75 0.935677071412404
76 0.936955491701762
77 0.937097549438477
78 0.935073395570119
79 0.93585463364919
80 0.939701696236928
81 0.937713066736857
82 0.9368963042895
83 0.93779593706131
84 0.937050183614095
85 0.93854167064031
86 0.936695079008738
87 0.937902450561523
88 0.938671867052714
89 0.936339974403381
90 0.938020825386047
91 0.938636362552643
92 0.937772254149119
93 0.938068171342214
94 0.938151041666667
95 0.938731054464976
96 0.938364108403524
97 0.939216395219167
98 0.939512332280477
99 0.938127378622691
100 0.937926133473714
101 0.938624521096547
102 0.937784095605214
103 0.93584281206131
104 0.934374988079071
105 0.938281257947286
106 0.937251428763072
107 0.939121703306834
108 0.938802103201548
109 0.938494324684143
110 0.939713537693024
111 0.939464966456095
112 0.939784566561381
113 0.938328603903453
114 0.941240529219309
115 0.93946494658788
116 0.939985791842143
117 0.938636382420858
118 0.939535995324453
119 0.940530300140381
120 0.939879258473714
121 0.939796388149261
122 0.939121703306834
123 0.939997633298238
124 0.93974906206131
125 0.941347062587738
126 0.939985791842143
127 0.939808249473572
128 0.94170218706131
129 0.941133995850881
130 0.940826237201691
131 0.941785037517548
132 0.940269887447357
133 0.940873583157857
134 0.941216846307119
135 0.939855595429738
136 0.941039303938548
137 0.939749042193095
138 0.940624992052714
139 0.941607475280762
140 0.941512783368429
141 0.940435608228048
142 0.940530320008596
143 0.942992428938548
144 0.9435014128685
145 0.945383528868357
146 0.94532436132431
147 0.946141103903453
148 0.946070075035095
149 0.946212112903595
150 0.945419033368429
151 0.945975383122762
152 0.945845186710358
153 0.945655783017476
154 0.946129262447357
155 0.946721136569977
156 0.945880671342214
157 0.946105579535166
158 0.946721116701762
159 0.946768462657928
160 0.946117420991262
161 0.94695786635081
162 0.946543574333191
163 0.946531732877096
164 0.945939878622691
165 0.946590900421143
166 0.946259478727976
167 0.9459161957105
168 0.947052578131358
169 0.947064399719238
170 0.946294963359833
171 0.946579078833262
172 0.94695786635081
173 0.946508049964905
174 0.946342329184214
175 0.947194596131643
176 0.946981529394786
177 0.946957846482595
178 0.946898678938548
179 0.946768462657928
180 0.947005212306976
181 0.947443187236786
182 0.946981509526571
183 0.946768482526143
184 0.946981549263
185 0.946472545464834
186 0.947277466456095
187 0.947076221307119
188 0.947526017824809
189 0.947111745675405
190 0.947608888149261
191 0.946946024894714
192 0.946803987026215
193 0.947289288043976
194 0.946768462657928
195 0.946413358052572
196 0.947774628798167
197 0.946792145570119
198 0.947289307912191
199 0.947301129500071
200 0.947147250175476
201 0.947147250175476
202 0.946981529394786
203 0.947443187236786
204 0.947383999824524
205 0.947064399719238
206 0.947289288043976
207 0.948094228903453
208 0.946922341982524
209 0.947537859280904
210 0.946910500526428
211 0.946792145570119
212 0.946874996026357
213 0.947585244973501
214 0.947088062763214
215 0.947715441385905
216 0.946910500526428
217 0.947549700737
218 0.946673770745595
219 0.947833816210429
220 0.946898678938548
221 0.948046882947286
222 0.947644412517548
223 0.946496208508809
224 0.9471235871315
225 0.947064399719238
226 0.947585244973501
227 0.947691758473714
228 0.94757338364919
229 0.948023200035095
230 0.947360316912333
231 0.947312970956167
232 0.947064399719238
233 0.947857479254405
234 0.946638246377309
235 0.947668095429738
236 0.946922361850739
237 0.9478693207105
238 0.947490533192953
239 0.946780284245809
240 0.947194596131643
241 0.948129753271739
242 0.947088062763214
243 0.947810133298238
244 0.947833816210429
245 0.947549720605214
246 0.947336653868357
247 0.947324812412262
248 0.94770359992981
249 0.947135408719381
250 0.9474076628685
251 0.947218279043833
252 0.947490533192953
253 0.947336653868357
254 0.947585244973501
255 0.947312990824381
256 0.946389695008596
257 0.947277466456095
258 0.947289307912191
259 0.947218259175619
260 0.947443167368571
261 0.94770359992981
262 0.947455028692881
263 0.947206457455953
264 0.94711172580719
265 0.947265625
266 0.947218279043833
267 0.947845637798309
268 0.947052558263143
269 0.947372158368429
270 0.947360316912333
271 0.946874996026357
272 0.947727262973785
273 0.947881162166595
274 0.94757338364919
275 0.947230100631714
276 0.947206437587738
277 0.947857479254405
278 0.947668075561523
279 0.947478691736857
280 0.947762807210286
281 0.947632590929667
282 0.947999536991119
283 0.9474076628685
284 0.947727262973785
};
\addplot [line width=2.5pt, black, opacity=1.0]
table {%
1 0.198330968618393
2 0.344258993864059
3 0.510321974754333
4 0.649621208508809
5 0.76020359992981
6 0.769045929114024
7 0.794140617052714
8 0.817613641421
9 0.845667600631714
10 0.875816782315572
11 0.876834750175476
12 0.865767041842143
13 0.898094216982524
14 0.890340924263
15 0.899242420991262
16 0.912073850631714
17 0.898484845956167
18 0.908629258473714
19 0.909292121728261
20 0.908889671166738
21 0.912819604078929
22 0.921235799789429
23 0.926171898841858
24 0.923816283543905
25 0.914405763149261
26 0.906806329886119
27 0.912322421868642
28 0.930160999298096
29 0.928432762622833
30 0.925556361675262
31 0.920703132947286
32 0.926242907842
33 0.926893929640452
34 0.919377366701762
35 0.933676620324453
36 0.925224900245667
37 0.926976799964905
38 0.924313445885976
39 0.9271306792895
40 0.922632575035095
41 0.930042604605357
42 0.931273678938548
43 0.928515613079071
44 0.930468757947286
45 0.933984378973643
46 0.933960696061452
47 0.940068662166595
48 0.938648184140523
49 0.932836174964905
50 0.923200746377309
51 0.92292849222819
52 0.937334279219309
53 0.937760392824809
54 0.885688920815786
55 0.938210209210714
56 0.938742895921071
57 0.934221108754476
58 0.927083333333333
59 0.928432762622833
60 0.918039778868357
61 0.934812982877096
62 0.926420470078786
63 0.934753795464834
64 0.937215904394786
65 0.937381625175476
66 0.931924700737
67 0.932883520921071
68 0.934611737728119
69 0.934670945008596
70 0.937334259351095
71 0.939985791842143
72 0.943998575210571
73 0.943678975105286
74 0.942862232526143
75 0.9400568207105
76 0.934564391771952
77 0.934410512447357
78 0.929817716280619
79 0.929651995499929
80 0.9381036957105
81 0.932362675666809
82 0.933901528517405
83 0.935949325561523
84 0.933439870675405
85 0.935310145219167
86 0.936422824859619
87 0.941074808438619
88 0.939476807912191
89 0.943288365999858
90 0.942672828833262
91 0.941512783368429
92 0.941832383473714
93 0.941654821236928
94 0.941631158192953
95 0.939595182736715
96 0.9405184785525
97 0.942305862903595
98 0.942743837833405
99 0.942234853903453
100 0.942542612552643
101 0.945845166842143
102 0.944850842158
103 0.944803496201833
104 0.943169991175334
105 0.94441286722819
106 0.944850862026215
107 0.944921871026357
108 0.945099413394928
109 0.944732467333476
110 0.944176157315572
111 0.945714970429738
112 0.945999046166738
113 0.945087591807047
114 0.945430874824524
115 0.944495757420858
116 0.945987204710642
117 0.945880691210429
118 0.945750455061595
119 0.946508049964905
120 0.945987204710642
121 0.945146779219309
122 0.945643941561381
123 0.946342329184214
124 0.946034570535024
125 0.945916215578715
126 0.945549249649048
127 0.94501656293869
128 0.945667624473572
129 0.94546639919281
130 0.94621213277181
131 0.946010887622833
132 0.945359845956167
133 0.946152925491333
134 0.946389675140381
135 0.945726811885834
136 0.946034590403239
137 0.94607009490331
138 0.945833325386047
139 0.946093757947286
140 0.94622395435969
141 0.945868849754333
142 0.946129262447357
143 0.945975363254547
144 0.945359845956167
145 0.946046411991119
146 0.946851332982381
147 0.946318646272024
148 0.946650087833405
149 0.945963541666667
150 0.946673770745595
151 0.947017033894857
152 0.946046411991119
153 0.946934183438619
154 0.946638266245524
155 0.946401516596476
156 0.946957846482595
157 0.946448862552643
158 0.946377833684286
159 0.946472545464834
160 0.945608417193095
161 0.946531713008881
162 0.94635417064031
163 0.946259458859762
164 0.946792145570119
165 0.945785979429881
166 0.945714950561523
167 0.945395370324453
168 0.946129262447357
169 0.946164766947428
170 0.946780304114024
171 0.946922361850739
172 0.946531732877096
173 0.946543554464976
174 0.94635417064031
175 0.946496208508809
176 0.945904354254405
177 0.946271300315857
178 0.945561071236928
179 0.946756621201833
180 0.946472545464834
181 0.946993370850881
182 0.946330487728119
183 0.947289288043976
184 0.945857028166453
185 0.946389675140381
186 0.946342329184214
187 0.946247617403666
188 0.946176608403524
189 0.946638266245524
190 0.946567237377167
191 0.946555395921071
192 0.94622395435969
193 0.947099904219309
194 0.946188429991404
195 0.947005212306976
196 0.946697433789571
197 0.947017033894857
198 0.946886837482452
199 0.946815808614095
200 0.947194596131643
201 0.946863154570262
202 0.946330507596334
203 0.947028875350952
204 0.946721116701762
205 0.946519891421
206 0.947893003622691
207 0.946034590403239
208 0.946531713008881
209 0.947064399719238
210 0.946010887622833
211 0.947502374649048
212 0.946389675140381
213 0.946567237377167
214 0.946318666140238
215 0.946555395921071
216 0.946863154570262
217 0.946425179640452
218 0.947135408719381
219 0.946460704008738
220 0.94622395435969
221 0.946614583333333
222 0.946519871552785
223 0.946590900421143
224 0.946981549263
225 0.946058233579
226 0.946448882420858
227 0.946555395921071
228 0.9466619292895
229 0.946330487728119
230 0.94682765007019
231 0.947549700737
232 0.947088082631429
233 0.946306824684143
234 0.946567217508952
235 0.946460704008738
236 0.946638266245524
237 0.946922341982524
238 0.946685612201691
239 0.946697453657786
240 0.946721096833547
241 0.945856988430023
242 0.945714950561523
243 0.947052538394928
244 0.947182754675547
245 0.946673770745595
246 0.947679917017619
247 0.946709275245667
248 0.946223974227905
249 0.946886817614237
250 0.946602741877238
251 0.946212112903595
252 0.946874996026357
253 0.946638246377309
254 0.946567237377167
255 0.946472545464834
256 0.946389675140381
257 0.947088082631429
258 0.946756621201833
259 0.946685612201691
260 0.946046392122904
261 0.946946024894714
262 0.946756621201833
263 0.946081896622976
264 0.946330487728119
265 0.94592801729838
266 0.946579058965047
267 0.948129713535309
268 0.946425179640452
269 0.946863174438477
270 0.946946005026499
271 0.946744779745738
272 0.946579058965047
273 0.9471235871315
274 0.946673770745595
275 0.946614603201548
276 0.946318646272024
277 0.94725380341212
278 0.946010907491048
279 0.946330487728119
280 0.946969707806905
281 0.946673770745595
282 0.946567257245382
283 0.946342329184214
284 0.946389675140381
};
\addplot [line width=2.5pt, color1, opacity=1.0]
table {%
1 0.189926614363988
2 0.174277936418851
3 0.159588068723679
4 0.15629735092322
5 0.174384464820226
6 0.159126415848732
7 0.189382101098696
8 0.174597536524137
9 0.169874529043833
10 0.220916186769803
11 0.255172828833262
12 0.368513251344363
13 0.461328124006589
14 0.468300186097622
15 0.553267046809196
16 0.555871210992336
17 0.57559185475111
18 0.616323386629423
19 0.619685133298238
20 0.630184650421143
21 0.638944134116173
22 0.645359843969345
23 0.650390615065893
24 0.65387074649334
25 0.66363637149334
26 0.665909086664518
27 0.693323860565821
28 0.765672345956167
29 0.823686083157857
30 0.845087587833405
31 0.861008524894714
32 0.883629262447357
33 0.87696494658788
34 0.870016574859619
35 0.892365058263143
36 0.890234390894572
37 0.89514676729838
38 0.894933720429738
39 0.903172334035238
40 0.902521312236786
41 0.910712579886119
42 0.909067233403524
43 0.906439383824666
44 0.912464499473572
45 0.916832387447357
46 0.912831445535024
47 0.914535999298096
48 0.911434670289358
49 0.914512316385905
50 0.917767524719238
51 0.918667137622833
52 0.919436554114024
53 0.920762300491333
54 0.919164299964905
55 0.915281732877096
56 0.923330962657928
57 0.917732020219167
58 0.923413832982381
59 0.922431329886119
60 0.925733904043833
61 0.925437986850739
62 0.927355587482452
63 0.925804932912191
64 0.926574349403381
65 0.923638721307119
66 0.930575291315714
67 0.929403384526571
68 0.929936091105143
69 0.930042604605357
70 0.927935600280762
71 0.92924952507019
72 0.929971595605214
73 0.92969932158788
74 0.925378779570262
75 0.928231537342072
76 0.931107958157857
77 0.933783153692881
78 0.930030783017476
79 0.93148672580719
80 0.930480599403381
81 0.933049241701762
82 0.932765165964762
83 0.929273207982381
84 0.932528416315714
85 0.933037420113881
86 0.932220657666524
87 0.934978703657786
88 0.93464724222819
89 0.932516574859619
90 0.933925191561381
91 0.934836645921071
92 0.9344815214475
93 0.934150099754333
94 0.93420926729838
95 0.93449338277181
96 0.935286462306976
97 0.934872150421143
98 0.936695059140523
99 0.935937484105428
100 0.935830950737
101 0.934824804464976
102 0.934244771798452
103 0.934292137622833
104 0.934694608052572
105 0.935061554114024
106 0.93599667151769
107 0.935558716456095
108 0.937121212482452
109 0.933345178763072
110 0.934256633122762
111 0.934919496377309
112 0.935416678587596
113 0.936742424964905
114 0.93584281206131
115 0.936789770921071
116 0.935771783192953
117 0.93539297580719
118 0.935108919938405
119 0.936067720254262
120 0.936837116877238
121 0.934173782666524
122 0.935310125350952
123 0.934919516245524
124 0.936055858929952
125 0.936186075210571
126 0.936481992403666
127 0.935641566912333
128 0.937310596307119
129 0.937973479429881
130 0.937369783719381
131 0.936541199684143
132 0.938600838184357
133 0.937298774719238
134 0.937511841456095
135 0.937085707982381
136 0.937239567438761
137 0.937097529570262
138 0.937073846658071
139 0.936955491701762
140 0.936742424964905
141 0.934576213359833
142 0.933712124824524
143 0.936588525772095
144 0.93644650777181
145 0.936979154745738
146 0.934789299964905
147 0.93660036722819
148 0.937606533368429
149 0.937760432561239
150 0.936908145745595
151 0.936837116877238
152 0.938186566034953
153 0.937784075737
154 0.937050183614095
155 0.937464495499929
156 0.938518007596334
157 0.937997162342072
158 0.937866965929667
159 0.936221579710642
160 0.933960715929667
161 0.936813453833262
162 0.936422824859619
163 0.937535524368286
164 0.937251408894857
165 0.937949816385905
166 0.936186075210571
167 0.936860799789429
168 0.936991016070048
169 0.936493853727976
170 0.936943650245667
171 0.936138729254405
172 0.937239587306976
173 0.937417129675547
174 0.935736258824666
175 0.936055878798167
176 0.939263741175334
177 0.93706202507019
178 0.937784095605214
179 0.938222070535024
180 0.937535504500071
181 0.937263250350952
182 0.935582379500071
183 0.935771783192953
184 0.938044508298238
185 0.936777949333191
186 0.936067720254262
187 0.935511350631714
188 0.93780775864919
189 0.936339974403381
190 0.938151041666667
191 0.939204553763072
192 0.937346120675405
193 0.937997142473857
194 0.936505675315857
195 0.937997142473857
196 0.937523682912191
197 0.936576684315999
198 0.935274620850881
199 0.938399612903595
200 0.939026991526286
201 0.938198387622833
202 0.936055878798167
203 0.936257123947144
204 0.940329074859619
205 0.938707391421
206 0.937772254149119
207 0.938790241877238
208 0.938032686710358
209 0.936789790789286
210 0.938080012798309
211 0.937535504500071
212 0.93720406293869
213 0.938529829184214
214 0.936766107877096
215 0.937689403692881
216 0.936091403166453
217 0.937334279219309
218 0.938458800315857
219 0.938636362552643
220 0.938009003798167
221 0.937026500701904
222 0.939666191736857
223 0.938091854254405
224 0.937736729780833
225 0.939441283543905
226 0.936553021272024
227 0.937334259351095
228 0.939180870850881
229 0.938458820184072
230 0.938009003798167
231 0.938233911991119
232 0.938328584035238
233 0.938920458157857
234 0.9393110871315
235 0.938328603903453
236 0.93841145435969
237 0.939808229605357
238 0.936375478903453
239 0.939038832982381
240 0.937440812587738
241 0.937736749649048
242 0.940210700035095
243 0.938435137271881
244 0.938127358754476
245 0.940743347009023
246 0.938269416491191
247 0.939488649368286
248 0.935795466105143
249 0.935866494973501
250 0.937866946061452
251 0.938932299613953
252 0.937381625175476
253 0.937144875526428
254 0.939879278341929
255 0.937772254149119
256 0.937819600105286
257 0.938671847184499
258 0.939701696236928
259 0.93854167064031
260 0.937618374824524
261 0.938091854254405
262 0.938683708508809
263 0.938896795113881
264 0.937594691912333
265 0.937665720780691
266 0.9368963042895
267 0.938340445359548
268 0.937168558438619
269 0.939370254675547
270 0.935913840929667
271 0.937594691912333
272 0.938162883122762
273 0.938707391421
274 0.939074357350667
275 0.939240058263143
276 0.938648204008738
277 0.937535504500071
278 0.939583341280619
279 0.940553963184357
280 0.938577175140381
281 0.937263250350952
282 0.937665720780691
283 0.939429461956024
284 0.938944141070048
};
\addplot [line width=2.5pt, blue, opacity=1.0]
table {%
1 0.358718052506447
2 0.583416193723679
3 0.682528406381607
4 0.81731179356575
5 0.862659782171249
6 0.878284782171249
7 0.874662637710571
8 0.879847288131714
9 0.897638499736786
10 0.903870731592178
11 0.909818887710571
12 0.8976029753685
13 0.915056794881821
14 0.91301491856575
15 0.917631387710571
16 0.917258530855179
17 0.920827388763428
18 0.919282674789429
19 0.917915493249893
20 0.916654855012894
21 0.906853705644608
22 0.907723724842072
23 0.918892055749893
24 0.913121461868286
25 0.920028418302536
26 0.925390630960464
27 0.926278412342072
28 0.929811775684357
29 0.9296875
30 0.925035506486893
31 0.932368606328964
32 0.927787661552429
33 0.930752843618393
34 0.933451682329178
35 0.932634949684143
36 0.9268998503685
37 0.932155549526215
38 0.928409099578857
39 0.930149167776108
40 0.930948138237
41 0.927077412605286
42 0.930042594671249
43 0.930397719144821
44 0.934055387973785
45 0.925887793302536
46 0.926100850105286
47 0.578338079154491
48 0.90241476893425
49 0.929119318723679
50 0.928977251052856
51 0.926491469144821
52 0.930326700210571
53 0.935724437236786
54 0.931995749473572
55 0.930415481328964
56 0.932368606328964
57 0.933771312236786
58 0.931321024894714
59 0.93645241856575
60 0.935724437236786
61 0.932794719934464
62 0.936345875263214
63 0.932634949684143
64 0.931711673736572
65 0.933984369039536
66 0.932297587394714
67 0.935600131750107
68 0.935813218355179
69 0.937837362289429
70 0.937855124473572
71 0.938725143671036
72 0.935955256223679
73 0.939240038394928
74 0.937215924263
75 0.936700969934464
76 0.937748581171036
77 0.936363637447357
78 0.935582369565964
79 0.937198132276535
80 0.938796162605286
81 0.937002837657928
82 0.937801867723465
83 0.938103705644608
84 0.937393486499786
85 0.938458800315857
86 0.939186811447144
87 0.938210219144821
88 0.939506381750107
89 0.941139906644821
90 0.939719468355179
91 0.939080268144608
92 0.938512086868286
93 0.935919761657715
94 0.939275562763214
95 0.939754962921143
96 0.93924006819725
97 0.938387781381607
98 0.940767049789429
99 0.940323144197464
100 0.938778430223465
101 0.940127849578857
102 0.941619336605072
103 0.940589487552643
104 0.942151993513107
105 0.942347288131714
106 0.942684650421143
107 0.942276298999786
108 0.944016337394714
109 0.941512793302536
110 0.942862212657928
111 0.942666888237
112 0.942933231592178
113 0.939897000789642
114 0.940926849842072
115 0.942808955907822
116 0.942631393671036
117 0.943164050579071
118 0.944105118513107
119 0.943412631750107
120 0.942205250263214
121 0.93938210606575
122 0.938636362552643
123 0.940891325473785
124 0.938387781381607
125 0.937766343355179
126 0.939098000526428
127 0.935227274894714
128 0.939754962921143
129 0.938370019197464
130 0.939417630434036
131 0.935653418302536
132 0.938192456960678
133 0.9383345246315
134 0.93840554356575
135 0.941086649894714
136 0.939222306013107
137 0.940589487552643
138 0.940074563026428
139 0.940145611763
140 0.938867211341858
141 0.930184662342072
142 0.938068181276321
143 0.9453125
144 0.945134937763214
145 0.945703119039536
146 0.944602251052856
147 0.945028424263
148 0.944975137710571
149 0.944353669881821
150 0.9453125
151 0.943004280328751
152 0.942844450473785
153 0.945507794618607
154 0.945152699947357
155 0.944850832223892
156 0.943625688552856
157 0.944389194250107
158 0.944460213184357
159 0.944549024105072
160 0.945081681013107
161 0.945294737815857
162 0.944833099842072
163 0.944673299789429
164 0.943306118249893
165 0.944158375263214
166 0.943022012710571
167 0.94426491856575
168 0.943838775157928
169 0.944921880960464
170 0.942666918039322
171 0.945188194513321
172 0.94231179356575
173 0.941992163658142
174 0.938955992460251
175 0.939098000526428
176 0.940465182065964
177 0.939577430486679
178 0.9405717253685
179 0.942968755960464
180 0.941441774368286
181 0.938565343618393
182 0.94140625
183 0.941122144460678
184 0.9425248503685
185 0.940980136394501
186 0.939950287342072
187 0.943856507539749
188 0.9402876496315
189 0.941637068986893
190 0.940749287605286
191 0.939897030591965
192 0.940376430749893
193 0.942169725894928
194 0.941956669092178
195 0.940234392881393
196 0.940891325473785
197 0.939772725105286
198 0.9376420378685
199 0.941317468881607
200 0.94231179356575
201 0.938831657171249
202 0.942560374736786
203 0.942098736763
204 0.940678268671036
205 0.942950993776321
206 0.942542612552643
207 0.941530525684357
208 0.942294031381607
209 0.94259586930275
210 0.941725850105286
211 0.942187517881393
212 0.94140625
213 0.940767049789429
214 0.944087356328964
215 0.944708794355392
216 0.946821719408035
217 0.945827424526215
218 0.946377843618393
219 0.947283357381821
220 0.948597311973572
221 0.947212368249893
222 0.948224455118179
223 0.94845524430275
224 0.947993606328964
225 0.948046892881393
226 0.948206692934036
227 0.947159081697464
228 0.947780519723892
229 0.947389930486679
230 0.948401987552643
231 0.9483842253685
232 0.948561787605286
233 0.948579549789429
234 0.947833806276321
235 0.948757082223892
236 0.948597311973572
237 0.948686093091965
238 0.947798311710358
239 0.948206692934036
240 0.947727262973785
241 0.947017043828964
242 0.946626424789429
243 0.947372168302536
244 0.947105824947357
245 0.947851568460464
246 0.947869330644608
247 0.948029100894928
248 0.947958111763
249 0.947745025157928
250 0.94705256819725
251 0.947389930486679
252 0.947230100631714
253 0.946928262710571
254 0.947034806013107
255 0.947585225105286
256 0.945205986499786
257 0.945543318986893
258 0.946608662605286
259 0.945880681276321
260 0.944016337394714
261 0.94552555680275
262 0.945507824420929
263 0.945472300052643
264 0.943554699420929
265 0.946075975894928
266 0.944779843091965
267 0.945614337921143
268 0.943767756223679
269 0.9441938996315
270 0.943980842828751
271 0.944655537605286
272 0.943572461605072
273 0.945223718881607
274 0.944797575473785
275 0.944105118513107
276 0.945916205644608
277 0.944797605276108
278 0.943696737289429
279 0.944140642881393
280 0.94552555680275
281 0.945543348789215
282 0.943643480539322
283 0.94621804356575
284 0.945294737815857
};
\addplot [line width=2.5pt, red, opacity=1.0]
table {%
1 0.423212597767512
2 0.637428979078929
3 0.718181808789571
4 0.802355587482452
5 0.852130691210429
6 0.797147254149119
7 0.854640146096547
8 0.879237691561381
9 0.874147733052572
10 0.885168095429738
11 0.903539299964905
12 0.890980124473572
13 0.906652450561523
14 0.911138733228048
15 0.907232483228048
16 0.919282674789429
17 0.920762300491333
18 0.9158735871315
19 0.916370729605357
20 0.914429446061452
21 0.913352270921071
22 0.918809195359548
23 0.916252374649048
24 0.918726325035095
25 0.919779817263285
26 0.924751440684001
27 0.924928983052572
28 0.923307279745738
29 0.928136845429738
30 0.930184662342072
31 0.9200639128685
32 0.929971575737
33 0.929438928763072
34 0.931806345780691
35 0.929438908894857
36 0.929296871026357
37 0.929758528868357
38 0.932386358579
39 0.933948854605357
40 0.934019883473714
41 0.934765617052714
42 0.93388968706131
43 0.933475375175476
44 0.932623108228048
45 0.935700754324595
46 0.931131641070048
47 0.932836174964905
48 0.935014228026072
49 0.934990545113881
50 0.934422353903453
51 0.936008512973785
52 0.9310369292895
53 0.936233441034953
54 0.934268474578857
55 0.936825295289358
56 0.935759941736857
57 0.934209287166595
58 0.936375478903453
59 0.934552550315857
60 0.937713086605072
61 0.937144895394643
62 0.936801612377167
63 0.937997162342072
64 0.938222070535024
65 0.936304450035095
66 0.9375
67 0.936624050140381
68 0.938482483228048
69 0.937571009000142
70 0.936091363430023
71 0.935594220956167
72 0.937582850456238
73 0.940411925315857
74 0.938222070535024
75 0.938731054464976
76 0.940163354078929
77 0.940376420815786
78 0.940909067789714
79 0.940411945184072
80 0.942660987377167
81 0.940767029921214
82 0.938908616701762
83 0.941382586956024
84 0.940920929114024
85 0.940672357877096
86 0.941287875175476
87 0.940423786640167
88 0.941276033719381
89 0.942294041315714
90 0.943312029043833
91 0.940980116526286
92 0.941015621026357
93 0.941145837306976
94 0.943276524543762
95 0.942057291666667
96 0.943300187587738
97 0.941098471482595
98 0.943856537342072
99 0.942874054114024
100 0.942708333333333
101 0.9439630707105
102 0.943761845429738
103 0.943643470605214
104 0.944057762622833
105 0.944318195184072
106 0.945584754149119
107 0.944673299789429
108 0.945442696412404
109 0.944377362728119
110 0.943726321061452
111 0.944625933965047
112 0.944554924964905
113 0.945359826087952
114 0.945028404394786
115 0.945336182912191
116 0.945501903692881
117 0.945999046166738
118 0.944472054640452
119 0.946425199508667
120 0.945336163043976
121 0.946460704008738
122 0.945845166842143
123 0.946058233579
124 0.944507579008738
125 0.94530063867569
126 0.946792145570119
127 0.944294492403666
128 0.945525566736857
129 0.945549249649048
130 0.944815337657928
131 0.945016582806905
132 0.945182303587596
133 0.944685121377309
134 0.944661438465118
135 0.944543103377024
136 0.943892061710358
137 0.944069604078929
138 0.945134937763214
139 0.944448391596476
140 0.944602270921071
141 0.944152474403381
142 0.945596595605214
143 0.947005212306976
144 0.945916175842285
145 0.947206437587738
146 0.946425199508667
147 0.947798311710358
148 0.947691758473714
149 0.945667624473572
150 0.945738633473714
151 0.945265154043833
152 0.946401536464691
153 0.945703128973643
154 0.944696962833405
155 0.944910049438477
156 0.946188449859619
157 0.946484386920929
158 0.946709275245667
159 0.946650087833405
160 0.945726792017619
161 0.946792125701904
162 0.946780323982239
163 0.946046392122904
164 0.945845186710358
165 0.946070075035095
166 0.945916175842285
167 0.94711172580719
168 0.94546639919281
169 0.944945553938548
170 0.946294983228048
171 0.947786450386047
172 0.948141554991404
173 0.947277466456095
174 0.944424708684286
175 0.947253783543905
176 0.946709275245667
177 0.94636599222819
178 0.946886857350667
179 0.946496208508809
180 0.946401516596476
181 0.946034570535024
182 0.947490533192953
183 0.946081896622976
184 0.945785979429881
185 0.944022258122762
186 0.943892041842143
187 0.943998595078786
188 0.944637775421143
189 0.945028404394786
190 0.944116950035095
191 0.943868378798167
192 0.943406740824381
193 0.944732487201691
194 0.943347533543905
195 0.942578136920929
196 0.937464495499929
197 0.942909558614095
198 0.943454086780548
199 0.943560600280762
200 0.941891570885976
201 0.941879749298096
202 0.941098471482595
203 0.944519420464834
204 0.944187978903453
205 0.94455490509669
206 0.943383057912191
207 0.943915724754333
208 0.943915724754333
209 0.942980567614237
210 0.944590429464976
211 0.942660987377167
212 0.943619787693024
213 0.944483896096547
214 0.943998575210571
215 0.943915704886119
216 0.944235324859619
217 0.944318195184072
218 0.944614092508952
219 0.944850842158
220 0.944614112377167
221 0.945099453131358
222 0.944898207982381
223 0.945809682210286
224 0.945833325386047
225 0.945134957631429
226 0.945241490999858
227 0.945288817087809
228 0.946022748947144
229 0.945241471131643
230 0.944886366526286
231 0.945549249649048
232 0.944992919762929
233 0.946342329184214
234 0.945750494798025
235 0.945407211780548
236 0.945584774017334
237 0.94650806983312
238 0.945276975631714
239 0.945785979429881
240 0.945632080237071
241 0.945845146973928
242 0.946602741877238
243 0.944862683614095
244 0.946164766947428
245 0.946034550666809
246 0.945584754149119
247 0.944803496201833
248 0.945371667544047
249 0.945194125175476
250 0.945928037166595
251 0.944543083508809
252 0.945537408192953
253 0.945383528868357
254 0.946105599403381
255 0.946046392122904
256 0.945359845956167
257 0.944625953833262
258 0.945501903692881
259 0.945501883824666
260 0.944886366526286
261 0.94635417064031
262 0.944566766421
263 0.944472074508667
264 0.943974912166595
265 0.945714970429738
266 0.944969216982524
267 0.945300658543905
268 0.946590920289358
269 0.944957375526428
270 0.945774157842
271 0.945632100105286
272 0.944744328657786
273 0.945324341456095
274 0.945194145043691
275 0.945419033368429
276 0.94487452507019
277 0.945004721482595
278 0.945253312587738
279 0.945442716280619
280 0.946342329184214
};
\addplot [line width=2.5pt, color2, opacity=1.0]
table {%
2 0.505894899368286
4 0.778219699859619
6 0.853444596131643
8 0.860606074333191
10 0.873307307561239
12 0.886671404043833
14 0.891134003798167
16 0.900343298912048
18 0.910179932912191
20 0.910049716631571
22 0.916702171166738
24 0.916583796342214
26 0.921318650245667
28 0.921910524368286
30 0.926598032315572
32 0.928101321061452
34 0.930172820885976
36 0.928823391596476
38 0.932741483052572
40 0.932812492052714
42 0.934895833333333
44 0.930705507596334
46 0.933368841807047
48 0.937227745850881
50 0.938458800315857
52 0.937618354956309
54 0.935913821061452
56 0.936470170815786
58 0.937014679114024
60 0.938577175140381
62 0.937144875526428
64 0.939417600631714
66 0.940553983052572
68 0.939938445885976
70 0.939926604429881
72 0.937630196412404
74 0.940826237201691
76 0.940258045991262
78 0.940376420815786
80 0.93901515007019
82 0.939358433087667
84 0.941039303938548
86 0.939760903517405
88 0.941335240999858
90 0.940897266070048
92 0.939689854780833
94 0.940411925315857
96 0.940601348876953
98 0.940944592158
100 0.940849920113881
102 0.939725379149119
104 0.940021296342214
106 0.94050661722819
108 0.940482954184214
110 0.940234382947286
112 0.939618845780691
114 0.94109849135081
116 0.940115988254547
118 0.940719703833262
120 0.939535975456238
122 0.939962108929952
124 0.941642999649048
126 0.940838058789571
128 0.93989109992981
130 0.939772725105286
132 0.94035275777181
134 0.937038362026215
136 0.940755228201548
138 0.939334750175476
140 0.940932750701904
142 0.940293550491333
144 0.941370725631714
146 0.940317233403524
148 0.939240058263143
150 0.940980116526286
152 0.941974441210429
154 0.939050654570262
156 0.939713537693024
158 0.940068662166595
160 0.940731525421143
162 0.941015621026357
164 0.938660025596619
166 0.941299716631571
168 0.940258045991262
170 0.940340916315714
172 0.9395951628685
174 0.943027953306834
176 0.941725850105286
178 0.940151512622833
180 0.940660516421
182 0.941583812236786
184 0.941796859105428
186 0.941737691561381
188 0.941832383473714
190 0.940767049789429
192 0.940364559491475
194 0.941193183263143
196 0.941276033719381
198 0.940778891245524
200 0.938932279745738
202 0.941832363605499
204 0.938671867052714
206 0.940944592158
208 0.941418091456095
210 0.940139691034953
212 0.941737691561381
214 0.940672338008881
216 0.940826237201691
218 0.941571970780691
220 0.940482934315999
222 0.941808700561523
224 0.940329054991404
226 0.941607475280762
228 0.941903392473857
230 0.938908636569977
232 0.941477259000142
234 0.939843753973643
236 0.940849900245667
238 0.941453595956167
240 0.940624992052714
242 0.940873603026072
244 0.941157658894857
246 0.941110332806905
248 0.941193163394928
250 0.941666662693024
252 0.940376420815786
254 0.940684159596761
256 0.94140625
258 0.942779342333476
260 0.940021296342214
262 0.940116008122762
264 0.942412416140238
266 0.94140625
268 0.940778871377309
270 0.942305862903595
272 0.941891551017761
274 0.941962599754333
276 0.941536466280619
278 0.941145837306976
280 0.940814395745595
282 0.9420099457105
284 0.943158149719238
};
\end{axis}

\end{tikzpicture}

%% file: labpal/figure_data/performance_comparison/SVHN/SVHN_ResNet-20_test_accuracy.pgf
% This file was created by tikzplotlib v0.9.8.
\begin{tikzpicture}

\definecolor{color0}{rgb}{0.647058823529412,0.164705882352941,0.164705882352941}
\definecolor{color1}{rgb}{1,0.647058823529412,0}
\definecolor{color2}{rgb}{0.933333333333333,0.509803921568627,0.933333333333333}

\begin{axis}[
xmajorticks=false,
grid = major,
major grid style={dotted},
legend cell align={left},
legend style={
  fill opacity=0.8,
  draw opacity=1,
  text opacity=1,
  at={(0.91,0.5)},
  anchor=east,
  draw=white!80!black
},
minor xtick={},
minor ytick={},
reverse legend, legend cell align={left}, legend style={ fill opacity=0.6, draw opacity=1, text opacity=1, at={(0.94,0.3)}, anchor=east, draw=white!80!black},,
tick align=outside,
tick pos=left,
title={test accuracy SVHN ResNet-20},
width=10.5cm,height=8cm,,
x grid style={white!69.0196078431373!black},
xlabel={epoch with best val. acc.},
xmin=-0.055, xmax=0.055,
y grid style={white!69.0196078431373!black},
ylabel={test accuracy},
ymin=0.955, ymax=0.982,
ytick style={color=black},
minor y tick num=3,
ytick={0.96,0.97,0.975,0.98},
y tick label style={
/pgf/number format/.cd,
fixed,
fixed zerofill,
precision=3,
/tikz/.cd
},
]
\addplot [draw=green!50.1960784313725!black, fill=green!50.1960784313725!black, mark=-, only marks, mark options={scale=3},line width=3pt]
table{%
x  y
-0.006 0.978247006734212
0 -1
};
\addlegendentry{SGD : $\lambda$: 0.1, $\beta$: 0.9}
\addplot [draw=black, fill=black, mark=-, only marks, mark options={scale=3},line width=3pt]
table{%
x  y
0 0.97828342517217
0 -1
};
\addlegendentry{SLS : c: 0.1, $\beta$: 0.9, $\mu$: 0.1}
\addplot [draw=color2, fill=color2, mark=-, only marks, mark options={scale=3},line width=3pt]
table{%
x  y
0 0.964839200178782
0 -1
};
\addlegendentry{GOLSI : $c$: 0.99, $\eta$: 2.0, $\alpha$: 0.1, $\beta$: 0.4}
\addplot [draw=color1, fill=color1, mark=-, only marks, mark options={scale=3},line width=3pt]
table{%
x  y
0 0.968094408512115
0 -1
};
\addlegendentry{PAL : $\alpha$: 1.0, $\mu$: 1.0, $\beta$: 0.4 }
\addplot [draw=blue, fill=blue, mark=-, only marks, mark options={scale=3},line width=3pt]
table{%
x  y
0 0.977805376052856
0 -1
};
\addlegendentry{LABPAL-SGD: $|\mathbb{B}_a|$: 1280, $\beta$: 0.0, $\alpha$: 1.9, n: 1000}
\addplot [draw=red, fill=red, mark=-, only marks, mark options={scale=3},line width=3pt]
table{%
x  y
0 0.976840237776438
0 -1
};
\addlegendentry{LABPAL-NSGD: $|\mathbb{B}_a|$: 1280, $\beta$: 0.0, $\alpha$: 1.9, n: 1000}

\end{axis}

\end{tikzpicture}

%% file: labpal/figure_data/performance_comparison/SVHN/SVHN_MobileNet-V2_test_accuracy.pgf
% This file was created by tikzplotlib v0.9.8.
\begin{tikzpicture}

\definecolor{color1}{rgb}{1,0.647058823529412,0}
\definecolor{color2}{rgb}{0.933333333333333,0.509803921568627,0.933333333333333}

\begin{axis}[
xmajorticks=false,
grid = major,
major grid style={dotted},
minor xtick={},
minor ytick={},
reverse legend, legend cell align={left}, legend style={ fill opacity=0.8, draw opacity=1, text opacity=1, at={(0.91,0.3)}, anchor=east, draw=white!80!black},,
tick align=outside,
tick pos=left,
title={test accuracy SVHN MobileNet-V2},
width=10.5cm,height=8cm,,
x grid style={white!69.0196078431373!black},
xlabel={epoch with best val. acc.},
xmin=-0.055, xmax=0.055,
y grid style={white!69.0196078431373!black},
ylabel={test accuracy},
ymin=0.95, ymax=0.985,
ytick style={color=black},
minor y tick num=3,
y tick label style={
/pgf/number format/.cd,
fixed,
fixed zerofill,
precision=3,
/tikz/.cd
},
]
%\addlegendentry{PLS : $\alpha$: 0.0001, $c-{W}$: 0.2, c1: 0.005, $\beta$: 0.0}
\addplot [draw=green!50.1960784313725!black, fill=green!50.1960784313725!black, mark=-, only marks, mark options={scale=3,opacity=0.8},line width=3pt]
table{%
x  y
-0.006 0.980741918087006
0 -1
};
%\addlegendentry{SGD : $\lambda$: 0.1, $\beta$: 0.0}
\addplot [draw=color2, fill=color2, mark=-, only marks, mark options={scale=3},line width=3pt]
table{%
x  y
0 0.974381744861603
0 -1
};
%\addlegendentry{GOLSI : $c$: 0.9, $\eta$: 2.0, $\alpha$: 0.0001, $\beta$: 0.0}
\addplot [draw=black, fill=black, mark=-, only marks, mark options={scale=3,opacity=0.8},line width=3pt]
table{%
x  y
0 0.978583931922913
0 -1
};
%\addlegendentry{SLS : c: 0.1, $\beta$: 0.9, $\mu-{init}$: 0.1}
\addplot [draw=blue, fill=blue, mark=-, only marks, mark options={scale=3,opacity=0.8},line width=3pt]
table{%
x  y
0 0.978688617547353
0 -1
};
%\addlegendentry{LABPAL-SGD : $\epsilon$: 1.0, max-step-size: 2.0}
\addplot [draw=red, fill=red, mark=-, only marks, mark options={scale=3},line width=3pt]
table{%
x  y
0 0.981392979621887
0 -1
};
%\addlegendentry{LABPAL-NSGD : $\epsilon$: 1.0}
\addplot [draw=color1, fill=color1, mark=-, only marks, mark options={scale=3},line width=3pt]
table{%
x  y
0 0.976366738478343
0 -1
};
%\addlegendentry{PAL : $\alpha$: 1.66, $\mu$: 1.0}
\end{axis}

\end{tikzpicture}

%% file: labpal/figure_data/performance_comparison/SVHN/SVHN_DenseNet-121_test_accuracy.pgf
% This file was created by tikzplotlib v0.9.8.
\begin{tikzpicture}

\definecolor{color1}{rgb}{1,0.647058823529412,0}
\definecolor{color0}{rgb}{0.933333333333333,0.509803921568627,0.933333333333333}

\begin{axis}[
grid = major,
xmajorticks=false,
major grid style={dotted},
minor xtick={},
minor ytick={},
reverse legend, legend cell align={left}, legend style={ fill opacity=0.8, draw opacity=1, text opacity=1, at={(0.91,0.3)}, anchor=east, draw=white!80!black},,
%tick align=outside,
%tick pos=left,
title={test accuracy SVHN DenseNet-121},
width=10.5cm,height=8cm,,
x grid style={white!69.0196078431373!black},
xlabel={epoch with best val. acc.},
xmin=-0.055, xmax=0.055,
y grid style={white!69.0196078431373!black},
ylabel={test accuracy},
ymin=0.95, ymax=0.985,
minor y tick num=3,
ytick style={color=black},
y tick label style={
/pgf/number format/.cd,
fixed,
fixed zerofill,
precision=3,
/tikz/.cd
},
restrict y to domain*=0.9:0.99,
]
\addplot [draw=green!50.1960784313725!black, fill=green!50.1960784313725!black, mark=-, only marks, mark options={scale=3},line width=3pt]
table{%
x  y
-0.006 0.980045358339945
0 -1
};
%\addlegendentry{SGD : $\lambda$: 0.1, $\beta$: 0.0}
\addplot [draw=black, fill=black, mark=-, only marks, mark options={scale=3},line width=3pt]
table{%
x  y
-0.006 0.979225854078929
0 -1
};
%\addlegendentry{SLS : c: 0.1, $\beta$: 0.9, $\mu-{init}$: 0.1}
\addplot [draw=color0, fill=color0, mark=-, only marks, mark options={scale=3},line width=3pt]
table{%
x  y
0 0.970830619335175
0 -1
};
%\addlegendentry{GOLSI : $c$: 0.9, $\eta$: 2.0, $\alpha$: 0.0001, $\beta$: 0.0}
\addplot [draw=blue, fill=blue, mark=-, only marks, mark options={scale=3},line width=3pt]
table{%
x  y
0 0.979758501052856
0 -1
};
%\addlegendentry{LABPAL-SGD : $\epsilon$: 1.0}
%\addplot [draw=red, fill=red, mark=-, only marks, mark options={scale=3},line width=3pt]
%table{%
%x  y
%0 0.979205369949341
%0 -1
%};
%\addlegendentry{LABPAL-NSGD : $\epsilon$: 1.0}
\addplot [draw=color1, fill=color1, mark=-, only marks, mark options={scale=3},line width=3pt]
table{%
x  y
0 0.975761214892069
0 -1
};
%\addlegendentry{PAL : $\alpha$: 1.66, $\mu$: 1.0}
\end{axis}

\end{tikzpicture}

%% file: labpal/batch_size_32.tex
\begin{figure}[h!]
	\tikzsetfigurename{labpal_opt_comparison_bs32}
	\centering
	\def\scale{0.4}
	\begin{tabular}{ c c c}
		\scalebox{\scale}{\input{"labpal/figure_data/performance_comparison_low_batch/plots_bs32/CIFAR-10_ResNet-20_learning_rate.pgf"}}&	
		\scalebox{\scale}{\input{"labpal/figure_data/performance_comparison_low_batch/plots_bs32/CIFAR-10_MobileNet-V2_learning_rate.pgf"}}&
		\scalebox{\scale}{\input{"labpal/figure_data/performance_comparison_low_batch/plots_bs32/CIFAR-10_DenseNet-121_learning_rate.pgf"}}\\
		\scalebox{\scale}{\input{"labpal/figure_data/performance_comparison_low_batch/plots_bs32/CIFAR-10_ResNet-20_training_loss.pgf"}}&
		\scalebox{\scale}{\input{"labpal/figure_data/performance_comparison_low_batch/plots_bs32/CIFAR-10_MobileNet-V2_training_loss.pgf"}}&
		\scalebox{\scale}{\input{"labpal/figure_data/performance_comparison_low_batch/plots_bs32/CIFAR-10_DenseNet-121_training_loss.pgf"}}\\	
		\scalebox{\scale}{\input{"labpal/figure_data/performance_comparison_low_batch/plots_bs32/CIFAR-10_ResNet-20_validation_accuracy.pgf"}}&	
		\scalebox{\scale}{\input{"labpal/figure_data/performance_comparison_low_batch/plots_bs32/CIFAR-10_MobileNet-V2_validation_accuracy.pgf"}}&	
		\scalebox{\scale}{\input{"labpal/figure_data/performance_comparison_low_batch/plots_bs32/CIFAR-10_DenseNet-121_validation_accuracy.pgf"}}\\
		\scalebox{\scale}{\input{"labpal/figure_data/performance_comparison_low_batch/plots_bs32/CIFAR-10_ResNet-20_test_accuracy.pgf"}}&
		\scalebox{\scale}{\input{"labpal/figure_data/performance_comparison_low_batch/plots_bs32/CIFAR-10_MobileNet-V2_test_accuracy.pgf"}}&
		\scalebox{\scale}{\input{"labpal/figure_data/performance_comparison_low_batch/plots_bs32/CIFAR-10_DenseNet-121_test_accuracy.pgf"}}

	\end{tabular}
	\caption{Performance comparison of several models on CIFAR-10 with \textbf{batch size 32}. \textbf{The same hyperparameters are used as for batch size 128} (see Figure \ref{labpal_fig_optimizer_comparison_cifar10}).  Due to the batch size adaptation to keep the noise scale on a similar level the LABPAL approaches perform almost identical compared to batch size 128. The performance off all other line searches decreases; SGD still performs well. Not that training steps were increased by a factor of 4.} 
	\label{labpal_fig_opt_comparison_bs32}
\end{figure}

%% file: labpal/figure_data/performance_comparison_low_batch/plots_bs32/CIFAR-10_DenseNet-121_learning_rate.pgf
% This file was created by tikzplotlib v0.9.8.
\begin{tikzpicture}

\definecolor{color0}{rgb}{0.933333333333333,0.509803921568627,0.933333333333333}
\definecolor{color1}{rgb}{0.647058823529412,0.164705882352941,0.164705882352941}
\definecolor{color2}{rgb}{1,0.647058823529412,0}

\begin{axis}[
,
log basis y={10},
minor xtick={},
minor ytick={},
tick align=outside,
tick pos=left,
title={learning rate CIFAR-10 DenseNet-121},
width=10.5cm,height=8cm,grid=major,major grid style={dotted},
x grid style={white!69.0196078431373!black},
xlabel={steps},
xmin=-19949.95, xmax=418970.95,
xtick style={color=black},
xtick={-50000,0,50000,100000,150000,200000,250000,300000,350000,400000,450000},
y grid style={white!69.0196078431373!black},
ylabel={learning rate},
ymin=10e-12, ymax=10e5,
ymode=log,
ytick style={color=black},
ytick={1e-14,1e-12,1e-10,1e-08,1e-06,0.0001,0.01,1,100,10000,1000000}
]
\addplot [line width=2.5pt, color0, opacity=1.0]
table {%
1 0.100000001490116
2777 1.38721084594727
5534 1.60026228427887
8313 0.673967063426971
11046 0.982671856880188
13759 0.97798103094101
16469 1.15661287307739
19178 1.15527653694153
21866 0.403303474187851
24549 0.862769901752472
27252 0.806399405002594
29915 1.05899012088776
32609 1.39318013191223
35243 0.638906121253967
37865 0.711150884628296
40507 1.04813981056213
43156 1.89725399017334
45788 1.59754800796509
48423 1.22970163822174
51025 1.1757345199585
53645 0.95950448513031
56297 1.78477644920349
58912 0.821375668048859
61565 0.859544098377228
64177 1.46984827518463
66816 1.11813986301422
69442 0.880754292011261
72067 0.943892776966095
74683 1.12780153751373
77289 1.14587390422821
79908 0.903947174549103
82498 2.04232907295227
85097 1.2383748292923
87698 1.42974078655243
90307 0.90019816160202
92906 1.09290146827698
95498 0.690349996089935
98094 1.54164326190948
100704 0.933300018310547
103300 1.35156869888306
105906 0.743186891078949
108500 0.874322533607483
111074 0.790430128574371
113668 0.537707448005676
116236 1.77401387691498
118820 1.36024129390717
121375 0.757489264011383
123978 1.21643459796906
126576 0.885730087757111
129146 1.72258472442627
131736 1.23126828670502
134310 1.03684592247009
136899 0.974213123321533
139467 1.11130940914154
142048 1.20939469337463
144623 1.04247891902924
147199 1.93937695026398
149777 2.66032218933105
152363 1.53151440620422
154921 0.941340386867523
157516 0.856026887893677
160133 1.44167864322662
162712 0.892515122890472
165341 0.7824667096138
167913 2.8176589012146
170456 2.65221285820007
173027 1.37965846061707
175625 1.09883522987366
178182 1.65511834621429
180748 1.24703335762024
183352 0.931830644607544
185920 0.931965470314026
188552 1.47914910316467
191145 1.73799216747284
193695 2.92869019508362
196284 3.25437331199646
198872 1.09675788879395
201470 2.56773519515991
204048 1.32815742492676
206631 7.56103134155273
209217 2.07933902740479
211786 0.475111156702042
214347 0.821169912815094
216942 2.5433452129364
219511 1.94269287586212
222123 1.37682926654816
224680 2.2419102191925
227299 2.05890226364136
229874 1.01553797721863
232471 3.50288200378418
235119 0.933994054794312
237732 0.941258728504181
240353 1.46326804161072
242992 2.37866044044495
245596 1.16441607475281
248185 1.69062840938568
250765 1.31647396087646
253375 1.90005481243134
255981 1.09865009784698
258584 1.59387755393982
261230 1.07727456092834
263877 2.88259053230286
266509 0.911364018917084
269163 0.0754771530628204
271771 1.74833345413208
274386 3.00834226608276
276998 2.10016512870789
279655 2.95104050636292
282321 0.733536601066589
284952 1.11500477790833
287594 9.99999993922529e-09
289676 9.99999993922529e-09
291766 9.99999993922529e-09
293849 9.99999993922529e-09
295936 9.99999993922529e-09
298029 9.99999993922529e-09
300129 9.99999993922529e-09
302228 9.99999993922529e-09
304316 9.99999993922529e-09
306396 9.99999993922529e-09
308496 9.99999993922529e-09
310586 9.99999993922529e-09
312666 9.99999993922529e-09
314773 9.99999993922529e-09
316858 9.99999993922529e-09
318938 9.99999993922529e-09
321024 9.99999993922529e-09
323118 9.99999993922529e-09
325197 9.99999993922529e-09
327283 9.99999993922529e-09
329385 9.99999993922529e-09
331479 9.99999993922529e-09
333583 9.99999993922529e-09
335667 9.99999993922529e-09
337762 9.99999993922529e-09
339860 9.99999993922529e-09
341949 9.99999993922529e-09
344040 9.99999993922529e-09
346110 9.99999993922529e-09
348202 9.99999993922529e-09
350284 9.99999993922529e-09
352362 9.99999993922529e-09
354433 9.99999993922529e-09
356511 9.99999993922529e-09
358600 9.99999993922529e-09
360701 9.99999993922529e-09
362796 9.99999993922529e-09
364884 9.99999993922529e-09
366959 9.99999993922529e-09
369057 9.99999993922529e-09
371145 9.99999993922529e-09
373235 9.99999993922529e-09
375333 9.99999993922529e-09
377406 9.99999993922529e-09
379503 9.99999993922529e-09
381598 9.99999993922529e-09
383673 9.99999993922529e-09
385760 9.99999993922529e-09
387860 9.99999993922529e-09
389948 9.99999993922529e-09
392044 9.99999993922529e-09
394134 9.99999993922529e-09
396234 9.99999993922529e-09
398327 9.99999993922529e-09
};
\addplot [line width=2.5pt, color1, opacity=1.0]
table {%
1 0
2750 0.235584909717242
4345 0.483907725972434
5882 2.279399116834
7036 3.16166504224141
8095 5.02711582183838
9229 5.39130369822184
10477 6.01416190465291
11895 7.06610790888468
13180 7.06610790888468
14408 7.06610790888468
15635 8.53778060277303
17226 9.58156855901082
19026 10.341699441274
20249 6.5814646879832
21249 7.74469335873922
22249 7.74469335873922
};
\addplot [line width=2.5pt, red, opacity=1.0]
table {%
20 0.0544980149716139
1020 0.0506910849362612
2020 0.0543158892542124
3020 0.060392377898097
4020 0.0566706843674183
5020 0.117775462567806
6020 0.12196858599782
7020 0.128394942730665
8020 0.141924113035202
9020 0.175884395837784
10020 0.206749141216278
11020 0.225401423871517
12020 0.227550469338894
13020 0.280481263995171
14020 0.341429129242897
15020 0.298750698566437
16020 0.318221688270569
17020 0.364757254719734
18020 0.450377255678177
19020 0.524550274014473
20020 0.525976195931435
21020 0.481768816709518
22020 0.638998627662659
23020 0.499094501137733
24020 0.534931004047394
25020 0.709545522928238
26020 0.600711017847061
27020 0.722618728876114
28020 0.725976169109344
29020 0.597132921218872
30020 0.720901221036911
31020 0.664112687110901
32020 0.753188461065292
33020 0.738119006156921
34020 0.861665338277817
35020 0.937308520078659
36020 0.588200986385345
37020 0.893023192882538
38020 0.928048312664032
39020 0.834814816713333
40020 0.808730095624924
41020 1.16887751221657
42020 0.999584496021271
43020 1.01347017288208
44020 1.68295294046402
45020 0.937935888767242
46020 1.07334733009338
47020 1.10943418741226
48020 1.16749531030655
49020 0.856108099222183
50020 0.969639807939529
51020 1.06498426198959
52020 1.12018746137619
53020 0.892938107252121
54020 1.19116121530533
55020 1.03246659040451
56020 1.4866731762886
57020 0.948221564292908
58020 1.11470663547516
59020 0.945331990718842
60020 1.16339391469955
61020 1.41830319166183
62020 0.837665140628815
63020 0.923907101154327
64020 1.57855892181396
65020 1.13307163119316
66020 1.24502152204514
67020 1.49590241909027
68020 1.16353815793991
69020 1.17714151740074
70020 1.32987651228905
71020 1.42214274406433
72020 1.41488260030746
73020 0.970477759838104
74020 1.25347471237183
75020 1.2658851146698
76020 1.40183877944946
77020 2.863325715065
78020 0.939962863922119
79020 1.13991656899452
80020 1.31267940998077
81020 1.37495082616806
82020 1.07703340053558
83020 2.56691586971283
84020 1.21149170398712
85020 1.25721019506454
86020 1.44348973035812
87020 1.33590525388718
88020 2.01336467266083
89020 1.25598028302193
90020 1.63934510946274
91020 1.68771064281464
92020 1.70045709609985
93020 1.31559348106384
94020 1.231818318367
95020 1.2151605784893
96020 1.05396384000778
97020 2.9871738255024
98020 1.52562892436981
99020 1.46457633376122
100020 0.95894181728363
101020 1.6305850148201
102020 2.08083897829056
103020 1.25190359354019
104020 1.054782807827
105020 2.49318027496338
106020 1.87549182772636
107020 1.48218256235123
108020 7.68150451779366
109020 1.20406419038773
110020 2.40544003248215
111020 2.68577206134796
112020 1.07903897762299
113020 1.2845256626606
114020 0.94949147105217
115020 1.08556494116783
116020 1.06917163729668
117020 2.68213596940041
118020 2.40727126598358
119020 1.97788065671921
120020 4.19820049405098
121020 22.5955818295479
122020 0.671730041503906
123020 1.64331549406052
124020 4.00281429290771
125020 10.4167091846466
126020 3.87630766630173
127020 2.50006866455078
128020 2.30857864022255
129020 1.22363582253456
130020 0.501404777169228
131020 0.45368330180645
132020 0.652102589607239
133020 0.835157364606857
134020 3.26677703857422
135020 1.53494438529015
136020 1.07089371979237
137020 3.3534597158432
138020 3.30405205488205
139020 1.58276188373566
140020 1.36136132478714
141020 8.81350481510162
142020 1.57139927148819
143020 25.2165019512177
144020 2.20276245474815
145020 26.8779344558716
146020 0.995668217539787
147020 5.93280649185181
148020 0.768967136740685
149020 122.374958232045
150020 5.11698812246323
151020 45.8248252868652
152020 0.237415201961994
153020 0.501953125
154020 5.08758282661438
155020 1.71095716953278
156020 5.27484047412872
157020 3.05319835245609
158020 31.2534618377686
159020 13.7601322233677
160020 146.383432388306
161020 8.45943367481232
162020 0.146177414804697
163020 22.8955995887518
164020 59.8301870524883
165020 85638.759765625
166020 92.6777081489563
167020 4.27313431352377
168020 84492.8715209961
169020 0.165330708026886
170020 183.288891792297
171020 2.41027367115021
172020 10.841026801616
173020 7.76973259449005
174020 92301.0421761125
175020 6691.83382415771
176020 40.920910179615
177020 2.0829558596015
178020 255.955903701484
179020 3516.77516956627
180020 89998.0538311005
181020 17.6787750720978
182020 229.824555397034
183020 392.555220961571
184020 49.5032272338867
185020 8391.73775315285
186020 14.5451887845993
187020 1036.49456787109
188020 80177.4765625
189020 703.573748245835
190020 63529.1030055732
191020 65581.6623535156
192020 614.914772033691
193020 1399.4920501709
194020 16242.6298828125
195020 29158.5070953369
196020 38.9231830835342
197020 265.319005966187
198020 1882.75942102075
199020 3785.20044648647
200020 27879.7640380859
201020 1621.02860867977
202020 2341.15128707886
203020 6261.13108238578
204020 48576.9015960693
205020 48603.8238372803
206020 33.5829188674688
207020 48408.7444680929
208020 9567.0613078028
209020 355.377181321383
210020 37166.8542685509
211020 28.3438355177641
212020 37165.7390368283
213020 37169.0111227036
214020 7464.05934429169
215020 0.237154763191938
216020 4.05981934070587
217020 36436.1463827044
218020 3192.00507031381
219020 36436.5680989027
220020 36436.1065635532
221020 24.9681911468506
222020 1073.67025184631
223020 11802.5197801888
224020 36455.4887638092
225020 122.221031188965
226020 523.392916604877
227020 4376.80901670456
228020 32963.541451931
229020 32984.5287132263
230020 32956.1384314001
231020 32957.8531552553
232020 32957.7682137489
233020 34085.8014936447
234020 34076.651446104
235020 23570.1565246582
236020 864.715620279312
237020 5870.40997666121
238020 15.2466688379645
239020 34076.5957214534
240020 131.507175624371
241020 209.024852409959
242020 131721.604901748
243020 34195.2673386335
244020 599.72900390625
245020 34194.6300016791
246020 34246.4213905334
247020 74.461600586772
248020 3324.59766483307
249020 16484.5104217529
250020 16297.9508428574
251020 0.157956690527499
252020 16444.70262146
253020 134.038597106934
254020 91.7711322158575
255020 1882.95368288457
256020 16298.093179971
257020 16244.0653043985
258020 0.249437997117639
259020 16682.7010461092
260020 27.9623991549015
261020 16682.8236039281
262020 16683.0254164636
263020 2933.39756831527
264020 9633.04264748096
265020 790.135734558105
266020 2354.56158447266
267020 15284.9148330688
268020 8318.09414672852
269020 17.1305894851685
270020 15216.5984296799
271020 15451.6536102295
272020 15354.5191955566
273020 1.68590413033962
274020 1016.64998580515
275020 1662.92541122437
276020 15559.7034301758
277020 14171.4333820343
278020 0.591159909963608
279020 15260.8715633601
280020 61.2904844284058
281020 15260.9549271911
282020 1342.52739620209
283020 15692.9615859985
284020 15640.2869358063
285020 1543.23529268801
286020 15669.847328186
287020 15634.4199801534
288020 54694.4929199219
289020 458.104309082031
290020 18.5846214294434
291020 197.128097057343
292020 196.316458702087
293020 65947.501953125
294020 12082.6164250374
295020 252.521911621094
296020 207.499101042747
297020 17194.7075195312
298020 11935.8501815796
299020 16180.2357940674
300020 13287.1555175781
301020 16161.7452793121
302020 15678.2680969238
303020 16142.1730666161
304020 95.1618435829878
305020 16469.877532959
306020 386.586845606565
307020 16150.4439592361
308020 12699.4364700317
309020 1420.87037789822
310020 12058.3662675768
311020 16140.0150086582
312020 1051.17601329088
313020 16158.9737377167
314020 2564.67634417117
315020 499.655913114548
316020 4246.38896942139
317020 16039.9250928313
318020 16039.8933590651
319020 499.962692260742
320020 4.50503444671631
321020 41.2024140357971
322020 15147.5302848816
323020 23062.2880859375
324020 21652.228515625
325020 134.63497005403
326020 484.406982421875
327020 2949.94036364555
328020 16050.3511314392
329020 107745.069602966
330020 761.070602416992
331020 4530.02368164062
332020 23532.515625
333020 103069.319000244
334020 16320.3801879883
335020 19001.0004882812
336020 15639.1117610931
337020 8826.82690429688
338020 54882.609375
339020 9277.93635329604
340020 5741.54125976562
341020 37637.0703125
342020 15631.1140717715
343020 15646.4455718994
344020 465.554825782776
345020 4575.98460388184
346020 81481.3920449317
347020 23.5597231835127
348020 893.373291492462
349020 22093.3109690547
350020 16667.2784423828
351020 88637.9234619141
352020 102948.579101562
353020 15661.5957641602
354020 10186.7916717529
355020 15428.0619183481
356020 63729.8869628906
357020 492.995185852051
358020 90633.3916015625
359020 73862.0556640625
360020 15430.3820867538
361020 15428.0708711147
362020 102616.948242188
363020 24091.8505859375
364020 240.936249747872
365020 15096.2548571602
366020 16694.9354248047
367020 5662.31480407715
368020 74451.6171875
369020 7708.02444458008
370020 74557.1083984375
371020 37173.2138671875
372020 74557.1083984375
373020 15125.1816577911
374020 2311.50085449219
375020 21992.07421875
376020 74557.1083984375
377020 74557.1083984375
378020 62541.8312988281
379020 60474.0427856445
380020 150.082107543945
381020 59251.8058063425
382020 9986.19624638557
383020 74347.8720703125
384020 15367.802734375
385020 60062.0706787109
386020 59285.1295318604
387020 33054.0908203125
388020 74347.8720703125
389020 3016.89144897461
390020 341.990898132324
391020 74347.8720703125
392020 15483.0335388184
393020 74347.8720703125
394020 88.9294852614403
395020 16508.2813720703
396020 85.6393814086914
397020 15616.3755745888
398020 53591.1015625
399020 62261.466796875
};
\addplot [line width=2.5pt, color2, opacity=1.0]
table {%
2 0.0623696732024352
2002 2.8521430293719
4002 2.80771772066752
6002 2.10934221744537
8002 5.53824353218079
10002 2.16535111268361
12002 3.72148243586222
14002 3.72570443153381
16002 5.13554501533508
18002 3.75440621376038
20002 3.58941793441772
22002 2.83069588740667
24002 3.40243943532308
26002 2.57252317667007
28002 4.21447436014811
30002 1.96283562978109
32002 3.03545316060384
34002 3.13685043652852
36002 3.38537653287252
38002 3.06387921174367
40002 2.38904468218485
42002 3.52945033709208
44002 1.82444129387538
46002 2.3024609486262
48002 2.90450100104014
50002 1.99535282452901
52002 2.05859417716662
54002 1.21006976316373
56002 1.5117921034495
58002 2.55598851044973
60002 2.11390721797943
62002 1.7471094528834
64002 2.75682727495829
66002 1.32595018545787
68002 1.7887673775355
70002 1.55318876107534
72002 1.80467879772186
74002 1.87822461128235
76002 2.2381720940272
78002 1.60379175345103
80002 2.01481266816457
82002 1.07392591238022
84002 1.51752916971842
86002 1.26203634341558
88002 1.54924952983856
90002 1.277943054835
92002 1.51651702324549
94002 1.24283691247304
96002 1.92141052087148
98002 0.988869428634644
100002 1.01921153068542
102002 1.01621601978938
104002 1.2845952908198
106002 0.983288695414861
108002 1.09987314542135
110002 1.77665090560913
112002 1.28267272313436
114002 1.37258938948313
116002 1.82566010951996
118002 1.17869818210602
120002 1.09554813305537
122002 1.81438636779785
124002 1.03169500827789
126002 1.05195885896683
128002 1.99723293383916
130002 1.40263464053472
132002 1.14564088980357
134002 1.24003211657206
136002 1.16686060031255
138002 3.05071576436361
140002 1.4908371369044
142002 1.14773291349411
144002 1.1458881298701
146002 3.1065582036972
148002 1.29933772484461
150002 1.10556777318319
152002 1.75915977358818
154002 1.05029918750127
156002 5.21462168296178
158002 0.477855424086253
160002 1.34721054633458
162002 2.27647272745768
164002 1.19807318846385
166002 0.932421088218689
168002 0.541910285751025
170002 2.04665903250376
172002 2.50551984707514
174002 1.8195081949234
176002 2.2287202278773
178002 1.41230384508769
180002 3.41319364309311
182002 5.18060700098674
184002 3.17123254140218
186002 1.93754661083221
188002 1.59503936767578
190002 2.1969145933787
192002 0.784367342789968
194002 0.789279898007711
196002 0.835910618305206
198002 3.79894111553828
200002 4.58587241172791
202002 0.956696768601735
204002 1.6825248003006
206002 1.21305894851685
208002 2.27978285153707
210002 1.83060216903687
212002 5.41654102007548
214002 6.20732927322388
216002 4.24254369735718
218002 2.4130943218867
220002 3.16114844878515
222002 0.915796538194021
224002 0.898742437362671
226002 1.05133581161499
228002 2.60008760293325
230002 1.11645632982254
232002 1.50339901447296
234002 2.24126555522283
236002 0.883228758970896
238002 3.08848468462626
240002 1.94926549990972
242002 0.93618913491567
244002 5.5878381729126
246002 2.31108661492666
248002 1.01800239086151
250002 1.36005008220673
252002 2.14081064860026
254002 3.97190161546071
256002 2.68489807844162
258002 2.61391909917196
260002 3.8522390127182
262002 2.73476309577624
264002 1.32810996969541
266002 0.871612350145976
268002 2.21251275142034
270002 1.27858068545659
272002 4.44485851128896
274002 5.59790617227554
276002 4.10061691204707
278002 2.973978916804
280002 1.3393040895462
282002 4.1113575498263
284002 1.85389616092046
286002 2.25054359436035
288002 3.31631028652191
290002 1.82894893487295
292002 0.641110897064209
294002 2.84200127919515
296002 0.984409213066101
298002 0.875784079233805
300002 4.13979522387187
302002 1.25967820485433
304002 0.921518166859945
306002 3.27926882108053
308002 1.64634170134862
310002 2.28677868843079
312002 1.20418494939804
314002 3.13542755444845
316002 0.735578298568726
318002 3.79727919896444
320002 2.33287264903386
322002 0.589277565479279
324002 0.302757680416107
326002 1.9994576772054
328002 1.40248507261276
330002 2.13966675599416
332002 2.05940318107605
334002 5.73937400182088
336002 4.5270624756813
338002 1.29657697677612
340002 1.84940695762634
342002 9.67708750565847
344002 3.22831877072652
346002 0.697071929772695
348002 1.66028889020284
350002 0.727725982666016
352002 0.641598244508108
354002 1.10220354795456
356002 0.798902690410614
358002 4.33519568045934
360002 1.79733653863271
362002 2.04673361778259
364002 7.68455090125402
366002 2.94216016928355
368002 5.90727748473485
370002 1.50649237632751
372002 12.3695233662923
374002 1.00670659542084
376002 3.75359582901001
378002 1.29866156975428
380002 3.41389479239782
382002 1.17355269193649
384002 1.59191280603409
386002 2.70432813962301
388002 5.52570756276449
390002 1.45218481620153
392002 0.788146555423737
394002 26.1854600509008
396002 1.43707389136155
398002 2.1672588189443
};
\addplot [line width=2.5pt, green!50.1960784313725!black, opacity=1.0]
table {%
1 0.100000001490116
1001 0.100000001490116
2001 0.100000001490116
3001 0.100000001490116
4001 0.100000001490116
5001 0.100000001490116
6001 0.100000001490116
7001 0.100000001490116
8001 0.100000001490116
9001 0.100000001490116
10001 0.100000001490116
11001 0.100000001490116
12001 0.100000001490116
13001 0.100000001490116
14001 0.100000001490116
15001 0.100000001490116
16001 0.100000001490116
17001 0.100000001490116
18001 0.100000001490116
19001 0.100000001490116
20001 0.100000001490116
21001 0.100000001490116
22001 0.100000001490116
23001 0.100000001490116
24001 0.100000001490116
25001 0.100000001490116
26001 0.100000001490116
27001 0.100000001490116
28001 0.100000001490116
29001 0.100000001490116
30001 0.100000001490116
31001 0.100000001490116
32001 0.100000001490116
33001 0.100000001490116
34001 0.100000001490116
35001 0.100000001490116
36001 0.100000001490116
37001 0.100000001490116
38001 0.100000001490116
39001 0.100000001490116
40001 0.100000001490116
41001 0.100000001490116
42001 0.100000001490116
43001 0.100000001490116
44001 0.100000001490116
45001 0.100000001490116
46001 0.100000001490116
47001 0.100000001490116
48001 0.100000001490116
49001 0.100000001490116
50001 0.100000001490116
51001 0.100000001490116
52001 0.100000001490116
53001 0.100000001490116
54001 0.100000001490116
55001 0.100000001490116
56001 0.100000001490116
57001 0.100000001490116
58001 0.100000001490116
59001 0.100000001490116
60001 0.100000001490116
61001 0.100000001490116
62001 0.100000001490116
63001 0.100000001490116
64001 0.100000001490116
65001 0.100000001490116
66001 0.100000001490116
67001 0.100000001490116
68001 0.100000001490116
69001 0.100000001490116
70001 0.100000001490116
71001 0.100000001490116
72001 0.100000001490116
73001 0.100000001490116
74001 0.100000001490116
75001 0.100000001490116
76001 0.100000001490116
77001 0.100000001490116
78001 0.100000001490116
79001 0.100000001490116
80001 0.100000001490116
81001 0.100000001490116
82001 0.100000001490116
83001 0.100000001490116
84001 0.100000001490116
85001 0.100000001490116
86001 0.100000001490116
87001 0.100000001490116
88001 0.100000001490116
89001 0.100000001490116
90001 0.100000001490116
91001 0.100000001490116
92001 0.100000001490116
93001 0.100000001490116
94001 0.100000001490116
95001 0.100000001490116
96001 0.100000001490116
97001 0.100000001490116
98001 0.100000001490116
99001 0.100000001490116
100001 0.100000001490116
101001 0.100000001490116
102001 0.100000001490116
103001 0.100000001490116
104001 0.100000001490116
105001 0.100000001490116
106001 0.100000001490116
107001 0.100000001490116
108001 0.100000001490116
109001 0.100000001490116
110001 0.100000001490116
111001 0.100000001490116
112001 0.100000001490116
113001 0.100000001490116
114001 0.100000001490116
115001 0.100000001490116
116001 0.100000001490116
117001 0.100000001490116
118001 0.100000001490116
119001 0.100000001490116
120001 0.100000001490116
121001 0.100000001490116
122001 0.100000001490116
123001 0.100000001490116
124001 0.100000001490116
125001 0.100000001490116
126001 0.100000001490116
127001 0.100000001490116
128001 0.100000001490116
129001 0.100000001490116
130001 0.100000001490116
131001 0.100000001490116
132001 0.100000001490116
133001 0.100000001490116
134001 0.100000001490116
135001 0.100000001490116
136001 0.100000001490116
137001 0.100000001490116
138001 0.100000001490116
139001 0.100000001490116
140001 0.100000001490116
141001 0.100000001490116
142001 0.100000001490116
143001 0.100000001490116
144001 0.100000001490116
145001 0.100000001490116
146001 0.100000001490116
147001 0.100000001490116
148001 0.100000001490116
149001 0.100000001490116
150001 0.100000001490116
151001 0.100000001490116
152001 0.100000001490116
153001 0.100000001490116
154001 0.100000001490116
155001 0.100000001490116
156001 0.100000001490116
157001 0.100000001490116
158001 0.100000001490116
159001 0.100000001490116
160001 0.100000001490116
161001 0.100000001490116
162001 0.100000001490116
163001 0.100000001490116
164001 0.100000001490116
165001 0.100000001490116
166001 0.100000001490116
167001 0.100000001490116
168001 0.100000001490116
169001 0.100000001490116
170001 0.100000001490116
171001 0.100000001490116
172001 0.100000001490116
173001 0.100000001490116
174001 0.100000001490116
175001 0.100000001490116
176001 0.100000001490116
177001 0.100000001490116
178001 0.100000001490116
179001 0.100000001490116
180001 0.100000001490116
181001 0.100000001490116
182001 0.100000001490116
183001 0.100000001490116
184001 0.100000001490116
185001 0.100000001490116
186001 0.100000001490116
187001 0.100000001490116
188001 0.100000001490116
189001 0.100000001490116
190001 0.100000001490116
191001 0.100000001490116
192001 0.100000001490116
193001 0.100000001490116
194001 0.100000001490116
195001 0.100000001490116
196001 0.100000001490116
197001 0.100000001490116
198001 0.100000001490116
199001 0.100000001490116
200001 0.00999999977648258
201001 0.00999999977648258
202001 0.00999999977648258
203001 0.00999999977648258
204001 0.00999999977648258
205001 0.00999999977648258
206001 0.00999999977648258
207001 0.00999999977648258
208001 0.00999999977648258
209001 0.00999999977648258
210001 0.00999999977648258
211001 0.00999999977648258
212001 0.00999999977648258
213001 0.00999999977648258
214001 0.00999999977648258
215001 0.00999999977648258
216001 0.00999999977648258
217001 0.00999999977648258
218001 0.00999999977648258
219001 0.00999999977648258
220001 0.00999999977648258
221001 0.00999999977648258
222001 0.00999999977648258
223001 0.00999999977648258
224001 0.00999999977648258
225001 0.00999999977648258
226001 0.00999999977648258
227001 0.00999999977648258
228001 0.00999999977648258
229001 0.00999999977648258
230001 0.00999999977648258
231001 0.00999999977648258
232001 0.00999999977648258
233001 0.00999999977648258
234001 0.00999999977648258
235001 0.00999999977648258
236001 0.00999999977648258
237001 0.00999999977648258
238001 0.00999999977648258
239001 0.00999999977648258
240001 0.00999999977648258
241001 0.00999999977648258
242001 0.00999999977648258
243001 0.00999999977648258
244001 0.00999999977648258
245001 0.00999999977648258
246001 0.00999999977648258
247001 0.00999999977648258
248001 0.00999999977648258
249001 0.00999999977648258
250001 0.00999999977648258
251001 0.00999999977648258
252001 0.00999999977648258
253001 0.00999999977648258
254001 0.00999999977648258
255001 0.00999999977648258
256001 0.00999999977648258
257001 0.00999999977648258
258001 0.00999999977648258
259001 0.00999999977648258
260001 0.00999999977648258
261001 0.00999999977648258
262001 0.00999999977648258
263001 0.00999999977648258
264001 0.00999999977648258
265001 0.00999999977648258
266001 0.00999999977648258
267001 0.00999999977648258
268001 0.00999999977648258
269001 0.00999999977648258
270001 0.00999999977648258
271001 0.00999999977648258
272001 0.00999999977648258
273001 0.00999999977648258
274001 0.00999999977648258
275001 0.00999999977648258
276001 0.00999999977648258
277001 0.00999999977648258
278001 0.00999999977648258
279001 0.00999999977648258
280001 0.00999999977648258
281001 0.00999999977648258
282001 0.00999999977648258
283001 0.00999999977648258
284001 0.00999999977648258
285001 0.00999999977648258
286001 0.00999999977648258
287001 0.00999999977648258
288001 0.00999999977648258
289001 0.00999999977648258
290001 0.00999999977648258
291001 0.00999999977648258
292001 0.00999999977648258
293001 0.00999999977648258
294001 0.00999999977648258
295001 0.00999999977648258
296001 0.00999999977648258
297001 0.00999999977648258
298001 0.00999999977648258
299001 0.00999999977648258
300001 0.00100000004749745
301001 0.00100000004749745
302001 0.00100000004749745
303001 0.00100000004749745
304001 0.00100000004749745
305001 0.00100000004749745
306001 0.00100000004749745
307001 0.00100000004749745
308001 0.00100000004749745
309001 0.00100000004749745
310001 0.00100000004749745
311001 0.00100000004749745
312001 0.00100000004749745
313001 0.00100000004749745
314001 0.00100000004749745
315001 0.00100000004749745
316001 0.00100000004749745
317001 0.00100000004749745
318001 0.00100000004749745
319001 0.00100000004749745
320001 0.00100000004749745
321001 0.00100000004749745
322001 0.00100000004749745
323001 0.00100000004749745
324001 0.00100000004749745
325001 0.00100000004749745
326001 0.00100000004749745
327001 0.00100000004749745
328001 0.00100000004749745
329001 0.00100000004749745
330001 0.00100000004749745
331001 0.00100000004749745
332001 0.00100000004749745
333001 0.00100000004749745
334001 0.00100000004749745
335001 0.00100000004749745
336001 0.00100000004749745
337001 0.00100000004749745
338001 0.00100000004749745
339001 0.00100000004749745
340001 0.00100000004749745
341001 0.00100000004749745
342001 0.00100000004749745
343001 0.00100000004749745
344001 0.00100000004749745
345001 0.00100000004749745
346001 0.00100000004749745
347001 0.00100000004749745
348001 0.00100000004749745
349001 0.00100000004749745
350001 0.00100000004749745
351001 0.00100000004749745
352001 0.00100000004749745
353001 0.00100000004749745
354001 0.00100000004749745
355001 0.00100000004749745
356001 0.00100000004749745
357001 0.00100000004749745
358001 0.00100000004749745
359001 0.00100000004749745
360001 0.00100000004749745
361001 0.00100000004749745
362001 0.00100000004749745
363001 0.00100000004749745
364001 0.00100000004749745
365001 0.00100000004749745
366001 0.00100000004749745
367001 0.00100000004749745
368001 0.00100000004749745
369001 0.00100000004749745
370001 0.00100000004749745
371001 0.00100000004749745
372001 0.00100000004749745
373001 0.00100000004749745
374001 0.00100000004749745
375001 0.00100000004749745
376001 0.00100000004749745
377001 0.00100000004749745
378001 0.00100000004749745
379001 0.00100000004749745
380001 0.00100000004749745
381001 0.00100000004749745
382001 0.00100000004749745
383001 0.00100000004749745
384001 0.00100000004749745
385001 0.00100000004749745
386001 0.00100000004749745
387001 0.00100000004749745
388001 0.00100000004749745
389001 0.00100000004749745
390001 0.00100000004749745
391001 0.00100000004749745
392001 0.00100000004749745
393001 0.00100000004749745
394001 0.00100000004749745
395001 0.00100000004749745
396001 0.00100000004749745
397001 0.00100000004749745
398001 0.00100000004749745
399001 0.00100000004749745
};
\addplot [line width=2.5pt, black, opacity=1.0]
table {%
1 0.100049309432507
2003 0.110145437220732
4003 0.180330981810888
6006 0.167528743545214
8007 0.250847501059373
10012 0.254680325587591
12015 0.273427426815033
14021 0.256926750143369
16021 0.356071357925733
18038 0.198253805438677
20038 0.284858867526054
22038 0.336605399847031
24050 0.21327926715215
26050 0.265361602107684
28050 0.394464448094368
30055 0.355927427609762
32064 0.261550868550936
34091 0.197479120145241
36091 0.158324678738912
38091 0.160196110606194
40091 0.176820253332456
42091 0.201589127381643
44091 0.317005544900894
46127 0.11392661742866
48127 0.139191267391046
50127 0.109238486737013
52127 0.158596908052762
54127 0.130658303697904
56127 0.158385172486305
58129 0.182393381992976
60133 0.181026871005694
62140 0.148198634386063
64159 0.117754459381104
66161 0.101097490638494
68161 0.0860518415768941
70161 0.0996937900781631
72172 0.0824359009663264
74186 0.0737614097694556
76198 0.0645380175362031
78198 0.05097881394128
80198 0.0646308821936448
82198 0.0674620792269707
84198 0.0733252912759781
86209 0.0549356391032537
88209 0.0634409586588542
90214 0.0627023453513781
92216 0.0698856338858604
94222 0.0668484941124916
96228 0.0621479203303655
98233 0.0639765647550424
100237 0.0641244339446227
102241 0.0616404811541239
104246 0.0627236093084017
106260 0.0352582323054473
108260 0.0463958693047365
110263 0.0389114040881395
112271 0.0378595888614655
114271 0.0473836548626423
116282 0.0421886617938677
118282 0.0448288495341937
120289 0.042085754374663
122297 0.0369032050172488
124297 0.0441900255779425
126300 0.0466440891226133
128305 0.0438725203275681
130308 0.0459223141272863
132316 0.0414804257452488
134320 0.0422283050914605
136326 0.042586375027895
138328 0.0449770124008258
140336 0.0374638866633177
142340 0.0382084331164757
144344 0.0383447663237651
146350 0.0370828000207742
148354 0.0370318864782651
150360 0.0333752694229285
152365 0.0362283481905858
154367 0.0398564375936985
156375 0.0327004262556632
158379 0.0301728459695975
160386 0.0325002893805504
162387 0.0358621372530858
164392 0.0327461299796899
166398 0.0353002833823363
168404 0.0307140511771043
170410 0.0306078058977922
172412 0.0321514718234539
174417 0.0312831544627746
176422 0.033122081309557
178426 0.0333742555230856
180431 0.030633774275581
182437 0.0290973894298077
184442 0.0292305499315262
186445 0.0307543929666281
188451 0.029016533245643
190456 0.0280518761525551
192459 0.0308754853904247
194466 0.0253813005983829
196470 0.0294611901044846
198481 0.0221205862859885
200481 0.0261239688843489
202483 0.0314146845291058
204491 0.0285294217367967
206507 0.0110754923714395
208507 0.00433113520952411
210507 0.007090969026254
212507 0.0116093903916359
214522 0.00391336998175485
216522 0.0064070001346392
218522 0.0104895917244604
220526 0.011267626060975
222532 0.00980372976892368
224536 0.0105308927795597
226541 0.0101807924805444
228548 0.00797228828378061
230549 0.0117470647302401
232556 0.00919879105906478
234560 0.00988108480174634
236566 0.00859732865481706
238568 0.0114012472353714
240576 0.00803519226973957
242582 0.00699125541146012
244585 0.00834423552599917
246589 0.00896314532617609
248595 0.00779864814614499
250599 0.00837709012469606
252601 0.0111091807736815
254606 0.0107398554945087
256613 0.00841007444813638
258615 0.0111529231243025
260621 0.00970392862360881
262626 0.00938132034671927
264630 0.010077153045732
266636 0.00876792337890726
268647 0.00450472411472113
270647 0.00737516956543005
272648 0.0108672181945381
274654 0.00945534249399983
276660 0.00822689883160382
278664 0.00883710510223405
280668 0.00949257241362151
282672 0.0101966566195998
284679 0.00798471087368497
286682 0.00952994825139033
288688 0.00829181201674068
290876 5.71288814865678e-11
292890 2.84565744810653e-11
294890 4.65893290381899e-11
296906 3.82824535960502e-11
298906 4.72546296074553e-11
300906 4.60018793890793e-11
302913 3.85539985353622e-11
304923 3.18740664649135e-11
306923 3.3667413475158e-11
308940 1.82937666751148e-11
310940 2.99506993280592e-11
312940 3.55785164061576e-11
314940 4.5311461180364e-11
316962 4.43932279536411e-11
318964 1.48772990454793e-11
320964 2.43572315095086e-11
322964 2.73843719691043e-11
324964 4.06442587919109e-11
326969 3.88835323892704e-11
328973 3.83783074764388e-11
330978 4.50104409762305e-11
332978 4.21873097798212e-11
334985 3.51351413241385e-11
336994 2.59833685239038e-11
339006 2.91951989204546e-11
341011 3.18756236683537e-11
343011 2.9273632996818e-11
345011 4.50813200427354e-11
347019 2.55643226495147e-11
349020 3.50292671027116e-11
351020 4.31745079502536e-11
353021 4.70163462560717e-11
355033 3.3455928999393e-11
357039 3.47271014028427e-11
359044 4.76580037008419e-11
361050 4.04277173080928e-11
363052 5.65202421901982e-11
365053 3.9689484337968e-11
367053 4.86134882349276e-11
369058 5.05893417435122e-11
371074 3.52225326451701e-11
373074 3.62115719354517e-11
375074 5.35523708711239e-11
377095 2.98924108845429e-11
379095 3.73415939373493e-11
381095 3.38701462709867e-11
383096 3.919457177796e-11
385100 5.10003243340644e-11
387106 5.04541651513701e-11
389123 2.71511800311227e-11
391123 2.61193549427885e-11
393123 3.28869674548482e-11
395123 4.45776193735981e-11
397128 5.08713430176962e-11
399141 3.06426591277035e-11
};
\addplot [line width=2.5pt, blue, opacity=1.0]
table {%
20 0.0638142228126526
1020 0.052600059658289
2020 0.052600059658289
3020 0.052600059658289
4020 0.052600059658289
5020 0.103502348065376
6020 0.103502348065376
7020 0.103502348065376
8020 0.103502348065376
9020 0.150411993265152
10020 0.150411993265152
11020 0.150411993265152
12020 0.150411993265152
13020 0.198586046695709
14020 0.198586046695709
15020 0.198586046695709
16020 0.198586046695709
17020 0.220505401492119
18020 0.220505401492119
19020 0.220505401492119
20020 0.220505401492119
21020 0.251788139343262
22020 0.251788139343262
23020 0.251788139343262
24020 0.251788139343262
25020 0.285890191793442
26020 0.285890191793442
27020 0.285890191793442
28020 0.285890191793442
29020 0.315260767936707
30020 0.315260767936707
31020 0.315260767936707
32020 0.315260767936707
33020 0.359337031841278
34020 0.359337031841278
35020 0.359337031841278
36020 0.359337031841278
37020 0.364026635885239
38020 0.364026635885239
39020 0.364026635885239
40020 0.364026635885239
41020 0.378974288702011
42020 0.378974288702011
43020 0.378974288702011
44020 0.378974288702011
45020 0.386409074068069
46020 0.386409074068069
47020 0.386409074068069
48020 0.386409074068069
49020 0.378196477890015
50020 0.378196477890015
51020 0.378196477890015
52020 0.378196477890015
53020 0.378116577863693
54020 0.378116577863693
55020 0.378116577863693
56020 0.378116577863693
57020 0.38107642531395
58020 0.38107642531395
59020 0.38107642531395
60020 0.38107642531395
61020 0.413988530635834
62020 0.413988530635834
63020 0.413988530635834
64020 0.413988530635834
65020 0.426286607980728
66020 0.426286607980728
67020 0.426286607980728
68020 0.426286607980728
69020 0.440522402524948
70020 0.440522402524948
71020 0.440522402524948
72020 0.440522402524948
73020 0.439496099948883
74020 0.439496099948883
75020 0.439496099948883
76020 0.439496099948883
77020 0.440635532140732
78020 0.440635532140732
79020 0.440635532140732
80020 0.440635532140732
81020 0.445022970438004
82020 0.445022970438004
83020 0.445022970438004
84020 0.445022970438004
85020 0.433503717184067
86020 0.433503717184067
87020 0.433503717184067
88020 0.433503717184067
89020 0.413608700037003
90020 0.413608700037003
91020 0.413608700037003
92020 0.413608700037003
93020 0.387716591358185
94020 0.387716591358185
95020 0.387716591358185
96020 0.387716591358185
97020 0.373207420110703
98020 0.373207420110703
99020 0.373207420110703
100020 0.373207420110703
101020 0.358298808336258
102020 0.358298808336258
103020 0.358298808336258
104020 0.358298808336258
105020 0.335995823144913
106020 0.335995823144913
107020 0.335995823144913
108020 0.335995823144913
109020 0.340781182050705
110020 0.340781182050705
111020 0.340781182050705
112020 0.340781182050705
113020 0.329207897186279
114020 0.329207897186279
115020 0.329207897186279
116020 0.329207897186279
117020 0.309711575508118
118020 0.309711575508118
119020 0.309711575508118
120020 0.309711575508118
121020 0.284049391746521
122020 0.284049391746521
123020 0.284049391746521
124020 0.284049391746521
125020 0.286866247653961
126020 0.286866247653961
127020 0.286866247653961
128020 0.286866247653961
129020 0.269594728946686
130020 0.269594728946686
131020 0.269594728946686
132020 0.269594728946686
133020 0.258042484521866
134020 0.258042484521866
135020 0.258042484521866
136020 0.258042484521866
137020 0.311275273561478
138020 0.311275273561478
139020 0.311275273561478
140020 0.311275273561478
141020 0.303622901439667
142020 0.303622901439667
143020 0.303622901439667
144020 0.303622901439667
145020 0.298938423395157
146020 0.298938423395157
147020 0.298938423395157
148020 0.298938423395157
149020 0.307735949754715
150020 0.307735949754715
151020 0.307735949754715
152020 0.307735949754715
153020 0.294962525367737
154020 0.294962525367737
155020 0.294962525367737
156020 0.294962525367737
157020 0.329672127962112
158020 0.329672127962112
159020 0.329672127962112
160020 0.329672127962112
161020 0.319820284843445
162020 0.319820284843445
163020 0.319820284843445
164020 0.319820284843445
165020 0.32157438993454
166020 0.32157438993454
167020 0.32157438993454
168020 0.32157438993454
169020 0.330822736024857
170020 0.330822736024857
171020 0.330822736024857
172020 0.330822736024857
173020 0.426197111606598
174020 0.426197111606598
175020 0.426197111606598
176020 0.426197111606598
177020 0.425680637359619
178020 0.425680637359619
179020 0.425680637359619
180020 0.425680637359619
181020 0.456744968891144
182020 0.456744968891144
183020 0.456744968891144
184020 0.456744968891144
185020 0.45198717713356
186020 0.45198717713356
187020 0.45198717713356
188020 0.45198717713356
189020 0.457675993442535
190020 0.457675993442535
191020 0.457675993442535
192020 0.457675993442535
193020 0.467437714338303
194020 0.467437714338303
195020 0.467437714338303
196020 0.467437714338303
197020 0.412060350179672
198020 0.412060350179672
199020 0.412060350179672
200020 0.412060350179672
201020 0.410495311021805
202020 0.410495311021805
203020 0.410495311021805
204020 0.410495311021805
205020 0.410495311021805
206020 0.410495311021805
207020 0.410495311021805
208020 0.410495311021805
209020 0.408468753099442
210020 0.408468753099442
211020 0.408468753099442
212020 0.408468753099442
213020 0.408468753099442
214020 0.408468753099442
215020 0.408468753099442
216020 0.408468753099442
217020 0.405989646911621
218020 0.405989646911621
219020 0.405989646911621
220020 0.405989646911621
221020 0.405989646911621
222020 0.405989646911621
223020 0.405989646911621
224020 0.405989646911621
225020 0.452391743659973
226020 0.452391743659973
227020 0.452391743659973
228020 0.452391743659973
229020 0.452391743659973
230020 0.452391743659973
231020 0.452391743659973
232020 0.452391743659973
233020 0.451100260019302
234020 0.451100260019302
235020 0.451100260019302
236020 0.451100260019302
237020 0.451100260019302
238020 0.451100260019302
239020 0.451100260019302
240020 0.451100260019302
241020 0.473604559898376
242020 0.473604559898376
243020 0.473604559898376
244020 0.473604559898376
245020 0.473604559898376
246020 0.473604559898376
247020 0.473604559898376
248020 0.473604559898376
249020 0.654541373252869
250020 0.654541373252869
251020 0.654541373252869
252020 0.654541373252869
253020 0.654541373252869
254020 0.654541373252869
255020 0.654541373252869
256020 0.654541373252869
257020 0.63267594575882
258020 0.63267594575882
259020 0.63267594575882
260020 0.63267594575882
261020 0.63267594575882
262020 0.63267594575882
263020 0.63267594575882
264020 0.63267594575882
265020 0.582283675670624
266020 0.582283675670624
267020 0.582283675670624
268020 0.582283675670624
269020 0.582283675670624
270020 0.582283675670624
271020 0.582283675670624
272020 0.582283675670624
273020 0.729400277137756
274020 0.729400277137756
275020 0.729400277137756
276020 0.729400277137756
277020 0.729400277137756
278020 0.729400277137756
279020 0.729400277137756
280020 0.729400277137756
281020 0.702348172664642
282020 0.702348172664642
283020 0.702348172664642
284020 0.702348172664642
285020 0.702348172664642
286020 0.702348172664642
287020 0.702348172664642
288020 0.702348172664642
289020 0.691676735877991
290020 0.691676735877991
291020 0.691676735877991
292020 0.691676735877991
293020 0.691676735877991
294020 0.691676735877991
295020 0.691676735877991
296020 0.691676735877991
297020 0.68524181842804
298020 0.68524181842804
299020 0.68524181842804
300020 0.68524181842804
301020 0.695001006126404
302020 0.695001006126404
303020 0.695001006126404
304020 0.695001006126404
305020 0.695001006126404
306020 0.695001006126404
307020 0.695001006126404
308020 0.695001006126404
309020 0.695001006126404
310020 0.695001006126404
311020 0.695001006126404
312020 0.695001006126404
313020 0.695001006126404
314020 0.695001006126404
315020 0.695001006126404
316020 0.695001006126404
317020 0.757639408111572
318020 0.757639408111572
319020 0.757639408111572
320020 0.757639408111572
321020 0.757639408111572
322020 0.757639408111572
323020 0.757639408111572
324020 0.757639408111572
325020 0.757639408111572
326020 0.757639408111572
327020 0.757639408111572
328020 0.757639408111572
329020 0.757639408111572
330020 0.757639408111572
331020 0.757639408111572
332020 0.757639408111572
333020 1.00965082645416
334020 1.00965082645416
335020 1.00965082645416
336020 1.00965082645416
337020 1.00965082645416
338020 1.00965082645416
339020 1.00965082645416
340020 1.00965082645416
341020 1.00965082645416
342020 1.00965082645416
343020 1.00965082645416
344020 1.00965082645416
345020 1.00965082645416
346020 1.00965082645416
347020 1.00965082645416
348020 1.00965082645416
349020 1.06449055671692
350020 1.06449055671692
351020 1.06449055671692
352020 1.06449055671692
353020 1.06449055671692
354020 1.06449055671692
355020 1.06449055671692
356020 1.06449055671692
357020 1.06449055671692
358020 1.06449055671692
359020 1.06449055671692
360020 1.06449055671692
361020 1.06449055671692
362020 1.06449055671692
363020 1.06449055671692
364020 1.06449055671692
365020 1.15304410457611
366020 1.15304410457611
367020 1.15304410457611
368020 1.15304410457611
369020 1.15304410457611
370020 1.15304410457611
371020 1.15304410457611
372020 1.15304410457611
373020 1.15304410457611
374020 1.15304410457611
375020 1.15304410457611
376020 1.15304410457611
377020 1.15304410457611
378020 1.15304410457611
379020 1.15304410457611
380020 1.15304410457611
381020 1.17879939079285
382020 1.17879939079285
383020 1.17879939079285
384020 1.17879939079285
385020 1.17879939079285
386020 1.17879939079285
387020 1.17879939079285
388020 1.17879939079285
389020 1.17879939079285
390020 1.17879939079285
391020 1.17879939079285
392020 1.17879939079285
393020 1.17879939079285
394020 1.17879939079285
395020 1.17879939079285
396020 1.17879939079285
397020 1.36329913139343
398020 1.36329913139343
399020 1.36329913139343
};
\end{axis}

\end{tikzpicture}

%% file: labpal/figure_data/performance_comparison_low_batch/plots_bs32/CIFAR-10_DenseNet-121_training_loss.pgf
% This file was created by tikzplotlib v0.9.8.
\begin{tikzpicture}

\definecolor{color0}{rgb}{0.933333333333333,0.509803921568627,0.933333333333333}
\definecolor{color1}{rgb}{0.647058823529412,0.164705882352941,0.164705882352941}
\definecolor{color2}{rgb}{1,0.647058823529412,0}

\begin{axis}[
log basis y={10},
minor xtick={},
minor ytick={},
tick align=outside,
tick pos=left,
title={training loss CIFAR-10 DenseNet-121},
width=10.5cm,height=8cm,grid=major,major grid style={dotted},
x grid style={white!69.0196078431373!black},
xlabel={epochs},
xmin=-14.2, xmax=298.2,
xtick style={color=black},
xtick={-50,0,50,100,150,200,250,300},
y grid style={white!69.0196078431373!black},
ylabel={training loss},
ymin=0.00005, ymax=10,
ymode=log,
ytick style={color=black},
ytick={1e-05,0.0001,0.001,0.01,0.1,1,10,100}
]
\addplot [line width=2.5pt, color0, opacity=1.0]
table {%
0 2.37410378456116
1 2.11263704299927
2 1.6380740404129
3 1.47528421878815
4 1.32104563713074
5 1.21250212192535
6 1.11915969848633
7 1.06377351284027
8 1.0271178483963
9 0.950353980064392
10 0.902829110622406
11 0.866868734359741
12 0.814785957336426
13 0.782378554344177
14 0.746473789215088
15 0.719318091869354
16 0.682642757892609
17 0.669746220111847
18 0.625452697277069
19 0.625093698501587
20 0.591982662677765
21 0.585796117782593
22 0.567818164825439
23 0.564058780670166
24 0.520277142524719
25 0.51272439956665
26 0.500591218471527
27 0.496876180171967
28 0.487066805362701
29 0.464012235403061
30 0.466643810272217
31 0.447903871536255
32 0.435579746961594
33 0.434790104627609
34 0.420179516077042
35 0.403992772102356
36 0.39692884683609
37 0.397824347019196
38 0.382652938365936
39 0.369284778833389
40 0.374171257019043
41 0.364004284143448
42 0.353976160287857
43 0.348782926797867
44 0.337022513151169
45 0.347371011972427
46 0.332608789205551
47 0.319343358278275
48 0.316938042640686
49 0.313799321651459
50 0.308018445968628
51 0.297667473554611
52 0.296751707792282
53 0.295387178659439
54 0.277126461267471
55 0.278618514537811
56 0.281874507665634
57 0.260563313961029
58 0.261438608169556
59 0.262855172157288
60 0.249639734625816
61 0.255876004695892
62 0.254728108644485
63 0.23946587741375
64 0.244776144623756
65 0.229827255010605
66 0.231889382004738
67 0.229366451501846
68 0.225783914327621
69 0.229230925440788
70 0.219207957386971
71 0.207100912928581
72 0.21511422097683
73 0.206841856241226
74 0.207833006978035
75 0.204892709851265
76 0.203919619321823
77 0.192765980958939
78 0.195752695202827
79 0.19518306851387
80 0.191324934363365
81 0.181730896234512
82 0.180117279291153
83 0.185959547758102
84 0.185126706957817
85 0.179072842001915
86 0.183761581778526
87 0.18240462243557
88 0.176073297858238
89 0.171166598796844
90 0.170072808861732
91 0.165437966585159
92 0.162724271416664
93 0.163420364260674
94 0.1627546697855
95 0.167178347706795
96 0.161235839128494
97 0.151515513658524
98 0.153850764036179
99 0.148060336709023
100 0.147626936435699
101 0.141091048717499
102 0.141664296388626
103 0.140705898404121
104 0.138388484716415
105 0.135475888848305
106 0.135381758213043
107 0.138003662228584
108 0.136425539851189
109 0.129536405205727
110 0.129797950387001
111 0.128718122839928
112 0.123550407588482
113 0.126118406653404
114 0.128062397241592
115 0.12221197783947
116 0.119337506592274
117 0.118045724928379
118 0.119121447205544
119 0.116721741855145
120 0.112981975078583
121 0.108588449656963
122 0.11593709141016
123 0.117119327187538
124 0.114103816449642
125 0.107234343886375
126 0.108888164162636
127 0.105220526456833
128 0.106320552527905
129 0.103520736098289
130 0.109536655247211
131 0.100526131689548
132 0.10213914513588
133 0.109002485871315
134 0.101685337722301
135 0.0989820212125778
136 0.0955306887626648
137 0.102417662739754
138 0.0990636497735977
139 0.0996313914656639
140 0.0979161560535431
141 0.0979676768183708
142 0.0939944162964821
143 0.0925341099500656
144 0.0964850261807442
145 0.0855733454227448
146 0.0875023305416107
147 0.0906692817807198
148 0.0894964188337326
149 0.09044449031353
150 0.0873476788401604
151 0.0883870348334312
152 0.0873795375227928
153 0.083484873175621
154 0.0864651724696159
155 0.0834783241152763
156 0.0860597342252731
157 0.0862895399332047
158 0.0865291357040405
159 0.082659512758255
160 0.0839173048734665
161 0.0783865228295326
162 0.0783684328198433
163 0.0775708854198456
164 0.0848316475749016
165 0.0770766139030457
166 0.0785360485315323
167 0.0745567157864571
168 0.0739491134881973
169 0.0729325637221336
170 0.0771292224526405
171 0.0806656330823898
172 0.0697465538978577
173 0.073096752166748
174 0.0745043307542801
175 0.0687715783715248
176 0.0731827020645142
177 0.0752760544419289
178 0.0718123316764832
179 0.0698574334383011
180 0.0652920380234718
181 0.0656904205679893
182 0.0718336254358292
183 0.0696171522140503
184 0.0652277693152428
185 0.0720472857356071
186 0.0705771818757057
187 0.0716281458735466
188 0.0686608850955963
189 0.0690044611692429
190 0.0620111338794231
191 0.0670796409249306
192 0.0651621297001839
193 0.0678878501057625
194 0.0675215944647789
195 0.0721891149878502
196 0.0660820081830025
197 0.0650021433830261
198 0.0660555809736252
199 0.0573298372328281
200 0.0612388402223587
201 0.0594034567475319
202 0.061943419277668
203 0.0676857456564903
204 0.0608105137944221
205 0.0664009749889374
206 0.0668582022190094
207 0.0703759342432022
208 0.0689470395445824
209 0.0676162168383598
210 0.0671312063932419
211 0.0720856040716171
212 0.0700751692056656
213 0.0693287551403046
214 0.0721734613180161
215 0.0712512508034706
216 0.0644268617033958
217 0.0736377462744713
218 0.0700146704912186
219 0.0682844743132591
220 0.0701041743159294
221 0.0690644010901451
222 0.0670258551836014
223 0.0749773606657982
224 0.0613823719322681
225 0.0639325454831123
226 0.0761375799775124
227 0.0708170756697655
228 0.0620080269873142
229 0.0700271651148796
230 0.0655287653207779
231 0.07344601303339
232 0.0637142658233643
233 0.0688487440347672
234 0.0683649554848671
235 0.0680724754929543
236 0.0665686726570129
237 0.0680628195405006
238 0.0671367645263672
239 0.0659409388899803
240 0.0668112859129906
241 0.0754575580358505
242 0.0719610154628754
243 0.0696917474269867
244 0.0691147819161415
245 0.0627963617444038
246 0.0721295103430748
247 0.0719298049807549
248 0.072200782597065
249 0.069501981139183
250 0.0699902549386024
251 0.0682210549712181
252 0.0674367472529411
253 0.0706242620944977
254 0.0704904198646545
255 0.0616364665329456
256 0.0735472366213799
257 0.0648876652121544
258 0.0684820264577866
259 0.0678661689162254
260 0.0706706345081329
261 0.0644749477505684
262 0.0700266659259796
263 0.0727490410208702
264 0.0678906366229057
265 0.0733067467808723
266 0.0692681297659874
267 0.068360410630703
268 0.0713283941149712
269 0.0709889903664589
270 0.0700035989284515
271 0.0669423490762711
272 0.0703315064311028
273 0.0711235031485558
274 0.0675360187888145
275 0.0616366341710091
276 0.07014250010252
277 0.072284109890461
278 0.0679267644882202
279 0.0651350319385529
280 0.070469930768013
281 0.0684867203235626
282 0.0684430748224258
283 0.0713949576020241
284 0.0689654052257538
};
\addplot [line width=2.5pt, color1, opacity=1.0]
table {%
0 2.39626351992289
1 2.31298633416494
2 1.73022242387136
3 1.40801072120667
4 1.27516869703929
5 1.10289965073268
6 1.05999338626862
7 0.902894000212351
8 0.97656124830246
9 0.887552539507548
10 0.903146326541901
11 0.850688755512238
12 0.846082071463267
13 0.855353097120921
14 1.64496664206187
15 28.8386301994324
16 33.9075519243876
};
\addplot [line width=2.5pt, red, opacity=1.0]
table {%
0 2.37675476074219
1 1.64857721328735
2 1.29507499933243
3 1.14816075563431
4 1.11235743761063
5 0.961838603019714
6 0.861434817314148
7 0.811680555343628
8 0.728533715009689
9 0.685496151447296
10 0.661962568759918
11 0.600625276565552
12 0.564288973808289
13 0.536853551864624
14 0.493012145161629
15 0.466624408960342
16 0.44277848303318
17 0.414440810680389
18 0.39636093378067
19 0.375313475728035
20 0.353255793452263
21 0.335591107606888
22 0.321688339114189
23 0.306178316473961
24 0.291536316275597
25 0.28372710943222
26 0.266726300120354
27 0.255673334002495
28 0.241875506937504
29 0.232966355979443
30 0.233526140451431
31 0.221223592758179
32 0.210880182683468
33 0.199962697923183
34 0.192307718098164
35 0.183856017887592
36 0.174819700419903
37 0.165540680289268
38 0.162620939314365
39 0.151312649250031
40 0.147732257843018
41 0.136291801929474
42 0.134238548576832
43 0.127862319350243
44 0.127709425985813
45 0.125545926392078
46 0.124323956668377
47 0.119272939860821
48 0.116664983332157
49 0.112121976912022
50 0.105153497308493
51 0.104076094925404
52 0.10200385004282
53 0.0973434373736382
54 0.0922885350883007
55 0.0883377008140087
56 0.0854926742613316
57 0.0826465748250484
58 0.0791321322321892
59 0.0773719996213913
60 0.0763856917619705
61 0.0700041726231575
62 0.0665194690227509
63 0.0668139010667801
64 0.060053551569581
65 0.0597352460026741
66 0.0575285088270903
67 0.0528721082955599
68 0.0521121360361576
69 0.050421666353941
70 0.0506859347224236
71 0.0510488711297512
72 0.0457159392535686
73 0.042035786435008
74 0.0424703303724527
75 0.039855943992734
76 0.041644211858511
77 0.0376032348722219
78 0.0357116293162107
79 0.0349899847060442
80 0.0334995249286294
81 0.0331455320119858
82 0.0317954532802105
83 0.0316176703199744
84 0.0278573976829648
85 0.027004704810679
86 0.0279987752437592
87 0.025578242726624
88 0.0232548695057631
89 0.0236761448904872
90 0.0242253169417381
91 0.0235695438459516
92 0.0225327285006642
93 0.0212547667324543
94 0.0208977265283465
95 0.020347973331809
96 0.0192042710259557
97 0.0191643703728914
98 0.0180803453549743
99 0.0169533649459481
100 0.0172781893052161
101 0.0159954256378114
102 0.014878123998642
103 0.0154404970817268
104 0.0147972581908107
105 0.0142887993715703
106 0.0128386085852981
107 0.0126488138921559
108 0.0114268152974546
109 0.0118427779525518
110 0.00981175363995135
111 0.0112170102074742
112 0.00898808194324374
113 0.00827715708874166
114 0.0077409315854311
115 0.00674166926182806
116 0.00772818899713457
117 0.0060452411416918
118 0.00554008386097848
119 0.00465484929736704
120 0.00458318763412535
121 0.00501151499338448
122 0.00465204077772796
123 0.00485123042017221
124 0.00414777698460966
125 0.00464629614725709
126 0.00587483821436763
127 0.00710894679650664
128 0.00690408598165959
129 0.00672998791560531
130 0.00730227131862193
131 0.00682279991451651
132 0.00582085072528571
133 0.00683645578101277
134 0.00815653172321618
135 0.00712779944296926
136 0.00801606418099254
137 0.00685258337762207
138 0.00768562545999885
139 0.0073506438639015
140 0.00589120760560036
141 0.00702540192287415
142 0.0072471137100365
143 0.00819028803380206
144 0.00587932561757043
145 0.00652080186409876
146 0.00505780967068858
147 0.00631845841417089
148 0.00619821395957842
149 0.00575023575220257
150 0.00487843039445579
151 0.00459051737561822
152 0.00551979435840622
153 0.00478640315122902
154 0.00452321078046225
155 0.00444572709966451
156 0.00430225412128493
157 0.00529404447297566
158 0.00549428124213591
159 0.00508324685506523
160 0.00537151846219786
161 0.00503060140181333
162 0.00444437560508959
163 0.00522900524083525
164 0.00442313085659407
165 0.00420798582490534
166 0.00341693652444519
167 0.00613054563291371
168 0.00533491547685117
169 0.00438861473230645
170 0.00425638089654967
171 0.00471535057295114
172 0.00353316032123985
173 0.00397284253267571
174 0.0040264411654789
175 0.00497772057133261
176 0.00468547065975145
177 0.00418639027338941
178 0.00468842707050499
179 0.00404991761024576
180 0.00400119132609689
181 0.00417697126977146
182 0.00501634323154576
183 0.00385693412681576
184 0.00363840521458769
185 0.00447070418158546
186 0.00540949024434667
187 0.00518375939282123
188 0.00396936628385447
189 0.00310125810938189
190 0.00391742397187045
191 0.00478435853074188
192 0.00324971442387323
193 0.0046269376653072
194 0.00352028638008051
195 0.00390692023211159
196 0.00440295292355586
197 0.00366116303484887
198 0.00417755833768751
199 0.00429008578066714
200 0.00572159115108661
201 0.00461080920285895
202 0.00461809425542015
203 0.00486825569532812
204 0.00562621810240671
205 0.00463079672772437
206 0.00453636049496708
207 0.00346008803171571
208 0.00172111979918554
209 0.00273565513634821
210 0.00239753704227041
211 0.00203165048878873
212 0.00215188538459188
213 0.00200857882737182
214 0.00165411785155811
215 0.00252070776150504
216 0.001772319606971
217 0.00128739490901353
218 0.00132051253604004
219 0.00156080714077689
220 0.00106654400588013
221 0.00124826694809599
222 0.00197667644533794
223 0.00152817089110613
224 0.00141910897218622
225 0.00095996777235996
226 0.0015026582259452
227 0.000709353887941688
228 0.00120194576265931
229 0.000899737356576225
230 0.00147793878568336
231 0.000775041291490197
232 0.000964709455729462
233 0.000688250402163249
234 0.000641373510006815
235 0.000581663174671121
236 0.00122741915402003
237 0.00108412252302514
238 0.0005265553627396
239 0.000930863403482363
240 0.000830511410640611
241 0.000747217422031099
242 0.000680556950101163
243 0.000762474218390707
244 0.00121575156663312
245 0.00131448141473811
246 0.000875802823429694
247 0.000726610858691856
248 0.000721949113540177
249 0.000298292827210389
250 0.00043554114017752
251 0.000424643032602035
252 0.000708730738551822
253 0.000619938229647232
254 0.000259140579146333
255 0.000535289109848236
256 0.000635256852547172
257 0.000778673485910986
258 0.000466258474148162
259 0.000284863917840994
260 0.000741529547667596
261 0.000261058420164773
262 0.000280015086900676
263 0.000269365938947885
264 0.000243268194026314
265 0.000231449528655503
266 0.000144700588407431
267 0.000218018380110152
268 0.000281353346508695
269 0.000233483689953573
270 0.000303304248518543
271 0.000273610858130269
272 0.000693770482030232
273 0.000209826474019792
274 0.000691649216605583
275 0.000598735212406609
276 0.000262210581240652
277 0.000297429636702873
278 0.000164005428814562
279 0.0003037121205125
280 4.50768275186419e-05
281 0.000235315346799325
282 0.000124017560210632
283 3.70944544556551e-05
284 0.000226081499931752
};
\addplot [line width=2.5pt, color2, opacity=1.0]
table {%
0 2.39626344045003
2 1.87688219547272
4 1.4129806359609
6 1.21624294916789
8 1.06627655029297
10 0.933724562327067
12 0.837472995122274
14 0.75414111216863
16 0.690883636474609
18 0.635125736395518
20 0.596083482106527
22 0.551489194234212
24 0.506591856479645
26 0.476546088854472
28 0.44644820690155
30 0.423635284105937
32 0.395293106635412
34 0.372405846913656
36 0.349947174390157
38 0.330107609430949
40 0.313510457674662
42 0.293022046486537
44 0.273036589225133
46 0.256639321645101
48 0.23928656677405
50 0.226570819815
52 0.210712810357412
54 0.200800473491351
56 0.186768089731534
58 0.181551332275073
60 0.172013690074285
62 0.161719699700673
64 0.153558408220609
66 0.14469084640344
68 0.136465951800346
70 0.133290981252988
72 0.125356823205948
74 0.122060048083464
76 0.115909308195114
78 0.112886500855287
80 0.109556200603644
82 0.105834282934666
84 0.105476426581542
86 0.098239079117775
88 0.0940127074718475
90 0.0909688696265221
92 0.08976033081611
94 0.0871340334415436
96 0.0871948425968488
98 0.084761177500089
100 0.0828162108858426
102 0.0803949932257334
104 0.0793623551726341
106 0.0788272395730019
108 0.0757418473561605
110 0.0730082467198372
112 0.072931299606959
114 0.0718397299448649
116 0.0707266281048457
118 0.0716946572065353
120 0.068113607664903
122 0.0671483476956685
124 0.0671630774935087
126 0.0663158694903056
128 0.0669613281885783
130 0.0654914329449336
132 0.0633065936466058
134 0.0622921797136466
136 0.0620878425737222
138 0.0610797367990017
140 0.0621243032316367
142 0.0600693734983603
144 0.0602375740806262
146 0.0606185955305894
148 0.0601237528026104
150 0.0590185572703679
152 0.0580855247875055
154 0.0583429286877314
156 0.0573103614151478
158 0.0574890573819478
160 0.0555190021793048
162 0.0582619781295458
164 0.0551251868406932
166 0.0567280203104019
168 0.0558704137802124
170 0.0547384234766165
172 0.0524295630554358
174 0.0541295719643434
176 0.0543450204034646
178 0.0527076373497645
180 0.0528932002683481
182 0.0529249931375186
184 0.0534114614129066
186 0.051971223205328
188 0.0509791572888692
190 0.0510349832475185
192 0.0516814030706882
194 0.0518399216234684
196 0.0508369257052739
198 0.0504374504089355
200 0.0510235652327538
202 0.0501142839590708
204 0.0499248293538888
206 0.0497089053193728
208 0.0467302761971951
210 0.0485764443874359
212 0.0487895794212818
214 0.0483723965783914
216 0.0486689979831378
218 0.0488613831500212
220 0.047200166930755
222 0.0482702081402143
224 0.0488875632484754
226 0.0471639161308606
228 0.0484681266049544
230 0.0473245941102505
232 0.0476268194615841
234 0.0491940788924694
236 0.0465046900014083
238 0.0469561877350012
240 0.0460806613167127
242 0.045696634799242
244 0.0470238626003265
246 0.0448353178799152
248 0.0445202228923639
250 0.0462814345955849
252 0.0446730069816113
254 0.0446601447959741
256 0.0452622013787429
258 0.0440855249762535
260 0.0448491002122561
262 0.0445403195917606
264 0.043704545746247
266 0.0456239432096481
268 0.0444220850865046
270 0.0446657041708628
272 0.0456120806435744
274 0.0444565849999587
276 0.045073031137387
278 0.042328242212534
280 0.0444600271681945
282 0.0431299780805906
284 0.0438310119012992
};
\addplot [line width=2.5pt, green!50.1960784313725!black, opacity=1.0]
table {%
0 2.39626344045003
1 1.78972256183624
2 1.36282229423523
3 1.12831898530324
4 0.974613189697266
5 0.861735284328461
6 0.772989352544149
7 0.700220684210459
8 0.636089523633321
9 0.577987770239512
10 0.530751009782155
11 0.750534812609355
12 0.568761865297953
13 0.506119082371394
14 0.470687141021093
15 0.431860287984212
16 0.409071485201518
17 0.384376923243205
18 0.361323903004328
19 0.339971611897151
20 0.329176872968674
21 0.311972916126251
22 0.292898163199425
23 0.27970327436924
24 0.268758068482081
25 0.257272884249687
26 0.241504549980164
27 0.226218461990356
28 0.22194633881251
29 0.211715633670489
30 0.202668418486913
31 0.192113886276881
32 0.191775992512703
33 0.177710756659508
34 0.169373954335848
35 0.159675007065137
36 0.15314893424511
37 0.147856342295806
38 0.152271643280983
39 0.143918161590894
40 0.129762709140778
41 0.125341951847076
42 0.119219263394674
43 0.114640610913436
44 0.111561382810275
45 0.10940399269263
46 0.105227537453175
47 0.107513544460138
48 0.10036630431811
49 0.0931002597014109
50 0.096093513071537
51 0.0889641096194585
52 0.0869680816928546
53 0.0815296346942584
54 0.0774724582831065
55 0.0772656475504239
56 0.0717695864538352
57 0.0714014495412509
58 0.0675665264328321
59 0.0654593457778295
60 0.0643165931105614
61 0.062594381471475
62 0.0620390002926191
63 0.0609287284314632
64 0.0585265743235747
65 0.0553072517116865
66 0.0517689163486163
67 0.0517325500647227
68 0.0527713783085346
69 0.0523890828092893
70 0.0565719169874986
71 0.0508154419561227
72 0.0484893098473549
73 0.0464178870121638
74 0.0437311592201392
75 0.0413778300086657
76 0.0420920935769876
77 0.0424219767252604
78 0.0415647365152836
79 0.0381179464360078
80 0.0387256977458795
81 0.0375897300740083
82 0.0362627257903417
83 0.0364472543199857
84 0.0343459397554398
85 0.0341695565730333
86 0.0353613241265217
87 0.0344590445359548
88 0.0344475992023945
89 0.0349981511632601
90 0.0351407000174125
91 0.0320501755923033
92 0.0305248976995548
93 0.033811475460728
94 0.0303976275026798
95 0.0300538272907337
96 0.0286027447630962
97 0.0280592553317547
98 0.0284116100519896
99 0.028617275878787
100 0.0268909179915984
101 0.0269845991084973
102 0.0239747477074464
103 0.027983644977212
104 0.0262253185113271
105 0.0252522882074118
106 0.0246526269863049
107 0.0226894461860259
108 0.0225575106839339
109 0.0248044499506553
110 0.0234141902377208
111 0.0275038598726193
112 0.0254972241818905
113 0.0221395151068767
114 0.021486929928263
115 0.0227457980314891
116 0.0246358861525853
117 0.0225822149465481
118 0.0211030474553506
119 0.0221163344879945
120 0.0219655527422826
121 0.0193937762329976
122 0.0206891844669978
123 0.0202086015293996
124 0.0199549980461597
125 0.0216136487821738
126 0.0205090412249168
127 0.018317305482924
128 0.0179531040290991
129 0.0192053467035294
130 0.0192658534894387
131 0.0179242795954148
132 0.0184300933033228
133 0.0193750125666459
134 0.0172163341194391
135 0.0183844286948442
136 0.0187064036726952
137 0.0169946973522504
138 0.0186773377160231
139 0.0165626198674242
140 0.0202882286782066
141 0.0164776410286625
142 0.0185774378478527
143 0.0111080715432763
144 0.0062903338111937
145 0.00528087901572386
146 0.00435254691789548
147 0.00472982755551736
148 0.00382127240300179
149 0.00376631288478772
150 0.00327773415483534
151 0.00319448004787167
152 0.00275439846639832
153 0.0026234428708752
154 0.00270543922670186
155 0.00244298577308655
156 0.0028643513796851
157 0.00226397796844443
158 0.00232093882126113
159 0.00221668645584335
160 0.00221415946725756
161 0.00213717475223045
162 0.00207136718866726
163 0.00215875760962566
164 0.00184821182241042
165 0.00199318312418958
166 0.00175529322586954
167 0.00164439750369638
168 0.00180889825181415
169 0.00171988589378695
170 0.00156184813628594
171 0.00161886250134557
172 0.00163182259226839
173 0.00170554236198465
174 0.00121606892207637
175 0.00128610474833598
176 0.00157614145427942
177 0.00166332733351737
178 0.00134291779249907
179 0.00146303519917031
180 0.0014350328516836
181 0.00148166120440389
182 0.00135560911924889
183 0.00123606404910485
184 0.00153006861607234
185 0.00120386133979385
186 0.00130431912839413
187 0.00115686852950603
188 0.00129073208275562
189 0.00110069280102228
190 0.00116404052823782
191 0.00132300386515756
192 0.00121261656749994
193 0.00130956701468676
194 0.000931382082247486
195 0.00107657395225639
196 0.00103358402460193
197 0.00100283879631509
198 0.00127687237303083
199 0.00108092027949169
200 0.00109932439712187
201 0.00104659776358555
202 0.000855385694497575
203 0.001164885237813
204 0.00129212704875196
205 0.0010195753032652
206 0.0010481364831018
207 0.000861324350504825
208 0.000902576701870809
209 0.000964715067918102
210 0.000822980461331705
211 0.000799560841793815
212 0.000875040170891831
213 0.000902378310759862
214 0.00106710341060534
215 0.000638338873007645
216 0.000943954490746061
217 0.000765217060688883
218 0.00092939465927581
219 0.000808856178385516
220 0.000945712982987364
221 0.000882831382720421
222 0.000783140363637358
223 0.000701017212122679
224 0.000812020987117042
225 0.000901536162321766
226 0.000979284620067726
227 0.00076876274154832
228 0.000700666433355461
229 0.000721448954815666
230 0.000609824996596823
231 0.000856952188769355
232 0.00083800593468671
233 0.000990870777362337
234 0.000794479448813945
235 0.000842863577418029
236 0.000710113061359152
237 0.000831290904898196
238 0.00080240461587285
239 0.000629947287961841
240 0.000807955123794576
241 0.000873973913257942
242 0.000829973840154707
243 0.000899946957360953
244 0.000888264078336457
245 0.000858113053254783
246 0.000931892699251572
247 0.00075141117364789
248 0.000798734448229273
249 0.000934149565485617
250 0.000902997892505179
251 0.000908898868753264
252 0.000839932821691036
253 0.000888828653842211
254 0.000724031706340611
255 0.000719840395807599
256 0.00071695662336424
257 0.000704112035843233
258 0.000896736300395181
259 0.000754394723723332
260 0.000726674140120546
261 0.000728755490854383
262 0.000669368969586988
263 0.000774575610800336
264 0.000834043234741936
265 0.000783387241729846
266 0.000553477157761032
267 0.000727978457386295
268 0.000612443623443445
269 0.000745226085806886
270 0.000890790950506926
271 0.000690240121912211
272 0.000805184516745309
273 0.000797218080454816
274 0.000761382048949599
275 0.000591909978538752
276 0.000666923229194557
277 0.000723902485333383
278 0.000803112459834665
279 0.000739031975778441
280 0.000750272088528921
281 0.000815722276456654
282 0.000860522598183403
283 0.000966791141157349
284 0.000759344082325697
};
\addplot [line width=2.5pt, black, opacity=1.0]
table {%
0 2.39626344045003
1 1.83898735046387
2 1.48437094688416
3 1.2804350455602
4 1.08082282543182
5 0.879706144332886
6 0.81853703657786
7 0.758288701375326
8 0.715769588947296
9 0.672856271266937
10 0.65533310174942
11 0.597477932771047
12 0.57491546869278
13 0.487379978100459
14 0.466114381949107
15 0.471034745375315
16 0.471292495727539
17 0.438007295131683
18 0.405245661735535
19 0.392647981643677
20 0.419500788052877
21 0.393310934305191
22 0.372691829999288
23 0.338040252526601
24 0.300463368495305
25 0.314726253350576
26 0.276527399818103
27 0.260373199979464
28 0.255639602740606
29 0.240403021375338
30 0.234892105062803
31 0.252234846353531
32 0.268964404861132
33 0.251021906733513
34 0.220352053642273
35 0.218420013785362
36 0.200261826316516
37 0.186049039165179
38 0.197187334299088
39 0.181633238991102
40 0.172657072544098
41 0.179596811532974
42 0.174877668420474
43 0.1685471534729
44 0.175015643239021
45 0.154313335816065
46 0.140563433369001
47 0.123286510507266
48 0.119854936997096
49 0.104723806182543
50 0.104560621082783
51 0.105509102344513
52 0.101408794522285
53 0.0962884624799093
54 0.092836561302344
55 0.0818912610411644
56 0.0788090452551842
57 0.0752881268660228
58 0.0744097183148066
59 0.0717355671028296
60 0.0765940248966217
61 0.0693824042876561
62 0.0618867216010888
63 0.0618909349044164
64 0.0609830319881439
65 0.0596114943424861
66 0.0599449450771014
67 0.0549276955425739
68 0.0573075798650583
69 0.0534778088331223
70 0.051107553144296
71 0.0496834889054298
72 0.0498472464581331
73 0.0488326090077559
74 0.0477107713619868
75 0.0416989798347155
76 0.0351029392331839
77 0.0391983973483245
78 0.0353695458422104
79 0.0352070312947035
80 0.0342085702965657
81 0.0333458787451188
82 0.0333277508616447
83 0.0299616952737172
84 0.0317827512820562
85 0.0312016954024633
86 0.0326400293658177
87 0.0276316317419211
88 0.0272641635189454
89 0.029305142040054
90 0.0288523392130931
91 0.0306070918838183
92 0.0297137262920539
93 0.0276863556355238
94 0.02662811614573
95 0.023946641633908
96 0.0240092550714811
97 0.0251356003185113
98 0.0242232450594505
99 0.0235837095727523
100 0.0238749142736197
101 0.0224639239410559
102 0.0217766147106886
103 0.0212005308518807
104 0.0209768041968346
105 0.0189339633410176
106 0.0180496921141942
107 0.018576482621332
108 0.0195876856644948
109 0.0178427242984374
110 0.018701002933085
111 0.0188958688328664
112 0.0180316840608915
113 0.016491813895603
114 0.0167704746127129
115 0.0180599223822355
116 0.0169969868535797
117 0.0155791155993938
118 0.01580809460332
119 0.0152923294032613
120 0.0156804537400603
121 0.0146909731750687
122 0.0150770442560315
123 0.0152473434184988
124 0.0148623147979379
125 0.0135400798171759
126 0.0143767644961675
127 0.0144324619323015
128 0.0140467109158635
129 0.0125958000620206
130 0.0135690442596873
131 0.012668707408011
132 0.0130671573181947
133 0.0130045783395569
134 0.0128109625851115
135 0.0124737378209829
136 0.0120116922383507
137 0.0121722857778271
138 0.0118069667369127
139 0.0103377566362421
140 0.0104685090482235
141 0.0108193848282099
142 0.0114025097961227
143 0.0103490728264054
144 0.0111230077842871
145 0.0114141398419937
146 0.0107335802167654
147 0.0114593974625071
148 0.0104065596436461
149 0.0109037822112441
150 0.0104965396846334
151 0.0101787407572071
152 0.0111282328143716
153 0.00989433936774731
154 0.00990808568894863
155 0.0102268109718959
156 0.0103180259466171
157 0.0115435573582848
158 0.0103265137101213
159 0.00968445806453625
160 0.00986364297568798
161 0.00978424896796544
162 0.010205009020865
163 0.00983858729402224
164 0.00987185382594665
165 0.00987515660623709
166 0.0105480654165149
167 0.0107853589579463
168 0.0107527837778131
169 0.0100447610020638
170 0.00998762156814337
171 0.00984048936516047
172 0.00930030674984058
173 0.00857095886021852
174 0.00971334024022023
175 0.00920589407905936
176 0.00953757421423992
177 0.00973607444514831
178 0.00886458080882827
179 0.0085964606453975
180 0.00957827627037962
181 0.0100687028219303
182 0.00881774351000786
183 0.00935187035550674
184 0.00993056067576011
185 0.0099572679027915
186 0.00863751225794355
187 0.00876452665155133
188 0.00938731556137403
189 0.00916259301205476
190 0.00973642617464066
191 0.00918455344314376
192 0.00965813842291633
193 0.00897677506630619
194 0.00925579108297825
195 0.00912471953779459
196 0.00921907446657618
197 0.0090113611270984
198 0.00925164514531692
199 0.00870911466578642
200 0.00928613574554523
201 0.00948909375195702
202 0.00924566062167287
203 0.00980961105475823
204 0.00964390517522891
205 0.00897726478676001
206 0.00930472246060769
207 0.00921889549742142
208 0.00865506799891591
209 0.0093362374852101
210 0.00893551126743356
211 0.00867032430445155
212 0.00923420178393523
213 0.00937975166986386
214 0.00970557052642107
215 0.00935321285699805
216 0.00855449431886276
217 0.00938057200983167
218 0.008812817124029
219 0.00878123116369049
220 0.00955007039010525
221 0.009590703373154
222 0.0102076500964661
223 0.00971290251861016
224 0.00923472720508774
225 0.00923658286531766
226 0.00929919366414348
227 0.00781997495020429
228 0.00867360442255934
229 0.00905908861507972
230 0.00935904821380973
231 0.00965342468892535
232 0.00895037672792872
233 0.00889148407926162
234 0.0095907316232721
235 0.00906005284438531
236 0.00874758356561263
237 0.00961932384719451
238 0.0092491105509301
239 0.00920498852307598
240 0.00939537569259604
241 0.00919236118594805
242 0.00889503381525477
243 0.00836776879926523
244 0.0089919736298422
245 0.00902217626571655
246 0.00876908392334978
247 0.00933446021129688
248 0.00926066593577464
249 0.00916852289810777
250 0.00891814722369115
251 0.00886025993774335
252 0.00925353510926167
253 0.00958024927725395
254 0.00892758974805474
255 0.00937110899637143
256 0.00928344732771317
257 0.00857071935509642
258 0.00907185549537341
259 0.00925298671548565
260 0.00956503953784704
261 0.00974779017269611
262 0.0082779402534167
263 0.00870087090879679
264 0.00839168283467492
265 0.0096511662316819
266 0.00859858716527621
267 0.00934144109487534
268 0.00896870360399286
269 0.00893097743391991
270 0.00868961804856857
271 0.00819737029572328
272 0.00854533134649197
273 0.00911900540813804
274 0.00958553655073047
275 0.00901980527366201
276 0.00850648743410905
277 0.00876011078556379
278 0.00956025517856081
279 0.0085540193443497
280 0.00848833347360293
281 0.00958192146693667
282 0.0093281123166283
283 0.0090007803713282
284 0.00968271323169271
};
\addplot [line width=2.5pt, blue, opacity=1.0]
table {%
0 2.37940573692322
1 1.65909695625305
2 1.30895471572876
3 1.16220688819885
4 1.1232545375824
5 0.945475161075592
6 0.86469841003418
7 0.813531816005707
8 0.715623259544373
9 0.671172082424164
10 0.629938066005707
11 0.570214092731476
12 0.535140156745911
13 0.503196060657501
14 0.464444786310196
15 0.449390679597855
16 0.426727712154388
17 0.393736511468887
18 0.396019011735916
19 0.375557541847229
20 0.352615535259247
21 0.340631663799286
22 0.322317391633987
23 0.305553078651428
24 0.314988911151886
25 0.293889462947845
26 0.272793680429459
27 0.257703840732574
28 0.240296274423599
29 0.236505672335625
30 0.223531514406204
31 0.211379677057266
32 0.200253650546074
33 0.194090440869331
34 0.19049009680748
35 0.176037028431892
36 0.167642876505852
37 0.155891463160515
38 0.149712920188904
39 0.138005256652832
40 0.136203765869141
41 0.128093853592873
42 0.126941874623299
43 0.118113286793232
44 0.123213641345501
45 0.118359632790089
46 0.123807825148106
47 0.112654142081738
48 0.111990258097649
49 0.106165789067745
50 0.103254847228527
51 0.103062897920609
52 0.0952605679631233
53 0.0919298902153969
54 0.0894540101289749
55 0.0854624733328819
56 0.0799407586455345
57 0.078885979950428
58 0.0750455856323242
59 0.0777239799499512
60 0.074401430785656
61 0.0657056793570518
62 0.066330187022686
63 0.0568484775722027
64 0.0537789277732372
65 0.0535534806549549
66 0.0477948077023029
67 0.0438040494918823
68 0.041724544018507
69 0.0421046949923038
70 0.0350806526839733
71 0.0353999771177769
72 0.0324632935225964
73 0.0283533744513988
74 0.0301747508347034
75 0.0263671539723873
76 0.0223463904112577
77 0.0232630949467421
78 0.0253826230764389
79 0.0261377841234207
80 0.0205928292125463
81 0.0257228817790747
82 0.0204409379512072
83 0.0223258845508099
84 0.0185203794389963
85 0.0166095905005932
86 0.0170064829289913
87 0.0136190019547939
88 0.0132343266159296
89 0.013918518088758
90 0.01301861833781
91 0.0133505314588547
92 0.012294958345592
93 0.0110842902213335
94 0.00771815748885274
95 0.00768282124772668
96 0.00766144832596183
97 0.00807882007211447
98 0.009119912981987
99 0.0133982421830297
100 0.0155907617881894
101 0.0138128036633134
102 0.0106026483699679
103 0.0150828119367361
104 0.0134939197450876
105 0.012306360527873
106 0.0112414630129933
107 0.00959359761327505
108 0.0109397945925593
109 0.0119255753234029
110 0.00952022895216942
111 0.0110898073762655
112 0.00866518542170525
113 0.0114551475271583
114 0.0133885880932212
115 0.0141924908384681
116 0.0153147717937827
117 0.0158870909363031
118 0.0159097295254469
119 0.0163313206285238
120 0.0128845516592264
121 0.013954727910459
122 0.0115848872810602
123 0.0126663828268647
124 0.0322379879653454
125 0.035164549946785
126 0.0286850892007351
127 0.0254731476306915
128 0.0320697873830795
129 0.0255366880446672
130 0.0307187344878912
131 0.0388447642326355
132 0.0331179387867451
133 0.0311034712940454
134 0.0264345891773701
135 0.0289553198963404
136 0.0285171754658222
137 0.0274297259747982
138 0.0302684884518385
139 0.0273567587137222
140 0.0263028331100941
141 0.0193956103175879
142 0.0196457989513874
143 0.0115978168323636
144 0.00598032306879759
145 0.00555597152560949
146 0.00414482038468122
147 0.00417296541854739
148 0.0033235177397728
149 0.00440254295244813
150 0.00382270198315382
151 0.00321585894562304
152 0.00284100952558219
153 0.00298112723976374
154 0.00327480607666075
155 0.00289270770736039
156 0.00301620131358504
157 0.00291527691297233
158 0.00264259846881032
159 0.00219233357347548
160 0.00265914970077574
161 0.00265163066796958
162 0.00291538052260876
163 0.00247999257408082
164 0.00224973866716027
165 0.00195048726163805
166 0.00269099604338408
167 0.00268789450637996
168 0.00223829504102468
169 0.0026006274856627
170 0.00209134863689542
171 0.00221865903586149
172 0.00203714985400438
173 0.00158197793643922
174 0.00171082420274615
175 0.00209679175168276
176 0.00245589599944651
177 0.0029256388079375
178 0.00253228051587939
179 0.00376834021881223
180 0.00392685132101178
181 0.00656614685431123
182 0.00685637863352895
183 0.00604316545650363
184 0.00545625714585185
185 0.00468174507841468
186 0.00479677459225059
187 0.00643965695053339
188 0.00585136748850346
189 0.00545806344598532
190 0.00543634872883558
191 0.00546286534518003
192 0.0033715961035341
193 0.00418554805219173
194 0.00312045961618423
195 0.0109891118481755
196 0.0117347855120897
197 0.010102529078722
198 0.00713707599788904
199 0.0070814467035234
200 0.00695914775133133
201 0.00785143114626408
202 0.00644478900358081
203 0.00660829758271575
204 0.0096964742988348
205 0.0074354144744575
206 0.00800181180238724
207 0.00843614153563976
208 0.00601400388404727
209 0.00723905814811587
210 0.0073815225623548
211 0.00461882352828979
212 0.00512368977069855
213 0.00782530847936869
214 0.00445747980847955
215 0.00221119448542595
216 0.0028712775092572
217 0.00212316657416523
218 0.00157090311404318
219 0.00191772531252354
220 0.00133790005929768
221 0.00200707092881203
222 0.00130815804004669
223 0.000961707788519561
224 0.00116303539834917
225 0.00160697405226529
226 0.00162830005865544
227 0.00105683098081499
228 0.00150321621913463
229 0.00139385566581041
230 0.00136182294227183
231 0.000977277872152627
232 0.000900506274774671
233 0.000654860690701753
234 0.000857937207911164
235 0.00108459056355059
236 0.00120717438403517
237 0.00108443526551127
238 0.00117343233432621
239 0.00174257776234299
240 0.00150041934102774
241 0.00129616679623723
242 0.00189626822248101
243 0.00131020706612617
244 0.00124432949814945
245 0.00118683255277574
246 0.00149117002729326
247 0.00107368151657283
248 0.00113661913201213
249 0.00192149903159589
250 0.002307667164132
251 0.00206667953170836
252 0.00184683571569622
253 0.00149792479351163
254 0.00126563163939863
255 0.00148981169331819
256 0.000976882874965668
257 0.00113240024074912
258 0.00153700332157314
259 0.00100453814957291
260 0.00117384956683964
261 0.00193861650768667
262 0.00242181029170752
263 0.002935447730124
264 0.00162386242300272
265 0.00249976455233991
266 0.00211467198096216
267 0.00228032865561545
268 0.00226473086513579
269 0.00215456797741354
270 0.0020745221991092
271 0.00177911517675966
272 0.00132052612025291
273 0.00159692019224167
274 0.00135916646104306
275 0.00132603524252772
276 0.0016743796877563
277 0.00286517082713544
278 0.00246291188523173
279 0.00230782106518745
280 0.00225598271936178
281 0.00258808489888906
282 0.00325773819349706
283 0.00408711237832904
284 0.0062231719493866
};
\end{axis}

\end{tikzpicture}

%% file: labpal/figure_data/performance_comparison_low_batch/plots_bs32/CIFAR-10_DenseNet-121_validation_accuracy.pgf
% This file was created by tikzplotlib v0.9.8.
\begin{tikzpicture}

\definecolor{color0}{rgb}{0.933333333333333,0.509803921568627,0.933333333333333}
\definecolor{color1}{rgb}{0.647058823529412,0.164705882352941,0.164705882352941}
\definecolor{color2}{rgb}{1,0.647058823529412,0}

\begin{axis}[
minor xtick={},
minor ytick={},
tick align=outside,
tick pos=left,
title={validation accuracy CIFAR-10 DenseNet-121},
width=10.5cm,height=8cm,grid=major,major grid style={dotted},
x grid style={white!69.0196078431373!black},
xlabel={epochs},
xmin=-13.15, xmax=298.15,
xtick style={color=black},
xtick={-50,0,50,100,150,200,250,300},
y grid style={white!69.0196078431373!black},
ylabel={validation accuracy},
ymin=0.85, ymax=0.94,
ytick style={color=black},
ytick={0.85,0.86,0.87,0.88,0.89,0.9,0.91,0.92,0.93,0.94}
]
\addplot [line width=2.5pt, color0, opacity=1.0]
table {%
1 0.377003192901611
2 0.350761204957962
3 0.512019217014313
4 0.552684307098389
5 0.599959909915924
6 0.621594548225403
7 0.654847741127014
8 0.67988783121109
9 0.663461565971375
10 0.66386216878891
11 0.707131385803223
12 0.722355782985687
13 0.733173072338104
14 0.750801265239716
15 0.754006385803223
16 0.741185903549194
17 0.75741183757782
18 0.7734375
19 0.793669879436493
20 0.787660241127014
21 0.795673072338104
22 0.809294879436493
23 0.805288434028625
24 0.798277258872986
25 0.799879789352417
26 0.81991183757782
27 0.794471144676208
28 0.809895813465118
29 0.810096144676208
30 0.822916686534882
31 0.817307710647583
32 0.827323734760284
33 0.837339758872986
34 0.832331717014313
35 0.829126596450806
36 0.847355782985687
37 0.847956717014313
38 0.850160241127014
39 0.852163434028625
40 0.842748403549194
41 0.844751596450806
42 0.854967951774597
43 0.854967951774597
44 0.847556114196777
45 0.843149065971375
46 0.855568885803223
47 0.864783644676208
48 0.845953524112701
49 0.854366958141327
50 0.85897433757782
51 0.855969548225403
52 0.845953524112701
53 0.870993614196777
54 0.853165090084076
55 0.844951927661896
56 0.86678683757782
57 0.860176265239716
58 0.852964758872986
59 0.861578524112701
60 0.865184307098389
61 0.863381385803223
62 0.864983975887299
63 0.85897433757782
64 0.85957533121109
65 0.86738783121109
66 0.854567289352417
67 0.8671875
68 0.866185903549194
69 0.862379789352417
70 0.864983975887299
71 0.865384638309479
72 0.86678683757782
73 0.871995210647583
74 0.873397409915924
75 0.87479966878891
76 0.877003192901611
77 0.863782048225403
78 0.868589758872986
79 0.871794879436493
80 0.87479966878891
81 0.872996807098389
82 0.87540066242218
83 0.878605782985687
84 0.878605782985687
85 0.874399065971375
86 0.878405451774597
87 0.883413434028625
88 0.8828125
89 0.87479966878891
90 0.885616958141327
91 0.894230782985687
92 0.889423072338104
93 0.877604186534882
94 0.88301283121109
95 0.886017620563507
96 0.888020813465118
97 0.880608975887299
98 0.881810903549194
99 0.886217951774597
100 0.889423072338104
101 0.884415090084076
102 0.886418282985687
103 0.888621807098389
104 0.895432710647583
105 0.887820541858673
106 0.886418282985687
107 0.885817289352417
108 0.887019217014313
109 0.889623403549194
110 0.88301283121109
111 0.880608975887299
112 0.89042466878891
113 0.889823734760284
114 0.889623403549194
115 0.887620210647583
116 0.887219548225403
117 0.882211565971375
118 0.89102566242218
119 0.88261216878891
120 0.886818885803223
121 0.891626596450806
122 0.895232379436493
123 0.878205120563507
124 0.892628192901611
125 0.89102566242218
126 0.880208313465118
127 0.894030451774597
128 0.883413434028625
129 0.89022433757782
130 0.885016024112701
131 0.888421475887299
132 0.888621807098389
133 0.892828524112701
134 0.883814096450806
135 0.893229186534882
136 0.886618614196777
137 0.887219548225403
138 0.887219548225403
139 0.887419879436493
140 0.889623403549194
141 0.894631385803223
142 0.89082533121109
143 0.893629789352417
144 0.888421475887299
145 0.890625
146 0.878806114196777
147 0.888421475887299
148 0.893229186534882
149 0.89823716878891
150 0.883613765239716
151 0.889423072338104
152 0.896634638309479
153 0.884415090084076
154 0.899439096450806
155 0.891225934028625
156 0.893830120563507
157 0.886217951774597
158 0.888221144676208
159 0.876802861690521
160 0.887820541858673
161 0.886818885803223
162 0.893629789352417
163 0.883814096450806
164 0.893028855323792
165 0.892027258872986
166 0.896834909915924
167 0.900440692901611
168 0.886017620563507
169 0.892227590084076
170 0.885016024112701
171 0.891826927661896
172 0.89102566242218
173 0.893629789352417
174 0.883613765239716
175 0.897435903549194
176 0.893629789352417
177 0.894030451774597
178 0.897035241127014
179 0.884615361690521
180 0.899839758872986
181 0.895232379436493
182 0.890625
183 0.886618614196777
184 0.88321316242218
185 0.892227590084076
186 0.896634638309479
187 0.900641024112701
188 0.887219548225403
189 0.895432710647583
190 0.896834909915924
191 0.894831717014313
192 0.891626596450806
193 0.891626596450806
194 0.897435903549194
195 0.895633041858673
196 0.891426265239716
197 0.902043282985687
198 0.896634638309479
199 0.900440692901611
200 0.900240361690521
201 0.897035241127014
202 0.896033644676208
203 0.896233975887299
204 0.888221144676208
205 0.894831717014313
206 0.893429458141327
207 0.893028855323792
208 0.890625
209 0.892227590084076
210 0.888421475887299
211 0.893229186534882
212 0.894230782985687
213 0.884615361690521
214 0.897235572338104
215 0.890625
216 0.888221144676208
217 0.889022409915924
218 0.890625
219 0.89102566242218
220 0.889823734760284
221 0.889423072338104
222 0.892027258872986
223 0.886017620563507
224 0.893629789352417
225 0.891426265239716
226 0.895232379436493
227 0.893629789352417
228 0.889022409915924
229 0.897636234760284
230 0.886217951774597
231 0.896434307098389
232 0.888822138309479
233 0.891426265239716
234 0.89102566242218
235 0.890625
236 0.896634638309479
237 0.891426265239716
238 0.894230782985687
239 0.890024065971375
240 0.890024065971375
241 0.892828524112701
242 0.894431114196777
243 0.888621807098389
244 0.89022433757782
245 0.891826927661896
246 0.895432710647583
247 0.896834909915924
248 0.895633041858673
249 0.884214758872986
250 0.889423072338104
251 0.887219548225403
252 0.89022433757782
253 0.892227590084076
254 0.887019217014313
255 0.895432710647583
256 0.891225934028625
257 0.892227590084076
258 0.897235572338104
259 0.892628192901611
260 0.893429458141327
261 0.895232379436493
262 0.892027258872986
263 0.893229186534882
264 0.89823716878891
265 0.896634638309479
266 0.894230782985687
267 0.896233975887299
268 0.888020813465118
269 0.887820541858673
270 0.893629789352417
271 0.892027258872986
272 0.893229186534882
273 0.888221144676208
274 0.889222741127014
275 0.892427861690521
276 0.896033644676208
277 0.896834909915924
278 0.891826927661896
279 0.895032048225403
280 0.891426265239716
281 0.892628192901611
282 0.888221144676208
283 0.894631385803223
284 0.889423072338104
};
\addplot [line width=2.5pt, color1, opacity=1.0]
table {%
1 0.113381413122018
2 0.22228899349769
3 0.296407575408618
4 0.517962058385213
5 0.472155456741651
6 0.678285241127014
7 0.320512823760509
8 0.5172275553147
9 0.492588135103385
10 0.462406526009242
11 0.303218478957812
12 0.545539528131485
13 0.210403308272362
14 0.291933755079905
15 0.503739312291145
16 0.258279919624329
};
\addplot [line width=2.5pt, red, opacity=1.0]
table {%
1 0.493990391492844
2 0.566907048225403
3 0.579026460647583
4 0.620192319154739
5 0.695913463830948
6 0.683193117380142
7 0.70272433757782
8 0.762820512056351
9 0.733273237943649
10 0.761217951774597
11 0.795172274112701
12 0.780749201774597
13 0.802984774112701
14 0.800480753183365
15 0.814002394676208
16 0.828225165605545
17 0.833733975887299
18 0.843249201774597
19 0.84805691242218
20 0.853665858507156
21 0.850761204957962
22 0.849959939718246
23 0.852864563465118
24 0.853866189718246
25 0.866286039352417
26 0.866486400365829
27 0.872095346450806
28 0.871895045042038
29 0.871794879436493
30 0.876201927661896
31 0.867287665605545
32 0.879407048225403
33 0.881310105323792
34 0.882011204957962
35 0.881810903549194
36 0.869691520929337
37 0.887219548225403
38 0.888822108507156
39 0.887219548225403
40 0.887319713830948
41 0.884715557098389
42 0.894330948591232
43 0.890625
44 0.889923900365829
45 0.890324503183365
46 0.886117786169052
47 0.889923870563507
48 0.889423072338104
49 0.895532876253128
50 0.897435873746872
51 0.898737996816635
52 0.890625
53 0.888521641492844
54 0.898337364196777
55 0.900340527296066
56 0.889623403549194
57 0.896434277296066
58 0.893529653549194
59 0.899839729070663
60 0.89473158121109
61 0.896033674478531
62 0.900240391492844
63 0.901342153549194
64 0.899539262056351
65 0.902644217014313
66 0.903746008872986
67 0.900040060281754
68 0.903946310281754
69 0.906951129436493
70 0.901542454957962
71 0.904747605323792
72 0.909254789352417
73 0.907852590084076
74 0.904146641492844
75 0.908353358507156
76 0.903846174478531
77 0.907051295042038
78 0.908052861690521
79 0.90234375
80 0.908553689718246
81 0.906850963830948
82 0.909254789352417
83 0.909955948591232
84 0.911558508872986
85 0.905649036169052
86 0.908653855323792
87 0.915164262056351
88 0.91015625
89 0.907051265239716
90 0.909655451774597
91 0.91055691242218
92 0.909054487943649
93 0.914463132619858
94 0.91035658121109
95 0.910556882619858
96 0.911458343267441
97 0.911959111690521
98 0.912860572338104
99 0.9140625
100 0.913261234760284
101 0.913261204957962
102 0.913962334394455
103 0.917467951774597
104 0.913161039352417
105 0.919871777296066
106 0.917267620563507
107 0.919471144676208
108 0.918970346450806
109 0.919571310281754
110 0.920172274112701
111 0.917367786169052
112 0.918970346450806
113 0.920572936534882
114 0.920272439718246
115 0.919871807098389
116 0.919971972703934
117 0.918870180845261
118 0.921975135803223
119 0.920272439718246
120 0.919971942901611
121 0.922475963830948
122 0.923277229070663
123 0.925480753183365
124 0.924278855323792
125 0.923778057098389
126 0.919170677661896
127 0.922976762056351
128 0.921975165605545
129 0.923477590084076
130 0.921975165605545
131 0.920472741127014
132 0.919471144676208
133 0.920472741127014
134 0.919571310281754
135 0.925380617380142
136 0.918970346450806
137 0.92167466878891
138 0.921374201774597
139 0.920072108507156
140 0.920973539352417
141 0.920472770929337
142 0.925580948591232
143 0.922175496816635
144 0.92167466878891
145 0.923377394676208
146 0.925380617380142
147 0.921374201774597
148 0.922275632619858
149 0.922976762056351
150 0.923477560281754
151 0.924379020929337
152 0.924178689718246
153 0.922375798225403
154 0.923778057098389
155 0.920973569154739
156 0.924579352140427
157 0.922676265239716
158 0.92207533121109
159 0.920673072338104
160 0.921374201774597
161 0.920172274112701
162 0.924479156732559
163 0.921975135803223
164 0.922676295042038
165 0.923076927661896
166 0.920773237943649
167 0.921274065971375
168 0.921975135803223
169 0.926081746816635
170 0.924479156732559
171 0.921774834394455
172 0.922776430845261
173 0.926181882619858
174 0.924178689718246
175 0.920572936534882
176 0.921875
177 0.924879819154739
178 0.925180286169052
179 0.923577725887299
180 0.923878222703934
181 0.923577725887299
182 0.92578125
183 0.924679487943649
184 0.926282048225403
185 0.922576129436493
186 0.92227566242218
187 0.926081717014313
188 0.924579322338104
189 0.925080150365829
190 0.924679487943649
191 0.922876626253128
192 0.923577725887299
193 0.922976762056351
194 0.924379020929337
195 0.927383810281754
196 0.924178689718246
197 0.924779653549194
198 0.921574532985687
199 0.925881415605545
200 0.921875
201 0.921875
202 0.922676295042038
203 0.923778027296066
204 0.924078524112701
205 0.92167466878891
206 0.924679487943649
207 0.924679487943649
208 0.926282048225403
209 0.925280451774597
210 0.926282048225403
211 0.923778027296066
212 0.928485572338104
213 0.922175496816635
214 0.925080120563507
215 0.926181882619858
216 0.924479156732559
217 0.926782846450806
218 0.925280421972275
219 0.926482379436493
220 0.924879819154739
221 0.92598158121109
222 0.925380617380142
223 0.925080120563507
224 0.926983177661896
225 0.926382213830948
226 0.926081717014313
227 0.923277229070663
228 0.924078524112701
229 0.927383840084076
230 0.92558091878891
231 0.926081717014313
232 0.924679487943649
233 0.92618191242218
234 0.926081746816635
235 0.926682680845261
236 0.92598158121109
237 0.927884608507156
238 0.926282048225403
239 0.926181882619858
240 0.927083343267441
241 0.926682710647583
242 0.928084939718246
243 0.927083313465118
244 0.925881415605545
245 0.925881415605545
246 0.928485572338104
247 0.926582545042038
248 0.927483975887299
249 0.928685903549194
250 0.927984774112701
251 0.927183479070663
252 0.926081746816635
253 0.927183508872986
254 0.927984774112701
255 0.926682680845261
256 0.928886204957962
257 0.927684307098389
258 0.927483975887299
259 0.925681084394455
260 0.928685903549194
261 0.928685903549194
262 0.927083343267441
263 0.92538058757782
264 0.925981551408768
265 0.929186701774597
266 0.929286867380142
267 0.928485572338104
268 0.927383840084076
269 0.927584141492844
270 0.928285241127014
271 0.926482379436493
272 0.926482349634171
273 0.927283644676208
274 0.928385406732559
275 0.927283644676208
276 0.927784472703934
277 0.927483975887299
278 0.930989593267441
279 0.930689096450806
280 0.927984774112701
281 0.930288463830948
282 0.927684307098389
283 0.928385406732559
284 0.93008816242218
};
\addplot [line width=2.5pt, color2, opacity=1.0]
table {%
2 0.449385672807693
4 0.568977038065592
6 0.631477018197378
8 0.667467951774597
10 0.69511216878891
12 0.725293795267741
14 0.753939628601074
16 0.759682138760885
18 0.770032048225403
20 0.784188012282054
22 0.791466355323792
24 0.811298072338104
26 0.818108975887299
28 0.835202991962433
30 0.83059561252594
32 0.831463694572449
34 0.841947118441264
36 0.849626084168752
38 0.85650372505188
40 0.858640491962433
42 0.857171495755514
44 0.860777239004771
46 0.863181094328562
48 0.872262279192607
50 0.87560095389684
52 0.876135150591532
54 0.879273513952891
56 0.880208333333333
58 0.878872871398926
60 0.885016024112701
62 0.883213142553965
64 0.880008002122243
66 0.881543815135956
68 0.883279919624329
70 0.879807690779368
72 0.880208333333333
74 0.881477038065592
76 0.875467419624329
78 0.888822118441264
80 0.883947650591532
82 0.890558222929637
84 0.886885662873586
86 0.885750532150269
88 0.881944437821706
90 0.884548624356588
92 0.891893704732259
94 0.881744126478831
96 0.885216335455577
98 0.881410260995229
100 0.891159176826477
102 0.881076375643412
104 0.885149578253428
106 0.88054221868515
108 0.88488248984019
110 0.884949247042338
112 0.886752128601074
114 0.887753744920095
116 0.886217951774597
118 0.883680542310079
120 0.889289538065592
122 0.873597741127014
124 0.888287921746572
126 0.892227570215861
128 0.885950843493144
130 0.885616997877757
132 0.891693373521169
134 0.885216335455577
136 0.883747339248657
138 0.885416646798452
140 0.88775372505188
142 0.889489849408468
144 0.886752128601074
146 0.889356315135956
148 0.886485040187836
150 0.885750532150269
152 0.887086013952891
154 0.875934819380442
156 0.886551817258199
158 0.879273494084676
160 0.887419879436493
162 0.887686967849731
164 0.885950843493144
166 0.886952459812164
168 0.885817309220632
170 0.89329594373703
172 0.888488233089447
174 0.88568377494812
176 0.887954076131185
178 0.894898494084676
180 0.884615381558736
182 0.885950863361359
184 0.891159176826477
186 0.88488248984019
188 0.892027258872986
190 0.883480230967204
192 0.892160793145498
194 0.891292750835419
196 0.888488252957662
198 0.886885682741801
200 0.890892088413239
202 0.879941244920095
204 0.88341345389684
206 0.889155964056651
208 0.882411857446035
210 0.895833333333333
212 0.890624980131785
214 0.887686947981516
216 0.885750532150269
218 0.888688584168752
220 0.889222760995229
222 0.884281535943349
224 0.887820502122243
226 0.887820521990458
228 0.889957269032796
230 0.887820521990458
232 0.890691777070363
234 0.895165582497915
236 0.888488252957662
238 0.890224357446035
240 0.888020833333333
242 0.882745722929637
244 0.891025642553965
246 0.890691777070363
248 0.888955652713776
250 0.891693393389384
252 0.891159196694692
254 0.893763363361359
256 0.894230763117472
258 0.892227550347646
260 0.896167198816935
262 0.895900110403697
264 0.892895301183065
266 0.898771365483602
268 0.88488248984019
270 0.892160793145498
272 0.893830120563507
274 0.894965271155039
276 0.888621807098389
278 0.897369106610616
280 0.887620190779368
282 0.893696586290995
284 0.894163986047109
};
\addplot [line width=2.5pt, green!50.1960784313725!black, opacity=1.0]
table {%
1 0.462473293145498
2 0.556290070215861
3 0.620058755079905
4 0.679220100243886
5 0.702991445859273
6 0.730835994084676
7 0.76008282105128
8 0.78104966878891
9 0.793068905671438
10 0.792200843493144
11 0.762753744920095
12 0.793469548225403
13 0.786124467849731
14 0.819911857446035
15 0.82792466878891
16 0.832932670911153
17 0.845152239004771
18 0.842881937821706
19 0.845753212769826
20 0.855502128601074
21 0.849225421746572
22 0.857572118441264
23 0.857371807098389
24 0.86144498984019
25 0.863715271155039
26 0.862847208976746
27 0.867254257202148
28 0.870058755079905
29 0.874999980131785
30 0.870592931906382
31 0.87252938747406
32 0.872996787230174
33 0.87706998984019
34 0.869724889596303
35 0.871594548225403
36 0.870592951774597
37 0.881610572338104
38 0.873197138309479
39 0.879807690779368
40 0.873798092206319
41 0.880608975887299
42 0.875801265239716
43 0.883146365483602
44 0.880208333333333
45 0.882946034272512
46 0.881677369276682
47 0.87974093357722
48 0.883747319380442
49 0.884882469971975
50 0.881209949652354
51 0.883547008037567
52 0.883680542310079
53 0.883880893389384
54 0.883079588413239
55 0.891626616319021
56 0.884081184864044
57 0.8828125
58 0.887553413709005
59 0.887686967849731
60 0.885817309220632
61 0.887686967849731
62 0.888688564300537
63 0.892628212769826
64 0.888020833333333
65 0.89082533121109
66 0.893763343493144
67 0.889489849408468
68 0.892361104488373
69 0.885884086290995
70 0.885149578253428
71 0.891426285107931
72 0.89536593357722
73 0.894163986047109
74 0.892160773277283
75 0.892027239004771
76 0.888020833333333
77 0.889957249164581
78 0.890825311342875
79 0.895566244920095
80 0.890090823173523
81 0.893229166666667
82 0.892561435699463
83 0.893028855323792
84 0.893830120563507
85 0.894564628601074
86 0.89122595389684
87 0.89309561252594
88 0.891760150591532
89 0.885683755079905
90 0.893629809220632
91 0.892160813013713
92 0.890958885351817
93 0.892895301183065
94 0.890958865483602
95 0.894831736882528
96 0.889890472094218
97 0.895432690779368
98 0.888421495755514
99 0.895633021990458
100 0.890892108281453
101 0.894364317258199
102 0.897369126478831
103 0.88795405626297
104 0.893629809220632
105 0.891826927661896
106 0.892494658629099
107 0.889890491962433
108 0.893563032150269
109 0.887086013952891
110 0.892561435699463
111 0.891225973765055
112 0.892828524112701
113 0.892962058385213
114 0.89042466878891
115 0.896434287230174
116 0.890424688657125
117 0.889489849408468
118 0.890224357446035
119 0.893963674704234
120 0.894831736882528
121 0.89596688747406
122 0.894631425539652
123 0.890157580375671
124 0.896100421746572
125 0.897235572338104
126 0.895833333333333
127 0.899572650591532
128 0.895299136638641
129 0.89596688747406
130 0.896701375643412
131 0.893563032150269
132 0.89783654610316
133 0.898704588413239
134 0.895566244920095
135 0.897970080375671
136 0.892962078253428
137 0.896968483924866
138 0.886017640431722
139 0.892494658629099
140 0.889423072338104
141 0.894631405671438
142 0.895699799060822
143 0.905916134516398
144 0.905381957689921
145 0.907518704732259
146 0.906049688657125
147 0.905715803305308
148 0.905715823173523
149 0.906984508037567
150 0.907852570215861
151 0.909722208976746
152 0.909989317258199
153 0.909321586290995
154 0.906450311342875
155 0.909054478009542
156 0.907852570215861
157 0.90604966878891
158 0.908520301183065
159 0.908453524112701
160 0.908520301183065
161 0.910323182741801
162 0.908653835455577
163 0.907051304976145
164 0.90685095389684
165 0.910456736882528
166 0.912526706854502
167 0.910924136638641
168 0.910189628601074
169 0.912059287230174
170 0.909989317258199
171 0.911925752957662
172 0.910590271155039
173 0.910323182741801
174 0.913261214892069
175 0.908587058385213
176 0.910924136638641
177 0.907785793145498
178 0.911658664544423
179 0.91199251015981
180 0.91159188747406
181 0.908587078253428
182 0.91346154610316
183 0.913060903549194
184 0.911992530028025
185 0.913862188657125
186 0.911057690779368
187 0.909588674704234
188 0.911858975887299
189 0.909655451774597
190 0.910657048225403
191 0.913928945859273
192 0.912860572338104
193 0.907919347286224
194 0.910990913709005
195 0.90892094373703
196 0.90831998984019
197 0.911124467849731
198 0.910990913709005
199 0.913261214892069
200 0.909989317258199
201 0.912860572338104
202 0.910456736882528
203 0.911124467849731
204 0.912727038065592
205 0.912593483924866
206 0.913194457689921
207 0.91426283121109
208 0.910924156506856
209 0.910924136638641
210 0.913327972094218
211 0.910657048225403
212 0.911258002122243
213 0.909054478009542
214 0.913862188657125
215 0.913928965727488
216 0.913194457689921
217 0.910924136638641
218 0.908987700939178
219 0.916466335455577
220 0.913728634516398
221 0.91446312268575
222 0.909054478009542
223 0.913194457689921
224 0.91426283121109
225 0.909922540187836
226 0.910456736882528
227 0.910857359568278
228 0.912192841370901
229 0.913261214892069
230 0.911925733089447
231 0.910723825295766
232 0.912526706854502
233 0.912727038065592
234 0.910723825295766
235 0.913194437821706
236 0.916132469971975
237 0.911858975887299
238 0.913595100243886
239 0.911725421746572
240 0.914863785107931
241 0.911525110403697
242 0.913261214892069
243 0.90892094373703
244 0.912326395511627
245 0.910657048225403
246 0.911658644676208
247 0.911258002122243
248 0.912192841370901
249 0.911258021990458
250 0.913795391718547
251 0.911792198816935
252 0.911525090535482
253 0.910657048225403
254 0.910056074460348
255 0.91079060236613
256 0.913595080375671
257 0.911792198816935
258 0.911658644676208
259 0.911658644676208
260 0.912793795267741
261 0.912393152713776
262 0.909588674704234
263 0.910389959812164
264 0.913127660751343
265 0.913595080375671
266 0.91119126478831
267 0.908253192901611
268 0.910523494084676
269 0.908653835455577
270 0.910590271155039
271 0.914797008037567
272 0.915197650591532
273 0.912326375643412
274 0.912459949652354
275 0.912459929784139
276 0.911925752957662
277 0.911591867605845
278 0.911525110403697
279 0.914463142553965
280 0.910323202610016
281 0.910657048225403
282 0.911124467849731
283 0.910189648469289
284 0.915665050347646
};
\addplot [line width=2.5pt, black, opacity=1.0]
table {%
1 0.458333333333333
2 0.500801275173823
3 0.59602028131485
4 0.638087610403697
5 0.685096164544423
6 0.732905983924866
7 0.749465823173523
8 0.73110310236613
9 0.775106847286224
10 0.791666666666667
11 0.79440438747406
12 0.802283644676208
13 0.827590823173523
14 0.82605501015981
15 0.826989849408468
16 0.827991465727488
17 0.825921456019084
18 0.827791134516398
19 0.838207801183065
20 0.839009086290995
21 0.850828011830648
22 0.842080652713776
23 0.856436967849731
24 0.863314648469289
25 0.866786857446035
26 0.872863272825877
27 0.878806094328562
28 0.878472248713175
29 0.880876044432322
30 0.881944437821706
31 0.869324266910553
32 0.865050752957662
33 0.883346676826477
34 0.883413473765055
35 0.88034188747406
36 0.890357891718547
37 0.888822118441264
38 0.89022437731425
39 0.884081184864044
40 0.885616978009542
41 0.886885682741801
42 0.889823714892069
43 0.884615401426951
44 0.881076375643412
45 0.890691757202148
46 0.895365913709005
47 0.896634618441264
48 0.901308755079905
49 0.899505873521169
50 0.900307158629099
51 0.896434307098389
52 0.898971696694692
53 0.901509086290995
54 0.903445502122243
55 0.902510662873586
56 0.901909708976746
57 0.899172008037567
58 0.903912921746572
59 0.901241997877757
60 0.900440692901611
61 0.903645833333333
62 0.909455120563507
63 0.905649026234945
64 0.904781003793081
65 0.904647429784139
66 0.905315160751343
67 0.902510702610016
68 0.903378744920095
69 0.904847741127014
70 0.905181626478831
71 0.903712610403697
72 0.903044859568278
73 0.905315160751343
74 0.904246807098389
75 0.906784196694692
76 0.907719016075134
77 0.90625
78 0.904513895511627
79 0.905048072338104
80 0.908453524112701
81 0.906116445859273
82 0.908052881558736
83 0.909922540187836
84 0.903979698816935
85 0.905715803305308
86 0.906650642553965
87 0.908119658629099
88 0.90564904610316
89 0.909254789352417
90 0.905114849408468
91 0.907785793145498
92 0.907184819380442
93 0.905982911586761
94 0.909121255079905
95 0.906784196694692
96 0.90685095389684
97 0.909855763117472
98 0.908987700939178
99 0.907986124356588
100 0.908253212769826
101 0.908520301183065
102 0.905448714892069
103 0.910256405671438
104 0.910389959812164
105 0.90932156642278
106 0.910590271155039
107 0.909655431906382
108 0.910189628601074
109 0.910389959812164
110 0.907251596450806
111 0.909321586290995
112 0.909722208976746
113 0.909254809220632
114 0.909989317258199
115 0.910256405671438
116 0.909054497877757
117 0.908787389596303
118 0.906784176826477
119 0.910323182741801
120 0.908987720807393
121 0.908453524112701
122 0.909788986047109
123 0.910256405671438
124 0.908453524112701
125 0.909455140431722
126 0.910189648469289
127 0.907585461934408
128 0.907251596450806
129 0.911258021990458
130 0.910122871398926
131 0.910590271155039
132 0.910323202610016
133 0.91079060236613
134 0.910389959812164
135 0.912860592206319
136 0.90892094373703
137 0.912593483924866
138 0.91159188747406
139 0.910723825295766
140 0.909655451774597
141 0.912727018197378
142 0.912393172581991
143 0.912459929784139
144 0.910189628601074
145 0.910456736882528
146 0.907919347286224
147 0.908520301183065
148 0.914930542310079
149 0.908520301183065
150 0.910924136638641
151 0.911191244920095
152 0.912459929784139
153 0.911992530028025
154 0.913795411586761
155 0.91159188747406
156 0.90912127494812
157 0.910122871398926
158 0.913595080375671
159 0.910790582497915
160 0.911525130271912
161 0.910456736882528
162 0.91099093357722
163 0.911725421746572
164 0.911124447981516
165 0.910456736882528
166 0.909188032150269
167 0.910523494084676
168 0.910056094328562
169 0.910723825295766
170 0.910122851530711
171 0.910323162873586
172 0.908119658629099
173 0.912727018197378
174 0.914596676826477
175 0.909922560056051
176 0.909788986047109
177 0.909388363361359
178 0.910590291023254
179 0.911792198816935
180 0.912393172581991
181 0.909388363361359
182 0.910924136638641
183 0.910456736882528
184 0.910256385803223
185 0.911124447981516
186 0.912192841370901
187 0.911792198816935
188 0.911124447981516
189 0.908854166666667
190 0.911458313465118
191 0.909254809220632
192 0.913060903549194
193 0.909054478009542
194 0.912393172581991
195 0.909588674704234
196 0.906917730967204
197 0.912192841370901
198 0.910323202610016
199 0.912927349408468
200 0.910723825295766
201 0.913795391718547
202 0.911191244920095
203 0.912126084168752
204 0.910456717014313
205 0.912526706854502
206 0.910924136638641
207 0.909588654836019
208 0.912593483924866
209 0.911725421746572
210 0.910256405671438
211 0.910523494084676
212 0.91139155626297
213 0.909188032150269
214 0.911725421746572
215 0.911792198816935
216 0.909722228844961
217 0.908453524112701
218 0.911124447981516
219 0.912192841370901
220 0.910590271155039
221 0.912994126478831
222 0.913194437821706
223 0.909722228844961
224 0.910056074460348
225 0.91159188747406
226 0.91079060236613
227 0.910456736882528
228 0.913595080375671
229 0.909722208976746
230 0.91079060236613
231 0.911458333333333
232 0.91139155626297
233 0.912927329540253
234 0.912459949652354
235 0.911324779192607
236 0.911858975887299
237 0.911725421746572
238 0.908052901426951
239 0.909521877765656
240 0.91159188747406
241 0.911258021990458
242 0.908854166666667
243 0.909655451774597
244 0.909655431906382
245 0.913528323173523
246 0.913728634516398
247 0.909054497877757
248 0.91179221868515
249 0.910790582497915
250 0.90892094373703
251 0.910189648469289
252 0.912126064300537
253 0.909855782985687
254 0.909922560056051
255 0.909789005915324
256 0.910990913709005
257 0.911591867605845
258 0.910456717014313
259 0.909788986047109
260 0.913461526234945
261 0.911792198816935
262 0.910857359568278
263 0.913194457689921
264 0.913127660751343
265 0.912994126478831
266 0.911725421746572
267 0.910990913709005
268 0.912126064300537
269 0.911925752957662
270 0.911258021990458
271 0.909722208976746
272 0.911658644676208
273 0.909788986047109
274 0.912326375643412
275 0.910924136638641
276 0.912326395511627
277 0.910122871398926
278 0.912192841370901
279 0.911258002122243
280 0.910122851530711
281 0.911258002122243
282 0.909855782985687
283 0.911925752957662
284 0.912727018197378
};
\addplot [line width=2.5pt, blue, opacity=1.0]
table {%
1 0.501201927661896
2 0.571714758872986
3 0.521233975887299
4 0.647035241127014
5 0.690304458141327
6 0.626402258872986
7 0.720152258872986
8 0.762820541858673
9 0.769831717014313
10 0.789863765239716
11 0.796875
12 0.785456717014313
13 0.821113765239716
14 0.828926265239716
15 0.837540090084076
16 0.821514427661896
17 0.832131385803223
18 0.84415066242218
19 0.847355782985687
20 0.849559307098389
21 0.848758041858673
22 0.844951927661896
23 0.852964758872986
24 0.856971144676208
25 0.844951927661896
26 0.860376596450806
27 0.861378192901611
28 0.858774065971375
29 0.878405451774597
30 0.870392620563507
31 0.877203524112701
32 0.864182710647583
33 0.877604186534882
34 0.871794879436493
35 0.873597741127014
36 0.853165090084076
37 0.877203524112701
38 0.89102566242218
39 0.882011234760284
40 0.875
41 0.876602590084076
42 0.876201927661896
43 0.889222741127014
44 0.880608975887299
45 0.877003192901611
46 0.869791686534882
47 0.891225934028625
48 0.882011234760284
49 0.879807710647583
50 0.886818885803223
51 0.888822138309479
52 0.886818885803223
53 0.88241183757782
54 0.887219548225403
55 0.892427861690521
56 0.88261216878891
57 0.888421475887299
58 0.899639427661896
59 0.886618614196777
60 0.881810903549194
61 0.89022433757782
62 0.899238765239716
63 0.89823716878891
64 0.895232379436493
65 0.901442289352417
66 0.908052861690521
67 0.894030451774597
68 0.897235572338104
69 0.902243614196777
70 0.900841355323792
71 0.902844548225403
72 0.900440692901611
73 0.900440692901611
74 0.900440692901611
75 0.90604966878891
76 0.905448734760284
77 0.901442289352417
78 0.899038434028625
79 0.900040090084076
80 0.90665066242218
81 0.896233975887299
82 0.903245210647583
83 0.90604966878891
84 0.908253192901611
85 0.908854186534882
86 0.909455120563507
87 0.903846144676208
88 0.907051265239716
89 0.900240361690521
90 0.911858975887299
91 0.90604966878891
92 0.907251596450806
93 0.913261234760284
94 0.909254789352417
95 0.911258041858673
96 0.907852590084076
97 0.909054458141327
98 0.910056114196777
99 0.905248403549194
100 0.904046475887299
101 0.905248403549194
102 0.904847741127014
103 0.905448734760284
104 0.90645033121109
105 0.909254789352417
106 0.902644217014313
107 0.905448734760284
108 0.911057710647583
109 0.90604966878891
110 0.910657048225403
111 0.908653855323792
112 0.90645033121109
113 0.901442289352417
114 0.902844548225403
115 0.914663434028625
116 0.904046475887299
117 0.904647409915924
118 0.907451927661896
119 0.908052861690521
120 0.908453524112701
121 0.909254789352417
122 0.902443885803223
123 0.908453524112701
124 0.900841355323792
125 0.895232379436493
126 0.89823716878891
127 0.907251596450806
128 0.889623403549194
129 0.895232379436493
130 0.899238765239716
131 0.899639427661896
132 0.896233975887299
133 0.89823716878891
134 0.904647409915924
135 0.90584933757782
136 0.896434307098389
137 0.901842951774597
138 0.901442289352417
139 0.896834909915924
140 0.897636234760284
141 0.900841355323792
142 0.902243614196777
143 0.914863765239716
144 0.911458313465118
145 0.912860572338104
146 0.915464758872986
147 0.91386216878891
148 0.911458313465118
149 0.915865361690521
150 0.914863765239716
151 0.916065692901611
152 0.914663434028625
153 0.913261234760284
154 0.919471144676208
155 0.916866958141327
156 0.915464758872986
157 0.917467951774597
158 0.91366183757782
159 0.915865361690521
160 0.911057710647583
161 0.916666686534882
162 0.917668282985687
163 0.912660241127014
164 0.912660241127014
165 0.918669879436493
166 0.913461565971375
167 0.916065692901611
168 0.909054458141327
169 0.919671475887299
170 0.915464758872986
171 0.915865361690521
172 0.915665090084076
173 0.916666686534882
174 0.915264427661896
175 0.907451927661896
176 0.915064096450806
177 0.912459909915924
178 0.915665090084076
179 0.912459909915924
180 0.91446316242218
181 0.90665066242218
182 0.911458313465118
183 0.9140625
184 0.910256385803223
185 0.909655451774597
186 0.911057710647583
187 0.909455120563507
188 0.912259638309479
189 0.91446316242218
190 0.909655451774597
191 0.912059307098389
192 0.912059307098389
193 0.91366183757782
194 0.910456717014313
195 0.912059307098389
196 0.907652258872986
197 0.909254789352417
198 0.903846144676208
199 0.90665066242218
200 0.90625
201 0.910256385803223
202 0.911458313465118
203 0.907852590084076
204 0.907451927661896
205 0.909254789352417
206 0.907652258872986
207 0.909455120563507
208 0.904447138309479
209 0.910857379436493
210 0.908653855323792
211 0.911258041858673
212 0.913060903549194
213 0.912459909915924
214 0.918068885803223
215 0.920272409915924
216 0.914663434028625
217 0.921073734760284
218 0.914863765239716
219 0.918469548225403
220 0.920072138309479
221 0.919070541858673
222 0.916666686534882
223 0.919070541858673
224 0.919471144676208
225 0.91446316242218
226 0.918068885803223
227 0.918669879436493
228 0.916065692901611
229 0.919871807098389
230 0.917467951774597
231 0.918669879436493
232 0.920272409915924
233 0.916666686534882
234 0.920673072338104
235 0.920072138309479
236 0.916866958141327
237 0.922676265239716
238 0.917668282985687
239 0.919270813465118
240 0.918669879436493
241 0.91426283121109
242 0.91426283121109
243 0.916866958141327
244 0.919671475887299
245 0.920873403549194
246 0.917668282985687
247 0.916065692901611
248 0.915264427661896
249 0.919871807098389
250 0.91386216878891
251 0.91366183757782
252 0.920472741127014
253 0.916466355323792
254 0.917467951774597
255 0.916065692901611
256 0.914663434028625
257 0.921274065971375
258 0.916666686534882
259 0.915264427661896
260 0.916866958141327
261 0.915264427661896
262 0.915464758872986
263 0.920272409915924
264 0.909655451774597
265 0.917267620563507
266 0.913261234760284
267 0.919671475887299
268 0.916266024112701
269 0.915064096450806
270 0.917067289352417
271 0.917668282985687
272 0.919270813465118
273 0.918469548225403
274 0.91426283121109
275 0.919070541858673
276 0.919471144676208
277 0.912660241127014
278 0.918068885803223
279 0.914863765239716
280 0.910056114196777
281 0.911258041858673
282 0.911057710647583
283 0.912860572338104
284 0.912459909915924
};
\end{axis}

\end{tikzpicture}

%% file: labpal/figure_data/performance_comparison_low_batch/plots_bs32/CIFAR-10_ResNet-20_test_accuracy.pgf
% This file was created by tikzplotlib v0.9.8.
\begin{tikzpicture}

\definecolor{color0}{rgb}{0.933333333333333,0.509803921568627,0.933333333333333}
\definecolor{color1}{rgb}{1,0.647058823529412,0}

\begin{axis}[
legend cell align={left},
legend style={
  fill opacity=0.8,
  draw opacity=1,
  text opacity=1,
  at={(0.91,0.5)},
  anchor=east,
  draw=white!80!black
},
minor xtick={},
minor ytick={},
tick align=outside,
tick pos=left,
title={test accuracy CIFAR-10 ResNet-20},
width=10.5cm,height=8cm,grid=major,major grid style={dotted},
reverse legend, legend cell align={left}, legend style={ fill opacity=0.8, draw opacity=1, text opacity=1, at={(0.9,0.22)}, anchor=east, draw=white!80!black},
x grid style={white!69.0196078431373!black},
xlabel={epoch with best val. acc.},
xmin=-0.055, xmax=0.055,
xtick style={color=black},
xtick={-0.06,-0.04,-0.02,0,0.02,0.04,0.06},
y grid style={white!69.0196078431373!black},
ylabel={test accuracy},
ymin=0.87, ymax=0.915,
xmajorticks=false,
ytick style={color=black},
ytick={0.85,0.86,0.87,0.88,0.89,0.90,0.91,0.92,0.93,0.94}
]
\addplot [draw=color0, fill=color0,mark=-, only marks, mark options={scale=3},line width=3pt]
table{%
x  y
0 0.888755341370901
0 -1
};
\addlegendentry{GOLSI : $c$: 0.99, $\eta$: 2.0, $\alpha$: 0.1, $\beta$: 0.4}
\addplot [draw=color1, fill=color1,mark=-, only marks, mark options={scale=3},line width=3pt]
table{%
x  y
0 0.886051019032796
0 -1
};
\addlegendentry{PAL : $\alpha$: 1.0, $\beta$: 0.4, $\mu$: 1.0}
\addplot [
  draw=green!50.1960784313725!black,
  fill=green!50.1960784313725!black,
mark=-, only marks, mark options={scale=3},line width=3pt
]
table{%
x  y
0 0.904013077418009
0 -1
};
\addlegendentry{SGD : $\lambda$: 0.1, $\beta$: 0.9}
\addplot [draw=black, fill=black,mark=-, only marks, mark options={scale=3},line width=3pt]
table{%
x  y
0 0.896834949652354
0 -1
};
\addlegendentry{SLS : c: 0.1, $\beta$: 0.9, $\mu$: 0.1}
\addplot [draw=red, fill=red,mark=-, only marks, mark options={scale=3},line width=3pt]
table{%
x  y
0 0.902694314718246
0 -1
};
\addlegendentry{LABPAL-NSGD : $\epsilon$: 4.0 (others same as above)}
\addplot [draw=blue, fill=blue,mark=-, only marks, mark options={scale=3},line width=3pt]
table{%
x  y
0 0.910306483507156
0 -1
};
\addlegendentry{LABPAL-SGD : $\epsilon$: 4.0 (others same as above)}
\end{axis}

\end{tikzpicture}

%% file: labpal/figure_data/performance_comparison_low_batch/plots_bs32/CIFAR-10_MobileNet-V2_test_accuracy.pgf
% This file was created by tikzplotlib v0.9.8.
\begin{tikzpicture}

\definecolor{color0}{rgb}{0.933333333333333,0.509803921568627,0.933333333333333}
\definecolor{color1}{rgb}{1,0.647058823529412,0}

\begin{axis}[
minor xtick={},
minor ytick={},
tick align=outside,
tick pos=left,
title={test accuracy CIFAR-10 MobileNet-V2},
width=10.5cm,height=8cm,grid=major,major grid style={dotted},
x grid style={white!69.0196078431373!black},
xlabel={epoch with best val. acc.},
xmin=-0.055, xmax=0.055,
xtick style={color=black},
xtick={-0.06,-0.04,-0.02,0,0.02,0.04,0.06},
y grid style={white!69.0196078431373!black},
ylabel={test accuracy},
ymin=0.89, ymax=0.94,
xmajorticks=false,
ytick style={color=black},
ytick={0.85,0.86,0.87,0.88,0.89,0.90,0.91,0.92,0.93,0.94}
]
\addplot [draw=color0, fill=color0,mark=-, only marks, mark options={scale=3},line width=3pt]
table{%
x  y
0 0.914196054140727
0 -1
};
\addplot [draw=red, fill=red,mark=-, only marks, mark options={scale=3},line width=3pt]
table{%
x  y
0 0.936698734760284
0 -1
};
\addplot [draw=blue, fill=blue,mark=-, only marks, mark options={scale=3},line width=3pt]
table{%
x  y
0 0.927534073591232
0 -1
};
\addplot [draw=color1, fill=color1,mark=-, only marks, mark options={scale=3},line width=3pt]
table{%
x  y
-0.06 0.908954322338104
0 -1
};
\addplot [
  draw=green!50.1960784313725!black,
  fill=green!50.1960784313725!black,
mark=-, only marks, mark options={scale=3},line width=3pt
]
table{%
x  y
0 0.928685883680979
0 -1
};
\addplot [draw=black, fill=black,mark=-, only marks, mark options={scale=3},line width=3pt]
table{%
x  y
0 0.919471164544423
0 -1
};
\end{axis}

\end{tikzpicture}

%% file: labpal/figure_data/performance_comparison_low_batch/plots_bs32/CIFAR-10_DenseNet-121_test_accuracy.pgf
% This file was created by tikzplotlib v0.9.8.
\begin{tikzpicture}

\definecolor{color0}{rgb}{0.933333333333333,0.509803921568627,0.933333333333333}
\definecolor{color1}{rgb}{1,0.647058823529412,0}

\begin{axis}[
minor xtick={},
minor ytick={},
tick align=outside,
tick pos=left,
title={test accuracy CIFAR-10 DenseNet-121},
width=10.5cm,height=8cm,grid=major,major grid style={dotted},
x grid style={white!69.0196078431373!black},
xlabel={epoch with best val. acc.},
xmin=-0.055, xmax=0.055,
xtick style={color=black},
xtick={-0.06,-0.04,-0.02,0,0.02,0.04,0.06},
y grid style={white!69.0196078431373!black},
ylabel={test accuracy},
ymin=0.89, ymax=0.94,
xmajorticks=false,
ytick style={color=black},
ytick={0.85,0.86,0.87,0.88,0.89,0.90,0.91,0.92,0.93,0.94}
]
\addplot [draw=color0, fill=color0,mark=-, only marks, mark options={scale=3},line width=3pt]
table{%
x  y
0 0.897736370563507
0 -1
};
\addplot [draw=red, fill=red,mark=-, only marks, mark options={scale=3},line width=3pt]
table{%
x  y
0 0.930038064718246
0 -1
};
\addplot [draw=blue, fill=blue,mark=-, only marks, mark options={scale=3},line width=3pt]
table{%
x  y
0 0.918669879436493
0 -1
};
\addplot [draw=color1, fill=color1,mark=-, only marks, mark options={scale=3},line width=3pt]
table{%
x  y
0 0.899973273277283
0 -1
};
\addplot [
  draw=green!50.1960784313725!black,
  fill=green!50.1960784313725!black,
mark=-, only marks, mark options={scale=3},line width=3pt
]
table{%
x  y
0 0.914229432741801
0 -1
};
\addplot [draw=black, fill=black, mark=-, only marks, mark options={scale=3},line width=3pt]
table{%
x  y
0 0.911525110403697
0 -1
};
\end{axis}

\end{tikzpicture}

%% file: labpal/batch_size_8.tex
\begin{figure}[h!]
	\tikzsetfigurename{labpal_opt_comparison_bs32}
	\centering
	\def\scale{0.4}
	\begin{tabular}{ c c c}
		\scalebox{\scale}{\input{"labpal/figure_data/performance_comparison_low_batch/plots_bs8/CIFAR-10_ResNet-20_learning_rate.pgf"}}&	
		\scalebox{\scale}{\input{"labpal/figure_data/performance_comparison_low_batch/plots_bs8/CIFAR-10_MobileNet-V2_learning_rate.pgf"}}&
		\scalebox{\scale}{\input{"labpal/figure_data/performance_comparison_low_batch/plots_bs8/CIFAR-10_DenseNet-121_learning_rate.pgf"}}\\
		\scalebox{\scale}{\input{"labpal/figure_data/performance_comparison_low_batch/plots_bs8/CIFAR-10_ResNet-20_training_loss.pgf"}}&
		\scalebox{\scale}{\input{"labpal/figure_data/performance_comparison_low_batch/plots_bs8/CIFAR-10_MobileNet-V2_training_loss.pgf"}}&
		\scalebox{\scale}{\input{"labpal/figure_data/performance_comparison_low_batch/plots_bs8/CIFAR-10_DenseNet-121_training_loss.pgf"}}\\	
		\scalebox{\scale}{\input{"labpal/figure_data/performance_comparison_low_batch/plots_bs8/CIFAR-10_ResNet-20_validation_accuracy.pgf"}}&	
		\scalebox{\scale}{\input{"labpal/figure_data/performance_comparison_low_batch/plots_bs8/CIFAR-10_MobileNet-V2_validation_accuracy.pgf"}}&	
		\scalebox{\scale}{\input{"labpal/figure_data/performance_comparison_low_batch/plots_bs8/CIFAR-10_DenseNet-121_validation_accuracy.pgf"}}\\
		\scalebox{\scale}{\input{"labpal/figure_data/performance_comparison_low_batch/plots_bs8/CIFAR-10_ResNet-20_test_accuracy.pgf"}}&
		\scalebox{\scale}{\input{"labpal/figure_data/performance_comparison_low_batch/plots_bs8/CIFAR-10_MobileNet-V2_test_accuracy.pgf"}}&
		\scalebox{\scale}{\input{"labpal/figure_data/performance_comparison_low_batch/plots_bs8/CIFAR-10_DenseNet-121_test_accuracy.pgf"}}
		
	\end{tabular}
	\caption{Performance comparison of several models on CIFAR-10 with \textbf{batch size 8}. \textbf{The same hyperparameters are used as for batch size 128} (see Figure \ref{labpal_fig_optimizer_comparison_cifar10}).  Due to the batch size adaptation to keep the noise scale on a similar level the LABPAL approaches still perform well. PAL and PLS fail to optimize at all. SGD still performs well. SLS still shows good performance, however if training longer this will not hold since the learning rate schedules degenerated. Note that training steps were increased by a factor of 16.}
	\label{labpal_fig_opt_comparison_bs8}
\end{figure}

%% file: labpal/figure_data/performance_comparison_low_batch/plots_bs8/CIFAR-10_DenseNet-121_training_loss.pgf
% This file was created by tikzplotlib v0.9.8.
\begin{tikzpicture}

\definecolor{color0}{rgb}{0.933333333333333,0.509803921568627,0.933333333333333}
\definecolor{color1}{rgb}{0.647058823529412,0.164705882352941,0.164705882352941}
\definecolor{color2}{rgb}{1,0.647058823529412,0}

\begin{axis}[
log basis y={10},
minor xtick={},
minor ytick={},
tick align=outside,
tick pos=left,
title={training loss CIFAR-10 DenseNet-121},
width=10.5cm,height=8cm,grid=major,major grid style={dotted},
x grid style={white!69.0196078431373!black},
xlabel={epochs},
xmin=0, xmax=298,
xtick style={color=black},
xtick={0,50,100,150,200,250},
y grid style={white!69.0196078431373!black},
ylabel={training loss},
ymin=0.001, ymax=5,
ymode=log,
ytick style={color=black},
ytick={1e-05,0.0001,0.001,0.01,0.1,1,10,100}
]
\addplot [line width=2.5pt, color0, opacity=1.0]
table {%
0 2.32029390335083
1 2.22087669372559
2 1.84740817546844
3 1.6413449048996
4 1.50631499290466
5 1.41749954223633
6 1.34963762760162
7 1.26021993160248
8 1.21403527259827
9 1.1708517074585
10 1.13493061065674
11 1.08899974822998
12 1.04026734828949
13 0.990843117237091
14 0.973604679107666
15 0.916489362716675
16 0.891168534755707
17 0.869129776954651
18 0.843946814537048
19 0.820201635360718
20 0.797384738922119
21 0.760483384132385
22 0.750025749206543
23 0.746259570121765
24 0.717544078826904
25 0.702578485012054
26 0.679867565631866
27 0.663985908031464
28 0.652943909168243
29 0.641680717468262
30 0.637288391590118
31 0.62588095664978
32 0.616382300853729
33 0.596770286560059
34 0.589268267154694
35 0.588674247264862
36 0.590311884880066
37 0.559290051460266
38 0.547646224498749
39 0.547776699066162
40 0.536327660083771
41 0.516798794269562
42 0.515848934650421
43 0.507753133773804
44 0.494721591472626
45 0.498693376779556
46 0.501806795597076
47 0.480020195245743
48 0.471119850873947
49 0.482883751392365
50 0.458865314722061
51 0.447191089391708
52 0.451348453760147
53 0.442609667778015
54 0.434025228023529
55 0.44811624288559
56 0.441348224878311
57 0.425707221031189
58 0.424085229635239
59 0.419919520616531
60 0.402106791734695
61 0.413045048713684
62 0.411106199026108
63 0.40675088763237
64 0.391625106334686
65 0.394551455974579
66 0.394960761070251
67 0.371852099895477
68 0.376625627279282
69 0.37752303481102
70 0.374732971191406
71 0.363618046045303
};
\addplot [line width=2.5pt, color1, opacity=1.0]
table {%
0 2.35180600484212
1 3.14379914601644
2 3.08916330337524
3 3.05402644475301
4 3.01054072380066
5 2.9610714117686
6 2.92661794026693
7 2.88093900680542
8 2.8712416489919
9 2.78582739830017
10 2.82310756047567
11 2.80485399564107
12 2.77237486839294
13 2.76050933202108
14 2.72600269317627
15 2.76347335179647
16 2.71111726760864
17 2.71343851089478
18 2.68060557047526
19 2.68604930241903
20 2.68314623832703
21 2.66602420806885
22 2.74379769961039
23 2.68589345614115
24 2.69380640983582
25 2.62683494885763
26 2.59397657712301
27 2.65929587682088
28 2.63076035181681
29 2.66809264818827
30 2.60684951146444
31 2.65456946690877
32 2.62377111117045
33 2.59990739822388
34 2.59337504704793
35 2.64951539039612
36 2.62162446975708
37 2.56830549240112
38 2.59652312596639
39 2.60436813036601
40 2.56967663764954
41 2.59933360417684
42 2.56502532958984
43 2.5687108039856
44 2.59556158383687
45 2.58576520284017
46 2.56521193186442
47 2.57945243517558
48 2.57669226328532
49 2.57327723503113
50 2.55905532836914
51 2.5831507841746
52 2.57698551813761
53 2.54828667640686
54 2.61122059822083
55 2.56740800539652
56 2.57511385281881
57 2.58308323224386
58 2.54243167241414
59 2.57317105929057
60 2.57255880037944
61 2.52787605921427
62 2.60168862342834
63 2.532870054245
64 2.54153029123942
65 2.52817956606547
66 2.53267288208008
67 2.56527169545492
68 2.53231620788574
69 2.56632280349731
70 2.53400945663452
71 2.52646581331889
72 2.54170966148376
73 2.50894371668498
74 2.51763065656026
75 2.52059626579285
76 2.5623950958252
77 2.50773437817891
78 2.48315223058065
79 2.52590680122375
80 2.48219029108683
81 2.5080767472585
82 2.49456866582235
83 2.48386208216349
84 2.53477907180786
85 2.50324440002441
86 2.52453438440959
87 2.47254101435343
88 2.46147465705872
89 2.52145799001058
90 2.46312650044759
91 2.50312868754069
92 2.51008685429891
93 2.49149258931478
94 2.41794578234355
95 2.46011527379354
96 2.45294570922852
97 2.49935555458069
98 2.48364122708639
99 2.47616163889567
100 2.46544869740804
101 2.51313837369283
102 2.47806644439697
103 2.42890016237895
104 2.44274226824443
105 2.45336540540059
106 2.47961250940959
107 2.45609347025553
108 2.45884323120117
109 2.49012915293376
110 2.49828187624613
111 2.43287984530131
112 2.42305668195089
113 2.46120810508728
114 2.46494587262472
115 2.43177914619446
116 2.46849910418193
117 2.43828511238098
118 2.45112363497416
119 2.41607260704041
120 2.41550000508626
121 2.45220947265625
122 2.44303949673971
123 2.42786924044291
124 2.42625586191813
125 2.44704659779867
126 2.43060779571533
127 2.43143637975057
128 2.3975252310435
129 2.40266680717468
130 2.36775294939677
131 2.41940569877625
132 2.3963208993276
133 2.44509863853455
134 2.41857043902079
135 2.38205091158549
136 2.39687967300415
137 2.39906120300293
138 2.40908336639404
139 2.37803339958191
140 2.40911714235942
141 2.39197587966919
142 2.45926467577616
143 2.39591375986735
144 2.39527058601379
145 2.39713803927104
146 2.42283304532369
147 2.38546403249105
148 2.38755718866984
149 2.42463874816895
150 2.38666168848673
151 2.33824793497721
152 2.37540443738302
153 2.38713606198629
154 2.38651045163473
155 2.39069286982218
156 2.39112416903178
157 2.3739398320516
158 2.4432434240977
159 2.3810932636261
160 2.40247456232707
161 2.3893609046936
162 2.36787470181783
163 2.33121109008789
164 2.3554691473643
165 2.3504467010498
166 2.38010064760844
167 2.41524457931519
168 2.3123893737793
169 2.37310194969177
170 2.3718516031901
171 2.31420675913493
172 2.36497108141581
173 2.31905579566956
174 2.35774683952332
175 2.37546865145365
};
\addplot [line width=2.5pt, red, opacity=1.0]
table {%
0 2.36756217479706
1 1.77909904718399
2 1.47512096166611
3 1.33695882558823
4 1.27773779630661
5 1.13892596960068
6 1.04802614450455
7 1.01020646095276
8 0.912501782178879
9 0.863252252340317
10 0.832915514707565
11 0.765883803367615
12 0.745684117078781
13 0.721222430467606
14 0.663951724767685
15 0.639398902654648
16 0.600749343633652
17 0.555666834115982
18 0.533076792955399
19 0.498760491609573
20 0.472794130444527
21 0.461219325661659
22 0.431328967213631
23 0.40871062874794
24 0.388277471065521
25 0.367937475442886
26 0.349426925182343
27 0.339826449751854
28 0.325334101915359
29 0.315374255180359
30 0.302218481898308
31 0.291919752955437
32 0.277768328785896
33 0.273986324667931
34 0.257766515016556
35 0.252836614847183
36 0.244944110512733
37 0.234000630676746
38 0.228621460497379
39 0.217458717525005
40 0.212022736668587
41 0.205703109502792
42 0.20218738168478
43 0.194375596940517
44 0.195778355002403
45 0.190281882882118
46 0.187613435089588
47 0.18518240749836
48 0.183473169803619
49 0.177688226103783
50 0.174533843994141
51 0.171019025146961
52 0.166003786027431
53 0.160033762454987
54 0.157048530876637
55 0.147644102573395
56 0.143190756440163
57 0.138453617691994
58 0.132637657225132
59 0.130497641861439
60 0.128486640751362
61 0.120945230126381
62 0.11610072106123
63 0.112232245504856
64 0.109035938978195
65 0.106326781213284
66 0.0997803471982479
67 0.0978445746004581
68 0.0945041067898273
69 0.093835785984993
70 0.0897338539361954
71 0.0874436870217323
72 0.0839112289249897
73 0.0811057612299919
74 0.0805945284664631
75 0.0743167959153652
76 0.0730773881077766
77 0.0705637261271477
78 0.0697639919817448
79 0.0667486302554607
80 0.0654974319040775
81 0.0632410626858473
82 0.0621865801513195
83 0.060638315975666
84 0.0558610614389181
85 0.0530126821249723
86 0.0524138100445271
87 0.048540061339736
88 0.0465740468353033
89 0.0458089094609022
90 0.0436054170131683
91 0.041276965290308
92 0.0415099747478962
93 0.0380475893616676
94 0.0360720977187157
95 0.0352303143590689
96 0.0345207769423723
97 0.033826369792223
98 0.0295391147956252
99 0.0285131651908159
100 0.0306579265743494
101 0.0265544224530458
102 0.0248086694628
103 0.0237640906125307
104 0.0218867436051369
105 0.0235529979690909
106 0.0224270075559616
107 0.0229069916531444
108 0.0217050351202488
109 0.0214551649987698
110 0.0228666113689542
111 0.0227506654337049
112 0.0219325050711632
113 0.0190663672983646
114 0.0232910653576255
115 0.0211427924223244
116 0.0207564081065357
117 0.0185817242600024
118 0.0189776392653584
119 0.0191646716557443
120 0.0191133390180767
121 0.0187286692671478
122 0.0174781372770667
123 0.0169232632033527
124 0.0151667739264667
125 0.0156270554289222
126 0.0140890430193394
127 0.0157791078090668
128 0.0174172418192029
129 0.0153237720951438
130 0.013995046261698
131 0.0140097308903933
132 0.0147329745814204
133 0.0142815751023591
134 0.0142990737222135
135 0.0132699618116021
136 0.0145658357068896
137 0.0137768865097314
138 0.013734984677285
139 0.0141921797767282
140 0.0127256102859974
141 0.0123111959546804
142 0.0118645345792174
143 0.0102639538235962
144 0.00913857575505972
145 0.0097904596477747
146 0.0082017257809639
147 0.00933877471834421
148 0.00705517758615315
149 0.00828637555241585
150 0.0075127212330699
151 0.00746782100759447
152 0.00767735810950398
153 0.00668322877027094
154 0.00786927714943886
155 0.00834663072600961
156 0.00853964779525995
157 0.00851914123632014
158 0.00752747058868408
159 0.00747982668690383
160 0.00801920657977462
161 0.00754617294296622
162 0.00814796099439263
163 0.00785631570033729
164 0.0082954321987927
165 0.00685588968917727
166 0.00794239668175578
167 0.00648275390267372
168 0.00676650274544954
169 0.00615471438504755
170 0.00682158139534295
171 0.00582376285456121
172 0.00558310700580478
173 0.00573535147123039
174 0.00564677501097322
175 0.00531166966538876
176 0.00488430703990161
177 0.00530954287387431
178 0.00471134064719081
179 0.00447113951668143
180 0.00465441145934165
181 0.00477315881289542
182 0.00433505862019956
183 0.00469798303674906
184 0.0045956252142787
185 0.00395588064566255
186 0.00393134099431336
187 0.00466127949766815
188 0.00430485745891929
189 0.00366107781883329
190 0.00311157701071352
191 0.00404704629909247
192 0.00408614228945225
193 0.00421657075639814
194 0.00364901951979846
195 0.00309064355678856
196 0.00226877629756927
197 0.00329814886208624
198 0.00287372537422925
199 0.00364659435581416
200 0.00269873498473316
201 0.0021644844673574
202 0.00291581870988011
203 0.00316093617584556
204 0.00266042468138039
205 0.00267748872283846
206 0.00357624201569706
207 0.00273207470308989
208 0.00174113002140075
209 0.00265023612882942
210 0.002396329655312
211 0.0032903179526329
212 0.00241800310323015
213 0.00263336603529751
214 0.00235624320339411
215 0.00205208861734718
216 0.0022289605694823
217 0.00212521763751283
218 0.00176533454214223
219 0.00220408098539338
220 0.00231533270562068
221 0.00144415436079726
222 0.00196504394989461
223 0.00230690173339099
224 0.00171907362528145
225 0.00150413997471333
226 0.00143635226413608
227 0.00182243488961831
228 0.00290005025453866
229 0.00237695476971567
230 0.00163827362121083
231 0.00188271456863731
232 0.00174136727582663
233 0.00279121560743079
234 0.00183932826621458
235 0.00192318606423214
236 0.0018314523040317
237 0.00131266325479373
238 0.0025062903878279
239 0.0030254201265052
240 0.00189953512744978
241 0.00141858786810189
242 0.00174859224352986
243 0.00155664357589558
244 0.00141179678030312
245 0.0020127979805693
246 0.00183576089330018
247 0.00210196373518556
248 0.00141084566712379
249 0.00187631230801344
250 0.0019865480135195
251 0.00109389793942682
252 0.00127630823408253
253 0.00172887608641759
254 0.00140759965870529
255 0.00189345743274316
256 0.00203772983513772
257 0.001616001711227
258 0.00150386977475137
259 0.00115456487401389
260 0.00099641150154639
261 0.00177803839324042
262 0.00147289133747108
263 0.00111747591290623
264 0.00163954350864515
265 0.00195575517136604
266 0.00148122163955122
267 0.00132868211949244
268 0.0017059615929611
269 0.00136703148018569
270 0.00175914203282446
271 0.00152943070861511
272 0.00191737210843712
273 0.00111841404577717
274 0.0010576804052107
275 0.00114855641731992
276 0.00188724300824106
277 0.00166335859103128
278 0.00133909523719922
279 0.00174447597237304
280 0.0014521618722938
281 0.0018608087557368
282 0.00127711793174967
283 0.00155940180411562
284 0.00174172956030816
};
\addplot [line width=2.5pt, color2, opacity=1.0]
table {%
0 2.35180608431498
2 2.72525922457377
4 2.75334962209066
6 2.76703945795695
8 2.76961596806844
10 2.77130643526713
12 2.76731030146281
14 2.76600686709086
16 2.77211022377014
18 2.76941259702047
20 2.77235674858093
22 2.76745367050171
24 2.76889665921529
26 2.76924117406209
28 2.77046696345011
30 2.77433276176453
32 2.76842093467712
34 2.76774414380391
36 2.76605828603109
38 2.77317690849304
40 2.76854507128398
42 2.77139584223429
44 2.76900339126587
46 2.76847060521444
48 2.76802198092143
50 2.77062273025513
52 2.76603182156881
54 2.76698422431946
56 2.7702542146047
58 2.76735289891561
60 2.76488717397054
62 2.77018531163534
64 2.76941084861755
66 2.77144360542297
68 2.77099370956421
70 2.76788298288981
72 2.77001476287842
74 2.76978039741516
76 2.77064228057861
78 2.76867938041687
80 2.76889133453369
82 2.77272264162699
84 2.7666076819102
86 2.77045877774556
88 2.76835783322652
90 2.76740948359172
92 2.76623384157817
94 2.76485761006673
96 2.7688353061676
98 2.77075529098511
100 2.77358913421631
102 2.76804455121358
104 2.76997224489848
106 2.77000594139099
108 2.7721479733785
110 2.76844882965088
112 2.76786462465922
114 2.76912871996562
116 2.76619251569112
118 2.7689642906189
120 2.76628073056539
122 2.76724863052368
124 2.7689114411672
126 2.77106865247091
128 2.77091240882874
130 2.76984794934591
132 2.76694464683533
134 2.77588208516439
136 2.76842395464579
138 2.77048341433207
140 2.77491966883341
142 2.77164673805237
144 2.77146506309509
146 2.77104314168294
148 2.76775979995728
150 2.7702751159668
152 2.77168027559916
154 2.76798764864604
156 2.76943651835124
158 2.77152514457703
160 2.77104115486145
162 2.76857900619507
164 2.77201867103577
166 2.76872849464417
168 2.76793217658997
170 2.77062296867371
172 2.7717661857605
174 2.76993083953857
176 2.76758726437887
178 2.76973223686218
180 2.77047610282898
182 2.7691904703776
184 2.76979724566142
186 2.76978810628255
188 2.77057377497355
190 2.77130301793416
192 2.77062646547953
194 2.76914230982463
196 2.76867485046387
198 2.7748130162557
200 2.7655618985494
202 2.76992233594259
204 2.76913992563883
206 2.77149415016174
208 2.7692445119222
210 2.767689148585
212 2.7692277431488
214 2.77186886469523
216 2.76641893386841
218 2.76574452718099
220 2.77017958958944
222 2.76665925979614
224 2.76755428314209
226 2.76776830355326
228 2.77044987678528
230 2.77070609728495
232 2.77146116892497
234 2.77125422159831
236 2.77060254414876
238 2.76949612299601
240 2.76845526695251
242 2.77335731188456
244 2.77200889587402
246 2.76850994427999
248 2.76345483462016
250 2.76680938402812
252 2.76930999755859
254 2.77027448018392
256 2.76626539230347
258 2.77008724212646
260 2.77006268501282
262 2.76941935221354
264 2.77176833152771
266 2.7705557346344
268 2.76956089337667
270 2.76890190442403
272 2.76935863494873
274 2.77241007486979
276 2.7668133576711
278 2.76745255788167
280 2.76754784584045
282 2.76832048098246
284 2.7698392868042
};
\addplot [line width=2.5pt, green!50.1960784313725!black, opacity=1.0]
table {%
0 2.35180608431498
1 2.01174406210581
2 1.61356107393901
3 1.38405561447144
4 1.23473195234934
5 1.13847621281942
6 1.04904802640279
7 0.996365527311961
8 0.936375916004181
9 0.946788390477498
10 0.942063291867574
11 0.910185635089874
12 0.802714904149373
13 0.84507167339325
14 0.760688225428263
15 0.72413698832194
16 0.719552636146545
17 0.633876442909241
18 0.627672791481018
19 0.613428354263306
20 0.62579333782196
21 0.580629527568817
22 0.593408783276876
23 0.519863128662109
24 0.503638297319412
25 0.49144579966863
26 0.467456042766571
27 0.461527794599533
28 0.460991789897283
29 0.439393033583959
30 0.431223561366399
31 0.410083204507828
32 0.394335259993871
33 0.381900737682978
34 0.393701553344727
35 0.374424715836843
36 0.3604809542497
37 0.342837065458298
38 0.328810503085454
39 0.319543431202571
40 0.315882960955302
41 0.304654439290365
42 0.676127900679906
43 0.54068598151207
44 0.46358730395635
45 0.437227219343185
46 0.39290518561999
47 0.372014820575714
48 0.600247442722321
49 0.813105036815008
50 0.70852675040563
51 0.640128880739212
52 0.5905615290006
53 0.554811278978984
54 0.518490602572759
55 0.532656888167063
56 0.512117475271225
57 0.451132526000341
58 0.421095818281174
59 0.412751793861389
60 0.397845774888992
61 0.384525860349337
62 0.375991210341454
63 0.394905542333921
64 0.353473350405693
65 0.336235374212265
66 0.33136914173762
67 0.324738706151644
68 0.310199022293091
69 0.302629292011261
70 0.292200038830439
71 0.291528845826785
72 0.281488870580991
73 0.276544799407323
74 0.264292935530345
75 0.254439488053322
76 0.253412688771884
77 0.244474664330482
78 0.243208378553391
79 0.234716082612673
80 0.236717785398165
81 0.233553007245064
82 0.235133727391561
83 0.230431035161018
84 0.222327470779419
85 0.231957296530406
86 0.21115214129289
87 0.213718091448148
88 0.214590067664782
89 0.205918384095033
90 0.204311033089956
91 0.202886740366618
92 0.191175113121668
93 0.186684042215347
94 0.186111646393935
95 0.200679115951061
96 0.184741670886676
97 0.183884300291538
98 0.178570014735063
99 0.168781605859598
100 0.163596111039321
101 0.165564489861329
102 0.164718555907408
103 0.156908668577671
104 0.158978032569091
105 0.146804504096508
106 0.144444763660431
107 0.139879328509172
108 0.164725663761298
109 0.145148105919361
110 0.139644506076972
111 0.134026534855366
112 0.138585781057676
113 0.13979971408844
114 0.128732514878114
115 0.131514757871628
116 0.123837672173977
117 0.119098636011283
118 0.118678785860538
119 0.115668378770351
120 0.108855380366246
121 0.10946732511123
122 0.108858671039343
123 0.215380801508824
124 0.141837817927202
125 0.122522328048944
126 0.117175674686829
127 0.115807072569927
128 0.104527613768975
129 0.106903427590926
130 0.106141850352287
131 0.106319935371478
132 0.0988738561669985
133 0.0967007217307885
134 0.0964573187132676
135 0.0977736686666806
136 0.0985572449862957
137 0.105841040611267
138 0.10092156380415
139 0.0937126961847146
140 0.0915130277474721
141 0.0918896893660228
142 0.0909050926566124
143 0.0664211474359035
144 0.0479090288281441
145 0.0433379200597604
146 0.0391482897102833
147 0.0382336191833019
148 0.0361297614872456
149 0.0352350367854039
150 0.0331086435665687
151 0.0311969437946876
152 0.0303819328546524
153 0.030241282346348
154 0.0291591423253218
155 0.0283468244597316
156 0.0285969286536177
157 0.0281656207516789
158 0.0274212571481864
159 0.025458269752562
160 0.0256740736464659
161 0.0245245493327578
162 0.0260783253858487
163 0.0260925553739071
164 0.0236361135418216
165 0.023669441541036
166 0.0219106528287133
167 0.0224390476942062
168 0.0221909265965223
169 0.0223250289758046
170 0.0223014528552691
171 0.02117710839957
172 0.0214871518934766
173 0.0226901589582364
174 0.021164496584485
175 0.0217098553354541
176 0.0200706461134056
177 0.0208557948159675
178 0.019723839747409
179 0.0204668558823566
180 0.0192376403138041
181 0.0196286242765685
182 0.0183374909684062
183 0.0188067884494861
184 0.0194641787869235
185 0.0178240551613271
186 0.0180749166756868
187 0.0177869146379332
188 0.0174063734399776
189 0.0178385080459217
190 0.0167636801488698
191 0.0175132267177105
192 0.0156777240335941
193 0.0173381894516448
194 0.0163319689842562
195 0.0163429202511907
196 0.0161872315220535
197 0.0157964491906265
198 0.0160667446131508
199 0.0162271652370691
200 0.0156333834553758
201 0.0158236290638645
202 0.0159857631660998
203 0.0149750035877029
204 0.0152163111294309
205 0.015192883554846
206 0.0154825667850673
207 0.0147435651160777
208 0.0148780738624434
209 0.0156053933314979
210 0.0152598489075899
211 0.0143577777780592
212 0.0138538052948813
213 0.0137552934077879
214 0.0138613906068106
215 0.0131627412823339
216 0.0134540603806575
217 0.0133874132297933
218 0.0126716308295727
219 0.0132477777078748
220 0.0133370632926623
221 0.012951971963048
222 0.0127517233292262
223 0.012685000586013
224 0.0132457014794151
225 0.0128161837346852
226 0.0125667951069772
227 0.0122618450162311
228 0.0124211649720867
229 0.0122379772365093
230 0.0121518229910483
231 0.0119676330747704
232 0.0127793204349776
233 0.0125678977929056
234 0.0137382072086136
235 0.0127262148695687
236 0.0120731275528669
237 0.0129051275240878
238 0.0119775521258513
239 0.011778077421089
240 0.0121553091642757
241 0.0126691575472554
242 0.0125411554860572
243 0.0118333803645025
244 0.0125036329651872
245 0.01204259119307
246 0.0118468099584182
247 0.0123061315777401
248 0.0125213540159166
249 0.0121231749653816
250 0.0123627805151045
251 0.0119622270576656
252 0.011299277190119
253 0.0120868732531865
254 0.0123135355922083
255 0.0128281405195594
256 0.0118311326950788
257 0.0119783702927331
258 0.0124095506034791
259 0.0112930218068262
260 0.0112860185715059
261 0.0119313611648977
262 0.0114256461771826
263 0.01230655113856
264 0.0114727026472489
265 0.0119837981183082
266 0.011356809331725
267 0.0120368528490265
268 0.0123837591769795
269 0.0119339961869021
270 0.0111364672581355
271 0.011146028758958
272 0.0114189335145056
273 0.0115430907656749
274 0.012129426933825
275 0.0110065963429709
276 0.0110875399162372
277 0.0121711250394583
278 0.0119503790823122
279 0.0117650160876413
280 0.0114246193940441
281 0.0118088542173306
282 0.0114514545227091
283 0.0124652947609623
284 0.0107167057382564
};
\addplot [line width=2.5pt, black, opacity=1.0]
table {%
0 2.35180608431498
1 1.8691846927007
2 1.56177731355031
3 1.33559223016103
4 1.17114333311717
5 1.04439258575439
6 0.956198294957479
7 0.891874492168427
8 0.859271804491679
9 0.834835787614187
10 0.787875334421794
11 0.759759982426961
12 0.687613010406494
13 0.653897285461426
14 0.626794576644897
15 0.582857708136241
16 0.530479192733765
17 0.472110003232956
18 0.470487833023071
19 0.44288315375646
20 0.434924443562826
21 0.410404761632284
22 0.382390558719635
23 0.358091255029043
24 0.347370564937592
25 0.329325169324875
26 0.311886529127757
27 0.305761903524399
28 0.28262135386467
29 0.285126874844233
30 0.261642396450043
31 0.25245334704717
32 0.236510788400968
33 0.227297951777776
34 0.214436714847883
35 0.204151704907417
36 0.192059874534607
37 0.187156518300374
38 0.181323776642481
39 0.174402852853139
40 0.170414144794146
41 0.15439228216807
42 0.153546154499054
43 0.153672258059184
44 0.148514782389005
45 0.142364809910456
46 0.138451342781385
47 0.130136410395304
48 0.120091003676256
49 0.120542081693808
50 0.116912022233009
51 0.117027218143145
52 0.115801073610783
53 0.106855973601341
54 0.10547503332297
55 0.10048675040404
56 0.100431648393472
57 0.0956327666838964
58 0.0960001200437546
59 0.0949204663435618
60 0.094520536561807
61 0.0941499471664429
62 0.0920211275418599
63 0.0972691848874092
64 0.0958826715747515
65 0.0966385106245677
66 0.0942780027786891
67 0.0938491473595301
68 0.0958142032225927
69 0.0923845693469048
70 0.0950541769464811
71 0.0919338588913282
72 0.0956561093529065
73 0.0987043256560961
74 0.095600351691246
75 0.091866301993529
76 0.0935849919915199
77 0.0965552975734075
78 0.0935945510864258
79 0.0932044436534246
80 0.0965274199843407
81 0.100528252621492
82 0.0968123748898506
83 0.0952958116928736
84 0.0965692823131879
85 0.0944064036011696
86 0.0933989658951759
87 0.093577578663826
88 0.0944874485333761
89 0.0957335531711578
90 0.0955525661508242
91 0.0937970727682114
92 0.0963778868317604
93 0.0941156248251597
94 0.0946870222687721
95 0.0962034339706103
96 0.0947102705637614
97 0.0972176045179367
98 0.0947794293363889
99 0.0959986224770546
100 0.0937589233120282
101 0.0960426926612854
102 0.0920295665661494
103 0.0953596358497937
104 0.0977961172660192
105 0.0949604138731956
106 0.0937181462844213
107 0.0921362191438675
108 0.0964589367310206
109 0.0954421386122704
110 0.0960952465732892
111 0.0955826739470164
112 0.0964522461096446
113 0.0943304374814034
114 0.0968354841073354
115 0.0936938896775246
116 0.0958563064535459
117 0.0936615616083145
118 0.0956735288103421
119 0.0941105112433434
120 0.0942979107300441
121 0.094616820414861
122 0.0955236032605171
123 0.0967011724909147
124 0.0935294851660728
125 0.0971979697545369
126 0.0935295398036639
127 0.0976003433267276
128 0.0936930254101753
129 0.0973002761602402
130 0.0971403221289317
131 0.0940840939680735
132 0.0929996346433957
133 0.0936199973026911
134 0.0950298259655635
135 0.0928112640976906
136 0.0952599520484606
137 0.0942994380990664
138 0.0952581539750099
139 0.0953299626708031
140 0.0944283132751783
141 0.0965280781189601
142 0.0963508859276772
143 0.0941596428553263
144 0.0953806067506472
145 0.0959427654743195
146 0.0949604784448942
147 0.0964830840627352
148 0.0958764354387919
149 0.0972405423720678
150 0.0949310089151065
151 0.0951963116725286
152 0.0944107274214427
153 0.0947092697024345
154 0.0958159143726031
155 0.0952033624053001
156 0.09757050126791
157 0.0950393254558245
158 0.0938485289613406
159 0.0951922858754794
160 0.0952045222123464
161 0.0947701210776965
162 0.0953875531752904
163 0.0967509324351947
164 0.0956929971774419
165 0.0942496061325073
166 0.0932095895210902
167 0.0948038573066394
168 0.0941926017403603
169 0.0920972675085068
170 0.0952046637733777
171 0.0924566661318143
172 0.0945890222986539
173 0.093785119553407
174 0.0959529429674149
175 0.0961743916074435
176 0.0941059043010076
177 0.0930221701661746
178 0.0965433567762375
179 0.0943160603443782
180 0.0958176453908284
181 0.0938844904303551
182 0.0956430360674858
183 0.0952998921275139
184 0.0946658477187157
185 0.0942128499348958
186 0.0955079744259516
187 0.0954867030183474
188 0.0948441997170448
189 0.0979505280653636
190 0.0933894341190656
191 0.0939865236481031
192 0.0958986232678096
193 0.0934503972530365
194 0.0949579775333405
195 0.0954596425096194
196 0.0967240581909816
197 0.0933628901839256
198 0.0953317855795224
199 0.0965433791279793
200 0.096393716832002
201 0.0952459722757339
202 0.0947314277291298
203 0.0955084909995397
204 0.0918333182732264
205 0.0968308175603549
206 0.0964128002524376
207 0.0916046599547068
208 0.0928833782672882
209 0.0939381445447604
210 0.0932760611176491
211 0.0930332690477371
212 0.0975471114118894
213 0.0965120221177737
214 0.0967432583371798
215 0.0963052262862523
216 0.0945587580402692
217 0.0962222913901011
218 0.0948578317960103
219 0.0948931525150935
220 0.095672311882178
221 0.0941453923781713
222 0.0966614286104838
223 0.0951195483406385
224 0.0959577982624372
225 0.0950521528720856
226 0.0952186485131582
227 0.0960748742024104
228 0.0957487945755323
229 0.094690424700578
230 0.0932854935526848
231 0.0959059372544289
232 0.0945544391870499
233 0.0936786582072576
234 0.0934340357780457
235 0.0958301648497581
236 0.0966253876686096
237 0.096359021961689
238 0.0970443064967791
239 0.0944135660926501
240 0.0956839670737584
241 0.0963657274842262
242 0.0933238516251246
243 0.095818226536115
244 0.0943026120464007
245 0.0936112677057584
246 0.0936310489972432
247 0.0952786480387052
248 0.0967057521144549
249 0.0967086826761564
250 0.0922888790567716
251 0.093141774336497
252 0.096469243367513
253 0.0966140752037366
254 0.0939102421204249
255 0.0947577580809593
256 0.0956111053625743
257 0.0965681100885073
258 0.0963396380345027
259 0.094615047176679
260 0.0932970494031906
261 0.0941283752520879
262 0.0958544835448265
263 0.0975084428985914
264 0.0949036727348963
265 0.0965845808386803
266 0.0954271083076795
267 0.0960042426983515
268 0.0961351444323858
269 0.0945308456818263
270 0.093527485926946
271 0.0960934435327848
272 0.0948646689454715
273 0.0955162247021993
274 0.0945463726917903
275 0.0941724677880605
276 0.0955258533358574
277 0.0956552053491275
278 0.0975134571393331
279 0.0958500107129415
280 0.0957794735829035
281 0.0969027330478032
282 0.0906133775909742
283 0.0964616909623146
284 0.0936882123351097
};
\addplot [line width=2.5pt, blue, opacity=1.0]
table {%
0 2.35180608431498
1 1.80796694755554
2 1.50887978076935
3 1.36825978755951
4 1.29153903325399
5 1.14788206418355
6 1.06330176194509
7 1.01036828756332
8 0.908619781335195
9 0.861547609170278
10 0.822780688603719
11 0.752278566360474
12 0.726650714874268
13 0.690664033095042
14 0.636799156665802
15 0.622696101665497
16 0.591268837451935
17 0.545111934343974
18 0.541185061136881
19 0.50712322195371
20 0.477231790622075
21 0.471933533747991
22 0.443819274504979
23 0.419320831696192
24 0.412826736768087
25 0.393638680378596
26 0.378206898768743
27 0.370458483695984
28 0.353931307792664
29 0.335956166187922
30 0.326151976982753
31 0.315248688062032
32 0.304072976112366
33 0.296264380216599
34 0.280360062917074
35 0.271805584430695
36 0.262646734714508
37 0.249953364332517
38 0.248952428499858
39 0.243101204435031
40 0.231813833117485
41 0.224954307079315
42 0.21813990175724
43 0.205308839678764
44 0.208405807614326
45 0.205374399820964
46 0.201464474201202
47 0.198987076679866
48 0.1937482158343
49 0.185186564922333
50 0.184432501594226
51 0.180778031547864
52 0.178825194636981
53 0.176446760694186
54 0.165764773885409
55 0.16389969487985
56 0.159354199965795
57 0.151870563626289
58 0.151745095849037
59 0.148814246058464
60 0.141756763060888
61 0.137276229759057
62 0.129771123329798
63 0.126693402727445
64 0.126331192751726
65 0.121438510715961
66 0.116993469496568
67 0.110483246544997
68 0.106588741143545
69 0.105055332183838
70 0.0971752852201462
71 0.0937950536608696
72 0.0932905375957489
73 0.0901451582709948
74 0.0883788069089254
75 0.0834874287247658
76 0.0817427734533946
77 0.0766135330001513
78 0.0780886809031169
79 0.0717979321877162
80 0.0727316799263159
81 0.0634690883258979
82 0.0601724001268546
83 0.0596907213330269
84 0.0598664445181688
85 0.054612739632527
86 0.0555846864978472
87 0.053515770783027
88 0.0509429176648458
89 0.050315398722887
90 0.0472515809039275
91 0.0465845403571924
92 0.0453856314222018
93 0.0421218064924081
94 0.0424088065822919
95 0.0377889052033424
96 0.0378874490658442
97 0.0346833541989326
98 0.0353256594389677
99 0.0337570520738761
100 0.0355284499625365
101 0.0322738457471132
102 0.0287074601898591
103 0.0289003358532985
104 0.0278868526220322
105 0.0302931529780229
106 0.0280247504512469
107 0.0271437211583058
108 0.0265766233205795
109 0.0251119397580624
110 0.0269778650254011
111 0.0288127629707257
112 0.0242452335854371
113 0.023726986721158
114 0.0247047686328491
115 0.0226944843307137
116 0.0227799313142896
117 0.0212330709521969
118 0.0203333658476671
119 0.0186734811092416
120 0.0185890961438417
121 0.0192068271959821
122 0.0175915500149131
123 0.0179262133315206
124 0.0162179659431179
125 0.0157618628193935
126 0.0153776311005155
127 0.0167108538250128
128 0.0149060621236761
129 0.0129261035472155
130 0.0124490323166052
131 0.0126193454489112
132 0.0127470477794607
133 0.012483953187863
134 0.0129048839832346
135 0.0119956446190675
136 0.0117235795284311
137 0.0117370591809352
138 0.0112915192730725
139 0.0104165058583021
140 0.0102187062924107
141 0.0165871827242275
142 0.0243207725385825
143 0.0152291520498693
144 0.0092476112768054
145 0.00784439050282041
146 0.00655884637186925
147 0.00670165374564628
148 0.00724470537776748
149 0.00641116841385762
150 0.00601595523767173
151 0.00546567568865915
152 0.00517862794610361
153 0.00561713132386406
154 0.0055432990193367
155 0.00757355196401477
156 0.00869164375277857
157 0.009542233000199
158 0.0101466932489226
159 0.00924616726115346
160 0.00861373062555989
161 0.00945397590597471
162 0.00794458909270664
163 0.00957639096304774
164 0.00943653852057954
165 0.00828360766172409
166 0.00897054102582236
167 0.00937087174194554
168 0.00756430408606927
169 0.00860574018831054
170 0.00889894366264343
171 0.00858696449237565
172 0.00951204728335142
173 0.00996722575897972
174 0.00995798789275189
175 0.00924776719572643
176 0.00841479220738014
177 0.00894828295956055
178 0.00897842723255356
179 0.00953733183753987
180 0.0101890874405702
181 0.0101293739086638
182 0.00824696628842503
183 0.00811940998149415
184 0.00901632523164153
185 0.0098320534452796
186 0.0097544405143708
187 0.0102384120691568
188 0.00926189551440378
189 0.00926952925510705
190 0.00940048570434252
191 0.01013292837888
192 0.00985157210379839
193 0.009137341985479
194 0.00927478199203809
195 0.00988989338899652
196 0.00952143889541427
197 0.00991699254761139
198 0.0092130444633464
199 0.0106563771453997
200 0.0101280764987071
201 0.0119116903903584
202 0.0143408020958304
203 0.0119908570001523
204 0.0150638865306973
205 0.0128639377653599
206 0.0178141891956329
207 0.0225938949733973
208 0.0233877835174402
209 0.0206651737292608
210 0.019243446799616
211 0.0212644723554452
212 0.0228808112442493
213 0.0216280243669947
214 0.0163693381473422
215 0.00892429519444704
216 0.00788375098879139
217 0.00690641223142544
218 0.00640641308079163
219 0.00654852359245221
220 0.00529943391059836
221 0.00556641568740209
222 0.00471637405765553
223 0.00477596977725625
224 0.00438080898796519
225 0.00502146819295983
226 0.00469582489070793
227 0.00381876877509058
228 0.00437326193787158
229 0.00448959018103778
230 0.00459367712028325
231 0.00427214157146712
232 0.00415736297145486
233 0.00377247172097365
234 0.0035610101185739
235 0.00371105115239819
236 0.00372465047985315
237 0.00397729260536532
238 0.00395515918110808
239 0.00394220114685595
240 0.004267941811122
241 0.00393797325280805
242 0.0037123558189099
243 0.0038668888155371
244 0.00350026064552367
245 0.00390491991614302
246 0.00336989900097251
247 0.00350749725475907
248 0.00324468400018911
249 0.00366376445163041
250 0.00340111134573817
251 0.00304649110573033
252 0.00380726996809244
253 0.00355075462721288
254 0.00366514533137282
255 0.00368332144959519
256 0.00346127059310675
257 0.00390768027864397
258 0.00350971911878636
259 0.00339725702845802
260 0.00334811742262294
261 0.00345912913326174
262 0.00332317140419036
263 0.0036018870693321
264 0.00300648292371382
265 0.00313164561521262
266 0.00297094637062401
267 0.00329599572190394
268 0.00277639122214168
269 0.00345937814563513
270 0.00288026364675413
271 0.00266215003406008
272 0.00364009694506725
273 0.0034149889058123
274 0.00311983260326087
275 0.00316660661095132
276 0.00295653403736651
277 0.00279675303803136
278 0.00295897402490179
279 0.00300892093218863
280 0.00313338978836934
281 0.00284435580639789
282 0.00250306919527551
283 0.00271780078765005
284 0.00262914400082082
};
\end{axis}

\end{tikzpicture}

%% file: labpal/figure_data/performance_comparison_low_batch/plots_bs8/CIFAR-10_DenseNet-121_validation_accuracy.pgf
% This file was created by tikzplotlib v0.9.8.
\begin{tikzpicture}

\definecolor{color0}{rgb}{0.933333333333333,0.509803921568627,0.933333333333333}
\definecolor{color1}{rgb}{0.647058823529412,0.164705882352941,0.164705882352941}
\definecolor{color2}{rgb}{1,0.647058823529412,0}

\begin{axis}[
minor xtick={},
minor ytick={},
tick align=outside,
tick pos=left,
title={validation accuracy CIFAR-10 DenseNet-121},
width=10.5cm,height=8cm,grid=major,major grid style={dotted},
x grid style={white!69.0196078431373!black},
xlabel={epochs},
xmin=0, xmax=298,
xtick style={color=black},
xtick={0,50,100,150,200,250},
y grid style={white!69.0196078431373!black},
ylabel={validation accuracy},
ymin=0.85, ymax=0.94,
ytick style={color=black},
ytick={0.85,0.86,0.87,0.88,0.89,0.9,0.91,0.92,0.93,0.94}
]
\addplot [line width=2.5pt, color0, opacity=1.0]
table {%
1 0.30579999089241
2 0.351599991321564
3 0.463200002908707
4 0.499799996614456
5 0.514999985694885
6 0.554000020027161
7 0.560800015926361
8 0.579999983310699
9 0.609600007534027
10 0.613200008869171
11 0.585600018501282
12 0.631399989128113
13 0.630800008773804
14 0.657599985599518
15 0.629000008106232
16 0.69080001115799
17 0.693000018596649
18 0.698599994182587
19 0.717999994754791
20 0.70959997177124
21 0.7185999751091
22 0.717000007629395
23 0.737200021743774
24 0.756600022315979
25 0.732999980449677
26 0.736599981784821
27 0.752399981021881
28 0.763800024986267
29 0.773599982261658
30 0.776000022888184
31 0.769999980926514
32 0.761600017547607
33 0.7882000207901
34 0.742200016975403
35 0.765999972820282
36 0.785600006580353
37 0.763599991798401
38 0.785600006580353
39 0.77700001001358
40 0.771000027656555
41 0.77539998292923
42 0.792800009250641
43 0.786199986934662
44 0.750400006771088
45 0.806400001049042
46 0.779999971389771
47 0.823199987411499
48 0.804400026798248
49 0.799199998378754
50 0.786199986934662
51 0.819800019264221
52 0.783800005912781
53 0.793200016021729
54 0.801999986171722
55 0.771200001239777
56 0.79040002822876
57 0.80919998884201
58 0.810599982738495
59 0.777999997138977
60 0.787199974060059
61 0.832400023937225
62 0.835799992084503
63 0.817799985408783
64 0.821799993515015
65 0.840799987316132
66 0.83380001783371
67 0.843200027942657
68 0.829200029373169
69 0.838199973106384
70 0.844200015068054
71 0.805999994277954
};
\addplot [line width=2.5pt, color1, opacity=1.0]
table {%
1 0.0981999983390172
2 0.102066665887833
3 0.0967333341638247
4 0.100733334819476
5 0.0990666647752126
6 0.102199998994668
7 0.100066664318244
8 0.101799999674161
9 0.0998666658997536
10 0.0999999990065892
11 0.102666666110357
12 0.100466666122278
13 0.100466666122278
14 0.0964666654666265
15 0.102933332324028
16 0.0993333334724108
17 0.101199999451637
18 0.0986000001430511
19 0.108600000540415
20 0.101333332558473
21 0.103133330742518
22 0.101599998772144
23 0.10080000013113
24 0.0961333339413007
25 0.103600000341733
26 0.0977333337068558
27 0.103866666555405
28 0.0993999987840652
29 0.101333332558473
30 0.104266665875912
31 0.0989999994635582
32 0.102666666110357
33 0.104399998982747
34 0.0993333334724108
35 0.10266666362683
36 0.0993999987840652
37 0.0991333325703939
38 0.0973333319028219
39 0.0986000001430511
40 0.110466664036115
41 0.0953333328167597
42 0.103866664071878
43 0.100266665220261
44 0.0996666674812635
45 0.100733332335949
46 0.103133330742518
47 0.104066664973895
48 0.102733331422011
49 0.0977333337068558
50 0.100399998327096
51 0.103199998537699
52 0.0996000021696091
53 0.0961999992529551
54 0.0941333323717117
55 0.103399999439716
56 0.102133331199487
57 0.101266667246819
58 0.100666667024295
59 0.0991333325703939
60 0.0946666672825813
61 0.0993999987840652
62 0.0994000012675921
63 0.101933332780997
64 0.0986000001430511
65 0.0997999981045723
66 0.100333333015442
67 0.102533333003521
68 0.100133332113425
69 0.100199999908606
70 0.0996666649977366
71 0.10099999854962
72 0.098133330543836
73 0.100199999908606
74 0.0973333319028219
75 0.0995999996860822
76 0.112066666285197
77 0.103133330742518
78 0.100599999229113
79 0.10793333252271
80 0.101533330976963
81 0.100266665220261
82 0.0994666665792465
83 0.0994666665792465
84 0.102333332101504
85 0.094200000166893
86 0.100933330754439
87 0.108199998736382
88 0.0987333332498868
89 0.100733332335949
90 0.0969333325823148
91 0.101466665665309
92 0.101799997190634
93 0.100933330754439
94 0.100000001490116
95 0.097800001502037
96 0.100599999229113
97 0.0992000003655751
98 0.0997333327929179
99 0.0964000001549721
100 0.100733332335949
101 0.0961999992529551
102 0.100199997425079
103 0.103533332546552
104 0.102066665887833
105 0.106666666766008
106 0.100399998327096
107 0.10099999854962
108 0.0997333327929179
109 0.0979333346088727
110 0.103066665430864
111 0.0993333334724108
112 0.0992000003655751
113 0.102133333683014
114 0.101066663861275
115 0.10080000013113
116 0.0999999990065892
117 0.0980666652321815
118 0.102866664528847
119 0.102733333905538
120 0.100466666122278
121 0.102133333683014
122 0.0988000010450681
123 0.101866667469343
124 0.0997999981045723
125 0.0953333328167597
126 0.0992666656772296
127 0.104333331187566
128 0.101866664985816
129 0.102866667012374
130 0.101266664763292
131 0.100866665442785
132 0.108133333424727
133 0.105199997623761
134 0.102933332324028
135 0.0995333318909009
136 0.102399999896685
137 0.104599999884764
138 0.101400000353654
139 0.0995333318909009
140 0.0989999994635582
141 0.100133334596952
142 0.10099999854962
143 0.102266664306323
144 0.0993333334724108
145 0.0998666658997536
146 0.10173333187898
147 0.10266666362683
148 0.0957333321372668
149 0.100199997425079
150 0.0996666649977366
151 0.0885999997456868
152 0.111466663579146
153 0.101000001033147
154 0.0961333339413007
155 0.101799999674161
156 0.100333333015442
157 0.104333331187566
158 0.0997999981045723
159 0.113866664469242
160 0.102600000798702
161 0.0961333339413007
162 0.0983333339293798
163 0.100199997425079
164 0.103399999439716
165 0.103733330965042
166 0.0964666654666265
167 0.102133333683014
168 0.101333332558473
169 0.0988000010450681
170 0.102933332324028
171 0.0957333346207937
172 0.104266665875912
173 0.10099999854962
174 0.0993999987840652
175 0.1047333329916
};
\addplot [line width=2.5pt, red, opacity=1.0]
table {%
1 0.470399990677834
2 0.533300012350082
3 0.578600019216537
4 0.620099991559982
5 0.648299992084503
6 0.675200015306473
7 0.693200021982193
8 0.738299995660782
9 0.73879998922348
10 0.751899987459183
11 0.776799976825714
12 0.736600011587143
13 0.77240002155304
14 0.794699996709824
15 0.800799995660782
16 0.807500004768372
17 0.8158999979496
18 0.801899999380112
19 0.808799982070923
20 0.840700000524521
21 0.836199998855591
22 0.83840000629425
23 0.857600003480911
24 0.847100019454956
25 0.864199995994568
26 0.856099992990494
27 0.862499982118607
28 0.870800018310547
29 0.870799988508224
30 0.854900002479553
31 0.860199987888336
32 0.870800018310547
33 0.878200024366379
34 0.875500023365021
35 0.866699993610382
36 0.87389999628067
37 0.879599988460541
38 0.876399993896484
39 0.883700013160706
40 0.881600022315979
41 0.881300002336502
42 0.883500009775162
43 0.882699996232986
44 0.879400014877319
45 0.883199989795685
46 0.881099998950958
47 0.874200016260147
48 0.880299985408783
49 0.881999999284744
50 0.870799988508224
51 0.887799978256226
52 0.887100011110306
53 0.894700020551682
54 0.893500000238419
55 0.880899995565414
56 0.890100002288818
57 0.884100019931793
58 0.894199997186661
59 0.891400009393692
60 0.888300001621246
61 0.892599999904633
62 0.879200011491776
63 0.901500016450882
64 0.900499999523163
65 0.898999989032745
66 0.90090000629425
67 0.899100005626678
68 0.897599995136261
69 0.899399995803833
70 0.90090000629425
71 0.901600003242493
72 0.904699981212616
73 0.90149998664856
74 0.9060999751091
75 0.902900010347366
76 0.89750000834465
77 0.901800006628036
78 0.904400020837784
79 0.904799997806549
80 0.906499981880188
81 0.905600011348724
82 0.901699990034103
83 0.904500007629395
84 0.911200016736984
85 0.903400003910065
86 0.907999992370605
87 0.907499998807907
88 0.90939998626709
89 0.905400007963181
90 0.911100000143051
91 0.909299999475479
92 0.909899979829788
93 0.90719997882843
94 0.912699997425079
95 0.911399990320206
96 0.913700014352798
97 0.911900013685226
98 0.910699993371964
99 0.91389998793602
100 0.915899991989136
101 0.90829998254776
102 0.909600019454956
103 0.913699984550476
104 0.916299998760223
105 0.914700001478195
106 0.91770002245903
107 0.914900004863739
108 0.912999987602234
109 0.913599997758865
110 0.914899975061417
111 0.912400007247925
112 0.914800018072128
113 0.913800001144409
114 0.912299990653992
115 0.916900008916855
116 0.91609999537468
117 0.912699997425079
118 0.918599992990494
119 0.914600014686584
120 0.918699979782104
121 0.915100008249283
122 0.918899983167648
123 0.918799996376038
124 0.918000012636185
125 0.91839998960495
126 0.918500006198883
127 0.918099999427795
128 0.919400006532669
129 0.918899983167648
130 0.919099986553192
131 0.917499989271164
132 0.922499984502792
133 0.919699996709824
134 0.920100003480911
135 0.917999982833862
136 0.919900000095367
137 0.918499976396561
138 0.920899987220764
139 0.919200003147125
140 0.921199977397919
141 0.920499980449677
142 0.921000003814697
143 0.921400010585785
144 0.92059999704361
145 0.922800004482269
146 0.91949999332428
147 0.921600013971329
148 0.920400023460388
149 0.920700013637543
150 0.923799991607666
151 0.919300019741058
152 0.922799974679947
153 0.92069998383522
154 0.92179998755455
155 0.923699975013733
156 0.920000016689301
157 0.919299989938736
158 0.922399997711182
159 0.919699996709824
160 0.916700005531311
161 0.918900012969971
162 0.922000020742416
163 0.922600001096725
164 0.922800004482269
165 0.919799983501434
166 0.918300002813339
167 0.920800000429153
168 0.92220002412796
169 0.919999986886978
170 0.921399980783463
171 0.92059999704361
172 0.924199998378754
173 0.927299976348877
174 0.923500001430511
175 0.925200015306473
176 0.924199998378754
177 0.922299981117249
178 0.924699991941452
179 0.924499988555908
180 0.923299998044968
181 0.921000003814697
182 0.921499997377396
183 0.921399980783463
184 0.92399999499321
185 0.922199994325638
186 0.922800004482269
187 0.923700004816055
188 0.923799991607666
189 0.923299998044968
190 0.922399997711182
191 0.923799991607666
192 0.92560002207756
193 0.92519998550415
194 0.922199994325638
195 0.925200015306473
196 0.928300023078918
197 0.925799995660782
198 0.922399997711182
199 0.924299985170364
200 0.923500001430511
201 0.924100011587143
202 0.924699991941452
203 0.925300002098083
204 0.92629998922348
205 0.923600018024445
206 0.926100015640259
207 0.92629998922348
208 0.924199998378754
209 0.926399976015091
210 0.923899978399277
211 0.922499984502792
212 0.925299972295761
213 0.927900016307831
214 0.928299993276596
215 0.924499988555908
216 0.927300006151199
217 0.927100002765656
218 0.92560002207756
219 0.925300002098083
220 0.92509999871254
221 0.92620000243187
222 0.923900008201599
223 0.92509999871254
224 0.924400001764297
225 0.928000003099442
226 0.926499992609024
227 0.924400001764297
228 0.928000003099442
229 0.926499992609024
230 0.924700021743774
231 0.925799995660782
232 0.928300023078918
233 0.926899999380112
234 0.924899995326996
235 0.927700012922287
236 0.927699983119965
237 0.926599979400635
238 0.927300006151199
239 0.924899995326996
240 0.926599979400635
241 0.924699991941452
242 0.925700008869171
243 0.927300006151199
244 0.925699979066849
245 0.924199998378754
246 0.924899995326996
247 0.925900012254715
248 0.928299993276596
249 0.926799982786179
250 0.925899982452393
251 0.925799995660782
252 0.927700012922287
253 0.927100002765656
254 0.926499992609024
255 0.928799986839294
256 0.926400005817413
257 0.926400005817413
258 0.926799982786179
259 0.924899995326996
260 0.926600009202957
261 0.925799995660782
262 0.922399997711182
263 0.925300002098083
264 0.930399984121323
265 0.926800012588501
266 0.926899999380112
267 0.926300019025803
268 0.928100019693375
269 0.92620000243187
270 0.925799995660782
271 0.926599979400635
272 0.924699991941452
273 0.925000011920929
274 0.929900020360947
275 0.924899995326996
276 0.926799982786179
277 0.927100002765656
278 0.929499983787537
279 0.926099985837936
280 0.926699995994568
281 0.926699995994568
282 0.924899995326996
283 0.926699995994568
284 0.928500026464462
};
\addplot [line width=2.5pt, color2, opacity=1.0]
table {%
2 0.103799998760223
4 0.10226666678985
6 0.0992000003655751
8 0.100466666122278
10 0.100466668605804
12 0.101066666344802
14 0.101000001033147
16 0.0989333316683769
18 0.0975999981164932
20 0.0984666645526886
22 0.100266667703787
24 0.100666669507821
26 0.0993333334724108
28 0.100466666122278
30 0.100266665220261
32 0.10099999854962
34 0.0980666677157084
36 0.100533333917459
38 0.0964000001549721
40 0.0994666665792465
42 0.102933332324028
44 0.0986666679382324
46 0.100066664318244
48 0.094200000166893
50 0.104066664973895
52 0.102866667012374
54 0.0953333328167597
56 0.0972666665911674
58 0.100733332335949
60 0.100266667703787
62 0.100466666122278
64 0.0986666654547056
66 0.101600001255671
68 0.096600001056989
70 0.0968666672706604
72 0.0987999985615412
74 0.0968666672706604
76 0.101133331656456
78 0.102733333905538
80 0.0960666661461194
82 0.101399997870127
84 0.0961333339413007
86 0.101066666344802
88 0.0989999994635582
90 0.096600001056989
92 0.0961333339413007
94 0.0979999999205271
96 0.101000001033147
98 0.0984000017245611
100 0.0997333327929179
102 0.100933333237966
104 0.101133331656456
106 0.101266664763292
108 0.0995999996860822
110 0.101133331656456
112 0.0966666638851166
114 0.0965333332618078
116 0.100266665220261
118 0.103666665653388
120 0.098600002626578
122 0.100466666122278
124 0.101333332558473
126 0.0961999992529551
128 0.0998666658997536
130 0.103399999439716
132 0.0984666645526886
134 0.0991333350539207
136 0.0988666663567225
138 0.0972666665911674
140 0.0999999990065892
142 0.103866664071878
144 0.103866666555405
146 0.101400000353654
148 0.102933332324028
150 0.099333330988884
152 0.095400000611941
154 0.0981999983390172
156 0.0981333330273628
158 0.101666664083799
160 0.10246666520834
162 0.101066666344802
164 0.102533333003521
166 0.0981333330273628
168 0.0993333334724108
170 0.102266664306323
172 0.0999999990065892
174 0.0961999992529551
176 0.101066663861275
178 0.101599998772144
180 0.104266665875912
182 0.0994000012675921
184 0.101199999451637
186 0.0982666661341985
188 0.0989333316683769
190 0.100933333237966
192 0.0989333316683769
194 0.0997999981045723
196 0.0973333343863487
198 0.09740000218153
200 0.0980666652321815
202 0.0997333327929179
204 0.101266664763292
206 0.102866667012374
208 0.100399998327096
210 0.103866666555405
212 0.100266665220261
214 0.0997999981045723
216 0.0976666659116745
218 0.102133333683014
220 0.0980666652321815
222 0.100066666801771
224 0.101199999451637
226 0.100133334596952
228 0.0982000008225441
230 0.101933332780997
232 0.10099999854962
234 0.0963333323597908
236 0.0993999987840652
238 0.0991333325703939
240 0.101199999451637
242 0.101666666567326
244 0.102000000576178
246 0.0982666661341985
248 0.0974666650096575
250 0.100333333015442
252 0.0995333318909009
254 0.102399997413158
256 0.100466666122278
258 0.0973999996980031
260 0.102199998994668
262 0.0981333330273628
264 0.103466667234898
266 0.103133330742518
268 0.102399999896685
270 0.100666667024295
272 0.100333333015442
274 0.0990666647752126
276 0.0992666656772296
278 0.100599999229113
280 0.101333332558473
282 0.101400000353654
284 0.100333333015442
};
\addplot [line width=2.5pt, green!50.1960784313725!black, opacity=1.0]
table {%
1 0.364466667175293
2 0.489933321873347
3 0.567799985408783
4 0.59826668103536
5 0.619599997997284
6 0.63726665576299
7 0.628800014654795
8 0.689733326435089
9 0.660866677761078
10 0.638999988635381
11 0.697933336098989
12 0.719400008519491
13 0.705333332220713
14 0.736733337243398
15 0.740599989891052
16 0.771599988142649
17 0.78766667842865
18 0.750733315944672
19 0.77593332529068
20 0.796466668446859
21 0.712266663710276
22 0.783666670322418
23 0.809800008932749
24 0.790866653124491
25 0.785666664441427
26 0.799999992052714
27 0.777600010236104
28 0.795399983723958
29 0.795799990495046
30 0.816799998283386
31 0.818399985631307
32 0.817533334096273
33 0.829599996407827
34 0.831933339436849
35 0.806800007820129
36 0.842733323574066
37 0.849733332792918
38 0.828666667143504
39 0.834733347098033
40 0.825133343537649
41 0.846466680367788
42 0.751600007216136
43 0.796333332856496
44 0.801066676775614
45 0.803599993387858
46 0.823466658592224
47 0.824933330217997
48 0.659000019232432
49 0.707466671864192
50 0.741266667842865
51 0.732933322588603
52 0.779533326625824
53 0.789266665776571
54 0.79666668176651
55 0.767866671085358
56 0.798666656017303
57 0.816733340422312
58 0.79093333085378
59 0.815533339977264
60 0.803800006707509
61 0.825000007947286
62 0.798199991385142
63 0.827133317788442
64 0.830666661262512
65 0.837333341439565
66 0.831999977429708
67 0.828266680240631
68 0.848266661167145
69 0.847333331902822
70 0.841733336448669
71 0.844400008519491
72 0.856266657511393
73 0.8392666776975
74 0.854200005531311
75 0.853133320808411
76 0.845933338006338
77 0.85073333978653
78 0.853866656621297
79 0.844866673151652
80 0.851333340009054
81 0.84253333012263
82 0.842399994532267
83 0.861733317375183
84 0.842133323351542
85 0.856666664282481
86 0.84553333123525
87 0.846599996089935
88 0.854666670163473
89 0.838133335113525
90 0.851533313592275
91 0.857999980449677
92 0.856333335240682
93 0.859800000985463
94 0.846200009187063
95 0.852933327356974
96 0.84253333012263
97 0.831533332665761
98 0.860733330249786
99 0.855999986330668
100 0.863333324591319
101 0.856333335240682
102 0.86053333679835
103 0.864466647307078
104 0.870266656080882
105 0.863333344459534
106 0.859666645526886
107 0.856666684150696
108 0.871533354123433
109 0.867266654968262
110 0.874466677506765
111 0.864866673946381
112 0.861866652965546
113 0.85233332713445
114 0.86520000298818
115 0.875333348910014
116 0.870333313941956
117 0.857866684595744
118 0.872933328151703
119 0.878866652647654
120 0.875333329041799
121 0.870133340358734
122 0.86353333791097
123 0.870733320713043
124 0.867800017197927
125 0.877400000890096
126 0.864600002765656
127 0.878399988015493
128 0.876866678396861
129 0.868533333142599
130 0.875666658083598
131 0.860466678937276
132 0.87199999888738
133 0.874466677506765
134 0.869399984677633
135 0.86573334534963
136 0.865066667397817
137 0.874666670958201
138 0.868399997552236
139 0.869199991226196
140 0.874066670735677
141 0.867866675059001
142 0.862866659959157
143 0.881933331489563
144 0.878866672515869
145 0.882399996121724
146 0.888800005118052
147 0.88700000445048
148 0.886533339818319
149 0.876199980576833
150 0.889933327833811
151 0.885266661643982
152 0.886799991130829
153 0.892866671085358
154 0.899133324623108
155 0.889799992243449
156 0.896733323733012
157 0.892933328946432
158 0.888533333937327
159 0.884266674518585
160 0.895400007565816
161 0.882933338483175
162 0.890199999014537
163 0.895333329836527
164 0.892199993133545
165 0.896333336830139
166 0.898333330949148
167 0.896399994691213
168 0.89053334792455
169 0.895066658655802
170 0.884866654872894
171 0.895799994468689
172 0.88453334569931
173 0.895266671975454
174 0.88866666952769
175 0.896599988142649
176 0.896399994691213
177 0.897800008455912
178 0.893533329168955
179 0.888800005118052
180 0.892066657543182
181 0.89466667175293
182 0.890333334604899
183 0.89766667286555
184 0.898666659990946
185 0.899399995803833
186 0.899133324623108
187 0.896066685517629
188 0.896200001239777
189 0.889799992243449
190 0.898999989032745
191 0.896399994691213
192 0.896399994691213
193 0.891266663869222
194 0.89300000667572
195 0.898866673310598
196 0.902266681194305
197 0.897866666316986
198 0.898600002129873
199 0.898533324400584
200 0.897533337275187
201 0.898799975713094
202 0.893466651439667
203 0.900866667429606
204 0.899599989255269
205 0.903933346271515
206 0.896999994913737
207 0.900866667429606
208 0.900600016117096
209 0.896600008010864
210 0.896733343601227
211 0.893599987030029
212 0.897866666316986
213 0.891800006230672
214 0.892333348592122
215 0.901066660881042
216 0.889533340930939
217 0.898400008678436
218 0.901199996471405
219 0.903200010458628
220 0.890666683514913
221 0.894466678301493
222 0.903733332951864
223 0.903800010681152
224 0.892800013224284
225 0.894600013891856
226 0.898533344268799
227 0.898200015227
228 0.899600009123484
229 0.892800013224284
230 0.896733323733012
231 0.895533343156179
232 0.896200021107992
233 0.901733338832855
234 0.900266667207082
235 0.898400008678436
236 0.894733349482218
237 0.902933339277903
238 0.893466651439667
239 0.90146666765213
240 0.902133325735728
241 0.896933337052663
242 0.899399995803833
243 0.897266646226247
244 0.902133325735728
245 0.89520001411438
246 0.900533318519592
247 0.90013333161672
248 0.90119997660319
249 0.89766667286555
250 0.902399996916453
251 0.895600020885468
252 0.896066665649414
253 0.901266674200694
254 0.898133317629496
255 0.901599983374278
256 0.896199981371562
257 0.898533324400584
258 0.895599981149038
259 0.89546666542689
260 0.898600002129873
261 0.895199994246165
262 0.900733331839244
263 0.898933331171672
264 0.897666652997335
265 0.892133335272471
266 0.896133343378703
267 0.896466672420502
268 0.895066658655802
269 0.898533324400584
270 0.901266674200694
271 0.898733337720235
272 0.891933341821035
273 0.887600004673004
274 0.900733331839244
275 0.894733309745789
276 0.893933316071828
277 0.903066674868266
278 0.899399995803833
279 0.900333325068156
280 0.90146666765213
281 0.897266666094462
282 0.902066667874654
283 0.899933338165283
284 0.898866673310598
};
\addplot [line width=2.5pt, black, opacity=1.0]
table {%
1 0.439266661802928
2 0.568400005499522
3 0.635066668192546
4 0.655066668987274
5 0.679533322652181
6 0.705933332443237
7 0.705666681130727
8 0.734399994214376
9 0.749800006548564
10 0.762199997901917
11 0.775800009568532
12 0.762333313624064
13 0.788133323192596
14 0.816266675790151
15 0.809066673119863
16 0.836066683133443
17 0.837799986203512
18 0.8392666776975
19 0.844066679477692
20 0.853600005308787
21 0.848933339118958
22 0.862866679827372
23 0.86133333047231
24 0.864600002765656
25 0.871333340803782
26 0.872133334477743
27 0.875466664632161
28 0.875133335590363
29 0.878199994564056
30 0.877666672070821
31 0.887200017770131
32 0.881800015767415
33 0.88266666730245
34 0.887866655985514
35 0.888800005118052
36 0.892066677411397
37 0.890599985917409
38 0.891200006008148
39 0.898466686407725
40 0.895400007565816
41 0.89926666021347
42 0.895266652107239
43 0.895999987920125
44 0.899666666984558
45 0.902400016784668
46 0.895533343156179
47 0.900599996248881
48 0.900266667207082
49 0.899866660435995
50 0.898600002129873
51 0.902066667874654
52 0.903533339500427
53 0.904333313306173
54 0.903866648674011
55 0.904666682084401
56 0.904933333396912
57 0.903733332951864
58 0.906199991703033
59 0.904733339945475
60 0.9050000111262
61 0.904933333396912
62 0.901199996471405
63 0.902866661548615
64 0.904599984486898
65 0.905733327070872
66 0.904333333174388
67 0.904600004355113
68 0.908333341280619
69 0.903933346271515
70 0.903200010458628
71 0.905799984931946
72 0.905533333619436
73 0.904866655667623
74 0.903066674868266
75 0.905399998029073
76 0.905999998251597
77 0.906533320744832
78 0.90366667509079
79 0.90173331896464
80 0.905533333619436
81 0.904399991035461
82 0.905066668987274
83 0.904999991257985
84 0.904133339722951
85 0.903799990812937
86 0.901866674423218
87 0.9050000111262
88 0.904866655667623
89 0.904733339945475
90 0.903533319632212
91 0.902200003465017
92 0.904333333174388
93 0.905466675758362
94 0.907199998696645
95 0.904000004132589
96 0.90446666876475
97 0.907133340835571
98 0.907000005245209
99 0.905266662438711
100 0.906400005022685
101 0.905133346716563
102 0.903333346048991
103 0.902599990367889
104 0.907266656557719
105 0.90366667509079
106 0.903799990812937
107 0.906266669432322
108 0.90420001745224
109 0.906599998474121
110 0.90666667620341
111 0.904400010903676
112 0.903733332951864
113 0.903666655222575
114 0.905066668987274
115 0.905933340390523
116 0.907533327738444
117 0.902733325958252
118 0.903866668542226
119 0.905199984709422
120 0.902800003687541
121 0.905999998251597
122 0.901933312416077
123 0.906733334064484
124 0.90666667620341
125 0.904866675535838
126 0.902933339277903
127 0.903866668542226
128 0.906066675980886
129 0.904399991035461
130 0.904866655667623
131 0.906066675980886
132 0.905266662438711
133 0.905266662438711
134 0.90613333384196
135 0.904266675313314
136 0.906199991703033
137 0.907133340835571
138 0.907799998919169
139 0.902933339277903
140 0.905466675758362
141 0.903866648674011
142 0.905066668987274
143 0.906466643015544
144 0.902733325958252
145 0.905799984931946
146 0.904933333396912
147 0.906466662883759
148 0.902599990367889
149 0.902999997138977
150 0.905933340390523
151 0.9039333264033
152 0.907000005245209
153 0.90066667397817
154 0.902800003687541
155 0.903199990590413
156 0.905466675758362
157 0.902066667874654
158 0.906733334064484
159 0.905066668987274
160 0.90146666765213
161 0.903733313083649
162 0.90446666876475
163 0.903800010681152
164 0.903266668319702
165 0.903799990812937
166 0.903066655000051
167 0.907533327738444
168 0.903866688410441
169 0.904266675313314
170 0.903933346271515
171 0.902399996916453
172 0.905999998251597
173 0.905333340167999
174 0.905200004577637
175 0.906533340613047
176 0.905533333619436
177 0.905333340167999
178 0.907066663106283
179 0.903733352820078
180 0.905599991480509
181 0.905999998251597
182 0.906799991925557
183 0.905799984931946
184 0.901466687520345
185 0.903266668319702
186 0.905466675758362
187 0.906199991703033
188 0.907066663106283
189 0.904533326625824
190 0.903666655222575
191 0.905600011348724
192 0.904466648896535
193 0.902066667874654
194 0.905933320522308
195 0.904533326625824
196 0.90613333384196
197 0.908866663773855
198 0.905599991480509
199 0.903799990812937
200 0.901933352152506
201 0.903133352597555
202 0.903933346271515
203 0.906866669654846
204 0.907266676425934
205 0.906333327293396
206 0.907066663106283
207 0.904266655445099
208 0.9039333264033
209 0.901933352152506
210 0.904133339722951
211 0.905066668987274
212 0.905666649341583
213 0.905200004577637
214 0.905333340167999
215 0.904000004132589
216 0.904133339722951
217 0.904199997584025
218 0.906199991703033
219 0.90200001001358
220 0.903799990812937
221 0.905533313751221
222 0.904533326625824
223 0.902133325735728
224 0.905200004577637
225 0.904733339945475
226 0.901600003242493
227 0.903733332951864
228 0.904733339945475
229 0.904333333174388
230 0.906733334064484
231 0.904000004132589
232 0.905933340390523
233 0.904399991035461
234 0.90773332118988
235 0.901400009791056
236 0.907066663106283
237 0.905200004577637
238 0.904799997806549
239 0.905733327070872
240 0.906200011571248
241 0.904533346494039
242 0.906333347161611
243 0.906666656335195
244 0.90446666876475
245 0.903199990590413
246 0.904066661993662
247 0.905333320299784
248 0.901733338832855
249 0.905733327070872
250 0.906800011793772
251 0.903799990812937
252 0.905200004577637
253 0.905399998029073
254 0.90613333384196
255 0.904599984486898
256 0.903866668542226
257 0.904999991257985
258 0.906733314196269
259 0.905266662438711
260 0.907066663106283
261 0.900333325068156
262 0.907199998696645
263 0.903333326180776
264 0.903800010681152
265 0.904666682084401
266 0.903533339500427
267 0.907933334509532
268 0.903599997361501
269 0.906866669654846
270 0.904799997806549
271 0.905466655890147
272 0.906333347161611
273 0.905466655890147
274 0.905466675758362
275 0.906333327293396
276 0.903799990812937
277 0.903533339500427
278 0.906333347161611
279 0.904199997584025
280 0.903666655222575
281 0.904999991257985
282 0.901799996693929
283 0.905666669209798
284 0.905133326848348
};
\addplot [line width=2.5pt, blue, opacity=1.0]
table {%
1 0.451466659704844
2 0.535333335399628
3 0.559800008932749
4 0.619199991226196
5 0.642666677633921
6 0.67933334906896
7 0.689466675122579
8 0.73333332935969
9 0.72846664985021
10 0.763800005118052
11 0.773533324400584
12 0.752933343251546
13 0.787999987602234
14 0.804266671339671
15 0.812599996725718
16 0.810466667016347
17 0.823466658592224
18 0.823133329550425
19 0.824266672134399
20 0.838533341884613
21 0.826533317565918
22 0.84086666504542
23 0.844533324241638
24 0.838133335113525
25 0.846933325131734
26 0.840266684691111
27 0.854800005753835
28 0.860466678937276
29 0.848800003528595
30 0.85646665096283
31 0.855800012747447
32 0.859733323256175
33 0.860933323701223
34 0.867933332920074
35 0.872999986012777
36 0.870066662629445
37 0.872466663519541
38 0.874466677506765
39 0.876399993896484
40 0.8739333152771
41 0.876866658528646
42 0.877933343251546
43 0.875466664632161
44 0.879599988460541
45 0.88266666730245
46 0.879266659418742
47 0.874399999777476
48 0.879799981911977
49 0.882399996121724
50 0.885133345921834
51 0.885599990685781
52 0.880400002002716
53 0.881999989350637
54 0.887133340040843
55 0.878866672515869
56 0.884733339150747
57 0.884400010108948
58 0.890533328056335
59 0.888133327166239
60 0.88346666097641
61 0.891266663869222
62 0.889000018437703
63 0.892333328723907
64 0.887333333492279
65 0.888800005118052
66 0.899066666762034
67 0.891266663869222
68 0.895800014336904
69 0.894800007343292
70 0.894866685072581
71 0.898333330949148
72 0.89053334792455
73 0.890733341375987
74 0.901199996471405
75 0.899866660435995
76 0.899533331394196
77 0.895333329836527
78 0.897933344046275
79 0.900466660658518
80 0.900799989700317
81 0.900866667429606
82 0.905533333619436
83 0.903200010458628
84 0.902666668097178
85 0.904666662216187
86 0.900733351707458
87 0.903799990812937
88 0.903466661771139
89 0.900733351707458
90 0.902866661548615
91 0.900800009568532
92 0.904133319854736
93 0.906266669432322
94 0.909733335177104
95 0.906200011571248
96 0.909533321857452
97 0.911133329073588
98 0.908733328183492
99 0.906600018342336
100 0.908466657002767
101 0.907600005467733
102 0.907933334509532
103 0.910599986712138
104 0.909133354822795
105 0.910999993483225
106 0.909800012906392
107 0.9121333360672
108 0.91293332974116
109 0.912266671657562
110 0.911266664663951
111 0.910999993483225
112 0.912999987602234
113 0.911266684532166
114 0.912799994150797
115 0.912733316421509
116 0.915333350499471
117 0.910333355267843
118 0.913999994595846
119 0.91539998849233
120 0.914133330186208
121 0.910666664441427
122 0.913399994373322
123 0.913066645463308
124 0.912000020345052
125 0.913533329963684
126 0.910266677538554
127 0.913599987824758
128 0.914333323637644
129 0.913933336734772
130 0.915533324082693
131 0.917199989159902
132 0.91619998216629
133 0.916133344173431
134 0.913999994595846
135 0.915800015131632
136 0.917066673437754
137 0.916333337624868
138 0.915600001811981
139 0.919933319091797
140 0.916600008805593
141 0.912999987602234
142 0.913333336512248
143 0.919399996598562
144 0.918866654237111
145 0.921666661898295
146 0.920533339182536
147 0.920066674550374
148 0.917400002479553
149 0.918733338514964
150 0.918266673882802
151 0.921266655127207
152 0.920133352279663
153 0.921266655127207
154 0.920200010140737
155 0.917199989159902
156 0.920333325862885
157 0.91759999593099
158 0.920933345953623
159 0.915999988714854
160 0.919533332188924
161 0.915066679318746
162 0.920466661453247
163 0.918333331743876
164 0.917400002479553
165 0.917266666889191
166 0.91646667321523
167 0.917733331521352
168 0.918999989827474
169 0.917333324750265
170 0.918666660785675
171 0.917866667111715
172 0.914733350276947
173 0.914266645908356
174 0.916399995485942
175 0.915799995263418
176 0.919399996598562
177 0.915600001811981
178 0.915666659673055
179 0.917199989159902
180 0.919466654459635
181 0.915666659673055
182 0.918799976507823
183 0.917066653569539
184 0.916999995708466
185 0.917866667111715
186 0.919399976730347
187 0.918400009473165
188 0.920866668224335
189 0.915800015131632
190 0.915600001811981
191 0.915400008360545
192 0.918266673882802
193 0.917400002479553
194 0.919066667556763
195 0.917000015576681
196 0.918866674105326
197 0.916866679986318
198 0.916266659895579
199 0.916666666666667
200 0.915666659673055
201 0.91593333085378
202 0.917466660340627
203 0.913599987824758
204 0.912933349609375
205 0.915199995040894
206 0.911333342393239
207 0.904533326625824
208 0.906999985376994
209 0.911800007025401
210 0.910399993260702
211 0.909999986489614
212 0.907866676648458
213 0.912466684977214
214 0.914533336957296
215 0.91813333829244
216 0.918466667334239
217 0.919666667779287
218 0.919999996821086
219 0.922733326752981
220 0.919800003369649
221 0.920266648133596
222 0.921333332856496
223 0.919599990049998
224 0.920999983946482
225 0.917133331298828
226 0.921333332856496
227 0.920866668224335
228 0.922066648801168
229 0.920333345731099
230 0.921000003814697
231 0.921199997266134
232 0.920133332411448
233 0.922333339850108
234 0.919200003147125
235 0.921466668446859
236 0.920600016911825
237 0.921400010585785
238 0.922733326752981
239 0.922000010808309
240 0.920866668224335
241 0.920666674772898
242 0.92113333940506
243 0.923733333746592
244 0.920066654682159
245 0.920066674550374
246 0.922333339850108
247 0.92166668176651
248 0.921533346176147
249 0.920733332633972
250 0.922066668669383
251 0.920200010140737
252 0.9189333319664
253 0.920200010140737
254 0.92086664835612
255 0.919600009918213
256 0.922666668891907
257 0.922133326530457
258 0.919800003369649
259 0.922599991162618
260 0.921199997266134
261 0.917933344841003
262 0.921600004037221
263 0.922399997711182
264 0.923466662565867
265 0.921933313210805
266 0.923333326975505
267 0.921866655349731
268 0.920133332411448
269 0.920600016911825
270 0.922533333301544
271 0.920866668224335
272 0.923066675662994
273 0.920266668001811
274 0.920266668001811
275 0.921200017134349
276 0.921266674995422
277 0.919999996821086
278 0.920733312765757
279 0.920266668001811
280 0.919599990049998
281 0.918466667334239
282 0.9189333319664
283 0.919399996598562
284 0.921466668446859
};
\end{axis}

\end{tikzpicture}

%% file: labpal/figure_data/performance_comparison_low_batch/plots_bs8/CIFAR-10_ResNet-20_test_accuracy.pgf
% This file was created by tikzplotlib v0.9.8.
\begin{tikzpicture}

\definecolor{color0}{rgb}{0.933333333333333,0.509803921568627,0.933333333333333}
\definecolor{color1}{rgb}{1,0.647058823529412,0}

\begin{axis}[
legend cell align={left},
legend style={
  fill opacity=0.8,
  draw opacity=1,
  text opacity=1,
  at={(0.91,0.5)},
  anchor=east,
  draw=white!80!black
},
minor xtick={},
minor ytick={},
tick align=outside,
tick pos=left,
title={test accuracy CIFAR-10 ResNet-20},
width=10.5cm,height=8cm,grid=major,major grid style={dotted},
reverse legend, legend cell align={left}, legend style={ fill opacity=0.8, draw opacity=1, text opacity=1, at={(0.9,0.22)}, anchor=east, draw=white!80!black},,
x grid style={white!69.0196078431373!black},
xlabel={epoch with best val. acc.},
xmin=-0.055, xmax=0.055,
xtick style={color=black},
xtick={-0.06,-0.04,-0.02,0,0.02,0.04,0.06},
y grid style={white!69.0196078431373!black},
ylabel={test accuracy},
ymin=0.85, ymax=0.91,
xmajorticks=false,
ytick style={color=black},
ytick={0.82,0.83,0.84,0.85,0.86,0.87,0.88,0.89,0.90,0.91,0.92,0.93,0.94}
]
\addplot [draw=color0, fill=color0,mark=-, only marks, mark options={scale=3},line width=3pt]
table{%
x  y
0 0.875500003496806
0 -1
};
\addlegendentry{GOLSI : $c$: 0.99, $\eta$: 2.0, $\alpha$: 0.1, $\beta$: 0.4}
\addplot [draw=color1, fill=color1,mark=-, only marks, mark options={scale=3},line width=3pt]
table{%
x  y
0 0.100700000921885
0 -1
};
\addlegendentry{PAL : $\alpha$: 1.0, $\beta$: 0.4, $\mu$: 1.0}
\addplot [
  draw=green!50.1960784313725!black,
  fill=green!50.1960784313725!black,
mark=-, only marks, mark options={scale=3},line width=3pt]
table{%
x  y
-0.005 0.903533359368642
0 -1
};
\addlegendentry{SGD : $\lambda$: 0.1, $\beta$: 0.9}
\addplot [draw=black, fill=black,mark=-, only marks, mark options={scale=3},line width=3pt]
table{%
x  y
0 0.864933331807454
0 -1
};
\addlegendentry{SLS : c: 0.1, $\beta$: 0.9, $\mu$: 0.1}
\addplot [draw=red, fill=red,mark=-, only marks, mark options={scale=3},line width=3pt]
table{%
x  y
0 0.904266655445099
0 -1
};
\addlegendentry{LABPAL-NSGD : $\epsilon$: 4.0 (others same as above)}
\addplot [draw=blue, fill=blue,mark=-, only marks, mark options={scale=3},line width=3pt]
table{%
x  y
0 0.906750023365021
0 -1
};
\addlegendentry{LABPAL-SGD : $\epsilon$: 4.0 (others same as above)}
\end{axis}

\end{tikzpicture}

%% file: labpal/figure_data/performance_comparison_low_batch/plots_bs8/CIFAR-10_MobileNet-V2_test_accuracy.pgf
% This file was created by tikzplotlib v0.9.8.
\begin{tikzpicture}

\definecolor{color0}{rgb}{0.933333333333333,0.509803921568627,0.933333333333333}
\definecolor{color1}{rgb}{1,0.647058823529412,0}
\definecolor{color2}{rgb}{0.647058823529412,0.164705882352941,0.164705882352941}

\begin{axis}[
minor xtick={},
minor ytick={},
tick align=outside,
tick pos=left,
title={test accuracy CIFAR-10 MobileNet-V2},
width=10.5cm,height=8cm,grid=major,major grid style={dotted},
x grid style={white!69.0196078431373!black},
xlabel={epoch with best val. acc.},
xmin=-0.055, xmax=0.055,
xtick style={color=black},
xtick={-0.06,-0.04,-0.02,0,0.02,0.04,0.06},
y grid style={white!69.0196078431373!black},
ylabel={test accuracy},
ymin=0.84, ymax=0.94,
xmajorticks=false,
ytick style={color=black},
ytick={0.85,0.86,0.87,0.88,0.89,0.90,0.91,0.92,0.93,0.94}
]
\addplot [draw=color0, fill=color0,mark=-, only marks, mark options={scale=3},line width=3pt]
table{%
x  y
0 0.883233348528544
0 -1
};
\addplot [draw=red, fill=red,mark=-, only marks, mark options={scale=3},line width=3pt]
table{%
x  y
0 0.937099993228912
0 -1
};
\addplot [draw=blue, fill=blue,mark=-, only marks, mark options={scale=3},line width=3pt]
table{%
x  y
0 0.927933355172475
0 -1
};
\addplot [draw=color1, fill=color1,mark=-, only marks, mark options={scale=3},line width=3pt]
table{%
x  y
0 0.170999998847644
0 -1
};
\addplot [
  draw=green!50.1960784313725!black,
  fill=green!50.1960784313725!black,
mark=-, only marks, mark options={scale=3},line width=3pt
]
table{%
x  y
0 0.924900015195211
0 -1
};
\addplot [draw=black, fill=black,mark=-, only marks, mark options={scale=3},line width=3pt]
table{%
x  y
0 0.910733342170715
0 -1
};
\addplot [draw=color2, fill=color2, mark=-, only marks, mark options={scale=3},line width=3pt]
table{%
x  y
0 0.320449996739626
0 -1
};
\end{axis}

\end{tikzpicture}

%% file: labpal/figure_data/performance_comparison_low_batch/plots_bs8/CIFAR-10_DenseNet-121_test_accuracy.pgf
% This file was created by tikzplotlib v0.9.8.
\begin{tikzpicture}

\definecolor{color0}{rgb}{0.933333333333333,0.509803921568627,0.933333333333333}
\definecolor{color1}{rgb}{1,0.647058823529412,0}

\begin{axis}[
minor xtick={},
minor ytick={},
tick align=outside,
tick pos=left,
title={test accuracy CIFAR-10 DenseNet-121},
width=10.5cm,height=8cm,grid=major,major grid style={dotted},
x grid style={white!69.0196078431373!black},
xlabel={epoch with best val. acc.},
xmin=-0.055, xmax=0.055,
xtick style={color=black},
xtick={-0.06,-0.04,-0.02,0,0.02,0.04,0.06},
y grid style={white!69.0196078431373!black},
ylabel={test accuracy},
ymin=0.84, ymax=0.94,
xmajorticks=false,
ytick style={color=black},
ytick={0.85,0.86,0.87,0.88,0.89,0.90,0.91,0.92,0.93,0.94}
]
\addplot [draw=color0, fill=color0,mark=-, only marks, mark options={scale=3},line width=3pt]
table{%
x  y
0 0.847599983215332
0 -1
};
\addplot [draw=red, fill=red,mark=-, only marks, mark options={scale=3},line width=3pt]
table{%
x  y
0 0.931349992752075
0 -1
};
\addplot [draw=blue, fill=blue,mark=-, only marks, mark options={scale=3},line width=3pt]
table{%
x  y
0 0.923900008201599
0 -1
};
\addplot [draw=color1, fill=color1,mark=-, only marks, mark options={scale=3},line width=3pt]
table{%
x  y
0 0.106000001231829
0 -1
};
\addplot [
  draw=green!50.1960784313725!black,
  fill=green!50.1960784313725!black,
mark=-, only marks, mark options={scale=3},line width=3pt
]
table{%
x  y
0 0.906433343887329
0 -1
};
\addplot [draw=black, fill=black,mark=-, only marks, mark options={scale=3},line width=3pt]
table{%
x  y
-0.005 0.907866656780243
0 -1
};
\end{axis}

\end{tikzpicture}

%% file: apal_iclr.bbl
\begin{thebibliography}{37}
\providecommand{\natexlab}[1]{#1}
\providecommand{\url}[1]{\texttt{#1}}
\expandafter\ifx\csname urlstyle\endcsname\relax
  \providecommand{\doi}[1]{doi: #1}\else
  \providecommand{\doi}{doi: \begingroup \urlstyle{rm}\Url}\fi

\bibitem[Abadi et~al.(2015)Abadi, Agarwal, Barham, Brevdo, Chen, Citro,
  Corrado, Davis, Dean, Devin, Ghemawat, Goodfellow, Harp, Irving, Isard, Jia,
  Jozefowicz, Kaiser, Kudlur, Levenberg, Man\'{e}, Monga, Moore, Murray, Olah,
  Schuster, Shlens, Steiner, Sutskever, Talwar, Tucker, Vanhoucke, Vasudevan,
  Vi\'{e}gas, Vinyals, Warden, Wattenberg, Wicke, Yu, and Zheng]{Tensorflow}
Mart\'{\i}n Abadi, Ashish Agarwal, Paul Barham, Eugene Brevdo, Zhifeng Chen,
  Craig Citro, Greg~S. Corrado, Andy Davis, Jeffrey Dean, Matthieu Devin,
  Sanjay Ghemawat, Ian Goodfellow, Andrew Harp, Geoffrey Irving, Michael Isard,
  Yangqing Jia, Rafal Jozefowicz, Lukasz Kaiser, Manjunath Kudlur, Josh
  Levenberg, Dandelion Man\'{e}, Rajat Monga, Sherry Moore, Derek Murray, Chris
  Olah, Mike Schuster, Jonathon Shlens, Benoit Steiner, Ilya Sutskever, Kunal
  Talwar, Paul Tucker, Vincent Vanhoucke, Vijay Vasudevan, Fernanda Vi\'{e}gas,
  Oriol Vinyals, Pete Warden, Martin Wattenberg, Martin Wicke, Yuan Yu, and
  Xiaoqiang Zheng.
\newblock {TensorFlow}: Large-scale machine learning on heterogeneous systems,
  2015.
\newblock URL \url{https://www.tensorflow.org/}.
\newblock Software available from tensorflow.org.

\bibitem[Balles(2017)]{probabilisticLineSearchImpl}
Lukas Balles.
\newblock Probabilistic line search tensorflow implementation, 2017.
\newblock URL
  \url{https://github.com/ProbabilisticNumerics/probabilistic_line_search/commit/a83dfb0}.

\bibitem[Baydin et~al.(2018)Baydin, Cornish, Rubio, Schmidt, and
  Wood]{hypergradientdescent}
Atilim~Gunes Baydin, Robert Cornish, David~Martinez Rubio, Mark Schmidt, and
  Frank Wood.
\newblock Online learning rate adaptation with hypergradient descent.
\newblock \emph{ICLR}, 2018.

\bibitem[Berrada et~al.(2020)Berrada, Zisserman, and Kumar]{L4_alternative}
Leonard Berrada, Andrew Zisserman, and M.~Pawan Kumar.
\newblock Training neural networks for and by interpolation.
\newblock \emph{ICML}, 2020.

\bibitem[Bostrom \& Yudkowsky(2014)Bostrom and Yudkowsky]{bostrom2014ethics}
Nick Bostrom and Eliezer Yudkowsky.
\newblock The ethics of artificial intelligence.
\newblock \emph{The Cambridge handbook of artificial intelligence}, 1:\penalty0
  316--334, 2014.

\bibitem[Chae \& Wilke(2019)Chae and Wilke]{empericalLineSearchApproximations}
Younghwan Chae and Daniel~N. Wilke.
\newblock Empirical study towards understanding line search approximations for
  training neural networks.
\newblock \emph{arXiv}, 2019.

\bibitem[De et~al.(2016)De, Yadav, Jacobs, and Goldstein]{bigBatchSGD}
Soham De, Abhay~Kumar Yadav, David~W. Jacobs, and Tom Goldstein.
\newblock Big batch {SGD:} automated inference using adaptive batch sizes.
\newblock \emph{arXiv}, 2016.

\bibitem[Draxler et~al.(2018)Draxler, Veschgini, Salmhofer, and
  Hamprecht]{elasticband}
Felix Draxler, Kambis Veschgini, Manfred Salmhofer, and Fred~A. Hamprecht.
\newblock Essentially no barriers in neural network energy landscape.
\newblock \emph{ICML}, 2018.

\bibitem[Duchi et~al.(2011)Duchi, Hazan, and Singer]{adagrad}
John Duchi, Elad Hazan, and Yoram Singer.
\newblock Adaptive subgradient methods for online learning and stochastic
  optimization.
\newblock \emph{J. Mach. Learn. Res.}, 12:\penalty0 2121--2159, 2011.

\bibitem[Fort \& Jastrzebski(2019)Fort and Jastrzebski]{wedge_model}
Stanislav Fort and Stanislaw Jastrzebski.
\newblock Large scale structure of neural network loss landscapes.
\newblock \emph{NeurIPS}, 2019.

\bibitem[Goodfellow et~al.(2015)Goodfellow, Vinyals, and Saxe]{LinePlots}
Ian~J Goodfellow, Oriol Vinyals, and Andrew~M Saxe.
\newblock Qualitatively characterizing neural network optimization problems.
\newblock \emph{ICLR}, 2015.

\bibitem[He et~al.(2016)He, Zhang, Ren, and Sun]{resnet}
Kaiming He, Xiangyu Zhang, Shaoqing Ren, and Jian Sun.
\newblock Deep residual learning for image recognition.
\newblock \emph{CVPR}, 2016.

\bibitem[Huang et~al.(2017)Huang, Liu, Van Der~Maaten, and
  Weinberger]{denseNet}
Gao Huang, Zhuang Liu, Laurens Van Der~Maaten, and Kilian~Q. Weinberger.
\newblock Densely connected convolutional networks.
\newblock \emph{CVPR}, 2017.

\bibitem[Jorge \& Stephen(2006)Jorge and Stephen]{numerical_optimization}
Nocedal Jorge and Wright Stephen.
\newblock \emph{Numerical Optimization}.
\newblock Springer series in operations research. Springer, 2nd ed edition,
  2006.
\newblock ISBN 9780387303031,0387303030.

\bibitem[Kafka \& Wilke(2019)Kafka and Wilke]{gradientOnlyLineSearch}
Dominic Kafka and Daniel Wilke.
\newblock Gradient-only line searches: An alternative to probabilistic line
  searches.
\newblock \emph{arXiv}, 2019.

\bibitem[Kingma \& Ba(2015)Kingma and Ba]{adam}
Diederik~P. Kingma and Jimmy Ba.
\newblock Adam: {A} method for stochastic optimization.
\newblock \emph{ICLR}, 2015.

\bibitem[Krizhevsky \& Hinton(2009{\natexlab{a}})Krizhevsky and
  Hinton]{CIFAR-10}
Alex Krizhevsky and Geoffrey Hinton.
\newblock Learning multiple layers of features from tiny images.
\newblock Technical report, 2009{\natexlab{a}}.

\bibitem[Krizhevsky \& Hinton(2009{\natexlab{b}})Krizhevsky and
  Hinton]{CIFAR-100}
Alex Krizhevsky and Geoffrey Hinton.
\newblock Learning multiple layers of features from tiny images.
\newblock Technical report, 2009{\natexlab{b}}.

\bibitem[Li et~al.(2018)Li, Xu, Taylor, and
  Goldstein]{visualisationLossLandscape}
Hao Li, Zheng Xu, Gavin Taylor, and Tom Goldstein.
\newblock Visualizing the loss landscape of neural nets.
\newblock \emph{NeurIPS}, 2018.

\bibitem[Loshchilov \& Hutter(2017)Loshchilov and Hutter]{sgdwithwarmrestarts}
Ilya Loshchilov and Frank Hutter.
\newblock {SGDR:} stochastic gradient descent with warm restarts.
\newblock \emph{ICLR}, 2017.

\bibitem[Luenberger et~al.(1984)Luenberger, Ye, et~al.]{nonlinear_programming}
David~G Luenberger, Yinyu Ye, et~al.
\newblock \emph{Linear and nonlinear programming}, volume~2.
\newblock Springer, 1984.

\bibitem[Mahsereci \& Hennig(2017)Mahsereci and
  Hennig]{probabilisticLineSearch}
Maren Mahsereci and Philipp Hennig.
\newblock Probabilistic line searches for stochastic optimization.
\newblock \emph{J. Mach. Learn. Res.}, 18:\penalty0 119:1--119:59, 2017.

\bibitem[McCandlish et~al.(2018)McCandlish, Kaplan, Amodei, and
  Team]{ModelOfLargeBatchTraining}
Sam McCandlish, Jared Kaplan, Dario Amodei, and OpenAI~Dota Team.
\newblock An empirical model of large-batch training.
\newblock \emph{arXiv}, 2018.

\bibitem[Muehlhauser \& Helm(2012)Muehlhauser and
  Helm]{muehlhauser2012singularity}
Luke Muehlhauser and Louie Helm.
\newblock The singularity and machine ethics.
\newblock In \emph{Singularity Hypotheses}, pp.\  101--126. Springer, 2012.

\bibitem[Mutschler \& Zell(2020)Mutschler and Zell]{pal}
Maximus Mutschler and Andreas Zell.
\newblock Parabolic approximation line search for dnns.
\newblock \emph{NeurIPS}, 2020.

\bibitem[Mutschler \& Zell(2021)Mutschler and Zell]{line_analysis}
Maximus Mutschler and Andreas Zell.
\newblock Empirically explaining sgd from a line search perspective.
\newblock \emph{ICANN}, 2021.

\bibitem[Netzer et~al.(2011)Netzer, Wang, Coates, Bissacco, Wu, and Ng]{SVHN}
Yuval Netzer, Tao Wang, Adam Coates, Alessandro Bissacco, Bo~Wu, and Andrew~Y.
  Ng.
\newblock Reading digits in natural images with unsupervised feature learning.
\newblock \emph{NeurIPS Workshop}, 2011.

\bibitem[Paszke et~al.(2019)Paszke, Gross, Massa, Lerer, Bradbury, Chanan,
  Killeen, Lin, Gimelshein, Antiga, Desmaison, Kopf, Yang, DeVito, Raison,
  Tejani, Chilamkurthy, Steiner, Fang, Bai, and Chintala]{PyTorch}
Adam Paszke, Sam Gross, Francisco Massa, Adam Lerer, James Bradbury, Gregory
  Chanan, Trevor Killeen, Zeming Lin, Natalia Gimelshein, Luca Antiga, Alban
  Desmaison, Andreas Kopf, Edward Yang, Zachary DeVito, Martin Raison, Alykhan
  Tejani, Sasank Chilamkurthy, Benoit Steiner, Lu~Fang, Junjie Bai, and Soumith
  Chintala.
\newblock Pytorch: An imperative style, high-performance deep learning library.
\newblock \emph{NeurIPS}, 2019.

\bibitem[Robbins \& Monro(1951)Robbins and Monro]{grad_descent}
H.~Robbins and S.~Monro.
\newblock A stochastic approximation method.
\newblock \emph{Annals of Mathematical Statistics}, 22:\penalty0 400--407,
  1951.

\bibitem[Rolinek \& Martius(2018)Rolinek and Martius]{L4}
Michal Rolinek and Georg Martius.
\newblock L4: Practical loss-based stepsize adaptation for deep learning.
\newblock \emph{NeurIPS}, 2018.

\bibitem[Sandler et~al.(2018)Sandler, Howard, Zhu, Zhmoginov, and
  Chen]{mobilenet}
Mark Sandler, Andrew~G. Howard, Menglong Zhu, Andrey Zhmoginov, and Liang-Chieh
  Chen.
\newblock Mobilenetv2: Inverted residuals and linear bottlenecks.
\newblock \emph{CVPR}, 2018.

\bibitem[Smith(2017)]{cyclicallearningrates}
Leslie~N. Smith.
\newblock Cyclical learning rates for training neural networks.
\newblock \emph{WACV}, 2017.

\bibitem[Smith \& Le(2018)Smith and Le]{bayesianPerspectiveSGD}
Samuel~L Smith and Quoc~V Le.
\newblock A bayesian perspective on generalization and stochastic gradient
  descent.
\newblock \emph{ICLR}, 2018.

\bibitem[Smith et~al.(2018)Smith, Kindermans, Ying, and
  Le]{increase_batch_size}
Samuel~L. Smith, Pieter{-}Jan Kindermans, Chris Ying, and Quoc~V. Le.
\newblock Don't decay the learning rate, increase the batch size.
\newblock \emph{ICLR}, 2018.

\bibitem[Vaswani et~al.(2019)Vaswani, Mishkin, Laradji, Schmidt, Gidel, and
  Lacoste-Julien]{sls}
Sharan Vaswani, Aaron Mishkin, Issam Laradji, Mark Schmidt, Gauthier Gidel, and
  Simon Lacoste-Julien.
\newblock Painless stochastic gradient: Interpolation, line-search, and
  convergence rates.
\newblock \emph{NeurIPS}, 2019.

\bibitem[Xing et~al.(2018)Xing, Arpit, Tsirigotis, and Bengio]{walkwithsgd}
Chen Xing, Devansh Arpit, Christos Tsirigotis, and Yoshua Bengio.
\newblock A walk with sgd.
\newblock \emph{arXiv}, 2018.

\bibitem[Yudkowsky et~al.(2008)]{yudkowsky2008artificial}
Eliezer Yudkowsky et~al.
\newblock Artificial intelligence as a positive and negative factor in global
  risk.
\newblock \emph{Global catastrophic risks}, 1\penalty0 (303):\penalty0 184,
  2008.

\end{thebibliography}
